\documentclass[phd,ilcc,twoside,logo,11pt,fullspacing,parskip,plainprepages,sansheadings]{infthesis}

\pdfoutput=1

\usepackage[table,xcdraw]{xcolor}
\usepackage{todonotes}
\usepackage{soul}
\usepackage{natbib}

\usepackage{graphicx}

\usepackage{blindtext}
\usepackage[colorlinks]{hyperref}
\usepackage{csquotes}
\usepackage{dirtytalk}

\usepackage{fdsymbol}
\usepackage{environ}
\usepackage{subfig}
\usepackage{tabularx}

\usepackage{booktabs}

\usepackage{titlesec, blindtext, color}

\usepackage[T1]{fontenc}

\usepackage{textcomp}

\usepackage[scaled=1.00]{gillius2}

\usepackage{newpxtext}
\usepackage{newpxmath}

\def\hat{\mathaccent "705E\relax}

\def\tilde{\mathaccent "707E\relax}

\usepackage{bibentry}

\usepackage{csquotes}
\usepackage{svg}
\usepackage{varioref}

\titleformat{\section}
{\normalfont\Large\bfseries}{\thesection}{1em}{}
\titleformat{\subsection}
{\normalfont\large\bfseries}{\thesubsection}{1em}{}
\titleformat{\subsubsection}
{\normalfont\normalsize\bfseries}{\thesubsubsection}{1em}{}

\usepackage[activate={true,nocompatibility},final,tracking=true,protrusion=true,kerning=true,spacing=true,factor=1200,stretch=50,shrink=50]{microtype}
\microtypecontext{spacing=nonfrench}

\definecolor{darkmidnightblue}{rgb}{0.0, 0.2, 0.4}

\definecolor{annotatedcolor}{RGB}{127, 127, 127}
\definecolor{plotacolor}{RGB}{31, 119, 180}
\definecolor{plotbcolor}{RGB}{255, 127, 14}
\definecolor{plotccolor}{RGB}{44, 160, 44}
\definecolor{plotdcolor}{RGB}{214, 39, 40}

\definecolor{annotatedcolor}{RGB}{127, 127, 127}
\definecolor{surprisecolor}{RGB}{31, 119, 180}
\definecolor{suspensecolor}{RGB}{255, 127, 14}
\definecolor{surpriseentropycolor}{RGB}{44, 160, 44}
\definecolor{suspensestatecolor}{RGB}{214, 39, 40}

\definecolor{plotlyblue}{RGB}{31, 119, 180}
\definecolor{plotlyyellow}{RGB}{255, 127, 14}
\definecolor{plotlygreen}{RGB}{44, 160, 44}
\definecolor{plotlyred}{RGB}{214, 39, 40}

\hypersetup{
linkcolor=darkmidnightblue
,citecolor=darkmidnightblue
,filecolor=Red
,urlcolor=darkmidnightblue
,menucolor=Red
,runcolor=Red
}

\newcommand{\simplex}{\triangle}
\newcommand\pp{p}
\newcommand\p{\mathbf{\pp}}

\newcommand{\HHt}{\mathsf{H}^{\textsc{T}}}

\DeclareMathOperator*{\argmax}{\mathsf{argmax}}

\newcommand*\entmaxtext{entmax\xspace}

\newcommand*\aentmax[1][\alpha]{\mathop{\mathsf{#1}\textnormal{-}\mathsf{\entmaxtext}}}

\NewEnviron{myequation}{%
    \begin{equation}
    \scalebox{1.20}{$\BODY$}
    \end{equation}
    }

\title{Great Expectations:\\ Unsupervised Inference of Suspense, Surprise and Salience in Storytelling}
\author{David Wilmot}

\submityear{2022}

\abstract{%
  \textit{Suspense} and \textit{surprise} are crucial ingredients of narrative fiction, engaging readers and making compelling stories. Stories can be seen as sequences of events where some events are more or less \textit{salient}, or essential to a plot. While there is a vast theoretical literature, these notations are not computationally well understood. Most recent work employing machine learning methods for narrative understanding tasks has relied on supervised methods that typically require large volumes of annotated data. In contrast, the thesis uses the knowledge implicitly encoded in unsupervised and self-supervised models trained primarily on story data. By applying methods derived from narrative theory, suspense, surprise, and salience can be inferred directly.

We create a hierarchical rollout model to infer suspense. The model builds on a GPT language model with a sentence encoder and context encoder that tracks the story state. Two qualities are modelled: Surprise, a backwards-looking measure of how unexpected the current state is given the story so far, and uncertainty reduction, a forward-looking measure of how unexpected the continuation of the story is. Both
  can be computed either directly with probability distributions via theoretical work from psycholinguistics \citep{Hale:01,hale2003information}, or a combination of story state and probabilities based on theoretical work by \citet{ely2015suspense}. A gold standard corpus of annotations is collected to evaluate suspense. There are 100 validation and 100 test set stories, randomly sampled from the \textit{WritingPrompts} corpus. Each evaluation story has per sentence annotations by five annotators.   Evaluating against short stories annotated with human suspense judgements finds that Ely suspense is the best predictor of suspense. The method is also applied to identifying turning points in movie synopses.

The hierarchical rollout model relies on the GPT language model to generate a tree of continuation sentences for the story. We propose an alternative hierarchical model that relies on temporal VAE (Variational Autoencoder) methods allows the same surprise and suspense methods to be inferred directly in a latent vector space, saving substantial computation. Reevaluating the same tasks finds that surprise measures perform similarly, but suspense measures perform worse. The same model can also function as a planning system for story generation. The results demonstrate strong performance in automatic cloze and swapping evaluations. The human judgments show stories generated with a reranking model improve on a GPT-2 baseline.

Measuring event \textit{salience} is essential in the understanding of stories and is conceptually related to both surprise and suspense. The salience work is extended to longer-form, such as full-length novels and plays. It extends a recent unsupervised method for \textit{salience} detection method from \citet{otake-etal-2020-modeling} derived from Barthes' Cardinal Functions. The method is scaled to longer works by incorporating an external knowledgebase and adding a memory mechanism. The model is evaluated with a novel approach to derive salience annotations with chapter-aligned summaries from the Shmoop corpus for classic literary works. The evaluation against this data demonstrates that the \textit{salience} detection model improves performance over and above a non-knowledgebase and memory augmented language model.

Overall the results in this thesis support the claim that recent advances in unsupervised NLP models possess internal representations that can be usefully applied in story comprehension tasks. Furthermore, as the state of the art develops, the approach offers much promise in other areas as an alternative to supervised methods for domain-specific tasks. 
}

\usepackage{graphicx}
\graphicspath{ {./graphics/} }

\usepackage{lscape}

\begin{document}

\begin{preliminary}

\maketitle

\begin{laysummary}
Stories interest us not because they are a sequence of mundane and predictable events but because they have drama and tension. Crucial to creating dramatic and exciting stories are surprise and suspense. Likewise, certain events are key to the plot and more important than others. Importance is referred to as salience. Inferring suspense, surprise and salience are highly challenging for computational systems. It is difficult because all these elements require a strong comprehension of the characters and their motivations, places, changes over time, and the cause/effect of complex interactions. 

Recently advances in machine learning (often called deep learning) have substantially improved in many language-related tasks, including story comprehension and story writing. Most of these systems rely on supervision; that is, huge numbers of people need to tag large quantities of data to tell the system what to teach these systems. An example would be tagging which events are suspenseful. It is highly inflexible and costly. 

Instead, the thesis trains a series of deep learning models via only reading stories, a self-supervised (or unsupervised) system. Narrative theory methods (rules and procedures) are applied to the knowledge built into the deep learning models to directly infer salience, surprise, and salience in stories. Extensions add memory and external knowledge from story plots and from Wikipedia to infer salience on novels such as \textit{Great Expectations} and plays such as \textit{Macbeth}. Other work adapts the models as a planning system for generating new stories.

The thesis finds that applying the narrative theory to deep learning models can align with the typical reader. In follow up work, the insights could help improve computer models for tasks such as automatic story writing, assistance for writing, summarising or editing stories. Moreover, the approach of applying narrative theory to the inherent qualities built in a system that learns itself (self-supervised) from reading from books, watching videos, listening to audio is much cheaper and more adaptable to other domains and tasks. Progress is swift in improving self-supervised systems. As such, the thesis's relevance is that applying domain expertise with these systems may be a more productive approach in many areas of interest for applying machine learning. 
\end{laysummary}

\begin{acknowledgements}
    I want to thank my supervisors, Prof. Frank Keller and Prof. Mirella Lapata, for their guidance during the PhD. To Prof. Alex Lascarides and Dr David Bamman for examining the PhD during the viva and for feedback.

Outside of the PhD, I would like to thank those with the School of Informatics who have worked to organise regular seminar series and other events with invited speakers. They made a significant contribution to making the PhD a richer experience by learning of the fantastic research within the School and the broader community. The same for the organisers of the IRR and IRP modules which I enjoyed tutoring. For comments and feedback on papers and ideas that contributed to the PhD then, I would like to thank Elizabeth Nielsen and Ida Szubert.

It was also great to assist in organising events for EdIntelligence. Special thanks to Stefanie  Speichert as the main instigator and driver for EdIntelligence, and Mari Liis-Pedak, for her contributions to the We Need To Talk About AI series. For these events, I would also like to thank Kasia Kokowska for promotion and for providing logistical support and Prof. Melissa Terras and EFI for funding for some of these events.

I would like to thank my family and friends for their support during the last four years. Especially to my sister Emma Scruggs for work in proofreading the thesis and to the memory of my late father, Peter Wilmot. Amongst others not mentioned but valued, I would like to thank Ivana Balazevic, Carl Allen, Sigrid Hellan, Rachel Bawden, Dibayendu Dey, Carol Chermaz, Rowan Spear,  Imogen Morris, Michael Camilleri, Ionela Mocanu, Mathew Tibble, Emily Bartran, David Hodges, Morgan Ruth, Laura Berrisford,  Sabine Webber, Sameer Bansal,  Yumnah Mohammed, Martin Rüfenacht, Lucia Chen, and Naomi Saphra for their support, friendship, and discussions during the PhD.

Lastly, I would like to thank the EPSRC council for funding the PhD.
\end{acknowledgements}

\standarddeclaration


\microtypesetup{protrusion=false} 
\tableofcontents

\listoffigures
\listoftables
\microtypesetup{protrusion=true} 

\end{preliminary}


\nobibliography*
\bibliographystyle{acl_natbib}

\chapter{Introduction}

\label{chap:introduction}

\section{Storytelling}

 In \textit{the Storytelling Animal} \citep{gottschall2012storytelling}  compares the story to a flight simulator:  

\begin{displayquote}
Fiction is an ancient virtual reality technology that specializes in simulating human problems ... Like a flight simulator, fiction projects us into intense simulations of problems that run parallel to those we face in reality. And like a flight simulator, the main virtue of fiction is that we have a rich experience and don’t die at the end.
\end{displayquote}

Storytelling is then central to cognition and exploring problems. Gottsschall also describes stories as the social glue binding together culture and values. He gives the formula for a story as  $\text{Story} = \text{Character} + \text{Predicament} + \text{Attempted Extrication}$. While it is not intended as a serious structural theory, it neatly sums up the elements. A story is not simply a predictable series of events with one event following the other: Fill a kettle with water, boil the water, put tea bags into a teapot, pour water into a teapot, ... is a coherent and logical flow of events arranged in time order. As the formula suggests, a story needs to have a character the reader can engage with and some predicament they are put in and need to get out of. With the tea making example, we could turn it into a story by creating a predicament such as a teapot being dropped and spilling water onto the floor. The person making the tea has to scrabble around to clean up the water. Maybe it was a gift from someone they are making tea for, and they are now panicked that they broke it. There are lots of creative alternatives. In differing from \citet{forster1985aspects}, the thesis defines a story as what Forster would call a plotful story.\footnote{Forster defined a story as a sequence of events and being plotful as have causality. It has been conventional since to use the second plotful version as the definition of a story.} A plot is the same temporal sequence of events but has the reader asking \textit{And then?} and \textit{Why?} The reader needs to want to know what will happen next, and there needs to be causality.

Stories either for comprehension or generation tasks have long been of interest in the computational field; some background is reviewed in Chapter \ref{chap:backgroundml}. It is because of the challenge inherent in the form: Comprehending or generating coherent stories requires knowledge of everyday events; temporal orderings; causality; characters and their motivations; and concepts such as suspense, surprise or mystery that create dramatic tension.

\citet{rabkin1970slim} placed suspense as the central dynamic in making a story exciting and eventful rather than an unremarkable sequence of events. Inferring suspense, alongside surprise and a higher level salience concept with unsupervised methods, is the core of the thesis. I now introduce these concepts with illustrative examples that are used as reference and discussion points later in the thesis:

\subsection*{Mr Bean - Sandwich for Lunch\footnote{For the video - \url{https://youtu.be/_hxldVSvDXE}}} This is a sketch from Rowan Atkinson's Mr Bean. It's best to watch the video, but it goes as follows: Mr Bean sits down next to a man eating a prepackaged sandwich. Rather than eat a pre-made sandwich, he proceeds to get a loaf of bread out of his jacket before cutting it into slices. He then uses his credit card to spread butter on the bread. Getting a piece of lettuce out of his pocket, he washes it in a fountain before drying it in his sock, having removed his shoe. Following this, he kills live sardines from a jar by hitting them against a park bench, cracks pepper with his shoe, and makes tea in a hot water bottle before sneezing and dropping the sandwich and tea onto the floor.

As \citet{forster1985aspects} notes at its most basic level, the only merit of a story is wanting to know what happens next, and the only fault is not wanting to know. What makes this particular scene an engaging story while someone making a sandwich at home is not. One idea comes from the theory of \textit{narrative suspense} that is covered in more detail later. \citet{Gerrig1994ReadersAP} describe the reader as a problem solver. In this view of suspense, typically, there is a hero with a goal. Gerrig and Bernardo use the example from \textit{Casino Royale} where \textit{James Bond} 's goal is to escape a gun pointed and pressed against his back. However, in this case, the goal is to make a sandwich. The reader thinks on behalf of the hero. The writer manipulates paths to the solution, adding and removing them from the problem space to generate interest or tension. Suspense and curiosity are created when the quality and quantity of the paths to the goal are reduced while still maintaining the goal's possibility. Crucially, this depends on the reader's or viewer's notion of a solution path and their expectations. Unlike \textit{James Bond}, \textit{Mr Bean} creates comedic suspense and draws in the viewer in a situation when there should be none. When \textit{Mr Bean} pulls out a piece of lettuce, washes it in the fountains, and spin-dries it in his sock is an absurdity that stretches out the path to the goal and creates uncertainty over the outcome. This uncertainty, combined with the unusualness of the activities, forces the viewer to think about Bean's situation and delivers the comedy. The viewer feels that anything might happen, which increases the uncertainty about the events' paths creates a sense of suspense. Crucially, just having different paths is not enough. If different paths are conceivable, but they all lead to similar outcomes, it is not suspenseful. \textit{Mr Bean} is suspenseful because we can imagine his unusual actions could work and also could go spectacularly wrong;  the paths lead to different destinations of material consequence to \textit{Mr Bean}. While this example is unusual, the point is to illustrate that suspense is the viewer's engagement with the situation and struggles to achieve a goal or avoid unfortunate outcomes that are important. And not just the common sense ordering of events, i.e., making a sandwich, you must first slice it and then butter it. Nor does suspense and storytelling rely on grand themes such as Romance or War for its sustenance. 

\subsection*{Only Fools and Horses: Bar Scene \footnote{For the video - \url{https://youtu.be/63rcdLeXiU8}}} This another comedy example to introduce the idea of surprise. In this sketch, two characters, \textit{Del Boy} and \textit{Trigger}, are talking about chatting up women in an upmarket bar. \textit{Del Boy} is leaning against the bar. Unknown to him a barman has lifted the bar. \textit{Del Boy} leans and fall through the bar. It is a basic slapstick joke that lampoons the pretensions of the main character, but it also shows \textit{surprise}. Unlike the sandwich, we don't anticipate what will happen or are not thinking about it this way. The fall is sudden. The main difference between \textit{suspense} and \textit{surprise} is that it is something that is more suddenly different from expectations. With \textit{suspense} there is uncertainty about a divergence of outcomes of consequence, a sense of anticipation or foreboding. 

\subsection*{Great Expectations - \citet{dickens_1861}} This is a much more substantial work (at around 183K words). There will be many characters and subplots in any longer work like this. Comprehension, therefore, depends not only on knowing the likely order of basic events such as making a sandwich but also on remembering characters and their relationships, the places involved, the motivation and the consequences of actions. The novel Great Expectations is essentially an estranged and extended love story between two characters \textit{Pip} and \textit{Estella} who meet as children. The book as a whole has five main \citep{10.2307/44398838} suspenseful plots: \textbf{(1)} \textit{Pip} receives money from a mysterious benefactor; one of the main twists of the book is the benefactor is not as expected which changes the whole course of the story. \textbf{(2)} The ongoing and changing relationship between \textit{Pip} and \textit{Estella}. \textbf{(3)} The murder of \textit{Mrs Joe}. \textbf{(4)} \textit{Orlick's} attempt to kill \textit{Pip}. \textbf{(5)} \textit{Magwitch's} attempted escape at the end of the book. Alongside these main plots, other sub-plots and details inform the reader: The back-story of the convicts escaping across the moor at the beginning of a novel; the details of \textit{Miss Haversham's} situation and her jilted marriage; \textit{Pip} growing up in rural Essex and then experiencing the sights and sounds of London; the eccentric lawyer \textit{Wemmick} and his strange house and relationship with his mother; \textit{Magwitch} secretly being \textit{Estella's} father. As well as ordinary everyday events, Dickens builds up and resolves these sub-plots through many thousands of words of text. Both of length and complexity of the plots and subplots makes comprehension of longer stories a computationally extremely challenging task. The same could be said for generating stories with such plots. In the event indexing model \citep{1999-04086-005,Zwaan1995TheCO}, readers need to able to keep track of five distinct dimensions: time, space, protagonist, causality, and intentionality. It is both through having long-term commonsense knowledge and expectations of what typically happens and situational awareness in recalling details pertinent at the time. The \textit{Great Expectations} example is highly relevant to Chapter \ref{chap:backgroundml} in discussing the limitations of existing deep learning models for comprehension and generation tasks, and to Chapter \ref{chap:salience} that implements an episodic memory system.

\subsection*{Gone Girl} Gone Girl is a contemporary book and film. It is selected as an example because it deliberately subverts typical plots. The plot: New York-based writer, \textit{Nick Dunne} and his glamorous wife \textit{Amy}, seem blissfully married. \textit{Amy} goes missing on the couple's fifth wedding anniversary. The police find strong forensic evidence and marital problems pointing the finger at \textit{Nick}. It transpires later that \textit{Amy}, having found out about Nick's affair, faked her death and framed him for the murder. Amy ended up with \textit{Collings}, a wealthy old boyfriend, after being on the road and robbed for the money she had stolen from \textit{Nick}. \textit{Amy} realises she had made a mistake. She then murders \textit{Collings}, makes it look like he kidnapped and raped her, and returns home to \textit{Nick}. After \textit{Amy} returns, there are more suspicions from the FBI, who do not believe her story. But after more contrivances, an argument in which \textit{Nick} learns \textit{Amy} artificially inseminated herself, the movie finishes with them expecting a baby and reuniting in the media as the happy couple. It has some things in common with \textit{Great Expectations} in that there is an intense mystery component with the viewer's expectations thwarted. It also, like \textit{Great Expectations} involves events running concurrently in multiple places and messes with the temporal structure to induce the viewer's uncertainty. The \textit{Gone Girl} example is relevant since it's a story that breaks standard conventions with storytelling and the readers' expectations. The characters know things the reader does not and which other characters don't, so there is a mystery as the viewer tries to figure out what happened. \textit{Gone Girl} is also highly relevant for the concept of salience. The movie illustrates the concept of salience used in the thesis, derived from \citet{Barthes1966AnIT} \textit{cardinal functions} and extends the work of \citet{otake-etal-2020-modeling}. The essential concept is events are salient if removing them makes the story less comprehensible. \textit{Gone Girl} has long stretches of everyday activity with not much significant happening. They frame the story but are not central to it. However, some other events or clues such as the affair or details that suggest \textit{Nick} wasn't responsible for cleaning the supposed murder scene are key to the plot. Salience is about identifying the top-level plot events that are more significant than other events in the story.

\subsection*{The Hero and Heroine's Journey, Rocky and Pretty Woman} 
\begin{figure}[htbp]
\centering
\includegraphics[width=0.70\textwidth]{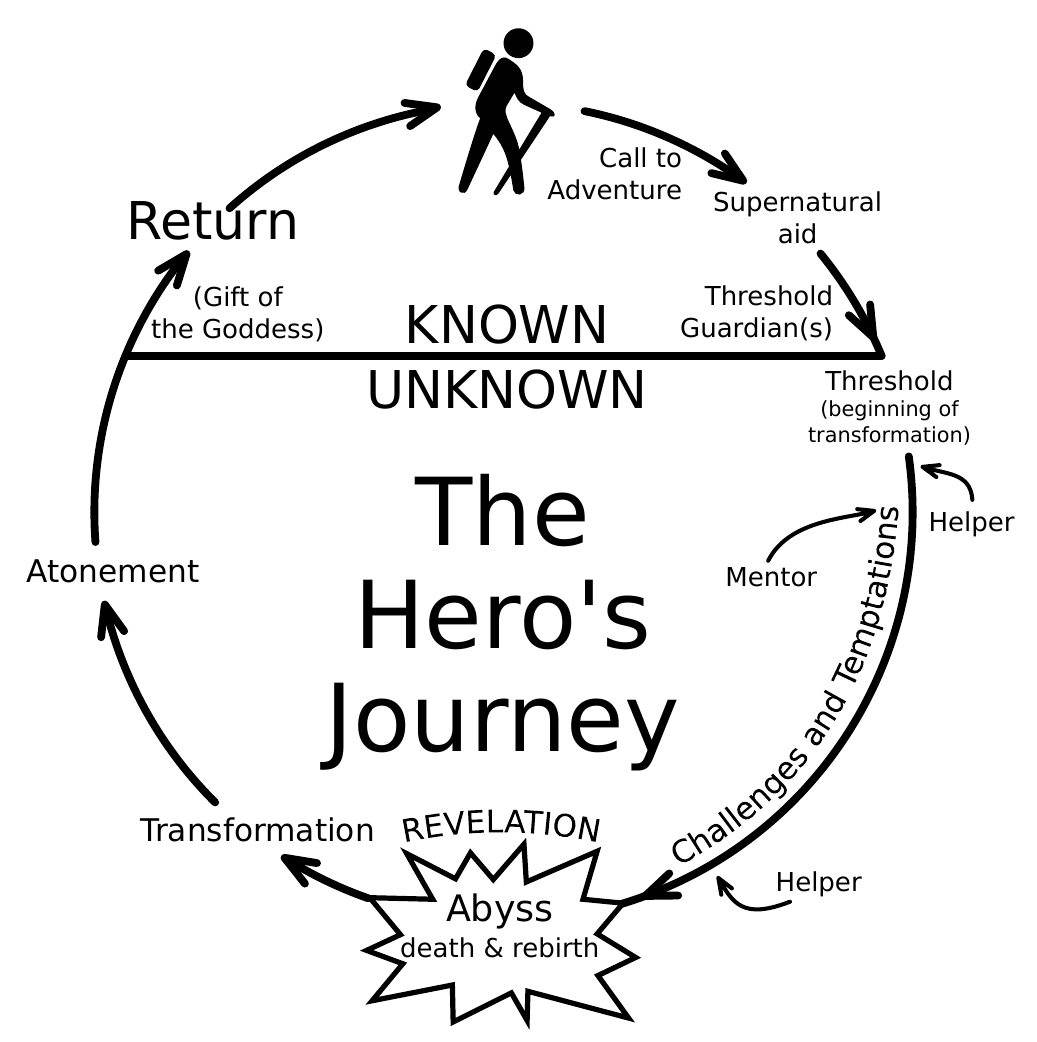}
\caption{The Hero's Journey from \citet{Luomala1949TheHW} and reproduced from \textit{Wikipedia}.}
 \label{fig:Hero'sjourney}
\end{figure}

The focus of the thesis is on a bottom-up understanding of salience and suspense rather than a top-down. Still, it has connections with another line of work that defines stereotypical story arcs. The classic example of this is \citet{Propp1958MorphologyOT} who expressed the \textit{Hero's journey} from classic Russian folk tales. Propp identifies 31 functional roles that make up these classic tales. These are split into four spheres: The first is introductory, sets the scene and includes functions such as \textit{interdiction} where the hero is warned of danger, and \textit{trickery} where a villain deceives a victim. The second starts the main quest and includes functions such as \textit{mediation} where the hero discovers the villainy or \textit{departure} where they embark on the journey or action to end the villainy. The third sphere is the central part of the story where there may be the \textit{acquisition} of a magical item,  \textit{struggle}, the hero or villain do battle, or \textit{victory}, the villain is defeated. The fourth sphere is a sometimes optional homecoming or epilogue where the hero is \textit{recognised}, or the villain may receive \textit{punishment}. While many of Propp's functions may seem genre-specific, the form has been highly influential in contemporary story writing with the theory elaborated by \citet{Luomala1949TheHW}, illustrated in Figure \ref{fig:Hero'sjourney}, and \citet{vogler1998writer}.\footnote{This booked coined the Heroe's Journey expression which was then backdated to Propp.}  The theory has been influential in Hollywood with films such as \textit{Star Wars}, \textit{The  Matrix}, \textit{The Wizard of Oz}, \textit{Toy Story}, \textit{Jaws}, \textit{True Grit}, \textit{Back to the Future} as well as non-western folk tales such as \textit{Mulan} or the \textit{Lion King}, or inspired by classic mythology such as \textit{The Lord of the Rings} use the form. The Hero's journey is more closely associated with male action narratives, but there is a version that has been proposed for a Heroine's Journey \citep{murdock1990heroine}. Traditional romantic narratives can also be interpreted as the relationship following a similar path to the hero on a quest. For example, in the classic fairytale \textit{Cinderella} there is the quest upon meeting the Prince. Challenges are thrown in the way of the relationship by the spell, which means that \textit{Cinderella} is unrecognised after leaving the ball and the villainous actions of the step-sisters, but \textit{Cinderella} eventually overcomes these barriers. Other romances may not have magic spells but usually have obstacles to overcome, such as family tensions or social status.

The two examples used in the thesis are \textit{Pretty Woman} and \textit{Rocky}. Pretty Woman described itself as a modern update of Cinderella. A prostitute (\textit{Vivian}) and a wealthy businessman (\textit{Edward}) meet when he picks her up as a lark. After \textit{Edward} hires \textit{Vivian} to stay with him for the weekend, the film follows the travails and hurdles they need to overcome between their different worlds to be together. \textit{Rocky} is the story of a small-time struggling Boxer who works on the side as a loan shark and debt collector and unexpectedly gets a chance to fight for the world title. The movie centres on his journey to fight for the title despite personal, financial, and emotional setbacks. Though being knocked down in the title fight, he recovers and breaks the ribs of the champion \textit{Apollo Creed}. Although he narrowly loses the fight, it proves to himself, the fans, and the media he is a worthy fighter and wins over \textit{Adrian}, the girl he has been dating. These Hero's journey type of stories have straightforward plotlines, don't have much mystery, and follow viewers' expectations. The relevance is part of the \textit{paradox of suspense} debate reviewed in Chapter \ref{chap:backgroundtheory}. \textit{Gone Girl} depends on breaking all viewers expectations to create suspense and dramatic interest. In contrast, Hero's journeys type plots can achieve suspense even if the view knows the ending in advance with high confidence. As with repeat viewing or reading, it questions the role of uncertainty in suspense or dramatic interest generally as being sure of what will happen means low uncertainty.

\section{Approach}

The examples are selected to illustrate the different ideas around surprise, suspense and knowledge in the thesis. These ideas are reviewed more thoroughly in Chapter \ref{chap:backgroundtheory} and where relevant with the main content chapters. The thesis computationally models the concepts of suspense, surprise, and salience. The approach is defined further in Chapter \ref{chap:backgroundml}. The core is that the concepts are conceptually about readers expectations; differences from what happens and readers expectations; or which events are necessary, and the story would be incoherent without. All can be modelled via heuristics applied to expectations over what will happen next. The most common supervised machine learning approach would label each of these desired properties and train a model to learn them directly. Instead, recent years has seen the advance of unsupervised or self-supervised models that are trained via inferring masked data. The models have demonstrated a high degree of capabilities across a range of tasks and domains. Unsupervised models can be trained by learning to predict future expectations. This thesis aligns the two by adapting the models to infer suspense, surprise, and salience directly without requiring a supervised model. Later in Chapter \ref{chap:salience} further ideas such as external knowledge and episodic memory are introduced to scale salience identification to longer works.

\section{Scope}

Common to all the experimental Chapters \ref{chap:rolloutsuspense}--\ref{chap:salience} if the computational inference of models trained on stories. The examples already given are illustrative of stories that are in scope for the thesis: \textit{Mr Bean} and \textit{Only Fools and Horses} are examples of physical comedies where there are unexpected turns of events. \textit{Rocky} and \textit{Pretty Woman} follow the archetype hero's arc of overcoming adversity. \textit{Gone Girl} and \textit{Great Expectations} are more mysterious plots where the viewer and reader respectively need to work out more for themselves. The thesis is focused on the plotful narrative stories, but there are other forms that could be considered stories or broader narrative forms that are excluded.

First, all the stories in the thesis are text only. Quite obviously, stories can come in various other forms, such as the oral tradition of many cultures. Within tribal and indigenous communities such as the Australian Aboriginals, stories such as the  \textit{Rainbow Serpent} or \textit{Seven Sisters} are often told through a mixture of an oral tradition, sand and cave paintings, dance, and pilgrimage to important sites. Stories can be as much spatial as temporal. Stories such as teaching or religious devotion involve experiences that are present in the Western tradition, for example,  \textit{Camino de Santiago} where there are sites, songs, icons and stained glass. Or from the Eastern tradition, such as the Buddhist pilgrimage to the Bodhi tree in Gaya.
\citet{gottschall2012storytelling} of the introductory quote argues these stories are central to human evolution and not only for devotional worship but also in learning skills, maintaining social relationships and community cohesion, learning hunting strategies or of climatic changes in the environment. An example of hunting stories for an aboriginal story can teach that hunters should cleanse themselves in tea tree oil before hunting. Or they can convey highly local and contextual messages, such as a particular location in a valley is good for hunting wallabies for a specific season or time of day. Stories are a way of incrementally teaching skills, culture, and social relations in an oral tradition. Clearly, these are not amenable to computational study and are not within scope.

There are plenty of other multimodal types of stories that are more common today, including theatre, puppet shows, opera, TV dramas, movies, and comics. Though TV dramas and movies are could apply the methods in the thesis with multimodal input such as speech, audio descriptions, image frames, etc, the thesis focuses exclusively on unimodal text to prove the narratology ideas in the thesis without the added complexity of multimodality.\footnote{Movies are included in the thesis but only with text scripts and synopses and not multimodal input.} The conclusion, Chapter \ref{chap:conclusion}, points to some future work for multimodality.

The chapter began by saying that it needs to be more than a series of events for a story to be plotful. The example given was filling a kettle and making a cup of tea, not being a story because a story needs jeopardy. A simple expected series of events does not qualify. So while comprehension every day expected sequences of events while necessary for storytelling, they are not in the scope of the thesis for evaluation. Often this form of learning can be called narrative learning, especially as part of commonsense reasoning, which is reviewed in Chapter \ref{chap:backgroundml}.

Other narrative forms are also outside the scope. One of them is news reporting. News reporting covers dramatic events such as murders, social conflict and natural disasters. Many of the topics and flows of events are the same in literary stories and news fiction. Many fictionalised stories are based on real events. The difference is how they are constructed. The stories in the thesis are based on the reader as problem solver idea. The writer is trying to lead the reader through the story making them wonder and try and infer what will happen next or infer why something has happened. It is why the \textit{Mr Bean} sketch can work as a story while he is just trying to make a sandwich. News reporting is different in that in typically follows a \textit{pyramid} structure where the most important information is given at the beginning of the report, following with filling in details in the subsequent sections. The most salient and suspenseful information is given at the beginning. The form is recapping what has happened. Hence the methods adopted in the thesis would not be expected to work. There are other stories such as \textit{Romeo and Juliet} where the entire plot synopses is contained in the prologue. However there is still dramatic interest in finding out how events unfold and engaging with the characters experiences which can create surprise and suspense. The prologue doesn't begin as a newspaper might -- \textit{Two young lovers committed suicide by poison, after being separated following one  lovers (Romeo) absconding following his murder of his lover's (Juliet's) cousin Tybalt, after a family altercation over the relationship.} Ideas of how something can be suspenseful when the reader may know what will happen is revisisted again in the \textit{paradox of suspense} discussion in Chapter \ref{chap:backgroundtheory}. One area of overlap is that liveblog news feed reporting has become more common where journalists post from real events as they are happening. As these have both uncertainty and jeopardy there could be some value in applying similar methods to live news reporting.

Creative forms such as poems and music are also not within scope. Poems and song lyrics often tell stories, sometimes they can be considered suspenseful as well. The reason for their exclusion is once again is because structure plays a strong role in both forms. Songs typically have repeating chorus elements, poems have rhyming schemes and metre. Songs obviously have the interaction between the music and lyrics as well. Because much of the form is related to structure the plots of narrative poems are expressed differently with repeating elements or made to meet the structural form. As such the methods in this thesis would not be expected to work on their own as expectations relate to the structure and language as well as the plot. Once again it is an area where ideas could be taken from the thesis for future work, and combined with some work that say could predict word rhymes or stresses for poetry, or sentiment classification for music say, as a small piece in a larger system.

Chapter is on \ref{chap:suspenseannotation} the annotation of the \textit{WritingPrompts} corpus for inferring suspense. The \textit{WritingPrompts} dataset is a collection of short stories written from prompts collated from a \textit{Reddit} creative writing forum. All of the stories are technically in scope as this is the dataset used for evaluation. However, while mainly of the stories are conventional because it is an open forum, some stories do not follow conventional narrative texts. An example is one of an old lady asked to recall what happened when she was a young child. One story from this prompts is just a series of unconnected memories such \textit{I remember standing on the grass and it raining}, \textit{we had cake and ice cream and my best friends birthday party}, \textit{my Grandpa's house had steep stairs and was dusty}. As noted, the models in the thesis of suspense and salience rely on the reader builds up expectations over what will happen next and consider outcomes. Stories like these that break conventions would not be expected to work for the model.

Another important aspect relates to cultural conventions in stories. As is argued in Chapter \ref{chap:backgroundtheory} the inference method in the thesis relies upon the reader's expectations as to what will happen. This is highly culturally contingent, so the models are not likely to transfer well outside the cultural norms and expectations of the training set. Out of domain could be in a different culture, period, or any other niche experience or interest. It applies as much to human readers as a computational model. For example, two of the four Chinese classical works of literature are \textit{Dream of the Red Chamber} and \textit{the Plum in the Golden Vase} (also translated as \textit{the Golden Lotus}). Even the names have a strong cultural association. The \textit{Red Chamber} is a reference to powerful officials, landowners and merchants richly decorating their bed chambers and offices in red lacquer work. Even before it was written, a Chinese reader would have guessed correctly that the book is about the struggle for power and ambition. Likewise, \textit{Plum in the Golden Vase} is a sexual metaphor suggesting sexual promiscuity, intrigue, gossip, rivalry and social climbing, which is what the novel is about. The cultural expectations are also highly different from what might be expected in a European tradition. It has been banned for periods as pornographic. Outside the culture, though, these associations will be completely missed, and that's without considering the more detailed aspects of the culture beyond the title. The same could be said of religious stories or niche fantasy or sci-fi, or Mills and Boon style romance novels that rely on the expectations of the familiar reader. Models in Chapters \ref{chap:tdvaesuspense} and \ref{chap:salience} are deliberately trained on a broader set of story datasets to try and improve the generalisation of the model. But more generally, it should not be expected that the methods in the thesis will transfer well out of domains they are trained on, as the methods rely on strong comprehension, which in storytelling is high domain-specific.

Lastly are the technical limitations of the methods within the thesis. The methods presented in Chapter \ref{chap:rolloutsuspense} and Chapter \ref{chap:tdvaesuspense} are both models that infer suspense and surprise in short stories (from \textit{WritingPrompts}. The method is outlined in detail in the respective chapters. However, the basic approach with suspense is to roll out expectations over what will happen next at the sentence level. The first of these chapters' methods infer suspense by generating a tree of concrete continuations. The second infers suspense by sampling from a latent vector space of possible future states. Although the second method has an order of magnitude less complexity, it is still comparatively slow to calculate. This makes it only feasible to run on shorter texts of up to a couple of hundred sentences, but not long books or screenplays, which is why the \textit{WritingPrompts} corpus was chosen. The secondary analysis also runs the method on synopses of longer movies. Overall the method is limited in scope to shorter works. However, there is a second bigger limitation in scope. The method is only feasible to run a few sentences ahead, which means the expectations for calculating suspense and surprise are local to the scene or particular point of the story. The point is that it means the model of suspense is more local and limited in scope than the general definition introduced in Chapter \ref{chap:rolloutsuspense}.

Chapter \ref{chap:salience} employs a computational much simpler method to infer salience. The chapter also augments a transformer-based language model with episodic memory. These allow salience to be inferred on longer works, including \text{Great Expectations} which is nearly two-hundred thousand words or the screenplays for feature-length movies such as \textit{Rocky} or \textit{Gone Girl}. While the method scales to longer work, the inference of salience is on the output context of the LM, which limits the range the inference operates on to a dozen or sentences. While longer than suspense inference, it still limits the scope of the salience within these much longer works to the local scenes within the story. For both the suspense and salience technical limitation, there are further discussions in the relevant chapters and some suggestions for how these limitations could be addressed with future work in the conclusion, Chapter \ref{chap:salience}.

\section{Thesis Chapters}

The main content chapters are as follows:

\begin{itemize}

\item \textbf{Chapter \ref{chap:backgroundtheory}: Suspense, Surprise and Salience –} Central to the thesis are implementing narratology and theory-derived models from the literature. Introduces the key concepts of suspense, surprise, and salience from theoretical work in narratology and cognitive science. The chapter also delves into the relevance of memory and knowledge within storytelling.

\item \textbf{Chapter \ref{chap:backgroundml}: Machine Learning in Story Comprehension –} Stories as fundamental human experience have been studied since the earliest days of computation. The chapter introduces a brief history of methods that have been applied both for story generation and comprehension tasks. The methods parallel the evolution of AI. Earlier methods are primarily slot filling and rule-based. More recent methods have been based on machine learning and automatically infer patterns from data. In recent years there has been a massive trend toward unsupervised (or self-supervised) latent space models that have achieved state of the art results across domains and tasks. The chapter reviews the power of the models as well as the limitations and proposes a thesis approach based on combing these unsupervised models with the discussed narrative theory.

\item \textbf{Chapter \ref{chap:suspenseannotation}: Suspense Annotation –} Although all the methods for inferring suspense in the thesis are unsupervised, evaluation is against human judgement. Two evaluation datasets are outlined. Both are collected on stories from the \textit{WritingPrompts} dataset of creative short stories. The first dataset annotates surprise and suspense for the whole story but fails to achieve a high inter-annotator agreement. The second dataset is a sentence by sentence annotation used for evaluation in the following chapters. Both the data collection and quality control process is explained in detail.

\item \textbf{Chapter \ref{chap:rolloutsuspense}: Hierarchical Suspense Rollout –} Implements the suspense and surprise theories introduced in earlier chapters. The model is a hierarchical unsupervised model on top of a base GPT language model. The model infers surprise and suspense via heuristic rules from theoretical models. The inference is derived from generating concrete continuations at each point in the story. Inference uses the probability distribution and/or the divergence in the outcome of a latent state to infer surprise and suspense. The results show a good correlation with human annotations of suspense. The second evaluation of movie synopses' turning points demonstrates that the method can also correlate with a top-down model of plot structure. 

\item \textbf{Chapter \ref{chap:tdvaesuspense}: Temporal VAE Suspense –} An alternative approach is taken to the same suspense methods. Generating concrete alternatives is computationally expensive and not feasible either for looking far ahead or longer works. Latent state models that forecast temporal state directly in the latent vector state have demonstrated promising results in domains such as computer vision and simulated real-world advantages. Directly sampling from future states in the model simplifies the complexity of the implement surprise and suspense models. Rerunning on the same evaluations finds that surprise performs similarly, but suspense is substantially worse. On the turning points movie synopsis evaluation, the performance is similar.

\item \textbf{Chapter \ref{chap:tdvaegeneration}: Temporal VAE Generation – } The temporal VAE model is capable of forecasting future latent states. Training a GPT-2 model to condition on the latent states from the temporal VAE model can operate as a planning system for generating stories. Analysis of cloze and swapping tasks demonstrates that the model has strong technical performance for predicting coherence. As a story generation system, it can be employed for reranking generated alternatives, or the latent state can be conditioned on for generation. The ranking method outperforms the ranking model baseline and the best other non-VAE reranking systems. The conditioning model slightly degrades performance. Further analysis suggests that latent conditioning while being strong on technical measures, reduces the generated text's diversity and makes the stories less interesting.

\item \textbf{Chapter \ref{chap:salience}: Knowledge and Salience – } Infers salience on long-form stories (books, plays, and screenplays). Salience is calculated with a method derived from Barthes' Cardinal Functions. Chapters \ref{chap:rolloutsuspense}-\ref{chap:tdvaesuspense} are only on shorter form stories, the language models could hold the whole story in context. The model augments a language model with a dense vector retrieval method that accesses knowledge from a knowledgebase of plot knowledge or Wikipedia. The knowledgebase is extended to act as episodic memory, allowing early references to similar situations in the story to be retrieved and conditioned on. Variations on Barthes' Cardinal Functions are experimented with, including vector-based equivalents, modelling as a difference between an experienced reader and naive one, weighting the salience according to the sentiment level, and swap-based. Evaluation is on automatically aligned silver labels from Shmoop, a corpus of classic literary texts. The method surpasses baselines in identifying salience. The knowledgebase and memory are shown to be key to the improvement.

\end{itemize}

\section{Publications}

The following papers have been published at major conferences or as pre-prints from the results of the thesis. \textbf{(1)} on inferring suspense presents the annotations from Chapter \ref{chap:suspenseannotation} as well as the model and evaluation from Chapter \ref{chap:rolloutsuspense}. \textbf{(2)} the the temporal VAE model and presents the story generation work from Chapter \ref{chap:tdvaegeneration}. \textbf{(3)} is the salience work with the knowledgebase and episodic memory model from Chapter \ref{chap:salience}. 

\begin{enumerate}
\item \bibentry{wilmot-keller-2020-modelling}

\item \bibentry{DBLP:journals/corr/abs-2109-06807}

\item \bibentry{wilmot-keller-2021-memory}
\end{enumerate}
\chapter{Suspense, Surprise and Salience}

\label{chap:backgroundtheory}

\section{Introduction}

This chapter looks into ideas from narratology and cognitive science that inform the models used for later comprehension and generation tasks. The first part considers what makes the story more than just a series of events, particularly why some events are more salient than others. The second part reviews ideas of suspense and surprise in storytelling, the paradoxes and conflicts of existing ideas, and how these ideas have been operationalised in computer models. This chapter concludes by considering ideas from cognitive narratology on memory and knowledge and how these apply to the computational modelling of suspense and salience.

\section{Salience}

\citet{Propp1958MorphologyOT} models the plot of traditional folktales using a fixed set of \textit{functions} that represent state changes and actions between characters. The main saliency work of this thesis builds on formal structural analysis and ideas of \textit{eventfulness}, that is, identifying what makes stories interesting. The primary approach to modelling \textit{salience} in this thesis is derived from Barthes' Cardinal Functions (BCF; \citealt{Barthes1966AnIT}) and extends recent work to operationalise this by \citet{otake-etal-2020-modeling}.  

BCF (Barthes Cardinal Functions, \citealt{Barthes1966AnIT}) are hinge events that cannot be deleted without altering the story; they are the decision points between alternative consequential paths. BCF can be thought of as a generalisation of the work of Propp's functions. Whereas Propp defines his functions as specific important actions in Russian Folktales, BCF extends this to encompass functionally important parts of any narrative text.  The basic theory behind Barthes' structural analysis is that a narrative can be decomposed into functional units; there are no useless parts.   Each unit has a function and type. Functions are operations that change the state of the narrative; they are usually actions undertaken by the main characters in the story but may be naturally occurring events. Within this scheme, there are two main types of functions: cardinal and catalysers. Cardinal functions are \textit{nuclei} that can be are referred to as \textit{dispatchers} in that they drive the stories direction. They change the state of the narrative world and open or close possibilities. Catalysers fill in the blanks between the cardinal functions and contain a limited amount of risk and uncertainty. Essentially they are trivial incidents of descriptions. Taking the \textit{Mr Bean} example, the cardinal functions are the main actions that advance the narrative, such as taking the loaf of bread out of his jacket and cutting with the scissors; it is progress made toward making the sandwich. The catalysers are the incidental actions linking them together. In \textit{Great Expectations}, the first chapter sees \textit{Pip} grabbed by an escaped convict on the moor and sent off to steal food and tools to help the convict free himself from his chains on pain of death. In this setting, the cardinal functions are these described main events. Catalysers are the in-between events such as the convict walking and the sound of the chains rattling or the breakfast being made while \textit{Pip} is preparing to acquire the food. For example, a cardinal event from the opening chapter would be \textit{"Hold your noise!" cried a terrible voice, as a man started up from among the graves at the side of the church porch. "Keep still, you little devil, or I'll cut your throat!"}, while a catalyser would be \textit{my sister had a trenchant way of cutting our bread and butter for us}. The second is an operative action but doesn't have the same status in the plot as the first.

Within Barthes' theory, there is another non-functional class called \textit{indices}. Pure indices are implicit and are descriptive text. Examples for \textit{Great Expectations} would include descriptions of the moor, of the churchyard, the convict's clothes and condition. Indices express more of an atmosphere or mood in the text, for example, from \textit{Great Expectations, Chap. 1} --- \textit{and that the dark flat wilderness beyond the churchyard, intersected with dikes and mounds and gates, with scattered cattle feeding on it, was the marshes; and that the low leaden line beyond was the river; and that the distant savage lair from which the wind was rushing was the sea; and that the small bundle of shivers growing afraid of it all and beginning to cry, was Pip}.  Another class of index is explicit and referred to as an \textit{Informant}. \textit{Informants} are factual; they are the means of identifying the character and placing them in time and space. For example, again in \textit{Great Expectations, Chap. 45} --- \textit{turning from the Temple gate as soon as I had read the warning, I made the best of my way to Fleet-street, and there got a late hackney chariot and drove to the Hummums in Covent Garden}. Notice that both sentences are indices or informants and contain actions that can be either catalysers or cardinal functions as well. The two can occur together within a single sentence.

In the Barthes' theory, functional relations are composable and hierarchical, with top-level cardinal functions serving as the high-level plot and central narrative structure. The catalysers, indices and informants are branches and leaves within this structure. Following Otake et al., we treat the cardinal functions or BCF as the most salient events in a story, corresponding to the primary plot. The main emphasis of the later machine learning work on the thesis is on cardinal versus non-cardinal functions using unsupervised methods rather than the whole hierarchical structure. Still, the theory is important for grounding the work.

The model has been defined in terms of Barthes' definitions, but there are several other closely related models from narrative theory, and the model connects to discourse theory. \citet{chatman1980story} splits \textit{events} in a story into \textit{kernels} and \textit{satellites}. The distinction between \textit{kernels} and \textit{satellites} is that a \textit{satellite} can be removed without changing the logic of the plot, whereas \textit{kernels} cannot. This is identical to the distinction between Cardinal Functions and all the other classes in Barthes' model. \citet{bal2009narratology} have a three-level model of stories: \textit{fabula} the logical and temporarily ordered sequence of events; \textit{story} which is \textit{fabula} presented from a particular point of view and may contain non-temporal ordering; and the \textit{text} which is surface-level realisations of the story. The high-level \textit{fabula} has similarities with \textit{cardinal functions} and \textit{kernels}; the primary difference is that this model is more of an abstraction from the text whereas \textit{cardinal functions} and \textit{kernels} are more demarcating the events within the text and so share more similarities to discourse analysis. Other narrative models such as by \citet{prince1973grammar} and \citet{10.2307/1772410} have a similar concept of a plot \textit{kernel}. All of the models discussed in this section fall within the formalist tradition.  \citet{pavel1985poetics} also produced a formal grammar of the story plot. However, this model is far more similar to the ideas discussed in this thesis on suspense. Rather than just seeing a story as a series of events, Pavel models stories as a graph representing the challenges to the characters and the actions required to meet the challenges, with paths to complete or fail to complete. This is similar to the idea of the reader as a problem solver and connects with ideas of suspense in the next section. \citet{10.2307/468420} proposes a model of the story where there is an \textit{action structure} separated from the realisation. The emphasis on the \textit{action structure} such as \textit{state changes} and \textit{causal chains} shares much with Pavel's model. All the discussed theories have been earlier structural approaches from \textit{narratology}. Still, there is comparable earlier work using similar ideas for computer-generated stories from \citet{s15516709cog0403}, \citet{RUMELHART1975211} and \citet{bringsjord1999artificial}. 

There is much in common between this line of structural narrative modelling and discourse theory. Rhetorical Structure Theory (RST; \citealt{Mann1988RhetoricalST}) defines hierarchical graphs of relations between \textit{discourse units} and these are defined in terms of \textit{nucleas} and \textit{satellite} operations. Like the narratology models discussed, one of the key distinctions is that the text is less comprehensible if the \textit{nuclei} are removed. \citet{hobbs1990literature} defines discourse in terms of \textit{subordinating} and \textit{coordinating} relations. Segmented Discourse Representation Theory (SDRT; \citealt{Asher2005LogicsOC}) also uses a \textit{subordinating} and \textit{coordinating} relation. \textit{Coordination} continues the discourse at the same level and includes types such as \textit{Narration}, \textit{Explanation}, \textit{Elaboration}, \textit{Background}, \textit{Contrast} and \textit{Parallel} to define different progressions. \textit{Subordination} includes relations such as \textit{Elaboration} or \textit{Explanation} and changes the granularity of description to a lower-level. Again the overlap with the literary models is striking. Indeed \citet{Bateman2014AMD} use the SDRT model more or less wholesale for a multimodal grammar of visual narratives, such as comics or film. The main difference between the discourse and the narrative models is that usually, the discourse models have more tightly defined types and are used to analyse shorter text of a few paragraphs. The narrative models are more concerned with longer running entire stories as the unit of analysis. Also they often use a coarse grouping such as a scene from a movie or paragraph instead of just a sentence. However, the  \textit{salience} model in this thesis uses the sentence level. The emphasis, with the narrative models is more semantically on the major actions of the main protagonists of the work and their challenges. While a discourse representation may have many top-level relations, a model such as Barthes' or Chatman's would place only some of these as \textit{cardinal functions} or \textit{nuclei} depending on their overall significance to the story.

The discussed structural models have the concept of \textit{cardinal} or \textit{nucleus} events that can be said to be more \textit{salient} than other events and form the core of the plot. However, the previous definition only defined this as the story being changed when this event was deleted. It didn't specify the characteristics that make an event \textit{cardinal}. \textit{Eventfulness} \citep{huhn2010eventfulness,huhn2014event,Schmid+2008+17+34,Schmid+2017+229+246} is a concept that extends the normal notion of an event (i.e. change of state with an agent and a patient) into something more significant in a story. Specifically, Schmid's concept of \textit{eventfulness} requires it have to have properties, with the first two being the most important:

\begin{enumerate}
  \item \textbf{Relevance:} The event must represent a significant change in the narrative world. It can't be a peripheral event. For example, in \textit{Rocky} the champion \textit{Apollo Creed} doing a sparring session for the cameras isn't that relevant. Whereas, him announcing that we will look for a new challenger as his opponent is injured, as this gives \textit{Rocky} his chance.
  \item \textbf{Unpredictability:} Deviation from what is expected in the narrative world. Note that the narrative world say for a fairytale or sci-fi, may have very different representations from modern-day reality, or from a historical work or a work set in another culture. Unpredictability is from the character's and the narrative worlds perspective and not the reader's.\footnote{Schub refers to the expectations in the narrative world as the \textit{Doxa}, it is the individual and collective assumptions and expectations of that world. The readers perspective can be thought of as a script. It is recipe or set of instructions they expect to follow from previous experience.} As is illustrated by the Hero's journey, plots can have stereotypical roles which the audience expects but aren't so in the narrative world. In the world of \textit{Rocky} is is expected he will beaten up badly or be paid to take a fall, but the viewer of Hollywood movies will expect him to rise to the challenge.
  \item \textbf{Persistence:} The event is consequential for the thought and action of the character within the narrated world. \textit{Apollo Creed} needing a new opponent and choosing \textit{Rocky} would be a good example.
  \item \textbf{Irreversability:} The effect of the event is persistent in the narrative world. Once \textit{Rocky} is chosen as the challenger it has an irreversible effect. He thinks of himself as a contender and not of quitting and working as a loan shark or in the docks. If he'd offered a different small time fight against another journeyman, the pattern of small time fights and the sideline work would continue.
  \item \textbf{Non-Iteratively:} The change happens once. Regular events that occur repeatedly in the characters everyday life are not eventful. For example the daily runs and bagwork in the meat factory are famous scenes from the movie but they more give atmosphere and show a passage of time than being eventful. 
\end{enumerate}

\textit{Irreversability} and \textit{Non-Iteratively} are binary qualities, but the others are all on a scale with the intuition being that the higher the \textit{Relevance}, \textit{Unpredictability}, and \textit{Persistence} the more \textit{eventfulness} there is. Unlike the deletion concept, which is about comprehension, \textit{eventfulness} is about a model of changes in the narrative world.

To give a further example for \textit{Pretty Women}: An \textit{eventful} incident would be when \textit{Edward} following a night together with \textit{Vivian}, hires her for a week to pretend to be his girlfriend. This is \textit{unexpected}, \textit{relevant} in the world, and has a big \textit{effect} as it starts the change of their relationship from a prostitututional one to a romantic one. Contrast this with the early scenes of \textit{Vivian} working on the streets of Los Angeles at the beginning, which meets none of the criteria and is intended to establish the character. As well as the existing narrative theory models mentioned, this distinction between \text{eventful} and not \textit{eventful} matches \citet{Lotman1977TheSO}'s distinction between a type I event (\textit{non-eventful}) and a type II event (\textit{eventful}).  Lotman defines the difference as an event where the hero transgresses a seemingly impenetrable barrier between two distinct worlds. This describes both the \textit{Rocky} and \textit{Pretty Woman} examples given, and aligns with \textit{Schmid's} \textit{eventfulness}.

The central concept of \textit{salience} used in the experimental work in this thesis is the deletion concept described using Barthes' Cardinal Function. Further technical work comparing the model used with other related new salience work is in the next chapter. The \textit{eventfulness} concept, while not directly used in the \textit{salience} chapter, is an essential link to suspense and has been introduced because it is important later in discussing the limitations of the experimental results and directions for future research.

\section{Surprise and Suspense}

This section follows on from the discussion of salience to cover another key ingredient of narrative storytelling, \textit{surprise} and \textit{suspense}. The salient events as outlined by the BCF are the critical events of the narrative on which the plot hangs. It distinguishes the significant events from the incidental or the descriptive text that fills in the narrative. \textit{Suspense} and \textit{surpise} are different and are crucial to creating interest, and anticipation in storytelling is key to enjoyment \citep{oliver1993exploring} and immersion in the story \citep{hsu14}. Modelling \textit{suspense} and \textit{surprise} using an unsupervised model is the second key theme of this thesis. This section introduces \textit{suspense} and \textit{surprise} building on the idea of salience. It then considers the debate in the literature around the \textit{paradox of suspense}. It then operationalises this suspense model with the model used for the ML (Machine Learning) models later in the thesis. 

An intuition of suspense was introduced with the idea of the reader as a problem solver. In reader as a problem solver that \textit{suspense} is concerned with both \textit{uncertainty} and \textit{jeopardy} over an outcome where the reader tries to imagine a solution path through the challenges faced by the character. This idea of suspense is related to both surprise, mystery and salient events in the story. The following example from \citet{boulter2007writing} and describes a situation in a story illustrating the concepts:

\begin{enumerate}
  \item{A woman is sitting at a kitchen table quietly. Suddenly a man with an axe bursts in.}
 \item{Before the man bursts in, we can see him approaching the door, but the woman is unaware of his presence.}
\item{The man bursts in with the axe and then goes and kisses the woman.}
\end{enumerate}

First, all of the variations would be considered salient events since they are central to the plot in the story. The difference is how the reader is guided through the process. \textbf{(1)} is an example of surprise. The reader thinks it is a typical domestic scene. Then there is a sudden increase in tension and alarm as the man comes in with an axe. \textit{Surprise} is characterised by a sudden shift in expectations without warning. \textbf{(2)} is \textit{suspense}. This is often thought of as the \textit{Alfred Hitchcock} \textit{Psycho} version of suspense; the reader knows something that the woman at the table doesn't know about the axeman. In \textit{eventfulness} one of the ideas is that something may be eventful if it is expected to the reader but unexpected to the character. The reader puts themselves in the character's shoes. An idea related to this is explored later in work on knowledge and memory. The important point is the reader is aware of something the character is not, which creates both uncertainty and jeopardy. Taking the reader as a problem-solver approach because the woman is unaware also makes it harder to solve how she will escape the predicament and increase her concern. In \textbf{(3)} we expect that the axeman will attack the woman but then suddenly, something happens which is unexpected to the reader. This is \textit{mystery}. It is the reverse of the previous suspense example; in this case, the characters know something that the reader doesn't. All three are important techniques in storytelling.

\textit{Gone Girl} makes use of all these concepts. The film first creates a lot of suspense when Amy, the wife, is missing. We wonder what has happened to her. Is she alive? If so, where? If not, who did it? The film creates suspense by creating a trail of possible clues that leads us to imagine what has happened and creates a sense of concern for Amy. Then the suspense turns toward the husband, Nick and whether or not he could have been involved in the disappearance and suspected murder. There is then a sudden surprise point in the film where we realise that Amy is still alive and deliberately acted to frame her husband. This creates a tremendous sense of mystery for the viewer as to why this has happened. Amy has tried to frame Nick, but Nick also doesn't seem too concerned by her disappearance. The mystery is gradually solved by back-filling the story of Nick's affair, and Amy's feeling of neglect and disappointment that Nick is a failure compared to her wealthy college boyfriend. 

\textit{Great Expectations} also makes heavy use of all these plot devices. For example, \textit{Pip}, a relatively poor young man is introduced to a young girl, \textit{Estella}, who is under the care of the wealthy but eccentric \textit{Miss Haversham}.  The suspense is created in the early part of the book by the strange interactions between \textit{Pip} and \textit{Estella} mediated by \textit{Miss Haversham}. When \textit{Pip} receives money from an anonymous benefactor, it is assumed to the reader and \textit{Pip} to be intended to raise him as a gentleman for \textit{Estella}. \textit{Dickens} always leaves ambiguity to keep the anticipation in the story. Surprise occurs much later when the convict \textit{Abel} \textit{Magwitch} appears, and we learn that he was the mysterious benefactor. This changes entirely the reader and the characters understanding of what has gone before. \textit{Magwitch} is an unnamed convict who appears at the start of the novel but is assumed to have exited the story, having been transported to Australia. In the novel, there is also a lot of mystery. One such example is \textit{Miss Haversham's} oddities, and bitterness are caused by having been jilted by a conman. \textit{Pip} gradually unravels during the novel that this conman, \textit{Compeyson}, had been a convict that had escaped together with and then fought with \textit{Magwitch} at the beginning of the book. They had been partners in crime who had fallen out. Unlike with suspense, there is not a particular threat the reader is concerned with. Instead, the information is revealed in pieces that we try to put together throughout the novel.

The discussed examples are to illustrate the different concepts of suspense, surprise, and mystery. \citet{boyd2009origin} describes the art of storytelling, including all these techniques, as the battle for the attention of the reader. This thesis is primarily concerned with suspense and surprise as well as the discussed salience. As well as \citet{Gerrig1994ReadersAP} several other literary theorists have framed suspense as one of problem-solving. \citet{scholes1974structuralism} describes Barthes' analysis idea as being about \textit{enigma}, \textit{delay} and \textit{disclosure} where a question is posed to the reader and this slow revealing of the answer is what creates suspense. \citet{ryan1991possible} frames the idea as one where the reader identifies with a character, then challenges and constraints are imposed on them, which narrows the range of possible answers, hence creating suspense. \citet{rabkin1970slim} also describes suspense as predicting outcomes for questions posed by the text.

 Relevant to the discussion around \textit{suspense} is although it is widely referred to in the literature and seen as a central concept in narrative and cognitive work, there are contradictory views about what precisely it is. There are two main components to the suspense ideas discussed: The first is some outcome at stake. The second is there is uncertainty over that outcome based on the reader's expectations of what will happen. The conflicting views in the literature about suspense is called the \textit{Paradox of Suspense} debate \citep{prieto1998paradox,yanal1996paradox}. The paradox of suspense is as follows:

\begin{enumerate}
  \item The traditional view of suspense is that it requires uncertainty.
  \item Knowledge of the outcome of a story would seem to indicate there is no uncertainty.
\item Suspense is an experience that has been observed on repeat viewings when obviously there is no uncertainty of outcome. Even if the reader doesn't know what will happen there are plenty of stories such as \textit{Romeo or Juliet} where the prologue gives away the whole plot. The story starts from the end and retells earlier events, so the outcome is known. The story is based on historical events likely to be known to the reader.\footnote{ As an example, a recent historical movie called \textit{Munich: The Edge of War}, it is about British Prime Minister's Neville Chamberlain's effort to secure a peace treaty with Hitler in Munich, in 1938. Although the exact events of the plot are unknown to viewers there won't be many wondering whether Chamberlain's attempts to prevent Hitler going to war succeeded.}  There are also genres such as in the hero's journey where although the characters and circumstances are new the plots can be fairly formulaic. 
\end{enumerate}

The paradox is that these three cannot all be true as they are inherently contradictory. Some authors emphasise that uncertainty over outcomes is important in suspense (e.g. \citealt{o2013computational,zillmann1996psychology, abbott2008cambridge}). Other authors claim
that suspense can exist without uncertainty
(e.g. \citealt{smuts2008desire,hoeken2000suspense,gerrig1989suspense}). \citet{Smuts2009-SMUTPO-2} says that each of these constraints can be rejected in turns, and there have been four proposed solutions: \citet{carroll_2001} idea of \textit{entertained uncertainty} rejects the first premise. It is similar to the discussed ideas with \textit{eventfulness} in that the reader puts themselves into the shoes of the main protagonist. Taking this view, someone watching \textit{Gone Girl} or \textit{Rocky} for the nth time, may still experience suspense as they feel for the character. It could even be said that knowing what will happen might heighten the suspense as they know what danger awaits the character. The \textit{Desire-Frustration} theory \citep{smuts2008desire} argues that it is the desire for the outcome that is frustrated by challenges that create the suspense and not the uncertainty, and so also rejects the first premise. The difference between these two is that in Caroll, the reader is imagining (and feeling) uncertainty on behalf of the character, whereas in Smut's they are still in their shoes but not uncertain. With Smut's is in the \textit{desire} in wanting them to achieve their goal and created by the obstacles on the path to achieving it. \citet{Gerrig1997ISTA}'s argument is for \textit{moment-to-moment forgetting} which is a rejection of the third premise. This is more of a cognitive explanation in that it argues that the memory limitations mean we are not used to repeating events, so they may feel uncertain even if we have read or seen them before. It is a failure of memory to recall knowledge correctly. This may be especially true for long and intricate plots such as \textit{Great Expectations} or \textit{Gone Girl}.  \citet{yanal1999paradoxes}'s argument is essentially to deny the third by saying that it is not suspense. Two different phenomena are confused for each other, and one is suspense, which depends on uncertainty, and the other is a separate concept.

There has been recent cognitive work to try and test this paradox of suspense. \citet{delatorre2018confronting} study evaluating repeat readings of a story. The results are provisional but seem to suggest that uncertainty affects the readers judgement of the story, but is distinct from \textit{suspense}. \citet{grady_sara} conducted a repeat study with movie clips and found that suspense occurs with repeat viewings but the response to stimuli declines. While there are no clear answers in the literature, the vital point is that both outcome and uncertainty may be considered different factors with unclear influences, and their interaction is complex. Therefore, in the computational inference, both uncertainty and outcome are modelled in separate measures and also combined.

\section{Models of Suspense}
\label{sec:suspense}

\label{sec:def}

This thesis aims to model suspense in computational terms, with the
ultimate goal of making it deployable in NLP systems that analyse or
generate narrative fiction. The model to operationalise suspense and surprise is from \citet{ely2015suspense}. Ely developed the model as a general model of suspense in the economics field. Intuitively the approach is to model suspense as a variance over future expectations; this adopts both the outcome and uncertainty aspects of the discussed literary theory of suspense. All of the events leading to the current point produce a belief as to what the current state is, as does the agent's prior knowledge and experiences. The bigger the divergence between this expected future state and other imagined alternative future paths, the bigger suspense is. \citet{lehne2015toward} have a related but different model of suspense to Ely et al. It is derived from cognitive reading and music listening studies. In Lehne and Koelsch's model suspense (called tension) is the distance between the expected future best outcome and worst outcome. To use a paths metaphor, if Ely is the expected average distance of all the paths diverging from the current point, a distribution over the possible onward journeys, then Lehne and Koelsch is the distance between the furthest paths end points from each other. This thesis concentrates on Ely, but as they are clearly conceptually related Lehne and Koelsch model is worth exploring in future work. Ely's surprise is modelled as the difference between the expected and actual outcome. The Ely paper is wide-ranging and applies these ideas to sports, political contests, gambling, games shows, and murder mysteries (though only theoretically, there is no experimental inference). Relating the sports example to storytelling, consider tennis. In tennis, not all points are equal in determining the outcome of the match. For example, if someone is 5-4 up in a set and serves at deuce (40-40), then two points, either way, results either in the set being won for one player or the set being equal at 5-5. The single points have a big impact on the outcome of the match. If someone is 5-1 up in a set and 40-0 then a couple of points, either way, will have a small impact on the outcome. Through Monte Carlo simulation, tennis games can be analysed for how suspenseful they are, with suspense being how likely the match's outcome is to be in the balance. In experimental work \citep{DBLP:conf/cogsci/LiBG18,Li2019TheCM,Li2021ExpectationsAF}, the Ely model of suspense corresponds with human judgement with the card game \textit{Blackjack} and other games. These papers independently test hypotheses that model uncertainty and closeness to a negative outcome (losing) to model suspense. They find that the combined model performs better than both independently, and so both components seem to be important. The links to narrative suspense are the same components described in problem-solving to avoid a negative outcome, or uncertainty over outcome applies to suspense in storytelling. The more challenging part is that space of possibilities in open-domain storytelling is much larger and more complicated than a simple tennis score or blackjack. This model is formally introduced in Chapter \ref{chap:rolloutsuspense}.

An alternative conceptual approach starts from the assumption that concepts
developed in psycholinguistics to model human language processing at
the word level \citep{Hale:01, hale2006uncertainty} can be generalised
to the story level to capture suspense, the Hale model. Rather than model each subsequent word, each subsequent sentence is modelled. Unlike Ely, the Hale model is based only on probability of what will happen, and not a combination of probability and states representing different outcomes. Intuitively, Hale suspense is a measure of how much the current context changes the entropy of subsequent continuations. If the current context is more influential it is more suspenseful. As per Ely it is also possible to adopt the same approach looking backwards as surprise. 

Both Ely and Hale models can be applied as: Surprise, a backwards-looking measure of how unexpected the current
state is given the story so far. Suspense, a
forward-looking measure of how unexpected the continuation of the
story is. Operationalising suspense both with the Ely and Hale notions of suspense and surprise with an explicit model is tackled in Chapters \ref{chap:rolloutsuspense} and \ref{chap:tdvaesuspense}.

\section{Knowledge and Memory}

The third theme of the thesis is memory and knowledge. The last practical chapter in this generation adapts the RAG model (Retrieval Augmented Generation; \citealt{NEURIPS2020_6b493230}) to act as an external knowledgebase and memory mechanism to infer salience and improve the performance of deep learning models that only use parametric knowledge. The technical advantage of doing this, which is covered in Chapter \ref{chap:backgroundml}, is knowledge encoded into model weights is limited by the number of weights, whereas external knowledge isn't. There is, however, a much more relevant cognitive rationale for the approach.  

This thesis hypothesises that external knowledge is as crucial for story understanding and generation as the factual question answering domains where the external knowledge and memory have been most commonly applied. To take several of the introductory examples: In the comedy sketch \textit{Mr Bean} the comedy relies on the absurdity of his actions in making the sandwich compared to how people typically make it at home or buy it and bring it to the park. So the comedy relies on being at odds with typical mental scripts or plans \citep{schank1975scripts,schank1977scripts} for how we accomplish tasks. In describing both salience and events, the idea of reader expectation was crucial. As in \textit{Mr Bean}, this relies on expectations of what typically happens in real life but also on expectations that are typical in narratives. This applies particularly to genre fiction such as \textit{Romance} and or \textit{Action}. As per our stereotypical Hero's journeys of \textit{Rocky} or \textit{Pretty Women} the viewer probably knows what to expect most of the time from watching similar movies. This aspect of knowledge doesn't apply only to understanding scripts or event orders that happen in stories. There is also a broader cultural knowledge in for example, \textit{Rocky} or \textit{Mr Bean}, so might a well-known \textit{Great Expectations} character such as \textit{Miss Haversham} or \textit{Magwitch}, have broader cultural associations. The same could be true of places such as \textit{Dickensian London}; or of fantasy or sci-fi worlds such as \textit{Narnia}, \textit{Gotham City}, or the \textit{Star Wars} universe; or stories set in a distant time or culture. Stories have a rich cultural dimension beyond just a sequence of events to be understood by viewers or readers. They can be incomprehensible to someone who does not have the required knowledge to understand the world. This is particularly the case to someone reading a work set in an unfamiliar culture or as part of a long-running fantasy/sci-fi world the reader may be unfamiliar with. Going back to the \textit{eventfulness} idea, the norms of the world in which a story is set is important, whether fictional or real, as is script knowledge of everyday events.

Memory is another crucial component of narrative understanding. Consider the more complicated plots of \textit{Gone Girl} or \textit{Great Expectations} where the reader or viewer is asked to keep track of many different characters and events to make sense of the narrative. These stories also play with expectations by containing false paths that require recalling much earlier passages and situations to think about what has and will happen. These two examples include mystery as a vital part of their plot, but memory and recall are hugely important generally in storytelling.

An essential base for this approach is work on stereotypical frames or scripts \citep{schank1975scripts,schank1977scripts} reduce the complexity of understanding the narrative. Whereas formal structure requires parts of a story entirely according to a set scheme, stereotypical knowledge learns simplifications and repeating patterns. Scripts allow the reader to build a mental model and expectations of what will happen by remembering a relatively small number of textual cues. \citet{10.2307/1773182} and \citet{emmott1997narrative} borrows from linguistics theory and use the idea of \textit{frames} based on episodic memory for understanding narratives. \textit{Frames} are stereotypical episodes from experience that are related to a given context and are used to make sense of events. The relationship between a context and a frame is many-to-many. As per Zwaan's indexing model, contexts are Spatio-temporal and connected with the entities in the scene, and the story gives signals which switch the context. To give an example, the opening scene of \textit{Great Expectations} takes place on bleak coastal moors of Essex near a remote Churchyard. The escaped convict \textit{Magwitch} grabs hold of the young boy \textit{Pip}. The reader in the introduction will recall from memory episodes where they themselves or from a previous story have encountered a similar context --- a windswept seascape, a flat moor, a quiet country churchyard. When the convict appears, even though it's not mentioned, the reader still infers that both \textit{Pip} and \textit{Magwitch} are in the same landscape, using the same term as Emmott, the context remains active.  When the convict appears in chains, the reader expects violence from recalling other incidents of stories where there has been a criminal in chains. The recall is based on the primacy of the events in the context and the recency of events. The important change from previous structuralism is the limitations and functions of the reader's memory and knowledge that are crucial, approximations and not abstract plot forms. The recall is based on pattern matching of the context. This thesis's comprehension modelling uses a similar approach, but instead replaces human memory with neural vectors that represent episodic knowledge and memory to infer salience. Dickens uses this to make suggestions and create interest or suspense. For example, after the first two chapters, the convict character \textit{Magwich} (who is unnamed at the beginning) disappears for a long part of the novel. In the beginning, \textit{Pip} helps get him a file and whittles for removing his chains. When \textit{Magwitch} appears much later in Chapter 39, he is described but not identified and is holding a file. Since the file played a key part earlier in the story, the reader immediately thinks of the start of the book and creates a question in their mind of whether there is a connection. Then the man is revealed later to be the same one. At the beginning of the book \textit{Magwitch} also demands food from \textit{Pip} else he would \textit{cut out} his liver. This makes \textit{Pip} feels uneasy and creates a sense of hostility towards \textit{Magwitch}, as it reminds him and the reader of the violence at the beginning of the book. This example emphasises it is not just obvious plot points, because recall is based on close heuristic pattern matching, storytellers can use seemingly innocuous details to heighten the drama. When applied to literature, one feature of this frames approach is not just a single mental model. Rather as discussed with the \textit{eventfulness} concept the reader makes a mental model \citep{Schmid20176578,GERRIG2010} of each of the main characters as well as an overall concept for the narrative world as a whole.

There has been more recent work on these ideas that have used psychological reading studies to examine these ideas. For example, \textit{depth of processing} and \textit{levels of attention} \citep{10.2307/j.ctt1ddr7zh.5,10.2307/j.ctt1ddr7zh.6,sanford2012mind} of the reader is highly variable in terms of attention to detail, mental processes are highly dynamic, and attention shifts quickly. There has been more general work proposing narrative as the core component of memory and knowledge for general cognition \citep{leon2016architecture,szilas2015towards}. Related cognitive work describes narrative memory in terms of \textit{causal relations}
\citep{nla.cat-vn5410210,TRABASSO1985612}. Readers use causal events to connect events in a narrative, and these are central to how memories are represented and recalled. Casual events are events where \textit{A} has to occur for \textit{B} to follow. This aligns with Barthes' idea of \textit{cardinal functions} or \textit{nuclei} events as those that cannot be deleted without altering the story, and this gives them a central role in the narrative. \citet{van2000role} finds writers when writing continuations for existing works strongly favour causal relations. This casual idea is explored in the knowledge and memory salience work via modelling knowledge through high-level plot \textit{fragments} or \textit{vignettes} of the major causal events in the stories.

Of most relevance in this line of work is that while storytelling can be described in terms of formal theory, the experience of reading depends on cognitive processes that are far looser and more approximate. This conceptually has more in common with the neural approaches described in the next chapter and operationalised in the later evaluated model than more formal rule-based structural approaches.

\chapter{Machine Learning in Story Comprehension}

\label{chap:backgroundml}

\section{Introduction}

The last chapter reviewed the concepts of suspense, salience, and important links between cognitive processes and comprehension of narratives. This chapter looks at the recent deep learning models and applications relevant to story comprehension and generation tasks. First, it reviews more traditional AI planning approaches. Then it covers more recent neural methods, including LSTM, VAEs and transformers, their qualities, and applications. This chapter reviews an underexplored area using vector spaces directly for analysis and inference. The chapter concludes by examining the weaknesses of supervised approaches and how integrating unsupervised methods with theory could provide a more productive approach to modelling narrative concepts.

\section{Symbolic Event Systems}

\textit{Structural} models such as those by Barthes' or Chatham are a natural fit for computational since they typically define specific roles on types of events, and relations between events. As such they are amenable to pattern match, slot filling, and reasoning methods that dominated earlier AI research methods. A story can be described as a hierarchical graph of events. These hierarchies can be divided into \textit{cardinal functions} or \textit{nuclei} that are the important function that progress the plot, and subordinate \textit{catalyser functions} or \textit{satellites} that fills in the gaps. This distinction depends on the readers psychological understanding of the narrative world and what constitutes an important change, which in turn depends on the readers mental model of the main characters, their goals and the challenges they face in achieving them. Given this formal theory, it's not surprising that earlier computational work on stories both for comprehension and generation has developed trying to formalise event graphs and/or character goals and challenges.

\begin{figure}[htbp]
\centering
\includegraphics[width=\textwidth]{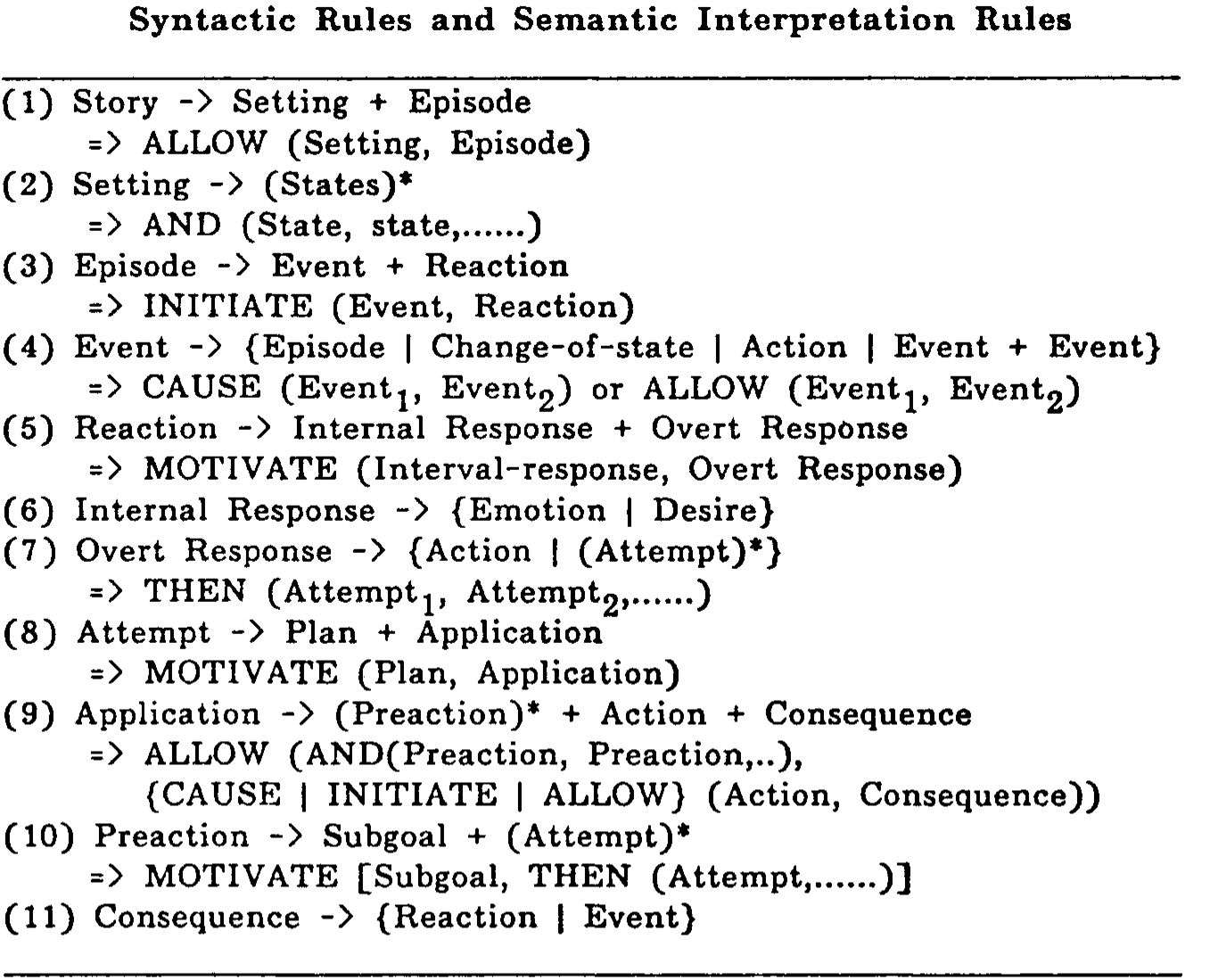}
\caption{Story grammar reproduced from \citet{RUMELHART1975211}}
 \label{fig:rumelhartgrammar}
\end{figure}

In story generation, the initial focus of research, unlike most of the current deep learning systems, was on non-learned systems. \citet{RUMELHART1975211} and \citet{THORNDYKE197777} defined a formal grammar of events that define a stereotypical story plot. Figure \ref{fig:rumelhartgrammar} shows the rules of story grammar. The grammar formally defines a story using \textit{Settings} and \textit{Epsiodes}. \textit{Episodes} are defined as  consisting of \textit{Events} and \textit{Reactions}. \textit{Reactions} motivate \textit{responses} that may motivate \textit{plans} and \textit{applications}. \textit{Applications} have \textit{actions} (that may require contain sub-actions and goals) and \textit{consequences}. \textit{Consequences} recursively define \textit{events} and \textit{reactions}. The scheme allows a hierarchical composable narrative based on actions taken to fulfil goals, which may cause reactions that require new actions or change goals.

\begin{figure}[htbp]
\centering
\includegraphics[width=\textwidth]{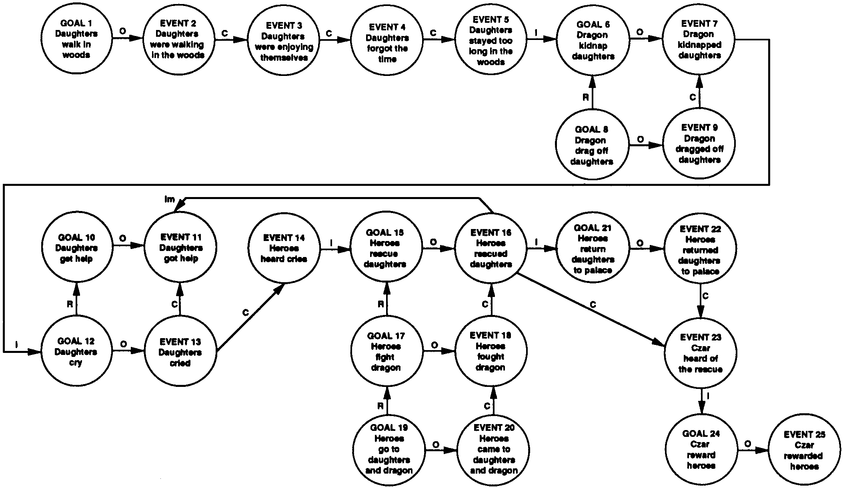}
\caption{QUEST conceptual graph from \citet{Graesser1991QuestionAI} of the plot from \textit{the Czar and His Daughters.}}
 \label{fig:quest_graph}
\end{figure}

\begin{figure}[htbp]
\centering
\includegraphics[width=\textwidth]{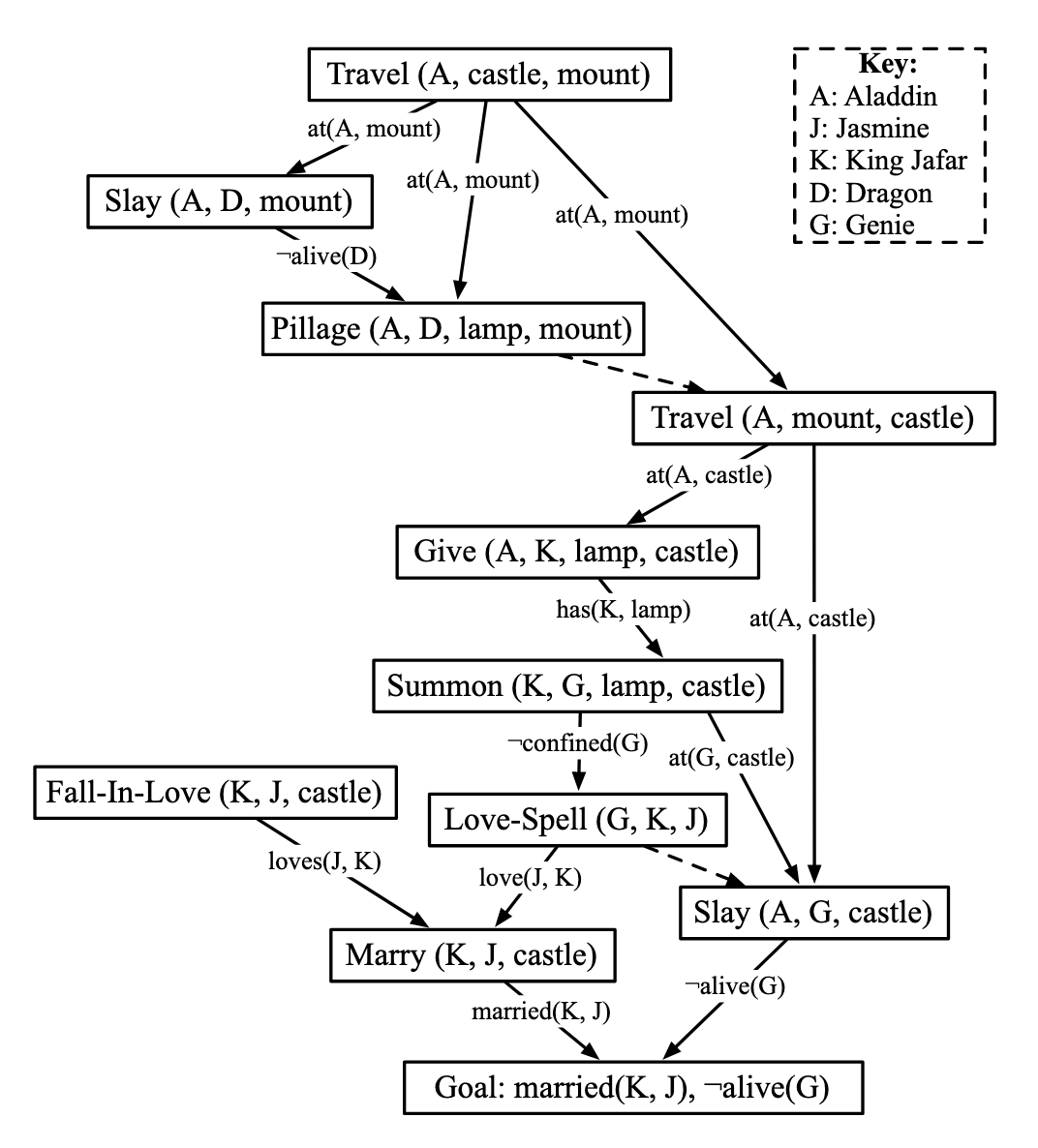}
\caption{A story plan reproduced from \citet{10.5555/1946417.1946422} from \textit{Aladdin}}
 \label{fig:reidlpocl}
\end{figure}

The story grammar approach was critiqued by \citet{blackgrammars} because the rewriting grammar approach could not generate all stories while also being capable of generating non-stories; these are similar criticisms to generative syntax grammars as the model is derived from. \citet{Black1980STORYUA} see stories as a form of problem-solving and about real-world entities and their goals.  \citet{s15516709cog0403} proposed approach is based on the semantic understanding of real-world events rather than linguistic structures as in story-grammars. This approach can be thought of as a cognitive planning system where characters have goals, hierarchies and paths to realising them. It closely related to ideas of cognitive narratology outlined in Chapter \ref{chap:backgroundtheory}. QUEST \citep{Graesser1991QuestionAI} is a question answering system based on similar concepts, see figure \ref{fig:quest_graph} for an example graph. \textsc{QUEST} uses these directed causal graphs to narrow down the answers to questions for story comprehension rather than story generation. \textsc{Universe} \citep{Lebowitz1985StorytellingAP} developed this planning approach into a system that could generate soap operas using a hierarchical planning approach by decomposing sub-goals using schema fragments that represent parts of a story. Goals have pre-conditions, and so a greedy planning system can chain together these prerequisites via state transitions into a graph that achieves the goal. Skipping forward to more recent work taking this approach \textsc{Fabulist} \citep{10.5555/1946417.1946422} tries to work around the problem with earlier planning systems which is that decomposing top-down goal systems produces unrealistic characters and plots because characters goals are often in conflict and not aligned. \textsc{Fabulist} instead uses a IPOCL (Intent-based Partial Order Causal Link) planning mechanism that allows for multiple goal and intent hierarchies for each character as well as for the overall plot. The story is written using a multistep process by iterating over possible actions and evaluating whether they resolve \textit{flaws} in reaching the goals and intents of the various characters and the overall story to produce a coherent narrative. See figure \ref{fig:reidlpocl} for an example.

Intuitively these planning approaches seems to embody the \textit{reader as problem solver} notion. Taking the example of the \textit{Great Expectations} opening: \textit{Magwitch} is an escaped convict who wants to getaway. What he needs to escape is to get his manacles off. To to do this, he threatens to kill \textit{Pip} if he doesn't help him. This fear causes a new goal for \textit{Pip} to save his skin. This leads to the actions of \textit{Pip} running off to steal tools from his Blacksmith Uncle and food from his house to help him.   Returning to \textit{Magwitch} enables him to break his chains and escape temporarily. This act of kindness from \textit{Pip} causes the reaction of \textit{Magwitch} to have gratitude for \textit{Pip}. This leads him later to become the anonymous benefactor to \textit{Pip}, a crucial suspense point later in the novel. Later in the book \textit{Miss Haversham} then pretends to be this anonymous benefactor to manipulate \textit{Pip} for her intentions. This seems to embody stories as a planning system by intuitively modelling the complex dynamic interactions of a story's intents, goals, and actions for different characters.

Although these rule-based symbolic methods seem intuitively appealing, there are limitations to using them for the tasks in this thesis. The main comprehension tasks of this thesis are on inferring salience and suspense in general story texts of any genre. Salience is inferred on longer book and movie script length works. \textsc{QUEST} is one example where a graph structure built on an explicit model with goals, pre-conditions, and events can be used for a comprehension task, in this case, question answering. In the computational suspense literature \textit{Dramatis} \citep{DBLP:conf/aaai/ONeillR14} and \textit{Suspenser} \citep{Cheong2015SuspenserAS} are the most relevant systems. These are rule-based story generation systems that rely on a similar concept of the reader as a problem solver. In \textit{Dramatis} suspense is defined as a plan to escape a negative outcome. A planning system manipulates the path to avoid the negative outcome with the increased cost or a more elaborate path corresponding to more suspense. Both these systems rely on an explicit symbolic approach.  It's theoretically possible that such as rule system could also be adapted to evaluate suspense or salience on existing stories. According to the given schemas, one idea would be to pattern match against the text to build the plot graphs from the bottom-up with the characters, goals, sub-goals, etc. The graph could then be used to infer from paths which of the events in the story are salient or how suspenseful the actual path of the story is compared to plausible alternatives. The problem is that any method that relies on hand-crafted rules of complexity is unfeasible beyond a minimal domain. Stories encompass a huge variety of circumstances. From the examples used in this thesis \textit{Mr Bean}, \textit{Only Fools and Horses}, \textit{Rocky}, \textit{Pretty Women}, \textit{Gone Girl}, and \textit{Great Expectations} encompass as enormous variety of circumstances. The symbolic rules need to cover these only, and alternatives include story-specific plot points and understanding everyday events such as getting a taxi to a destination. hen there are fantasy, sci-fi, or historical worlds that may contain any number of novel elements that don't exist or don't prefer the real world. As a result, this is an intractable task, so recent planning approaches have focused on learning event structure.

\section{Learning Event Systems}

One way around this problem is to try and learn chains of narrative events. More complex symbolic structures, hierarchies of goals, pre-conditions, and actions would be hard to learn, but there has been success in learning simple causal chains in narratives. \citet{chambers-jurafsky-2008-unsupervised} learns unsupervised narrative chains. The method preprocesses a dataset by extracting named entities and their coreferences to identify the principal protagonists. The model takes sentences to split into events which are the main verbs. The goal is to learn a partial order of tuples for the character and the event. This is done via clustering the verbs using the PMI (Pointwise Mutual Information) of event counts to determine the most frequent events across a whole dataset. A multi-stage pipeline based on an SVM (Support Vector Machine, \citealt{DBLP:journals/ml/CortesV95}) and trained classifier is used to learn the statistical properties of event structure to determine the temporal ordering of events. Clustering on the  graph of all events can then produce discrete narrative event chains (as well as a global chain) for typical scripts such as a \textit{restaurant}, a \textit{court prosecution}, or \textit{employment}. The chains have alternatives paths for example, in a \textit{court prosecution}: \textit{charged} $
\rightarrow$ \textit{indicted} $
\rightarrow$ (\textit{convicted} or \textit{acquitted}). \citet{chambers-jurafsky-2009-unsupervised} extends this work to add FrameNet \citep{10.3115/980845.980860} style semantic role label slots \textit{(subject, object, prep)} and clustering over all the event slots rather than just the verb. One downside is that the evaluation is only on a cloze task and is not applied to downstream comprehension or generation tasks. 

For story generation, \citet{mcintyre-lapata-2009-learning} extract event chains from a corpus in a similar manner as Chambers and Jurafsky, although the event chains are more localized. This knowledgebase is used as part of a beam search to compose together events into a story using a mutual information metric. Then a surface level realiser is used to generate fulls sentences from the abstract event plan. \citet{mcintyre-lapata-2010-plot} uses a similar approach but apply an evolutional algorithm rather than a beam search. These event chains are learnt within a relatively narrow domain of fairytales, and the models are only used to generate short stories of five sentences. \textsc{SCHEHERAZADE} \citep{10.5555/2891460.2891543} rather than learn from an existing corpus crowdsourced stories on a given topic from AMT (Amazon Mechanical Turk).  \textsc{SCHEHERAZADE}'s improvements focus on being able to determine optional or mutually exclusive events. This is a problem for Chambers and Jurafsky in that their event chains often cannot determine mutually exclusive events. For example, in the learned \textit{court prosecution} chain both \textit{acquittal} and \textit{conviction} imply \textit{sentencing} which isn't how courts work. Optional events are also important since a typical event chain such as a \textit{court case} will have lots of events that may or may not occur depending on the type of case, as will most sequences of events that occur in stories. Once again, this graph method is based on mutual information and is used to generate story plans, although not full-text stories.  \citet{pichotta-mooney-2014-statistical} use a similar mutual information approach to learn general scripts or chains of events but not specifically for stories. Unlike the other mentioned research, they focus on dependencies between named entities rather than counts of events and handle multi-protagonists better than other methods. The Chambers and Jurafsky, or McIntyre and Lapata methods often can mix up named entities and produce noisy chains. 

The problem with \textsc{SCHEHERAZADE} and other similar methods is they depend on collecting a huge number of stories on a single subject, with variations of the same plot. With mutual information methods a significant overlap is needed to calculate mutual exclusion, optional events, or temporal order. The need for lots of similar examples is also why the domain is limited to \textit{fairy tales} in Mcintyre and Lapata. This substitutes the problem of building a huge amount of hand-crafted rules requiring many examples of similar events of interest for training. Again, this is not feasible if the goal is to apply methods to a text of any genre and open-domain works where there may be a few or single examples of a given event chain. Performance is also not strong as in \citet{pichotta-mooney-2014-statistical} the recall at top-10 is only $24.5$. The problem with these methods is that they also tend to be brittle.  Matching of verbs or more complicated event slots such as in Pichotta et al. is noisy. It can easily overlook synonyms or paraphrases, or overmatch with homonyms or confuse named entities. This can lead to error-prone chains that don't match real-world events, that mix topics and protagonists. More recent models have instead tried to combine symbolic rule-based methods with recent neural networks to improve the performance of event chain extraction or implicitly try and learn this as part of a deep learning model. These hybrid methods will be returned to later in the chapter.

The last two sections have introduced symbolic rule-based systems for story comprehension and generation tasks. This thesis has three basic models: A hierarchical model built on top of a neural language model, a variational model, and an encoder/decoder with external knowledge and memory. The approach taken in all three is the opposite of those described in both of these sections. Rather than encode scripts into an explicit symbolic rule structure like much of the recent trend in deep learning the approach is to learn implicit weights within a latent vector space. The technical details of vector spaces are in the following sub-sections, but there are several justifications for the approach related to symbolic systems.

Take the examples used in this thesis. With more formulaic \textit{Hero's Journeys} stories such as \textit{Rocky} and \textit{Pretty Woman}, it's possible to imagine that the basic plot templates can be codified as Propp tried to. However, the Disney animated movies the \textit{Lion King} and \textit{Cinderella} can also be described as following the \textit{Hero's Journey} plot structure. Although it is possible to align plot elements at a high level with the same role, the contexts differ wildly. Manually defining rules at this high level is entirely impractical. Any statistical learning approach that relies on surface-level lexical features will also struggle to identify more abstract causal patterns. In practice, though the stories may also follow a general template at the surface level of the text, it is particularities of the narrative world (characters, places, motivations, and intent) that are important in writing or comprehending a coherent and exciting story. At the low level, there is an almost unlimited number of plausible scripts. 

The thesis's approach is that semantic understanding via more generalisable vector spaces is more promising than learning enormous graphs of possible events and the rules that connect them. Taking the two simplest comedy examples: In the \textit{Only Fools and Horses} falling at the bar example, this kind of slapstick comedy is tough to codify into rules. Falling can be slapstick in many situations, but it would be tricky to define when it would be an optional event. With \textit{Mr Bean} it's unlikely that washing lettuce in a fountain or drying it in a sock would be captured in an event chain, but it captures an essential part of the comedy. \textit{Gone Girl} and \textit{Great Expectations} are more elaborate in their plot than most other works. Still, they and the discussed theory from the last chapter emphasise that stories are often about surprising and unusual situations that challenges expectations. In both generation and comprehension, one of the central challenges of storytelling is that it needs both expected events and novelty. \citet{Schmid+2017+229+246} discusses this in \textit{Eventfulness and Repetitiveness: Two Aesthetics of Storytelling}. While there is much work required to improve deep learning methods, they hold the best promise at the moment for modelling semantic meaning. Before considering the relevant deep learning methods and applications, a short note on alternative architectures for story comprehension and generation. 

\section{Case Based Methods}

\begin{figure}[tb]
\centering
\includegraphics[width=0.5\textwidth]{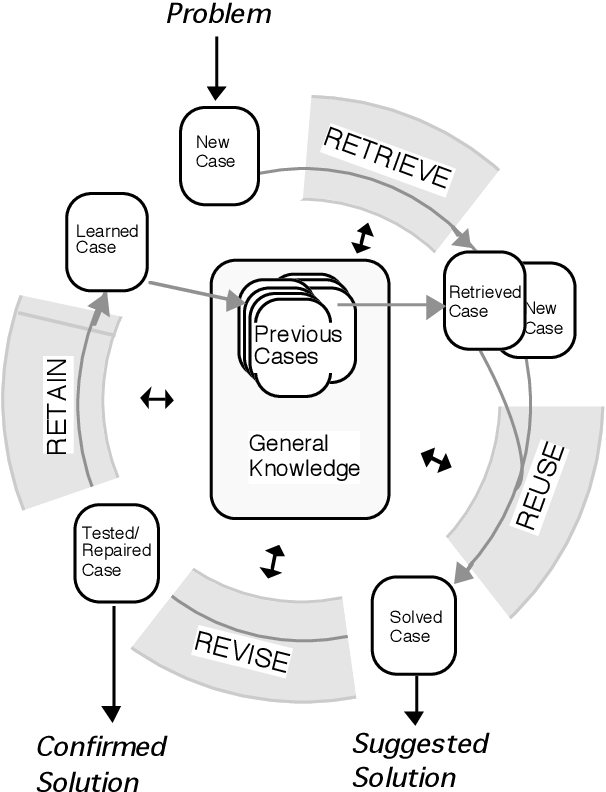}
\caption{Case based reasoning reproduced from \citet{Aamodt1994CaseBasedRF}.}
 \label{fig:casereasoning}
\end{figure}

Instead of a rule-based inference approach, the knowledge representation approach in this thesis is based on a traditional case-based reasoning \citep{Aamodt1994CaseBasedRF} updated to use new neural models. Figure \ref{fig:casereasoning} shows a typical case-based reasoning architecture. Unlike a rule-based approach that is resolved through inductive inference, case-based reasoning relies on retrieving similar cases from memory and then using these as a template for understanding the situation or a solution. Within a story context, whether for reading or generation, text can be added as a case to the knowledgebase and then recalled for later reasoning about the story.

The method more closely resembles the episodic recall orientated cognitive narratology research. To illustrate the concept when applied to stories, when we first encounter \textit{Rocky} boxing in a rough venue called the \textit{Cambria Fight Club}, relevant cases that might come to mind is Boxing and what typically happens at a boxing match, the punches themselves, sweat and blood, the referee, the shouting of the crowd, but also detail of the environment the Boxing is taking place in. Comparing recalled cases with what is being watched, the viewer may infer from the grotty dimly venue and a small rough-looking crowd and the brawling style of boxing that the fighter they are watching is a  small-time, struggling one.  This is in contrast to the movie introducing the champion \textit{Apollo Creed} in a glitzy venue with flashing paparazzi bulbs and being fawned over by broadcast journalists; we infer he is successful. Neither the failure of one or the success of the other is stated. Rather we are meant to imply from different cues in the situation:  A venue with faded and peeling posters and paint is somewhere that recalls other worse off or poor circumstances. The paparazzi suggest a celebrity or someone important. The point of this example is that traditional case-based reasoning usually relies on looser and fuzzier matching between the properties of recalled cases and the given situation, and this matches the looser cures and associations required for comprehending stories.

There's quite a body of work of case-based methods being used for story generation. \textsc{Minstrel} \citep{turner1993minstrel} is an early well-known example. The framework is defined as Transform-Recall-Adapt Methods (TRAMs). Prompts are transformed into search criteria used to recall cases, adapted for the specific circumstances to generate a coherent continuation. The intuition behind this is that retrieving, combining and adapting multiple cases from knowledge produces a creative leap from existing knowledge that predefined rules do not and is also more similar to cognitive processes. The rationale for adapting the RAG (Retrieval Augmented Generation; \citealt{NEURIPS2020_6b493230}) is similar. \textit{ProtoPropp} \citep{Gervs2005StoryPG} uses cases defined by Propp functional fairytale roles. \citet{riedl2008story} and \citet{Riedl2010StoryPC} use fragments of existing story plots called vignettes and integrate them into a story generation system that also combines these recalled vignettes with partial ordering planning. \citet{Zhu2010TowardsAS} and \citet{Zhu2014ShallIC} refer to the approach as analogy-based story generation. The emphasis is on the common knowledge rather than on plot fragments; knowledge such as \textit{Honeysuckle is in the backyard}, \textit{Beehives are by Honeysuckle}, or \textit{Bees are Dangerous}. The system links user input such as \textit{As a child, Julian would often sniff the Honeysuckle in the backyard} with this common knowledge in a graphical structure. It then transforms it into a target situation that generates a continuation of the user-provided prompt.

Intuitively, case-based reasoning is a more appealing way of modelling narratives and aligns more with cognitive narratology ideas on how readers or viewers comprehend stories than highly formalised symbolic planning systems. However, the referenced case-based reasoning models all share many of the same limitations. All the mentioned systems are non-learning. This means that there needs to be a vast number of hand-engineered rules for a given domain, making a broad domain impracticable. The reviewed systems also have highly structured knowledge representations that require specific template forms and slot filling. This method also requires structured transformation to convert natural text input as a user prompt into a structure retrieved from the knowledgebase. Typically retrieved knowledge then needs to be joined with the context, usually using some graph algorithm that matches shared properties. There is a transformation using typed slot filling to make the cases relevant to the story context. Rather than use this structured approach, the ideal would be to use a raw representation and have a model learn to represent and retrieve these cases as and when needed. This thesis uses neural approaches based on transformers that are introduced in later chapters. Perhaps surprisingly, most of the work in this area has focused on the question and factual domain. The advantage of these methods is that raw text can be used as a knowledgebase to represent plot vignettes or factual knowledge. The training process can learn to retrieve the most salient cases and how best to incorporate this knowledge in story generation or comprehension.  The advantage of this is that it's generic in only requiring text as a source. It's also flexible in the representation as it's not always obvious what the most relevant knowledge and a hand-engineered system must decide apriori. Going back to the \textit{Rocky} reference, seemingly innocuous cues can be the most important in the situation.  An example referenced in the last chapter was how the character of \textit{Magwitch} first met \textit{Pip} as an escaped convict while threatening him to get some whittles and a file to cut off his chains. When he appears some 20+ years later in the source of the story, he is holding and using a file, which suggests who he is even when little else is given to infer this. This is a literary trick by the author. However, while episodic memory in stories is often about the previous appearances of the same protagonists, there are subtler cues that wouldn't be part of conventional templates. These cues can be as much about the general situation and mood as the specific actions of the main protagonists. The decaying club the fight takes place in at the start of \textit{Rocky} or the foggy country churchyard on the marshes at the start of \textit{Great Expectations} can trigger relevant recalls in memories as much as explicit actions. Learned representations at least have more of a chance of picking up these nuances while also being more flexible in how this knowledge is applied. The literature in this section shows a long history of using these methods in creative story writing. The primary use of these methods in the thesis is comprehension. As has been covered, though, the comprehension models are based on expectations, so the same recall informs these expectations and notions of suspense, surprise, and salience.

\section{Language Models}

The heart of the methods in this thesis are built on top of language models. Earlier language models Eqn. \ref{eqn:wordcondprob} shows the concept of a conditional language model for a word $w$ for a token index $t \in T$, defines the conditional probabilities of a word conditioned on all the previous words using a chain rule. N-gram language models are defined over continuous  sequences of n items from a sequence of words, \textit{bigrams} $P(w_{t}\mid w_{t-1})$, \textit{trigrams} $P(w_{t}\mid w_{t-1},w_{t-2})$, \textit{quadgrams} $P(w_{t}\mid w_{t-1},w_{t-2},w_{t-3})$, etc. For example a \textit{trigram} count for the text \textit{ memorable raw afternoon} then the probability calculation is $P(\mathrm{afternoon} \mid \mathrm{memorable}, \mathrm{raw} ) P(\mathrm{raw} \mid \mathrm{memorable}) P(\mathrm{memorable})$. The apparent limitation with statistical n-gram language models are mainly sparse, and most n-grams will have zero counts and zero probability.  These statistical language models have proved effective especially with large volumes of data as argued by \citet{halevy2009unreasonable}. To redress this, smoothing models such as interpolation by \citet{JelMer80} or backoff by \citet{katz1987estimation} are employed to ensure non-zero probabilities. This thesis will skip over a survey of methods since they are not relevant except to they say they are an attempt to handle the problem of sparsity.

\begin{myequation}
P(w_1^T)=\prod _{t=1}^{T}P(w_{t}\mid w_1^{t-1})
\label{eqn:wordcondprob}
\end{myequation}

The heart of the methods used in this thesis are built on top of neural language models. The modern incarnation of neural language models (LM) started with \citet{10.5555/944919.944966}. The essential problem addressed by neural language models is the curse dimensionality. In an n-gram language, the number of parameters required is a permutation of the length of the n-gram and the vocabulary needed. The neural LM uses a feed-forward neural network to learn representations for words in a much smaller lower dimensionality real vector space, $\mathbf{R}^m$ where $m$ is the dimensionality. Training is by using gradient descent to maximise the log-likelihood probability of the sequence of word tokens. The reduced feature space of $m$, which in this early LM is 30, 60, or 100, compresses the information for each word reducing sparsity. The intuition behind this is that the model can learn to distribute the representation across features so that similar words are closer to each other within the vector space. Learning is also structured in a self-supervised way. The representations can be learnt only from the text data rather than requiring explicit annotations that may be infeasible for large and wide-ranging datasets and domains. As an aside, although this language model has been surpassed by much larger and improved architectures to be discussed, recent work had found that feed-forward networks can still produce reasonably competitive LMs \citep{sun-iyyer-2021-revisiting}.

Though the models used are more sophisticated than this LM, the essential rationale is the same. In the review of symbolic rule-based systems and case-based reasoning systems, it was argued that one of the main weaknesses of these systems is they rely on brittle matches. They typically match slots against approved types such as part-of-speech tags, word lists, named entities, etc. The major improvement needed is more semantic understanding that can model the important aspects of narrative comprehension. At the word level, neural LM models often promise that similar concepts will be close to each other in the semantic space, for example, with \textit{Rocky} the sport of \textit{Boxing} will be closely related to \textit{Glove}, \textit{Fight}, \textit{Ring}, \textit{Sweat} whereas words related to the glamorous clothing in \textit{Pretty woman} such as \textit{Dress}, \textit{Hat}, \textit{Necklace}, etc would be expected to be distant from \textit{Boxing} but close in geometric space to each other.

\begin{figure}[htbp]
\centering
\includegraphics[width=\textwidth]{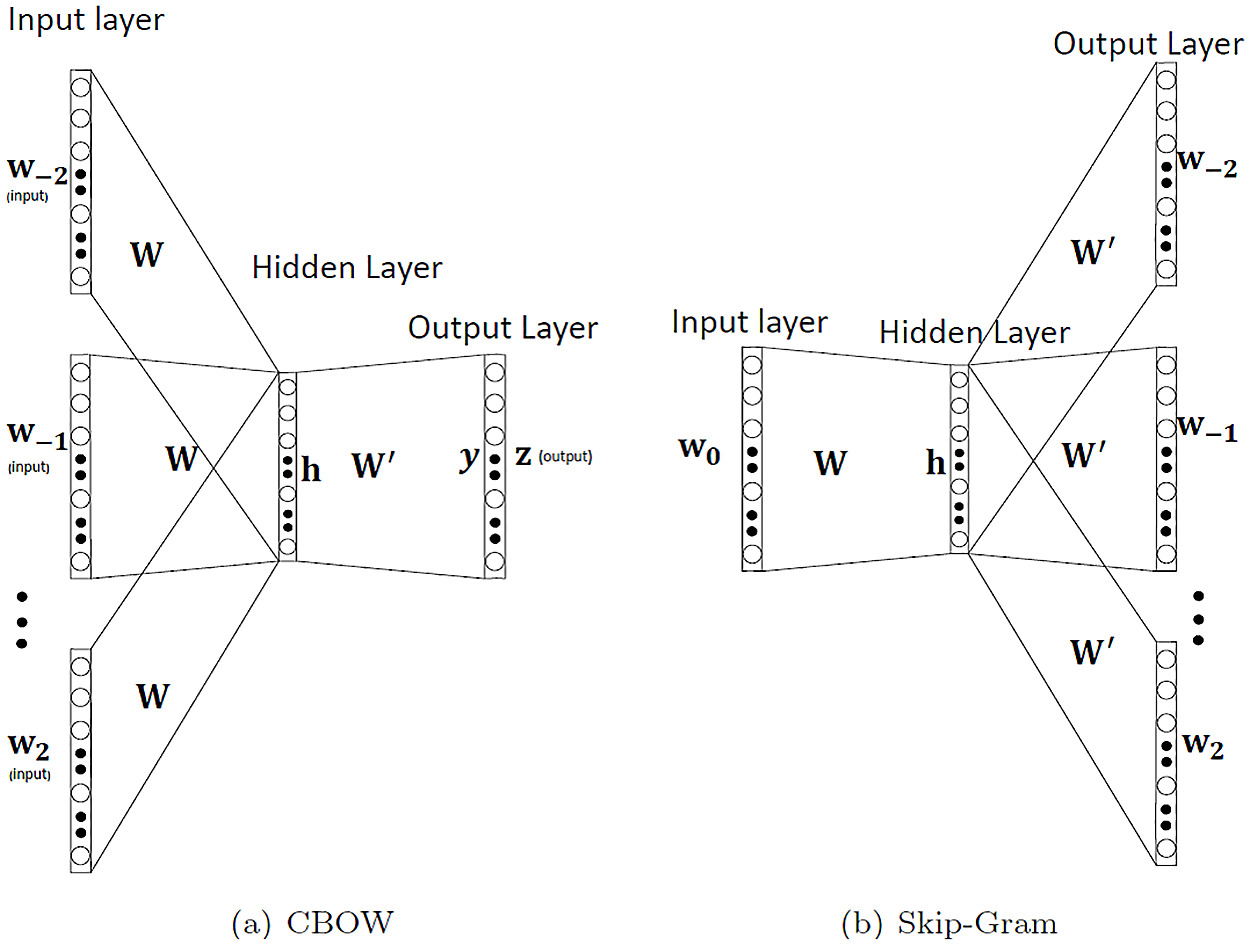}
\caption{\textit{word2vec} models reproduced from \citet{DBLP:journals/corr/abs-1301-3781}}
 \label{fig:skipgramcbow}
\end{figure}

\begin{myequation}
J_\theta = \frac{1}{T}\sum\limits_{t=1}^T\ \text{log} \space P(w_t \: | \: w_{t-n} , \cdots , w_{t-1}, w_{t+1}, \cdots , w_{t+n})
\label{eqn:cbow}
\end{myequation}

\begin{myequation}
J_\theta = \frac{1}{T}\sum\limits_{t=1}^T\ \sum\limits_{-n \leq j \leq n , \neq 0} \text{log} \space P(w_{t+j} \: | \: w_t)
\label{eqn:skipgram}
\end{myequation}

Neural LMs were extended later by \citet{mikolov2010recurrent} and \citet{mikolov2011extensions} to employ RNN (Recurrent Neural Networks; \citealt{rumelhart1985learning}). The first model to attract widespread attention was \textit{word2vec} \citep{DBLP:journals/corr/abs-1301-3781} which uses a multi-layer feed-forward architecture (see Figure \ref{fig:skipgramcbow}). The model in the case of the (a) CBOW model learns to predict a missing word from a context of a window of $n$ words on each side, while (b) skip-gram is the opposite in being given only a single context word and predicting multiple window words on each side. This objective is defined in eqns. \ref{eqn:cbow} and \ref{eqn:skipgram}. Though surpassed by much newer models used in this thesis, the main insight is that by predicting missing tokens that are masked out, the model learns a representation of meaning. There are variants such as classifying correct or incorrect alternatives or reranking \footnote{There are other approaches such as \textsc{Glove} \citep{pennington-etal-2014-glove} that learns embeddings via pointwise mutual information ratios, but that isn't relevant to the rest of this thesis.}, but the general strategy is to learn meaning via inferring masked parts of the text. To go back to the \textit{Rocky} example because \textit{boxing} often includes \textit{gloves} this trains the representation to be close to each other in semantic space. Because words can be used in multiple contexts \textit{gloves} will also be close in the vector space to other words such as \textit{dress} or \textit{hat} mentioned as part of the \textit{Pretty Woman} scene. To be effective, the model needs to encode information onto different dimensions to encapsulate these different usages so that the use in a sporting context can be distinguished from the glamorous dressing up one. Note that other related words will distinguish the context, such as \textit{fight}, \textit{punch}, \textit{knockdown} or \textit{necklace}, \textit{elegant}, \textit{boulevard}. Any distance measures can be used to measure similarity but it has been more common to use cosine distances which consider the angular similarity of vectors and not magnitude, or some form of geometric distance such as Euclidean (L2) and Manhattan (L1) distance. \citet{10.5555/2999792.2999959} and \citet{mikolov-etal-2013-linguistic} observe useful geometric patterns in learnt embeddings representations. For example, there are similar directional projections between countries and their capital cities such as \textit{China} and \textit{Beijing}, and \textit{Portugal} and \textit{Lisbon}. These semantic spaces also support the composition between vectors. For example, if the vectors \textit{Russia} $+$ \textit{River}  $=$ \textit{Volga River}, while more famously \textit{King} $+$ \textit{Women}  $=$ \textit{Queen}. This analogy work shows that embedding representations can encode different aspects spatially into alternate dimensions of a vector. Whereas typically, rule-based systems align based on templates, slots, and matching word types/patterns/lists, word embeddings represent meaning through distances and relative positions in the vector space. These compositional \textit{distributed semantics} or \textit{semantic space} models have been formalised by those including \citet{clark2008compositional}, \citet{DBLP:journals/corr/abs-1003-4394} , and \citet{clark2015vector}; this work has defined how vector spaces and operations can map to category theory and predicate calculus and so linguistic relations can be supported by underlying vector representations. Symbolic relations become operators on the vector space, allowing meaning to be represented by dimensions and transforms on the space. A variety of empirical work by \citet{pado2007dependency}, \citet{mitchell2008vector}, \citet{mitchell-lapata-2009-language}, \citet{mitchell2010composition}, \citet{baroni-lenci-2010-distributional}, \citet{grefenstette-2013-towards} and \citet{fyshe-etal-2013-documents} have demonstrated this composition applies to a variety of tasks. These compositions scale to phrases and sentences. The literature discussed represents earlier neural language models such as \textit{word2vec} that popularised the approach and work on distributional semantics that were often based on non-neural methods such as PMI (Pointwise Mutual Information). There is a link between these as \citet{DBLP:conf/nips/LevyG14}, \citet{Li2015WordER} and \citet{NEURIPS2019_23755432} analyse \textit{word2vec} as a method to factorise a matrix and the negative sampling loss function of \textit{word2vec} encodes PMI. 
Using semantic vectors to understand narrative progression is also a more complicated task than this linguistic phenomenon. Nevertheless, the motive and the advantages are the same as the transformer based methods described in the next section. \citet{clark2015vector} outlines the advantages of semantic space models over hand-built ontologies as being a) being created automatically from the data; b) they can represent gradations of meaning, and c) they map well to cognitive sensitivities to distributional data. 

\section{Transformer Language Models}

The last section covered \textit{word2vec} which popularised neural language models and compositional \textit{semantic spaces} and formalised how vectors can encode meanings and use vector operations to compose longer phrases. All the neural LMs used in this thesis are based on the now dominant transformer model created by \citet{NIPS2017_3f5ee243}. There is much intermediate work between \textit{word2vec} and \textit{transformer} LM models used in this thesis that use RNN (Recurrent Neural Network; for example \citealt{Peters:2018}) or CNN (Convolutional Neural Network; e.g. \citealt{Dauphin2017LanguageMW}) architectures. This review will skip over much of this as all the experimental work relies on transformers as they demonstrate the strongest recent performance for language models.

The motive for using language models such as \textit{word2vec} is to automatically encode semantic information into vector space to produce word representations. However, one limitation with \textit{word2vec} models is that it's limited to a dictionary of simple word tokens or deliberately chosen bi-word or tri-word tokens for selected noun or verb phrases. The discussed vectors composition models demonstrate that vector spaces can compose semantic relations across longer phrases and sentences. This earlier work was often limited to standard vector operations such as addition or tensor multiplication, possibly with some weightings in composition. The main advantage in a \textit{transformer} model or the RNN, CNN alternative languages models is they can learn more complicated linguistics or semantic interactions and compositions dynamically. Just as a symbolic rule system will have higher-order rules, likewise a hierarchical \textit{transformer} (or other) can, in principle, learn high-order transformations of the space to represent higher-order concepts and relations. These semantic space representations for longer units such as events, sentences, paragraphs, or whole documents.

\begin{figure}[htbp]
\centering
\includegraphics[width=0.7\textwidth]{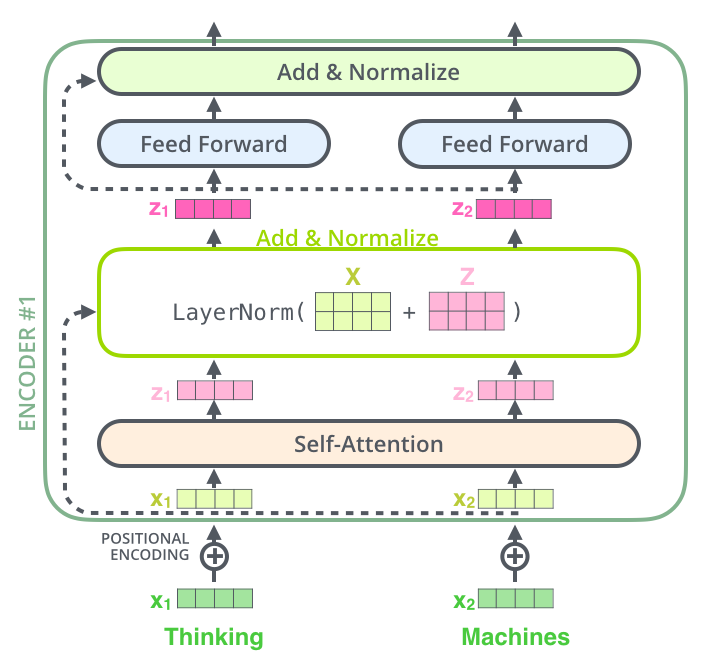}
\caption{Transformer attention block reproduced from \citet{JayAlammar.2018}}
 \label{fig:transformerblock}
\end{figure}

The core of \textit{transformer} models is the attention block which is illustrated in figures \ref{fig:transformerblock}. A \textit{transformer} takes an input token representation $x$. A variety of schemes have been used, but generally, most work uses a \textit{SentencePiece} style byte-encoding \citep{kudo-richardson-2018-sentencepiece}, a development of BPE (Byte-Pair Encoding; \citealt{Gage1994ANA}). Subword tokens are efficiently encoded onto a more limited vocabulary with the benefit that it makes output layers across the whole vocab more computationally tractable. Throughout the thesis, the default encoding of the pre-trained models LMs are used. This shall be called a wordpiece. $x$ tokens are passed through a stacked number of layers, with each layer consisting of multiple attention heads and each layer merging the results of the heads by concatenating the heads together and multiplying another learned weight matrix. The core of this is the self-attention unit. Because the self-attention mechanism does not capture order, it is customary to use additional positional embeddings in capturing either the relative or absolute position of a wordpiece within the context window.  In self-attention, each workpiece attends to all the other wordpieces in a given context window size, making the number of comparisons $n*n$ or complexity of $O(n^2)$ for a single head. Each attention head has three sets of weights and representations $W_qX = Q$ for the query, $W_k$ for the key and $W_v$ for the value weights. The output values vectors for applying an input vectors is $XW_q = Q$, $XW_k = K$, and $XW_v = V$. See eqn. (\ref{eqn:selfattention}) for the most commonly used scaled dot product attention. 

\begin{myequation}
Z = softmax\left(\frac{QK^T}{\sqrt{d_k}}\right) V
\label{eqn:selfattention}
\end{myequation}

$d$ is the dimension of the vectors, and so $\sqrt{d_k}$ is just normalising for vector dimensions and magnitudes. The main part of self-attention is applying a probability distribution via the softmax over all the other wordpieces in context and then using that for a weighted aggregated attention $Z$ over all the $V$ input vectors. The net result is that the model can learn which other wordpieces are relevant for a given position and how each workpiece token should be represented. Multiple heads allow for different $W_q$, $W_q$ and $W_v$ to learn to focus on different linguistics or semantic aspects. Stacking layers allows each layer in a transformer to learn from the abstractions from the layers below and potentially learn higher-level concepts. This makes the transformer a powerful latent space compositional model. There are other details such as the layer normalisation where the input $X$ is passed through and recombined with $Z$ and feedforward networks used to combine features that are skipped over. This review will not go into more technical depth as better sources will be given for this on \textit{transformers}, but will instead review the rationale behind inferring the salience and suspense ideas discussed in the last chapter, or story generation planning with models based on \textit{transformer} LMs or hierarchical models based on LMs.

\begin{figure}[htbp]
\centering
\includegraphics[width=\textwidth]{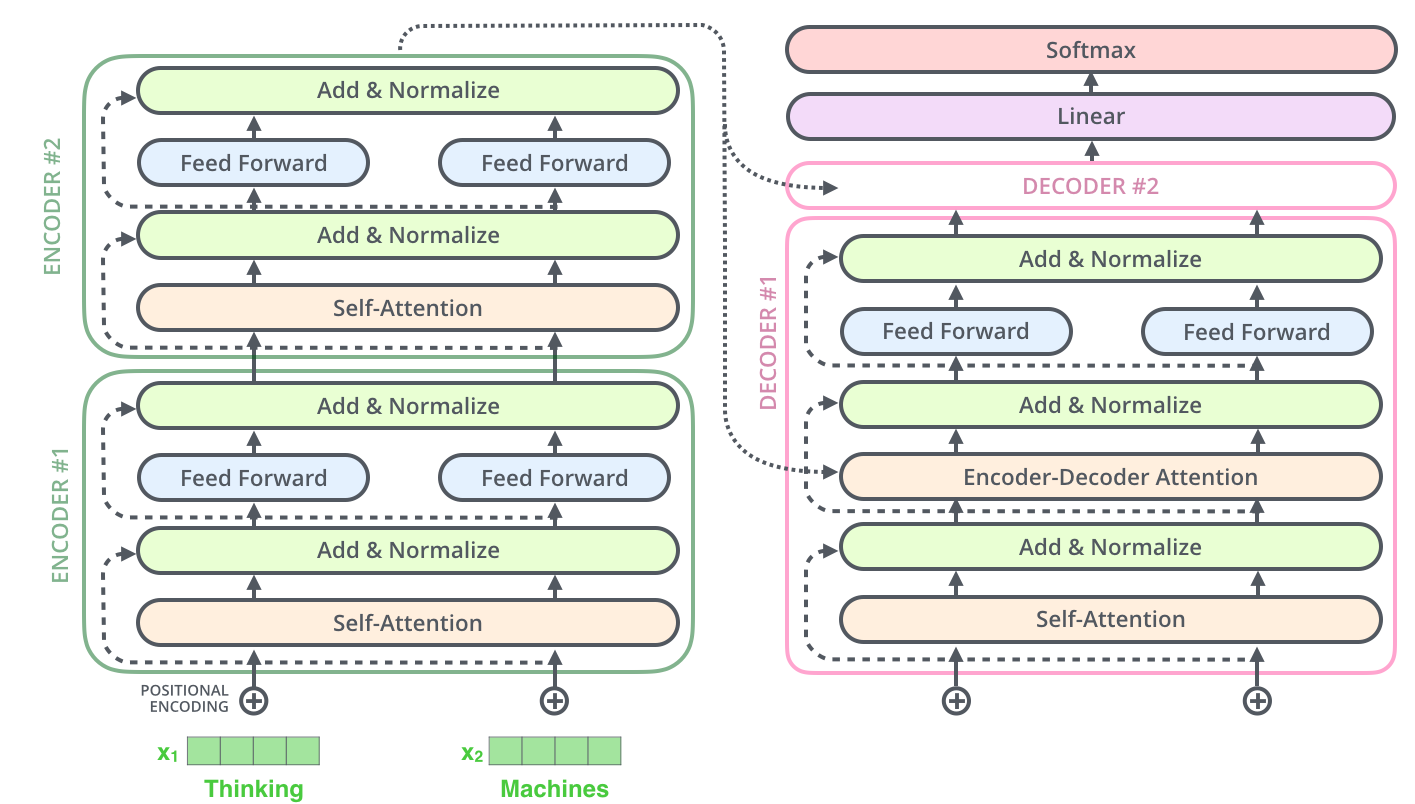}
\caption{Encoder/decoder architecture reproduced from \citet{JayAlammar.2018}}
 \label{fig:transformerencoderdecoder}
\end{figure}

Figure \ref{fig:transformerencoderdecoder} shows the Encoder-Decoder architecture introduced by \citet{NIPS2017_3f5ee243}. In this model \textit{transformer} layers are stacked up in the encoder with encoder representations from an input sequence. The decoder has a cross-attention mechanism to attend to any of the wordpiece tokens in the encoder and a separate attention mechanism to attend to tokens in the output. This architecture defines a universal sequence to sequence model where input text is mapped to output text; this is a common model in machine translation where the input could represent French, and the output could represent Chinese, for example.  The model also fits a range of other tasks such as natural question answering where the input could be the question and the output the written answer, or summarisation where the input could be the whole text and the output the summary. In training, the model learns to predict the output sequence and then can be used at inference time to generate the translation by iteratively generating a token. Or it can perform some other inference of current output, such as measuring the perplexity to gauge how expected the output is, which could, for instance, be used for a ranking task to select the most likely output. This combined Encoder-Decoder model is only one commonly used architecture. A decoder only model as exemplified by GPT 1-3 (Generative Pre-Training; \citealt{radford2018improving,radford2019language,NEURIPS2020_1457c0d6}) has been widely used and has achieved the state of the art results in text generation. Like \textit{word2vec} these LMs are trained via self-supervision. For a GPT language model the loss is auto-regressive in that the LM predicts the next wordpiece token out of all possible tokens. The loss is the cross-entropy loss of the predictions versus the wordpiece label. The losses for the base LMs of the thesis models are covered in more detail later in experimental chapters. A model can be an encoder only; \textsc{BERT} \citep{devlin-etal-2019-bert} and \textsc{RoBERTA} \citep{Liu2019RoBERTaAR} are  two highly influential encoder only models. In \textsc{BERT} the basic training loss is based on predicting randomly masked tokens. \textsc{BERT} also uses a sentence to sentence loss function that learns by predicting the next correct sentence from an incorrect sentence. The fundamentals of training \textit{transformer} LM models is the same as \textit{word2vec} in that it relies on the LM learning to inferring structural patterns or masked tokens in the text. For \textit{Skip-gram} this is inferring neighbour words in a window; for \textsc{BERT} this is inferring missing masked tokens. As per the discussed \textit{word2vec} literature, these filling in the gaps or structural manipulations allows the model to build representations that can capture lexical or semantic content. The core of much research in this area is combining or varying training objectives and tailoring the architecture to achieve improved performance on particular tasks or datasets. More recent \textit{transformer} models have moved back to the Encoder-Decoder models with \textsc{BART} \citep{lewis-etal-2020-bart}, with both \textsc{BERT} and GPT masking, and auto-regressive losses applied to multiple tasks. T5 \citep{JMLR:v21:20-074} models a variety of NLP tasks as a sequence-to-sequence learning task. Both models demonstrate strong performance in a variety of tasks and also the general trend in deep-learning to train for multiple tasks to improve performance for all the trained for tasks over specialised single-task models. \citet{hendrycks-etal-2020-pretrained} finds pre-trained LMs on larger and diverse datasets, trained on multiple tasks, can improve out of distribution robustness, and hence performance in other domains and tasks. The reasoning for improved multitask learning performance predates current neural models \citep{Caruana1998MultitaskL}: Noise patterns differ across tasks, and so a model that learns multiple representations simultaneously is more likely to learn a general representation and average out the noise; hence it acts as a form of data augmentation. It also allows the model to differentiate relevant from irrelevant features. Multitask learning also favours representations that can be shared across tasks and hence more generalisable, and can also regularise the model by preventing overfitting to the data. Most of these arguments also apply to the diversity of the domain and dataset as well as the task, which is why the trend has been to pre-train on more diverse datasets and tasks.
The overall strong performance of \textit{transformer} models, including all those mentioned on multitask benchmarks such as GLUE \citep{Wang2018GLUEAM}, SuperGLUE \citep{Wang2019SuperGLUEAS}, SQUAD 2.0 \citep{lee-etal-2020-squad2} demonstrates the capabilities of \textit{transformers} on a wide range of tasks. In the last couple of years, there has been an enormous number of variant architectures and approaches; these won't all be reviewed. To give a couple of examples, there are variants such as {S}pan{BERT} \citep{joshi-etal-2020-spanbert} that infers spans rather than single tokens and thus performs strongly in coreference and relation extraction tasks.  \textsc{ELECTRA} \citep{clark2020electra} that improves performance vis-a-vis BERT via reframing training as a discriminatory rather than generational infilling task. Some recent models have taken these ideas to extremes: Switch transformers  \citep{DBLP:journals/corr/abs-2101-03961} that use switching between transformers network as a mixture-of-experts to scale models to trillions of parameters; Hypergrid \citep{DBLP:journals/corr/abs-2007-05891} that scales multitask learning via a grid of adaptive input and output transformers to a greater variety of tasks; visual transformers \citep{DBLP:journals/corr/abs-2006-03677} that adapt the transformer model to work with image data and allow it to be applied to multimodal tasks; UniT \citep{DBLP:journals/corr/abs-2102-10772} that combined some HyperGrid and multimodal transformer features for flexible multimodal multitask learning. There are also more recent architectures that have employed various mechanisms to reduce the computational complexity of the \textit{transformer} to work on longer context sequences, see \citet{DBLP:journals/corr/abs-2009-06732} and \citet{DBLP:journals/corr/abs-2103-14636} for reviews of the approaches. This is most relevant to chapter \ref{chap:salience} which uses a memory mechanism for longer texts and is discussed further then. This thesis sticks with variants of GPT and BART by fine-tuning on domain-specific and task-specific objectives as these are the most relevant and suitable models to adapt, given the computational resources available for the given task.

\section{Power of Vector Representations and Composition}

As discussed in the last chapter, comprehending stories and generation requires strong semantic and causal knowledge of how both the world works and how narrative worlds work. Thus it represents a tough challenge for any computational model. For a \textit{transformer} to be expected to model salience or suspense or generate coherent plans for stories, they must be capable of learning to model complex causal knowledge in the same way as the reviewed symbolic and case-based rule systems attempted to in more limited domains. This requires that \textit{transformers} have the computational power to do this and that the latent semantic space can encapsulate this knowledge. There are also clear limitations to the models used in this thesis and their ability to model such knowledge and these limitations will be discussed. 

On the theoretical power of models, there was much earlier work \citep{SIEGELMANN199177,SIEGELMANN1994331,siegelmann1995computational,kremer1995computational} on simpler RNNs that demonstrated they could be Turing complete. However, this required the assumptions that the whole input would be read in before any inference and that there be \textit{unbounded time}, and \textit{infinite precision}. Clearly, the second and third are not practical, and the first is not desirable. Later empirical work \citep{weiss-etal-2018-practical} on the RNN successor LSTM with peepholes \citep{10.1162/neco.1997.9.8.1735,gers2000recurrent} and \citet{weiss-etal-2018-practical} demonstrated that they were capable of unbounded counting with limited precision; this is linguistically equivalent to recognising context-free or context-sensitive language. In related work  \citet{DBLP:journals/corr/abs-1906-06349} prove that a sister architecture of the LSTM, the GRU \citep{cho-etal-2014-learning}, is with a single layer reading one token at a time is at least as powerful as pushdown automata with arbitrary precision or deterministic finite automata with fixed precision. These capabilities increase with the stacking of layers; this demonstrates these recurrent models are capable of powerful hierarchical composition and structure. While \textit{transformers} are the main basis for the models used for this thesis, LSTMs are used in the upper layers of hierarchical models and as part of the VAE (Variational Autoencoder) model employed later, and so the capabilities are relevant.

Similar theoretical insights have been applied to \textit{transformer} models. \citet{perez2021attention} prove that a \textit{transformer} can be Turing complete but this requires \textit{hard attention}, and additionally arbitrary precision. \citet{bhattamishra-etal-2020-computational} provides an alternative proof, and an analysis that shows the residual connection in the default self-attention block is also necessary.  The \textit{hard attention} is required for the proofs because, otherwise, softmax can produce irrational numbers. These proofs also require positional encoding as without this \textit{transformers} are unable to reason about ordering. However, arbitrary precision is not practical as typically models need fixed precision to train and infer with high-performance GPU or equivalent accelerated hardware. Universal Transformers \citep{DBLP:conf/iclr/DehghaniGVUK19} are an alternative architecture that can be Turing complete with fixed precision but requires an extra recurrent transition function, which isn't used by the most popular pre-trained models or those in this thesis.
\citet{DBLP:conf/iclr/YunBRRK20} proves that the \textit{transformer} is a universal sequence to sequence approximator; this is important for practical downstream application domains. In related theoretical work \citet{DBLP:conf/icml/HronBSN20} demonstrates that \textit{transformers} approximate \textit{Gaussian processes} as the number of heads increase towards infinity.  On the cautionary side, \citet{hahn-2020-theoretical} finds theoretical limitations in modelling periodic finite-state languages, or hierarchical structure, unless the number of layers or heads increases with input length. Hahn suggests that in practice, with these limits, as shown by the success of these models, these tasks may be simpler than formal theoretical models require. There are human cognitive limits on recursion, which limits its practical use and beyond which humans struggle to comprehend, and so the theoretical limits of \textit{transformers} may align well with more limited human cognition as per \citet{parker2017cue} cue based memory model. This is also closer to previously mentioned cognitive narratology ideas where the limitations of memory and processing are more important than in the formal structural traditions.
In practice, then the main pre-trained \textit{transformers} used in this thesis with fixed precision are not Turing complete. Nevertheless, theoretically they are powerful models in being able to model complex states and structures. This power increases with the depth and heads used in the models. The power of the stacked model is illustrated by \citet{tenney-etal-2019-bert} whose analysis of BERT using probes is that different layers of the architectures recreate the \textit{classical NLP pipeline}; POS (Part-of-Speech) tagging, parsing, NER (Named Entity Recognition), semantic roles, and coreferences are roughly grouped into layers. The \textit{transformer} acts as a pipeline so that hierarchical, more complex representations build from simpler abstractions at each layer. Other similar analysis on attention \citep{clark-etal-2019-bert,vig-belinkov-2019-analyzing} supports this and also the notion that attention patterns vary across layers according to the role, that attention heads are often specialised to particular lexical or semantic properties, and that lower-down attention and focus is broader, and this narrows further up the hierarchy, again suggesting a pipeline. The many existing analyses of BERT and related \textit{transformers} has been called BERTology \citep{rogers-etal-2020-primer}; Rogers et al. reviews literature of many of the representational qualities of BERT and the issues around how the architecture, task, training, and deployment can affect models. The theoretical power combined with the mentioned state-of-the-art results on many practical tasks justifies that they can learn the relations needed for the theoretical models discussed in Chapter \ref{chap:backgroundtheory}. While it also preserves the potential advantages over symbolic systems in better modelling looser gradations of meaning and being learnable from data.

Compositional has been used in the functional sense of combining representations using the attention mechanism of the transformer or the recursive state mechanisms of LSTMs. For the same reason, these LMs are also often called contextual because each word embedding isn't standalone but depends on the context word around it via the compositional mechanisms of the model. Hence, every one of the same wordpiece representations will be different, which is unlike \textit{word2vec} where each representation of the same word such as \textit{float} will be the same. In a contextual model \textit{float} could be very different depending on whether it was a \textit{duck} or \textit{ship} floating on the water, the monetary \textit{float} of the pound against the dollar, \textit{float} a business idea, or a market \textit{float} that sells its wares, or a tennis player who \textit{floats} a forehand down the line. The power of the models comes from learning to adapt the representations from the neighbouring text, and rather than more weakly covering all senses.

Earlier \textit{word2vec} demonstrated the capability to represent syntactic, lexical, and semantic information in a latent vector representations. The \textit{transformer} models are in theory, and practice, far more powerful, and there has been much more recent analysis of these models. This is important for the thesis because story comprehension requires a much higher-level understanding of events, characters and causality. The challenge is that more complicated relational information, such as when two characters mentioned are the same or the intent behind the action, needs to be represented in the vector space via the composition capabilities outlined. Although the dimensionality of the vector can increase, the information compression required will be a lot higher to encode it within a dense space. The methods later directly use vector distance measurements to model salience or suspense, so it is important that relevant semantic concepts can be represented well in vector models. For \textit{transformer} models, there are several different ways of extracting sentence vectors. Most common with BERT or GPT style models is to take the delimiter tokens, which for BERT are explicitly seen as sentence representations or average across a layer (usually the last or second last) of the embeddings. This allows \textit{transformer} models to represent a whole sentence or a custom span in a single vector, which should, like word embeddings, encapsulate meaning in vector space. Some alternative non-transformers models have tried to learn sentence embeddings directly, such as Skip-Thoughts \citep{NIPS2015_5950} that is modelled on the negative sampling method of \textit{word2vec} scaled to the sentence level. InferSent \citep{conneau-etal-2017-supervised} and  USE (Universal Sentence Encoder; \citealt{cer-etal-2018-universal}) both primarily use a supervised training method on entailment datasets. Other architectures use similar training objectives with \textit{transformers} such as Sentence-BERT \citep{reimers-gurevych-2019-sentence}. S-BERT \citep{DBLP:conf/aaai/0001WZLZZZ20} that takes a different approach to improve sentence representations by extracting and augmenting BERT with SRL (Semantic Role Labelling), and additional fusion layers to enrich representations. None of these models are directly used, but the later work in the thesis borrows from these papers to improve sentence representations as part of a hierarchical model.

\citet{conneau-etal-2018-cram} develops a suite of probes for testing the linguistic properties of sentence embeddings, including testing whether models are sensitive to legal word orderings, can infer passing tree depth, tenses, number of subjects, or numbers of objects, for example. SentEval \citep{conneau-kiela-2018-senteval} is a suite of broader semantic tests for sentence embeddings that test entailment, paraphrasing, semantic similarity, sentiment and opinions. The results from the strongest models are currently around $0.9$ out of $1.0$ on these benchmarks. Both sets of tests show that sentence representations from both LMs and dedicated sentence embedding models perform well, indicating these models implicitly learn both significant lexical and semantic meaning as part of training. \citet{zhu-etal-2018-exploring} and \citet{Perone2018EvaluationOS} find similar strong results for Infersent and USE on higher-level semantic tasks, and   \citet{krasnowska-kieras-wroblewska-2019-empirical} on linguistic features.   \citet{tenney2018what} concludes that on semantic tasks, the gains are modest. However, on tasks such as POS tagging, dependency parsing labelling, NER, SRL, coreference resolution, relation classification, or semantic proto-roles, the performance increases are considerable with \textit{transformer} model. All of these tasks require more complicated compositional reasoning demonstrating the increased capabilities of \textit{transformers} in high-level modelling processing. Similarly,  \citet{peters-etal-2018-dissecting} also finds that there are compositional or contextual benefits to syntactic and semantic tasks. \citet{petroni-etal-2019-language} analyses a variety of recent large \textit{transformer} LMs encoded to act as knowledgebase. That is, to implicitly encode factual and commonsense relations. Factual relations, for example, that \textit{Lisbon} is the  \textit{capital} of \textit{Portugal}, and \textit{Charles Dickens} was an \textit{author}. Commonsense relations include conceptual knowledge such as \textit{Lions} and \textit{Tigers} are both type of \textit{cats}, and \textit{cats} are \textit{carnivores}. It can include knowledge about everyday objects such as \textit{coffee} is \textit{poured} into a \textit{mug}, and a \textit{mug} is \textit{put on} a \textit{table}, and that other combinations don't work. Commonsense knowledge includes commonsense sequences of events such as person x's car is broken so they want to repair it. In order to repair it they require a mechanic. The mechanic requires the equipment for the specific repair, and want to be paid, etc. It also includes social interactions such as with the example movie, \textit{Gone Girl}, \textit{Amy} is \textit{angry} her husband is having an affair with his student, and she decides to plot to take revenge. As outlined in the last chapter, all of these are crucial for story comprehension. \citet{heinzerling-inui-2021-language} paper also supports the idea that LMs can be effective knowledgebases. Although this work is concerned with factual knowledge, the argument is that \textit{transformer} LMs can store and query complex relational graph structures. As seen with the discussion on symbolic reasoning, this is also the classic way of modelling commonsense computationally and how it is modelled in the formalist narrative theory. There is a technical limit: knowledge needs to be encoded directly onto the models' parameters, which means more knowledge requires more parameters. Chapter \ref{chap:salience} adapts recent theories on using external knowledgebases with \textit{transformers} to bypass this limitation for longer narrative works.

\section{Limitations of Transformers and Representations}

The reviewed literature so far has showcased both the power of \textit{transformers} theoretically, in empirical benchmarks, and their potential to represent complex semantic concepts and relations. The thesis uses these models throughout the experimental work. While there are strong reasons why these models are suitable, some weaknesses place an upper bound on performance. \citet{ribeiro-etal-2020-beyond} finds that pre-trained \textit{transformers} often have gaps in their representation for more complicated forms of antonym negation, coreferences, or temporal ordering. These failures are not always general, as overall performance can still be good. Still, there's are often unexpected failings, particularly where examples of the form are not present in the training data. There are problems with the high-performance benchmarks, such as SuperGLUE or SQuAD. Because LMs infer general rules from the text, then they can learn behaviour from spurious correlations. \citet{geirhos2020shortcut} refers to this as shortcut learning where the model can perform well on synthetic tasks based on surface-level features but without deeper understanding. It is a common problem in most deep learning architectures and extends to non NLP tasks such as image recognition. \citet{niven-kao-2019-probing} demonstrates with an adversarial approach that it's straightforward to create counterexamples that can fool \textit{transformer} LMs by manipulating logical reasoning arguments. This lack of robustness is often a problem when models infer on out of domain data. An example of this would be a multimodal dataset that contains images and text describing surfers and sailing yachts every time a beach is mentioned; models can then hallucinate surfers whenever a beach or sea occurs, regardless of whether it's applicable. Geirhos et al. argue this is because the deep-learning models will always take the shortest path to the solution, the architecture, training data, loss function, and optimisation all play a huge part in how susceptible models are to spurious correlations. On the COPA (Choice of Plausible Alternatives; \citep{roemmele_choice_2011,gordon-etal-2012-semeval}) common sense reasoning task, \citet{kavumba-etal-2019-choosing} finds that \textsc{ROBERTa} is far less sensitive to these surface-level spurious correlations than BERT. This suggests it's not the architecture that causes this and it's more down to the training data, the training objective, and other training optimisation steps than the underlying architecture. This is relevant to later chapters that justify the training setup and datasets for the models.
As mentioned earlier, this is one of the reasons for multitask learning and training on diverse datasets. Also important are losses that don't stop learning when the answer is found, as this is more likely to lead the model to rely only on a few cues that may be erroneous.
Geirhos et al. also counter that this is not a uniquely deep learning problem as human and biological cognition also takes a shortcut of cues. It is something that occurs in the story examples used in the thesis. For example, \textit{Gone Girl} relies on the viewer picking up surface-level impressions that \textit{Ben} must have killed \textit{Amy} because of their marital issues, and so we ignore cues that hasn't happened. 

The problems with these spurious correlations can also be in the construction of a dataset or task. One relevant example is the ROC cloze story task \citep{mostafazadeh-etal-2016-corpus} where a model needs to be able to predict the correct ending for an incorrect one following a four-sentence prompt.  Later work by  \citet{cai-etal-2017-pay} and \citet{schwartz-etal-2017-effect} found it was possible to do reasonably well on the task without even considering the story prompt. The reason for this is the way the dataset was constructed is that the crowd workers who wrote the incorrect endings were different from those who wrote the story prompt and correct endings. Stylistically the incorrect endings were just linguistically different from the correct ones, so a model was able to infer the correct endings with just surface-level features and no requirement to comprehend the story at all. This is a widespread problem with synthetic datasets and tasks. Datasets such as QuAIL \citep{Rogers_Kovaleva_Downey_Rumshisky_2020} in Question Answering have tried to address these issues, but these steps are often quite complicated, and it's easy to introduce artefacts inadvertently. In QuAIL the performance of the best models is $30\%$ lower than similar datasets without these steps, emphasising the effects of these artefacts is substantial. This is one motivation for this thesis to avoid artificial tasks and training on diverse and large corpora. It is also a benefit of an unsupervised approach combined with inference rules for predicting comprehension labels. Taking a supervised approach would necessitate annotating a gold standard corpus. The costs involved in this necessitate using a smaller dataset. The small dataset and human gold labelling increase the risk of artificially introducing artefacts that would exaggerate performance as the model would learn them in training and then reproduce them on the test set. A silver labelling or distant supervision (for example \citet{min-etal-2013-distant}) approach where a model trained on small dataset labels a larger one automatically can increase the diversity of the data but risks that these surface features will be the basis of labelling on the larger dataset, introducing noise, and hence the risk of exaggerating performance.

Spurious correlations are not the main challenge for using pre-trained  LMs or hierarchical models built on top of them. This is the long-known symbolic grounding problem \citep{harnad1990symbol}. The problem is that the language model or other ML models learn from purely self-referential text with no meaning in the real world outside the language. This is a massive problem for language in general and in particular story comprehension and generation tasks. In some tasks such as machine translation, it is more practical to learn close synonyms in both domains purely from bilingual texts. NER supervised datasets can learn the patterns and cues associated with people's names or locations. In contrast, stories are open domain and describe a wide variety of circumstances that occur in the real world and the physical world, and the motivations and thoughts of people within the world. Another factor is that in stories, typically, the reader is asked to fill in many gaps and assumptions in what is happening, and communication focuses on necessary details \citep{abbott2008cambridge}. While grounding is usually thought of as many computational problems, mainly in storytelling, there are strong cultural elements. For example, though with \textit{Rocky} and \textit{Pretty Woman} the overall plot may be stereotypical, there are lots of elements that would be hard for someone unfamiliar with the US to understand. The same would be true of \textit{Great Expectations} for someone unfamiliar with Victorian England, or for the reverse cultural shift, a Chinese classic such as \textit{Dream of the Red Chamber} or an old Bollywood classic movie such as \textit{Mother India}.

These problems are observed in real large \textit{transformer} LMs.  \citet{Forbes2019DoNL} test LMs understanding real world interactions, for example \textit{a blender needs to be plugged in to function}, or \textit{coffee can be poured into a cup}, or \textit{boots can be laced up}. They find that LMs can learn the real world physical interactions well, but only if the interactions are explicitly in the training data. For physical interactions not in the dataset the LMs perform much worse. The model does not generalise to unseen interactions and objects. For example, would a \textit{spatula require electricity}? A model could easily infer that it did if it identified it as a kitchen device and had information that a blender, toaster, and oven did. The problem is that physical interactions are often left unstated in text, and so LM's knowledge of the physical world, can thus have spurious representations. \citet{zhou2020evaluating} finds that LMs can model commonsense reasoning but perform poorly when multiple steps are required; these chains of reasoning, as reviewed, are common in storytelling. In considering LMs as knowledgebases \citet{DBLP:journals/corr/abs-1911-03681} found that LMs are able to learn general rules well, such as people with \textit{French} names were typically \textit{French}, or \textit{Italian} names \textit{Italians}. This is clearly generally a desirable property but had a problem when concrete knowledge is required; the \textit{American} actor  \textit{Leanordo DiCaprio} could easily be assumed to be \textit{Italian}. \citet{DBLP:journals/corr/abs-1907-13528}'s psycholinguistic analysis suggests that LMs, while capable, have difficulty with more complicated causal event roles and inference. A given an example, someone \textit{who is kissed has red on their face because of X}; the correct answer is \textit{lipstick}, but the model may infer \textit{mascara}.  This is more of a grounding problem than in the capabilities of the model as there are not examples explaining the relationship between these in the dataset, such as \textit{lipstick} goes on the \textit{lip}, \textit{lipstick} is often \textit{red}, a \textit{kiss} uses the \textit{lips}, \textit{lipstick} can transfer from one surface to another. These things are rarely contained in written corpora, though, so the model learns a more general sense that makeup is associated with faces and beauty, but without the causal link.

Three problems have been identified for using \textit{transformers} related to the \textit{grounding}: a) Learning from text-only misses much of the real-world physical and social interactions that ground language. b) LMs will not learn relations well without concrete examples, especially reasoning chains about common sense and abstract knowledge. c) Models are limited in the amount of knowledge they can represent by the number of parameters. 

\section{Overcoming Limitations}

\begin{figure}[htbp]
\centering
\includegraphics[width=\textwidth]{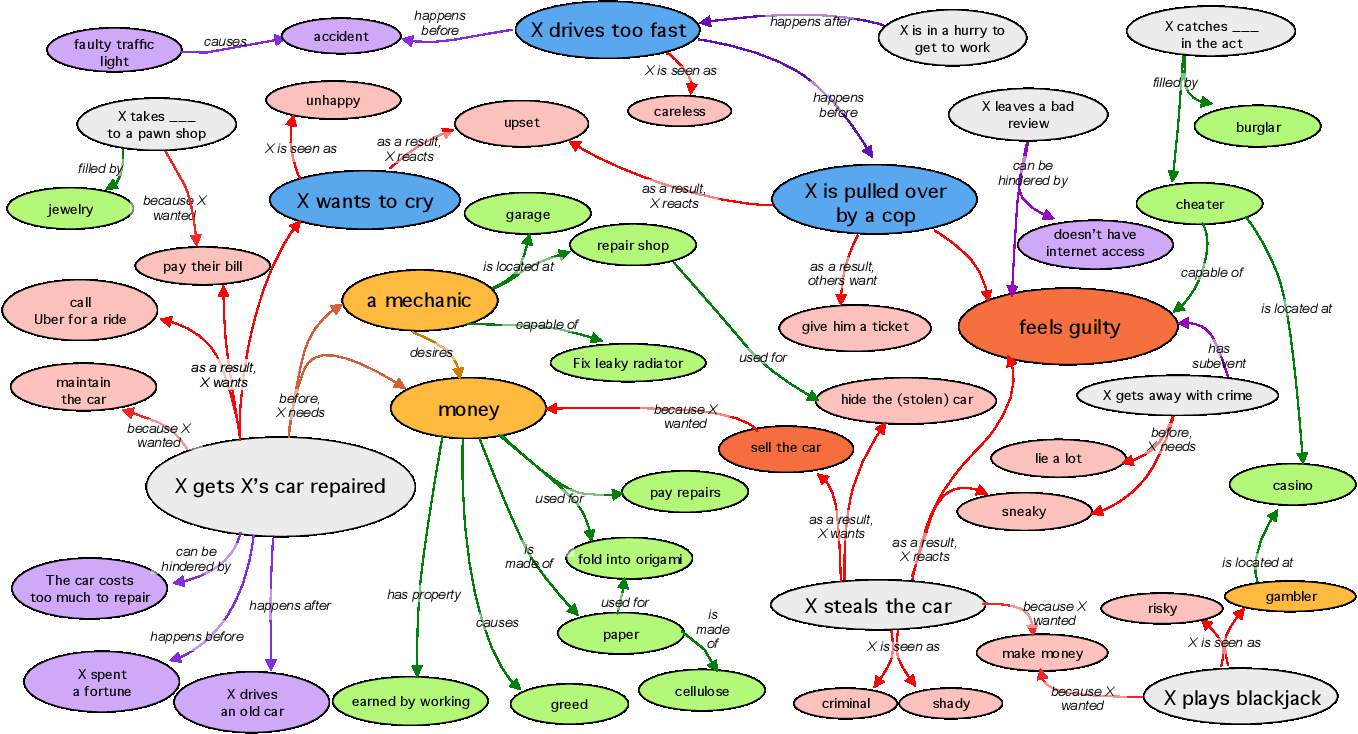}
\caption{Part of a car knowledgegraph from ATOMIC2020 \citep{Hwang2021COMETATOMIC2O}}
 \label{fig:atomic}
\end{figure}

 I shall now introduce a few approaches to these problems and analyse the relevance to the aims of this thesis, which are comprehension and generation on open domain stories. One common recent approach to address these first two problems is to encode commonsense knowledge in structured datasets that can be used directly for inference or, more recently, to train models or as part of hybrid models that use external knowledgebases. ConceptNet \citep{conceptnet1,conceptnet2} is a long-running knowledge graph that stores words and thesaurus-like related words or synonyms but also the spatial, physical, social, temporal affordances of those objects. For example, \textit{poodle} is a type of \textit{Dog}, \textit{Dogs bark}, are used for \textit{fetching sticks} or \textit{guarding a property}, etc. For NLI (Natural Language Inference), there are the  SNLI \citep{bowman-etal-2015-large} and
MultiNLI datasets \citep{N18-1101} that contains relations between statements and whether or not they entail or contradict each other, and so represent simplified reasoning chains. More recently, there have been datasets that try to incorporate more complex real-world causal chains. SWAG \citep{zellers-etal-2018-swag} is a multi-modal dataset that teaches models to infer the consequences to events such as a \textit{kayak flips upside down}, or \textit{sugar is added to the bowl to make butter cream}. A more recent {PIGL}e{T} datase \citep{zellers-etal-2021-piglet} has extended this idea by using a simulation environment AI2-Thor \citep{ai2thor,manipulathor} (based on the \textit{Unreal} game engine) to learn the properties of real-world physical interaction. For example a \textit{vase} has the property \textit{isBroken} that becomes \textit{true} after it is thrown onto a table. These datasets describe everyday events or physical interactions. Other datasets tie together social interactions, events, and physical entities into causal chains. \textsc{ATOMIC} \citep{DBLP:conf/aaai/SapBABLRRSC19} and \textsc{ATOMIC2020} \citep{Hwang2021COMETATOMIC2O} are two versions of the same dataset, see figure \ref{fig:atomic} for an extract from a car causal chain where red represents social, purple events, and green physical world interactions. GLUCOSE \citep{mostafazadeh-etal-2020-glucose} is a dataset of story-specific casual events. For factual knowledge, there are knowledgebases from extracted Wikipedia dumps such as the TF-IDF key-value store used for question answering Wizard of Wikipedia \citep{DBLP:conf/iclr/DinanRSFAW19}, or a recent conversion of Wikipedia into a structured graph
WikiGraphs \citep{wang-etal-2021-wikigraphs}. There have also been attempts to integrate many of these into a single knowledgebase, such as  \citet{ilievski2021cskg}.

All of the previously mentioned knowledge bases are different pieces of the knowledge puzzle.  Most of the cited papers have baselines where they use these datasets to train an LM. The most common and most straightforward in these works is that used by \textsc{COMET} \citep{bosselut-etal-2019-comet} on the \textsc{ATOMIC} datasets to feed text into a \textit{transformers} model with special text tokens to encode structured graph relations. The intuition is that the \textit{transformer} will learn to use these to structure relational representations. This further supports the idea that these models can learn this reasoning but need appropriate training examples and structure to do so. There are plenty of other too numerous variations that directly incorporate the graph structure via Graph neural network methods or other inference methods.  See  \citet{9416312} for a review of methods in structured knowledge graphs or \citet{DBLP:journals/corr/abs-2010-04389} for methods of combining knowledge with the domain-specific story generation. 

This thesis analyses in the TD-VAE chapters experiments with using NLI datasets and ATOMIC to improve the sentence representations within hierarchical models. Later analysis of results in this thesis suggests this was not entirely successful. However, the main reason for including this brief review is the general approach of this thesis does not adopt these methods. A major weakness when reviewing the symbolic approaches was that it is extremely difficult to construct these rules for anything but a tiny domain. This thesis is concerned with open-ended storytelling by definition, which can cover any subject or period of events. Crowdsourcing has made it feasible to collect more extensive datasets, although at the expense of possible noise and extra quality control steps. The problem in building a commonsense knowledgebase that describes all events and interactions that could occur is an unbounded problem.  Collecting datasets is an expensive and time-consuming endeavour that will only cover a tiny fraction of the domain. For example, a \textit{whittles} was mentioned earlier for the \textit{Great Expectations}, \textit{whittles} is an archaic name for a knife for carving (or whittling) wood. It is highly unlikely that a common-sense reasoning dataset would contain a rule that \textit{prisoner leg irons} has an \textit{affordance} that it can be \textit{cut open} by \textit{whittles}. This is an archaic example, but there are more or less infinite similar examples that could be applied to open domain stories. For example, \textit{Rocky slips on a tattered robe}, physically is the robe something someone is likely to slip on? Or in \textit{Gone Girl} where the plot explicitly relies on the causal expectation (though incorrect) that a woman's disappearance is most likely because her husband killed her, which is derived from social expectations driven by real and fictional crime . In foreign cultures such as Chinese classic \textit{Dream of the Red Chamber}, there are different social norms, such as a wealthy man is expected to have many wives and concubines. This could conflict with a western common sense graph about marriages. This is without going into fantasy or sci-fi that may have Vampires, Fairies with magic potions, warp drives, or time travel. Generally, the approach shifts the problem from symbolic systems that require complex rules to infer directly to creating training examples that cover the domain for training ML (Machine Learning) models.

 Some of the ideas are used from this line of work, as mentioned with the ATOMIC and NLI. Also, Chapter \ref{chap:salience} on salience that learns to use Wikipedia \footnote{Factual knowledge is easier to capture comprehensively because of large and well-maintained sites such as Wikipedia.} and an extracted plots knowledgebase to assist with factual and plot knowledge with a neural case-based reasoning model. The approach is deliberately unstructured in allowing the model to learn the most salient features in the context. As has been reviewed in the last chapter it is often seemingly innocuous cues and mood that drive stories and these can be lost in pre-structuring and pre-processing. The general approach is to train on diverse and large corpora of real stories and tune the architecture and training objective to extract the most from the data. While noting in the analysis, there may be problems because of the grounding issue. Though some of the datasets may be beneficial to train on within their domain, that is quite limited to specific knowledge chosen by the creators of the datasets. The last chapter introduced many cognitive ideas that have the concept of the reader model, storytelling as an unfolding problem that the reader uncovers. All the models incorporate a sequential reading pattern in training and inference to best replicate that process rather than on a complex engineered pipeline with many preprocessing steps; to favour implicit learning over complex engineering. However, graph neural network methods on more structure representation are interesting for further work and may provide performance benefits. Though even here there are theoretical insights that \textit{attention} can be understood as a special case of \textit{convolutional graph methods} \citep{DBLP:journals/corr/abs-1905-01289,joshi2020transformers}. The difference being that \textit{attention} is implicitly learning a graph structure from a fully connected graph, whereas graph methods are working off a predetermined representation. So the choice may be a  cost-benefit analysis as to whether the structure focuses on salient information and helps model performance above and beyond the complexity cost of implementing it.

 There are several other approaches to grounding language beyond structured knowledgebases and training data. One other obvious problem with text only, training apart, is visual knowledge; the model doesn't know what anything looks like beyond adjectives commonly applied in descriptions. There have been improvements in recent multimodal models: CLIP \citep{DBLP:journals/corr/abs-2103-00020} that semantically link a huge corpus of aligned text and images. DALL-E \citep{pmlr-v139-ramesh21a} that can generate images from text input, or vice-versa text descriptions from input images. UNiT \citep{DBLP:journals/corr/abs-2102-10772} that models a variety of NLP only, vision only and multimodal tasks with a shared encoder, input encoder for the different modalities and separate decoder for each task. This leads to improved performance in tasks across all the modalities. Similarly, multitask learning can improve results for all tasks, multimodalities can improve performance even when one modality is no longer involved. The vision can improve performance slightly on NLP tasks; although the modalities are different, there are likely shared elements in the representations that may help ground the language. It could be imagined if a model was co-trained on multimodal data. When the boxing club in \textit{Rocky} is mentioned, the representation could capture the grotty atmosphere of the club through similar photos of boxing but also bars, casinos and other seedier nightlife locations. This would also implicitly ground the images back to the co-occurring language and events that occur in these places, which may help even with text comprehension. Lots of these newer multimodal multitask \textit{transformers} are relatively recent and post much of the work done on this thesis, and so it would be left for future work. They also obviously have great potential for multimodal forms of narrative such as movies or comics. For example, the case-based reasoning model in the salience chapter could easily be adapted  to retrieve images or screenshots from movies.\footnote{Like UNiT this could be done by having separate encoders for text and images while sharing the retrieval encoders and decoder for inferring or generating output.} The model could attend to the encoded representation of both the dialogue and the images of the movie in either comprehension or generation tasks.
 
There are other grounding strategies within storytelling. Another limitation of relying on a text-only dataset is there is a problem understanding the physical world. A simple example is that the same person cannot be in two places at once so that someone in a story is not likely to be in \textit{London} or \textit{Sydney} in the same. It applies to objects as well. For example, someone can only take a sip from their coffee if they had a coffee, to begin with. LM's can hallucinate things from nowhere because they don't have a real-world model of where objects are. Within story narratives, there has been a line of work that has tackled this. \textsc{LIGHT} is a popular online text role-playing game, based on popular Dungeons and Dragon game, modelled by \citet{Martin2018DungeonsAD}.  A Facebook team constructed a dataset \citep{urbanek-etal-2019-learning} with several follow-up papers including \citet{DBLP:conf/aaai/FanURDQKPKRSW20}, \citet{ammanabrolu-etal-2021-motivate} and \citet{DBLP:journals/corr/abs-2008-08076}.  Both of these are interactive text games in which characters have personas such as a \textit{King} or an \textit{Elf}. These characters carry objects such as a \textit{bucket} or a \textit{sword}, and some locations also contains objects such as a \textit{throne} or a \textit{well}. The intuition behind this is to ground understanding and generation in the story's real-world locations and possibilities. So that the \textit{Elf} can draw water from the well using the \textit{bucket} only if they are in a location that contains the \textit{well} and are in possession of the \textit{bucket}. Interestingly the main framework from these papers has been just to concatenate these context elements with special tokens, as discussed earlier, to assist \textit{transformers} and related architectures and implicitly learn relations between these aspects. The relatively decent performance of these models again suggests that inferring from these characters intentions and objects usage is within the model's capabilities if these can detail can be provided in training data. Once again, the issue with this is that this is of a limited domain and cannot be applied to open-ended storytelling.  \citet{DBLP:journals/corr/abs-2008-08076} extended the \textsc{LIGHT} system so it could be open in the sense that bots could play along in interactive stories, and hence extend the domain. Because the system works only on text, this could be extended to any domain to model. Unfortunately, the huge resources required to collect a dataset with similar structures in an open domain are beyond this thesis. 

\section{Related Work}

There are other lines of work on character development in narratives, comprehension, and \textit{neuro-symbolic} methods that are relevant but not directly related to any specific chapter. This chapter concludes by discussing these methods as they are largely orthogonal to the experimental work in this thesis, and there is much scope for follow-up and hybrid work.

The approach taken in this thesis is mainly to consider a story as a whole as a reader model. Simply put, the model reads a story one sentence or block at a time and infers what will happen next. Comprehension of different characters and their motivations to improve those predictions are implicit within the models. A long line of separate work focuses on analysing characters' motivations and development, another critical component of storytelling to this thesis's plot-focused ideas. Plot and character-based methods are complementary approaches where there is much scope for future work.  \citet{bamman-etal-2013-learning} and \citet{bamman-etal-2014-bayesian} develop a model to extract the personas of characters from movie scripts and books, respectively. There are more recent neural attention-based models \citep{chu-etal-2018-learning}, or BERT with external memory \citep{vijayaraghavan-etal-2020-dapper}. These papers use more traditional word count hierarchical topic models combined with a named entity and coreference to detect a cluster of words or phrases associated with characters. These learnt personas identify stereotypical character types, such as \textit{Rocky} the rough but ultimately good hero, or \textit{Amy's}  the scheming psychopath from \textit{Gone Girl}. Characters can then be defined as a mixture of these personas, which helps analyse individual works of fiction and typical character types across fiction. These are static representations, but of course, the whole point of storytelling, as exemplified in the \textit{eventfulness} concept, is that characters and their circumstances change and is about the dynamic relationship between characters. \citet{DBLP:conf/AAAI/ChaturvediID17}, with similar methods to Banman et al., modelled the evolving relationships between characters with text where characters co-occur. \citet{iyyer-etal-2016-feuding} developed a neural model that tracks how with relationship trajectories change over time; like later work in this thesis, these changes are modelled with cosine and other vector distances. Outside of storytelling \citet{han-etal-2019-permanent}, with a similar approach, modelled the changing relations between countries with news data. \citet{frermann-szarvas-2017-inducing} developed a multi-cluster approach with a multi-view that creates clusters for individual characters and also relations when they co-occur; the loss induces the views to be as different as possible, so they capture the unique qualities of the character and the relations. Gaussian processes \citep{DBLP:journals/corr/abs-1804-04164,HannahKim-BoyangLi-Actors-2019}  have also been used to model character personas.  Characters also play an important role in detecting the significance of events; \citet{gorinski-lapata-2015-movie} and \citet{moviegraphs}. Both rely on the intuition that the main characters typically are together in important scenes or that sub-graphs of characters that often occur together can mark sub-plot strands. As far as neural models go, there are several LMs that have incorporated named entities as a specific construct into training, including EntityNLM \citep{ji-etal-2017-dynamic}, EntNet \citep{DBLP:conf/iclr/HenaffWSBL17}, and Reference-Aware LM \citep{yang-etal-2017-reference}, though not specifically for story comprehension. There is also \citet{puduppully-etal-2019-data} who specifically model entities for data-to-text generation. \citet{Lee2020StoryEL} in more recent work proposes \textit{Char2Vec} and \textit{Story2Vec} that extends graph embedding methods to model vector representations for individual characters and the story as a whole; the models are, in principle, similar to \textit{word2vec} in inferring missing edges representing relationships in the graph. Another novel recent work is by \citet{sims-bamman-2020-measuring}. This is inspired by earlier work in building communication networks between literary characters by \citet{10.5555/1858681.1858696}. Sims et al. infer how information is communicated between characters through a corpus of 5000 classic stories. Rather than just how often characters appear together, they study the propagation of specific tuples to track A told B who told C, for example A telling B \textit{Miss Haversham} died and B tells C \textit{She Died} so the graph can infer C knows this piece of information. So enabling the model to produce a graph with the strength of propagation between characters can be analysed with standard graph techniques such as centrality. It is evaluated on LitBank  \citep{sims-etal-2019-literary,bamman-etal-2019-annotated} dataset.\footnote{LitBank is a dataset that contains 100 classic and contemporary fiction works with metadata, and processed events, NER, coreferences, POS and quotations. Unfortunately, these later annotations are only for the first 2000 words of the work. Many of these works are far longer.}

As per simulation-based methods, there is an intuitive appeal to these methods. Story planning can be naturally thought of as an ordered series of interactions between characters. This suits a graph structure and can be thought of as linking the five indices from \citet{1999-04086-005} and \citet{Zwaan1995TheCO} indexing mode into a graph. A cognitive debate is beyond the scope of this thesis which is about a more natural representation of story comprehension: The episodic lookup indexing model for Zwaan, which aligns closely with the salience work later in the thesis, or a graph structure is more representative. From narrative theory there's a difference between a plot-centric view as per this thesis and a character-centric one as per this literature. Neither one is correct. Both are orthogonal and complimentary. Another significant difference between the models in this thesis and graph methods is that the models read in order where all the cited papers preprocess the graph structure upfront. As emphasised by the \textit{reader as a problem solver} paradigm, storytelling is a dynamic cognitive process where the reader working out what may happen is essential, which is relevant to surprise, suspense, and salience. What's important in the overall picture may not tell all as there are cues and prompts that may trigger suspense or surprise (false lead) but not be necessary to the overall plot, or small details that seem innocuous but become important later.
An upfront constructed graph is strong from the point of view of analysing a whole plot by graph centrality or other means. Important events and characters can be known but retrospectively. It is very different from the reading experience. One workaround is to use a directed graph, so only previously seen information is seen, which is not a natural fit. Another option is to construct the graph dynamically, but the preprocessing pipelines make this complicated. The cited papers differ but in the typical pipeline to construct the input graphs for these models are as follows: The story need to be segmented into sentences or events; key named entities need to be identified; simplified event information extracted maybe with an SRL model; coreferences are resolved; speech may be extracted; entities participating in the same event are linked together. Each of these is typically a separate model that may need fine-tuning. They are also not perfect, especially away from trained data which inevitably introduces noise and errors. Thus constructing a graph inline is a substantial engineering effort, especially in an open domain and for a large corpus dataset for training. These tasks are often more complicated and require more engineering to perform well than first appearances suggest. \footnote{For example, state of the art coreference systems usually only work over a fixed-sized window of tokens, so a second level is needed to join together blocks from across chunks of text. It is a complex problem to identify when two characters are involved in the same action. There can be long strings of pronouns to disentangle, or another character that can be referred to that is not involved. There are difficult, tough choices about telling if characters are together, e.g. in the same extracted event, in the same sentence, direct speech from one to the other, in a text window, or the same longer block of text.} In addition, recent research by \citet{li-etal-2021-implicit} has demonstrated that LMs such as BART and T5 can model evolving entity states and relations and the dynamics of situations. As per the discussed causal chains discussion, much of what is made explicit in the graph model may be captured explicitly in \textit{transformers}. The general trend with graph structures is to move from statistical counting methods and traditional graph algorithms to neural models that encode representations in the vector space and then measure similarity and perform operations, which aligns with the earlier discussion on methods. Highly structured graph representations are an alternative paradigm. There is much scope for a hybrid work between these approaches discussed further in the future work section of the conclusion. 
 
The basic comprehension methods in this thesis are \textit{neuro-symbolic} methods; that is, combining inference processes with underlying ML models to produce a hybrid architecture.  This thesis covers not only the reasoning for suspense, surprise and salience but also the case-based retrieval system, and the latent state story planning {Neuro-symbolic} methods have been suggested as \textit{the third wave of AI} \citep{DBLP:journals/corr/abs-2012-05876}. It is far too big of an area to go too in-depth, but I will now touch on a few recent examples with domain relevance which are methods in story generation and more complex question answering.

PlotMachines \citep{rashkin-etal-2020-plotmachines} is a text generation system that aims to improve text generation by keeping track of whether elements of a plot have been fulfilled. Trained on \textit{WikiPlots} - plots scraped from Wikipedia - the system maintains a memory of generated text and then attends both to this and the \textit{plot outline}. In this way, the model can learn to follow a plot when generating text by knowing which parts have already been generated. The case-based retrieval model in the salience chapter uses a related approach that can condition on a knowledgebase and memory from previously seen passages. Although the focus is on inferring salience, the same model can generate stories with more dynamic knowledgebase access than a fixed plan. CAST  \citep{DBLP:journals/corr/abs-2105-01311} is a generation system that incorporates common sense knowledge. It uses a language model to generate text by sampling; an ATOMIC \citep{DBLP:conf/aaai/SapBABLRRSC19} trained commonsense reasoning system COMET \citep{bosselut-etal-2019-comet} generates candidate common sense inferences. These can be for multiple characters representing slots in the generated candidate. These are scored against the generated continuations for plausibility in an iterative process that generates the text. \citet{Martin2018EventRF} develops a text generation system that has a two-stage pipeline where one model learns to generate \textit{event2event} sequences that learn from simplified event representations of sentences. A second model then realises each in surface output text. C2PO \citep{Ammanabrolu_Cheung_Broniec_Riedl_2021} builds on the event abstraction idea but combines this with more traditional reasoning. COMET is not used to filter existing text. Instead, it joins together to start and end states with a partial order planning process that iteratively generates pre and post-condition events. These events are joined into a plot graph with distance-based vector representation matching similarity. The most common sense path, where the intuition is most coherent and plausible, can be found through the graph. This event plot path can then be used to generate concrete stories.

Simple factual question answering is of the kind \textit{What is the capital city of France?}, the answer \textit{Paris} is highly likely to co-occur together with \textit{city}, \textit{capital} or other related terms. So these kinds of questions are relatively simple statistical problems for models to learn. In story comprehension or generation understanding, common sense events are essential, as is being able to track the state of multiple entities and draw conclusions of multiple steps. NarrativeQA \citep{kocisky-etal-2018-narrativeqa} asks real-world questions about narratives such as \textit{Where does the trip begin on the east coast?
}, \textit{what is the relationship between Estella and Miss Haversham?}, \textit{Who is Walter reunited with?}. These questions test general comprehension and may be possible with a simple statistical lookup but often requires more sophisticated reasoning such as understanding a trip starts near the beginning of the book or the relationship is adopted daughter though this may not be stated directly but rather need inference over multiple sentences. BookQA \citep{angelidis-etal-2019-book} is similar in answering real comprehension questions such as \textit{Who became Emily's guardian after her father's death?} or \textit{Who owns the castle in which Emily is imprisoned?} Again, in many cases, these may be stated in the text. Still, in other times there can be multiple steps and sources from different points in the story, for example, the fact of \textit{Emily} being held in a \textit{castle} can be in a different part of the story from who owns the \textit{castle}. In both these datasets, the answer may be a simple match or require more reasoning, the DROP dataset  \citep{dua-etal-2019-drop} is specifically designed to have questions that requires reasoning over multiple paragraphs, so there are chains of reference any model has to be able to follow with such questions such as \textit{Where did Charles travel to first, Castile or Barcelona?} or \textit{What were the 3 villages that people were killed in?}.  CosmosQA \citep{huang-etal-2019-cosmos} is a commonsense reasoning dataset with questions such as \textit{what's a possible reason the writer needed someone to dress him every morning?} and \textit{What would have happened to the woman if the staff at the hospital were doing their job properly?}. The dataset asks about causes, effects, how the state of entities changes, and counterfactuals. Not specifically on narratives, but LogiQA \citep{ijcai2020-501} is a question answering dataset on short reasoning. Together these datasets provide far more of a challenge than simple factual answering and are more relevant to storytelling. These datasets are better tests for a model's ability to perform reasoning about entities, causes and effects that are a necessary part of storytelling. In the performance of these leaderboards, one thing to notice is that \textit{transformer} benchmarks, especially encoder-decoder models such as \textit{T5} and \textit{BART} perform well with fine-tuning. UNICORN \citep{Lourie2021UNICORNOR}, a multitask trained model based on T5, performs strongly across eight commonsense reasoning datasets (RAINBOW), including CosmosQA, and is as of writing state-of-the-art. UNIFIEDQA \citep{khashabi-etal-2020-unifiedqa} is a similar multitask setup also based on \textit{T5} that is state of the art for a variety of benchmarks, including BooksQA. It supports the idea that these models can learn these reasoning tasks, model dynamic state changes, ordering and causation with suitable training examples. Masque \citep{nishida-etal-2019-multi} is a hybrid model with a \textit{transformer} that encodes, attends to and reranks input passages to feed the most relevant passages into a pointer decoder that can reference spans of answers in the relevant passages. On NarrativeQA, Masque was state of the art until surpassed by AnswerBART \citep{DBLP:journals/corr/abs-2103-06500} . AnswerBART is conceptually similar to RAG \citep{lewis-etal-2020-bart}, the basis for the salience work on longer texts. AnswerBART is designed more for interactive dialogue and contains a passage reranker for passages in the text rather than RAG, which matches against an open-domain knowledge base.  Both RAG and AnswerBART are conceptually similar to REALM \citep{Guu2020RetrievalAL}, Wizard-of-Wikipedia \citep{DBLP:conf/iclr/DinanRSFAW19}
Hard EM \citep{min-etal-2019-discrete}, SpanSeqGen \citep{min-etal-2020-ambigqa}, and Fusion-in-Decoder \citep{izacard-grave-2021-leveraging} that have achieved state-of-the-art results in factual domains such as answering natural language questions. These models are discussed in more depth later. Still, the main feature is that they all use a learnt retrieval mechanism to retrieve from a knowledgebase or passages in the text and then use neural attention, a reranking mechanism or a combination to integrate knowledge from multiple sources. This integrated knowledge can then be conditioned to generate, select or point to the relevant answers. There are other similar case-based reasoning systems such as CBR-KBQA \citep{DBLP:journals/corr/abs-2104-08762} that use neural representations to query and reason about structure symbolic semantic web queries. Another trend is towards graph-based reasoning methods: QDGAT \citep{chen-etal-2020-question} is state of the art on the DROP dataset. QDGAT builds on Roberta representations and explicitly constructed a directed graph structure that distinguishes entity types such as entities, dates, or numbers. QDGAT then runs a reasoning module on top of this graph structure. This aligns with other similar work by \citet{ding-etal-2019-cognitive}, \citet{tu-etal-2019-multi}, or \citet{DBLP:conf/AAAI/TuHW0HZ20} that use an underlying \textit{transformer} model with rules to construct a graph. Unlike QDGAT, rather than higher-level reasoning modules, these graphs are fed into a GNN (Graph Neural Network) that infers the answering directly and can provide higher-order graph-level vector representations. This work is also related to GraphRetriever \citep{DBLP:journals/corr/abs-1911-03868} or PathRetriever \citep{Asai2020Learning} that use an external knowledgebase based on Wikipedia that's similar to RAG or REALM but preprocess with explicit graph structures rather than the raw text.

The discussion on the latest story generation systems and the more challenging QA datasets brings to attention a few points. First newer neural network architectures can perform well on these tasks including \textit{transformers}, but also \textit{CNN}, \textit{RNN}, and \textit{GNN} architectures. Third, both areas, theoretical and empirical work on vector representations reinforce that these models can capture the dynamic of changing entity states and the cause and effect of storytelling, with noted limitations. Second, both areas can improve performance by integrating neuro-symbolic logic built on top of the vector representations via reasoning, reranking, filtering, or graph-based algorithms. There is, however, a trade-off in the amount of engineering required: Some systems favour neural only representations applying logic on top. With other systems, there is are more complicated hand-engineered pipelines, especially with building complex graph structures. Then there are in-between approaches such as those models that create hand-engineered structures such as graphs out of neural representations and then use this as an input to a higher-level graph module. The bias in this thesis is to favour implicit stacked hierarchical models. In the leaderboards of the various benchmarks in all the areas covered by this chapter, there does not seem a clear winner with seemingly competitive models from the super-large only transformer to complex hand-engineered hierarchical architecture. The onus should be on the more complicated engineered models to justify the benefits given the extra costs in developing, training and integrating the components.  In both areas, there is a trend to integrate common sense and external knowledge either via direct training or neural models that can access and attend to or integrate this external knowledge, aligning with the later work on salience. One note is that some of the newer models from 2020 and 2021 reviewed in this chapter postdate the suspense and surprise work. As such, while they are theoretically and practically interesting, results are evaluated against suitable benchmarks available at the time. Notwithstanding, the trend generally supports the direction taken in this thesis and points the direction to future research in the conclusion.

\section{Thesis Approach}

This chapter has reviewed the base models used throughout the thesis. This section explains the approach for how these models implement the ideas in the last chapter.  The common paradigm for using LMs on downstream tasks has been called Pretraining-Agnostic Identically Distributed (PAID, also called the few-shot learner paradigm ): \textbf{(1)} Train a pre-trained LM, \textbf{(2)} Fine-tune on the downstream task, \textbf{(3)} Evaluating on the in-domain test set. This thesis adopts this basic pattern. The \textit{transformer} models GPT-2 and BART large are fine-tuned with the hierarchical models above them with story domain datasets. The training environments are described in the relevant chapters. However, as a general rule, all experimental work was restricted to be trained on up to 4x 12GB GPUs. As such, the models selected need to be trainable on these GPUs with a reasonable batch time in a reasonable time and with hierarchical layers of memory mechanisms that add additional memory requirements. The billion-plus parameter models such as T5 Large, GPT-3, Switch-U, Transformer-XL are not feasible, and the focus is on tailoring the architecture to the task. There are methods to adapt existing large LMs such as PPLM (Plug and Play Language Model; \citealt{DBLP:conf/iclr/DathathriMLHFMY20} without fine-tuning. This was tried unsuccessfully with the TD-VAE work (in Chapter \ref{chap:tdvaegeneration}), but the general method used throughout the thesis is the more standard fine-tuning for the task approach. \citet{schick-schutze-2021-just} also finds that smaller language models can be competitive with much larger models with optimisation of the training objectives.  A basic training evaluation is performed in-domain with perplexity, loss, top-K accuracy, and other standards measures. \citet{linzen-2020-accelerate} is critical of this practice, arguing that in-domain testing hurts the generalisation of the model. The test set may be randomly selected, but it still contains the same domain biases as the dataset, leading to overestimating performance and problems with generalisation. Linzen's remedy is more comparable general task leader boards to provide out of domain evaluation for comparison across models. In contrast, the one advantage of an unsupervised model with inference rules to infer measures such as suspense and salience is that the labels for these are unconnected to the training. There's less danger of learning surface-level features that overestimate performance failing to generalise. Though widely adopted across a range of downstream tasks there are still a lot of challenges ahead with these models; \citet{bommasani2021opportunities} analyse the risk and opportunities in detail of what they call \textit{foundation models}.

The most widespread method of applying deep learning in the cited literature is supervised learning. In this paradigm, suspense, surprise, salience, or other measures could be learnt by annotating whole stories at the sentence or block level and then training models on this dataset. This is standard for a range of tasks such as POS tagging, NER, co-references, etc.  Instead, this thesis takes an unsupervised approach. Unsupervised methods don't require annotated labels and can be trained directly from data via techniques such as contrasting learning, predicting missing data, or reconstructing input. By reusing the representations of unsupervised deep learning models with simple reasoning derived from the theory in the last chapter, with the addition of reusing the same models for planning in story generation. There are a number of practical and theoretical reasons for rejecting a supervised approach and for adopting the unsupervised one. The problems with supervision:

\begin{itemize}
  \item \textbf{Subjectivity:} Lots of traditional supervised tasks are relatively narrow. In NLP, this could be highlighting named entities, or with images, it could be labelling known categories. Generally, it would be expected there would be a high agreement between annotators. There are areas of NLP annotation that are looser and where there may be more noise in annotations, such as entailment or factual datasets where there are more likely to be mistakes. Storytelling annotation for salience, surprise and suspense are far more subjective. Subjectivity is obviously a problem for evaluation as well but is more of an issue for training since it will build errors directly into the model. Even when the guidelines are well-written in piloted, there would be a lower inter-annotator agreement. Collecting a Gold standard dataset would thus require a larger number of annotators per story with training.
\item \textbf{Synthetic Datasets: } As mentioned earlier with the ROC corpus example, the initial release had a problem with data collection.  A small theoretical dataset that presented a cloze task with more/less suspenseful elements could be an option, or a ranking task. There is a high possibility with a synthetic dataset of creating unrealistic stories, and a model trained on them would not generalise and risks anomalies that would exaggerate model performance. Hence the focus of this thesis is on learning from real-world short and longer-form stories.
\item \textbf{Dataset Size:} Generally, deep learning methods need a large amount of data to train on in the thousands of examples. These are easily obtained with unsupervised learning but not with supervised labels. The suspense work uses the WritingPrompts \citep{fan-etal-2018-hierarchical} dataset of $300K$ stories which are an average length of $53$ sentences. Labelling greater than $>10K$ stories for training with multiple high-quality annotations would be far beyond the budget scope for this thesis. It would also cover a small part of the dataset and make overfitting more likely.  Even if feasible, it would be a poor use of resources. For the later work on salience, it becomes worse. Salience inference is on long book-length texts. \textit{Great Expectations}, though longer than most books, is close to $200K$ words. Sentence by sentence salience annotation is ridiculous for works of these lengths. \citet{Papalampidi2019MoviePA} for example, annotates circa $200$ movie scripts with $5$ turning points (key plot points) per movie at the scene level, and this is still a costly undertaking.
\item \textbf{Narrowness:} This is another problem besides cost. Salience, surprise, or suspense are only narrow measures of story properties. Many other plot measures could be significant, such as Propp style functions, turning points, affect changes to characters goals or situations, and tracking object or communication state. Labelling all of these is an unscalable undertaking. Each set of labels is also suitable only for training for that specific task. On the other hand, unsupervised methods contain much relevant knowledge to be exploited for inferring across tasks.
\item \textbf{Weak Supervision:} There are alternatives such as Distant Supervision \citep{mintz-etal-2009-distant} and frameworks such as Snorkel \citep{ratner2017snorkel} for automating the process. Distant supervision is a method that commonly involves incrementally labelling Silver standard data and then training a classifier. This is a valid alternative approach, but even this would require bootstrapping of a substantial amount of initial data to be viable.

\end{itemize}

The positive case for an unsupervised approach is far more compelling:

\begin{itemize}
  \item \textbf{Derived Theory: } As reviewed in the last chapter, there is much relevant narrative theory. While the theory and analysis is often complex, the practical reasoning requiring to infer it is often not. For example, the \textit{Ely} \textit{suspense} is a variance of future expectations, whilst \textit{surprise} is a difference between expectations and what happens. \textit{Barthes' Cardinal Functions} are the events that change the story when removed. With story generation, many plots can be described as derivatives of previous cases, given the changed context. As such, models can, potentially, infer these with relatively simple heuristic rules from unsupervised ML models trained on story data. The philosophy behind all the experimental work in this thesis is to link these models with theory in a more principled way. The intuition behind this is this takes the best advantage of the implicit semantic knowledge learnt by models. It also more directly tests the theory within narratology. The method can be thought of as a \textit{neuro-symbolic} method \citep{DBLP:journals/corr/abs-2012-05876} where neural methods are applied with more traditional reasoning, though the reasoning is relatively simple.
\item \textbf{Power of ML Models:} The early part of this chapter covered more traditional AI symbolic rule and case systems. These models are similar in principle to the formal structural narrative tradition. As has been argued earlier in the chapter, both theoretical and empirical work demonstrates that ML models can learn similar structural relations, though with the noted grounding concern. The intuition is then that combining theory-based heuristics with the neural ML model will effectively predict narrative metrics such as surprise, suspense, and salience. These neural ML models can also learn directly from data and, as argued earlier, are not as brittle as symbolic methods and better at representing gradients of meaning.
  \item \textbf{Cognitive Relevance:} The classic formalist narratives mirror closely traditional symbolic causal rule chains. As has been reviewed, there has been a shift in narratology from formal causal chains to more cognitive memory approaches. However, there are significant differences between current ML models and neuroscience models. The brain isn't retrieving memories from large dictionaries with the highest dot-product match between vectors or sampling from Gaussians with variational methods. Nevertheless, as reviewed earlier, vector space models correspond more closely with cognitive ideas of meaning. For example, other cognitive work on storytelling \citep{sap-etal-2020-recollection} has shown that various deep learning memory models have power in explaining cognitively the difference between recalled and imagined stories.
  \item \textbf{Abundance of data:} The opposite of the reason for avoiding supervised learning is that story data are abundant in books, web stories, or movie scripts currently available. Far more than any tiny supervised dataset that would be feasible. As discussed earlier, having a larger and more diverse training dataset can help avoid overfitting and improve generalisation.
   \item \textbf{Universality:} Another advantage with inference heuristics applied on top of unsupervised models is they are more broadly applicable. A supervised method may learn cues that are only applicable in a domain, and performance can be lower when transferred across domains. Domain adaptation is more straightforward if it only requires domain-specific data. For example, this thesis trains on data from movies, short web forum stories, contemporary free fiction, and historical fiction. Cues and learned plot knowledge for contemporary fiction would be very different from historical fiction. So it would be likely a supervised system trained only on one would suffer on the other. This would also happen if extended to further domains beyond the thesis such as plays, children's literature, Bollywood movies, or more factual narrative domains such as news corpora, biography or histories. 
\end{itemize}

This chapter has reviewed earlier logical and case methods and their applications in story generation and comprehension tasks. It has introduced vector space models from simpler word embeddings models to more recent large \textit{transformer} models. It has reviewed the theoretical and practical work on these models that demonstrate the capability to represent narrative concepts and argued for an unsupervised rather than a supervised approach instead exploiting the implicit knowledge captured by the deep learning model with heuristic rules aligned with theory. Finally, related areas have been reviewed which shows a general convergence on the advocated themes. The next chapter is on the annotation of a dataset for evaluating suspense in short stories.

\chapter{Suspense Annotation}

\label{chap:suspenseannotation}

\section{Introduction}

While the models of suspense and surprise in this thesis can be learnt without supervised labels, the same isn't true with evaluation. Evaluation requires that suspense be compared to actual human judgements on suspense. This chapter outlines the annotation that were used to evaluate suspense in Chapter \ref{chap:rolloutsuspense} and Chapter \ref{chap:tdvaesuspense}.

There isn't a suitable corpus of suspense annotations. The closest is by \citet{delatorre2018confronting}. Delatorre et al. whose approach is to annotate a single-story passage by passage on a 9 point scale. The main purpose of this study is to manipulate this story to test alternative hypotheses about suspense via variations of the single story. An example is an alternative version of the story where key plot developments are revealed at the beginning rather than the end to test how important uncertainty of the outcome is. This approach of annotating a single story, even with variations, is insufficient to evaluate unsupervised suspense models. Evaluation requires a broader statistical base with a diversity of writers and genres. The final evaluation dataset in the thesis draws inspiration from this approach by annotating multiple stories sentence by sentence. The main difference is the annotation employs a relative scale rather than an absolute one. 

In subsequent work \citet{DBLP:journals/kbs/DelatorreLST20} has applied a predictive model to a larger set, employing a similar manipulation approach by asking participants to rate the suspense for ten different outcomes per story. The work also measures the correlation between these suspense judgements and show a link between these and electromyography (EMG) physiological reactions. \citet{DBLP:journals/access/DelatorreLH21} develops a system that modifies the suspense in examples through a method that changes the emotional valence of terms and actions by modifying a template. Both of these works were published after the annotation in this chapter was performed. One thing to note is the approach taken is to filter a subset of overall stories that may be suspenseful. This is the approach behind the unsuccessful whole story annotation to follow. Also, both these papers focus on the emotional valence of words and actions, particularly negative outcomes. While this thesis was not influenced directly, the motivation is shared with the \textit{Impact} adjusted models in the next chapter. The \textit{Impact} adjusted metrics approximate the importance of state via \citet{ely2015suspense}'s $\alpha$ extension

\section{Dataset}

The annotation corpus chosen for both whole story and sentence by sentence annotation is \textit{WritingPrompts} \citep{fan-etal-2018-hierarchical}. \textit{WritingPrompts} is a corpus of short stories sourced from a Reddit forum of the same name \href{https://www.reddit.com/r/WritingPrompts/}{r/WritingPrompts}. The forum consists of circa $300$K stories pre-split $80$\%, $10$\%, and $10$\% in training validation and test datasets. The dataset comes from a creative writing forum that consists of topic threads with a given prompt, for example, \textit{[WP] The format The Mage, the Warrior, and the Priest}, or \textit{[WP] You had a really bad break up with your ex… 300 years ago. Neither of you realised the other was immortal until you met again while grocery shopping.} Forum users respond to the topic and write their own short stories. For annotating suspense, this dataset has a number of advantages:

\begin{itemize}
  \item \textbf{Length:} Annotation is expensive and particularly the sentence by sentence annotation detailed later in the chapter. Real book-length stories or even published short stories of half a dozen pages would be too long to be feasible to annotate. Annotation requires that the annotators read the story, comprehend it, and make judgements about the stories. An evaluation requires that there is reasonable agreement between annotators and the quality of annotations can be validated. The average length of a \textit{WritingPrompts} story is $53$ sentences which means the stories are long enough to contain a reasonable plot length beyond the five sentences of ROC \citep{mostafazadeh-etal-2016-corpus} while still being feasible to annotate.
 \item \textbf{Natural Stories:} An alternative would be to extract short summaries of books or movies from either Wikipedia or other relevant web sources such as IMDB. The problem with these summaries is while they contain the full plot, summaries typically contain just a description of the main events. Often they are written in an abstract way. They lack many modalities of real stories, such as dialogue or description of scenes. Although shortened, \textit{WritingPrompts} stories have these dynamics, which are important aspects of real storytelling and will influence suspense and surprise. Synthetic stories could be collected via crowdsourcing, but as discussed in Chapter \ref{chap:backgroundml}, there is a high danger of artificially creating biases and shortcuts. Typically crowdworkers are motivated to work quickly and not to write better stories, whereas with \textit{WritingPrompts} contributors write them because of interest and so they are likely to be better stories.    
\item \textbf{Topicality:} Because of the format of the forum the dataset collected from the corpus also has multiple examples for a given prompt. It also contains a diverse range of subjects, writers, plots, and writing styles.
\end{itemize}

There are also disadvantages:

\begin{itemize}
  \item \textbf{Lack of Metadata:} One disadvantage, as opposed to some plot summary sources, is the lack of metadata about stories. Metadata that that could be used to filter more typically suspenseful genres such as  \textit{thriller} or \textit{horror}.
 \item \textbf{Slant:} The other issue is more subjective. While the stories are often creative, there is a slant towards popular online genres such as fantasy or sci-fi. The forum is intended for creative writing exercises. Hence, some of the prompts and written stories can be unusual.
\end{itemize}

Slant to specifying the scope from the introduction (Chapter \ref{chap:introduction}) of stories as being strongly culturally, medium and genre-specific. The models in the thesis are about the reader's expectations. So naturally, these will not be expected to transfer. The current chapter annotates the \textit{WritingPrompts} dataset for evaluation of suspense. \textit{WritingPrompts} as it's an online creative, the stories can be quite unusual as the writers are trying to be more imaginative with their stories. Often the prompts suggest unusual situations. Yet, at the same time, the writers draw a lot from popular works, especially fantasy, sci-fi and classics. Some stories can have a fan-fiction feel, but they try to be creative with the situations. Finally, because they are amateur writers, there can both be some excellent and bad quality writing. These make the corpus more niche than mainstream fiction and more variable in theme, writing style, and quality. Overall, these properties mean the stories are less likely to follow traditional plots and make the stories more challenging both for human readers and for computational models.

To address the lack of metadata some exploratory work was done on topic modelling on \textit{WritingPrompts}. With the \textit{Gensim} library \citep{rehurek2011gensim} both LDA (Latent Dirichlet Allocation; \citealt{NIPS2010_71f6278d}) topic modelling and hierarchical non-parametric  \citep{DBLP:journals/jmlr/WangPB11} topic models were applied. The intuition behind this is that topic models would assist with analysing the datasets and which stories may be deemed more or less surprising or suspenseful. Such as to test if there are correlations between say \textit{thriller} or \textit{murder mystery} type topics in either the annotations or the inferences from the model.
Unfortunately, all methods produced poorly defined clusters and are therefore not part of further analysis. 

Through both the whole story annotations and sentence by sentence annotations the following pre-processing and selection process was applied:

\begin{enumerate}
  \item \textbf{Validation set and Test set Stories:} Annotated examples are taken from the predefined splits from the dataset.
  \item \textbf{Cleanup:} {Whatthelang} for detecting language and filtering out all non-English texts.\footnote{\url{https://github.com/indix/whatthelang}, the library is built on \textit{Fasttext} classifiers \citep{joulin2017bag}} As it's a web scraped dataset, \textit{WritingPrompts} can have some strange line breaks, page breaks and repeating punctuation. Some cleanup is performed for these, which is matched by that used in the training of models. The processing stops, for example, are a long list of * demarcating a paragraph being recognised as a sentence which can cause issues with inference and annotation.
\item \textbf{Filter for length:} The test set is filtered, so it contains only stories between $25$ and $75$ sentences after the above processing. The rationale is that this is the typical suitable length for both the corpus stories and for annotation. It excludes brief comments or suggestions for short and not fully written out stories and some unusual but extremely long stories that can be hundreds of sentences and be too long to annotate. 
\item \textbf{Shuffle:} Those remaining validation and test set stories were shuffled and the first $500$ were taken of each and split into blocks of $100$. For annotation, the first $100$ of each have been annotated. The remainder has not been used for human evaluation but have been for automated evaluations such as cloze tasks or BLEU score. The rationale behind a randomised set, is first, to be representative. The second is the Ely theory is broader than some genre-specific ideas about suspense and so should apply to varying degrees to all narratives. Hence the decision to use a random subset for evaluation rather than to cherry-pick from stories that may be expected to have higher degrees of suspense based on genre.
\end{enumerate}

This section will conclude with an example of a \textit{WritingPrompts} and the sentence segmentation; this example is plotted in the following chapter with annotator judgements and model predictions of suspense.

\begin{enumerate}
\setcounter{enumi}{-1}
\item As I finished up my research on Alligator breeding habits for a story I was tasked with writing , a bell began to ring loudly throughout the office .
\item I could feel the sound vibrating off the cubicle walls .
\item I looked over my cubicle wall to ask a co - worker what the bell was for .
\item I watched as he calmly opened his desk drawer , to reveal a small armory .
\item There were multiple handguns , knives and magazines and other assorted weapons neatly stashed away .
\item `` What the hell is that for ? ''
\item I questioned loudly , and nervously .
\item The man looked me in the eyes , and pointed his handgun at my face .
\item I saw my life flash before my eyes , and could n't understand what circumstances had arisen to put me in this position .
\item I heard the gun fire , and the sound of the shot rang through my ears .
\item I heard something hit the ground loudly behind me .
\item I turned to see the woman who had hired me yesterday , lying in a pool of blood on the floor .
\item She was holding a rifle in her arms .
\item I looked back at the man who had apparently just saved my life .
\item He seemed to be about 40 or so , well built , muscular and had a scar down the right side of his face that went from his forehead down to his beard .
\item `` She liked to go after the new hires '' he explained in a deep voice .
\item `` She hires the ones she wants to kill ''
\item I was n't sure what to make of this , but my thoughts were cut off by the sounds of screaming throughout the building .
\item `` What 's happening ''
\item I asked , barely able to look my savior in the eyes .
\item `` You survive today , and you 'll receive a bonus of \$5,000 and your salary will be raised 5 \% ''
\item I cut the man off .
\item `` What does that ? ''
\item He continued to speak , while motioning me to stop taking .
\item `` I 'll keep you alive , if you give me your bonus and half your raise 
\item He finished .
\item I just nodded , still unable to understand the position I was in .
\item He grabbed my arm so hard I thought it would break , and pulled me over the cubicle wall , and under his desk .
\item Then , he placed a gun in my hand .
\item `` The safety is on , and it 's fully loaded with one in the chamber ''
\item He said , pointing to the safety switch .
\item The weapon felt heavy in my hand , I flicked the safety off with my thumb and gripped the gun tightly .
\item The man looked down at his watch .
\item `` 45 minutes to go ''
\end{enumerate}

\section{Whole Story Annotation}

\begin{figure}[htbp]
\centering
\includegraphics[width=1.0\textwidth]{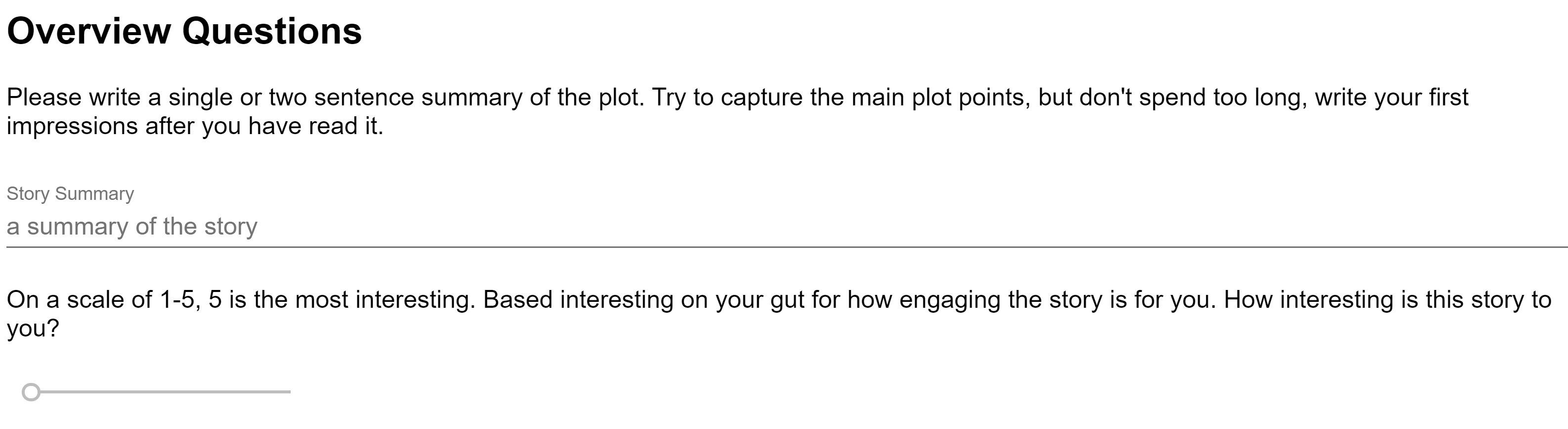}
\caption{Whole sentence annotation questions.}
 \label{fig:wholeannoverview}
\end{figure}

There were two phases of annotation. The first of which is whole story annotation, and the second is sentence by sentence suspense annotation. Both annotations exercises are intended to be complimentary. The \textit{whole story} annotation is annotating the randomly sampled WP stories with judgements about surprise, suspense and metadata. An overview of the data collected:

\begin{itemize}
 \item \textbf{The Story Presented: } Annotators are presented with the whole story in a single block of text. 
  \item \textbf{Description:} The first question asks for a summary of the story, shown in Figure \ref{fig:wholeannoverview}. There are two motives for this question: The first is quality control in that it is a way to test annotators have read and comprehended the story. The question is intended as one filter for approving or rejecting annotators. The second is an intention that the summary may be valuable in a secondary task such as providing training data for summarisation.
 \item \textbf{Story Sentiment:} Is on a five-point scale between strong negative sentiment (i.e. tragedy) and strongly positive (i.e. uplifting and happy). In some of the reviewed theories of suspense there is a stronger link between possible negative outcomes and suspense, and so this is to test whether there is a correlation.
  \item \textbf{Genre:} Also shown in Figure \ref{fig:wholeannoverview}. As noted, the unsupervised topic modelling was unsuccessful in identifying cohesive topics. A question is asked on Genre with the categories taken from the Wikipedia classification for literary stories with an \textit{Other} category. The motive for collecting is to better understand the genres of stories within the dataset, and as a secondary form of analysis to correlate with suspense and surprise. Do particular genres such as \textit{Thriller} have more suspense, for example?
 \item \textbf{Reader Suspense:} Questions are shown to ask annotators to rate the level of suspense, surprise and how interesting the story is on a scale of 1-5. The annotator phrasing is italicised:
\begin{itemize}
 \item \textbf{Interest:} \textit{Intuitively how interesting do you find the story. This is how much do you enjoy reading it. It is not some measure of literary merit, but a gut feel for how much you like it.} Interest is included to test correlations between how good the reader thinks the story is and other metrics such as suspense and surprise, i.e. does more suspense make a better story?
 \item \textbf{Suspense:} \textit{A state or feeling of excited or anxious uncertainty about what may happen. Suspense is forward-looking uncertainty. It is not just that the outcome is uncertain, but also the outcome must be of consequence for the story as a whole, i.e. important.}
 \item \textbf{Surprise:} \textit{An unexpected event that happens. Unlike suspense, there isn't anticipation it will happen.}
 \item \textbf{Emotional Resonance:} \textit{This is an intuitive sense of how strong the emotional connection is with reading the story or to the characters depending on the question. The difference between this is the level of emotional intensity and not whether positive or negative.} Once again, this is trying to separate the emotional content of the stories with surprise and suspense. As has been noted this is an important influence in the Delatorre papers. 
\end{itemize}
\item \textbf{Additional instructions are provided to annotators:}
\begin{itemize}
	\item \textbf{Suspense vs Surprise:} Imagine a family sitting around having dinner having a conversation, a bomb goes without warning. This is surprise, there isn't a warning. If there are cues that something bad might happen that builds the tension, and then the bomb goes off, this is suspense. When judging the question, please rate the story plot as a whole by considering the most crucial plot events. 
\item \textbf{Readers vs Character Perspective:} There are two sets of 3 similar questions from a reader's perspective and a character's perspective. For readers, answer the questions based on your own impressions and view. The characters more complicated: For these questions, imagine yourself in the shoes of the main characters; what do they think about happens in the story. Are the events surprising or suspenseful to them even if they may not be to you. For example, in a romance, it seems obvious that the characters will get together, but there may be a high degree of uncertainty to one or more of the main characters.
 \end{itemize}
\end{itemize}

\begin{figure}[htbp]
\centering
\includegraphics[width=1.0\textwidth]{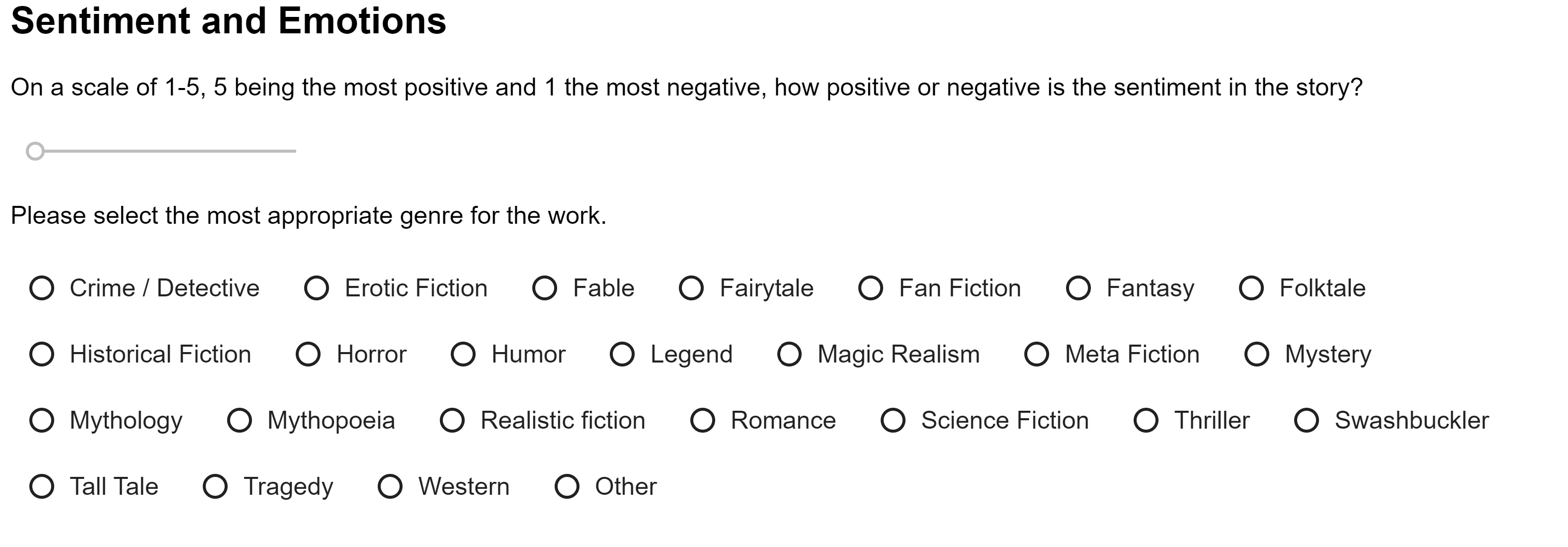}
\caption{Sentiment and genre questions from the whole sentence annotation.}
 \label{fig:wholeanngenre}
\end{figure}

\begin{figure}[htbp]
\centering
\includegraphics[width=1.0\textwidth]{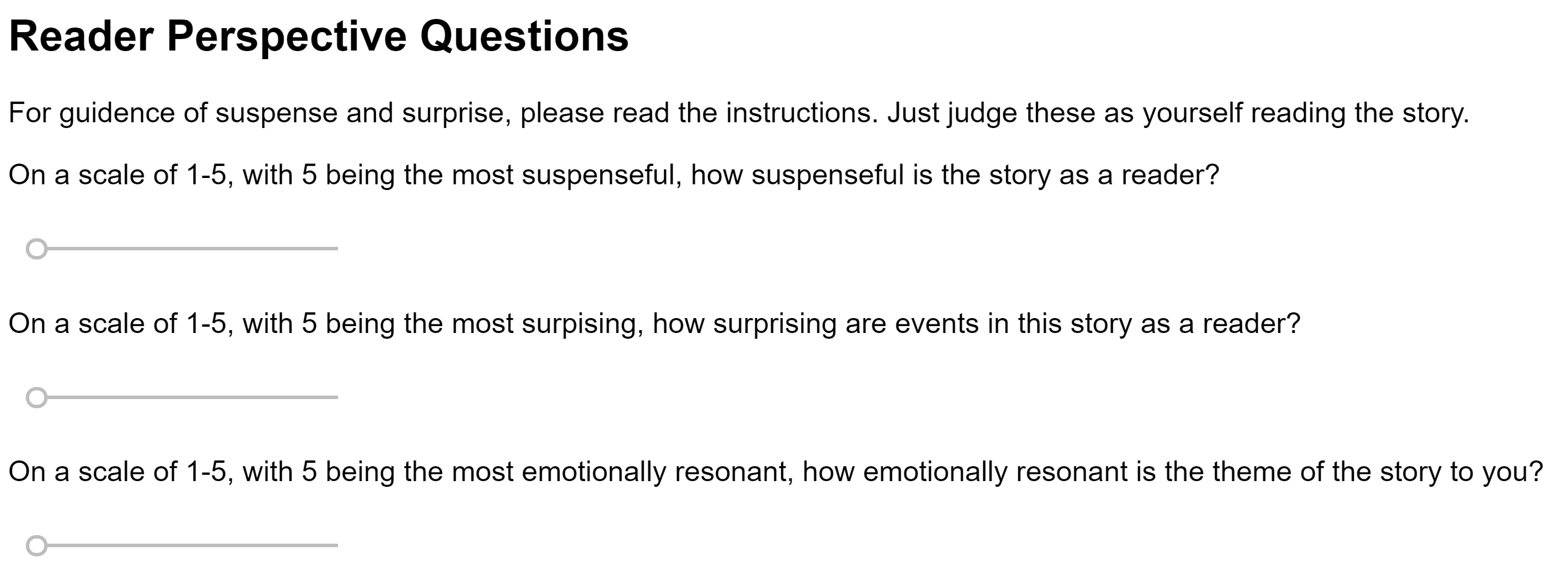}
\caption{Reader perspective questions from the whole sentence annotation.}
 \label{fig:wholeannreader}
\end{figure}

\begin{figure}[htbp]
\centering
\includegraphics[width=1.0\textwidth]{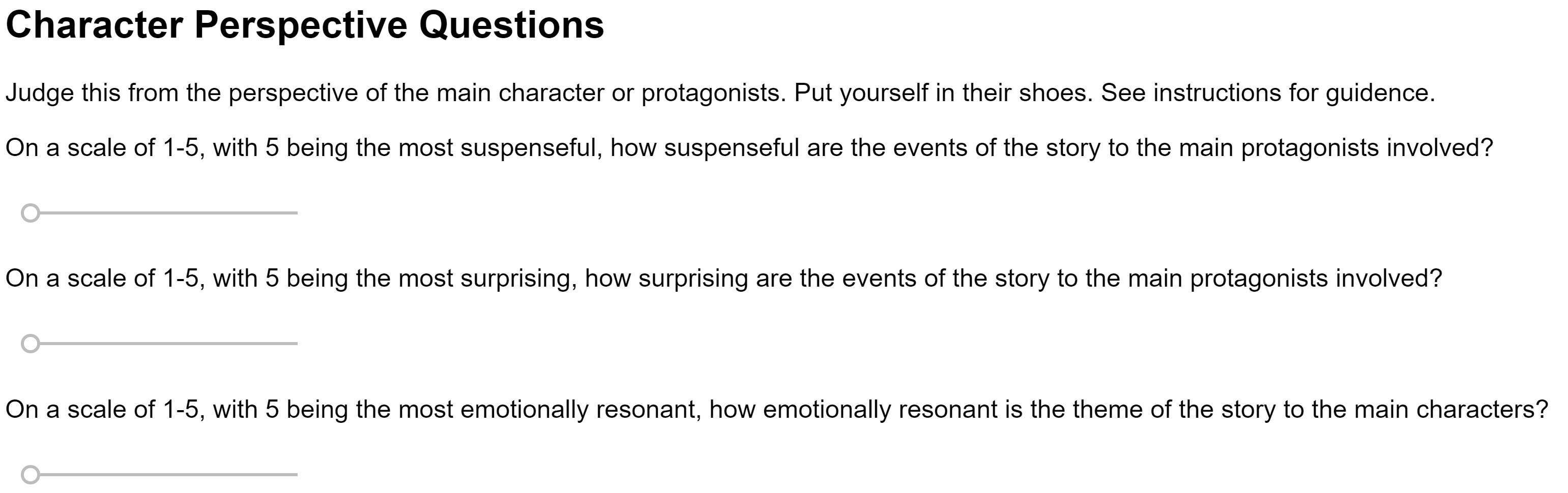}
\caption{Character perspective questions from the whole sentence annotation.}
 \label{fig:wholeanncharacter}
\end{figure}

The advantage of questions that evaluate a whole story is that they are relatively quick for annotators to do compared with making sentence-by-sentence judgements. The whole story annotation thus allows the collecting of more detailed questions that would be burdensome for an individual annotator on a per sentence basis. On the other hand, it's not clear how easy suspense or surprise are to judge at the level of a whole story versus for parts of the plot. The core of the annotation are the three questions on surprise, suspense and emotional resonance. The scale is a five-point one as it was deemed that this would be the finest granularity annotators would be easily able to assess. The main design decision was to split up the questions into reader's questions and character's perspectives. The rationale for splitting these ties into the reviewed suspense theory that suggests that reader's feel suspense on behalf of the character's; splitting the questions tests whether there is really a difference between readers and characters expectations. 

A second relevant point to the data collection is that it should not be expected there would be a Gold standard. Stories are inherently subjective, and so a story someone finds fascinating and full of suspense may be tedious and predictable for someone else. Therefore legitimate disagreements are to be expected in the annotations. It should also be expected this disagreement will act as an upper bound on the predictions, as naturally, the models performance can not surpass disagreement amongst annotators. The following is the collection process from AMT (Amazon Mechanical Turk):

\begin{enumerate}
	\item \textbf{Recruitment:} MTurk crowd workers are recruited from English speaking countries. The payment rate for the average assignment, following a small trial, is the British minimum wage. 
	\item \textbf{Collections:} $5$ annotations were collected for each of the $500$ stories. The data collection was in several batches.
	\item \textbf{Assignment:} The tasks are on the Amazon Mechanical Turk (AMT) platform. AMT randomly assigns stories from an uncompleted batch as the are requested by annotators, and each can do as few or as many as they want. The consequence is that each of the $5$ annotations per story will have been completed by a different subset of annotators.
	\item \textbf{Filtering:} Only annotators with a $>98\%$ approval rate and with over $1000$ tasks completed were allowed to accept the task. The AMT \textit{Master Workers} qualification was turned on, which demands a 20\% premium and is the AMT algorithmic filter for workers with consistently high approval rates across a wide range of tasks.
	\item \textbf{Quality Control:} Because the judgements are inherently subjective it is hard to put in checker questions to verify quality as would be typical in a factual annotation.  The main quality control was a combination of checking the summaries written for each story with the times taken, looking for unusually quick annotators and or annotators that hadn't seemed to have read the story. Generally, the responses' quality was reasonable, and annotators were only excluded if they repeatedly gave answers that suggested they hadn't read or comprehended the story, about $12$\% of the annotators. For these rejected tasks further annotations were completed.
\end{enumerate}

\begin{figure}[htbp]
\centering
\includegraphics[width=1.0\textwidth]{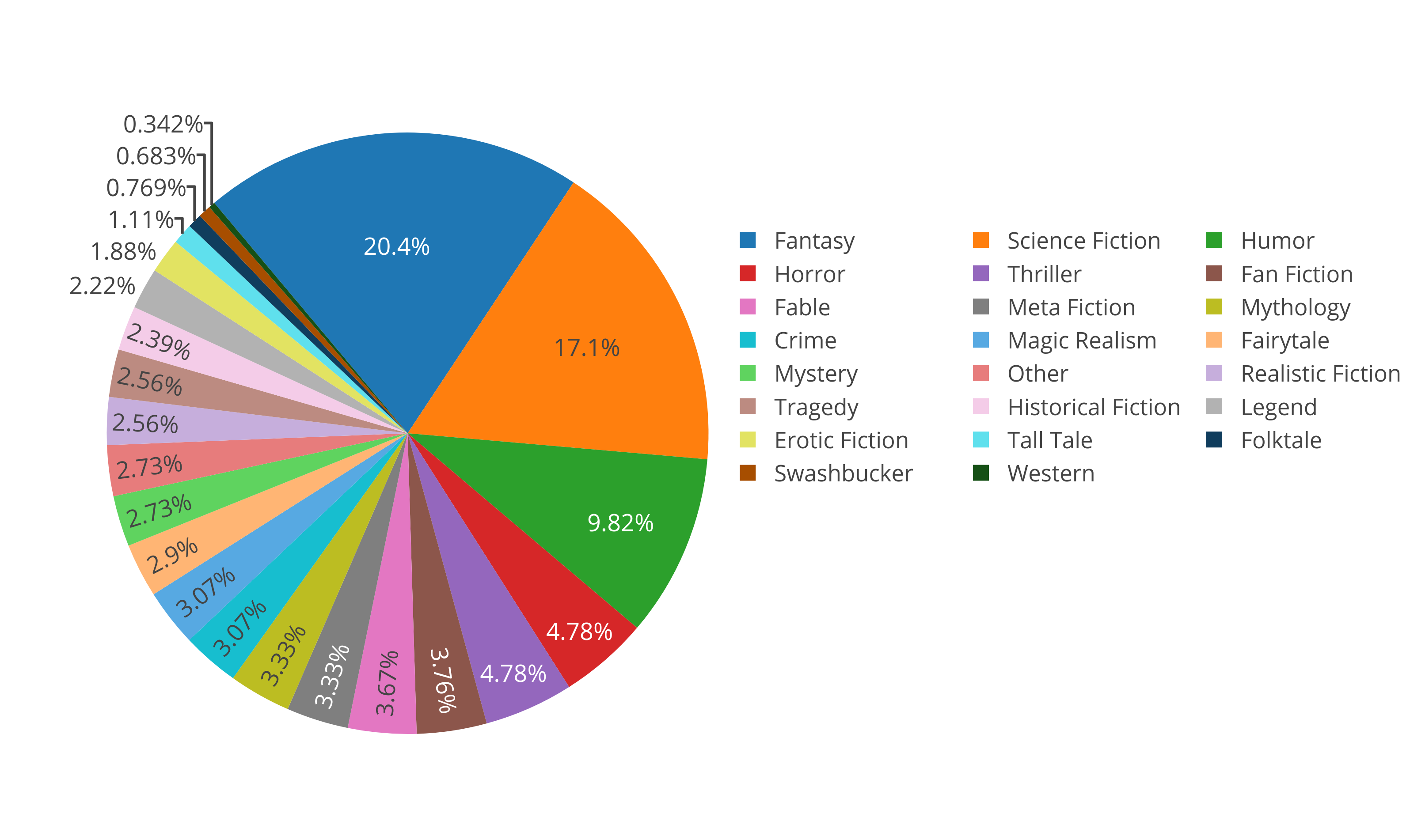}
\caption{The annotated genres of the WritingPrompts $500$ randomly sampled test set stories. The chart shows all approved annotators judgements and not a consensus judgement per story.}
 \label{fig:wholeanngenres}
\end{figure}

\begin{figure}[htbp]
\centering
\includegraphics[width=1.0\textwidth]{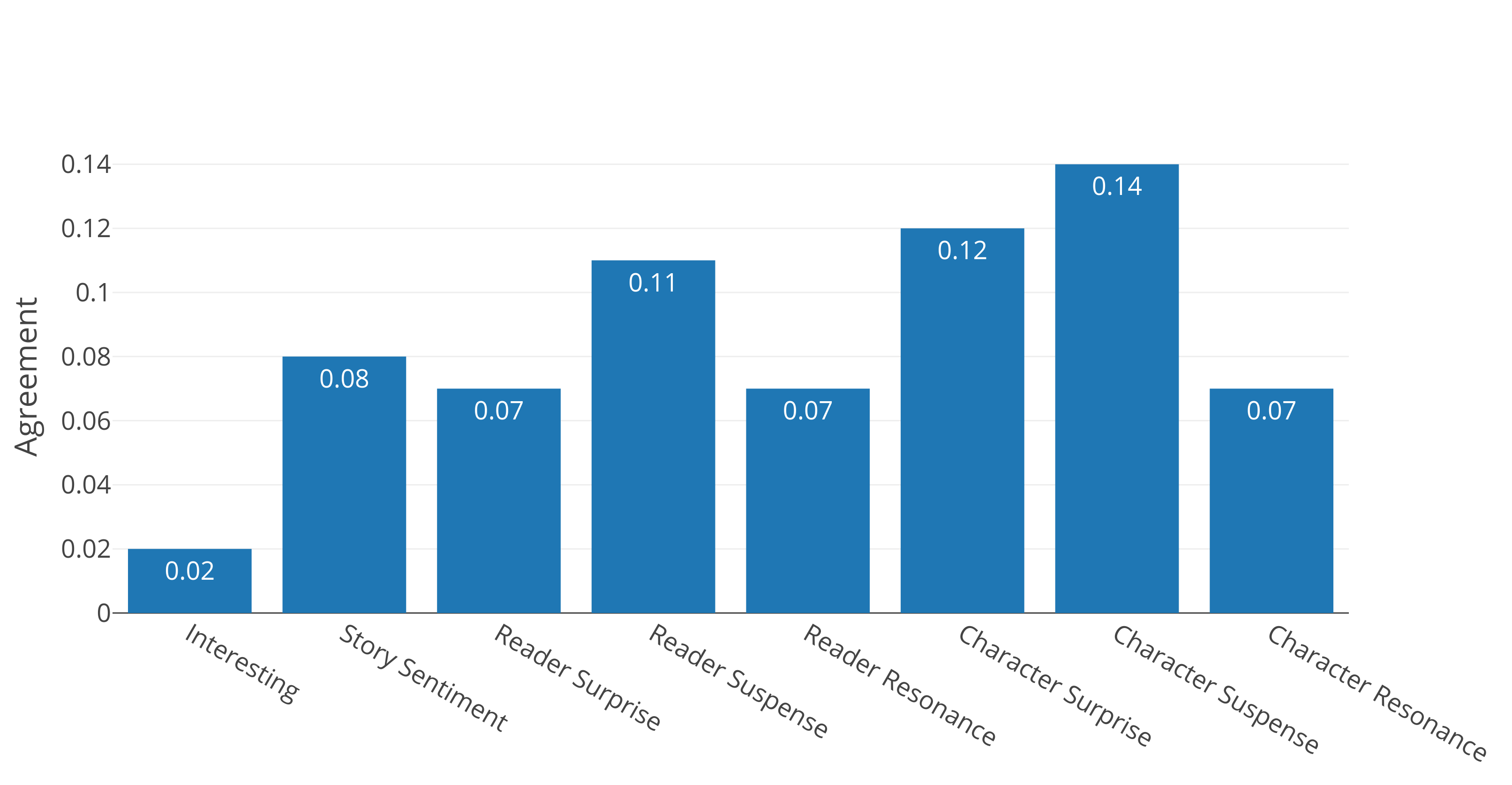}
\caption{The ordinal questions from the annotation with inter-annotator agreement according to \citet{krippendorff2011agreement} $\alpha$. The scale is $-1$ (perfect disagreement) to $1$ (perfect agreement) with $0$ representing random.}
 \label{fig:wholeannagreement}
\end{figure}

The analysis: See Figure \ref{fig:wholeanngenres} genres for a breakdown of the genres, which is skewed towards \textit{Fantasy}, \textit{Science Fiction}, and \textit{Humor} genres. This is perhaps not unsurprising for an online web corpus. There is, however, a major issue with the annotations, and that is inter-annotator agreement on all questions is extremely low. Krippendorff's $\alpha$ \citep{krippendorff2011agreement} shows the inter-annotator agreement on the main questions in Figure \ref{fig:wholeannagreement}. Krippendorff's $\alpha$ is preferred to alternatives such as Scott's $\pi$ \citep{scott1955reliability} or Cohen's $\kappa$ \citep{Cohen1968WeightedKN} as it handles ordinal data and tiebreaks better than these other measures. It is also better at handling gaps in the data, which is the case here as each annotator can choose how many tasks they perform and these are randomly allocated, so each individual story is annotated by a random combination of $5$ different MTurk workers.

\begin{figure}[htbp]
\centering
\includegraphics[width=1.0\textwidth]{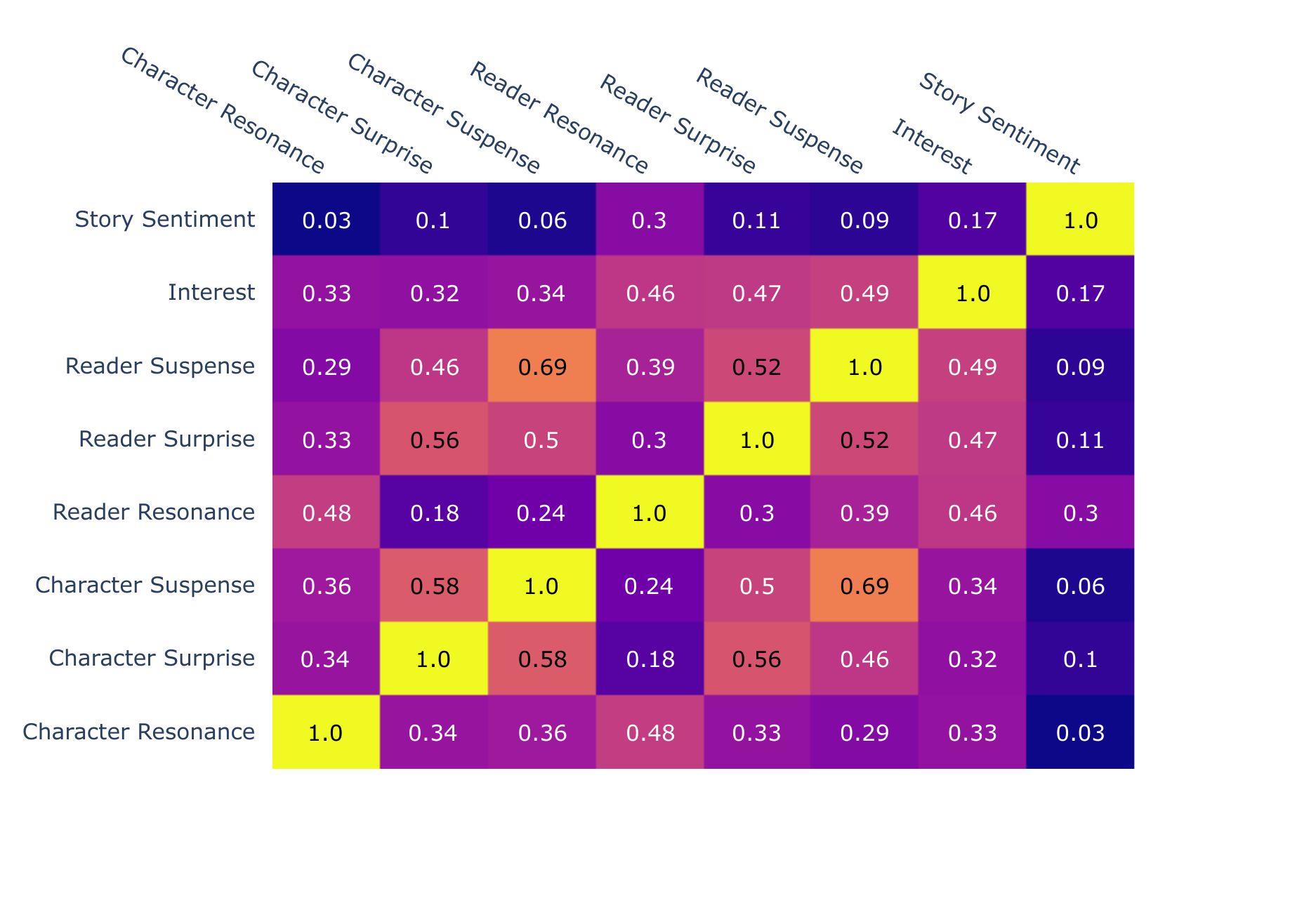}
\caption{The Spearman $\rho$ correlation between the main questions in the whole sentence annotation.  }
 \label{fig:wholeanncorrelation}
\end{figure}

The correlations are low and not that much better than random, Figure \ref{fig:wholeanncorrelation} shows the correlation. The quality of the summaries provided by annotators are generally good, and there's not much reason to expect there has been cheating. Therefore it's likely that the outcome reflects the inability to legitimately agree on the answers. Not reported in the thesis, but the sparse pairwise agreement between annotators has also been calculated. The pairwise agreement is not random but appears to suggest a number of groups that are in strong agreement with each other and disagreement with others, which would reflect an underlying difference in judgement. This disagreement makes the data collected unusable for inference of suspense with models. The reason for the difference is probably a combination of three factors: First, as noted, the stories are diverse in quality, writing style, and theme, so it may be natural to have divergence in judgement. Second, the same applies to the annotators who will have diverse backgrounds. Third, it may be much harder to make a judgement at the whole story level. The first two are inherent, the third can be controlled. In the sentence by sentence annotation steps are taken to make the exercise less subjective and repeatable in order to increase inter-annotator agreement.

One possible option was to take higher suspense genres with a higher agreement between annotators and use only this subset of genres. The top five genres were analysed separately. Note these genres include \textit{Horror}, \textit{Crime}, and \textit{Mystery}, three genres that are typically seen as more suspenseful. The results aren't reported, but both the agreement and correlation between the following results are largely the same as the dataset.

Despite the disagreement, there are still some possible insights from the dataset. Figure \ref{fig:wholeanncorrelation} is a heatmap with Spearman's $\rho$ correlation between question answers. The main relevant points for the sentence by sentence annotation: There is a reasonably strong correlation between the reader and the character questions and their equivalents. But there are still substantial differences suggesting that annotators can tell the difference between suspense or surprise from the character and reader perspective. There are positive correlations across surprise, suspense and emotional resonance and with general interest in the story. This applies to both reader and character versions of the questions, although the correlation is higher with reader questions in the $0.4x$ range. Perhaps surprisingly, there is little relationship between the overall story sentiment and any of the other questions. Whereas the resonance questions is an absolute measure, the whole story one is negative to positive. As reviewed in Chapter \ref{chap:backgroundtheory}, some theories of suspense took negative outcomes as having stronger weight than positive possibilities. The cited studies in this chapter by Delatorre et al. rely on this, as do the text generation systems \textit{Dramatis} \citep{DBLP:conf/aaai/ONeillR14} and \textit{Suspenser} \citep{Cheong2015SuspenserAS}. This is perhaps another problem with the whole story annotation. It may not be that negative events are still strongly associated with suspense, but this is hard to evaluate as a single measure over a whole story. After all, the typical narrative may have the hero imperilled at some point, and have a death of a secondary character, but have a happy ending, making the sentiment unclear to judge for the story as a whole.

\section{Sentence by Sentence Annotation}

The problems with the whole story annotation have also guided the sentence by sentence evaluation. First, it emphasises the necessity of low-level annotations in the first place to model suspense dynamics throughout the story. Second, the decision was taken to focus on the character perspective of suspense. The rationale is that the agreements are slightly higher. It is also simpler to write instructions to judge suspense neutrally if the reader is trying to judge from the character's perspective rather than their viewpoint. If a character is in danger in an action sequence, it's more likely that it will be judged as being suspenseful from a character perspective. Readers without this guidance may find it suspenseful if they are invested in the story or just boring if they are not engaged. Thus, the character perspective is more likely to produce a higher agreement, especially with improved instructions and training, even if the improvement in the whole story annotation is modest. 

\begin{figure}[tb]
\centering
\includegraphics[trim={0.5cm 0.5cm 2.0cm 2.5cm},clip,width=1.0\textwidth]{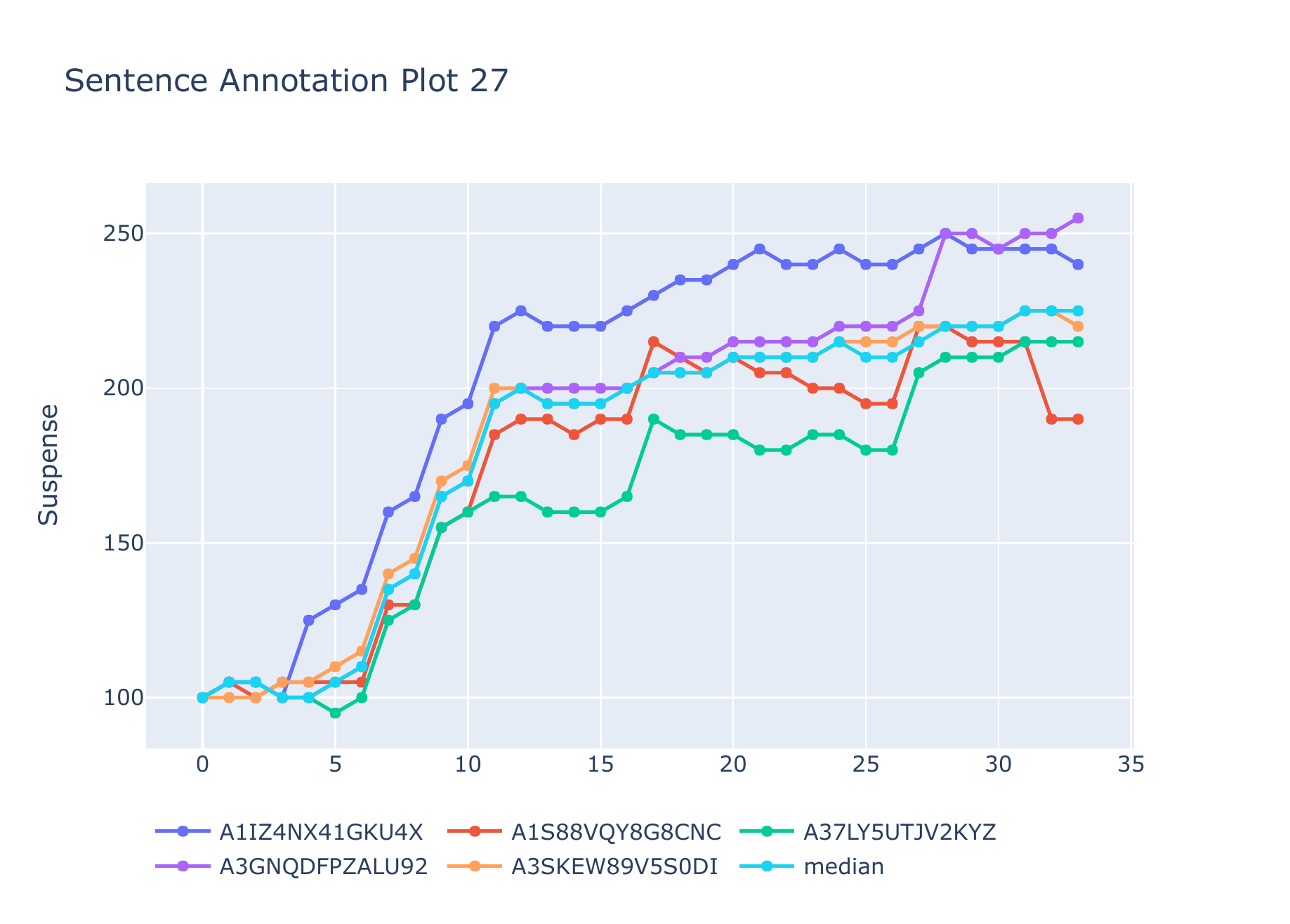}
\caption{An example of sentence by sentence annotations plotted earlier in this chapter, the story on the Crocodile. The plot shows how the relative judgements for each sentence, along the $x$ axis, with the relative suspense judgements converted into a suspense arc on the $y$ axis. This plot adds or subtracts $5$ for \textit{Increase} or \textit{Decrease} and $25$ for the big versions of each. The values are illustrative to show the plots arc more clearly; how these values are set for evaluation is discussed in the following chapter.}
 \label{fig:sentbysentannotatorplot}
\end{figure}

\begin{figure}[tb]
\centering
\includegraphics[trim={0.5cm 0.5cm 2.0cm 2.5cm},clip,width=1.0\textwidth]{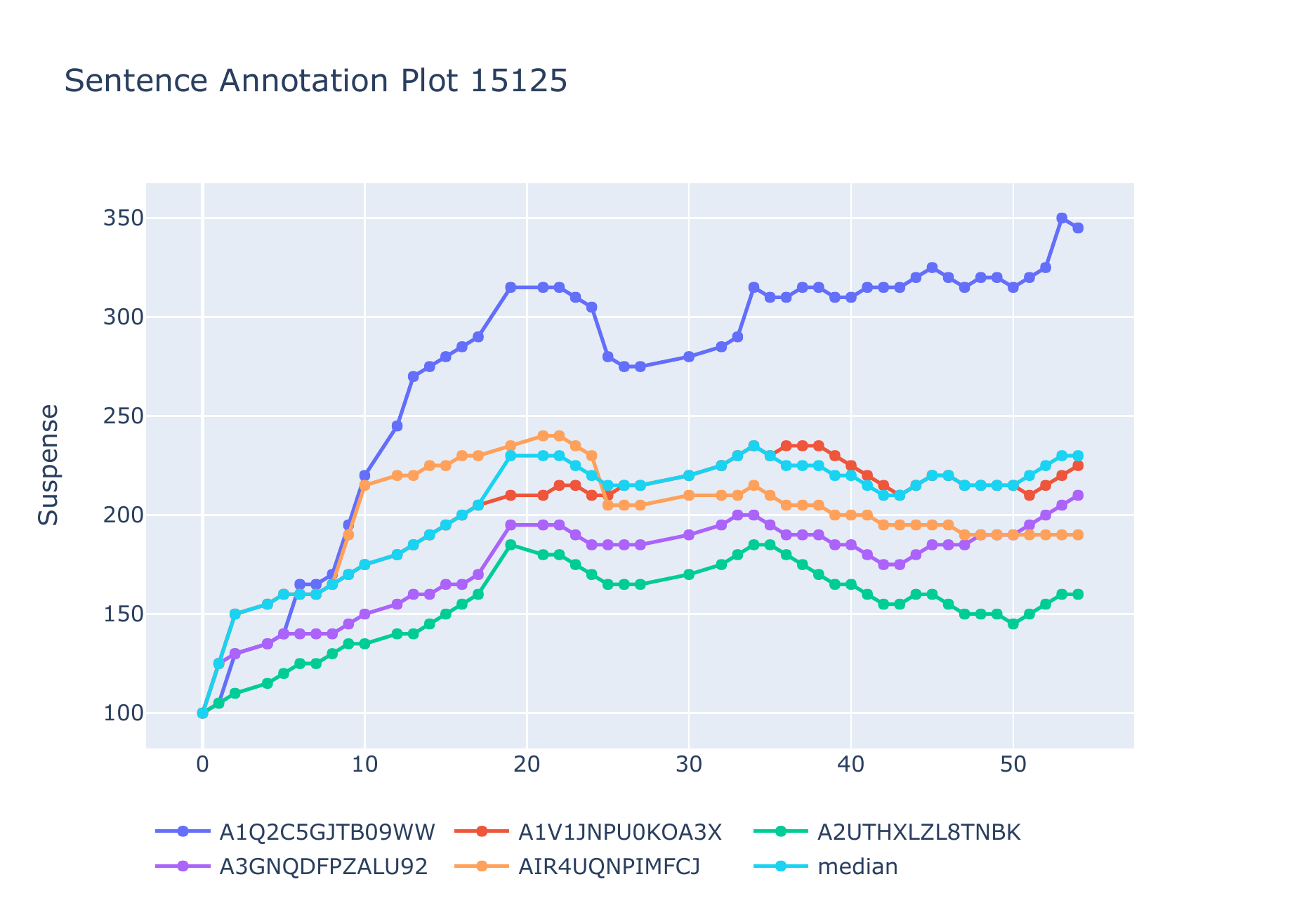}
\caption{Another example of an annotator plot. This one illustrates that there are sometimes outlier annotators with a completely different trajectory to the others. Also while in general there is an increase in suspense there are annotated stories that don't fit this pattern and are flatter.}
 \label{fig:sentbysentannotatorplotoutlier}
\end{figure}

\begin{figure}[tb]
\centering
\includegraphics[trim={0.5cm 0.5cm 2.0cm 2.5cm},clip,width=1.0\textwidth]{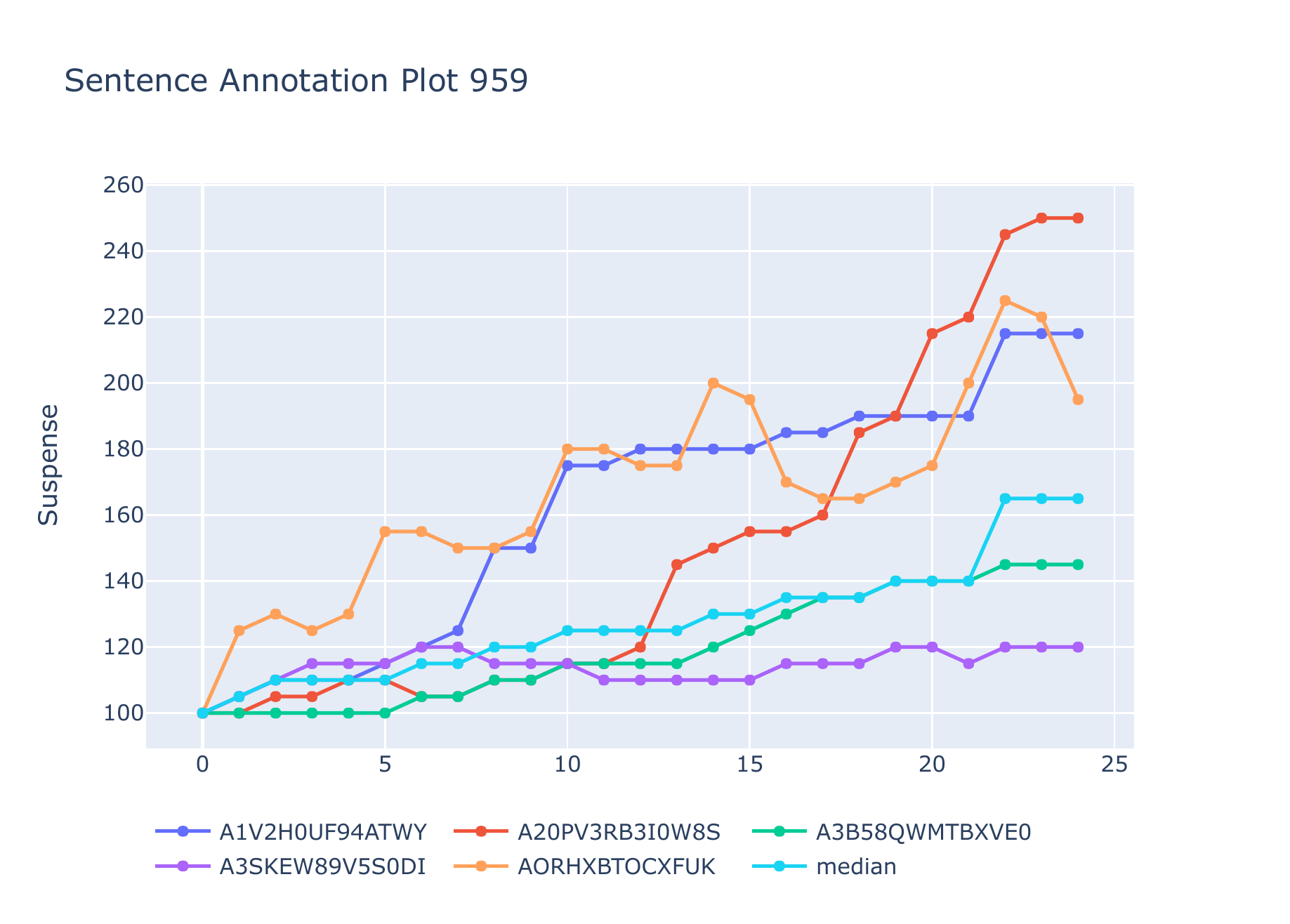}
\caption{While in most stories there is good agreement for some, there are smaller numbers such as this where there is a complete divergence of opinion. The stories are randomly sampled from \textit{WritingPrompts} which is creative writing forum. Low inter-annotator seems to correspond with stories that have unusual structure or plot, or try to do something non conventional. So the low-agreement seems to be a function of annotators just not knowing how to interpret these more unusual stories.}
 \label{fig:sentbysentannotatorplotdivergence}
\end{figure}

The annotation is performed at the sentence by sentence level to maintain a low-level judgement about suspense. An individual sentence is a small unit in a story, but when judged in the existing context of the story, it can substantially shift the level of suspense. Sentence by sentence annotation requires that the annotator comprehend the story as they are reading and be able to recall details from earlier. If there were multiple questions per sentence it would become immensely time consuming, and the annotators would lose focus, and find it much harder to follow the plot of the story, reducing annotation quality. Therefore, only character suspense is judged per question, with summary questions at the end. Annotating each sentence with a single mouse click or keyboard button press maintains the flow of reading. One major design change from \citet{delatorre2018confronting} is to use relative judgements that correspond to a five-point scale that corresponds to a \textit{Large Decrease}, \textit{Decrease}, \textit{Same}, \textit{Increase}, and \textit{Large Increase}. The rationale for a relative rather than absolute scale is the intuition is that it is easier to judge how the level changes rather than make a single absolute judgement. The relative judgements can be converted posthoc into a plot via adding or subtracting the relative judgements from an initial fixed point, which is shown in Figure \ref{fig:sentbysentannotatorplot}. The evaluation can then be modelled as an averaged pairwise rank correlation between each approved annotation and the inference curve for each model's metrics. The two correlations reported are Kendall's $\tau$ and Spearman's $\rho$. This tests whether suspenseful parts of the story correspond with human judgement and low suspense parts. Rank correlation is preferred for evaluation as this ranking judgement is more stable than fitting an absolute value for suspense. An alternative to converting the category judgement to a suspense arc would have been to try and convert the suspense plots generated in categories based on steepness between sentences. The problem with the approach is that it is really the arc of the story that is important not individual shifts between sentences. Hence comparing the overall arc or shape of suspense is more appropriate.

\begin{figure}[tb]
\centering
\includegraphics[width=1.0\textwidth]{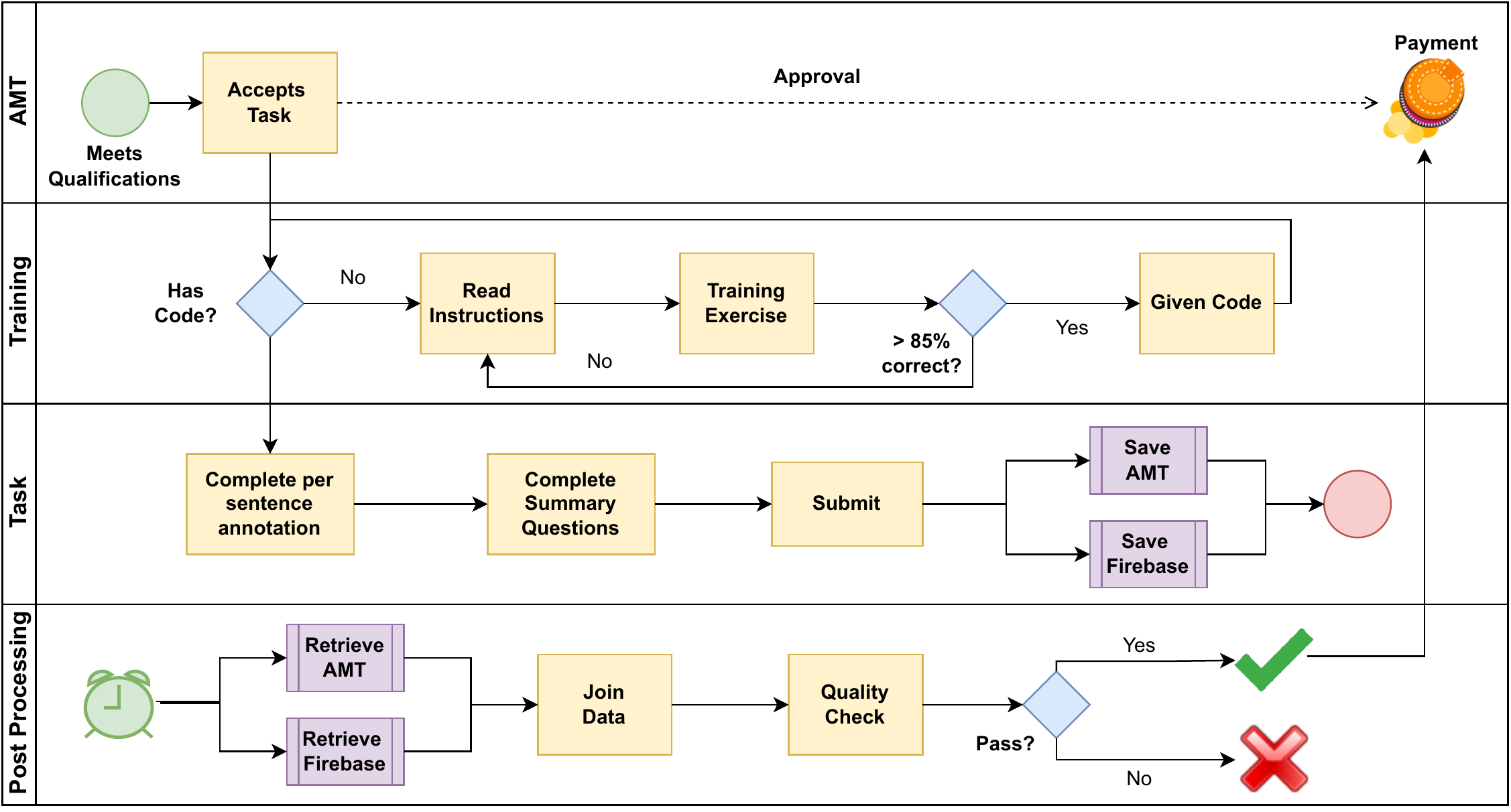}
\caption{Main per task annotator workflow for a AMT Hit for the sentence-by-sentence suspense annotation task.}
 \label{fig:sentannworkflow}
\end{figure}

\begin{figure}[htbp]
\centering
\includegraphics[width=1.0\textwidth]{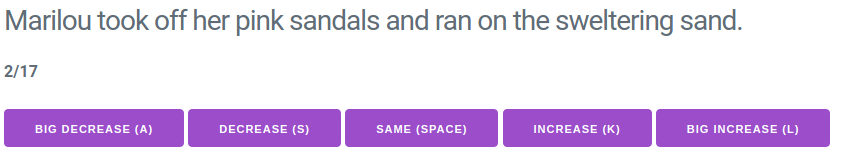}
\caption{An example illustrating the sentence by sentence reader.}
 \label{fig:sentannex1}
\end{figure}

\begin{figure}[htbp]
\centering
\includegraphics[width=1.0\textwidth]{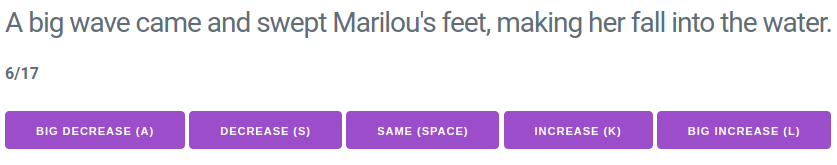}
\caption{A second trainer sentence.}
 \label{fig:sentannex2}
\end{figure}

\begin{figure}[htbp]
\centering
\includegraphics[width=1.0\textwidth]{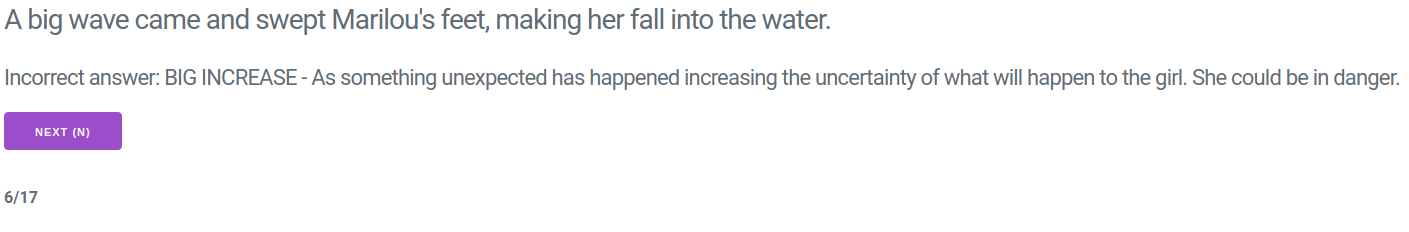}
\caption{The same sentence when the wrong option is selected in training giving feedback as to why another choice should have been made.}
 \label{fig:sentannexfailure}
\end{figure}

\begin{figure}[htbp]
\centering
\includegraphics[width=1.0\textwidth]{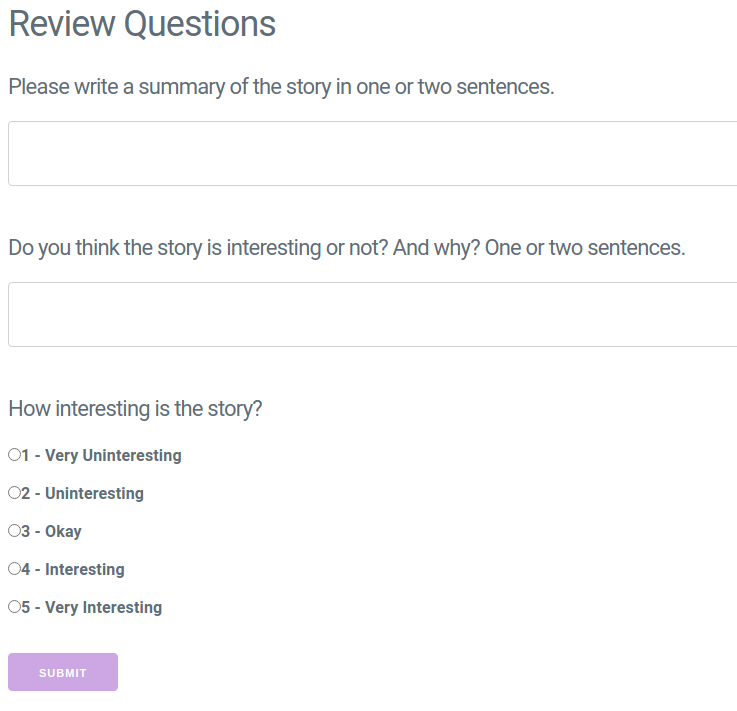}
\caption{The summary questions for the sentence by sentence annotation.}
 \label{fig:sentannexsummary}
\end{figure}

Another change identified by the problems of the whole story annotation is the need to improve training and quality control of annotations to increase inter-annotator agreement. The approach is to have a qualifying training exercise with examples and a test to ensure that only annotators who understand the task well can complete exercises, and similarly more careful quality filtering of submitted tasks to keep the quality high. The overall workflow is in Figure \ref{fig:sentannworkflow}. It should be noted that the workflow shows a single task flow; if a task completed by an annotator is rejected for quality reasons then the task is still in the pool and will be picked up by another annotator until all tasks are completed. Five annotations are collected for each story and $100$ stories from the validation set and $100$ from the test set are annotated. The annotated set is smaller and per sentence judgements make the process slower and hence more expensive. However, as will be reported in following chapters this dataset size is big enough to find strong correlations with high confidence values. The workflow proceeds as follows:

\begin{itemize}
  \item \textbf{AMT:} The workflow that takes place inside Amazon Mechanical Turk, which are the qualifying conditions for being able to accept the task, and then approval or rejection and payment.
	\begin{itemize}
		\item \textbf{Accepts Tasks:} In order to accept the tasks in the first place. AMT annotators must have a $> 98$\% acceptance rate, have completed over $1000$ tasks, and be Master Workers.
		\item \textbf{Payment:} Happens asynchronously following the later offline quality checks.
	\end{itemize}
 \item \textbf{Training:} Because of the requirement for sentence by sentence reading which is not supported directly in AMT, both training and the task are in a separate standalone tasks integrated into AMT.\footnote{The code from the app is at \url{https://github.com/dwlmt/sent-by-sent-ann/}} The training task is the same as the real annotation task, except the annotator gets feedback on errors and needs to pass a threshold to move onto the real task. The aim is to ensure they have read and understood the instructions correctly.
	\begin{itemize}
		\item \textbf{Has Code:} Access to the task is via a secret passcode. The annotator can only get the code by successfully completing training. 
		\item \textbf{Read Instructions:} The annotator is required to read instructions which are given later in this chapter. The guide suspense annotations for a sample story are in Table \ref{tab:guide_annotations}.
		\item \textbf{Training Exercise:} For training, annotators are taken through the story example given in Table \ref{tab:train_annotations} sentence by sentence. For each sentence, they are asked to rate the suspense change; screenshots are in Figure \ref{fig:sentannex1} and Figure \ref{fig:sentannex2}. The layout is the same as the real task, except that in training, if the annotator selects an incorrect category, it tells them what they should have selected and why, as in Figure \ref{fig:sentannexfailure}. While the task is subjective and a few should be questioned at the margins, it focuses the annotators on what is expected in the task.

\item \textbf{Pass or Fail:} At the end of the training story, the annotator is presented with their final score. If it is over 85\% they are given the code to annotate the real story and can use it for subsequent tasks. If they fail, they can return to the start page and retry if they will. The app doesn't restrict this. However, in practice, nearly everyone who didn't pass gave up. 
	\end{itemize}
 \item \textbf{Task:} Again completed in the standalone app that displays one sentence at a time for the reader and asks for a suspense judgement.
\begin{itemize}
	\item \textbf{Complete per sentence annotation:} The annotation is the same as the training exercise, except obviously, there is no feedback on what the answer should be. The UI allows the annotator to quickly read the sentence and make a decision via a mouse click or keyboard shortcut while also seeing how far into the story they are. As well as the recorded suspense score, per sentence time is also recorded. The time is another quality check to identify annotators who may be reading too quickly and, hence, are more likely to be cheating.
	\item \textbf{Complete Summary Questions}: Like the whole story annotation, there are review questions, these are shown in Figure \ref{fig:sentannexsummary}. As well as the same question on how interesting the story is, and a summary for quality control, there is a question asking for a rationale behind why they rated the story as they did. This question is partly about comprehension again and a better understanding of what criteria annotators use when judging a story. After submission, the annotator is returned to AMT and can accept another story as a task if they wish.
\end{itemize}
\item \textbf{Post Processing:} Is processing the resulting data, calculating annotating quality measures, reviewing and either approving or rejecting.
\begin{itemize}
	\item \textbf{Join Data:} This is a simple technical step of joining data held by AMT and the annotations held by the external app in the cloud database Firebase.
	\item \textbf{Quality Check:} Three criteria were used to flag annotators for review post the task agreement for the stories they annotated: If the Kripdendorf's $\alpha$ agreement averaged less than $\alpha < 0.35$ against all the other annotators who annotated the same story. If there were suspiciously low reading times (mean RT $<1$ second per sentence). Or alternatively, if the answer to either the summary question or the rationale questions are poor and don't suggest they comprehended the story or have a good reason for the justification. For all the flagged tasks, all of the annotations done by the annotator were reviewed and either rejected or accepted as a whole on a combination of these factors. For the rejected annotators, they were excluded from completing future tasks. The review was completed regularly during the task, so the annotators could be excluded before completing too many. A few excluded annotators were still paid because they were borderline and probably tried to complete the task as requested but had not understood the task, or had rushed it.
	\item \textbf{Payment:} \$1 was paid per annotator per story, plus the 20\% Master Worker premium, working out at just over the minimum wage per story.
\end{itemize}
\end{itemize}

As described in the annotation process, with the AMT pre-filtering on who the task was accessible to, the training exercise and post-processing filtering considerable attention has been paid to the quality of the annotations. The inter-annotator agreement with Krippendorff $\alpha$
 was $0.52$ and $0.57$ for the
validation and test sets, respectively. The agreement is a big increase on the whole story annotation. The $\alpha$ scale is from $-1$ perfect disagreement through $0$, which is random and to $1$, perfect agreement. So agreement of greater than $0.5$ represents a reasonable agreement especially given the subjective nature of the task. The $\alpha$ inter-annotator agreement matches the exact category chosen by each annotator. 

As has been seen in Figure \ref{fig:sentbysentannotatorplot} the agreement can produce similar suspense arcs, though there are different judgements in individual sentences. Which emphasises why it is the arc of suspense that is important rather than single sentence judgements. Generally, as per this Figure, there is a general upward trajectory in suspense followed by a plateau at the end or a slight fall. This Figure is typical of the type of agreement that is common across the dataset, but it is worth mentioning exceptions. For example, Figure \ref{fig:sentbysentannotatorplotoutlier} shows a plot where the general agreement is good, but there is one outlier. This figure also shows a far flatter story without a pronounced peak. The story is about a murder and the murder happens at the beginning, and the rest of the story is reflections on the murder by the murderer. The curve does seem to reflect the long meandering dilemmas and self-reflection of the murder, rather than the intense peak of action of a typical murder story.

There are also cases where there is widespread disagreement, as in Figure \ref{fig:sentbysentannotatorplotdivergence}. The prompt for this story is -- \textit{Your friends had always pestered you to get an account on that godforsaken app, Tinder.} The story is essentially doubt and fear about going through the steps of signing up for online dating. It is noticeable that, as per this example, where the subject matter is unusual, and the plot doesn't follow a traditional path, agreement and the suspense arc alignment are lower. It is easy to imagine some annotators will find the indecision of deciding which photos to upload to Tinder suspenseful if they have worried about similar things themselves and maybe have relevant online dating experiences, and others deciding that the events are inconsequential and hence suspense doesn't change much. The relevant point is the decision to select randomly entails that there will be more and less clearcut cases. The advantage is it means the model predictions are judged against a much more representative set of stories and not tailored towards suspense which may overestimate performance. However, it does mean that the evaluation shouldn't be thought of as a more conventional NLP task such as Named Entity Recognition. Usually, in NER it is pretty clear whether the entity is a person, corporation or a place, and there isn't much disagreement, so there can be a Gold standard. With suspense, there are naturally disagreements, and therefore, the test for model predictions should be a broader agreement rather than directly aligning with a correct label. The evaluation method against the model is detailed in the next chapter. Still, the broader principle is to align each model prediction arc pairwise with each annotator, rather than the median or consensus, to represent the real divergence of opinion on suspense.

The actual instructions given to the Mechanical Turk Annotators are in Appendix \ref{appendex:annotationinstructions}. The Gold standard guide to the annotation scheme used in the instructions is in Table~\ref{tab:guide_annotations}. Table \ref{tab:train_annotations} contains the text and the correct answers from training; annotators are required to pass with greater than 85\% to continue to the real task.

\begin{table*}[t!]
\centering
\begin{tabularx}{1.0\textwidth}{ |l|X| }
\toprule
\textbf{Annotation} & \textbf{Sentence}                                                                                                                                                                                                                     \\ \midrule
NA                  & Clancy Marguerian, 154, private first class of the 150 + army , sits in his foxhole.                                                                                                                                                  \\
Increase            & Tired cold, wet and hungry, the only thing preventing him from laying down his rifle and walking towards the enemy lines in surrender is the knowledge that however bad he has it here, life as a 50 - 100 POW is surely much worse . \\
Increase            & He's fighting to keep his eyes open and his rifle ready when the mortar shells start landing near him.                                                                                                                                \\
Same                & He hunkers lower.                                                                                                                                                                                                                     \\
Increase            & After a few minutes under the barrage, Marguerian hears hurried footsteps, a grunt, and a thud as a soldier leaps into the foxhole.                                                                                                   \\
Same                & The man's uniform is tan , he must be a 50 - 100 .                                                                                                                                                                                    \\
Big Increase        & The two men snarl and grab at each other , grappling in the small foxhole .                                                                                                                                                            \\
Same                & Abruptly, their faces come together.                                                                                                                                                                                                  \\
Decrease            & ``Clancy?''                                                                                                                                                                                                                           \\
Decrease            & ``Rob?''                                                                                                                                                                                                                              \\
Big Decrease        & Rob Hall, 97, Corporal in the 50 - 100 army grins, as the situation turns from life or death struggle, to a meeting of two college friends.                                                                                           \\
Decrease            & He lets go of Marguerian's collar.                                                                                                                                                                                                    \\
Same                & `` Holy shit Clancy , you're the last person I expected to see here ''                                                                                                                                                                \\
Same                & `` Yeah '' `` Shit man , I didn't think I'd ever see Mr. volunteers every saturday morning at the food shelf' , not after The Reorganization at least ''                                                                              \\
Same                & ``Yeah Rob , it is something isn't it ''                                                                                                                                                                                              \\
Decrease            & `` Man , I'm sorry, I tried to kill you there''.                                                                                                                                                                             \\ \bottomrule
\end{tabularx}
\caption{The instruction example given to MTurk Workers trying to get a qualification. The annotation labels are the recommended labels. This is an extract from a validation set WritingPrompts story.}

\label{tab:guide_annotations}
\end{table*}

\begin{table*}[t!]
\centering
\begin{tabularx}{1.0\textwidth}{ |l|X| }
\toprule
\textbf{Annotation} & \textbf{Sentence}                                                                                                                                                                                                                     \\ \midrule
NA  &   Ah, The Beach! \\ 
Increase & Marilou took off her pink sandals and ran on the sweltering sand. \\
Increase & She tiptoed into the water, giggling as a wave washed her legs. \\
Same & She looked back at the big red and white umbrella and waved at her mother, who was getting a much needed suntan. \\
Same & Her mother smiled and waved with two fingers, holding a peach in one hand and a bottle of ice cold lemonade in the other. \\
Big Increase & A big wave came and swept Marilou's feet, making her fall into the water. \\
Big Increase & She tried to scream but swallowed a gulp of salty water instead. \\
Increase & She didn't know which way was up and which was down. \\
Increase & Her sweeping hand touched something with a tiny claw in the sand. She screamed bubbles and tried hard to swim. \\
Decrease & The wave receded, dumping the startled girl upon the soft, wet beach. \\
Decrease & Marilou coughed and rubbed her eyes, struggling to get back up. \\
Same & Grains of sand in her blue swimsuit scratched her skin. \\
Increase & She looked back toward her mother, who was standing up with a worried frown partly covered by sunglasses. \\
Decrease & Marilou ran back to the big umbrella, tiptoeing through the hot sand. \\
Same & She smelled of salt and seaweeds. \\
Decrease & She took her mother's big, warm hand with her cold, wet one. \\
Big Decrease & Come play with me in the waves, Mommy! This is the bestest vacation ever! \\
\bottomrule
\end{tabularx}
\caption{This is the training example that MTurk workers must get $85$\% on to receive a code to be able to complete annotations on real stories. Both this as the previous instruction example are relatively straightforward to make it clear what the annotation task is.}

\label{tab:train_annotations}
\end{table*}

\chapter{Hierarchical Suspense Rollout}

\label{chap:rolloutsuspense}

\section{Introduction}

Chapter \ref{chap:backgroundtheory} introduced background theory on suspense and surprise. The thesis aims to implement models of suspense and surprise from  \citet{ely2015suspense} and from \citet{Hale:01, hale2006uncertainty}. The primary task is inferring suspense on the annotated dataset from Chapter \ref{chap:suspenseannotation}, and a second related task is introduced from \citet{Papalampidi2019MoviePA} to infer the turning points of Movie summaries.

As reviewed in the background material in Chapter \ref{chap:backgroundtheory}, the crucial benefit of both Ely's and Hale's models is they depend only on the probability of a given state and outcome states. To recap, the general approach is to reproduce a state with sentence embeddings in the vector space, and probability can be a distribution over possible continuations. The advantage is that both can be inferred from hierarchical unsupervised models that can benefit from recent developments in deep learning without requiring an expensive annotated dataset. The next section will introduce the implemented theories in more detail. The subsequent section will elaborate on the unsupervised models for inferring suspense and surprise. It is followed by a method to infer suspense reported in this chapter by generating a tree of alternatives with a language model. Chapter \ref{chap:tdvaesuspense} presents an alternative model that models suspense directly in the latent vector space with variational methods.

\section{Suspense}

In order to formalise measures of suspense, it is assumed that a story
consists of a sequence of sentences. These sentences are processed one
by one, and the sentence at the current timepoint $t$ is represented
by an embedding $e_t$. Each embedding is associated with a
probability $P(e_t)$. Continuations of the story are represented by a
set of possible next sentences, whose embeddings are denoted
by~$e^i_{t+1}$.

The first measure considered is surprise \citep{Hale:01},
which in the psycholinguistic literature has been successfully used to
predict word-based processing effort \citep{Demberg:Keller:08a,
  roark-etal-2009-deriving, DBLP:journals/corr/abs-1810-11481,
  DBLP:conf/cogsci/SchijndelL18}. Surprise is a backward-looking
predictor: it measures how unexpected the current word is given the
words that preceded it (i.e.,~the left context).  Hale formalises
surprise as the negative log of the conditional probability of the
current word. For stories, surprise is computed over sentences. Intuitively, there is more surprise if the sentence is less predictable. As our
sentence embeddings $e_t$ include information about the left context
$e_1,\dots,e_{t-1}$, \emph{Hale surprise} which can be written as in eqn. \ref{eqn:con_surprise}:
\begin{myequation}
\begin{aligned}
S^\text{Hale}_t = -\log P(e_t)
\label{eqn:con_surprise}
\end{aligned}
\end{myequation}
An alternative measure for predicting word-by-word processing effort
used in psycholinguistics is entropy reduction
\citep{hale2006uncertainty}. This measure is forward-looking: it
captures how much the current word changes our expectations about the
words that will be encountered next (i.e.,~the right context). Entropy is computed at the story level, i.e., over sentences instead of
over words. Given a probability distribution over possible next
sentences $P(e^i_{t+1})$, the model calculates the entropy of that
distribution. Entropy reduction is the change of that entropy from one
sentence to the next in eqn. \ref{eqn:ent_surprise}:
\begin{myequation}
\begin{aligned}
H_t = - \sum_{i}{P(e^i_{t+1}) \log P(e^i_{t+1})} \\ 
U^\text{Hale}_t = H_{t-1} - H_t
\label{eqn:ent_surprise}
\end{aligned}
\end{myequation}
Note that the model follows \citet{frank2013uncertainty} in computing entropy
over surface strings, rather than over parse states as in Hale's
original formulation. Hale entropy reduction is based entirely on uncertainty. The relevance for suspense is that the measure is essentially how much information is conveyed by a particular sentence for predicting subsequent ones. As such, the hypothesis with this metric is that more informant sentences are more suspenseful. It overlaps with the Barthes' Cardinal Functions salience detection method outlined in Chapter \ref{chap:salience}.

As has been proposed in \citet{ely2015suspense}, this definition can be extended from uncertainty to include outcome. Representing the state at time $t$ as
$e_t$, \emph{Ely surprise} is defined as:

\begin{myequation}
\begin{aligned}
S_t^\text{Ely} = (e_{t} - e_{t-1})^2
\label{eqn:ely_surp}
\end{aligned}
\end{myequation}
Ely et al.'s approach can be adapted for modelling suspense in stories
if when it is that each sentence changes the state (the
characters, places, events in a story,~etc.). States $e_t$ then become
sentence embeddings, rather than beliefs in end states, and Ely
surprise is the distance between the current embedding $e_t$ and the
previous embedding~$e_{t-1}$. This chapter reports L2 and Cosine distance; like \cite{DBLP:conf/cogsci/LiBG18} we experiment
with information gain and KL divergence (and more obscure distances such as Earth Mover/Wasserstein distance) but found worse performance.\footnote{The main thing is that in principle any vector distance can be used.} Just like Hale surprise, Ely surprise
models backwards-looking prediction, but over representations, rather
than over probabilities. $S_t^\text{Ely}$ and $U_t^\text{Ely}$ can be thought of as the converse of $S^\text{Hale}_t$ and  $U^\text{Hale}_t$. In the reviewed background in Chapter \ref{chap:backgroundtheory} it was discussed that in the suspense and surprise literature, uncertainty and consequential outcomes were the two hypothesised components. $S^\text{Hale}_t$ and  $U^\text{Hale}_t$ model only uncertainty and $S^\text{Hale}_t$ and  $U^\text{Hale}_t$ outcome. For  surprise, $S^\text{Hale}_t$, the metric essentially measures how big shifts are between the expected context state and what really happens in the story.

Ely et al. also introduced a measure of forward-looking prediction,
which they define as the expected difference between the current
state $e_t$ and the next state~$e_{t+1}$, in eqn. \ref{eqn:ely_susp}:
\begin{myequation}
\begin{aligned}
U^\text{Ely}_t = \mathop{\mathbf{E}}[(e_t - e^i_{t+1})^2] \\
=  \sum_{i} P(e^i_{t+1})(e_t - e^i_{t+1})^2
\label{eqn:ely_susp}
\end{aligned}
\end{myequation}
This is closely related to Hale entropy reduction, but again the
entropy is computed over states (sentence embeddings in this case),
rather than over probability distributions. Intuitively, this measure
captures how much the uncertainty about the rest of the story is
reduced by the current sentence. These are forward-looking
measures in Equations~(\ref{eqn:ent_surprise})
and~ define as \textit{Hale} and \textit{Ely} uncertainty reduction
respectively. Crucially, the measure encapsulates both uncertainties over the possible future paths and how much they diverge from the current context state.

Ely et al. also suggest versions of their measures in which each state
is weighted by a value $\alpha_t$, thus accounting for the fact that
some states may be more inherently suspenseful than others, in eqn. \ref{eqn:ely_surp_ext}:
\begin{myequation}
\begin{aligned}
S_{t}^{\alpha\text{Ely}} = \alpha_t(e_{t} - e_{t-1})^2\\
U^{\alpha\text{Ely}}_t = \mathop{\mathbf{E}}[\alpha_{t+1}(e_t - e^i_{t+1})^2]
\end{aligned}
\label{eqn:ely_surp_ext}
\end{myequation}
With $\alpha$ sentences with high emotional valence are more
suspenseful, as emotional involvement heightens reader's experience of
suspense. This can be captured in Ely et al.'s framework by assigning
the $\alpha$s the scores of a sentiment classifier. This is implemented in the \textit{Impact} models defined later. Ely et al. also specify a $\beta$ extension variable for suspense. $\beta$ is an adjustment factor in temporal position in the game, story, sport, etc. The intuition behind this is the reader could become more emotionally invested as a story continues and hence suspense will increase. This is not part of the thesis models but is mentioned as it is discussed later. The main issue for excluding it is the parameter could only realistically be trained via a supervised approach and this would induce a bias that could easily exaggerate performance against a small testset of annotations.

This section has defined a version of suspense and surprise based on Ely. Different measures encompass different views from the paradox of suspense debate and align with recent cognitive studies and story generation systems.  The challenge of applying the Ely model is there needs to be a way of representing the space of possible outcomes. There are two approaches: The first is using a language model to generate alternative continuations and then use vector embeddings as a proxy for differences in the outcome. One difference from the original formulation of Ely et al. is a Bayesian final outcome state. The hypothesis behind the hierarchical rollout model in this chapter is that the sentence embedding states can act as an approximation. Unlike a tennis match where either A or B will win there are infinite final end states in the story. The latent space of the sentence embedding serves as a stand-in and so the further the latent representation diverges with plausible continuations, then the more suspense or surprise there is. As reviewed in Chapter \ref{chap:backgroundml}, the latent spaces can represent a range of linguistic and semantic features. The intention is thus that they should represent a meaningful proxy.

The other model is to directly use a temporal VAE model to sample directly over a latent space for continuations. The second model is covered in Chapter \ref{chap:tdvaesuspense}. The practical model work uses an unsupervised approach based on recent deep learning models to infer suspense and surprise. The advantage is that it is both more principled and avoids the enormous cost and time required to annotate a suitable corpus to train a supervised model.

In tying up this section, I will connect the Ely model with the examples, starting with the simplest in \textit{Mr Bean} and the \textit{Only Fools and Horses} bar sketch. In the \textit{Only Fools and Horses} sketch, we expect the two characters will continue talking at a bar. Instead, the slapstick comedy fall is unexpected. This corresponds with the Ely surprise idea of surprise, where surprise is a difference of state between what is expected and what happens; the fall is a complete shift from what the viewer expects carrying on the conversation. \textit{Mr Bean} models suspense in that the strange behaviour, such as washing lettuce, creates a variance in expectations about what might happen for the viewer. This encompasses the divergence of expectations about what might happen that fits with Ely suspense. One problem with a belief over expectations is, in theory, anything might happen. The space of possibilities is infinite. In the \textit{Mr Bean} example, it is the odd behaviour that asks the viewer to imagine what might happen, which creates comedic suspense. It goes back to the discussion of a \textit{reader as a problem solver} or \textit{eventfulness} in the reader placing themselves in the author's shoes; it is the cognitive imagination that is important at the moment and not infinite possibilities. The \textit{Dramatis} \citep{DBLP:conf/aaai/ONeillR14} story generation paper makes a similar point that it is the best heuristic plan that is important in suspense and not all possible plans. The aim of the model in this chapter or the variational one is to be able to sample or generate a representative set of reasonable continuations given the context, and not exhaustively calculate all possibilities. The aim of the model is primarily one of inferring suspense rather than cognitive plausibility. However given the memory and processing limitations of human cognition, if the Ely models holds cognitively, then the most prominent continuations in the reader's mind are likely the most influential in feelings of suspense.

\section{Method}

This section shows how to compute the surprise and suspense (uncertainty
reduction) measures introduced in the previous section. This involves
building a model that processes stories sentence by sentence, and
assigns each sentence an embedding that encodes the sentence and its
preceding context, as well as a probability. These outputs can then be
used to compute a surprise and suspense values for the sentence.

Furthermore, the model needs to be able to generate a set of possible
next sentences (story continuations), each with an embedding and a
probability. Generating upcoming sentences is potentially very
computationally expensive since the number of continuations grows
exponentially with the number of future time steps. As an
alternative, the model can therefore sample possible next sentences from a
corpus and use the model to assign them embeddings and
probabilities. Both of these approaches will produce sets of upcoming
sentences, which can compute uncertainty
reduction. So far only next sentences have been discussed,
but uncertainty reduction can be computed using
longer rollouts of several sentences, and constructing a tree of generated continuations.

Embeddings implicitly encode information about the situation
 \citep{conneau-etal-2018-cram,DBLP:conf/nips/LevyG14}. The
 advantages: It is theoretically guided by literary and cognitive
 theory. Success of language models such as ELMO
 \citep{peters-etal-2018-deep} and BERT \citep{devlin-etal-2019-bert}
 have shown that highly powerful models can be trained
 auto-regressively, which eliminates much of the bottleneck on available
 training data, and also lends itself to cheaply to broader application
 in a way a supervised model does not.
 
 The overall approach leverages contextualised language models, which
are a powerful tool in NLP when pretrained on large amounts of text
and fine-tuned on a specific task \citep{peters-etal-2018-deep,
  devlin-etal-2019-bert}. Specifically, the model employs Generative Pre-Training
(GPT, \citealp{radford2018improving}), a model which has proved
successful in generation tasks \citep{radford2019language,See2019DoMP}.

\subsection{Hierarchical Model} 

Previous work found that hierarchical models show strong performance
in story generation \citep{fan-etal-2018-hierarchical} and
understanding tasks \citep{cai-etal-2017-pay}. The language model and
hierarchical encoders are uni-directional, which matches the
incremental way in which human readers process stories when they
experience suspense.

\begin{figure}[t]
\centering
\includegraphics[width=1.0\textwidth]{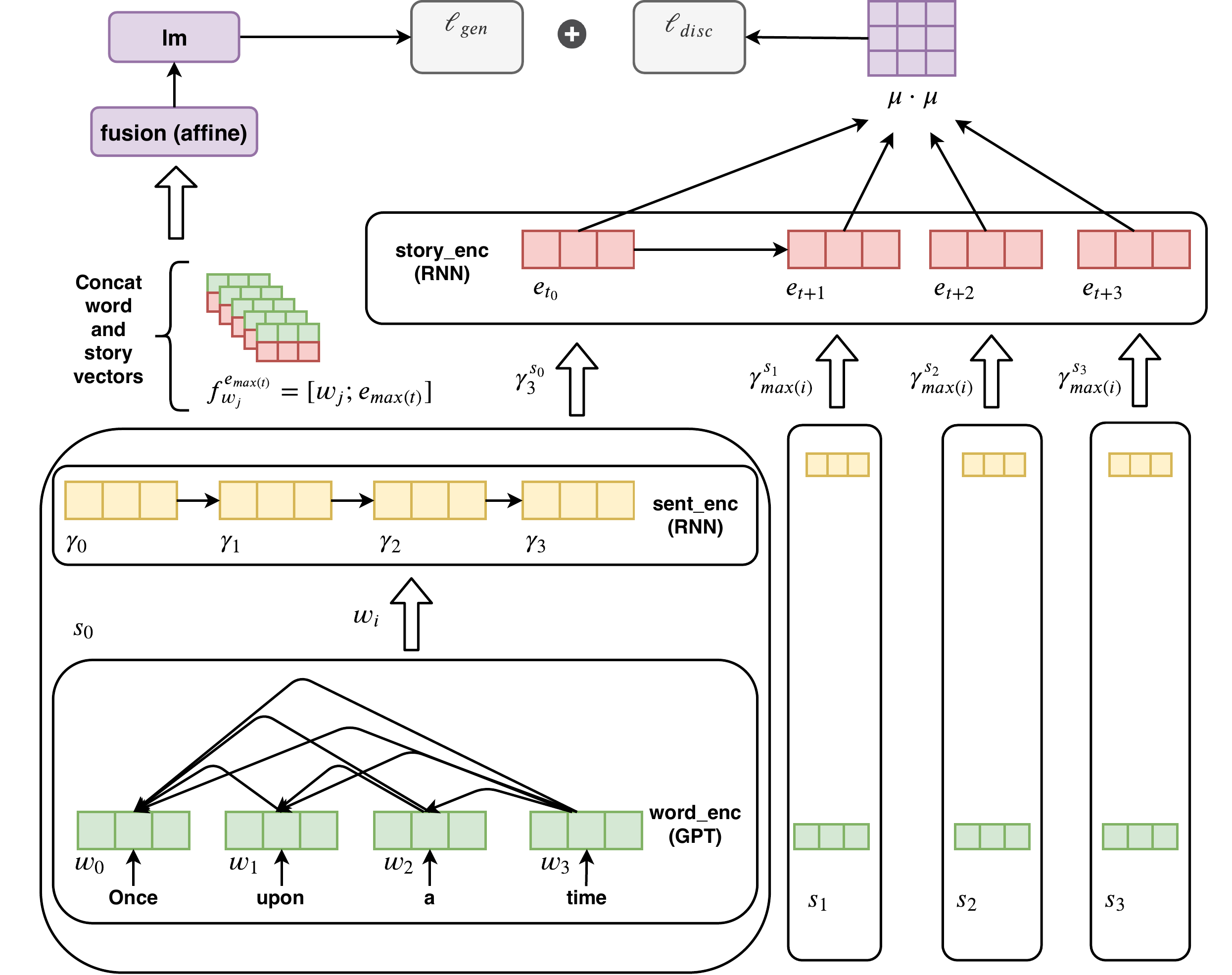}
\caption{Architecture of the hierarchical rollout model. The \emph{word\_enc} is a GPT model for generation, \emph{sent\_enc} is the middle tier sentence encoder,
  and \emph{story\_enc} is an autoregressive RNN for modelling the story context state.}
 \label{fig:rollout_suspense_arch}
\end{figure}

Figure~\ref{fig:rollout_suspense_arch} depicts the architecture of our
hierarchical model.\footnote{Model code and scripts for evaluation are
  available at
  \url{https://github.com/dwlmt/Story-Untangling/tree/acl-2020-dec-submission}}
It builds a chain of representations that anticipates what will come
next in a story, allowing us to infer measures of suspense. For a
given sentence, GPT is the word encoder (\emph{word\_enc} in
Figure~\ref{fig:rollout_suspense_arch}) which turns each word in a sentence
into a word embedding~$w_i$. An RNN (\emph{sent\_enc}) converts the word embeddings of the sentences into a sentence
embedding~$\gamma_i$. Each sentence is represented by the hidden state
of its last word, which is then fed into a second RNN
(\emph{story\_enc}) that computes a story embedding. The overall story
representation is the hidden state of its last sentence. Crucially,
this model also gives us $e_t$, a contextualised representation of the
current sentence at point~$t$ in the story, to compute surprise and
uncertainty reduction.

The intuition behind the architecture is that the model needs: Capability to generate plausible continuations. At the sentence level, there must be a semantically relevant story state. At the top level, the model needs to discriminate probabilities over possible continuations whether sampled from the corpus or generated. Each of the layers in the models carries out one of these functions. The base GPT model can generate continuations for a given story conditioned on the story's context so far. The sentence encoder is trained to compress the whole sentence into a vector to represent the particular sentence. The top-level RNN compresses the entire state of the story up to that point into a single vector. The probability of each continuation is calculated by the cosine similarity or dot product of the existing story context state against the hypothetical ones with a softmax. The rationale is that the top-level RNN will learn to retain a state that maximises similarity with the state update informed by hypothetical continuations. This effectively makes the model an autoregressive model since the RNN is trained to predict the changes that the $t+1$ sentence will make. Keeping the probability calculation as a function of two vectors (e.g. dot product) also means all the calculations are in the vector space. An alternative would have been to encode the dot product maximisation against the sentence representation  $s_{t+1}$ rather than $e_{t+1}$; in practice, this didn't work as well against either the suspense or movie turning points evaluation and is not reported. An alternative that was briefly trialled is to have a direct discriminator that directly takes either the $s_{t+1}$ or $s_{t+1}$ state and then either takes a fixed $n$ number of possible continuations or scores a likelihood story between $1.0$ and $0.0$ for a single continuation. Neither approach looked promising from initial trials but could be explored in future work.

\subsection{Loss Functions} 

To obtain the discriminatory loss $\ell_{\text{disc}}$ for a particular
sentence~$s$ in a batch, the model computes the dot product of all the story embeddings $e$ in the batch, and then takes the
cross-entropy across the batch with the correct next sentence, in eqn. \ref{eqn:quickthoughts_loss_hier}:\footnote{Cosine similarity and minimising L2 distance was also tried but performed worse in training.}
\begin{myequation}
\begin{aligned}
\ell_{\text{disc}}(e_{t+1}^{i=s}) = -\log \frac{\exp(e_{t+1}^{i=s} \cdot e_{t})}{\sum_i \exp(e_{t+1}^{i}  \cdot e_{t})}
\label{eqn:quickthoughts_loss_hier}
\end{aligned}
\end{myequation}
Modelled on Quick Thoughts \citep{DBLP:conf/iclr/LogeswaranL18}, this forces the
model to maximise the dot product of the correct next sentence versus
other sentences in the same story, and negative examples from other
stories, and so encourages representations that anticipate what
happens next.

Model training includes a conditional LM loss $\ell_{\text{gen}}$ to improve the quality of the sentences generated by the model. The model concatenates the word representations $w_j$ for all word embeddings in the latest
sentence with the latest story embedding $e_{\max(t)}$. This is run
through affine ELU \citep{DBLP:journals/corr/ClevertUH15} layers to produce enriched word embedding
representations, analogous to the Deep Fusion model
\citep{DBLP:journals/corr/GulcehreFXCBLBS15}, with story state instead
of a translation model. It is a late fusion model, fusing together the sentence state with the GPT before the decoder that predicts the output wordpiece tokens. The related Cold Fusion \citep{DBLP:conf/interspeech/SriramJSC18} approach
 proved inferior. The generative loss in eqn. ~(\ref{eqn:gen_loss}) is a standard LM
loss, where $w_j$ is the GPT word embeddings from the sentence and
$e_{\max(t)}$ is the story context that each word is concatenated
with:
\begin{myequation}
\ell_{\text{gen}} = - \sum_{j} \log P({w_j | w_{j-1}, w_{j-2}, \dots; e_{\max(t)})}
\label{eqn:gen_loss}
\end{myequation}
The overall loss is $\ell_{\text{disc}} + \ell_{\text{gen}}$.

\subsection{Inference} 

From the model measures of surprise and uncertainty reduction
introduced in Section~\ref{sec:def} are computed using the output of the story
encoder \emph{story\_enc}. In addition to the contextualised sentence
embeddings $e_t$, this requires their probabilities $P(e_t)$, and a
distribution over alternative continuations~$P(e^i_{t+1})$.

The inference implements a recursive beam search over a tree of future sentences
in the story, looking between one and three sentences ahead
(rollout). The probability is calculated using the same method as the
discriminatory loss, but with the cosine similarity rather than the dot product
of the embeddings $e_t$ and $e^i_{t+1}$ fed into a softmax
function. Cosine outperformed dot product on inference. The reason is that dot product seems to concentrate the probability distribution on a small number of the most likely continuations whereas cosine is more distributed; a broader distribution means less likely sampled continuations are weighted more highly in the suspense judgement.

 In corpus sampling random sentences $n$ are sampled from the corpus and used as continuations. In the more sophisticated generation, sentence are generated using top-k sampling ($k=50$ for trials) using the GPT LM and the approach used by \citet{radford2019language}. Top-k sampling \citep{holtzman-etal-2018-learning} is randomly sampling each next wordpiece token from only the most probable $k$ rather than a softmax over all possible wordpieces. The intuition behind it is it concentrates the probability mass on the most likely continuations, and so the text is much less likely to degenerate, and is thus more likely to be coherent. Top-k sampling can produce more diverse and better results than a complicated decoder beam search decoders \citep{See2019DoMP}. The sampling decoder includes up to the last $300$ word tokens of context, and because of the fusion with the story context all of these tokens are enriched with information for the $\mu$ sentence vectors from the respective point in the story.

\subsection{Training Details}

\textit{WritingPrompts} \citep{fan-etal-2018-hierarchical} is the dataset used to train the model. \textit{WritingPrompts} is a
corpus of circa 300k short stories from the \textit{/r/WritingPrompts}
subreddit. These stories were created as an exercise in creative
writing, resulting in stories that are interesting, natural, and of
suitable length. The original split of the data into 90\% train, 10\%
validations, and 10\% test sets was used. All the suspense annotated stories are the validation and test sets so it means the model is not trained on any stories used for suspense evaluation. The same preprocessing steps were applied to all the stories in the \textit{WritingPrompts} corpus as detailed in Chapter \ref{chap:suspenseannotation} on the suspense annotations. As discussed in Chapter \ref{chap:introduction} and in the previous chapter \textit{WritingPrompts} has a particular slant. Therefore models such as this chapter trained in the corpus will learn from the particularities of the corpus. Nevertheless, the secondary study on movie turning points shows that while this may limit overall transferability, the model can still apply moderately well to more conventional movie plots.

The training used SGD with Nesterov momentum
\citep{sutskever2013importance} with a learning rate of $0.01$ and a
momentum of $0.9$. Models were run with early stopping based on the
mean of the accuracies of training tasks. For each batch, $50$ sentence blocks from two
different stories were chosen to ensure that the negative examples in
the discriminatory loss include easy (other stories) and difficult
(same story) sentences.

\begin{table}[tb]
\centering
\begin{tabular}{@{}lllll@{}}
\toprule
                   & \textbf{GRU} & \textbf{LSTM}  \\ \midrule
Loss      & 5.84       & 5.90      \\
Discriminatory  Acc.  & 0.55       & 0.54        \\
Discriminatory Acc. $k=10$ & 0.68       & 0.68         \\
Generative Acc.    & 0.37       & 0.46        \\
Generative Acc. $k=10$  & 0.85       & 0.85       \\
Cosine Similarity       & 0.48       & 0.50       \\
L2 Distance        & 1.73       & 1.59    \\
Number of Epochs    & 4          & 2      \\ \bottomrule
\end{tabular}
\caption{For accuracy the baseline probability is 1 in 99; $k=10$ is
  the accuracy of the top 10 sentences of the batch. From the best
  epoch of training on the WritingPrompts development set.}
\label{tab:tech_results}
\end{table}

The model used the existing pretrained GPT weights but fine-tuned the encoder and
decoder weights on our task. For the RNN components of our
hierarchical model: Models are trained and evaluated on with both GRU
\citep{chung2015gated} and LSTM
\citep{10.1162/neco.1997.9.8.1735} variants. The GRU model had
two layers in both \emph{sen\_enc} and \emph{story\_enc}; the LSTM
model had four layers each in \emph{sen\_enc} and
\emph{story\_enc}.\footnote{Four layer versions of the GRU are trained but not reported as results are similar to the two layer version, and the two-layer version better represents a smaller model.} Both had two fusion layers and the size of the
hidden layers for both model variants was $768$. Results are presented
of both variants on the tasks of sentence generation and sentence
discrimination in Table~\ref{tab:tech_results}. Both perform
similarly, with slightly worse loss for the LSTM variant, but faster
training and better generation accuracy. The LSTM variant picks out the correct sentence 54\% of the
time and generates it 46\% of the time, GRU is similar.\footnote{The discriminator accuracy is just if the real next sentence in the batch has the highest probability. The generation accuracy is judged on whether the average perplexity is lower for the following sentence than the others in the batch.} There are typically $100$ sentences in a batch from two stories and so picking the correct sentence the majority of the time represents strong performance.

At inference time, a set of story continuations are produced either by random sampling or by generation. Random sampling means that $n$
sentences were selected from the corpus and used as continuations. Random sampling is used since sentences in the same story are already in the context and cannot be used. It is also a simpler baseline than generating alternatives. For
generation, sentences were generated using top-$k$ sampling (with $k=50$) using the GPT language model and the approach of
\citet{radford2019language}, which generates better output than beam search \citep{holtzman-etal-2018-learning} and can outperform a decoder \citep{See2019DoMP}. For generation, up to $300$ words were conditioned on  as context, enriched with the story sentence embeddings from the corresponding points in the story. For rollouts of one sentence, we generated $100$ possibilities at each step; for rollouts of two, $50$ possibilities and rollouts of three, $25$ possibilities. This keeps what is an expensive inference process manageable.

\subsection{Importance} 

Variants of surprise and suspense follow from Ely et al. in evaluating weighted versions $S_t^{\alpha\text{Ely}}$ and
$U_t^{\alpha\text{Ely}}$ (see eqn.~(\ref{eqn:ely_surp_ext})). $\alpha_t$ is calculated by taking the sentiment scores assigned
by the VADER sentiment classifier \citep{Hutto_Gilbert_2014} to each
sentence and multiplying them by $1.0$ for positive sentiment and
$2.0$ for negative sentiment. VADER was chosen as it is fast to run on the CPU and so can easily be applied to large batches of generated sentences necessitated by the inference process. The stronger negative weighting reflects the observation that negative consequences can be more important than positive ones in suspense \citep{o2013computational}, or from prospect theory \citep{kahneman2013prospect}. Other weightings such as equal weighting were tried but performed slightly worse and so aren't reported. The intuition behind the Ely $\alpha$ adjusted definitions is that divergence of the state is one factor in suspense, but the state's importance magnifies or diminishes. For example, it might be that in a thriller such as \textit{Gone Girl} scenes related to the murder are more suspenseful. This is partly because there is uncertainty and divergence of the outcome, but it's also because more is at stake in the situation. The same could be said for the fight scenes in \textit{Rocky}, or the romantic moments in \textit{Pretty Woman}. The main model only captures the divergence and relies on the implicit representation of the sentences for the state. The $\alpha$ adjusted version attempts to approximate this through a simple sentiment metric. It will magnify the suspense or surprise where there is strong sentiment. The hypothesis is that stronger absolute sentiment is more likely to correlate with dramatic events and hence more suspenseful parts of the story; the scaling effect will thus improve the predictive power of the Ely model.

Ely also defines a secondary $\beta$ extension, which is position-dependent. The intuition behind it is that the position in the sequence is important to suspense, which is particularly applicable to story plots. The problem is that there is no straightforward way to learn an appropriate position weighting unsupervised. Experimenting with a trained latent weight on position, especially with a small training dataset, risks fitting the model to the data and exaggerating performance. As such, the $\beta$ extension is left for future work.

\subsection{Baselines} 

A number of baselines are evaluated as alternatives to surprise and
uncertainty reduction derived from our hierarchical model. These
baselines also reflect how much change occurs from one sentence to the
next in a story: WordOverlap is the Jaccard \citep{jaccard} similarity between the two
sentences, GloveSim is the cosine similarity between the averaged
Glove \citep{pennington-etal-2014-glove} word embeddings of the two
sentences and GPTSim is the cosine similarity between the GPT
embeddings of the two sentences. The rationale for the baselines is they are simpler approximations for how much a story changes from sentence to sentence. If there were a simple heuristic pattern such as sentence embeddings becoming further apart at certain points in the story or in more suspenseful passages, the baselines would be expected to pick up this pattern. The difference between GPTSim and GloveSim, with Ely surprise inferred with the LSTM model, is that the baselines are modelling only sentence to sentence changes, whereas Ely surprise is the difference taking into account all earlier context compressed into embeddings. As such the Ely version is more global. The $\alpha$ baseline is the weighted VADER sentiment score; there is a possibility that absolute sentiment could correspond with suspense as it might be thought more emotional events, positive or negative, are suspenseful.

\section{Experiments}

\subsection{Narrative Suspense}

The following narrative suspense evaluation is based on the annotated dataset described in Chapter \ref{chap:suspenseannotation}. The annotator judgements are relative (amount of decrease/increase in
suspense from sentence to sentence), but the model predictions are
absolute values. The model predictions could be put into discrete categories, but this would fail to capture the overall
arc of the story. Instead, annotations convert the relative judgements into
absolute suspense values, where $J_t = j_1 + \dots + j_t$ is the
absolute value for sentence $t$ and $j_1, \dots, j_t$ are the relative
judgements for sentences~1 to $t$. $-0.2$ for Big Decrease,
$-0.1$ for Decrease, $0$ for Same, $0.1$ for Increase, and $0.2$ for
Big Increase.\footnote{These values were fitted with predictions (or
  cross-worker annotation) using 5-fold cross validation and an L1
  loss to optimise the mapping. A constraint is placed so that Same is
  $0$, increases are positive and decreases are negative with a
  minimum $0.05$ distance between.} Both the absolute suspense
judgements and the model predictions are normalised by converting them
to $z$-scores.

\begin{figure}[tb]
\includegraphics[trim={0.5cm 0.5cm 2.5cm 2.5cm},clip,width=1.0\textwidth]{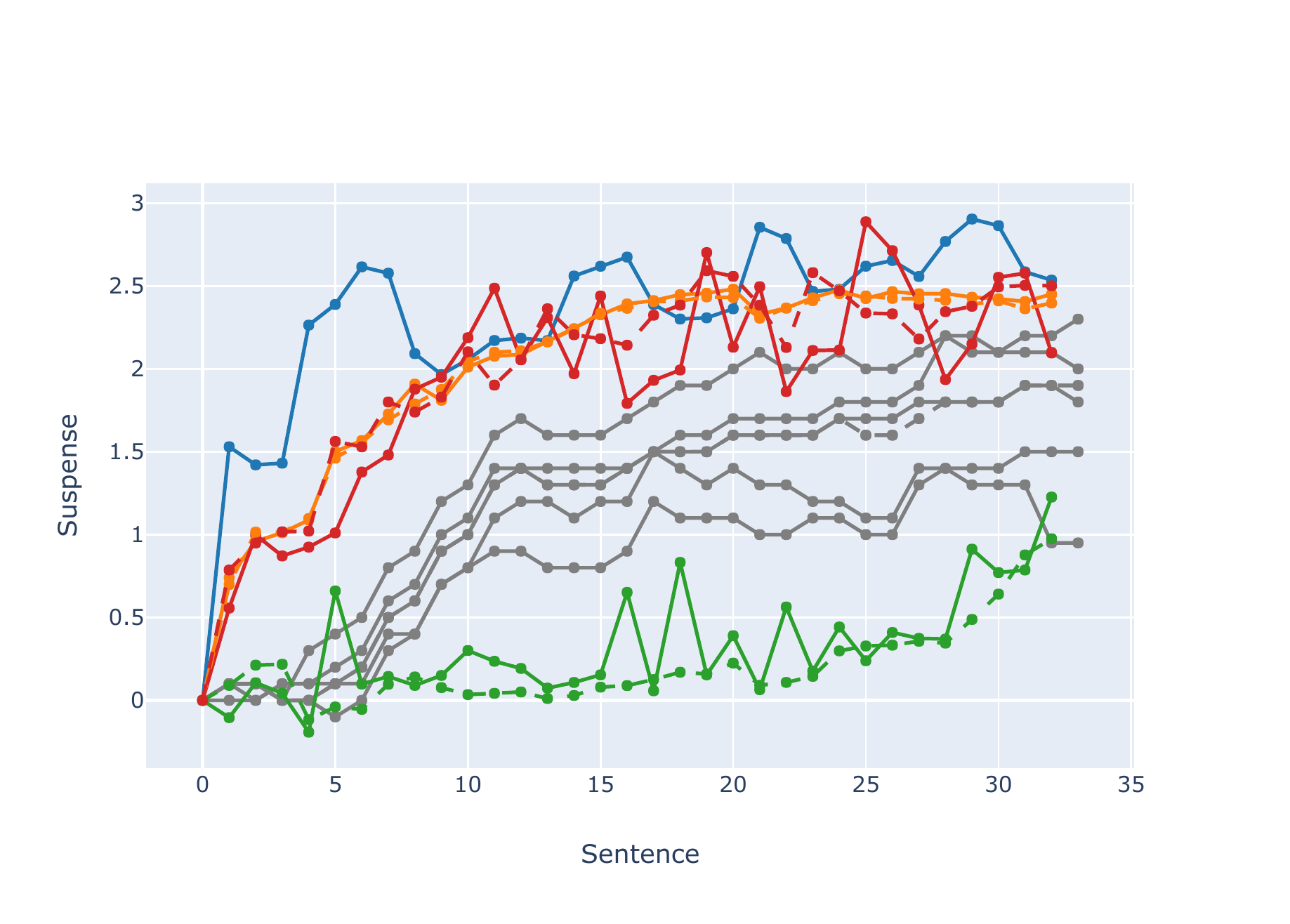}
\centering
\caption{Story 27,
\textbf{\textcolor{annotatedcolor}{Human}},
\textbf{\textcolor{surpriseentropycolor}{$S^{\text{Hale}}$}},
\textbf{\textcolor{surprisecolor}{$S^{\text{Ely}}$}},
\textbf{\textcolor{suspensecolor}{$U^{\text{Ely}}$}},
\textbf{\textcolor{suspensestatecolor}{$U^{\alpha\text{Ely}}$.}} Solid
lines: generated alternative continuations, dashed lines: sampled
alternative continuations.}
\label{fig:wp_27_main}
\end{figure}

\begin{figure}[tb]
\includegraphics[trim={0.5cm 0.5cm 2.5cm 2.5cm},clip,width=1.0\textwidth]{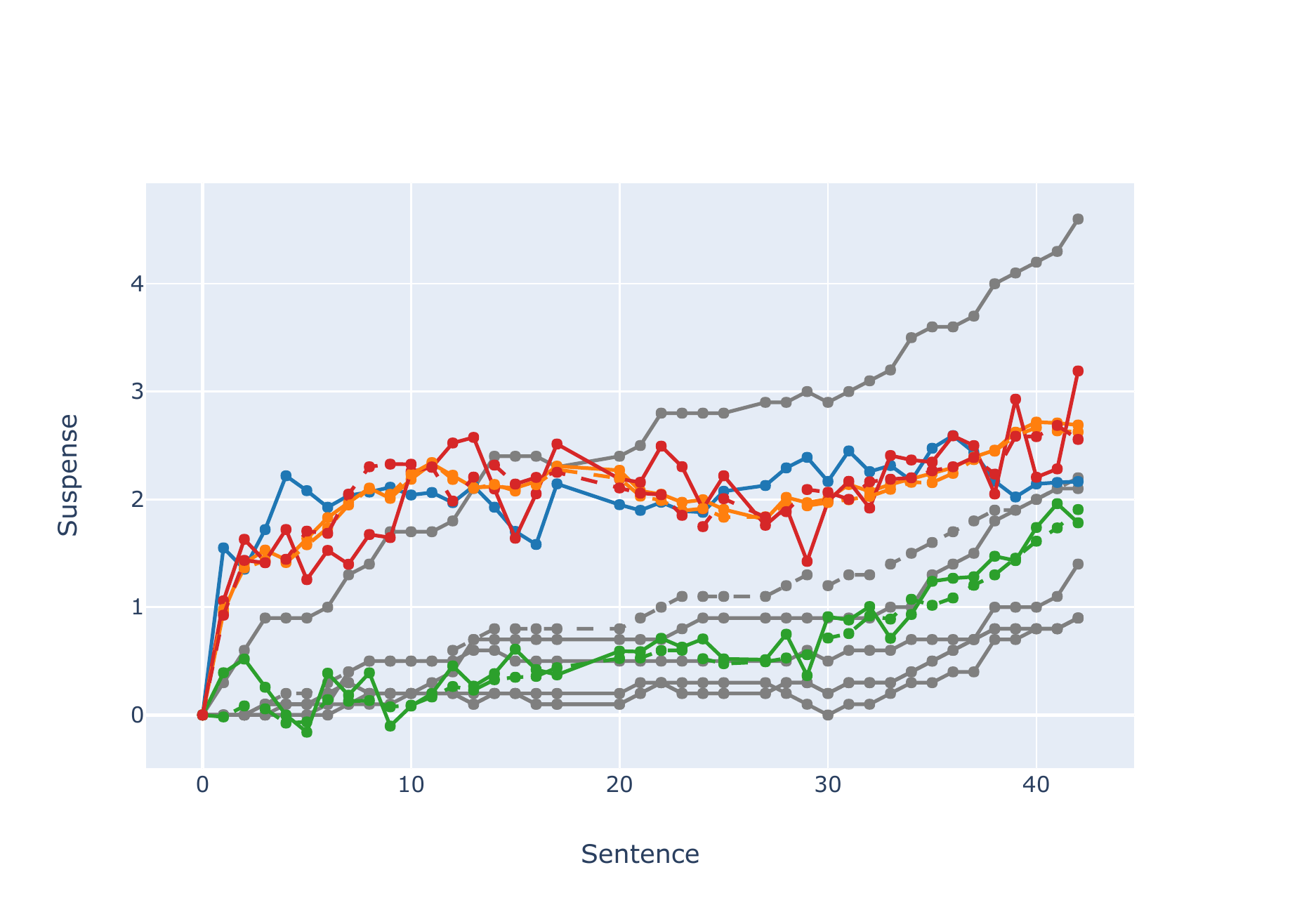}
\centering
\caption{Story 2066,
\textbf{\textcolor{annotatedcolor}{Human}},
\textbf{\textcolor{surpriseentropycolor}{$S^{\text{Hale}}$}},
\textbf{\textcolor{surprisecolor}{$S^{\text{Ely}}$}},
\textbf{\textcolor{suspensecolor}{$U^{\text{Ely}}$}},
\textbf{\textcolor{suspensestatecolor}{$U^{\alpha\text{Ely}}$.}} Solid
lines: generated alternative continuations, dashed lines: sampled
alternative continuations.}
\label{fig:wp_2066_main}
\end{figure}

\begin{figure}[tb]
\includegraphics[trim={0.5cm 0.5cm 2.5cm 2.5cm},clip,width=1.0\textwidth]{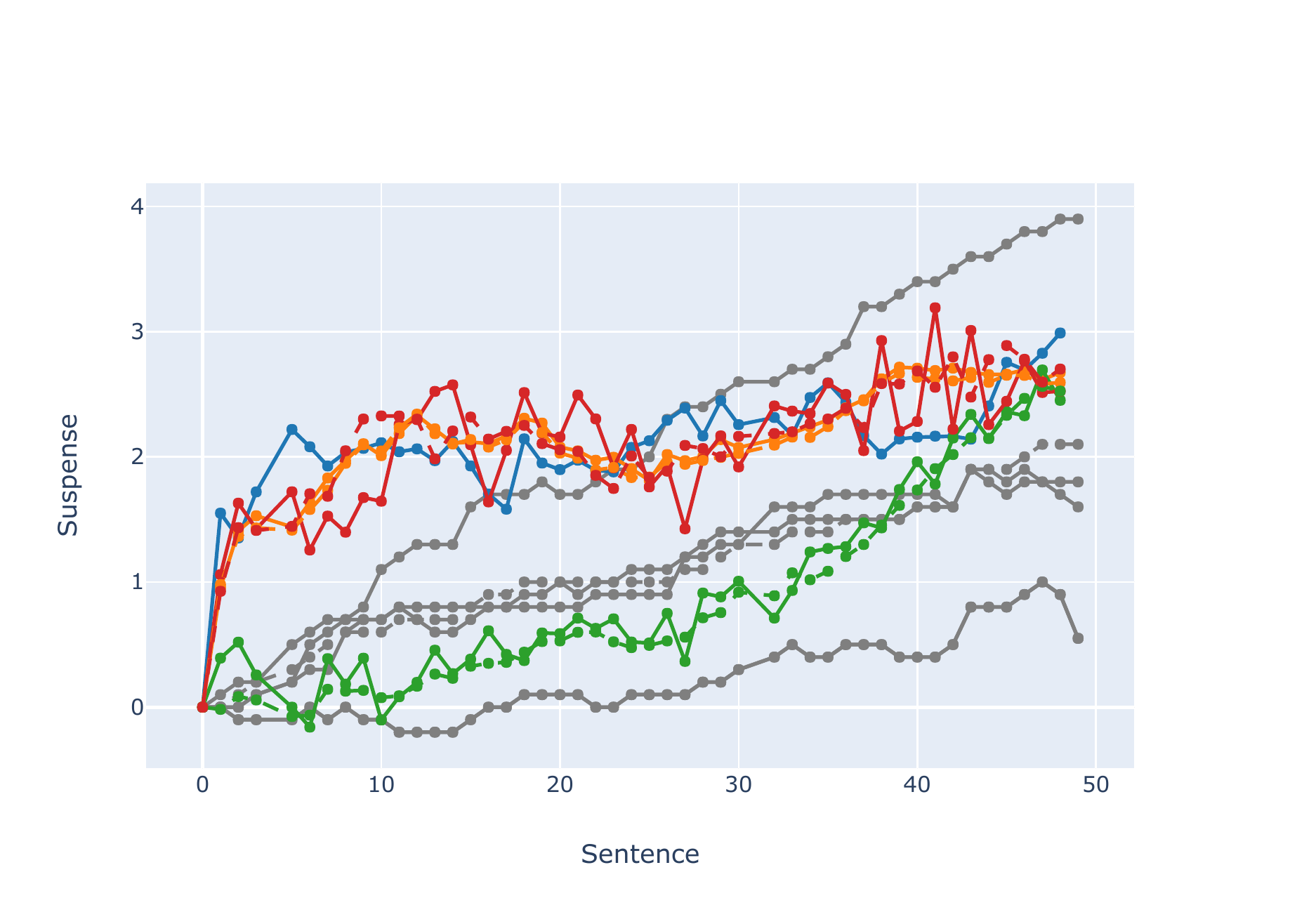}
\centering
\caption{Story 3203,
\textbf{\textcolor{annotatedcolor}{Human}},
\textbf{\textcolor{surpriseentropycolor}{$S^{\text{Hale}}$}},
\textbf{\textcolor{surprisecolor}{$S^{\text{Ely}}$}},
\textbf{\textcolor{suspensecolor}{$U^{\text{Ely}}$}},
\textbf{\textcolor{suspensestatecolor}{$U^{\alpha\text{Ely}}$.}} Solid
lines: generated alternative continuations, dashed lines: sampled
alternative continuations.}
\label{fig:wp_3203_main}
\end{figure}

To compare model predictions and absolute suspense values, two rank correlation methods,
Spearman's $\rho$ \citep{sen1968estimates} and Kendall's $\tau$
\citep{kendall1975rank}, are employed. Rank correlation is preferred because we are
interested in whether human annotators and models view the same part
of the story as more or less suspenseful. Rank correlation methods are good at detecting trends. Spearman's $\rho$  and Kendall's $\tau$ are preferred over the more standard Pearson as they better handle ordinal data with tiebreaks, as is the case with the suspense annotations. 

Several alternative approaches were considered that are commonly applied to time series evaluation. Cross-correlation \citep{stoica2005spectral} is effectively the normalized sliding dot product of one sequence in relation to another. As per the normal dot product, the agreement will be higher if both signals are higher in either direction. One advantage over rank correlation with the typical methods is it allows a lag between either sequence to be defined and test if sequences align with an offset. However, with suspense, we don't want this lag alignment as it is the exact alignment that is the aim. Also, as reviewed in the background to the annotations in Chapter \ref{chap:suspenseannotation}, rank correlation is preferred since it is the relative position that is most important. Cross-correlation would give more weighting to higher or lower values. A second approach is cointegration \citep{murray1994drunk}. The metaphor employed by Murray for cointegration is that of a drunk and their dog; the drunk may stagger around and not follow a straight path to their destination, and the dog may move about them. However, they can be cointegrated if, once the noise from both of their steps are removed, they are stationary in relation to another and follow the same path. Cointegration tests if two non-stationary processes are stationary in relation to each other. The potential advantage is that it is less prone to spurious correlations than other methods on time series. Some experiments were made applying the Engle-Granger method  \citep{10.2307/1913236} with the StatsModels library \citep{seabold2010statsmodels}. Unfortunately, applying this method was unstable, because cointegration is a fitted model. Typically cointegration methods are applied to much longer sequences on economic or environmental data. It is likely that the length of the sequences and the limited amount of annotations makes the method unstable. Even using a hand-picked example where the arcs of annotation and predictions are near identical, the model can fail to converge and produce a low statistical confidence value. 

The results compute $\rho$ and $\tau$ between the model predictions and the
judgements of each of the annotators (i.e.,~five times for five
annotators), and then take the average. We then average these values
again over the 100~stories in the test or development sets. As the
human upper bound, evaluation is computed as the mean pairwise correlation of the
five annotators. A comparison against the median judgements was also experimented with instead of pairwise against all the annotations. The problem with the median, though, is because the most common judgement is \textit{Same} it produces a flatter suspense trajectory with longer spans of sentences that have the same suspense value, which makes it worse for comparing with rank correlation methods because of ties.

Just as in the previous chapters, the annotators judgements can be plotted to show suspense and surprise against human judgement. Figures \ref{fig:wp_27_main}, \ref{fig:wp_2066_main}, and \ref{fig:wp_3203_main} show plots of the main Ely suspense and surprise plots for both $\alpha$ adjusted and non-adjusted variants.  The plots all have an initial jump for the first few sentences; this is an artefact over the LSTM with a limited context and doesn't impact the overall alignment with suspense.

The key for the models is as follows: $U$ represents suspense, and $S$ represents surprise. The two different approaches represented by Hale (uncertainty reduction) and Ely both are displayed as suffixes of the same name. There are two approaches to selecting alternative continuations: They can be randomly sampled from the corpus, which is a suffix of $\text{Cor}$, or generated with the GPT model, which has a suffix of $\text{\text{gen}}$. The impact adjustment for the $\alpha$ Ely extension is marked as $\alpha$. For example $U^{\text{Ely}}$-Gen is suspense, with the Ely method and generating continuations with GPT.

All the reported results are based on L1 similarity between vectors. In principle, the method works with any vector distance metric. A variety of others were tried, including Cosine distance, L2, and Wasserstein distance. However, the other methods generally performed worse and so only L1 is reported.\footnote{It should be noted that although the distance between vectors is based on L1 in these results as discussed earlier in the architecture the probability distribution over future continuations is based on Cosine.} 

The tables to follow also specify a \textit{Roll} values which is the number of levels of sentences of rollout in the tree. Fewer results are presented for multiple levels of rollout partly because it is extremely computationally expensive to rollout multiple sentences; it is one weakness with this approach compared to that of Chapter \ref{chap:tdvaesuspense}. Secondly, performance generally deteriorates, and so the rollout is given just to illustrate the pattern.

\begin{table}[ht!]
\centering
\begin{tabular}{@{}ll@{}ccc@{}}
\toprule
\textbf{Prediction}  & \textbf{Model} & \textbf{Roll} & \textbf{$\tau$ $\uparrow$}           & \textbf{$\rho$ $\uparrow$}           \\ \midrule
Human                &                &               & \textbf{.553}                        & \textbf{.614}                        \\ \midrule
Baselines            & WordOverlap    & 1             & .017                                 & .026                                 \\
                     & GloveSim       & 1             & .017                                 & .029                                 \\
                     & GPTSim         & 1             & .021                                 & .031                                 \\
                     & $\alpha$       & 1             & \textbf{.024}                        & \textbf{.036}                        \\ \midrule
$S^\text{Hale}$-Gen   & GRU            & 1             & .145                                 & .182                                 \\
                     & LSTM           & 1             & \textbf{.434}                        & \textbf{.529}                        \\ \midrule
$S^\text{Hale}$-Cor   & GRU            & 1             & .177                                 & .214                                 \\
                     & LSTM           & 1             & \textbf{.580}                        & \textbf{.675}                        \\ \midrule
$U^\text{Hale}$-Gen   & GRU            & 1             & \textbf{.036}                        & \textbf{.055}                        \\
                     & LSTM           & 1             & .009                                 & .016                                 \\ \midrule
$U^\text{Hale}$-Cor   & GRU            & 1             & .048                                 & .050                                 \\
                     & LSTM           & 1             & \textbf{.066}                        & \textbf{.094}                        \\ \midrule
$S^{\text{Ely}}$        & GRU            & 1             & \textbf{.484}                        & \textbf{.607}                        \\
                     & LSTM           & 1             & .427                                 & .539                                 \\ \midrule
$S^{\alpha\text{Ely}}$   & GRU            & 1             & .089                                 & .123                                 \\
                     & LSTM           & 1             & \textbf{.115}                        & \textbf{.156}                        \\  \bottomrule
\end{tabular}
\caption{Validation set results for WritingPrompts for generated
  (Gen) or corpus sampled (Cor) alternative continuations; $\alpha$
  indicates sentiment weighting. \textbf{Bold:} Best model in a given category.} 
\label{tab:dev_wp_res}
\end{table}

\begin{table}[ht!]
\centering
\begin{tabular}{@{}ll@{}ccc@{}}
\toprule
\textbf{Prediction}  & \textbf{Model} & \textbf{Roll} & \textbf{$\tau$ $\uparrow$}           & \textbf{$\rho$ $\uparrow$}           \\ \midrule
Human                &                &               & \textbf{.553}                        & \textbf{.614}                        \\ \midrule
Baselines            & WordOverlap    & 1             & .017                                 & .026                                 \\
                     & GloveSim       & 1             & .017                                 & .029                                 \\
                     & GPTSim         & 1             & .021                                 & .031                                 \\
                     & $\alpha$       & 1             & \textbf{.024}                        & \textbf{.036}                        \\ \midrule

$U^{\text{Ely}}$-Gen   & GRU             & 1             & .241                                 & .161                                 \\
                     &                & 2             & .304                                 & .399                                 \\
                     & LSTM           & 1             & {\color[HTML]{9A0000} \textbf{.610}} & {\color[HTML]{9A0000} \textbf{.698}} \\
                     &                & 2             & .393                                 & .494                                 \\ \midrule
$U^{\text{Ely}}$-Cor    & GRU            & 1             & .229                                 & .264                                 \\
                     &                & 2             & .512                                 & .625                                 \\
                     &                & 3             & .515                                 & .606                                 \\
                     & LSTM           & 1             & \textbf{.594}                        & \textbf{.678}                        \\
                     &                & 2             & .564                                 & .651                                 \\
                     &                & 3             & .555                                 & .645                                 \\ \midrule
$U^{\alpha\text{Ely}}$-Gen & GRU          & 1             & .216                                 & .124                                 \\
                     &                & 2             & .219                                 & .216                                 \\
                     & LSTM           & 1             & \textbf{.474}                        & \textbf{.604}                        \\
                     &                & 2             & .316                                 & .418                                 \\ \midrule
$U^{\alpha\text{Ely}}$-Cor & GRU          & 1             & .205                                 & .254                                 \\
                     &                & 2             & .365                                 & .470                                 \\
                     & LSTM           & 1             & \textbf{.535}                        & \textbf{.642}                        \\
                     &                & 2             & .425                                 & .534                                 \\ \bottomrule
\end{tabular}
\caption{Validation set results for WritingPrompts for the split out $ U^{\text{Ely}} $ variants. $\alpha$ indicates sentiment weighting.  \textbf{Bold:} Best model in a given category;  {\color[HTML]{9A0000}\textbf{Red:}} Best model overall.}
\label{tab:dev_wp_res_ely}
\end{table}

The validation set results are presented split across two Table~\ref{tab:dev_wp_res} with the baselines, the Hale results, and Ely surprise. Table~\ref{tab:dev_wp_res_ely} repeats the baselines and has the Ely suspense. On the validation set all baselines perform poorly. Random is not reported, but as would be expected, it is around 0.0. The simple WordOverlap, GloveSim and GPTSim in increasing sophistication attempts to model if there are just simple of increasing or decreasing vector distances that correlate with suspense. In addition, the $\alpha$ absolute measure of suspense also correlates poorly. Both suggest there is not a simple pattern, such as sentence vectors gradually getting further away when the story gets more suspenseful or vice versa. The $\alpha$ baseline suggests there is no direct relationship between sentimentality and suspense. This baseline weights negative sentences twice as high as positive. Some of the reviewed background material suggested a strong negative sentiment (fear, anger, heartache, etc.) would correlate with suspense, but the relationship is less clear. 

The main difference between the hierarchical rollout model and the baselines is that the baselines have no context, whereas the hierarchical model has the context of the story. The ability to aggregate across the story state when predicting the next sentence should give the hierarchical models advantages. The Hale models, unlike Ely, rely purely on the probability of continuations and not the vector state. Overall the Hale surprise $S^\text{Hale}$ performs well, reaching a maximum $\rho$ of .675 on the validation set. Hale uncertainty reduction $U^\text{Hale}$, however, performs consistently poorly. 

The Ely metric combines both probabilities with the outcome vector state. Ely surprise $S^\text{Ely}$ also performs well, reaching a similar value as Hale surprise. Overall, Ely uncertainty reduction $U^\text{Ely}$ is the strongest performer, with $\rho = .698$, numerically outperforming the human upper bound. Surpassing the human measure should be taken with a pinch of salt. The human judgements are variable with a moderate level of agreement. As noted the task is inherently a subjective task and so it shouldn't be expected that either human annotators would always agree with each other, or that extremely high agreement with model predictions would be possible.  Nevertheless, it is still a good performance.

Some other trends are clear from the validation set: Apart from in one case, the LSTM model surpasses the GRU, which is expected as the gating mechanism is more sophisticated. The $\alpha$ weighting degrades performance slightly. It is perhaps surprising as it was hypothesised that strong sentiment would correlate with more dramatic events. It would more closely align with suspense by magnifying the underlying suspense and surprise. 

From the plots earlier, though, one noticeable function of the $\alpha$ adjustment is that it makes the plots noisier, which would be the reason for degrading performance. Also, the multistep rollout slightly degrades performance. 

Generally, the generation methods for the main $U^{\text{Ely}}$ method is stronger than the corpus variant. The generation method should be better since generating natural continuations should be more informative. On other metrics, the picture is mixed, and the corpus approach works better with Hale surprise. One explanation for this is that it is essentially the probability of the correct next sentence versus randomly picked samples. It is similar in principle to negative examples often used during loss training. It is a more straightforward task than the generation version since this version must determine the correct next sentence from GPT generated continuations which would be much harder than random. 

The intuition for rolling out more than one sentence ahead is that suspense is about divergence in outcomes, so looking further ahead should improve performance. One reason for performance not improving is that it would depend on the GPT model being able to develop a plot coherently over multiple sentences. It represents more of a challenge and hints at some suggestions for future work discussed later, inspired by more recent story generation systems.

\begin{table}[tb]
\centering
\begin{tabular}{@{}lrr@{}}
\toprule
\textbf{Prediction}        & \multicolumn{1}{c}{$\tau$ $\uparrow$}                   & \multicolumn{1}{c}{$\rho$ $\uparrow$}                     \\ \midrule
Human                      & {\color[HTML]{9A0000} \textbf{.652}} (.039)     & {\color[HTML]{9A0000} \textbf{.711}} (.033)     \\ \midrule
$S^\text{Hale}$-Gen               & .407                                 (.089)     & .495                                 (.081)     \\
$S^\text{Hale}$-Cor               & \textbf{.454}                        (.085)     & \textbf{.523}                        (.079)     \\
$U^\text{Hale}$-Gen               & .036                                 (.102)     & .051                                 (.102)     \\
$U^\text{Hale}$-Cor               & .061                                 (.100)     & .088                                 (.101)     \\ \midrule
$S^{\text{Ely}}$              & .391                                 (.092)     & .504                                 (.082)     \\
$U^{\text{Ely}}$-Gen          & \textbf{.620}                         (.067)     & \textbf{.710}                        (.053)     \\
$U^{\text{Ely}}$-Cor          & .605                                  (.069)     & .693                                 (.056)     \\
$U^{\alpha\text{Ely}}$-Gen & .450                                 (.085)     & .580                                 (.072)     \\
$U^{\alpha\text{Ely}}$-Cor & .538                                 (.077)     & .646                                 (.064)     \\
\bottomrule
\end{tabular}
\caption{Test set results for WritingPrompts for generated
  (Gen) or corpus sampled (Cor) continuations. LSTM with rollout 
  one; brackets: confidence intervals.}
\label{tab:test_wp_res}
\end{table}

For the test set results in Table~\ref{tab:test_wp_res} the
upper and lower confidence bounds were computed using the Fisher
$Z$-transformation ($p < 0.05$). Only the best models on the validation set with the LSTM model reported. On the test set, $U^{\text{Ely}}$ again is the best measure, with a correlation statistically
indistinguishable from human performance (based on CIs). The pattern on the test set matches the validation set across the various metrics. Overall, the correlations are higher on the test set than the validation set, presumably reflecting the higher human upper bound of inter-annotator agreement.

Overall, the conclusion is that the hierarchical architecture successfully
models human suspense judgements on the WritingPrompts dataset. The
overall best predictor is $U^{\text{Ely}}$, uncertainty reduction
computed over story representations. This measure combines the
probability of continuation ($S^\text{Hale}$) with the distance between
story embeddings ($S^{\text{Ely}}$), which are both good predictors in
their own right. This finding supports the theoretical claim that
suspense is an expectation over the change in future states of a game
or a story, as advanced by \citet{ely2015suspense}, and can be modelled by the hierarchical model of this chapter. More analysis, including caveats to the findings and limitations, will be discussed after presenting results for a secondary task, movie turning points. 

\subsection{Movie Turning Points}

An interesting question is whether the peaks in suspense in a story
correspond to important narrative events. Such events are sometimes
called turning points (TPs) and occur at certain positions in a movie
according to screenwriting theory \citep{cutting2016narrative}. A corpus
of movie synopses annotated with turning points is available in the
form of the TRIPOD dataset \citep{Papalampidi2019MoviePA}. The turning points scheme is from \citet{freytag1894freytag}, who, rather than have specific functions and spheres, splits plots into a pyramid of dramatic tension that increases with \textit{Exposition}, \textit{Inciting Incident}, \textit{Rising Action}, \textit{Crisis}, then to a \textit{Climax}, and then finishing with \textit{Falling Action}, and \textit{Denouncement} (the ending). Between each of these stages are essential turning points that are the most important parts of the plot. The Freytag plot model has been used as part of a supervised approach to identify the turning points of stories and assist with summarisation \citep{Papalampidi2019MoviePA,papalampidi-etal-2020-screenplay}. It is possible therefore to test if surprise or suspense from the Hale or Ely models predict turning points in TRIPOD. The tripod dataset consists of 99 annotated movies across a training and test set. As the model is trained on a corpus of short stories, this will also serve as an out-of-domain evaluation.

\begin{figure}[htbp]
\centering
\includegraphics[trim={1.5cm 0.5cm 2.5cm 2.5cm},clip,width=1.0\textwidth]{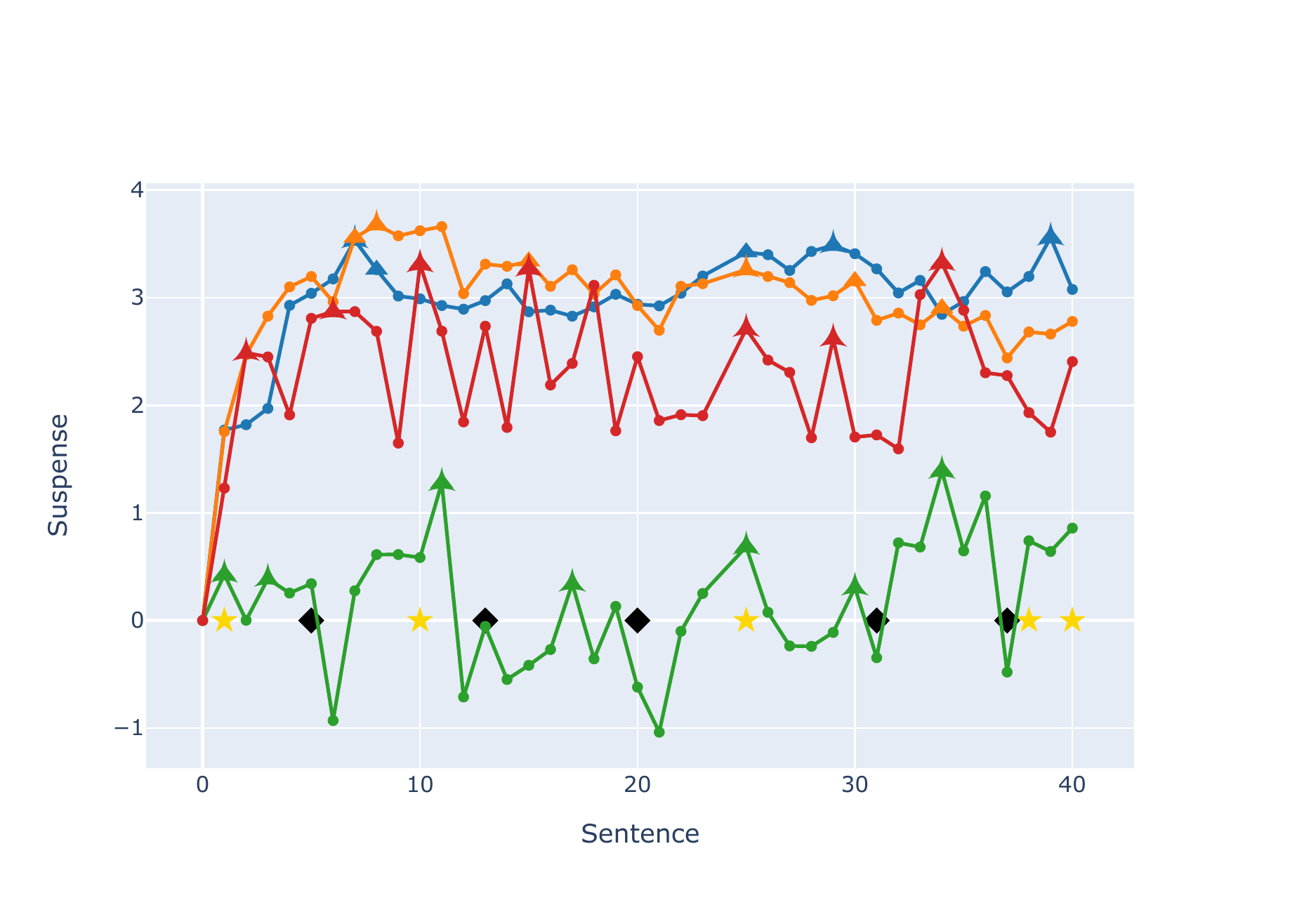}
\caption{\href{https://www.imdb.com/title/tt0100405/}{Pretty Woman},
 \textbf{\textcolor{surpriseentropycolor}{$S^{\text{Hale}}$}}, \textbf{\textcolor{surprisecolor}{$S^{\text{Ely}}$}}, \textbf{\textcolor{suspensecolor}{$U^{\text{Ely}}$}}, \textbf{\textcolor{suspensestatecolor}{$U^{\alpha\text{Ely}}$}}, $\medblackdiamond$ theory baseline, {\color{yellow} $\medstar$} TP annotations, triangles are
  predicted TPs.}
\label{fig:tp_train_36}
\end{figure}

As an illustration of the dataset and task, the following is the full text for the synopsis of the film \href{https://www.imdb.com/title/tt0100405/}{Pretty Woman} plotted in Figure~\ref{fig:tp_train_36}. As in the previous evaluation all the metrics are scaled with standard scaling and centred. The predictions in the plots will be discussed later in this section.  \citet{Papalampidi2019MoviePA} recasts Freytag's schemas with the following turning points (TPs): 1.~Opportunity, 2.~Change of Plans, 3.~Point of no Return, 4.~Major Setback, and 5.~Climax. The dataset annotates these turning points at the sentence level with synopses and at the scene level within the movie script. All evaluation is entirely on the synopses. The following is the text for \textit{Pretty Woman}, the Gold standard turning points are marked in the text:

\begin{enumerate}
\item  \textit{\textbf{Opportunity:}} \textbf{Edward Lewis (Gere), a successful businessman and "corporate raider", takes a detour on Hollywood Boulevard to ask for directions.}
\item  Receiving little help, he encounters a prostitute named Vivian Ward (Roberts) who is willing to assist him in getting to his destination.
\item  The morning after, Edward hires Vivian to stay with him for a week as an escort for social events. 
\item  Vivian advises him that it "will cost him," and Edward agrees to give her \$3,000 and access to his credit cards. 
\item  Vivian then goes shopping on Rodeo Drive, only to be snubbed by saleswomen who disdain her because of her unsophisticated appearance. 
\item  Initially, hotel manager Barnard Thompson (Hector Elizondo) is also somewhat taken aback. 
\item  But he relents and decides to help her buy a dress, even coaching her on dinner etiquette. 
\item  Edward returns and is visibly amazed by Vivian's transformation. 
\item  The business dinner does not end well, however, with Edward making clear his intention to dismantle Morse's corporation once it was bought, close down the shipyard which Morse spent 40 years building, and sell the land for real estate. 
\item  \textit{\textbf{Change of Plans:}} \textbf{Morse and his grandson abandon their dinner in anger, while Edward remains preoccupied with the deal afterward.} 
\item   Back at the hotel, Edward reveals to Vivian that he had not spoken to his recently deceased father for 14 and half years.
\item  Later that night, the two make love on the grand piano in the hotel lounge. 
\item  The next morning, Vivian tells Edward about the snubbing that took place the day before. 
\item  Edward takes Vivian on a shopping spree. 
\item  Vivian then returns, carrying all the bags, to the shop that had snubbed her, telling the salesgirls they had made a big mistake. 
\item  The following day, Edward takes Vivian to a polo match where he is interested in networking for his business deal. 
\item  While Vivian chats with David Morse, the grandson of the man involved in Edward's latest deal, Philip Stuckey (Edward's attorney) wonders if she is a spy. 
\item  Edward re-assures him by telling him how they met, and Philip (Jason Alexander) then approaches Vivian and offers to hire her once she is finished with Edward, inadvertently insulting her. 
\item  When they return to the hotel, she is furious with Edward for telling Phillip about her. 
\item  She plans to leave, but he apologizes and persuades her to see out the week. 
\item  Edward leaves work early the next day and takes a breath-taking Vivian on a date to the opera in San Francisco in his private jet. 
\item  She is clearly moved by the opera (which is La Traviata, whose plot deals with a rich man tragically falling in love with a courtesan). 
\item  While playing chess with Edward after returning, Vivian persuades him to take the next day off. 
\item  \textit{\textbf{Point of No Return:}} \textbf{They spend the entire day together, and then have sex, in a personal rather than professional way.}
\item  Just before she falls asleep, Vivian admits that she's in love with Edward.
\item  Over breakfast, Edward offers to put Vivian up in an apartment so he can continue seeing her. 
\item  She feels insulted and says this is not the "fairy tale" she wants. 
\item  He then goes off to work without resolving the situation. 
\item  Vivian's friend, Kit De Luca (Laura San Giacomo), comes to the hotel and realizes that Vivian is in love with Edward. 
\item  Edward meets with Mr. Morse, about to close the deal, and changes his mind at the last minute. 
\item  His time with Vivian has shown him another way of living and working, taking time off and enjoying activities for which he initially had little time. 
\item  As a result, his strong interest towards his business is put aside. 
\item  He decides that he would rather help Morse than take over his company. 
\item  Furious, Philip goes to the hotel to confront Edward, but only finds Vivian there. 
\item  He blames her for changing Edward and tries to rape her. 
\item  Edward arrives in time to stop Philip, chastising him for his greed and ordering him to leave the room. 
\item  Edward tends to Vivian and tries to persuade her to stay with him because she wants to, not because he's paying her. 
\item   \textit{\textbf{Major Setback:}} \textbf{She refuses once again and returns to the apartment she shares with Kit, preparing to leave for San Francisco to earn a G.E.D. in the hopes of a better life.}
\item  Edward gets into the car with the chauffeur that took her home. 
\item  \textit{\textbf{Climax:}} \textbf{Instead of going to the airport, he goes to her apartment arriving accompanied by music from La Traviata.}

\end{enumerate}

 From \citet{thompson1999storytelling} it is hypothesised that the typical Hollywood movie contains these turning points at predefined points as plotted in Figure \ref{fig:tp_train_36} with the black diamonds. There are strong links but also differences between the approach and the suspense ideas reviewed in Chapter \ref{chap:backgroundtheory}: To start with the differences, the most obvious is that Ely is a bottom-up model of suspense, as were most of the discussed theories of suspense. As has been reviewed with the Ely theory, the essential feature is uncertainty about the divergence of outcomes, especially if these are highly consequential states. This view of suspense suggests suspenseful events can occur at any point in the story. In contrast, turning points shares that a story branches at different directions at these points. However, while dramatic events can occur anywhere, top-down events are expected at specific points in the story. They are also important at the level of the whole plot rather than more locally as suspense is. Papalampidi et al.'s annotations identify where these important points arise and are generally close to where theory suggests they should be. Inferring suspense, as in the WritingPrompts results, is thus an interesting comparison as it tests whether the unsupervised model can also infer the related turning points, compete against a supervised model, and transfer to a different domain.

The best performing model from Papalampidi et al. is TAM (Topic Aware Model). TAM is a hierarchical bi-directional model. The bottom layer of the TAM model is whole sentences encoded into vectors with USE (Universal Sentence Encoder; \citealt{cer-etal-2018-universal}). After this, a directional LSTM layer concatenates the left and right contexts. On top of the LSTM is a \textit{context interaction} layer that combines the left interactions of the left and right context for each sentence representation via cosine similarity (or alternative distance metrics) to model the interactions and as a means of implicitly marking segmentation of scenes. Additional enhancements provide named entity metadata and $5$ per turning point encoders that feed into merging layers that can be fed into a classifier that attempts to predict each turning point individually. The relevance is that this is a complicated hierarchical model in terms of engineering pipelines comparable to the hierarchical rollout model. The model is also bi-directional, which again should be an advantage as the model is able to learn if an event is a turning point when 
the main difference is that the model is supervised on the TRIPOD dataset, which should give it a substantial advantage over an unsupervised model. A simplified CAM (Context-Aware Model) is the same model without the \textit{context interaction} entity or per turning point encoders. The method of inferring turning points is just to use the values of the output classifier to identify the most likely candidate sentences for synopses. In addition to TAM and CAM, the theory baseline is reported, which is just if the turning points are placed in the most likely positions according to screenplay theory. The theory positions are just the expected percentage position specified that the turning points should occur at by \citet{thompson1999storytelling}.

\begin{figure}[tb]
\includegraphics[trim={1.5cm 0.5cm 2.5cm 2.5cm},clip,width=1.0\textwidth]{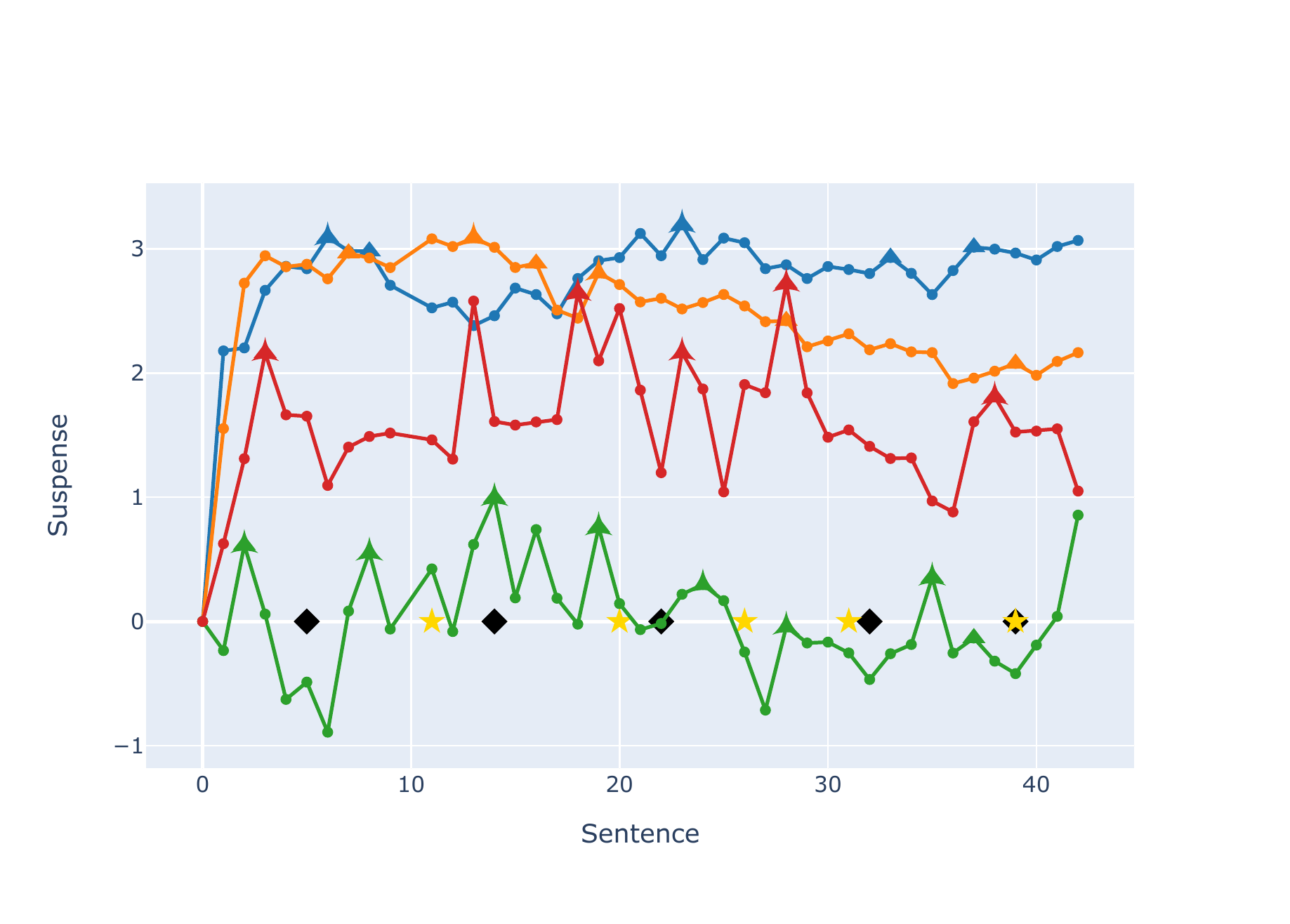}
\caption{{\href{https://www.imdb.com/title/tt1010048/}{Slumdog Millionaire}},
  \textbf{\textcolor{surpriseentropycolor}{$S^{\text{Hale}}$}},
  \textbf{\textcolor{surprisecolor}{$S^{\text{Ely}}$}},
  \textbf{\textcolor{suspensecolor}{$U^{\text{Ely}}$}},
  \textbf{\textcolor{suspensestatecolor}{$U^{\alpha\text{Ely}}$}},
  $\medblackdiamond$ theory baseline, $ \color{yellow} \medstar$~TP annotations, triangles are
  predicted TPs.}
\label{fig:tp_exp_2}
\end{figure}

\begin{figure}[t]
\centering
\includegraphics[trim={1.5cm 0.5cm 2.5cm 2.5cm},clip,width=1.0\textwidth]{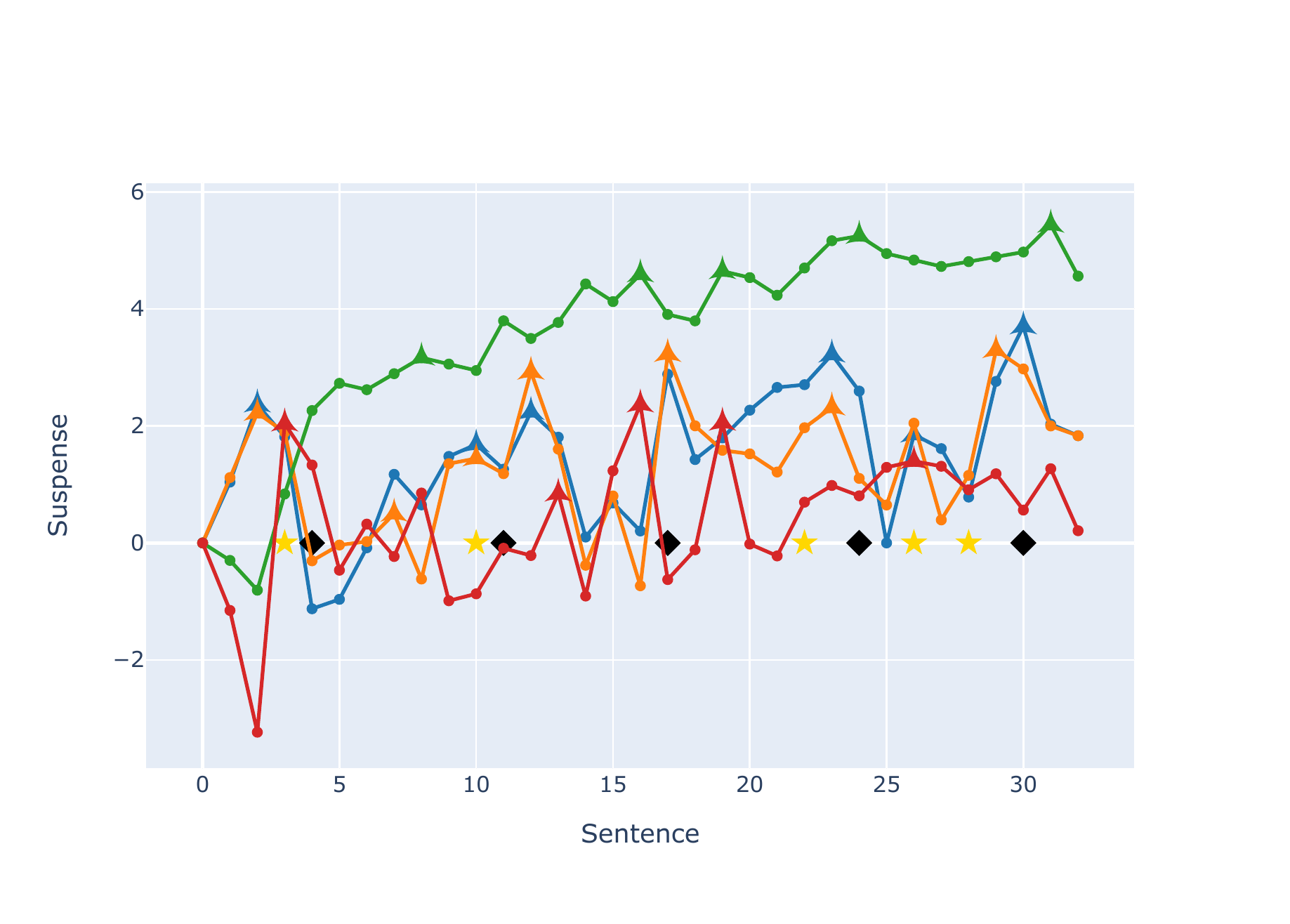}
\caption{\href{https://www.imdb.com/title/tt0073195//}{Jaws}, \textbf{\textcolor{surpriseentropycolor}{$S^{\text{Hale}}$}}, \textbf{\textcolor{surprisecolor}{$S^{\text{Ely}}$}}, \textbf{\textcolor{suspensecolor}{$U^{\text{Ely}}$}}, \textbf{\textcolor{suspensestatecolor}{$U^{\alpha\text{Ely}}$}}, $\medblackdiamond$ theory baseline, {\color{yellow} $\medstar$} TP annotations, triangles are
  predicted TPs.}
\label{fig:tp_exp_3}
\end{figure}

Figure \ref{fig:tp_exp_2} and Figure \ref{fig:tp_exp_3} show two more example plots with the main suspense metrics plotted against the turning points for movies. Upwards arrows identify the peaks identified by SciPy's \textit{find peaks} function, which doesn't identify the maximum values along the Y-axis but instead finds more local maxima with lower values on either side. \footnote{The peaks in the plot are filtered only to show the most prominent peaks by a threshold to keep the diagrams readable and not all those identified. } \textit{Prominence} which is defined in the SciPy algorithm as the \textit{as the vertical distance between the peak and its lowest contour line} is used to define the magnitude of a peak for identifying turning points. The rationale is that turning points are important changes in a plot with acts of the story. Just selecting the highest values in a story that builds to a dramatic conclusion would just lead to all the values near the climax being selected. The difference between the application of the suspense model and the TAM model is that TAM has specific classifier heads for each turning point. In contrast, the hierarchical rollout only has a single output value per sentence metric. Therefore categorising peaks as TP1, TP2, etc. is based on position.  

Papalampidi et al. have three methods of evaluation. TA (Total Agreement) is whether a predicted turning point matches exactly the predicted turning point. PA (Partial Agreement) allows a small window where it can be close. However, only the Annotation Distance is presented in this thesis, which is in eqn. (\ref{eqn:distance_eval}):

\begin{myequation}
\begin{aligned}
 d[p_i, tp_i] = \frac{1}{N}|p_i - tp_{i}|
\label{eqn:distance_eval}
\end{aligned}
\end{myequation}

Given the predicted turning point $p_i$ and the Gold standard annotated turning point $tp_i$, the equation is the mean distance normalised by the synopsis length $N$. The reason for preferring the method: There are only $84$ training set movies and $15$ test set and given only $5$ turning points, exact agreement can be noisy. A poor predictor that randomly hits a few turning points but does badly on most turning points could score higher than a model that's generally close but not exactly correct. Therefore average distance is a better evaluation for the task. 

\begin{table}[tb]
\centering
\begin{tabular}{@{}lcc@{}}
\toprule
            & \multicolumn{1}{c}{\textbf{Dev $D$} $\downarrow$}            & \multicolumn{1}{c}{\textbf{Test $D$} $\downarrow$}                    \\ \midrule
Human       & Not reported          & \color[HTML]{9A0000}{\textbf{4.30 (3.43)}}         \\ \midrule
Theory Baseline & 9.65 (0.94)          & 7.47 (3.42)          \\
Random & - & 37.8 (25.3) \\
CAM         & 7.44 (8.09) & - \\
TAM         & \color[HTML]{9A0000}{\textbf{7.11 (7.98)}} & 6.80 (5.19) \\ \midrule
WordOverlap          & 13.9 (1.45)          & 12.7 (3.13)         \\
GloveSim          & \textbf{10.2 (2.74)}          & \textbf{10.4 (2.54)}         \\ 
GPTSim          & 16.8 (1.47)       & 18.1 (4.71)         \\ 
$\alpha$          & 11.3 (1.24)         & 11.2 (2.67)         \\  \midrule
$S^\text{Hale}$-Gen        & \textbf{8.27 (1.68)}          & \textbf{8.72 (2.27)}          \\
$U^\text{Hale}$-Gen        & 10.9 (1.02)          & 10.7 (3.66)          \\
\midrule
$S^{\text{Ely}}$    & 9.54 (1.56)          & 9.01 (1.92)          \\
$S^{\alpha\text{Ely}}$       & 9.95 (1.78)          & 9.54 (2.76)          \\
$U^{\text{Ely}}$-Gen       & 8.75 (1.76)          & 8.38 (1.53)          \\
$U^{\text{Ely}}$-Cor       & 8.74 (2.76)         & 8.50 (1.69)          \\
$U^{\alpha\text{Ely}}$-Gen   & 8.80 (1.61) & 7.84 (3.34) \\
$U^{\alpha\text{Ely}}$-Cor    &\textbf{8.61 (1.68)} & \textbf{7.78 (1.61)} \\ \bottomrule
\end{tabular}
\caption{TP prediction on the TRIPOD
  validation and test sets. $D$ is the normalised distance to the
  gold standard. The standard deviation is in brackets.}
\label{tab:turn_point_res}
\end{table}

To select the turning points with the hierarchical rollout model in an unconstrained way would be to select the five biggest peaks from across the whole story. Papalampidi et al. instead constrain the evaluation so that the turning points are only selected from a window calculated from the standard deviations of each turning point in the annotation, so our evaluation matches this approach.

The results on the TRIPOD training and test
sets are reported in Table~\ref{tab:turn_point_res}. Both are reported due to the small number of synopses in TRIPOD. Secondly, unlike with the original paper's models, there is no supervision in the case of the training set. The unsupervised model is disadvantaged on the training set.\footnote{There were some further trials in trialling to finetune on the TRIPOD training set, but this didn't improve performance, and so the results aren't reported.} The results reported are a single level of the rollout from the best LSTM suspense model. The best performing hierarchical rollout model on the test set with $D = 7.78$ is $U^{\alpha\text{Ely}}$-Cor, with $U^{\alpha\text{Ely}}$-Gen is only slightly worse. Performance on the training set is slightly worse for each. The $S^\text{Hale}$-Gen model also does well on both training and test sets. Compared to the CAM and TAM models, the models are slightly worse and significantly worse than human performance. Crucially though, the best suspense measures are relatively close in performance to the supervised models, which would be expected to do better. They also perform far better than any of the baselines or random performance. The overall result is moderately successful. It supports the idea that there is an overlap between suspense and the concept of the turning points, that the overlap extends to the implemented models and that the model has predictive power. As noted by Papalampidi et al., turning points often mark the boundaries or segments between sequences of events in the story. These boundaries will naturally produce more surprise or suspense with the Ely or Hale definitions. This is because changes in direction will naturally increase uncertainty. In the case of suspense generation, variants would be likely to produce more divergent text. In the case of surprise, the expectation would differ more from reality. In all these cases, the prediction would be more likely to produce a peak and so indicate a turning point. However, as a cautionary note, the model is still far away from human performance, and clearly, there is much scope for improvement and further work.

\section{Conclusion}

The overall findings suggest that by implementing concepts from
psycholinguistic and economic theory, the hierarchical models can predict reasonably well human judgements
of suspense in storytelling. That uncertainty reduction
($U^\text{Ely}$) outperforms probability-only ($S^\text{Hale}$) and
state-only ($S^\text{Ely}$) surprise suggests that, while
consequential state change is of primary importance for suspense,
the probability distribution over the states is also a necessary
factor. Uncertainty reduction, therefore, captures the view of suspense
as reducing paths to the desired outcome, with more significant shifts
as the story progresses \citep{DBLP:conf/aaai/ONeillR14,
  ely2015suspense, perreault2018universal}. It is more in line with
the \citet{smuts2008desire} Desire-Frustration view of suspense, where
uncertainty is secondary. Secondly, as noted with the movie turning points task, there is some promise for predicting more formulaic and top-down structures.  This work has demonstrated that hierarchical neural language models
are not only promising for generation but can also model suspense
in a way that corresponds to human judgement. It also manifests the
advantages of applying an unsupervised model combined with models
from theory rather than relying on large amounts of hand-coded
training data examples. As outlined in Chapter \ref{chap:suspenseannotation} the process required to collect even the smaller number of annotations for evaluation is relatively expensive. The cost would be exorbitant for a training set that would require thousands of annotations. That is only for short stories without considering the additional cost of extending annotations to longer works such as screenplays or novels.
  
Strong psycholinguistic claims about suspense are difficult to make
due to several weaknesses in the approach. The proposed model does not have a higher-level
understanding of event structure or can be expected to plan long-term. Since the Ely suspense depends on generating significant and plausible longer-term future events in a story, this is a problem. It is likely why multi-step sentence rollout performs worse as GPT is not able to develop the plot beyond a few sentences. The recent success of language models in wide-ranging NLP tasks
\cite[e.g.,][]{radford2019language} has shown those language models are
capable of learning semantically rich information implicitly. However, in-text generation, \citet{fan-etal-2019-strategies} amongst others have found that explicitly incorporating coreference and structured event
representations into generation produce more coherent generated
text. The literature overlaps with story planning and all the techniques discussed in the background and future work section of Chapter \ref{chap:tdvaegeneration} on text generation. Generating plausible longer-term continuations also requires better comprehension of real-world common sense events, as was reviewed in Chapter \ref{chap:backgroundml}. One relatively straightforward enhancement would be to extract simplified events from sentences, say the key verb plus subject and object. Coreferences could also be replaced with codes such as \textit{ent0}, \textit{ent1}. The intuition is that with simplified representations, the model would be able to develop better longer-term plots as per story planning systems. While simplified events are likely to improve, it is left for future work as there is another more pressing problem with the rollout model.

The more severe issue with rolling future continuations with concrete text generation is complexity. If, hypothetically, it were possible to make a system that could generate a distribution of plausible continuations that are as good as the best writers stretching far ahead, then there would still be a problem. It is especially the case if, instead of a couple of sentences or events, the model needed to look $10$ or $20$  ahead. An exhaustive search has a complexity of $\mathcal{O}(\left|\mathcal{Y}\right|^{T'})$ with $\mathcal{Y}$ being the number of events or sentences to be generated, and ${T'}$ is the number of them to look ahead. It is clearly impractical when a large neural language model might take a second or two to generate a handful of events or sentences. Even with a beam search with complexity $\mathcal{O(\mathcal{K}\left|\mathcal{Y}\right|T')}$ where $\mathcal{K}$ is the beam is impractical if there needs to be a beam of $50$ or $100$ to maintain diversity in the continuations. The complexity, while making it infeasible to look further ahead, also makes it infeasible to run on long stories such as full-length screenplays or novels. 

There is also a second problem: While the method presented in this Chapter directly implements the Ely concept, it is far from cognitively plausible. If the method was replicated with humans by, say, a creative writing group, then at each step in the story, each writer in the group would be asked to write a continuation. Then suspense would be how much they diverged, plus with the $\alpha$ extension, the significance of each of the end states. It is highly improbable that the typical reader of a novel or a viewer of a movie thinks implicitly or explicitly about a large number of possible continuations of a story to the end with a Bayesian expectation over each one occurring. While Ely is conceptually valid, there are practical and cognitive issues. It is far from a problem with Ely alone; for example, \citet{castricato-etal-2021-towards} has proposed a recent model of narrative understanding that overlaps somewhat with the intuitions behind the Ely model. The Castricato et al. model is one where reader uncertainty is measured over time during a story. The reader uncertainty is measured by the number of parallel world states (generated by a concrete model) that are plausible with what is known. For example, with the mysterious benefactor in \textit{Great Expectations}, there could be many possibilities, and until revealed, the world state would have contained many of them. Computationally, this is far more complicated than the Ely rollout as its parallel world states must be generated and maintained, not just short-term future possibilities. Another problem is that, in the case of the benefactor in the story, \textit{Pip} is entirely sure who the benefactor is, and not a distribution over alternatives, the plot then twists. He finds out he is wrong, as does the reader, who assumes the same, which creates much drama. It is hypothesised the same applies to Ely and that likely approximation of what will happen is more critical than a complete distribution over exhaustive possibilities. Therefore, the next Chapter simplifies the rollout with variational approximations in the vector space to reduce the complexity of the implementation and make it more intuitive. 

Another point of note from the results is the failure of the $\alpha$ extension to include the importance of state within the suspense and surprise metrics. It makes intuitive sense that more important states in terms of consequences would impact suspense and surprise within a story. Absolute sentiment doesn't seem a good measure for this and introduces noise. There are no more practical experiments with $\alpha$ in the thesis as the focus of the next chapter is comparing alternative methods for the base Ely surprise models. Nevertheless, it is an important area of improvement for both variants of the model. One idea in this space for future work: Typically, the most important events will occur in the synopses or summaries. This is the approach taken in the evaluation of Chapter \ref{chap:salience} with the Shmoop corpus \citep{DBLP:journals/corr/abs-1912-13082} of aligned summaries. Therefore an approach might be to label events occurring in the synopsis as important and those not mentioned unimportant and then train a model that could be applied as the $\alpha$ in Ely. The idea is revisited in the Conclusion, Chapter \ref{chap:conclusion}, of the thesis.

\chapter{Temporal VAE Suspense}

\label{chap:tdvaesuspense}

\section{Introduction}

The last chapter proposed a method for calculating \textit{suspense} and \textit{surprise} based on Ely's definitions. To recap, Ely's definition of \textit{surprise} intuitively is the difference between the expected state and the actual state. \textit{Suspense} is the expected variance of future states. The Ely model uses Bayesian expectations over final states. There are few finite end states in a tennis match or a hand in Poker, whereas, in open domain story-telling, there is a potentially unlimited number of states. So instead, the hierarchical rollout model from Chapter \ref{chap:rolloutsuspense} generates continuations with the implicit state-space learned by taking a vector distance over future continuations. This method comes with a number of limitations:

\begin{enumerate}
   \item \textbf{Computational Complexity:} The main practical issue is computational complexity. Even with modern GPUs generating a batch of $100$ or so sentences can take $30-60$ seconds. The suspense reported earlier only used a single sentence rollout. The code supports a tree rollout as a beam search over a tree. However, in practice, this was enormously computational slow for even a depth of $2$ or $3$ sentence continuations and didn't make much difference to results. This complexity has several consequences:
    \begin{itemize}
        \item \textbf{Longer Works:} The \textit{WritingPrompts} evaluation was only on works of $25 - 75$ sentences. It is not feasible to use this \textit{suspense} method for really long works of fiction such as movies scripts, or novels. Calculating suspense at each sentence positions can take a minute or so with GPU acceleration or a few minutes if it is a multistep rollout. A long novel such as \textit{Great Expectations} can have $10000$ sentences and so could take a couple of weeks to calculate on a single GPU.
        \item \textbf{Local Suspense:} The most obvious consequence of this is that \textit{suspense} is localised to looking only at the next sentence, or with a tree rollout, a few sentences. This hasn't been tested because of the mentioned practicality of running on longer works. However, suppose the expected state is only a short distance ahead of a few sentences. In that case, the \textit{suspense} inferred is likely to be local to the discourse of the scene rather the plot as a whole.\footnote{This is also relevant to the later work on \textit{salience}.} This may be less important for short stories as per \textit{WritingPrompts} as necessitated by short story writing they have highly compressed plots, with fast progression. However, it is far more likely to be relevant to longer works. This approach will probably lead to identifying important inflexion points and changes of directions in dialogue, scenes, and local discourse units but not necessarily suspense for the plot as a whole.
    \end{itemize}
        \item \textbf{Plausibility:} The limitation with the approach is that GPT-2 (or another LM) needs to be able to generate plausible continuations to advance the plot. The sentence representations also need to encode meaningful semantic plot information so that vector distance corresponds with similar relative differences in the outcome of the plot; shorter vector distances represent similar plot states and longer distances bigger shifts in plot. While there are huge advantages brought forward by big LMs, as reviewed earlier in the thesis, there are clear limitations with existing models in terms of grounding, casualty, etc.
\end{enumerate}

Plausibility depends on understanding the plot and how the story would develop, and how characters perceive events. The Ely model of suspense depends on a posterior distribution over possible outcomes. So it depends on the model being able to generate plausible alternatives in the way a human reader would expect them to happen. This also requires some degree of comprehension of how characters would act in a given situation. This overlaps heavily with the challenges of story generation and more complicated comprehension tasks such as natural question answering that require reasoning over multiple passages (or multihop QA). As well as the background material already reviewed, relevant ideas from this literature will be discussed in future work.
This chapter is concerned with an alternative model that looks to tackle the first problem, computational complexity. The definition of \textit{suspense} given by Ely when used with an L2 distance is variance over future states; in the model the encoded sentence embeddings from the generated continuations represented these states. What is needed is a possibility to generate alternative plausible narrative paths. To return to the examples. At the beginning of \textit{Great Expectations} when \textit{Pip} is grabbed by as yet unnamed \textit{Magwitch} different possibilities arise: That \textit{Pip} will be badly hurt; that \textit{Pip} will be killed; he helps \textit{Magwitch}; betrays \textit{Magwitch} to the authorities; a third party intervenes, etc. The key to strong performance with the Ely model is being able to infer these continuations and estimate a probability distribution of these scenarios. As per the discussion in the last chapter, the scenarios don't need to be exhaustive. For example, it's also possible to imagine \textit{Magwitch} or the church getting struck by lightning in this situation. The main thing is to tell when there's likely to be more or fewer consequential divergent paths. Generating more divergent outcomes in situations of uncertainty or jeopardy means higher Ely suspense, and vice versa, in a highly predictable and non-consequential situation, less suspense.

\begin{figure}[htbp]
\centering
\includegraphics[trim=0 140 0 0,clip,width=0.80\textwidth]{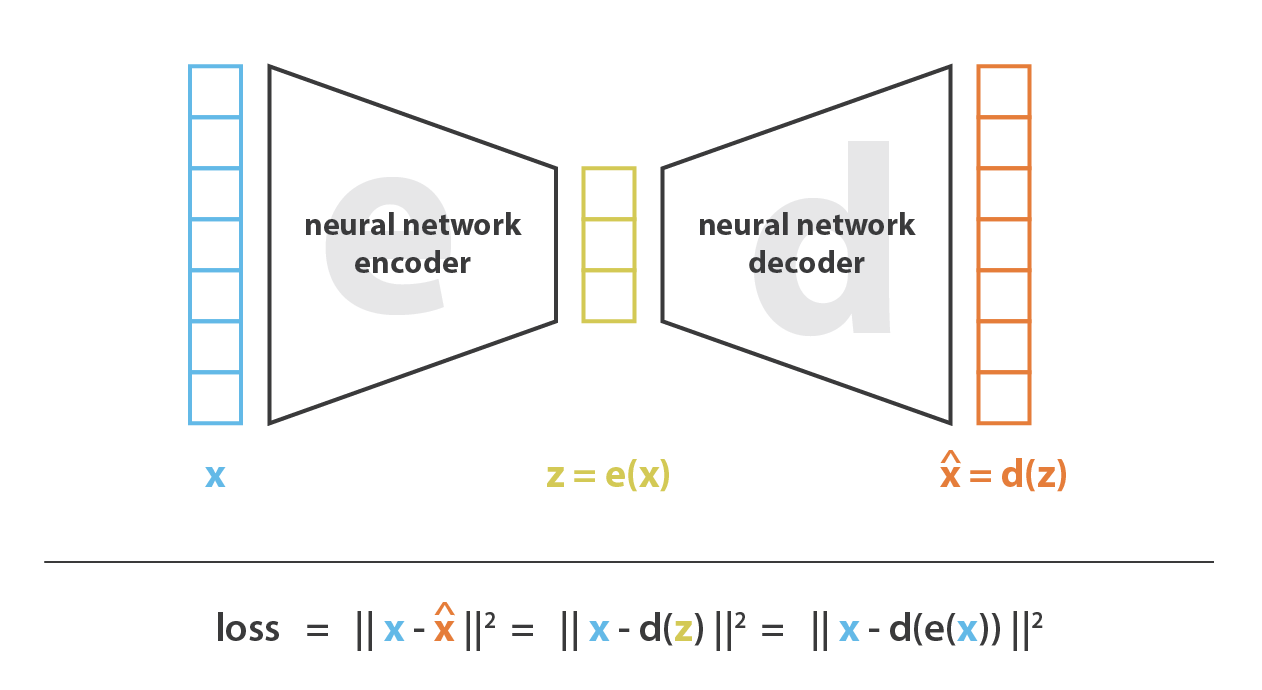}
\caption{Simplified VAE architecture reproduced from \citet{rocca_2021}.}
 \label{fig:vae}
\end{figure}

The model proposed in this thesis is based on VAEs (Variational Auto-encoders; \citealt{DBLP:journals/corr/KingmaW13}). VAEs are a variational Bayesian generative ML method. Figure \ref{fig:vae} illustrates the basic architecture. There are two main, usually multilayer, feedforward networks: An \textit{encoder} and \textit{decoder}. The encoder takes a given input $x$ that can be any vector input and map it to a latent space. It is normal to project into a lower-dimensional space and reduce the number of dimensions; this is a form of \textit{lossy compression}. The intuition behind doing this is similar to a more traditional dimensionality reduction technique such as PCA (Principal Component Analysis; \citealt{doi:10.1080/14786440109462720}). The motive is first to disentangle the noise from the signal, the independent variable from the dependent ones, while preserving as much underlying structure in the representation. The decoder reconstructs the original input to reconstruct the original $x$. The aim is to minimise the difference between the original $x$ and the reconstructed $\hat{x}$. In terms of the number of dimensions, there is often a tradeoff between minimising the reconstruction loss and the desired compression.

\begin{figure}[htbp]
\centering
\includegraphics[width=0.80\textwidth]{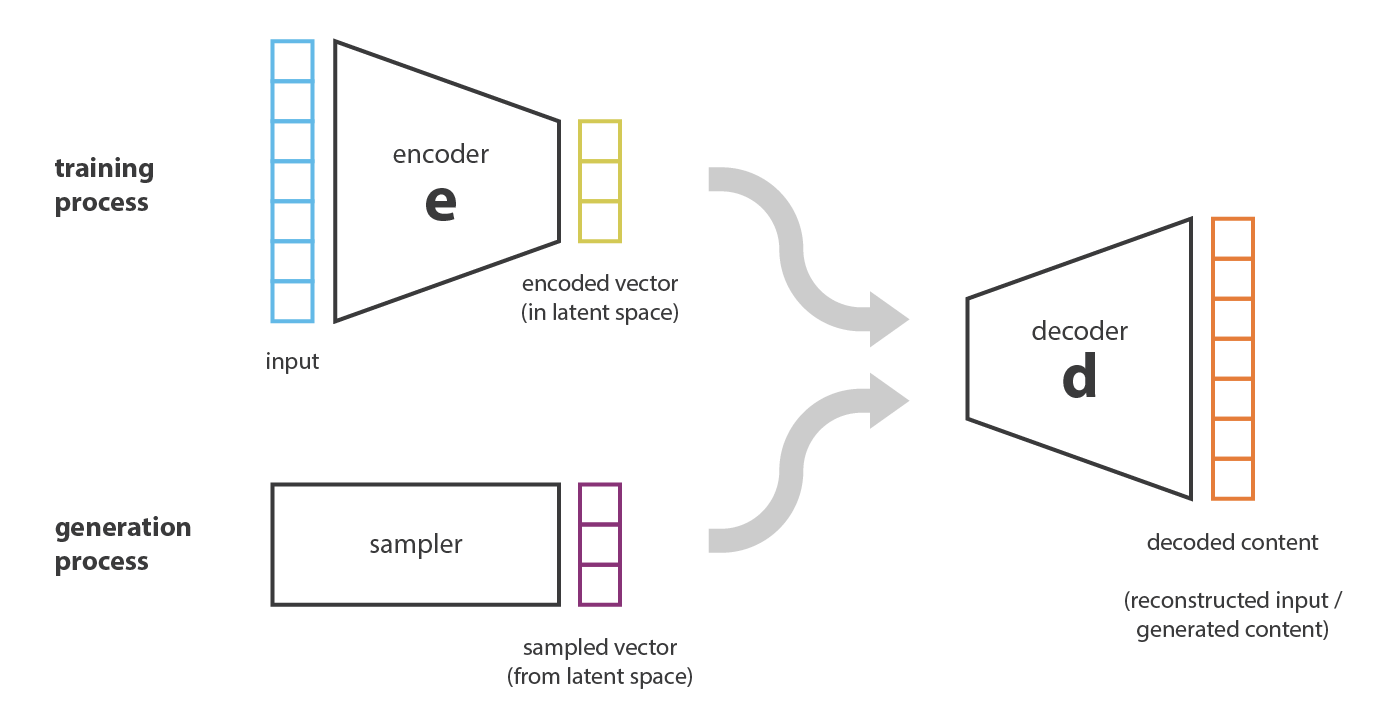}
\caption{The VAE architecture illustrating the separate encoder and sampling path, reproduced from \citet{rocca_2021}.}
 \label{fig:vaesampler}
\end{figure}

Everything described so far is identical to a conventional autoencoder \citep{Hinton2011TransformingA}. The goal is to learn an encoder and decoder that project onto a lower-dimensional and reconstruct the input. There are many variants, such as denoising autoencoders \citep{lu2013speech} that try to improve the generalisability of representations by corrupting (or adding noise) to the input so the autoencoder. The noise aids in being able to separate the noise from the signal in the learn representations. The primary difference between a conventional autoencoder and a VAE, illustrated in Figure \ref{fig:vaesampler}, is there is a separate path for sampling that excludes the encoder. In a VAE  the latent vector represents pairs of means $\mu$ and variances $\rho$ in a Gaussian distribution. \footnote{Other distributions are possible as well but it is standard to use Gaussians} When sampling directly from these dimensions with a standard Gaussian $z \sim \mu(0,1)$, decoder weights will transform into a representation approximating the input $x$. Initially most practical work for VAEs was image-based so that the output might be digits, human faces, or Imagenet classes depending on the dataset and task. The method should apply to any data that can be represented in vector space. The model is creating examples of the original input space $x$. The model can be used to efficiently compute a MAP (Maximum a Posteriori) estimate for the parameters of the model $ \theta $, estimate the hidden variables $z$ given an input $x$, or probabilistically marginalise over the input $x$.  Efficient estimation is key since mathematically marginalising over the posterior distribution is computationally intractable, and Monte Carlo methods are typically too slow as each output sampled requires a loop. The relevance, especially to \textit{suspense}, is that Ely's definition is a variance over future expectations. It is posterior over the expected future state.  Sampling from a VAE gives a similar posterior. The vanilla VAE is not suitable since the sampling can only directly sample a posterior over the input space. Later temporal extensions are introduced in this chapter that rely on VAE concepts as a foundation. The key concepts are the same for the thesis, though, in that the VAE provides a variational Bayes method for approximating a posterior distribution that aligns with ideas from Ely. The second issue is that, unlike an image where the input can be a flattened representation of the image pixels, an input representation for inferring suspense has to be semantically meaningful for the plot and represented in the vector space. This is also discussed later in the chapter. Before moving on to conditional and temporal models, I will briefly review how these more simple VAE models are trained.

Mathematically if we describe $X$ as observed data, $Z$ as an unobserved variable, which has been defined in the VAE has correspond to the input and latent representations. $Q$ is the approximated distribution over $Z$, which is optimised to approximate the true posterior $P(\mathbf{Z}|\mathbf{X})$. Our model produces $Q$, which is learnable, whereas $P$ intractable and unknown.  The ELBO (Evidence Lower Bound), eqn. \ref{eqn:elbo}, where $H$ is the entropy and $H(Q;P(X,Z))$ is the cross-entropy. This derivation, taken from \citet{yang2017understanding} is shown in eqn. \ref{eqn:elboderivation}.

\begin{equation}
\ell(X)=H(Q)-H(Q;P(X,Z))=-\sum _{\mathbf {Z} }Q(\mathbf {Z} )\log Q(\mathbf {Z} )+\sum _{\mathbf {Z} }Q(\mathbf {Z} )\log P(\mathbf {Z} ,\mathbf {X} )
\label{eqn:elbo}
\end{equation}

\begin{equation}
\begin{aligned}D_{\mathrm {KL} }(Q\parallel P(Z|X))&=\sum _{\mathbf {Z} }Q(\mathbf {Z} )[\log {\frac {Q(\mathbf {Z} )P(\mathbf {X} )}{P(\mathbf {Z} ,\mathbf {X} )}}]\\D_{\mathrm {KL} }(Q\parallel P)&=\sum _{\mathbf {Z} }Q(\mathbf {Z} )[\log {\frac {Q(\mathbf {Z} )}{P(\mathbf {Z} ,\mathbf {X} )}}+\log P(\mathbf {X} )]\\D_{\mathrm {KL} }(Q\parallel P)&=\sum _{\mathbf {Z} }Q(\mathbf {Z} )\log Q(\mathbf {Z} )-\sum _{\mathbf {Z} }Q(\mathbf {Z} )\log P(\mathbf {Z} ,\mathbf {X} )+\log P(\mathbf {X} )\\\log P(\mathbf {X} )-D_{\mathrm {KL} }(Q\parallel P)&=\sum _{\mathbf {Z} }Q(\mathbf {Z} )\log P(\mathbf {Z} ,\mathbf {X} )-\sum _{\mathbf {Z} }Q(\mathbf {Z} )\log Q(\mathbf {Z} )=\ell(X)\end{aligned}
\label{eqn:elboderivation}
\end{equation}

Maximising the lower bound of these functions minimises $D_{\mathrm{KL}} (Q\parallel P)$. $D_{\mathrm {KL}}(Q\parallel P)$ is the \textit{Kullback-Leibler} divergence between the true posterior and the estimated one. So minimising the difference between the two will make the model generate posterior distributions closer to the true one and so learn to approximate the true posterior distribution from the data. Note that this form is preferred over $D_{\mathrm{KL}(P \parallel Q)}$ as this has a behaviour called \textit{zero-avoiding}, that will lead to training $Q$ spreading out its probability mass to cover non-zero probabilities of $P$. In practice, this spreads the probability mass too thinly over low probability points. In contrast $ D_{\mathrm{KL}(Q \parallel P)} $ encourages \textit{zero-forcing} as it encourages the probability mass of $Q$ to be zero where $P$ is zero. Metaphorically, $D_{\mathrm {KL}(Q\parallel P)}$ concentrates the mass on islands of probability, leaving gaps; this is desirable with ML applications since we typically want to model the main stereotypical classes of data well rather than badly fitting all of the data. When this is parameterized as a loss for a VAE, see eqn. \ref{eqn:vaeloss}, for a single data point and dimension, where $\theta$ are the parameters of the model. As is typical for training, the loss can be averaged or summed across all dimensions for a minibatch.

\begin{myequation}
 \ell({\theta})= -E_{\mathbf {z} \sim q(\mathbf {z|x} )}(\log(p_{\theta }(\mathbf {x\mid z} )))+D_{KL}(q(\mathbf {z\mid x} )\parallel p_{\theta }(\mathbf {z} ))
\label{eqn:vaeloss}
\end{myequation}

The two parts to the loss can be thought of as a \textit{reconstruction loss} and a \textit{regularization loss}. The first part is a binary-cross entropy reconstruction loss that pushes the reconstructed $\hat{x}$ data points to be as similar to the input. The second loss functions as a regularizer; it pushes the distribution to be a Gaussian prior, which keeps the distribution around $0$ with datapoints smoothly distributed around this. Unlike a regular autoencoder, the Bayesian properties of a VAE enforce a distribution that also applies when sampling over the posterior. Though not presented in this formulation, both sides can be weighted by parameters to prioritize one loss component. For the purposes of modelling Ely, this is desirable as we want the sampled distribution to match the Bayesian prior. This section has only briefly reviewed the VAE, \citet{Doersch2016TutorialOV} review the mathematics and principles more thoroughly. It has also touched on only the base VAEs where there are many variants and extensions. Some of these are discussed later as possible improvements concerning future work. 

One final relevant note is that although VAE can be powerful, it is complicated to train effectively.  VAEs like other similar generative models can be prone to posterior collapse \citep{Lucas2019UnderstandingPC}. Posterior collapse can be described simply when the posterior is trained to be an uninformative prior that ignores the latent variables $z$. The consequence of this is the reconstructed $\hat{x}$ can become almost independent and reflects a generic and not the real one, which undermines the predictive power of the model. Posterior collapse has been linked to the weak decoders or noisy inputs where the latent space has been unable to learn to input well.  \citet{DBLP:conf/nips/LucasTG019}  find that is not, as has previously been suggested, down to the ELBO loss but instead comes from spurious local minima that prevent the latent space learning meaningful state. \citet{dai2020the} agrees with this conclusion and suggests the problem lies more directly with the loss surfaces of deep autoencoder network rather than inherently the variational bayes aspect of the model. The intuitive explanation is that the VAE becomes stuck, cannot use the latent space to reconstruct the input with a reduced error, and then starts to ignore the latent space. The topic is introduced not for a theoretical discussion but to highlight that VAE models are both powerful and difficult to train. This has practical implications in terms of setup for more complicated temporal variational models to follow. It is also is relevant to discussing the experimental results from this suspense chapter and the following one on generation in terms of possible causes for negative results and pointing a direction to future work.

\section{Related Generative Architectures}

There have been extensions to the VAE that have tried to improve the architecture such as $\beta$-VAE \citep{DBLP:conf/iclr/HigginsMPBGBML17} that reformulates the VAE to disentangle the latent space better but are still fundamentally the same concept. CVAEs (Conditional Variational Autoencoder, \citealt{DBLP:conf/nips/SohnLY15}) are more controllable and condition on a latent vector. A learnt latent vector usually from a codebook is concatenated with the latent representation and conditioned on by the decoder. The original work was image-focused, the CVAE could condition on a simple digit or a more complex class of objects such as \textit{Dog}, \textit{Boat}, or \textit{Lamp}. For example in TGVAE (Topic Guided VAE, \citealt{wang-etal-2019-topic}) where news topics such as \textit{crime} or movie genres such as \textit{horror} can be conditioned on to generate text from a prompt guided and specific to the conditioned topic. \citet{li-etal-2018-generating-classical} trains a CVAE model that instead learns an encoding for a title of a Chinese poem. The title encodings are concatenated with the VAE with the intuition that the generated poem will be more closely based on the title. The implementation is different from this thesis, but the principle idea is the same both for modelling \textit{suspense} and text generation in the following chapter. We don't just want to sample of a Bayesian posterior from a space of possible stories. Instead, at each stage, the models need to generate this distribution conditional on the story up until that point.

\begin{figure}[htbp]
\centering
\includegraphics[width=0.95\textwidth]{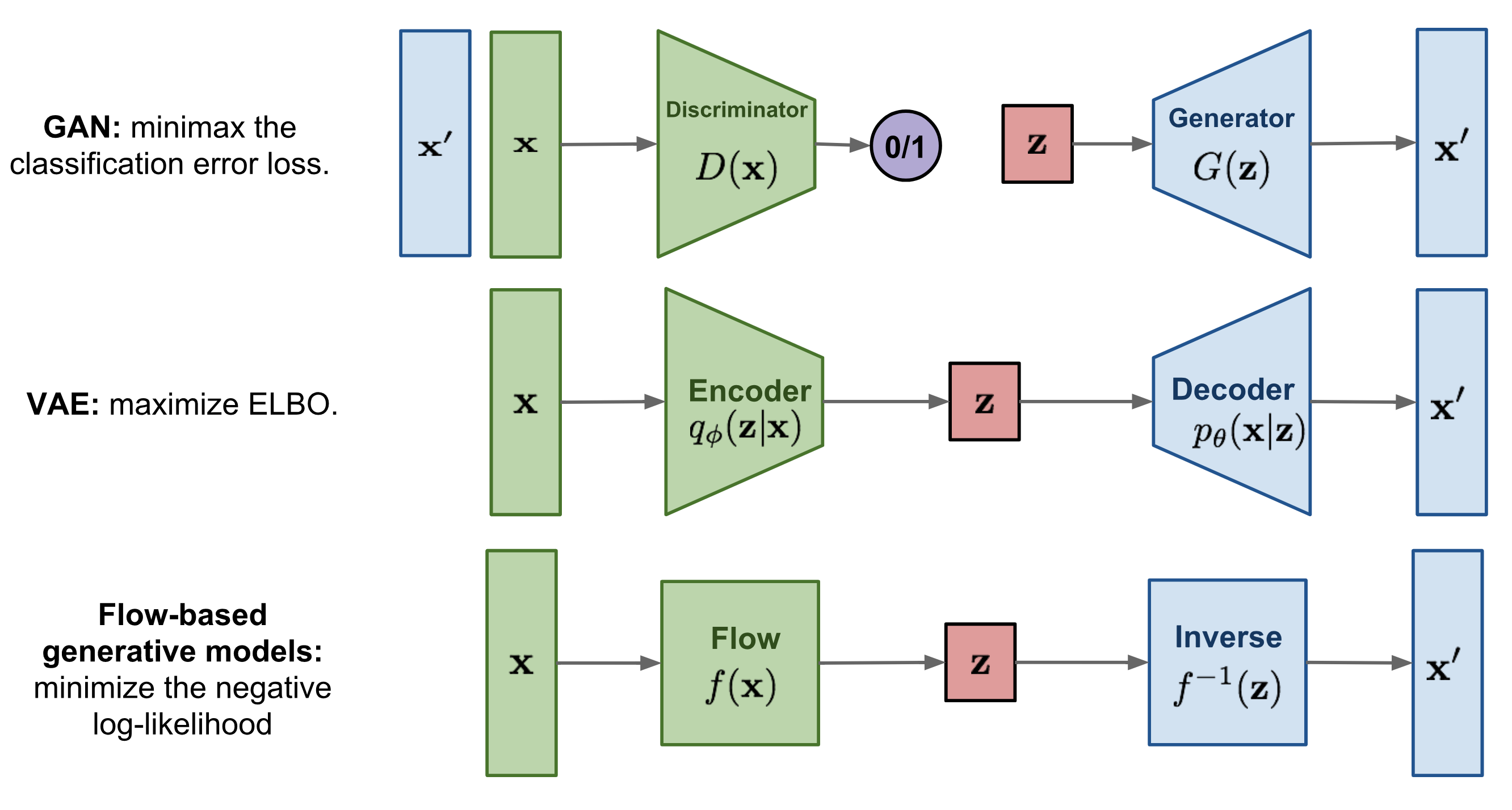}
\caption{A comparison of three recent neural generative model architectures reproduced from \citet{weng2018flow}.}
 \label{fig:threegenerativemodels}
\end{figure}

Before moving on to discussing state-space models and the specific details of the model used both for the suspense work in this chapter and following one is it is worth touching on three recent neural generative models for comparison. Figure \ref{fig:threegenerativemodels} illustrates the three architectures. So far the review has focused on the middle architecture VAEs. GAN (Generative Adversarial Network, \citealt{DBLP:conf/nips/GoodfellowPMXWOCB14}) attracted a lot of attention after outperforming VAEs again largely on computer vision-related tasks. The primary difference is that instead of an encoder, the model has a discriminator. The generator will sample from a latent space like the VAE decoder. Learning isn't via reconstructing the original $x$. Rather the generator learns by trying to fool the discriminator. In turn, the discriminator learns to predict the correct examples from the incorrect examples. Each then gradually learns to iteratively to become better at generating $x$ examples or telling the fake from the real data point. Adversarial training is not mutually incompatible with VAEs as per VAE-GAN \citep{DBLP:conf/icml/LarsenSLW16} where adversarial training can be incorporated into a VAE architecture. GANs can also be difficult to train if there is an imbalance in either's capability; if one has a too easy a task to fool or detect the other then the model can fail to train. More principally, the GAN doesn't have the Bayesian inference property desirable to default in the generative models so it is a less desirable architecture for modelling Ely.

The third architecture are \textit{normalizing flows} \citep{weng2018flow,DBLP:journals/pami/KobyzevPB21}. The issue with VAEs is that they rely on the underlying prior of a Gaussian distribution. In reality, lots of real-world problems do not fit this distribution. The intuition behind normalizing flows is by stacking a series of invertible functions with trainable weights. Stacking layers allows each layer to modify the distribution fed into the lower layers and hence allows the modelling of more complex distrubutions. The inverse (decoder), as the name suggests, can be the inverse of the flow that creates the latent space $z$. The approach has several advantages: It can be more flexible in modelling a distribution than a predefined distribution such as a Gaussian. The training loss is much simpler as it just becomes the negative log-likelihood of the reconstructed $x$. Like a VAE $x$ examples can be generated by sampling from the latent space with only the inverse transformations. One limitation is that the functions need to be relatively limited since it needs to be feasible to calculate the inverse function. Flow-based models tend to be computationally expensive relative to VAE's, especially during training. RealNVP \citep{DBLP:conf/iclr/DinhSB17}, NICE \citep{DBLP:journals/corr/DinhKB14}, and Glow \citep{DBLP:conf/nips/KingmaD18}, in order of developments, have all been proven powerful in learning normalised distributions. As per the GANs, Normalised Flows aren't necessarily mutually exclusive to VAEs, F-VAEs \citep{DBLP:journals/corr/abs-1809-05861} is a model that attempts to incorporate conditional flows into VAEs, but this is still a developing area. \citet{DBLP:journals/corr/abs-2006-00866} show that they can learn to be directed Bayesian networks. Each layer of transformation entangles the probability distribution further and allows it to represent a graphical structure between latent variables. Just like VAEs so far, the flow models discussed have only been about learning a distribution over the original space $x$. The next section touches on the state-space models for modelling suspense that are subsequently also used in the next chapter for text generation. In this context, autoregressive normalised flow models are touched on for comparison. The following chapter also discusses normalising flow models concerning future work and overcoming the weaknesses in the experimental results for the TD-VAE model.

\section{State Space and Autoregressive Models}

To recap, various generative models have been reviewed, including the focus of this and the next chapter, which is VAE based methods. VAEs reconstruct $x$ examples from the projected latent space $z$. Either can be used to generate $x$ by sampling from the latent $z$, or projecting from $x$ into a latent space $z$. As the $z$ space usually has a reduced dimensionality, the advantage is it allows the model to learn a compressed representation that best represents the data structure. As noted, though, these models only represent data in the original space. A generative model for Ely needs to be conditionally autoregressive to infer forward to future timesteps based on previous ones. For example, if each $x$ is a sentence, then the model should predict the following sentence $t$ as conditionally $(x_{t+1} | x_{t}, x_{t-1}, x_{t-2}, ... )$ . Ideally as well the model should be represented as a latent space model, not in the observation space; this is referred to as a \textit{state-space} model \citep{koller2009probabilistic}. We don't simply want just directly to infer from an input domain $x_t$ to a future domain $x_{t+1}$ because any input of events or sentences in the story domain is highly complicated. Just as in a VAE, the model should compress the space to capture the most salient features for the task in a reduced number of dimensions. By removing the noise, the task of inferring forward to $x+{t+1}$ should be more straightforward. The goal would be to have latent representations representing the most pertinent plot points encoded into latent vectors $z$, not the irrelevant details. Also, the rationale of this approach was to computationally simplify the approach from having to generate a tree of concrete alternatives, so reducing computational complexity is a requirement of any alternative model. To sum up the desired requirements of the model are:

\begin{itemize}
  \item \textbf{Bayesian:} Aligns closely with the Bayesian notions of expectations over states and is able to sample a distribution. The relevance of being Bayesian is the original formulation of Ely expectations is over Bayesian outcomes, and so the work in Chapter \ref{chap:rolloutsuspense} was a simplification of this requirement, and so VAE is in principle closer to Ely's proposed methods.
  \item \textbf{Conditionally Autoregressive:} Need to infer a future state, conditioned on the story up until that point. 
 \item \textbf{Latent Space:} Works on reduced dimensionality space, a state-space model.
  \item \textbf{Reduced Complexity:} Must have reduced computational complexity and be more scalable than the tree generation method.
\item \textbf{Longer Range:} The tree generation method only work, for the number of sentences generated ahead, which, as noted, is only feasibly 1-3 even for shorter works. Ideally, the model would be able to more easily project further into the future as events further ahead are more likely to lead to suspenseful divergences of the outcome.
\end{itemize}

Within the literature, there are numerous autoregressive generative models: MADE \citep{pmlr-v37-germain15} is an early autoregressive approach that made it more feasible by employing binary masks to simplify the ordering and hence make it more efficient. 
Wavenet \citep{DBLP:conf/ssw/OordDZSVGKSK16} is a model specifically for the autoregressive prediction of audio. PixelRNN \citep{DBLP:conf/icml/OordKK16} as the name suggests is for autoregressively predicting successive frames of images which can include filling in masked areas. \citet{DBLP:conf/iclr/ChiappaRWM17} model interactive environments such as Atari games. 
There are also more general normalized autoregressive flow models such as IAF \citep{DBLP:conf/nips/KingmaSJCCSW16} and MAF \citep{DBLP:conf/nips/PapamakariosMP17} that demonstrate strong performance in conditional density estimation tasks. As these are autoregressive models, density estimation of the posterior naturally corresponds to Ely's expectations over future states if the vector space represents the state as per the last chapter. As far as the criteria go they only meet the first two of being conditionally autoregressive and Bayesian. None of them is explicitly Bayesian, although, as noted earlier, flow-based models can be thought of in these terms. These models also work directly in the observation space and not an abstracted latent space and, as a result, are often computationally complex. Mapping a higher dimensional space is inevitably more complicated than work in an abstracted and reduced dimensionality space. Also, these models are only capable of predicting the next $x_{t+1}$. This means that if the model wanted to infer 5 sentences ahead it would need to infer iteratively $x_{t+1}$, $x_{t+2}$, $x_{t+3}$, $x_{t+4}$, $x_{t+5}$. While this would be cheaper than generating sets of concrete sentences with GPT-2 as the approach would require sampling, it would still be relatively expensive. The focus for a lot of these models in the literature has been video or interactive game environments where this process of reconstructing frames iteratively is similarly expensive. The other problem with this step by step rollout is that errors in the rollout process will be magnified at each step, degrading the generation's quality.

 A latent space model that predicts autoregressively directly in the latent space rather than the observation space has the benefit of reduced complexity because they can use simplified representations focused on the most salient features of the task. The model in the thesis is TD-VAE \citep{gregor2018temporal}, both for this chapter on suspense and surprise and the following, which conditions on the model for story generation as part of a planning process. There are alternative latent state-space autoregressive models \citep{DBLP:journals/corr/Graves13,DBLP:journals/corr/ChungKDGCB15,DBLP:journals/corr/abs-1804-01523,Doerr2018ProbabilisticRS,DBLP:conf/nips/HaS18}. These models while sharing much in common with TD-VAE only allow generation of the next $x$. The key differentiator is that TD-VAE supports \textit{jumpy} which means it can directly infer several steps ahead while inferring the steps in-between. All of the cited papers are in the video and simulated environment frame. This potentially removes the rollout problem where each timestep needs to be calculated iteratively, which is slow and can magnify errors. The advantage in a simulated environment like a game is obvious as time is essential; the agent needs to take action and not wait for a rollout. In \textit{suspense} the intuition is that the model could sample directly up to say 10 sentences ahead in the latent space, and these samples would act as the same variance of expectations as the concrete sentence generated in the last chapter. In theory, this would allow suspense to be calculated further than possible with the tree generation approach. The inference would also be much cheaper because of the \textit{jumping}, the latent state-space, and that it as a VAE model is computationally simple to infer with compared to some other generative models. The \textit{jumpy} model would be optional since the model also supports step-by-step rollout. The model is elaborated further in the section, but as of the time this work was completed, it was the only model that met all the desired requirements to align with Ely. The model and forward-looking suspense can also model surprise by, as per the previous chapter, comparing expectations of the future state with the actual state at time $t$.

The TD-VAE model is not unique in making \textit{jumpy} predictions. Because the autoregressive is a known issue with a wide range of especially video related and in environment simulation reinforcement learning tasks. The state of the art in this area is Clockwork VAE \citep{DBLP:journals/corr/abs-2102-09532}. It should be noted that Clockwork VAE is more recent than the work done on suspense with TD-VAE, though. It, represents a different approach than the one employed to TD-VAE. TD-VAE makes predictions over gaps directly. The Clockwork VAE instead uses a hierarchy of VAE layer and latent representations that sample from each other. Each layer is on a different time cadence each layer and skips over a a decreasing number of steps. Hence, in video prediction, the intuition is that the lower layers will learn expected bigger changes while the top levels will learn finer frame-to-frame changes. The consequences are that it allows better overall predictions by having the different layers of inferring different levels of changes while allowing \textit{skipping} over intermediate frames; jumping over multiple frames at a time by only using the VAE on the required cadence. Clockwork VAE builds on similar ideas as, for example, VideoFlow \citep{DBLP:conf/iclr/KumarBEFLDK20} and \citet{DBLP:conf/nips/KimAB19}. This alternative approach is worth exploring for comprehension tasks such as in this chapter and as a planning mechanism for a generation as per the next chapter. However, the TD-VAE is preferred at least for initial work because it is simpler. Also, in a video or game environment, it is more straightforward to pick out meaningful time intervals for each layer. It doesn't seem as picking out every $n$ sentence makes much sense in the story domain. In stories, as is covered in the theory from Chapter \ref{chap:backgroundtheory}, stories can be thought of simply in terms of more \textit{salient} \textit{nuclei} that drive forward the plot and \textit{satellites} that fill in the details. Predefined regular intervals won't match \textit{nuclei}. The idea is revisited in the future work section of Chapter \ref{chap:conclusion}.

One short note is that all of the source models haven't been developed in the NLP domain. VAE and generational models such as the earlier example have been used for text generation. However, the driving force behind much of the recent development of these more advanced models has been in the vision and simulated environment space. The use of such models in NLP has been limited. As of the time the suspense work in this chapter and the generation work in the next chapter was completed, this approach was pretty novel. This is surprising because these models were developed to deal with long-term predictions in complex domains via latent abstract states. As reviewed in Chapter \ref{chap:backgroundml}, more recent LMs and sentence or longer form vector representation have been shown to represent a wide range of semantic properties well and transfer successfully to a range of tasks. So it would seem natural to combine the two and use these latent generative models with representations from \textit{transformers} and apply them to more complex tasks such as suspense inference or planning for text generation or other similar tasks. As will be discussed in Chapter \ref{chap:tdvaegeneration}, the more common approach has been to use semantic role labelling or keyphrase extraction to pick out the most salient information to represent events ahead of time. Then these simplified representations can be fed into neural models. Unlike latent space models such as TD-VAE, these models need to decide on a fixed representation apriori and cannot learn one dynamically for the task and domain.

The technical detail will follow in the next section but briefly to tie together the TD-VAE model with the Ely concept of \textit{suspense} and \textit{suprise}. The original Ely concept is one of Bayesian expectation over outcomes over the future state for \textit{suspense}. With \textit{surprise} it is the reverse, and it is an expectation from a previous time period versus the real outcome. Rather than take the concrete states from hypothetical future GPT-2 continuations, the approach is to use the TD-VAE to sample future states. As TD-VAE is Bayesian, this will aligns with both Ely and is comparable with the GPT-2 generation method. It also, like the GPT-2 generation method, works directly on the latent space to represent outcomes. The theoretical advantages are that it is mathematically much simpler to sample from TD-VAE. The \textit{jumpy} predictions should make it easier to calculate \textit{suspense} a few sentences ahead. TD-VAE and similar state-space models have demonstrated strong performance in other complicated temporal domains and so may prove powerful in inferring future latent states and, hence, directly improve the GPT-2 generation method.

The following technical constraints are relevant to applying the TD-VAE model:

\begin{itemize}
  \item \textbf{Meaningful Sentence Representations:} The TD-VAE model works in a latent semantic space. Unlike, say, with an image with raw pixels as an obvious input, there is no direct vector, so sentences or events need to be encoded into semantically rich representation as an input to the TD-VAE model. Note that the input for TD-VAE needs to be fixed and the architecture requires that the gradient cannot be back-propagated through the input representations. This is because otherwise the TD-VAE losses can be minimised, making the sentence representations degenerate. 
\item \textbf{Plot Development:} The model needs to capture temporal shifts, which in stories is plot development. It will impact the choice of layers, training regime, etc.
\end{itemize}

\section{Methods}

\subsection{Approach}

\begin{figure}[htbp]
  \centering
  \includegraphics[trim=0 70 130 0,clip,width=1.0\textwidth]{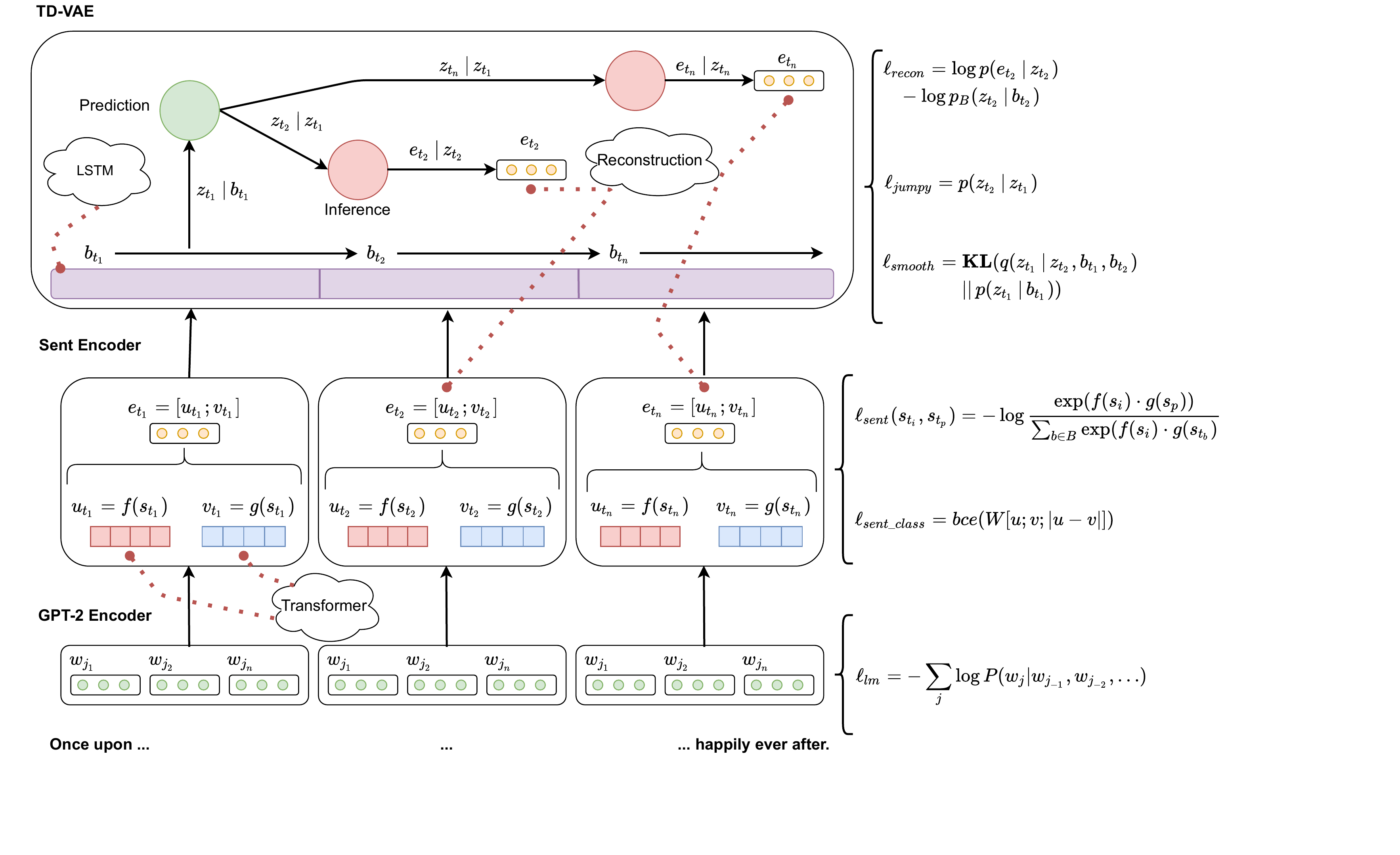}
  \caption{TD-VAE hierarchical architecture showing the multi-layer architecture: The base layer is GPT-2. From this, a sentence encoder infers sentence embeddings. The top-level is TD-VAE which learns to infer and reconstruct future sentence embedding states. It should be noted that that the sentence embeddings are concatenated when fed into the TD-VAE loss, but the respective loss maximises the dot product.}
  \label{fig:tdvaehierarchy}
\end{figure}

Our overall approach is to build a hierarchical model on a base GPT-2 \citep{radford2019language}. On top of GPT-2 is a sentence encoder based on transformers that build rich sentence representations. The top level of the model is the TD-VAE, which learns to draw inference between the sentence time steps in the story. Figure~\ref{fig:tdvaehierarchy} illustrates our architecture.\footnote{Code and configuration for this paper are available on Github at \url{https://github.com/dwlmt/knowledgeable-stories}.}

The justification for GPT-2 is to combine the model for inferring suspense with that of text generation. Chapter \ref{chap:tdvaegeneration} shows how the same model can be conditioned on as part of a text generation planning system with the same model. The rationale behind sharing a common model is, first, that the Ely ideas of suspense involves projecting forward latent vectors as an expectation over the future state. If a model can do this, then these future $t$ projected vectors can also be conditioned on for text generation, and so there is a strong conceptual overlap. The second reason is the sentence representations can benefit from GPT-2 finetuning and conditioning generation to improve the sentence representations. Third, the loss of the model is based on the one in the last chapter for sentence representations, so it provides a more direct comparison with the results of the previous chapter. Alternatively, another standard sentence model could be plugged into this sentence representation layer such as Sentence-BERT \citep{reimers-gurevych-2019-sentence}, InferSent \citep{conneau-etal-2017-supervised}, or taking a sentence representation from BERT \citep{devlin-etal-2019-bert}. However, the preferred approach is, for the reasons given, to integrate the training into an end-to-end, and to support text generation in Chapter \ref{chap:tdvaegeneration}.

\subsection{TD-VAE}

\begin{figure}[htbp]
  \centering
  \includegraphics[trim=0 0 0 0,clip,width=1.00\textwidth]{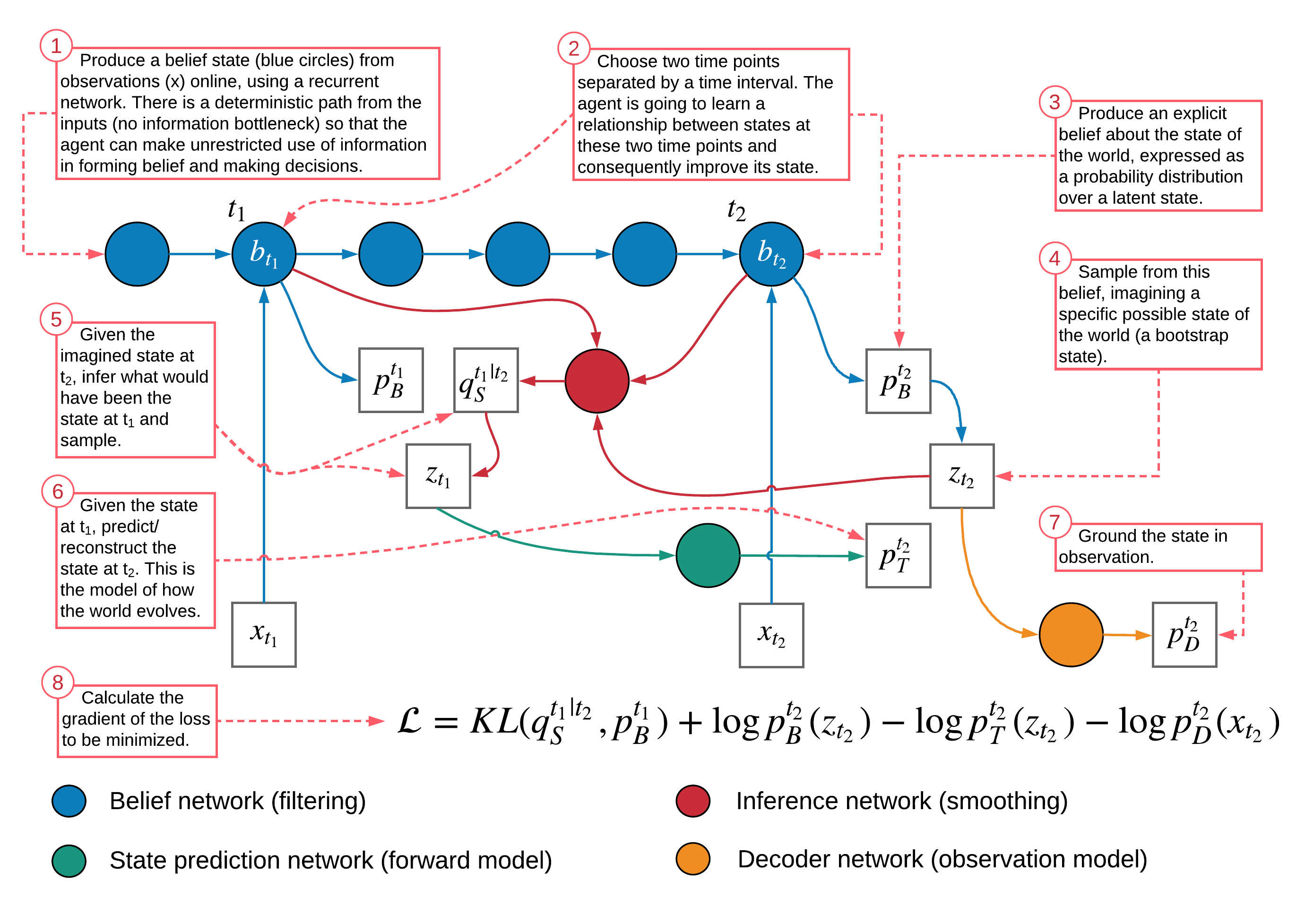}
  \caption{This is the main TD-VAE standalone architecture reproduced from \citet{gregor2018temporal}. The architecture is built on GPT-2 at the lowest layer for processing sentence inputs. On top of this is a layer for learning sentence embeddings from the lower GPT-2 layer. On top of this is the TD-VAE layer that can temporally project forward the latent states which is used in this chapter for modelling suspense, and the following for text generation.}
  \label{fig:tdvae}
\end{figure}

The top TD-VAE layer in the architecture infers a future time step $t_2$ from an earlier time step $t_1$ (see Figure~\ref{fig:tdvae}) for the main flow of the TD-VAE model. TD-VAE enables jumpy state predictions, projecting forward multiple-time steps of latent vectors from each position, conditioning from $t_1$ to $t_n$. The following describes the inference process: The input is the sentence embedding $e_t$, which encapsulates the world's state as the reader observes it. These sentence representations are compressed into beliefs $b_{t_1}$ and $b_{t_2}$ using a stacked unidirectional LSTM. These beliefs are expectations about the state of the world in the future, i.e., they model what the reader expects to happen. The difference between the beliefs and the sentence representation is that the sentence representations only represent the standalone sentence. The beliefs are the final state form the LSTM. As such they aggregate across all the sentence representations and represent an integrated state at a particular point $t$ in the story. These expectations aren't mapped directly to a future sentence state but via an abstract latent state $z_t$. The model can sample using variational inference from the latent distribution $p_B(z_{t_2}\,|\,b_{t_2})$ to guess the latent state for a future point in the story, and given the future latent state, reconstruct the sentence embedding at this future point in the story, $p(e_{t_2}\,|\,z_{t_2})$. This future state will be sampled to infer suspense, the same process in the following chapter as a planning mechanism for projecting forward state to a future time. A linear layer can them reconstruct $x_{t_2}$ from the latent state $z_{t_2}$. 

The diagram shows a single layer, but TD-VAE can be stacked (as is illustrated by Figure~\ref{fig:tdvaedeepnet}). In this hierarchical extension, each temporal VAE component is conditioned on the beliefs from the LSTM but can also sample from the VAE of the layer below. Stacking layers as per most deep learning architectures allows both the LSTM and VAE components to learn abstractions from the layers below. Thus, in theory, the higher layers should learn higher levels and more abstract concepts. The reconstructed $x$ in the original paper is a single linear neural network layer that connects the output for the VAE component back to the original latent space. In the stacked version, this linear layer then applies across all of the VAE components outputs. This means in practice if there are multiple layers, the latent layer will be larger than the reconstructed $x$. The intuition behind this is that the model can learn to draw from features from multiple layers to reconstruct the future $x$. A change from the original paper is that the model in the thesis allows multiple linear layers with different non-linear functions (the final one is always linear). The rationale for this is that there are significant number 3-6 TD-VAE layers. It is better to use a pyramid layer structure to reduce the dimensionality of the representation rather than going directly to the reduced dimensionality space. The main components of the TD-VAE model are:

\begin{figure}[htbp]
  \centering
  \includegraphics[trim=0 0 0 0,clip,width=1.0\textwidth]{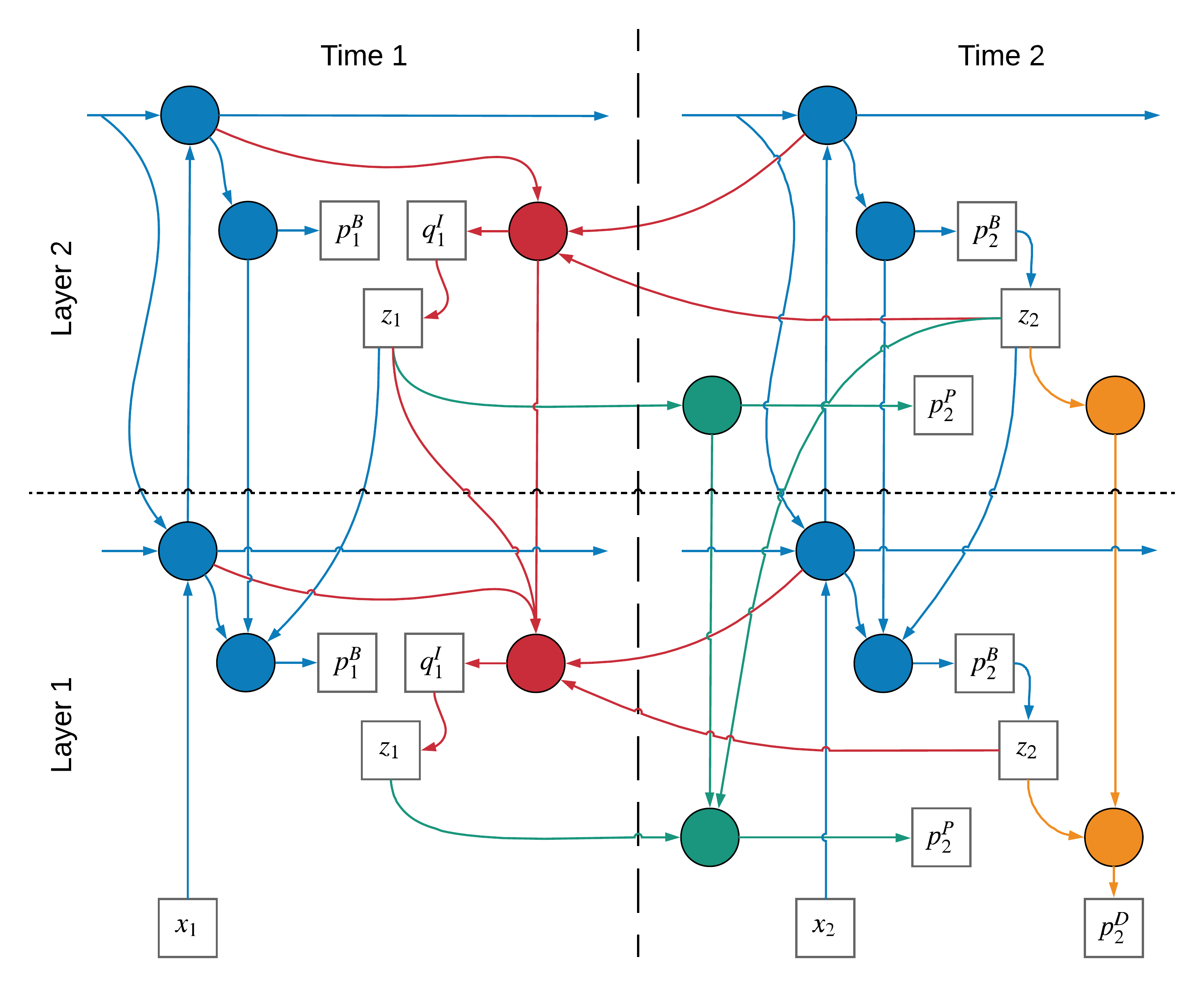}
  \caption{The hierarchical version of the model \citet{gregor2018temporal}. The main feature of note is when stacking TD-VAE layers the VAE samples in the reverse direction from the LSTM, so layer 0 will be the upper-most layer of the LSTM. The hierarchical version allows each TD-VAE receive information from the LSTM and so learn directly from the LSTM from earlier timepoints and sample from VAEs at a lower level of abstraction.}
  \label{fig:tdvaedeepnet}
\end{figure}

\begin{itemize}
  \item \textbf{Belief Network:} As stated, the belief network is just a stacked LSTM network. The input in this model are the encoded sentence vectors. The role of the belief network is to learn from up to $t$ sentences up until that point in the story. $t$ will be limited by the memory constraints of the batch size at inference time. LSTMs are perform well in time series tasks, making them a suitable choice for this component. The second sub-component of the belief network will be called the $Z$ posterior belief. The first LSTM component can only produce a fixed output latent vector. This posterior belief network uses a VAE component to create a distribution over the belief state at the current timestep.
  \item \textbf{State Prediction Network:} The state prediction component is from VAE. The architecture is vanilla with a single layer of weights transforming the input into a reduced dimension latent space and a layer of weights to transform from the latent space into the output. It is the temporal component that learns to predict future contexts from the earlier one. The job of this network is to infer a future latent state $Z_2$ from $Z_1$.
  \item \textbf{Inference Network:}  Although called the inference network, this is not used when only making forward predictions. Rather it infers backwards from what the state is at a future state to what it must have been at the earlier time point. The intuition behind this inference network (or smoothing) network is that at the later timepoint $t_2$ we know the state at the story from sampling the belief posterior because, during training, this is known. Instead, the model works backwards to estimate what the state should have been at $t_1$. This estimated $t_1$ can be used directly to train the temporal state prediction network. Separately by minimizing the divergence (the KL loss) between the belief network sampled posterior at $t_1$ and the estimate of what it should have been looking retrospectively, it will train the belief network to anticipate the future state. Likewise, the reconstruction loss can be on the actual latent state and not an estimated future state at a future time. It reuses the same VAE components as the state prediction and posterior state networks.
\item \textbf{Decoder Network:} Is the dense feedforward network that converts from the latent state $Z_{t+n}$ back into  $X_{t+n}$. This should be a straightforward mapping between the compressed latent representation and the input space. 
\end{itemize}

The training pathway which is an extended description of the descriptions in Figure~\ref{fig:tdvae}:

\begin{enumerate}
  \item \textbf{Produce Belief State:} This is just running the LSTM over the input context sentence compressed into vectors. It is called a belief network because the LSTM should, in theory, compress the context sentences into a belief about the state of the narrative up to that point. 
  \item \textbf{Choose a Time Interval: } Because TD-VAE allows \textit{jumpy} predictions it needs to pick a future $t$ to project.  \footnote{Other shorter ranges were tested and will be discussed in the training setup section.} These are the time windows the model needs to reconstruct and how far ahead to infer.
\item \textbf{Explicit Belief:} The VAE prediction network component produces a concrete state for time $t_2$, which is the future state of the LSTM output at $t_2$. It is a belief in the VAE sense that all the vector dimensions are structured as a probability distribution output between $0.0 - 1.0$, not in the sense that there are multiple $P_B^{t_2}$ latent states with a probability attached to each.
\item \textbf{Sampling Latent State:}  The model samples $Z_{t_2}$ from the $Z$ posterior belief network. 
\item \textbf{Infer Earlier Latent State:} With the inference network inferring from $Z_{t_2}$ what the earlier $Z_{t_1}$ state would have been, and sample from this state.
\item \textbf{Predict Temporal State:} The estimated $Z_{t_1}$ provides the input to predict the future target $z_{t_2}$ for the state prediction network.
\item \textbf{Ground Latent State:} Grounds the latent state at $Z_{t_2}$ by maximising it's reconstruction $x_2 | z_2$. The process is training the decoder to reconstruct vectors in the original input space from the latent state.
\item \textbf{Train Loss:} The final step is to back-propagate the loss, the equations follow.
\end{enumerate}

The process can seem complicated, but the primary rationale is to separate each LSTM, and VAE components have separate input and output targets that aren't dependent on noisy outputs from other modules. The non-training pipeline is far more simple: The input is run through the belief network LSTM, which is sampled by the posterior VAE network into the latent space $z_{t_1}$, the state prediction network then infers the future $z_{t_2}$ latent state, and the decoder network reconstructs this as $x_{t_2}$. The model can either sample multiple latent vectors through the whole process as a posterior over the whole network or reconcile a single  $z_{t_1}$ and then pass this forward through the network.

The $x$ input is redefined as $e$ from now on and in the overall architecture because TD-VAE represents the raw input. Still, in the hierarchical model, it is the middle layer representing the encoded sentence state. For the rest of the model, $x$ refers to the token sentence inputs to the GPT-2 model.

The reconstruction loss of the \textit{decoder network} is given in Figure (\ref{eqn:tdvae_recon}). It is designed to maximise the sentence embedding reconstruction probability given the latent state at that time step. The second part constrains this, a bottleneck which prevents too much of the concrete sentence state $e_{t_2}$ being encoded in the latent state $z_{t_2}$: 
\begin{myequation}
\ell_{\text{recon}} = \log p(e_{t_2}\,|\,z_{t_2})  \\
- \log p_B(z_{t_2}\,|\,b_{t_2})
\label{eqn:tdvae_recon}
\end{myequation}
To predict $t_2$, the model should estimate the state of the world at $t_1$ for $t_2$ to have happened. To do that a \textit{smoothing} distribution is calculated $q(z_{t_1}|z_{t_2}, b_{t_1}, b_{t_2})$. Eqn.  (\ref{eqn:tdvae_jumpy}) is the \textit{transition} distribution that projects the latent state forward into the future; it is the jumpy prediction. The \textit{state prediction} can be optimised by maximising the probability of the future latent state given the current one:

\begin{myequation}
\ell_{\text{jumpy}} = p(z_{t_2}\,|\,z_{t_1})
\label{eqn:tdvae_jumpy}
\end{myequation}

Finally (\ref{eqn:tdvae_smoothing}) is the KL divergence between the smoothed state and what could have been known at $t_1$. Minimising this divergence indirectly trains the \textit{belief} distribution to learn what the state of the world should be at $t_1$ to anticipate it at $t_2$:
\begin{myequation}
\ell_{\text{smooth}} = \mathbf{KL}(q(z_{t_1}\,|\,z_{t_2}, b_{t_1}, b_{t_2})\,||\,p(z_{t_1}\,|\,b_{t_1}))
\label{eqn:tdvae_smoothing}
\end{myequation}
Combining these losses produces the overall loss:
\begin{myequation}
\begin{split}
\ell({t_1, t_2}) = \mathbf{E}[
  \log p_D(e_{t_2}|z_{t_2}) \\
  + \log p_B(z_{t_1}|b_{t_1}) \\
  + \log p_T(z_{t_2}|z_{t_1}) \\
  - \log p_B(z_{t_2}|b_{t_2}) \\
- \log p_S(z_{t_1}|z_{t_2}, b_{t_1}, b_{t_2})]
\end{split}
\label{eqn:tdvae}
\end{myequation}

TD-VAE provides the backbone for generating Bayesian expectations over future states. But that depends on having meaningful sentence representations that encode the semantic information of the story. The sentence encodings build on the encodings from Chapter \ref{chap:rolloutsuspense}.

\subsection{Sentence Encoding}

In the middle layer of the architecture (see Figure~\ref{fig:tdvaehierarchy}), we encode a sentence representation for each sentence in the batch. The sentence representations are based primarily on Quick-Thoughts \citep{DBLP:conf/iclr/LogeswaranL18}, inspired by Skip-Thought \citep{NIPS2015_5950}. For the sentence representations, let there be two encoded vectors from two functions $u = f(s)$ and $v = g(s)$, with $s$ being the sequence of the word tokens representing the sentence output by GPT-2. $f(s)$ and $g(s)$ can be any functions that can convert a sequence of inputs into a single vector. Unlike Quick-Thoughts, $f(s)$ and $g(s)$ are stacked autoregressive transformers \citep{NIPS2017_3f5ee243} that use embeddings to encode position. Some experiments were tried with LSTM and GRU alternatives, as well some simple bag of words averaging tests, but all these proved inferior and so all the reported suspense results and text generation results in the following chapter use transformers. The weights are not shared between the two encoders. To produce a single vector per sentence $u$ and $v$ are concatenated across the embeddings from the GPT-2 layer.\footnote{Mean pooling was all experimented with but proved inferior and is not in any reported results.} This loss is identical to one in the previous chapter to keep consistency across comparisons. Like the previous chapters, all the non-neighbouring sentences in the batch are used as a negative example for incorrect sentences. For a batch of sentences, the loss is for a given candidate sentence; $s^{\text{cand}}$ is the negative log probability of the dot product between $f(s_{t_{2}})$ and $g(s_{t_{2}})$, normalised by all the other sentences in the batch, indexed by $i$, see~(\ref{eqn:quickthoughts_loss}). For each sentence in the batch there are two correct candidates $s_{t_{-1}}$ and $s_{t_{1}}$, given by the target position $s_{t_{p}}$ in~(\ref{eqn:quickthoughts_loss}), the previous and following sentence. 

\begin{myequation}
\begin{aligned}
\ell_{\text{sent}}(s_{t_i},s_{t_p}
) = -\log \frac{\exp(f(s_{i}) \cdot g(s_{p}))}{\sum_{b\in B} \exp(f(s_i) \cdot g(s_{t_{b}})}
\label{eqn:quickthoughts_loss}
\end{aligned}
\end{myequation}

The intuition behind this loss is it encourages vectors to be similar to their neighbours while dissimilar to sentences further away in the batch. The loss fits within the contrastive learning paradigm \citep{weng2021contrastive} which has proved successful across a wide range of machine learning domains and tasks. The training batch in this corpus is $100$ sentences per story maximum. \footnote{This will be padded if the end of the story is finished before the end of the batch.} There are $2$ stories per batch. Models such as S-BERT \citep{reimers-gurevych-2019-sentence} that rely on supervised datasets training such as NLI have been proven to perform better on general semantic benchmarks. A similar loss is an optional extension to the sentence encodings described later. However, the disadvantage of this is they lack the domain specificity of a Quick-Thoughts type loss. These losses also lack the localisation desirable with suspense or story planning. As with the reviewed background theory in Chapter \ref{chap:backgroundml}, stories can be thought of as having \textit{nuclei} or \textit{cardinal functions} as the main drivers of the plot, and \textit{satellites} or \textit{catalysers} that fill in the detail in a hierarchical structure. While the loss has no concept of hierarchy, these hierarchically related units are likely to occur together. So a loss that encourages similarity amongst neighbouring sentences and dissimilarity amongst more distant sentences is likely to capture some of this structure. The similarity between these local clusters and the structure within a story is more of a priority than a vector representation that models topics more accurately, which may be more suitable in an information retrieval context. Negative examples from the same story encourage intra-story differentiation along position, plot and discourse. Multiple stories in a batch gives the loss a story with a different theme and plot as a negative example encouraging improved inter-story differentiation in representations. If one story is romance and another is a crime thriller, it would be expected that the loss will also learn topicality.

The sentence encoders are $u = f(s)$ and $v = g(s)$. The $f$ and $g$ functions are transformers to produce an output for each input wordpiece token and not a single vector representation for $u$ or $v$. To accomplish this, pooling is used. Two methods are tried: \textit{mean-pooling} by averaging across the top layers output representations from the output of the transformer. The second is \textit{final-pooling} which takes the output of the final word piece token from the top layer of the transformer. In \textit{final-pooling} the final token acts as a class token and can learn to attend as necessary. The transformer is the standard one from Pytorch with positional embeddings and the default ReLU activation function.

As noted, though the qualities of the Quick-Thoughts loss are desirable, supervised approaches trained on NLI datasets have advantages.  InferSent \citep{conneau-etal-2017-supervised} pioneered an approach where a sentence embedding can be trained with an entailment dataset. InferSent is fed a triple consisting of a premise, a sentence entailed by the premise, and one that contradicts, it is neutral or unrelated. The ranking loss maximises the margin between the entailed and non-entailed sentences using a classifier head-on top encoders $u$ and $v$ run on the pairs of sentences. Each pair of sentences receives a similarity sore between $0.0-1.0$ via the classification head. The loss encourages entailed sentences to be close to one another in the vector space and dissimilar sentences further apart. It is a desirable loss extension for the sentence vector loss. Entailment encodes semantic property from what will be encoded by a Quick-Thoughts loss which should improve the representations' overall performance, even if the dataset's domain differs from storytelling. The model follows more closely from S-BERT \citep{reimers-gurevych-2019-sentence}, the vector is joined by concatenating the vectors thus $[u;v;|u - v|]$. Operations such as $|u - v|$ have been shown to improve performance by representing the absolute values of the difference between vectors. InferSent also concatenates $|u \cdot v|$, but this was not found helpful in S-BERT and is not part of the model in the thesis. InferSent uses a 3-way output classification for each relation type.

The loss in the thesis follows from the InferSent and S-BERT approach. On top of the concatenated vectors is a single layer linear dense head for classifying the relations between the statements as being entailed, neutral, or contradictions. NLI is trained with a standard cross-entropy loss in eqn. (\ref{eqn:crossentropy_loss}) where the class in a one-hot encoding of the relation and $x = (W[u;v;|u - v|])$, the outputs from the sentence encoders. $W$ is the dense layer $W \in \mathbf{R}^{3n \cdot k}$ where $n$ are the dimensions of $u$ and $v$, and $k$ is the number of classes, for NLI this is $3$.

\begin{myequation}
\begin{aligned}
\ell(x, class) = -\log(\frac{\exp(x[class])}{\sum_j \exp(x[j])})
                       = -x[class] + \log(\sum_j \exp(x[j]))
\label{eqn:crossentropy_loss}
\end{aligned}
\end{myequation}

The rationale is that the classification head will jointly fine-tune both encoders $f$ and $g$ to distinguish between the relation types. Thus the loss will encode entailment knowledge between the anchor text and the hypothesis, or that entailed, neutral, and contradictory pairs will be encoded differently in the semantic space. The datasets trained on are SNLI \citep{bowman-etal-2015-large} (570K pairs) and MultiNLI \citep{N18-1101} (433k pairs), examples from SNLI are in table \ref{tab:nliexamples}. The datasets are shuffled so that similar examples are not together in the batches.  

\begin{table}[]
\begin{tabular}{ccc}
\hline
{ \textbf{Text}}                                                                                                 & { \textbf{Judgments}} & { \textbf{Hypothesis}}                                                                                              \\ \hline
{ \begin{tabular}[c]{@{}c@{}}A man inspects the uniform of a \\ figure in some East Asian country.\end{tabular}} & { contradiction}      & { The man is sleeping.}                                                                                             \\ \hline
{ An older and younger man smiling.}                                                                             & { neutral}            & { \begin{tabular}[c]{@{}c@{}}Two men are smiling and \\ laughing at the cats \\ playing on the floor.\end{tabular}} \\ \hline
{ \begin{tabular}[c]{@{}c@{}}A black race car starts up \\ in front of a crowd of people.\end{tabular}}          & { contradiction}      & { \begin{tabular}[c]{@{}c@{}}A man is driving \\ down a lonely road.\end{tabular}}                                  \\ \hline
{ \begin{tabular}[c]{@{}c@{}}A soccer game with \\ multiple males playing.\end{tabular}}                         & { entailment}         & { \begin{tabular}[c]{@{}c@{}}Some men are \\ playing a sport.\end{tabular}}                                         \\ \hline
{ \begin{tabular}[c]{@{}c@{}}A smiling costumed woman\\  is holding an umbrella.\end{tabular}}                   & { neutral}            & { \begin{tabular}[c]{@{}c@{}}A happy woman in a fairy\\  costume holds an umbrella.\end{tabular}}                   \\ \hline
\end{tabular}
\caption{Examples of NLI relations from \citet{bowman-etal-2015-large}.}
\label{tab:nliexamples}
\end{table}

NLI is about the formal entailment of one statement to another. It differs from some of the more recent common sense datasets that judge whether one statement entails or contradicts another is also about human motivations or social consequences. Table  \ref{tab:atomicexamples} gives an example of what happens when someone is sent to prison. Everyday interactions from the ATOMIC such as \textit{PersonX puts \_ in the fridge}, \textit{PersonX puts arm around PersonY's shoulder}, or \textit{PersonX quits smoking} contain unspecified causes (both social and physical), attributes and consequences. Typically these are implicit in the texts and are therefore absent for typical LMs.  Storytelling is about these everyday events. So the rationale for training on a common-sense dataset is that by training on these largely missing factors that are often absent from stories, the representations will learn to encode them, which will help with both comprehension and generation of stories. It should be noted that though this model is only trained on the original ATOMIC, the updated version ATOMIC2020 \citep{Hwang2021COMETATOMIC2O}, and GLUCOSE \citep{mostafazadeh-etal-2020-glucose} which is specifically for storytelling, both released after work on this model, would also be relevant and could be treated in the same way.

\begin{table}[]
\begin{tabular}{lc}
 \hline
\multicolumn{2}{c}{\textbf{PersonX puts PersonY in prison}}                                                                                                                   \\ \hline
\multicolumn{2}{l}{\textbf{Causes PersonX}}                                                                                                                                   \\ \hline
{Because PersonX wanted (xIntent):}    & {\begin{tabular}[c]{@{}c@{}}to maintain law and order; \\ to punish him; to punish person y;\\  to realize person y mistake\end{tabular}} \\ \hline
{Before, PersonX needed (xNeed):}      & { \begin{tabular}[c]{@{}c@{}}to arrest PersonY; \\ to try PersonY; \\ to prosecute PersonY; \\ to close the case\end{tabular}}             \\ \hline

{\textbf{Attributes of PersonX}}       & {}                                                                                                                                        \\ \hline
{PersonX is seen as (xAttr):}          & { \begin{tabular}[c]{@{}c@{}}just; lawful; justified; \\ mean; cruel; spiteful\end{tabular}}                      \\ \hline

{\textbf{Effects on PersonX}}       & {}                                                                                                                              \\ \hline
{As a result, PersonX feels (xReact):} & { \begin{tabular}[c]{@{}c@{}}satisfied for doing his/her duty; \\ sad for him; good; satisfied\end{tabular}}                      \\ \hline
As a result, PersonX wants (xWant):                         & \begin{tabular}[c]{@{}c@{}}to forget about PersonY; \\ to celebrate; \\ to keep PersonY in prison; \\ to find another case\end{tabular}                        \\ \hline
Has an effect on PersonX (xEffect):                         & \begin{tabular}[c]{@{}c@{}}gets petitioned; is hated; \\ stress level decreases; \\ body becomes relaxes; \\ gets threatened; \\ receives blame\end{tabular}   \\ \hline
\textbf{Effects on others}                                  & \textbf{}                                                                                                                                                      \\ \hline
As a result, others feel (oReact):                          & \begin{tabular}[c]{@{}c@{}}angry at personx for putting him/her in prison; \\ distressed; bad; worried\end{tabular}                                   \\ \hline
As a result, others want (oWant):                           & \begin{tabular}[c]{@{}c@{}}to get out of prison; \\ to not do anything stupid\end{tabular}                                                                     \\ \hline
Has an effect on others (oEffect):                          & \begin{tabular}[c]{@{}c@{}}blames x; threatens x; \\ body tenses; stress level increases\end{tabular}                                                          \\ \hline
\end{tabular}
\caption{ATOMIC examples from \citet{DBLP:conf/aaai/SapBABLRRSC19}.}
\label{tab:atomicexamples}
\end{table}

\citet{bosselut-etal-2019-comet} and \citet{Hwang2021COMETATOMIC2O} train common reasoning with a transformer by concatenating the strings into the form \textit{PersonX puts PersonY in prison $\langle$oWant$\rangle$ to get out of prison}. The transformer learns and conditions on the relation \textit{oWant}. The model can be trained with the standard GPT negative log-likelihood loss. The model in this thesis, by contrast, requires that the common sense knowledge be encoded into sentence vectors via the encoders \textit{f} and \textit{g}. The knowledge needs to be encoded in the relations between $u$ and $v$, not a single transformer. The approach is to split each relation into an \textit{f} input  \textit{PersonX puts PersonY in prison} and a \textit{g} input \textit{to get out of prison}. If there are multiple consequence clauses for each relation, they are split into a different training example. The dense classification layer then predicts from the concatenated vectors. The setup allows common sense reasoning with the exact same loss as the NLI datasets described. Like the NLI datasets, a single dense layer leads to a classifier with the dimensions of the relation types in ATOMIC. With ATOMIC $k = 8$ as there are more relation types. The loss is the cross-entropy loss when predicting the binary relations type. Like NLI training, the classifier head will change the $u$ and $v$ sentence representations to encapsulate the input sentence differences between relation types between each encoder.  Thus, in theory, the sentence embeddings will learn to represent the common sense relations within the ATOMIC dataset. Each event representing the example table is flattened out and batched together, with each event shuffled across the dataset. ATOMIC has $300$K events and $877$K relational clauses.

The complete sentence loss conceptually is $\ell_{\text{sent}} + \ell_{\text{snli}} + \ell_{\text{atomic}} + \ell_{\text{cond}}$. The $\ell_{\text{cond}}$ loss is not defined in this Chapter but instead in Section \ref{sec:conditionalgeneration}. $\ell_{\text{cond}}$ is to enable text generation in GPT-2 to be conditioned on a latent vector for the purposes of story generation. However it also acts as a form of autoencoder on the sentence representations, improving the sentence representations, and hence is beneficial for the comprehension tasks in this chapter.

The $\ell_{\text{snli}}$ and $\ell_{\text{atomic}}$ require different dataset to train. The approach taken is to randomly sample from the training datasets according to a probability in the training configuration. When an NLI dataset batch is sampled, the NLI loss is trained, and when the ATOMIC dataset is sampled the respective ATOMIC one is trained.  $\ell_{\text{sent}}$ and $\ell_{\text{cond}}$ can both be run on all the narrative databases but in practice it is not practical because of limited memory. Training randomly samples which of $\ell_{\text{cond}}$ and $\ell_{\text{sent}}$ to run in a batch if narrative dataset batch is sampled, and they are both enabled to the training configuration.

This section has reviewed multiple sentence losses. The Quick-Thought inspired loss referred to as $\ell_\textrm{\text{sent}}$ in figure \ref{fig:tdvaehierarchy}, and the NLI and ATOMIC losses are referred to as $\ell_\textrm{sent\_class}$. Together these losses should induce sentence representations that model the local context within a story and are enriched with entailment and common sense knowledge. The rationale is combined these represent a meaningful semantic state from which both to model suspense or, in the following chapter, to generate text.

This sub-section will conclude by touching on some failed exploratory tests for sentence losses. One characteristic that would be expected to be encoded in story text is the position of the sentence. From the reviewed narrative theory, for example, from \citet{freytag1894freytag} it is argued there are specific turning points in the story. It might be expected then that the position of a sentence within a story would encode important semantic information. For example, the beginning would introduce characters and the scenes. In contrast, around 80\% through the work, there might be expected to be a major setback such as a violent event or a romantic breakup. Even with surface-level lexical features, there are likely to be differences between the two. The intuition with trying to learn position is that it encodes information about where they typically occur in a story into the sentence representations. Suppose sentences are more likely or less likely to occur in different temporal positions. In that case, the sentence representation will encode wherein the typical plot events occur, which will be valuable in inferring suspense and story planning. Several approaches were tried, including a classifier to predict the absolute position of the sentence in the story directly between $0.0$ and $1.0$. A second approach bucketed the story in contiguous blocks of  $5$, $10$, or $20$ buckets and then, like the NLI or Atomic training, used a classification head to predict which bucket the sentence is within. However, all the mentioned variants failed to train correctly and caused degeneration of the sentence embeddings. As a result, this loss was not part of any of the evaluated models. There are other possible ways to encode temporality; one would be to train random pairs and a train a loss to infer which one was earlier. Nevertheless, it is surprising that the approaches trialled didn't work. \citet{Papalampidi2019MoviePA} has found similar embeddings can learn to predict turning points. While predicting precise turning points is different from a position, it would be reasonable to expect there is some overlap. This isn't discussed or evaluated further in the thesis. However, given the power of transformers in learning patterns, the model would be expected to learn some aspect of the position even if from surface forms and not underlying meaning. It cannot easily either suggest a weakness in the model. Alternatively, that positionality is not as important as Freytag like-minded theories suggest and that typical plot events are more fluid in where they occur. Either would be an interesting follow-up research question. 

Another approach to improving the sentence representations is to use an existing model with silver labels. The idea is to label training data with silver labels from a supervised model, and while noisy, the teacher model provides supervision to improve the representations of the unsupervised one. An initial idea was to label with sentiment polarity from VADER. \citep{Hutto_Gilbert_2014}; the classification head would then try and target the VADER labels with the polarity scores being split into five categories from strongly negative, negative, neutral, positive and strongly positive. This loss is not used in any reported models because it largely made no impact of the sentence embeddings. When added to existing losses, the loss quickly moves to a low loss and high accuracy, and then there is little training signal. The fast training suggests that the Quick-Thoughts are other losses that are encoding sentiment already, which make sense given that sentiment will be important to judging neighbouring sentences from other sentences in the batch. Another variant attempted loss, which as to use a classifier to predict the source dataset to encode metadata about the form of narrative, also proved unsuccessful. There are other variants of the approach of applying a teacher model such as SemBERT \citep{DBLP:conf/aaai/0001WZLZZZ20} that augments the raw text with semantic role labels to assist the sentence encoder with learning meaning by providing pointers. A similar approach could be taken with NER or coreferences, but these are at the cost of lots of additional preprocessing and are left to future work.

\subsection{Language Model}

The language model is Generative Pre-Training~2 (GPT-2; \citealt{radford2019language}), using the \textit{medium} model with $345$ million parameters as a base. For text generation, an auto-regressive language model is preferred to a bidirectional and masked one, such as BERT \citep{devlin-etal-2019-bert}. One of the goals is to combine the modelling of suspense and surprise with generation hence the preference for a left to right autoregressive LM. The architecture, however, would support bi-directional LM models if only modelling for suspense and surprise. Or the LM could be replaced completely by just sentence models such as the mentioned S-BERT. This would be a comparison for future work. However, part of the motive for including the same lower-level LM as the lower level approach and a similar sentence embedding layer is to make the comparison with the alternative generation-based suspense method in the previous chapter more valid. There is a secondary advantage in that narrative text is typically different from much of the text these language models are trained on, which often includes large amounts of web crawled text, Wikipedia text and news sources. Fine-tuning the LM makes the LM more relevant to the domain and should provide a stronger base for improved sentence representations in the layer above. The model is fine-tuned using the original LM loss in~(\ref{eqn:lm_loss}) on the story datasets. $w_j$ is a particular encoded word piece, $w$, in a passage of text.
\begin{myequation}
\ell_{\text{lm}} = - \sum_{j} \log P({w_j | w_{j_{-1}}, w_{j_{-2}}, \dots)}
\label{eqn:lm_loss}
\end{myequation}
Training GPT-2 works best on long blocks of text, whereas the hierarchical model, outlined later, encodes sentences as vectors and trains predictions based on the whole story. To train GPT-2 efficiently, the language model training is run distinct from the hierarchical training, with the training process alternating per batch between the two. This allows GPT-2 to be fine-tuned with a longer context of $1024$ word tokens.

\subsection{Alternative Losses}

Several alternate approaches were tried with sentence embeddings such as \textit{soft labels} or \textit{label smoothing} \citep{DBLP:conf/cvpr/SzegedyVISW16}. The notion behind \textit{label smoothing} \citep{NEURIPS2019_f1748d6b} is that one-hot coded hard labels like those proposed by the losses in this section can become overconfident in their predictions, and this hurts the generalisation of the model. \text{Label smoothing} distributes the probability mass around, so rather than having one target label of $1.0$ and the rest are zeroes, the incorrect labels are given small values, say, $0.1$ or $0.01$ depending on the number of classes. The correct label is then $1.0$ minus the sum of the incorrect labels. The rationale is that these smoothed labels provide a regularisation effect during training. \textit{Label Smoothing} can be easily applied to all the losses in this section; models were trained, and the configuration is in the Github repository. However, this produced no better results either for this chapter or the following and sometimes slightly worse results. As a result, these smoothed models are not reported. \citet{NEURIPS2019_f1748d6b} find that though smoothing works in some cases because it encourages labels of the same class to form tight classes, it also loses information between classes which they say has hurt model distilling. The described losses, such as the Quick-Thoughts one or predicting labels, are already looser in that neighbours will differ more and not be as clear as, say, distinguishing a cat or a dog or an image classification task. So the possible loss of information between classes may be why this does not work for the sentence vector representations.

Other variants trialled for the Quick-Thoughts inspired $\ell_\textrm{\text{sent}}$ were to make the loss wider. Quick-Thoughts only trains on the two immediate neighbours before and after with equal weighting for each. Instead, it may be desirable only to train on the next sentence. Or it may be desirable to train $2$ or $3$ as targets and the next $2$ or $3$ with varying weighting. The intuition is similar to label smoothing in that the neighbouring sentences are most relevant. However, neighbours further away are still relevant and so should be encouraged to be somewhat similar in vector representations. A scheme for each target was implemented that allows each target label to be weighted differently. Targets windows of up to $3$ on either side of the target were experimented with, as was both linear weighting and weighting nearer sentences higher. Weightings were tried that weighted the second nearest neighbour half or a third as the closest and the one beyond that half of a third again. Unfortunately, as per soft labels using both validation losses and the SentEval evaluation, these alternative schemes all degraded performance and aren't reported.

\subsection{Datasets}

The story datasets used for training consist of long and short story datasets. As noted in earlier chapters on suspense, storytelling is highly context-specific. Since the evaluation in this chapter is the same, the points aren't repeated. However, to address the limitation in addition to \textit{WritingPrompts} only models, a variety of other shorter and longer-form story datasets are trained on. The intuition behind this is that training on a more diverse set of stories improves the models' capabilities to generalise, and so the downstream task performance on suspense. The short story datasets consist of:

\begin{itemize}
   \item \textbf{WritingPrompts:} From \citet{fan-etal-2018-hierarchical} as per the last chapter is used as the default training dataset. The suspense evaluation is performed only against this dataset with the same dev/validation/test splits per the last chapter.
  \item \textbf{CMU Books:} \citet{DBLP:journals/corr/abs-1305-1319}
 is circa $15$K book plots extracted from Wikipedia. Plot summaries are of a similar length to the WritingPrompts stories but are not as diverse in topic.
 \item \textbf{ROC Cloze:} \citet{mostafazadeh-etal-2016-corpus} set of crowd sourced short stories of only $6$ sentences.
 \item \textbf{CMU Movies:} \citet{bamman-etal-2013-learning} is a corpus of circa $45$K movie plot summaries likewise extracted from Wikipedia. The Books and Movie corpora increase the diversity of the short stories the model is trained on.
\item \textbf{CBT (Children's Book Test):} \citet{DBLP:journals/corr/HillBCW15} is a corpus consisting of chapters from Children's literature. The original CBT was created as a comprehension test and has questions to determine best whether a machine reading system can follow the details of a plot. Children's literature is typically simpler in structure to adult works and about different themes, so the inclusion of this dataset is intended to increase the diversity of the narrative sources. The corpus has $108$ books in total.
\end{itemize}

In practice, models trained on multiple datasets performed better on automatic evaluation than single dataset trained models, and so only these results are presented in four experiments. For training, we use the existing dataset split for WritingPrompts.

While all these short story works are valid and models have been trained only on these corpora. One of the limitations of short stories of $50$ odd sentences is that the story needs to progress quickly and so they lack much of the detail and colour found in longer works. Many common plot elements such as extended descriptive narration of a scene or extended dialogue are often missing from these datasets. As are smaller everyday events which might make up much of what would be called common-sense knowledge. To address this the model is trained on longer form datasets:

\begin{itemize}
   \item \textbf{Shmoop:} \citet{DBLP:journals/corr/abs-1912-13082} is a dataset created from the online study guide.\footnote{The online study guide is at \url{https://www.shmoop.com/}
}. As it consists of classic book chapters aligned with summaries. Only the full text is trained on and not the summaries. The aligned summaries are part of the evaluation for Chapter \ref{chap:salience} on salience that explicitly tries to model salience in full-length books. The Shmoop corpus consists of $226$ books split into $7234$ chapters.
 \item \textbf{Books Corpus:} A replica version of the dataset from \citet{Zhu2015AligningBA}. The original was pulled for copyright reasons. A replica recreates the dataset from the original source \href{https://www.smashwords.com/}{Smashwords}, an online self-publishing platform. The books are largely contemporary fiction with a high degree of emphasis on genre fiction such as \textit{Romance}, \textit{Fantasy}, \textit{Crime} or \textit{Historical}. The corpus consists of circa $18$K works.
\item \textbf{Film Corpus:} \citet{Lin2011AllTW} is a corpus that consists of $862$ film scripts mainly from Hollywood.
\end{itemize}

The existing splits are kept for all the datasets with a predefined split, such as WritingPrompts, for training, validation, and testing. For all other datasets, the training examples are shuffled with a fixed seed (for consistency) and split by story according to a training ($80$\%), validation ($10$\%), test ($10$\%) split.

The above-described datasets allow the model to be trained on a mixture of short stories and long stories from various genres. The rationale for increasing the story genres the model is trained on is it improves the diversity of the data. As reviewed in the background chapters, increased diversity in datasets leads to the improved overall performance of many tasks. Together, the datasets cover short-form creative writing, book and movie summaries, children's fiction, extended contemporary fiction, classic books, plays, and movie scripts to represent a broad range of the narrative domain. The training configuration was split into three groups: Models only trained on WritingPrompts, those only on the short-story forms, and those trained on all. As per the last chapter, the evaluation is only on WritingPrompts, so the longer stories are more supplementary.

One issue of training with so many datasets is there are comparatively huge differences between the size of the datasets. For example, Schoop and Books Corpus consist of thousands of chapters from full-length novels, while CBT is a small set of Children's short stories. The danger is that the small datasets will be iterated far more quickly than larger datasets which risks overfitting on all the small datasets. An approach was taken to randomly sample a dataset batch in each training iteration to address this issue. The weightings are in the full configurations on Github, but the rough approach is sample small datasets less frequently to avoid overfitting them. Small is judged as being the discrete number of stories and not the length of text as the evaluation is on shorter stories, so larger corpora shouldn't dominate training.

\subsection{Implementation and Training Details}

The models are implemented on the AllenNLP framework \citep{gardner-etal-2018-allennlp} on Pytorch \citep{NEURIPS2019_9015}. The final evaluated models used the finetuned GPT-2 medium model for the LM layer, six layers of transformer with a $1024$ embedding size (matching GPT-2) for the sentence encoding top layers of the respective models: LSTM/Transformer/TD-VAE. The primary TD-VAE model has $485$M tunable parameters, circa $345$M are from GPT-2.

During training, batches of four blocks of $100$ sentences for each story were used. This was chosen primarily by balancing having a reasonable length context with the size of the model and fitting within the architectural constraints of training the model on $4$ $12$GB GPUs. For TD-VAE follows the source paper in picking $n$ (default $100$ samples per training batch from $t_1$ where $t_2$ representing a uniformly sampled range between $1$ and a max of $5n$ in experiments. 

SGD with Nesterov momentum \citep{DBLP:conf/ijcnn/BotevLB17} was used to train the model with a learning rate of $0.01$ and momentum of $0.9$. Models were trained with 10,000 batches per epoch (note this is not the whole dataset). Each of these epochs where the model fails to improve on the loss against the validation halved the learning rate. The rationale is that it gradually reduces the learning rate when the learning plateaus. Alternatives such as ADAM were tried but often proved unstable during training, so a simpler SGD plus Momentum method is preferred.

One additional training detail is that the model is too large to train all the losses at once using 12GB Nvidia GPUs and with a reasonable batch size even when layers of the hierarchy are split across GPUs to save memory. As a result, the training takes a rotation approach where the LM, sentence losses, and top-level TD-VAE losses are trained individually in a cycle of training batches. Cycling the training allows a large model to be trained with limited memory. Of course, in principle, all these losses could be trained simultaneously rather than in turn; the decision is for pragmatic reasons. In sentence specific loss batches such as NLI and ATOMIC, both the LM and sentence loss are trained simultaneously. The main memory usage which requires cycling is having a large batch of context for the hierarchical model and the sample of TD-VAE; both NLI and ATOMIC consist of only pairs of sentences without a long context and so doesn't have this constraint.

For the TD-VAE training per block, the model randomly samples $100$ example states to reconstruct where the target state is randomly sampled between $t+1$ and $t+n$ sentence ahead using a linear sampling scheme. Various different training schemes were tried from $n = 1$, which is an autoregressive model, up to $n = 7$. The most common configurations trained were $4$ and $7$, meaning the TD-VAE model is trying to predict $1 - 5$ and $1 - 9$ sentences into the future. As per the original paper, the $200$ samples are taken across the whole input context batch. So the whole batch of context sentences has $200$ samples and not each input sentence in the batch.

The preprocessing pipeline for the training datasets is as follows. It is the same as Chapter \ref{chap:suspenseannotation} except in step two where there were more cleanup rules for the additional datasets:

\begin{enumerate}
  \item \textbf{Langauge Check:} \href{https://github.com/indix/whatthelang}{Whatthelang} for detecting language and filtering out all non English texts. 
\item \textbf{Punctuation and Formatting Fixes:} Movie scripts and WritingPrompts, in particular, can have some strange line breaks, page breaks and repeating punctuation, and so cleanup is needed.
\item \textbf{Sentence Splitting:} Sentence splitting performed using the same sentence splitting approach as the previous chapter with \href{https://spacy.io/}{Spacy} 2.0.
\end{enumerate}

The main suspense evaluated models have the following parameter configurations. Multiple alternative configurations were tested using various combinations of layer and dimensions sizes, loss functions, source datasets, and training regime setup. The selection of the full evaluated models is based on validation losses and top k accuracy against the validation dataset. Cutting the number of evaluated models is necessitated by training and conducting downstream evaluations such as SentEval and Suspense in terms of computational resources and the cost of human evaluation for story generation in Chapter \ref{chap:tdvaegeneration}. The chosen model sizing also take account of the computational resources available. The evaluated configurations all have the same base structure:

\begin{itemize}
  \item \textbf{LM:} The GPT-2 medium model with pre-trained weights.\footnote{The only change is to add two tokens to the vocab \textit{zBlank} and \textit{endofsentence} with a corresponding adding of two extra outputs to the decoder layer. \textit{zBlank} is a replacement for underscores in Atomic. \textit{endofsentence} is a specific end of sentence marker. It was intended to be a collection class token for collating the sentence representation for final token pooling in the sentence encoders but performed worse than averaging. So isn't in evaluated models.}
  \item \textbf{Sentence Level:} Transformer layers that implement a bidirectional mask. \footnote{As noted earlier other variants on transformers are supported by default in the code such as bag-of-embeddings, LSTMs, or CNNs but trials showed they were inferior.} For each $f$ and $g$ pair there are distinct weights. \footnote{Tied weights were experimented with but were inferior on sentence losses.} The transformer uses positional embeddings of $128$ dimensions:
	\begin{itemize}
		\item \textbf{Heads:} $16$ attention heads per layer.
		\item \textbf{Layers:} $5$ layers of transformer.
		\item \textbf{Dims}: There are $1024$ dimensions on all reported models keeping it the same as GPT-2 medium. 
	\end{itemize}
 \item \textbf{TD-VAE:} The total TD-VAE archictecture consists of $6$ layers. Of the components:
 	\begin{itemize}
	\item \textbf{Belief Network:} A vanilla LSTM with a dimensionality of $2048$. The model supports reducing the inpt via a input projection layer but all trials with this reduced performance.
	\item \textbf{Prediction Network:} The latent space has $512$ dims, half of for the $\sigma$ and the other half are for the $\rho$ of the Gaussian sampling. The dimensionality reduction is from $2048$ as the sentence embeddings from $f$ and $g$ are concatenated together.
	\item \textbf{Inference Network:} The same as the \textit{Prediction Network}.
	\item \textbf{Decoder Network:} The decoder network uses two RELU layers of $6144$ and $4096$ and a linear layer of $2048$ which reconstructs the original sentence embedding input before projection. This is projected from a dim size of $2048$ the internal $Z$ representation size multiplied by the number of layers.\footnote{As noted earlier, the model follows the architecture of the TD-VAE paper in that the reconstruction can select features from each layer and not only the top. The top only would be an alternative implementation, but this was not tried.} Pyramid stepping improves the reconstruction loss from a high dimensionality.
	\end{itemize}
\end{itemize}

\subsection{Ely Inference}

Equations \ref{eqn:ely_surp_rep} and \ref{eqn:ely_susp_rep} are the two Ely surpise and suspense reproduced from Chapter \ref{chap:rolloutsuspense}. 

\begin{myequation}
\begin{aligned}
S_t^\text{Ely} = (e_{t} - e_{t-1})^2
\label{eqn:ely_surp_rep}
\end{aligned}
\end{myequation}

\begin{myequation}
\begin{aligned}
U^\text{Ely}_t = \mathbf{E}[(e_t - e^i_{t+1})^2] \\
=  \sum_{i} P(e^i_{t+1})(e_t - e^i_{t+1})^2
\label{eqn:ely_susp_rep}
\end{aligned}
\end{myequation}

Now that the model has been defined, the main question is how the TD-VAE model can infer both surprise and suspense. The $e$ in these equations from the previous chapter represents the encoded sentence representations. In the context of the hierarchical model, the representation is the same, the sentence encoding, which is the input to TD-VAE. The requirement for modelling suspense is an expectation over the future state, which, as has been discussed, is intrinsic to the model. Surprise is comparing the expected state with a future state. With TD-VAE there are, however, multiple components with internal latent states that can be modelled. The TD-VAE model projects forward the reconstructed state $e_t$, the latent state $z_t$, and the belief state from the LSTM $b_t$. The $b_t$ states are the final LSTM encoder states and so lack Bayesian inference capabilities and is not suitable for modelling suspense. However, both the $z_t$ latent state and reconstructed state can be used to model surprise and suspense.  The model works either in the input or the latent space $z_t$ or the reconstructed $e_t$. Either way, the process for modelling suspense is sampling $i$ examples to $t$ timesteps ahead. Remember, the model supports more than sampling one step ahead, represented by $n$ and $i \in I$ an index into a set of samples. Suspense is then reformulated in the latent space from Ely as eqn. (\ref{eqn:ely_susp_z}).

\begin{myequation}
\begin{aligned}
U^\text{Ely Z}_t = \mathop{\mathbf{E}}[(z_t - z^i_{t+n})^2] 
\label{eqn:ely_susp_z}
\end{aligned}
\end{myequation}

Adapting this to TD-VAE is shown in eqn. (\ref{eqn:ely_susp_z_exp}). The two main changes: First, as per the previous chapter Ely is defined explicitly as a variance hence the squared component. This is the case with an L2 distance, but other distances such as L1, Cosine, or Wasserstein distance can be applied to any vector space measure. The \textit{dist} function generalises the equation. Second, the previous chapter had two components, the first of which generated concrete sentence continuations and a second that inferred a probability distribution over the continuations or tree of continuations. To recap, the intuition is more likely outcomes should be weighted more heavily than less likely ones. TD-VAE samples from a Gaussian distribution, and so all the samples then become equally weighted, which means all sampled outcomes are considered equally. TD-VAE directly models the posterior distribution over $z$, and so repeated sampling will lead to more samples clustered around more likely outcomes and makes it equivalent to the Ely definition.

\begin{myequation}
\begin{aligned}
U^\text{Ely Z}_t = \sum_{i \in I} \frac{1}{\textrm{count}(i \in I)}  \cdot \textrm{dist}(z_t - (z^i_{t+n}|z^i_{t}))
\label{eqn:ely_susp_z_exp}
\end{aligned}
\end{myequation}

The stated equations are modelled directly in expectations in the latent space, $z$. $(z^i_{t+n}|z^i_{t})$ is a notational simplification for the future latent state being projected from the earlier timestep. It is the forward inference process of TD-VAE. The definition is the same when run in the sentence embeddings space $e$, see eqn. (\ref{eqn:ely_susp_e_exp}). The difference is that the $z$ version is comparing the latent space $z_{t_1}$ with hypothetical $z_{t_1}$. The $e$ version takes it a step further and runs the decoder for each projected sample, and measures the expectation of the difference between $e_{t_1}$ and $e_{t_2}$. Both are valid interpretations of Ely but operate in different latent spaces and so are worth comparing. For the inference, $100$ samples are taken for each $t$ into the future. 

\begin{myequation}
\begin{aligned}
U^\text{Ely E}_t = \sum_{i \in I} \frac{1}{\textrm{count}(i \in I)}  \cdot \textrm{dist}(e_t - (e^i_{t+n}|e^i_{t}))
\label{eqn:ely_susp_e_exp}
\end{aligned}
\end{myequation}

A difference is that in the previous chapter, surprise was taken as the difference between the $e_{t}$ and $e_{t-1}$ the encoded sentences. This is a simplification because the model relied on generation and discrimination to model suspense. It doesn't directly autoregressively project forward the state to $t+1$. It couldn't directly model expectations at a future timestep for $e$ against the real encountered $e$. An autoregressive version of an LSTM model such as those reviewed at the beginning would work but so does the hierarchical TD-VAE model defined in this chapter. So surprise then becomes the projections versus the real state at time $t$ versus the actual state. Eqn. (\ref{eqn:ely_surp_z_exp}) shows the revised surprise where surprise is the distance between the expectations of the latent stats $z$ projected forward to a future timestep and the real latent vector at that timestep. Once again, the same can be done in the TD-VAE input space $e$ just by running the decoder and comparing the reconstruction, eqn. (\ref{eqn:ely_surp_e_exp}). 
Another functional difference in surprise from the previous chapter is because that TD-VAE can project forward multiple timesteps. So there can be a surprise from looking multiple steps ahead. 

\begin{myequation}
\begin{aligned}
S_t^\text{Ely Z} = \textrm{dist}(z_{t} ,  \mathop{\mathbf{E}}[(z^i_{t} | z^i_{t-n})]
\\
= \textrm{dist}(z_{t} , \sum_{i \in I} \frac{1}{\textrm{count}(i \in I)}  \cdot (z^i_{t} | z^i_{t-n})))
\label{eqn:ely_surp_z_exp}
\end{aligned}
\end{myequation}

\begin{myequation}
\begin{aligned}
S_t^\text{Ely E} = \textrm{dist}(e_{t} ,  \mathop{\mathbf{E}}[(e^i_{t} | e^i_{t-n})]
\\
= \textrm{dist}(e_{t} , \sum_{i \in I} \frac{1}{\textrm{count}(i \in I)}  \cdot (e^i_{t} | e^i_{t-n})))
\label{eqn:ely_surp_e_exp}
\end{aligned}
\end{myequation}

As per the last chapter, suspense is forward-looking, and surprise is backwards-looking. In the case of suspense, the hierarchical TD-VAE model samples a probability distribution over future states. If, as in the last chapter, those representations are meaningful and to the plot, then this is a plausible model of suspense. As per the discussed theory, if the point in the narrative has highly divergent outcomes, then the suspense will increase, and when they become more limited, they will decrease. Recall the divergence isn't between the samples of a future state but between the current state and future state. So the model isn't counting the number of paths that could be taken but how consequential each path is vis a viv the current situation. The difference is rather than rely on GPT-2 to generate these in concrete forms, it is sampled from the latent space. Surprise is formulated differently but more in line with Ely than the previous model. It is the expectations versus the reality of the situation so that if expectations in the future deviate more from reality, the measure will be higher.

Reconsidering the comedy example, \textit{Mr Bean} and \textit{Only Fools and Horses}. In only \textit{Only Fools and Horses}, when \textit{Del Boy} falls through the bar, it is unexpected. With surprise, the model will most likely expect the conversation to continue, and so translating to text would generate dialogue discussion. There may be an alternative, and it is a sampling process. The counter for the bar is lifted up just before, and a model might infer falling through is a possibility, but it will be a tiny number of the samples. Thus the actual results being very different from the sample distribution will lead to a high surprise score. The same situation would not apply to the suspense. Most of the expectations will carry on the conversation, so the distribution of samples will have narrower, and hence suspense lower. 

From \textit{Mr Bean}, we should expect a surprise from the unusual actions versus what we expect when someone eats a sandwich. This description is more illustrative of a human for suspense as it would likely require comprehension beyond the model in the thesis. Unexpected incidents happen in the sequence, such as cutting slices of bread with scissors or washing lettuce in his sock. After this, the situation is not the normal one of someone eating a sandwich in a park. In theory, the highly unusual events will condition the model to expect a greater variance in outcome since they suggest things are not normal. This will generate a greater variance over future outcomes from the sampled space with the current situation, which will increase suspense.  The \textit{Mr Bean} example is unusual in being an absurdist comedy. Similar signs of cues exist in most narratives. From the thriller \textit{Gone Girl}: One of the more dramatic and suspenseful scenes is where \textit{Amy} is staging her own disappearance. She makes a mess of the place to make it look like there has been a struggle. The voice-over is \textit{Amy} talking about her husband's affair. Both of these unusual events should increase the uncertainty of the model about what will happen. Compare this to if someone was cleaning the house; then there would be a low divergence of uncertainty as we should expect normal hoovering, dusting, etc., activities to continue. The scene ends with \textit{Amy} holding a hammer in front of her face, in the bathroom. The intuition is the context should increase the expectation of divergence from the high state, thus increasing suspense. After all, in the scene when she holds the hammer up to her face given the previous minute of action, we the human viewers do not expect her to fix a loose mirror in the bathroom. The context creates the impression that more dramatic events will happen, i.e. suspense.

This idea of suspense requires a complex causal chain which, as noted in the background material in Chapter \ref{chap:backgroundtheory} and Chapter \ref{chap:backgroundml}, deep learning models have some capacity for but also severe limitations. Just as in the last chapter, the rationale for the model is not that it can fully model these causal chains. But instead, there is some moderate level of modelling combined with some surface-level features that could correlate with the underlying model that would provide a reasonable inference of suspense and surprise.

\section{Experiments}

\subsection{Introduction}

The experimental work for the TD-VAE model has three parts. The first part is an assessment of the sentence vectors encoded by the middle tier of the hierarchy. The second and third parts are direct comparisons between TD-VAE models on the suspense annotated \textit{WritingPrompts} corpus and on \textit{TRIPOD} turning points. For both of these, all evaluations have the same measures and methods as in the previous chapter for direct comparison between the approaches.

\subsection{Sentence Representations}

One potential problem with adhoc latent sentence representations is, by definition, they are opaque. Despite best efforts to select the losses to induce appropriate properties and the losses training correctly, and performing well against the validation set, it isn't clear what the learnt sentence embeddings represent. It is crucial because the surprise and suspense method depends on projecting the latent vectors, so the method will only be expected to be relevant to the narrative.  To address this the sentence representations are evaluated with SentEval \citep{conneau-kiela-2018-senteval} and additional probing tasks \citep{conneau-etal-2018-cram}. SentEval is a suite of linguistic and semantic tasks designed to test sentence embeddings. To take one task, MR (Movie Reviews), is to classify movie reviews from the sentence text according to their star rating. The task has a given set of sentences; the TD-VAE model is adapted to output vector representations of arbitrary dimensions. The SentEval toolkit will then train a logistic regression model taking these vectors as input features against the tasks training set. The model is trained $n$ times per task across $n$ cross-validation folds, averaging across the test set being the model's score for a given task. The approach provides a test for whether the latent vector contains information that infers the goal of the task and hence, taking the example, whether the vectors can infer the positivity or negativity of movie reviews.  As well as testing the properties of the vectors, SentEval also allows a more direct comparison between models.

All reported tasks are with $5$ fold cross-validation, the Adam optimizer with the default learning rate $0.01$, up to $50$ epochs of training and early stopping after $3$ epochs of no improvement.

For SentEval the evaluated variants are based on combinations of the trained datasets and the sentence encoding losses. The key is as follows:

\begin{enumerate}
  \item Datasets
	\begin{enumerate}
  		\item \textbf{WP:} Trained on WritingPrompts.
		\item \textbf{ALL:} Trained on all datasets mentioned in the chapter apart from ATOMIC and the NLI datasets.
	\end{enumerate}
	\begin{enumerate}
		\item \textbf{QT:} The quick-thoughts inspired $\ell_{\text{sent}}$ loss.
		\item \textbf{C:} The conditioning loss $\ell_{\text{cond}}$.
		\item \textbf{KB:} With both $\ell_{\text{snli}}$ + $\ell_{\text{atomic}}$ the two commonsense knowledgebase training datasets.
	\end{enumerate}
\end{enumerate}

It should be noted that there were also models trained with only the short form datasets - WritingPrompts, CBT, and the CMU Movie and Book Summary datasets - but these performed similarly with the sentence validation loss accuracy to the WritingPrompts only model and so the results are omitted. 

The SentEval results are split into four tables, two for the semantic tasks in tables \ref{tab:senteval_var} and \ref{tab:senteval_var2} and for the linguistic  probing tasks in tables \ref{tab:senteval_probe} and \ref{tab:senteval_probe2}. The other baseline models are  Skip-Thoughts \citep{NIPS2015_5950}, BERT (using the [CLS] token as the sentence representation) \citep{devlin-etal-2019-bert}, S-BERT \citep{reimers-gurevych-2019-sentence}, InferSent \citep{conneau-etal-2017-supervised} for the semantic tasks. BERT [CLS] takes the [CLS] special token as the sentence representation, which is what it was designed for. For the linguistic probing tasks, it is Fasttext  \citep{joulin2017bag} and Skip-Thoughts \citep{NIPS2015_5950}. The task baselines have been selected to encompass relevant alternatives and not be comprehensive, but for the semantic tasks S-BERT was SOTA at the time these experiments were performed. Much weaker baselines such as averaging Word2Vec or Glove embeddings are not reported as they are much weaker across the board than any of these models.

The primary purpose of the analysis is not to establish a SOTA benchmark, but rather: First to understand what information the sentence losses is encoding into embeddings. Second, make sure the model is competitive, especially in semantic tasks, even if not SOTA. The third is to measure the differences caused by the disparate training sets and sentence loss combinations. State of the art is not expected because the sentence representations are mainly finetuned on domain-specific story datasets. In contrast, the other sentence models in the evaluation are generally trained on datasets from a broader range of domains. Likewise, the tasks are more general-purpose rather than being focused on stories.

\begin{table}[]
\centering
\begin{tabular}{cccccc}
\hline
\textbf{Model $\uparrow$} & \textbf{MR}   & \textbf{CR}   & \textbf{SUBJ} & \textbf{MPQA} & \textbf{SST-2}  \\ \hline
Random Embedding & 48.1 & 61.3 & 49.8 & 68.4 & 49.2 \\
Skip-Thought    & 79.4          & 83.1          & 93.7          & 89.3          & 82.9            \\
InferSent      & 81.1          & 86.3          & 92.4          & 90.2          & 84.6            \\
BERT {[}CLS{]} & 78.7          & 84.8          & 94.2          & 88.2          & 84.1         \\
S-BERT-NLI     & \textbf{84.9} & \textbf{90.1} & \textbf{94.5} & 90.3 & \textbf{90.6}           \\ \hline
WP+QT          & 79.1          & 84.8          & 84.4          & \textbf{93.0}          & 83.8                     \\
ALL+QT         & 77.0          & 82.8          & 82.6          & 92.6          & 78.8                  \\
WP+C           & 78.5          & 82.9          & 82.5          & 92.9          & 85.0                     \\
ALL+C          & 77.4          & 81.3          & 84.7          & 92.7          & 83.3                     \\
WP+KB+QT       & 78.8          & 83.9          & 84.8          & 92.6          & 82.7                     \\
ALL+KB+QT      & 78.9          & 83.4          & 85.6          & 92.7          & 82.3                     \\
WP+KB+QT+C     & 77.2          & 82.6          & 92.8          & 92.8          & 82.7                     \\
ALL+KB+QT+C    & 77.2          & 82.6          & 84.7          & 92.8          & 82.9                     \\ \hline
\end{tabular}
\caption{A selection of benchmarks from SentEval. The tasks are MR (Movie Review), CR (Product Review), SUBJ (Subjectivity Opinion), MPQA (Opinion Polarity), SST-2 (Binary Sentiment Analysis).}
\label{tab:senteval_var}
\end{table}

To analyse the results from the main SentEval tasks in Table \ref{tab:senteval_var}. Across the board, S-BERT (Sentence Bert) is the strongest across all these tasks that are primarily related to opinion and sentiment. With these tasks broadly speaking, the TD-VAE sentence embedding variants perform slightly better on some tasks than the other baselines, e.g. MPQA (Opinion Polarity), and slightly worse on the others, such as CR (Product Reviews).  On these tasks, there aren't many differences between any of the dataset or sentence loss variants. In particular, the Quick-Thoughts only variant the \textit{all datasets} trained model performs slightly worse. 

\begin{table}[]
\centering
\begin{tabular}{ccccc}
\hline
\textbf{Model $\uparrow$} & \textbf{TREC} & \textbf{MRPC} & \textbf{SICK-E} & \textbf{SICK-R} \\ \hline
Random Embedding & 18.0 & 64.35 & 54.9 & - \\
Skip-Thought             & 88.4          & 72.4          & 79.5          & 85.8          \\
InferSent               & 88.2          & 76.2          & \textbf{86.3} & \textbf{88.4} \\
BERT {[}CLS{]}    & 91.4          & 71.1          & -             & 42.6          \\
S-BERT-NLI    & \textbf{97.4} & 75.9          & 76.5          & 73.8          \\ \hline
WP+QT   & 90.4          & 68.7          & 69.7          & 31.5          \\
ALL+QT   & 86.4          & 64.5          & 65.2          & 36.4          \\
WP+C     & 88.2          & 67.1          & 73.9          & 39.1          \\
ALL+C     & 91.2          & 71.5          & 73.6          & 40.0          \\
WP+KB+QT       & 91.4          & 68.5          & 69.0          & 48.1          \\
ALL+KB+QT     & 89.2          & 75.7          & 68.9          & 48.8          \\
WP+KB+QT+C   & 88.4          & 79.9          & 75.4          & 60.0          \\
ALL+KB+QT+C    & 91.2          & \textbf{80.2} & 77.0          & 60.2          \\ \hline
\end{tabular}
\caption{A selection of benchmarks from SentEval. The tasks are TREC (Question Type Classification), MRPC (Paraphrase Detection), SICK-E (Natural Language Entailment), SICK-R (Semantic Text Relatedness).}
\label{tab:senteval_var2}
\end{table}
There are more significant differences in the second set of SentEval tasks in Table  \ref{tab:senteval_var} and Table \ref{tab:senteval_var2}. S-BERT is not the best performing model across the tasks, and Infersent is the best performing model on two of the tasks. The TD-VAE models sentence embedding is once again generally competitive with the baselines. One clear difference is that combining the losses improves performances on these losses. The best performing versions are both versions that combine all the sentence losses, and the best overall is the All dataset version. The tasks TREC (Question-time Classification), MRPC (Paraphrase Detection), SICK-E (Entailment), and SICK-R (Semantic Relatedness) are more directly related to storytelling. The previous table was more about opinion and sentiment. While sentiment is relevant to stories, the relatively worse scores for the TD-VAE vectors are on the review tasks, a pretty different domain. As reviewed in chapter \ref{chap:backgroundtheory}, tasks such as semantic relatedness and entailment are far more relevant, and it is where the combined losses improve performance. However, one concern is that with SICK-R, which is direct semantic relatedness, most of the variants substantially underperform most of the benchmarks. Only the BERT [CLS] is worse than the TD-VAE models sentence representations, and combining all losses is a substantial improvement. This point is revisited in the discussion from the current and related following generation chapter.

\begin{table}[]
\centering
\begin{tabular}{cccccc}
\hline
\textbf{Model $\uparrow$} & \textbf{SentLen}   & \textbf{WC}   & \textbf{TreeDepth}   & \textbf{TopConst}   & \textbf{BShift}   \\ \hline
Random Embedding & 17.1 & 0.12 & 17.6 & 4.7 & 50.2 \\
BoV-fasttext              & 66.6          & \textbf{91.6} & 27.1          & 68.1          & 50.8                \\
Skip-Thought               & 68.1          & 35.9          & 33.5          & \textbf{75.4} & 60.0                   \\
InferSent               & 71.7          & 87.3          & 41.6          & 70.5 & 65.1                   \\ 

\hline
WP+QT                     & 92.0          & 27.1          & 32.6          & 66.5          & 76.2             \\
ALL+QT                    & 92.7          & 24.0          & 31.9          & 66.9          & 73.3                  \\
WP+C                      & 93.3          & 37.7          & \textbf{34.8} & 75.1          & \textbf{80.2}  \\
ALL+C                     & 92.2          & 33.6          & 34.4          & 75.0          & 78.0          \\
WP+KB+QT                  & \textbf{93.5} & 26.3          & 32.5          & 68.5          & 75.9             \\
ALL+KB+QT                 & 93.4          & 26.2          & 32.2          & 67.2          & 74.7          \\
WP+KB+QT+C                & 92.4          & 32.8          & 34.1          & 72.8          & 78.1     \\
ALL+KB+QT+C               & 92.4          & 32.8          & 32.8          & 73.5          & 78.1     \\ \hline
\end{tabular}
\caption{SentEval linguistic probing tasks: The SentEval probing tasks : SentLen (Sentence Length), WC (Word Content Analysis), TreeDepth (Tree Depth), TopConst (Top Constituents Prediction), BShift (Word Order Analysis).}
\label{tab:senteval_probe}
\end{table}

\begin{table}[]
\centering
\begin{tabular}{cccccc}
\hline
\textbf{Model $\uparrow$} & \textbf{Tense} & \textbf{SubjNum} & \textbf{ObjNum} & \textbf{OMO} & \textbf{CoordInv} \\ \hline
Random Embedding & 50.1 & 50.8 & 50.4 & 51.3 & 51.4 \\
BoV-fasttext                       & \textbf{89.1} & 82.1           & 79.8           & 54.2          & 54.8          \\
Skip-Thought     & \textbf{89.1} & 80.5           & 77.1           & 55.6          & \textbf{67.7}          \\ 
InferSent               & 86.7          & 80.7          & 80.3          & \textbf{62.1} & 66.8                   \\ 
\hline
WP+QT     & 85.2          & 83.7           & 75.2           & 57.6          & 64.5          \\
ALL+QT    & 86.9          & 82.9           & 73.5           & 54.8          & 62.3          \\
WP+C     & 83.7          & 86.6           & \textbf{78.7}  & 59.6 & 65.7 \\
ALL+C    & 84.3          & \textbf{87.1}  & 78.4           & 58.3          & 65.7 \\
WP+KB+QT  & 87.8          & 84.9           & 74.5           & 56.7          & 65.1          \\
ALL+KB+QT     & 87.7          & 84.2           & 75.1           & 56.6          & 65.3          \\
WP+KB+QT+C  & 85.1          & 86.6           & 76.5           & 58.8          & 65.6          \\
ALL+KB+QT+C  & 85.7          & 86.4           & 76.5           & 57.6          & 65.6          \\ \hline
\end{tabular}
\caption{SentEval linguistic probing tasks:  Tense (Tense), SubjNum (Subject Number) , ObjNum (Object Number), OMO (Odd Man Out), CoordInv (Coordination Inversion). }
\label{tab:senteval_probe2}
\end{table}

The linguistic probing tasks in tables \ref{tab:senteval_probe} and \ref{tab:senteval_probe2} are less relevant to suspense inference. Still, both tables show that a substantial amount of linguistic knowledge is also being encoded into the vectors. The one concern is the WC (Word Content) task, where all of the variants perform poorly. The task is to predict the correct missing word from $1000$. For example, given a sentence \textit{Jayden said, " Elijah will take you over to the \_\_\_\_, where you can design your strategy} the SentEval would try and learn to infer \textit{Farm} which is one of the thousand possible outputs words in the classifier. It might be expected that sentence embedding with the $\ell_{\text{cond}}$ would perform well on the task. After all, the loss is projecting learning to condition on the sentence embedding when running the GPT-2 loss. The loss should encourage generating tokens which are in the real sentence, which in theory should encode information that would help with the task. It could be a result of the way the loss is setup whereby the latent vector is projected into the GPT-2 space at the start of the sentence, and therefore the loss guides the general direction of generation rather than specific words. Again this point is raised as it's relevant to the discussion of the related results. It should be noted, though, that although the task performance is much worse than the best sentence vector models, it is similar to Quick-Thoughts and much better than the random baseline, which is close to zero for this task.

Overall the sentence representations results show that the sentence losses induce competitive performance with other sentence representations across a range of semantic and linguistics tasks. Combining the sentence losses improves overall performance. As noted in the rationale for the model, further work could swap out the sentence representation and lower-level LM model for an alternative. However, such as model would lack the generation capability and crucially for the thesis, be less directly comparable to the models in the previous chapter.

\subsection{Surprise and Suspense}

This section of the chapter reports results comparing the TD-VAE model versus the hierarchical rollout model from Chapter \ref{chap:tdvaegeneration}. The following best baselines are carried forward for comparison from the previous chapter:

\begin{itemize}
  \item \textbf{Human:} Naturally, the human evaluation is carried forward.
\item \textbf{GPTSim:} Of the several baselines from the previous chapter, just chosen for representative comparison as all of the baselines were fairly close to random performance.
 \item $\mathbf{U^{\text{Gen}}}$: The Ely suspense measure with the Generation method. In the previous chapter it was given as $U^{\text{Ely}} - \text{Gen}$. The simplification is because the \textit{Hale} alternatives were inferior and so not worth comparing, and all the TD-VAE variants are derived from \textit{Ely}.
 \item $\mathbf{S^{\text{Gen}}}$: The surprise Ely metric. The Generation or Corpus method could have been compared, but the Generation is preferred for its slightly better performance overall. Both this and  $\mathbf{U^{\text{Gen}}}$ use a cosine distance in the vector space. All of the evaluation is also on the LSTM versions as these had the best performance.
\end{itemize}

The metric notation is more complicated for the TD-VAE model since the model can potentially look ahead multiple steps, and there are multiple vector spaces. All metrics are derived from the Ely concept of suspense and surprise. An example of the notation is: $S^{Z}_{t=-1}(\text{Cos)}$ . The $S$ represents surprise as per the previous chapter, and $U$ is suspense. The $t$ index is how many sentences before or after the metric in sentences; for surprise, it is negative because it is backwards-looking and positive for suspense as it is forward-looking. The exponent is either $Z$ for the internal latent vector space or $X$ for the reconstructed input space. The final dimension is the distance function inside the function, Cosine. In the results, Cosine and L2 are reported. Other distance metrics tried included L1, Jensen-Shannon distance, KL-Divergence, Earth Mover (or Wasserstein) distance. These results are not reported because either they are too similar to L2 and Cosine, as in the case of L1 or dot product respectively, or performed worse, as with the other distance metrics.\footnote{As discussed in the method section, one of the advantages of operating in the vector space with Ely definition of surprise and suspense is that any distance metric can be applied.}

\begin{figure}[htbp]
\centering
\includegraphics[trim={0.5cm 1.0cm 2.0cm 2.5cm},clip,width=1.0\textwidth]{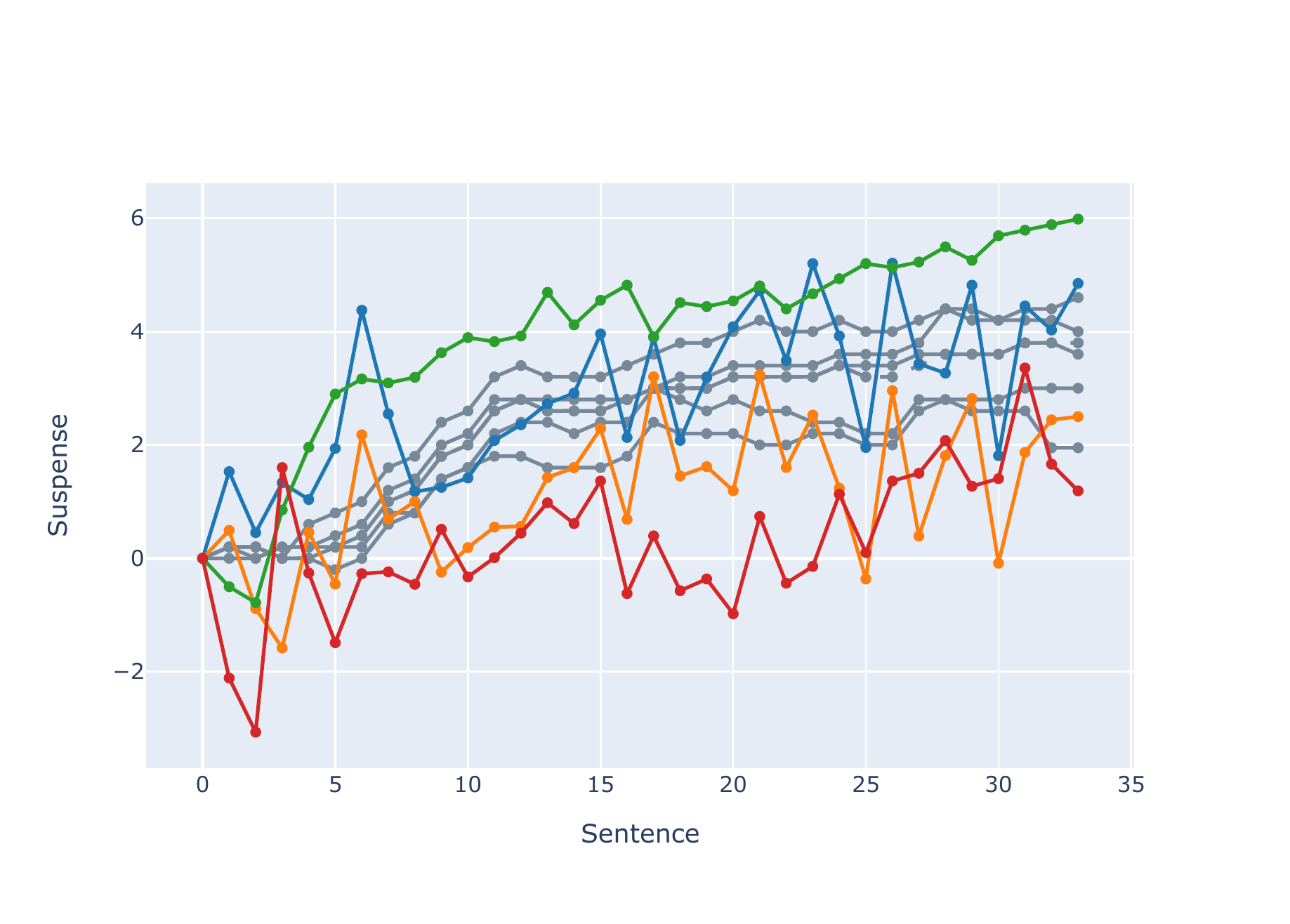}
\caption{Story \textit{27} plot comparing annotations with model predictions: \textbf{\textcolor{annotatedcolor}{Human}}
,\textbf{\textcolor{plotacolor}{$U^{Z}_{t=1}(\text{Cos)}$}}, \textbf{\textcolor{plotbcolor}{$U^{Z}_{t=1}(\text{L2)}$}}, \textbf{\textcolor{plotccolor}{$S^{Z}_{t=-1}(\text{Cos)}$}}, \textbf{\textcolor{plotdcolor}{$S^{Z}_{t=-1}(\text{L2)}$}}.}
 \label{fig:tdvaesuspense_27_plot}
\end{figure}

The models of suspense can also be plotted against the annotations for a story.  Figure \ref{fig:tdvaesuspense_27_plot} shows the annotations plotted against L2 and Cosine versions of suspense and surprise for the $Z$ space. As will be shown in the results to follow, the $Z$ space outperforms the $X$ space. The model is the \textit{All} variants where all TD-VAE layers are conditioned on to reconstruct the input $X$. This example was also used in Chapter \ref{chap:rolloutsuspense}, and is a case where there is strong annotator agreement and the more expected upward trajectory with a plateau or slight decline at the end.  The plot shows a  good match between surprise measures and the annotations in the example. Suspense is a lot more variable with a less clear correlation pattern with the annotations. This is a strong correlation example where the TD-VAE plots, other worse ones are plotted later. Nevertheless, the pattern between the plotted measures is repeated throughout the evaluation. Surprise in the latent space $Z$ correlates relatively well, whereas suspense performance is much weaker.

\begin{figure}[htbp]
\centering
\includegraphics[trim={0.5cm 0.0cm 0.5cm 2.5cm},clip,width=1.0\textwidth]{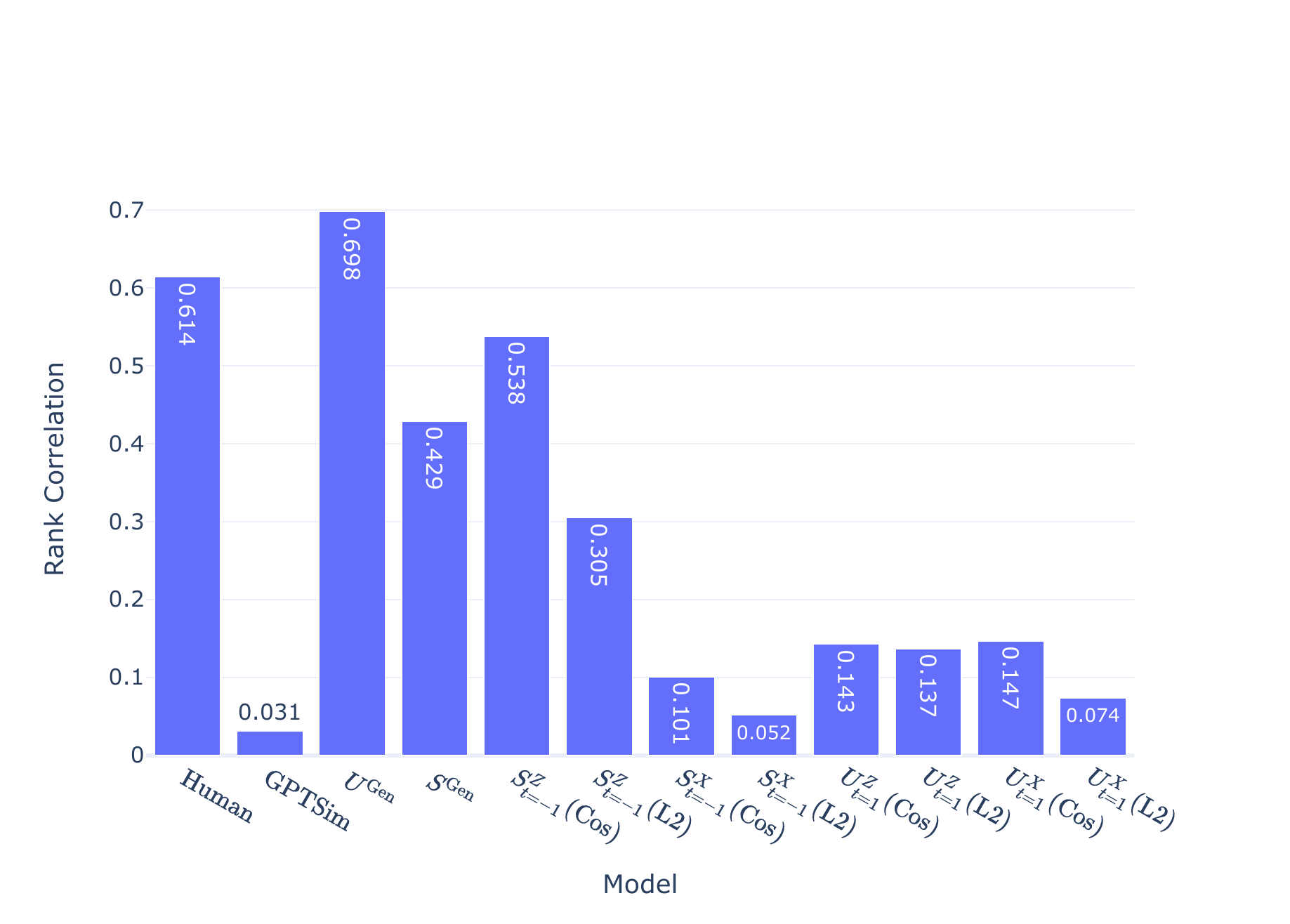}
\caption{Comparison between the base model and the hierarchical rollout results with Spearman's $\rho$. The model is the All layer version where $X$ is reconstructed from across all TD-VAE $Z$ layers.}
 \label{fig:tdvaesuspensesurprisebaselinespearman}
\end{figure}

\begin{figure}[htbp]
\centering
\includegraphics[trim={0.5cm 0.0cm 0.5cm 2.5cm},clip,width=1.0\textwidth]{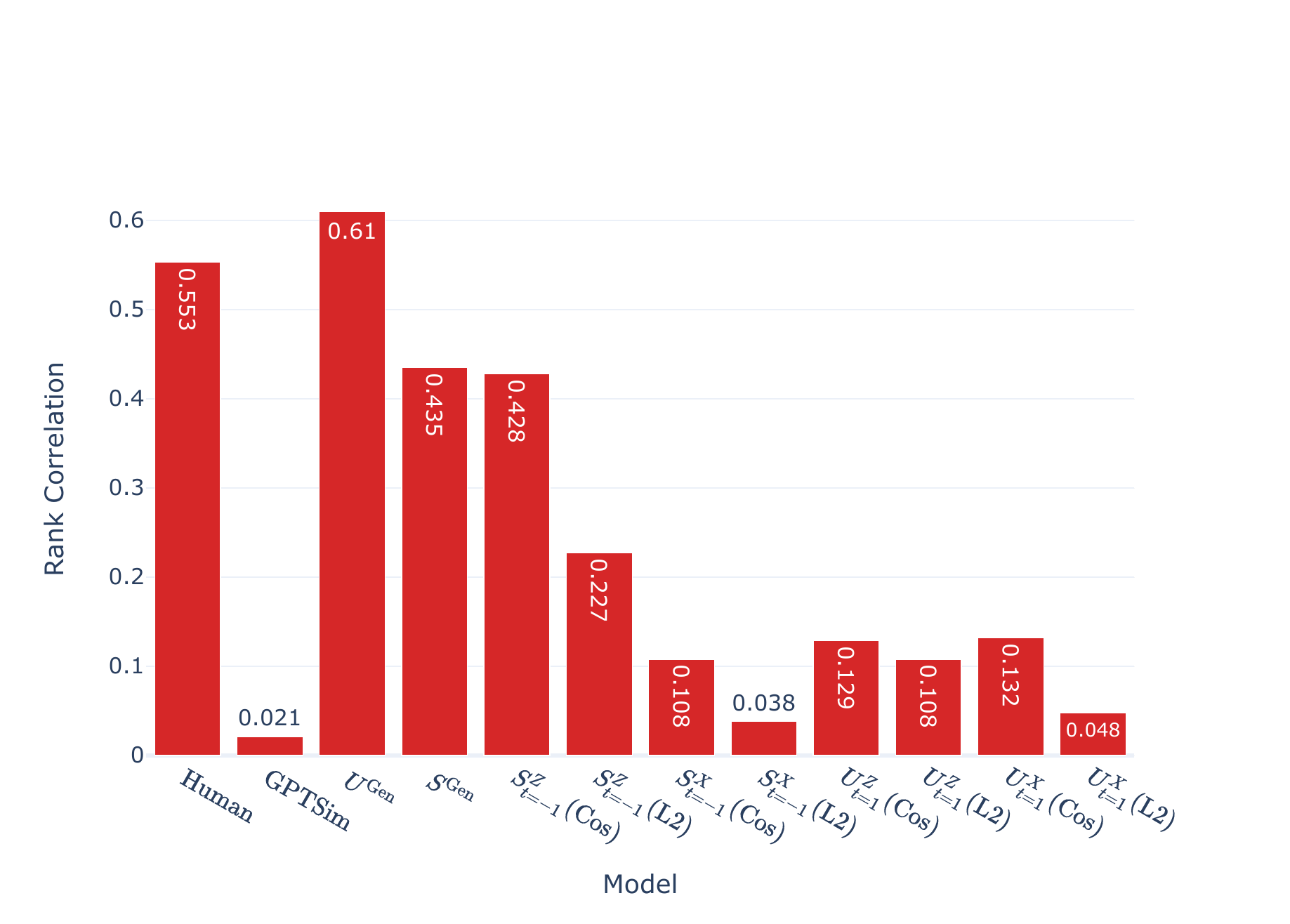}
\caption{Comparison between the base model and the hierarchical rollout results with Kendall's $\tau$. The model is the All layer version where $X$ is reconstructed from across all TD-VAE $Z$ layers.}
 \label{fig:tdvaesuspensesurprisebaselinekendall}
\end{figure}

The main results for the \textit{All} TD-VAE are shown in figure \ref{fig:tdvaesuspensesurprisebaselinespearman} for Spearman's $\rho$ and figure \ref{fig:tdvaesuspensesurprisebaselinekendall} and Kendall's $\tau$. The correlation method against both Kendall's $\tau$ and Spearman's $\rho$ is the same as the previous chapter, and the non-TD-VAE results presented are the best from the hierarchical rollout model. The only version that performs well is  $ S^{Z}_{t=-1}(\text{Cos}) $ which is better on Spearman than the equivalent surprise model and roughly the same on Kendall. The  $S^{Z}_{t=-1}(\text{L2)}$ version has a reasonable correlation but worse performance. Correlations of all results in the $X$ space are weak, as is the correlation between all suspense variants. It should be remembered thought that all of the baselines from the previous chapter including GPT-Sim, in the chart are close to random performance; so though suspense is weak it is a positive correlation.

\begin{figure}[htbp]
\centering
\includegraphics[trim={0.5cm 1.0cm 0.5cm 2.5cm},clip,width=1.0\textwidth]{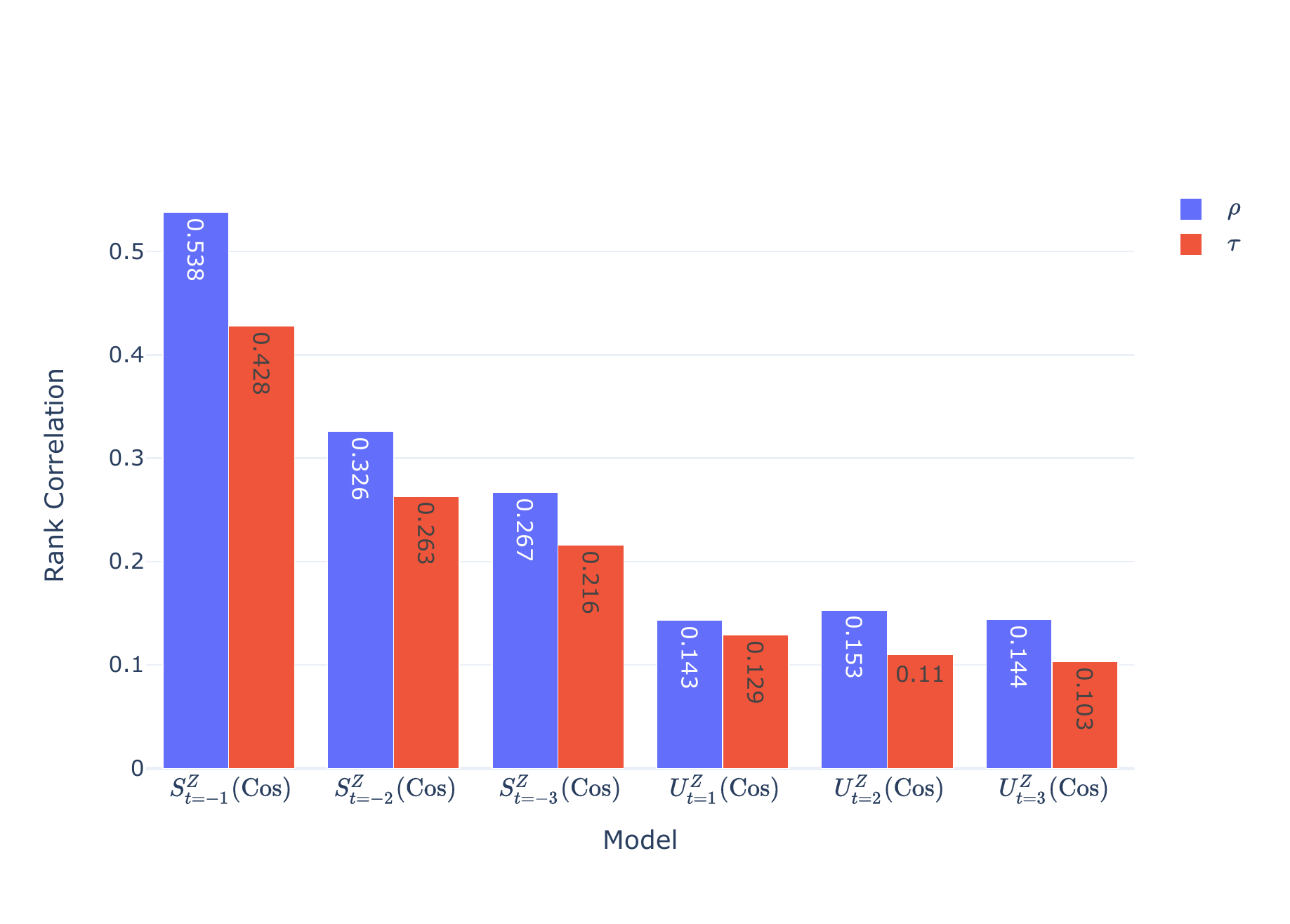}
\caption{The baseline model looking forwards and backwards multiple steps, notice that the indices for suspense and surprise are in different directions. Both have been calculated to $5$ steps but are excluded for brevity as a similar pattern continues.}
 \label{fig:tdvaesuspensesurprisemultistep}
\end{figure}

In addition to these two main models and inferring suspense or surprise one step ahead, it is possible to infer them multiple sentence steps because of the jumpy predictions of the TD-VAE model. For suspense, it means projecting forward the sentence representations multiple steps. For surprise, it means comparing the expectations over the projected vector to the vector representing the actual sentence in the story. Figure \ref{fig:tdvaesuspensesurprisemultistep} shows surprise and suspense metrics for the $Z$ for Cosine predicted $3$ steps ahead and the same behind for surprise. The real predictions extended to $t=5$ and $t=-5$ but this shows the same pattern as the first three and so is not reported. The results from these models show that the performance of surprise deteriorates the further surprise is calculated from the projection point.  This might be expected since it would be thought that predicting accurately further ahead would become more difficult, and so the projected differences would become less reliable. One positive point for the TD-VAE model is it would suggest a relationship between how accurately a model can predict ahead and correlation with human suspense judgements. On the other hand, with suspense, there is little difference between the correlation of the time steps. Unlike surprise, there is no comparison with the actual text. It is simply the TD-VAE projecting forward further to the future. The intuition behind why it might work is that suspense is about the divergence of outcomes. The further the model can predict the outcomes, the more consequential the outcomes will be, hence better-predicting suspense. The TD-VAE model should have an advantage in doing this because it doesn't require the more computationally complicated generation of a tree of sentences. 
 That results are largely similar for various $t$s further out suggests the model cannot learn the more meaningful plot shifts in suspense. It is also possible for the relatively weak results for suspense overall compared to the hierarchical rollout model. The weakness of the \textit{jumpy} predictions is relevant to the discussion on TD-VAE as a planning mechanism in the following chapter.
 
 \begin{figure}[htbp]
\centering
\includegraphics[trim={0.5cm 1.0cm 0.5cm 2.5cm},clip,width=1.0\textwidth]{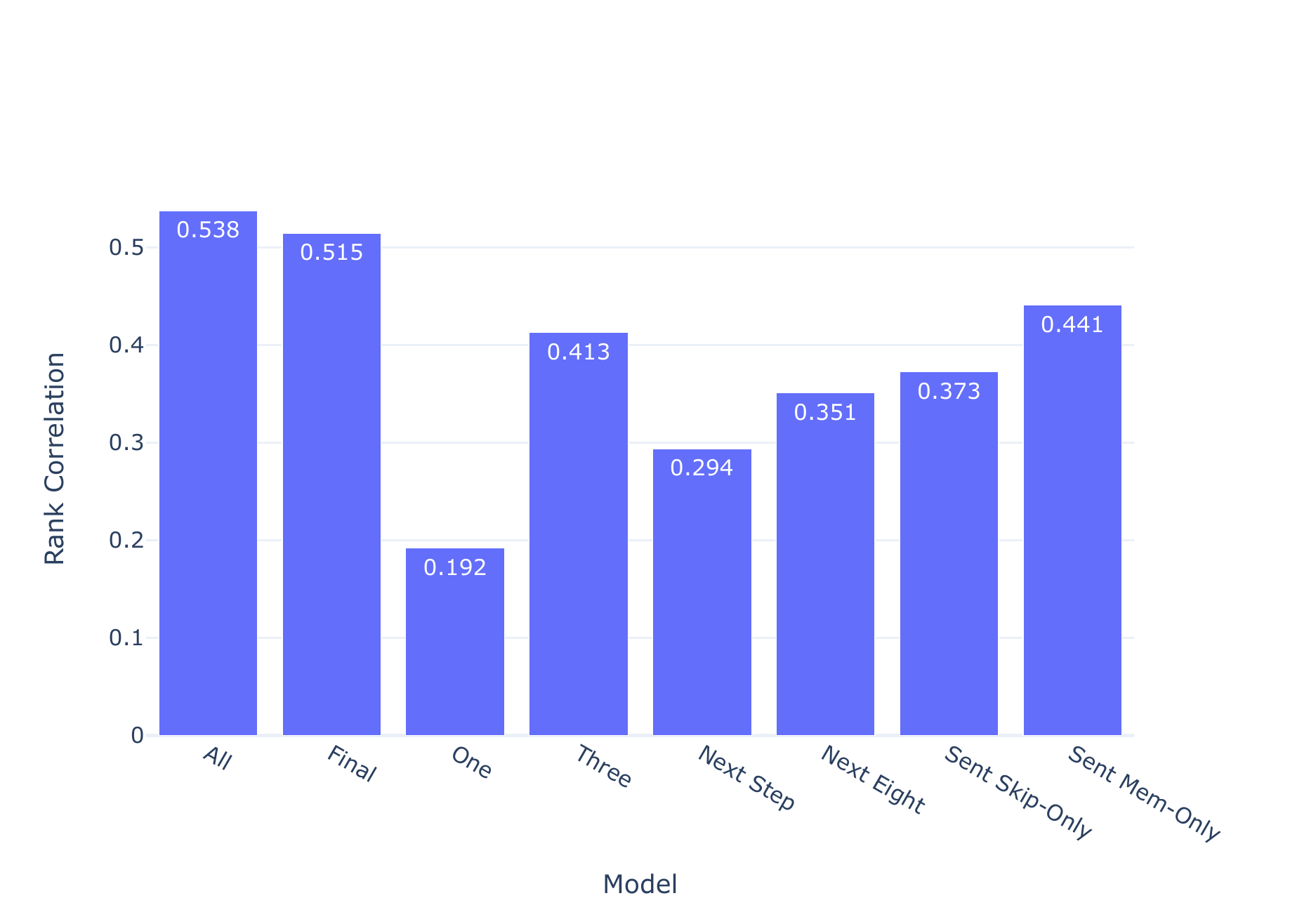}
\caption{Comparison between variants of the architecture on $S^{Z}_{t=-1}(\text{Cos)}$ with Spearman's $\rho$}
 \label{fig:tdvaesuspensesurprisecomparisonspearman}
\end{figure}

 \begin{figure}[htbp]
\centering
\includegraphics[trim={0.5cm 1.0cm 0.5cm 2.5cm},clip,width=1.0\textwidth]{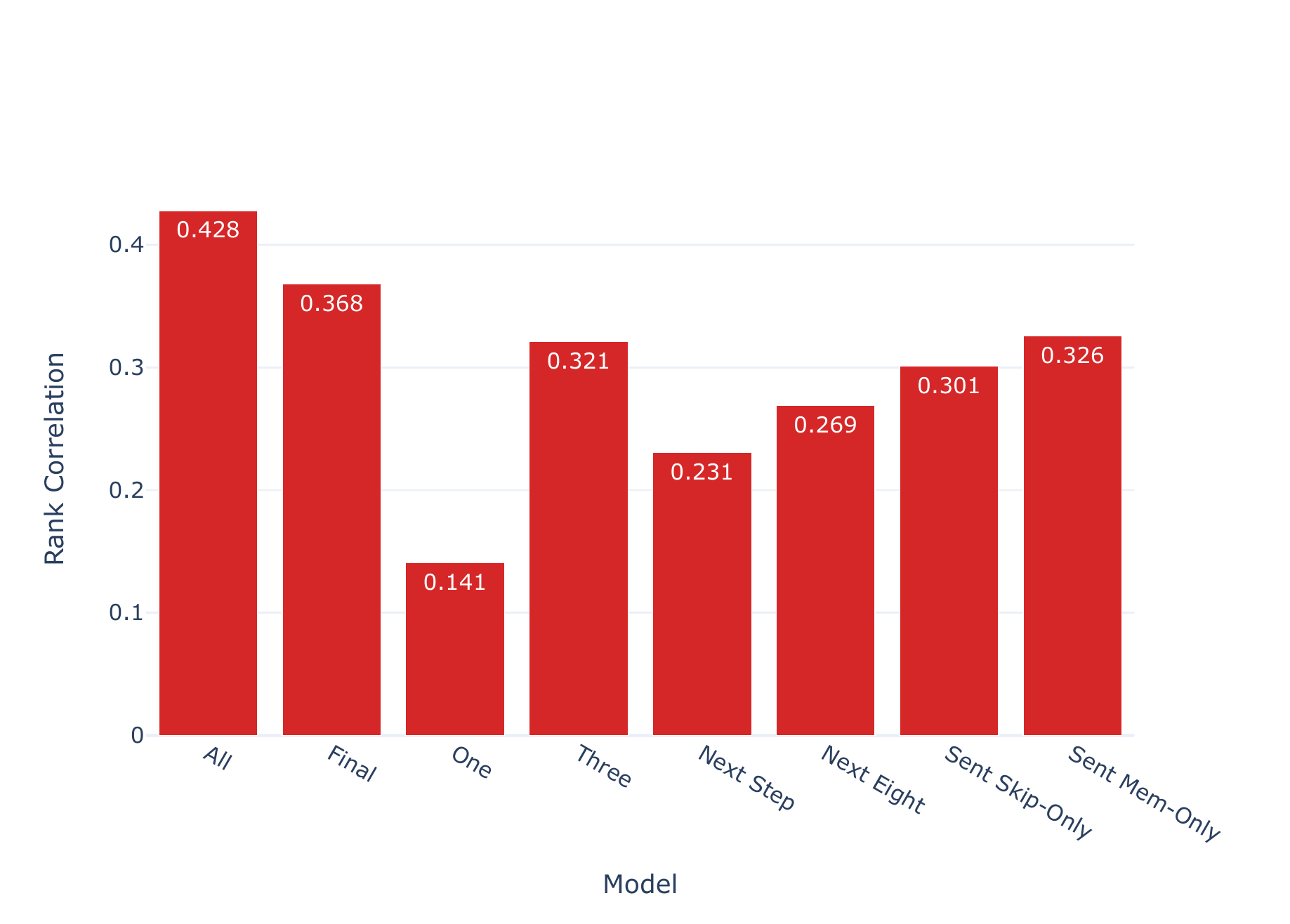}
\caption{Comparison between variants of the architecture on $S^{Z}_{t=-1}(\text{Cos)}$ with Kendall's $\tau$.}
 \label{fig:tdvaesuspensesurprisecomparisonkendall}
\end{figure}

Figure \ref{fig:tdvaesuspensesurprisecomparisonspearman} and figure \ref{fig:tdvaesuspensesurprisecomparisonkendall} shows a comparison between variants of the model that are as follows:

\begin{itemize}
  \item \textbf{All:} The main model with $6$ TD-VAE layers discussed earlier in the architecture section. 
  \item \textbf{Final:} The final model which is identical except only the top TD-VAE layer is used to reconstruct the original input $X$.
  \item \textbf{One:} Contains only a single TD-VAE layer.
  \item \textbf{Three:} Has $3$ TD-VAE layers with the \textit{All} configuration for reconstruction.
  \item \textbf{Next Step:} The TD-VAE layer is only trained to reconstruct the next step, there is not \textit{jumpy} prediction, it is autoregressive.
  \item \textbf{Next Eight:} The \textit{jumpy} prediction is trained up the $8$ sentences ahead rather than the default setting of $5$. 
  \item \textbf{Sent Skip-Only:} A model trained only with the $\ell_{\text{sent}}$
  \item \textbf{Sent Memory-Only:} A model trained only with the $\ell_{\text{cond}}$ loss sentence vector loss.
\end{itemize}

The purpose of these results is to ablate the various components of the model. The main model is expensive in terms of parameters because of the depth of the TD-VAE layers. The \textit{One} and \textit{Three} layer configurations naturally have a smaller number of parameters to test the effectiveness of the extra layers. One criticism of the \textit{jumpy} prediction is that it could be just inferior to an autoregressive model, \textit{Next Step} is TD-VAE operating purely as an autoregressive model. Likewise, although projection $5$ steps ahead seemed to perform best in initial training, the \textit{Next Eight} tests the performance of when the \textit{jumpy} prediction is extended further. While it is possible to report intermediate values such as looking $4$ steps ahead, or with $2$ layers ahead, any combination of the mentioned models, or others such as changing the dimensionality of embeddings training, would be prohibitive. This is because it takes up to a week with $4$ GPUs to train the model and substantial time to run inference over the validation and testset.

\begin{figure}[htbp]
\centering
\includegraphics[trim={0.5cm 0.0cm 0.5cm 2.5cm},clip,width=1.0\textwidth]{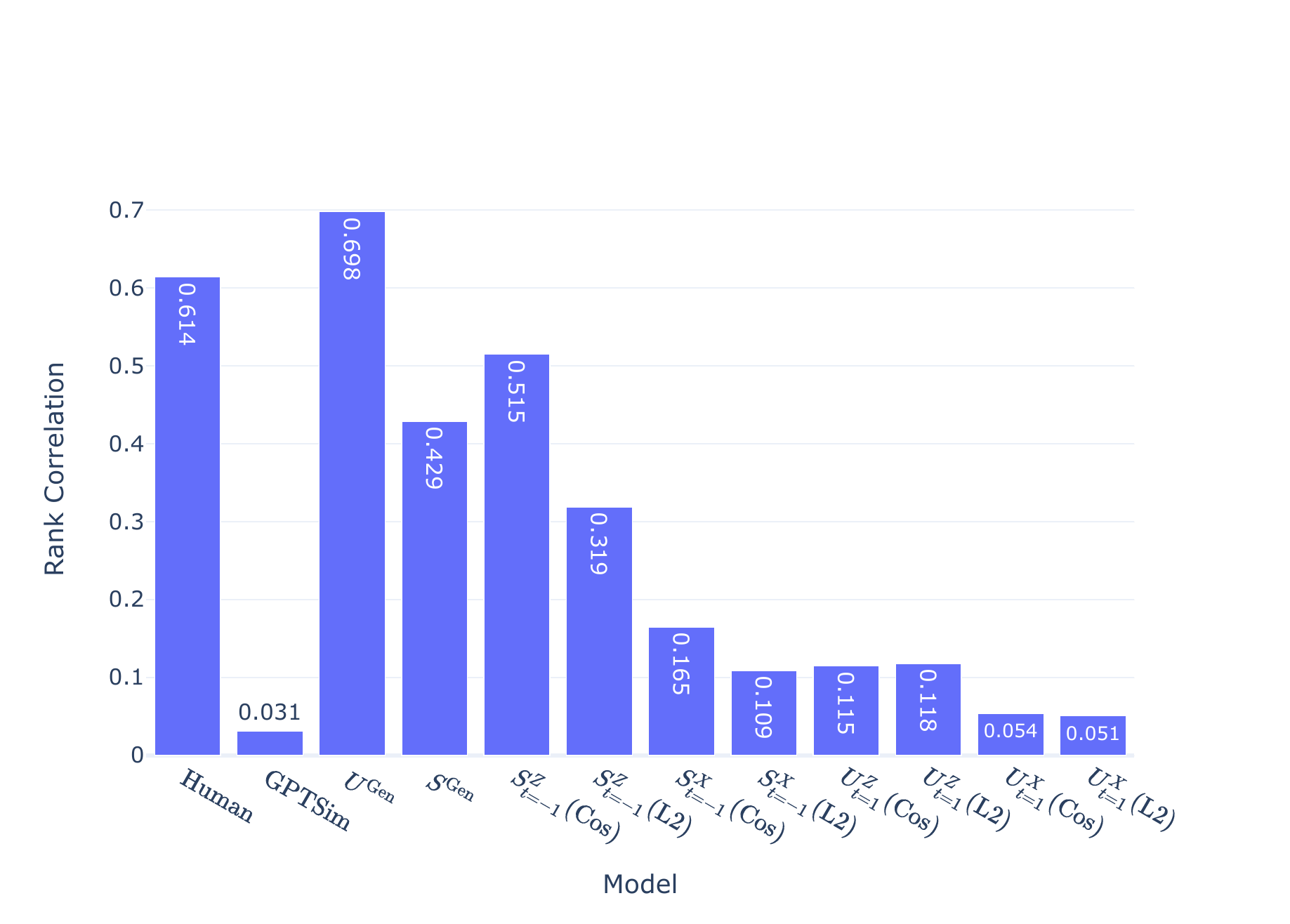}
\caption{Comparison between the base model and the hierarchical rollout results from the previous chapter with Spearman's $\rho$. The model is the Final layer version where only the final layer $Z$ is used to project forward to $X$.}
 \label{fig:tdvaesuspensesurprisefinalspearman}
\end{figure}

\begin{figure}[htbp]
\centering
\includegraphics[trim={0.5cm 0.0cm 0.5cm 2.5cm},clip,width=1.0\textwidth]{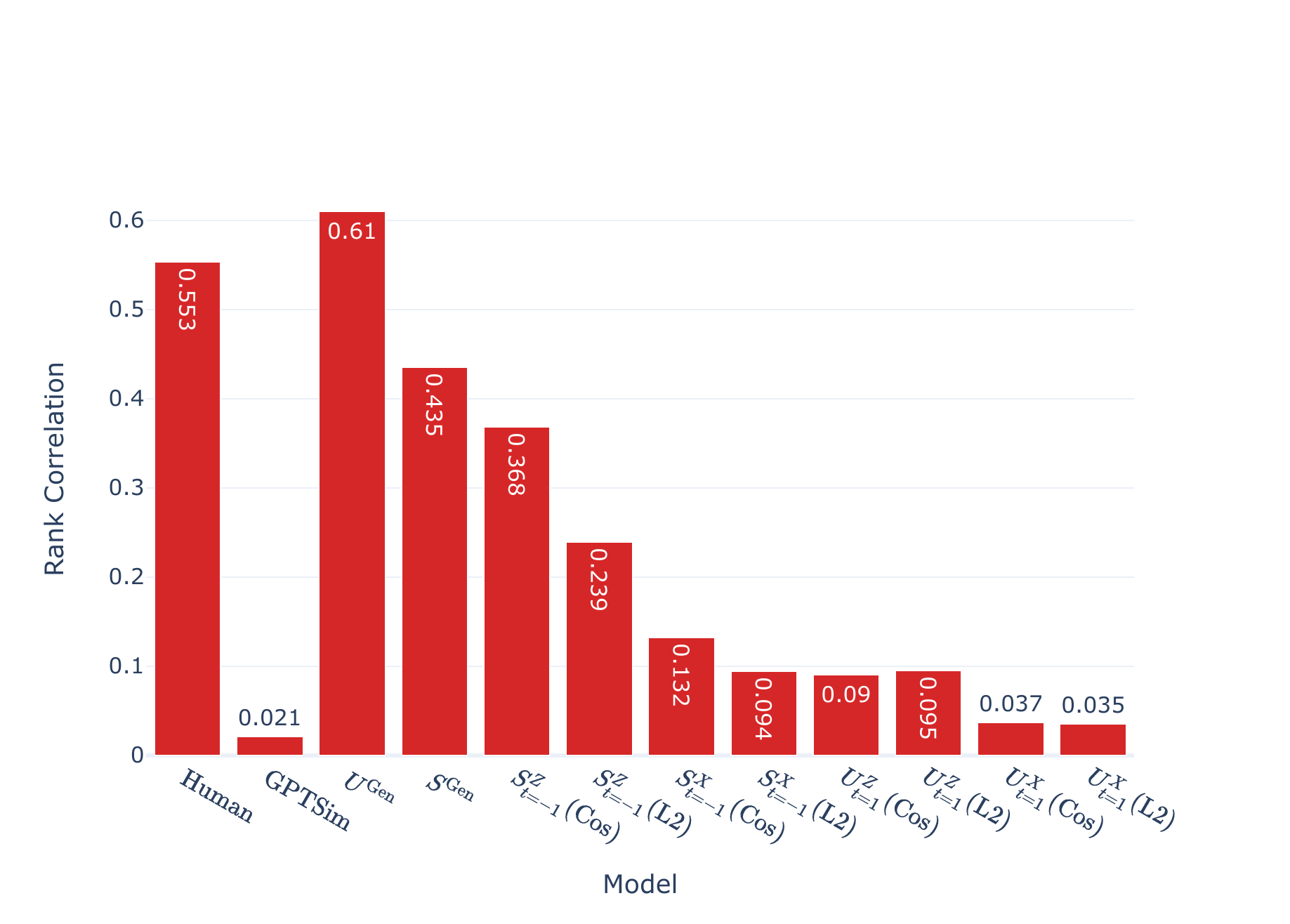}
\caption{Comparison between the base model and the hierarchical rollout results from the previous chapter with Spearman's $\rho$. The model is the Final layer version where only the final layer $Z$ is used to project forward to $X$.}
 \label{fig:tdvaesuspensesurprisefinalkendall}
\end{figure}

To summarise the results in Figure \ref{fig:tdvaesuspensesurprisefinalspearman} and Figure \ref{fig:tdvaesuspensesurprisefinalkendall} on the best performing surprise metric $S^{Z}_{t=-1}(\text{Cos)}$, all of the alternative model configurations perform worse. \textit{One} and \textit{Three} layers both perform worse, with one being worst expected if the increased number of layers improved model performance and also correlated with suspense judgement. Notably, the autoregressive \text{Next Step} performs considerably worse. One possibility for the lower overall performance of the TD-VAE model is it just doesn't work with \textit{jumpy} predictions. If this were the case, then \textit{Next Step} may improve performance, but it doesn't seem to be the case. Whereas, the \textit{Next Eight} configuration already degrades performance, which suggests for this task, $5$ is a reasonable compromise, and longer-range predictions also degrade performance. Finally, both the \textit{Skip-Only} and \textit{Mem-Only} sentence versions also degrade performance, which demonstrates that the combination of losses also has value on the end task.  Overall, these results show that the combination of elements in the main model is task-related and has an improved overall effect on model performance.

The main alternative model is the \textit{Final} model, where only the top TD-VAE layer is conditioned on to reconstruct the $X$. The respective charts are in Figure \ref{fig:tdvaesuspensesurprisefinalspearman} and Figure \ref{fig:tdvaesuspensesurprisefinalkendall}. The results for both of these models are similar to \textit{All} the versions except the \textit{Final} version is slightly worse. To recap the difference between the models is that \textit{All} reconstructs $x$ for all layers of VAE layers of the model whereas \textit{Final} only reconstructs from the top layer. Analysing the varying performance of the different metrics follows considering other variations of the model.

\begin{figure}[htbp]
\centering
\includegraphics[trim={0.5cm 1.0cm 0.5cm 2.5cm},clip,width=1.0\textwidth]{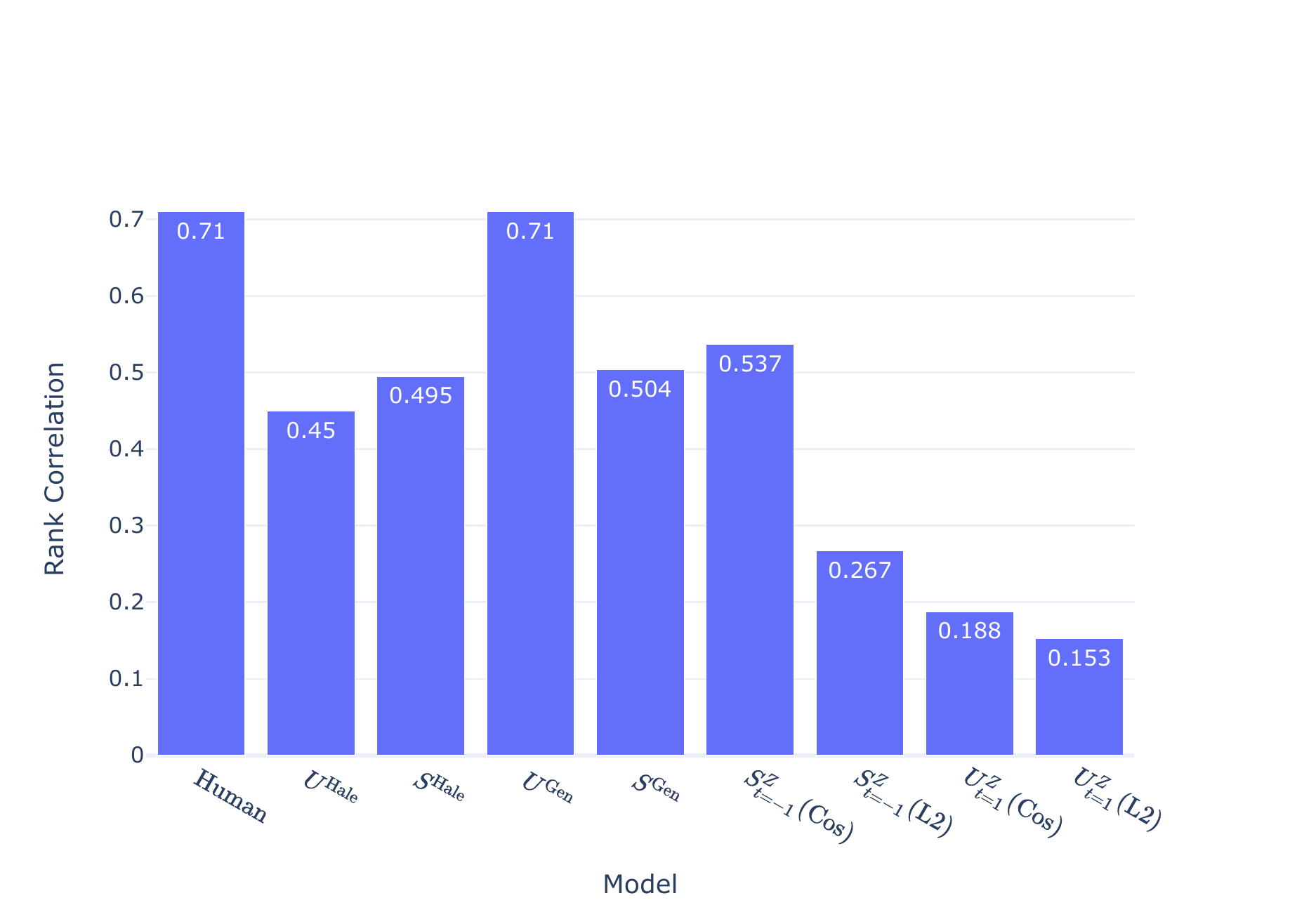}
\caption{Spearman's $\rho$ TD-VAE suspense evaluation for the testset.}
 \label{fig:tdvaesuspensesurprisewithheldspearman}
\end{figure}

\begin{figure}[htbp]
\centering
\includegraphics[trim={0.5cm 1.0cm 0.5cm 2.5cm},clip,width=1.0\textwidth]{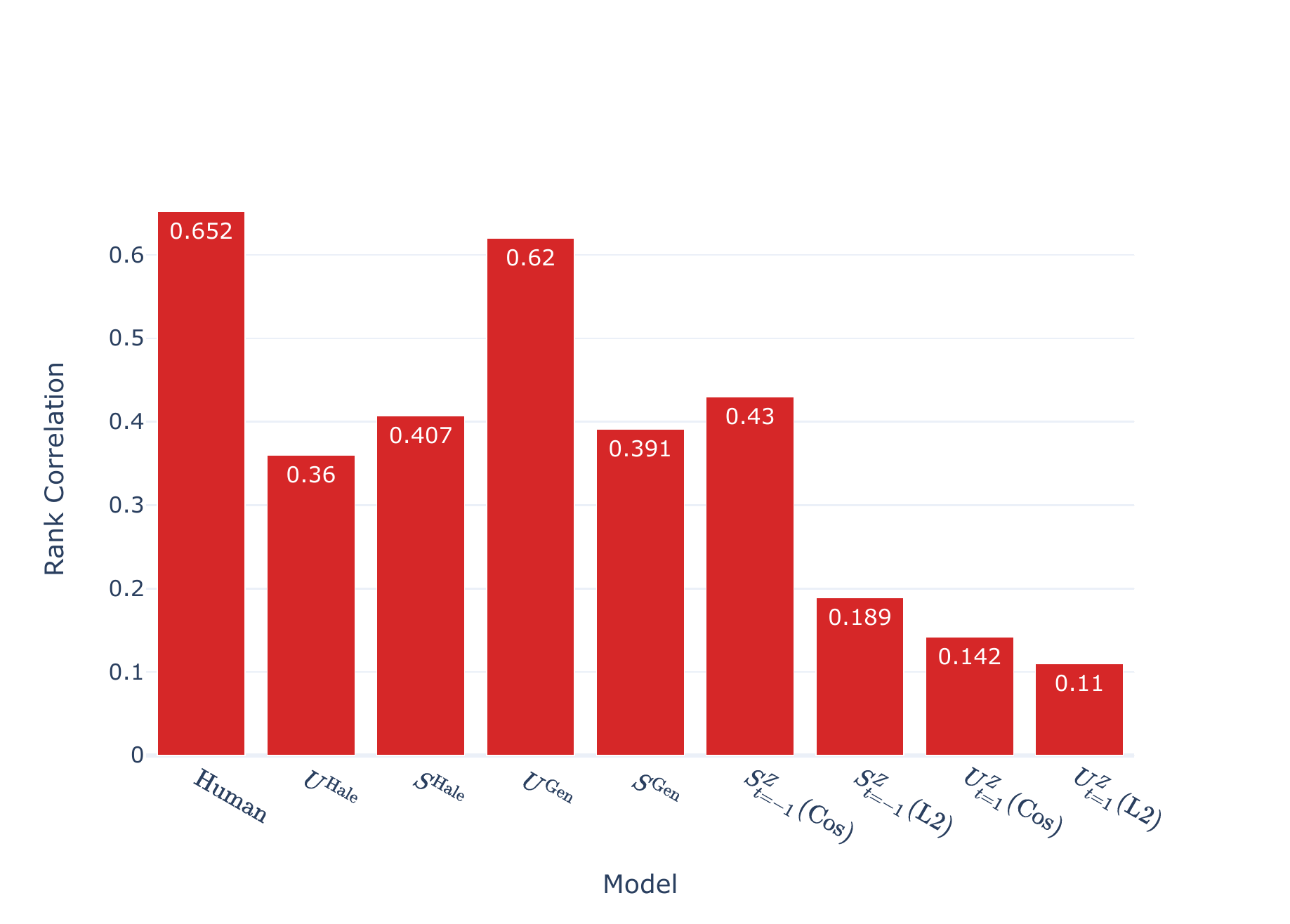}
\caption{Kendall's $\tau$ TD-VAE suspense evaluation for the testset.}
 \label{fig:tdvaesuspensesurprisewithheldkendall}
\end{figure}

All of the results thus far presented have been against the validation set. To complete the results, figures \ref{fig:tdvaesuspensesurprisewithheldspearman}  and \ref{fig:tdvaesuspensesurprisewithheldkendall} show the results of the main \textit{All} model against the test set. The general pattern of results is similar to the validation set. Generally, the results are higher, which is likely a result of the better agreement between annotators. As touched on in the previous chapter, this is likely a function of the Mechanical Turk process. Many of the annotators who completed the test set conducted after the validation set annotation did because they had already completed the training and been paid for the validation set tasks. So a larger group of the tasks were done by a smaller group of annotators who were already pre-approved, hence the higher inter-annotator agreement.

\begin{figure}[htbp]
\centering
\includegraphics[trim={0.5cm 1.0cm 2.0cm 2.5cm},clip,width=1.0\textwidth]{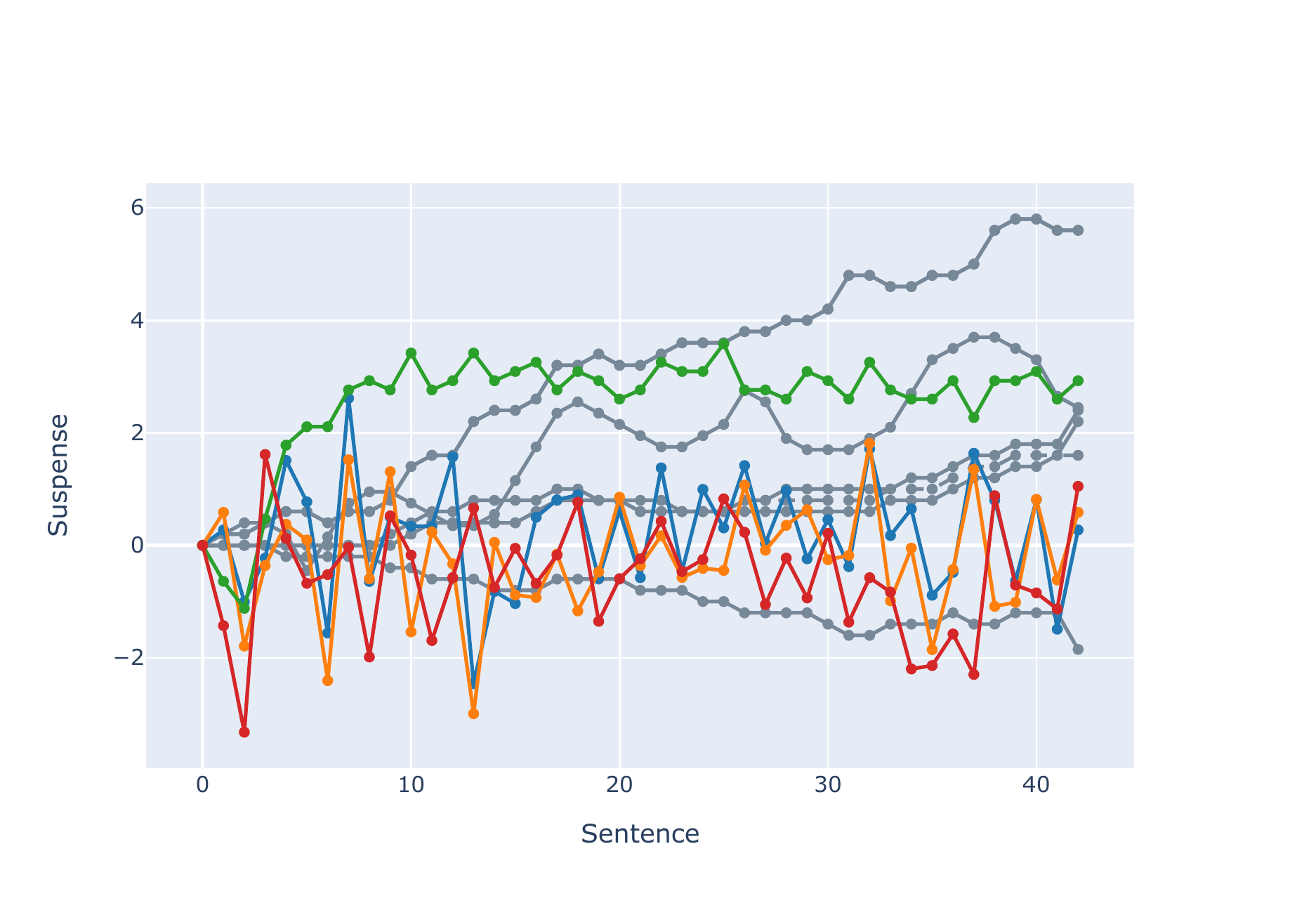}
\caption{Story \textit{900} plot comparing annotations with model predictions: \textbf{\textcolor{annotatedcolor}{Human}}
,\textbf{\textcolor{plotacolor}{$U^{Z}_{t=1}(\text{Cos)}$}}, \textbf{\textcolor{plotbcolor}{$U^{Z}_{t=1}(\text{L2)}$}}, \textbf{\textcolor{plotccolor}{$S^{Z}_{t=-1}(\text{Cos)}$}}, \textbf{\textcolor{plotdcolor}{$S^{Z}_{t=-1}(\text{L2)}$}}.}
 \label{fig:tdvaesuspense_900_plot}
\end{figure}

\begin{figure}[htbp]
\centering
\includegraphics[trim={0.5cm 1.0cm 2.0cm 2.5cm},clip,width=1.0\textwidth]{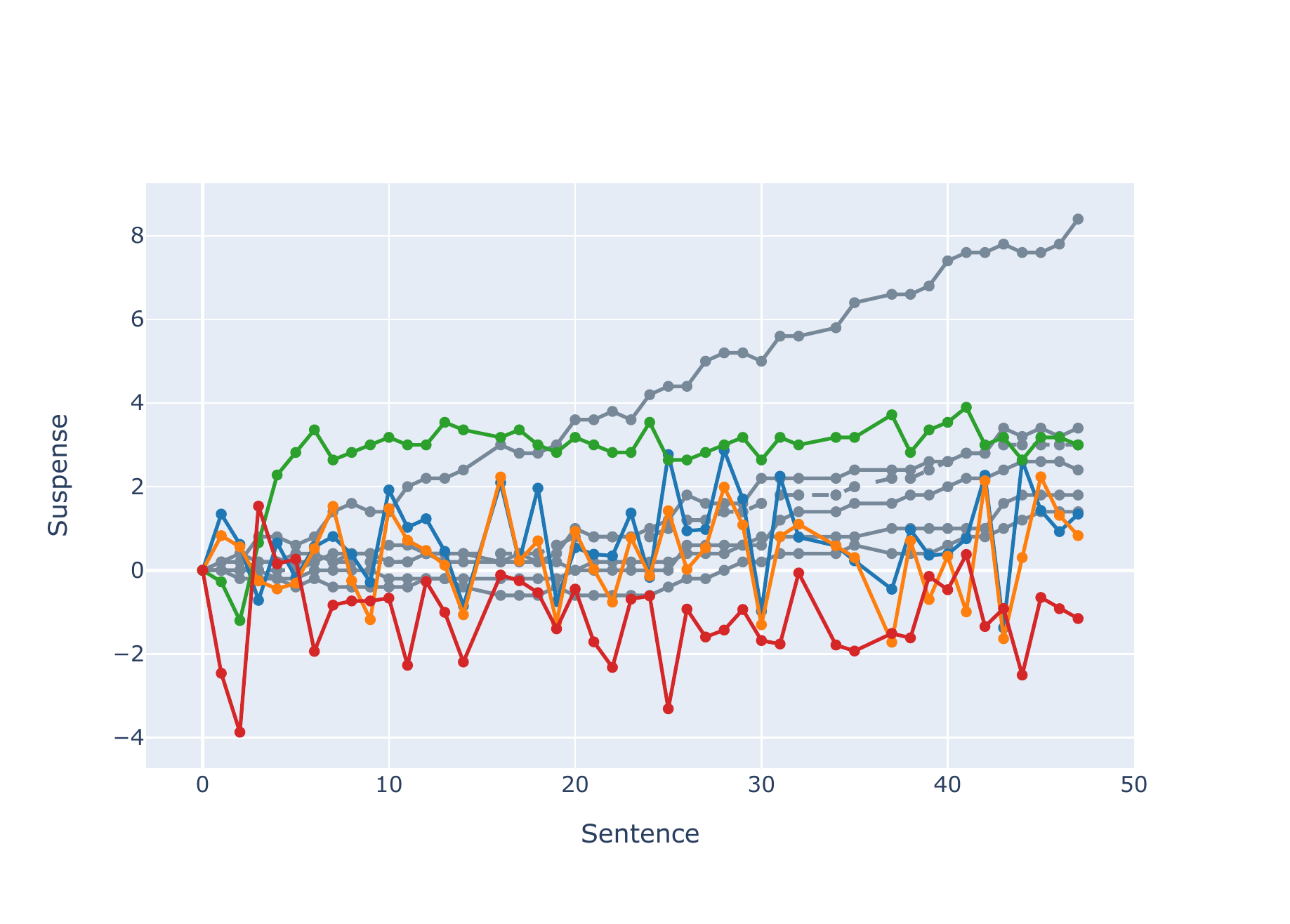}
\caption{Story \textit{6787} plot comparing annotations with model predictions: \textbf{\textcolor{annotatedcolor}{Human}}
,\textbf{\textcolor{plotacolor}{$U^{Z}_{t=1}(\text{Cos)}$}}, \textbf{\textcolor{plotbcolor}{$U^{Z}_{t=1}(\text{L2)}$}}, \textbf{\textcolor{plotccolor}{$S^{Z}_{t=-1}(\text{Cos)}$}}, \textbf{\textcolor{plotdcolor}{$S^{Z}_{t=-1}(\text{L2)}$}}.}
 \label{fig:tdvaesuspense_6787_plot}
\end{figure}

\begin{figure}[htbp]
\centering
\includegraphics[trim={0.5cm 1.0cm 2.0cm 2.5cm},clip,width=1.0\textwidth]{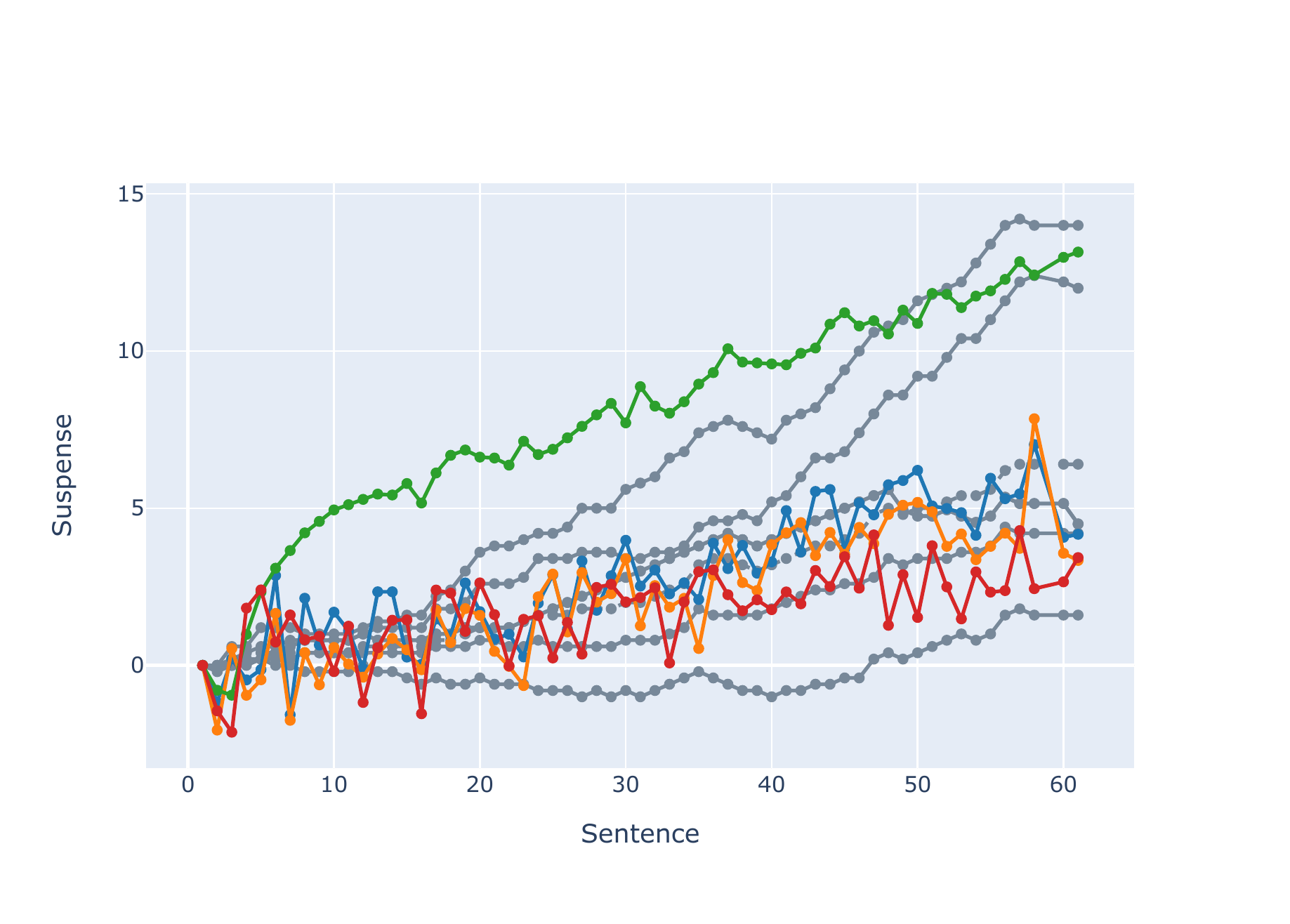}
\caption{Story \textit{14804} plot comparing annotations with model predictions: \textbf{\textcolor{annotatedcolor}{Human}}
,\textbf{\textcolor{plotacolor}{$U^{Z}_{t=1}(\text{Cos)}$}}, \textbf{\textcolor{plotbcolor}{$U^{Z}_{t=1}(\text{L2)}$}}, \textbf{\textcolor{plotccolor}{$S^{Z}_{t=-1}(\text{Cos)}$}}, \textbf{\textcolor{plotdcolor}{$S^{Z}_{t=-1}(\text{L2)}$}}.}
 \label{fig:tdvaesuspense_14804_plot}
\end{figure}

This section concludes by reviewing the overall suspense results for the TD-VAE model. To illustrate the points, there are three other additional more varied suspense plots: Figure \ref{fig:tdvaesuspense_900_plot} shows a weak correlation example where there is low inter-annotator correlation and also a divergence in inferred suspense and surprise. Figure \ref{fig:tdvaesuspense_6787_plot} where there is a clear annotator outlier and a much flatter suspense trajectory. Figure \ref{fig:tdvaesuspense_14804_plot} shows a plot where there is a general upward trend and more of a split in annotation judgement. 

In reviewing this section, the main conclusions are: Modelling in the reconstruction space $X$ generally performs poorly in modelling suspense. Suspense in the latent $Z$ is better than the baselines from the previous chapter but doesn't strongly model suspense and is certainly much worse than the hierarchical rollout model. One thing clearly noticeable in all the plots is there is far less of a general trend and that spikes are far more short term. Unlike the hierarchical rollout model, the models predict more local changes at the sentence level. This is clearer from reading the text, where increases in suspense often correlate to topic or changes rather than general increases in suspense. This ties the model more into the Movie Turning Points identification, in the next section, and the salience work in Chapter \ref{chap:salience}. The positive results are surprise in the $Z$ space and are comparable to surprise results from the hierarchical rollout model. The main difference between the two is that surprise by its definition is grounded in the story since it is the difference between expectations and what happens. As per Chapter \ref{chap:rolloutsuspense}, the difference in expectations correlates with suspense judgements. Roughly speaking, this means that in more suspenseful sections, the distance is greater, and so the model is further away in its prediction. This aligns with the reviewed theory from Chapter \ref{chap:backgroundtheory}, in that it would be expected in more suspenseful parts of the story there would be more uncertainty and also bigger changes of state. Like the hierarchical rollout, model cosine is the strongest distance measure. Speculatively, this could be because cosine distance is a measure of angles. Changes in plot directions are likely in more suspenseful events in a story, which may translate directly to the vector space. Likewise though not reported, dot product which is just an unscaled cosine with magnitude also performs well. The main issue with the results is that the main suspense measures do not perform well; this discussion is revisited at the end of the chapter.

\subsection{Movie Turning Points}

As per the previous chapter, a secondary evaluation is conducted on TRIPOD dataset on Movie Turning Points. Once again, the evaluation on TRIPOD  is motivated by the contrast provided by the task and dataset: The medium are movie plot summaries rather than short stories. More importantly, the task is about identifying specific points rather than general trends. As noted in the previous section, the plots of suspense in the TD-VAE are far spikier and less smooth than the hierarchical rollout model with the \textit{WritingPrompts} task. This could lead them to be bad models for the general trends of suspense, but spikes (or peaks) correlate well to important turning points in movie plots.

\begin{figure}[htbp]
\centering
\includegraphics[trim={0.5cm 1.0cm 2.0cm 2.5cm},clip,width=1.0\textwidth]{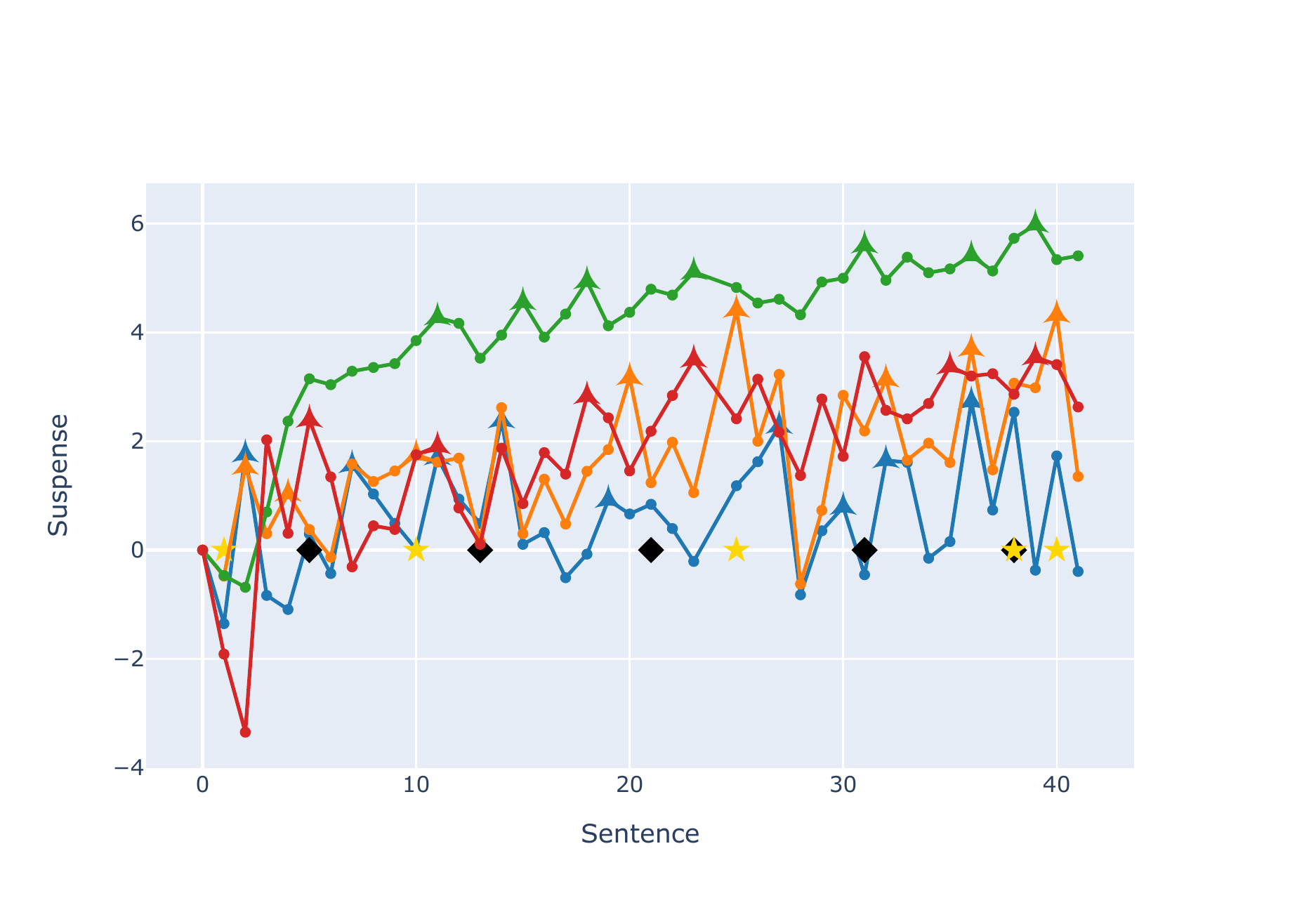}
\caption{\href{https://www.imdb.com/title/tt0100405/}{Pretty Woman} turning points plot: \textbf{\textcolor{plotacolor}{$U^{Z}_{t=1}(\text{Cos)}$}}, \textbf{\textcolor{plotbcolor}{$U^{Z}_{t=1}(\text{L2)}$}}, \textbf{\textcolor{plotccolor}{$S^{Z}_{t=-1}(\text{Cos)}$}}, \textbf{\textcolor{plotdcolor}{$S^{Z}_{t=-1}(\text{L2)}$}}, $\medblackdiamond$ theory baseline, {\color{yellow} $\medstar$} TP annotations. Upward triangles are identified peaks.}
 \label{fig:turningpoint_pretty_woman}
\end{figure}

\begin{figure}[htbp]
\centering
\includegraphics[trim={0.5cm 1.0cm 2.0cm 2.5cm},clip,width=1.0\textwidth]{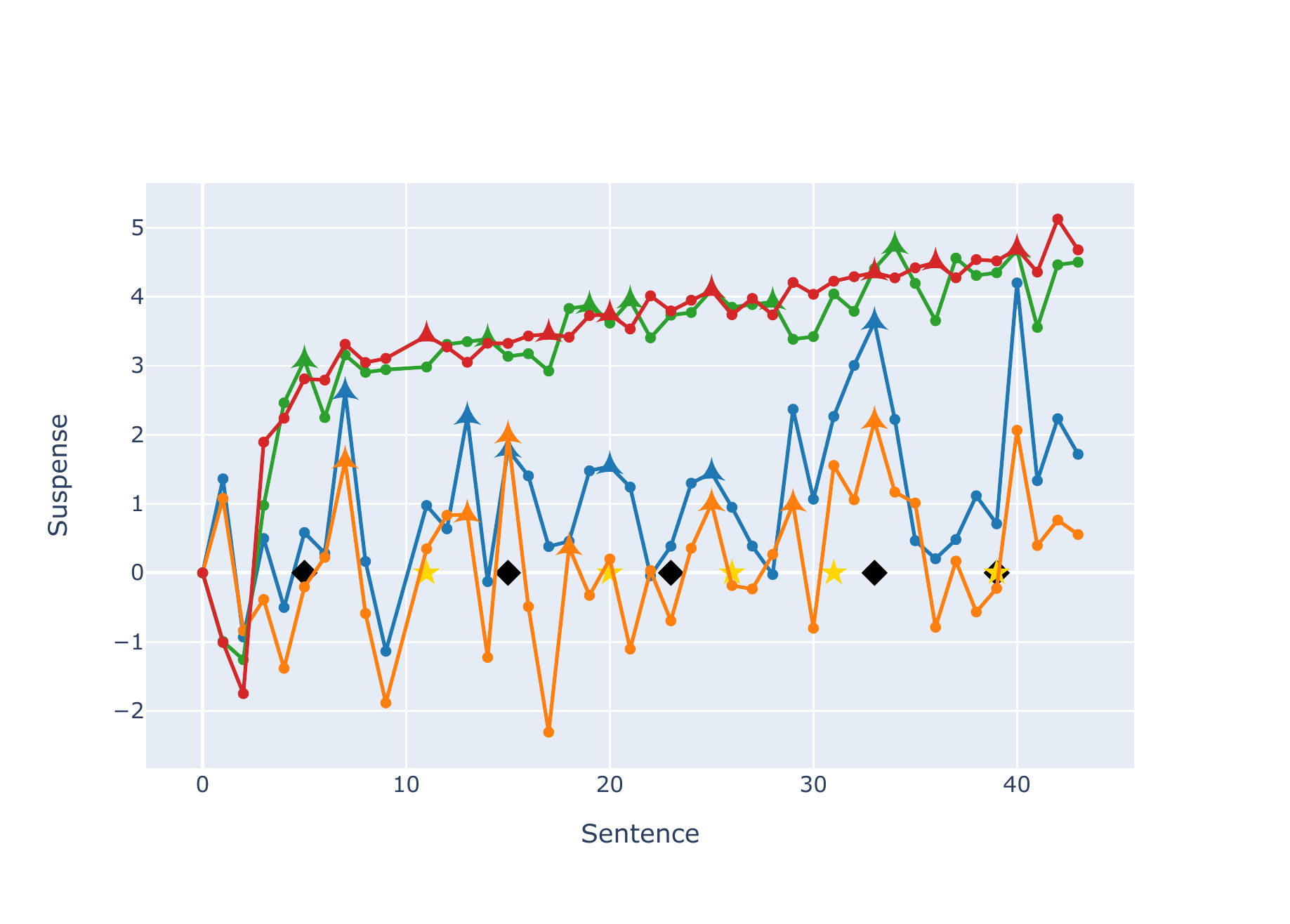}
\caption{{\href{https://www.imdb.com/title/tt1010048/}{Slumdog Millionaire}} turning points plot: \textbf{\textcolor{plotacolor}{$U^{Z}_{t=1}(\text{Cos)}$}}, \textbf{\textcolor{plotbcolor}{$U^{Z}_{t=1}(\text{L2)}$}}, \textbf{\textcolor{plotccolor}{$S^{Z}_{t=-1}(\text{Cos)}$}}, \textbf{\textcolor{plotdcolor}{$S^{Z}_{t=-1}(\text{L2)}$}}, $\medblackdiamond$ theory baseline, {\color{yellow} $\medstar$} TP annotations. Upward triangles are identified peaks.}
 \label{fig:turningpoint_slumdog}
\end{figure}

\begin{figure}[htbp]
\centering
\includegraphics[trim={0.5cm 1.0cm 2.0cm 2.5cm},clip,width=1.0\textwidth]{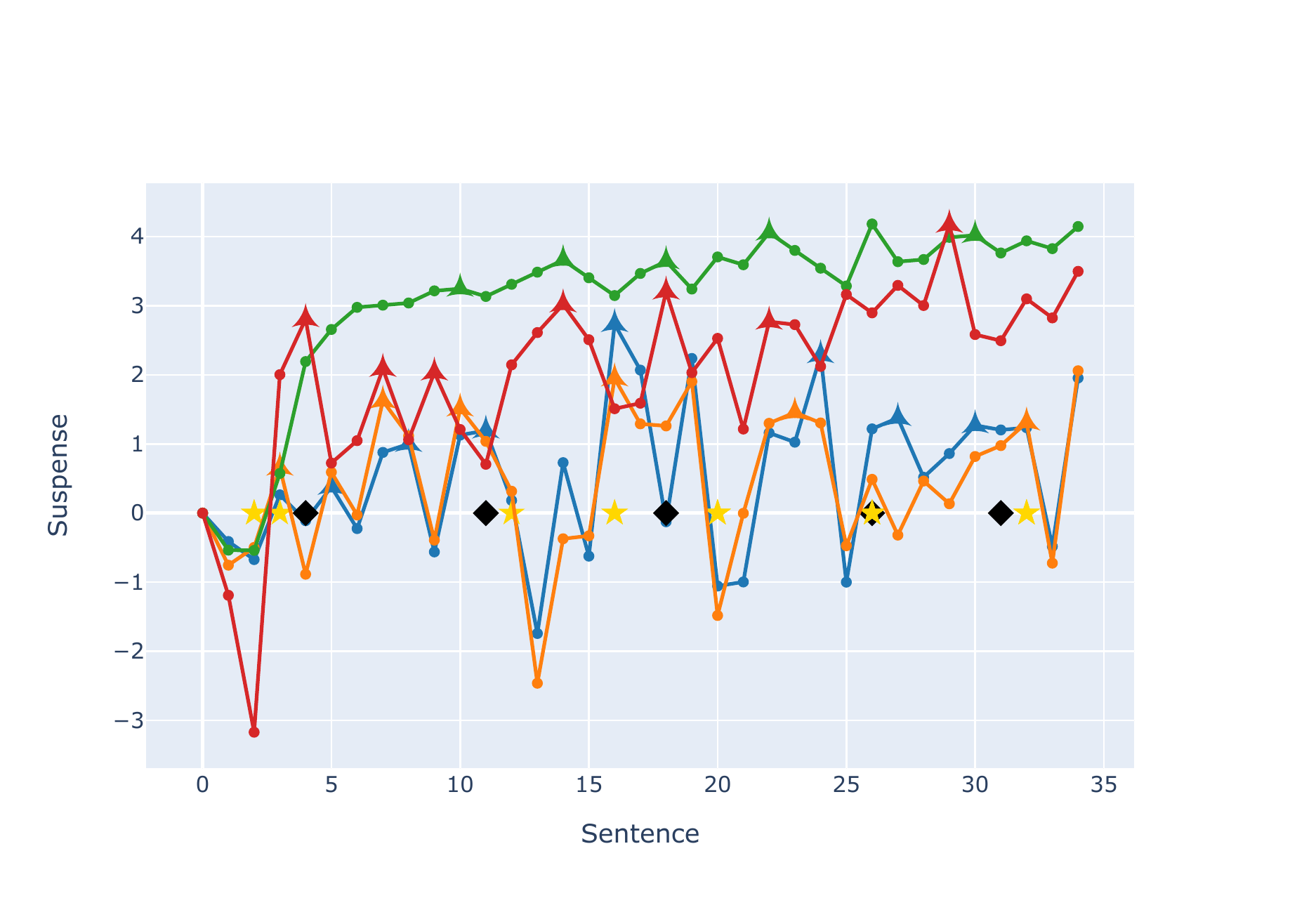}
\caption{\href{https://www.imdb.com/title/tt0073195/}{Jaws} turning points plot: \textbf{\textcolor{plotacolor}{$U^{Z}_{t=1}(\text{Cos)}$}}, \textbf{\textcolor{plotbcolor}{$U^{Z}_{t=1}(\text{L2)}$}}, \textbf{\textcolor{plotccolor}{$S^{Z}_{t=-1}(\text{Cos)}$}}, \textbf{\textcolor{plotdcolor}{$S^{Z}_{t=-1}(\text{L2)}$}}, $\medblackdiamond$ theory baseline, {\color{yellow} $\medstar$} TP annotations. Upward triangles are identified peaks.}
 \label{fig:turningpoint_jaws}
\end{figure}

To illustrate, the three figure represent plots from \textit{Pretty Woman} (figure \ref{fig:turningpoint_pretty_woman}) \textit{Slumdog Millionaire} (Figure \ref{fig:turningpoint_slumdog}), and \textit{Jaws} (Figure \ref{fig:turningpoint_jaws}). The three works represent distinctively different genres in terms of being a romance, uplifting drama and a classic thriller.  The key is the same as in the previous chapter: The black diamonds represent the \textit{theory baseline}, which is to recap where the theory would predict based purely from the \textit{Freytag} model. The \textit{Gold stars} are the consensus judgements for the turning points. The main plots are the same as in the suspense evaluation. The one difference is the \textit{find peaks} function from SciPy \citep{2020SciPy-NMeth} is used to show peaks with upward-facing triangles; this is illustrative and is intended to make it easier to compare the peaks against both the Gold annotation and the theory. 

As noted in the previous section, intuitively, the suspense plots, while not being predictive of suspense, seem to look better against the Gold standard annotations. Also, as noted in the last section looking at individual examples of suspense output, it appears that suspense is more likely to shift with local discourse changes. These changes can be a change of topic or sudden dramatic action. Hence there is a reason to think the model may work better with turning points.

\begin{figure}[htbp]
\centering
\includegraphics[trim={0.5cm 1.0cm 1.5cm 2.5cm},clip,width=1.0\textwidth]{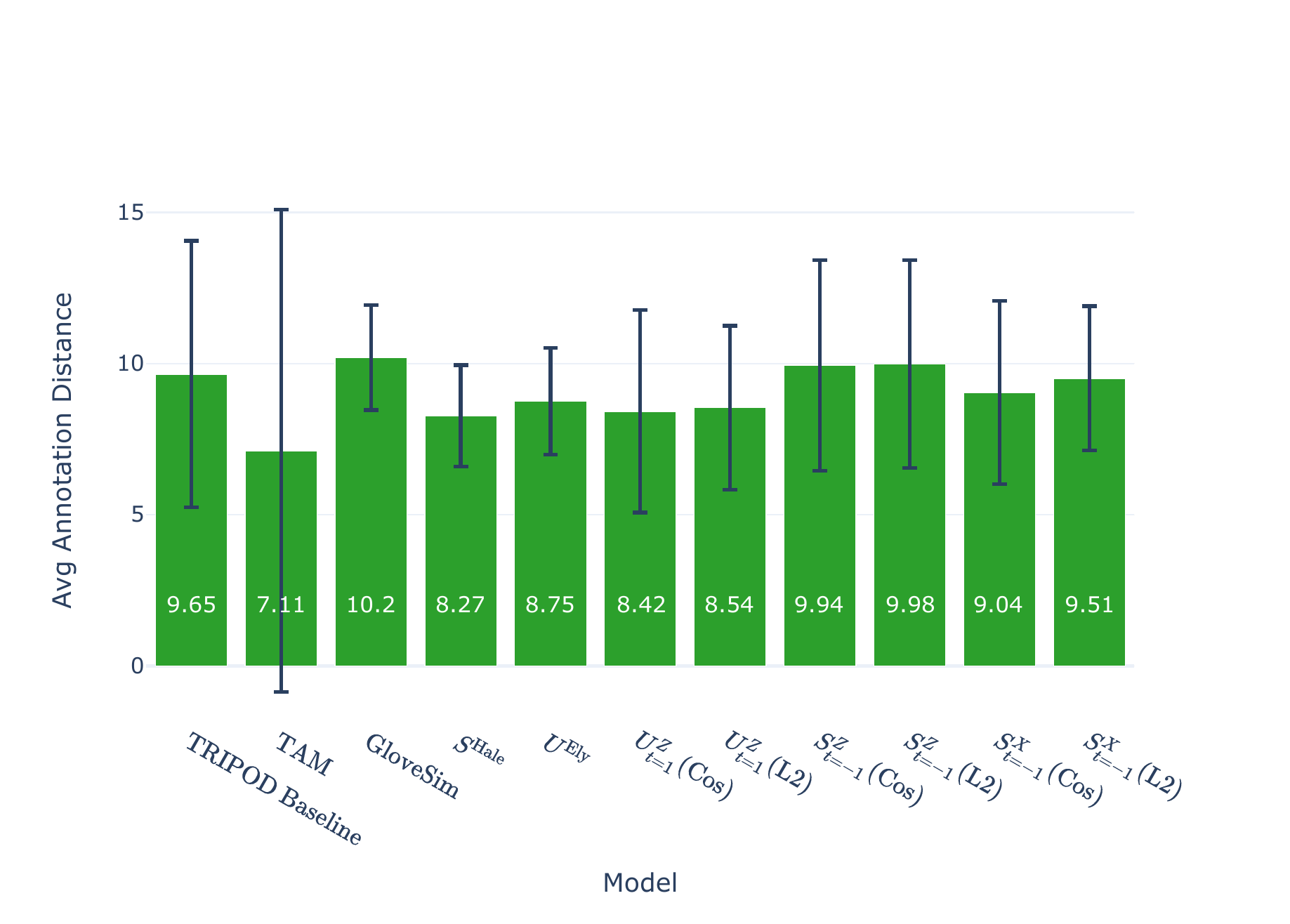}
\caption{The turning points result against the trainingset. The error bars are the standard deviation. Random not shown but published as $37.79$ ($25.33$)}
 \label{fig:turningpointstrain}
\end{figure}

\begin{figure}[htbp]
\centering
\includegraphics[trim={0.5cm 1.0cm 1.5cm 2.5cm},clip,width=1.0\textwidth]{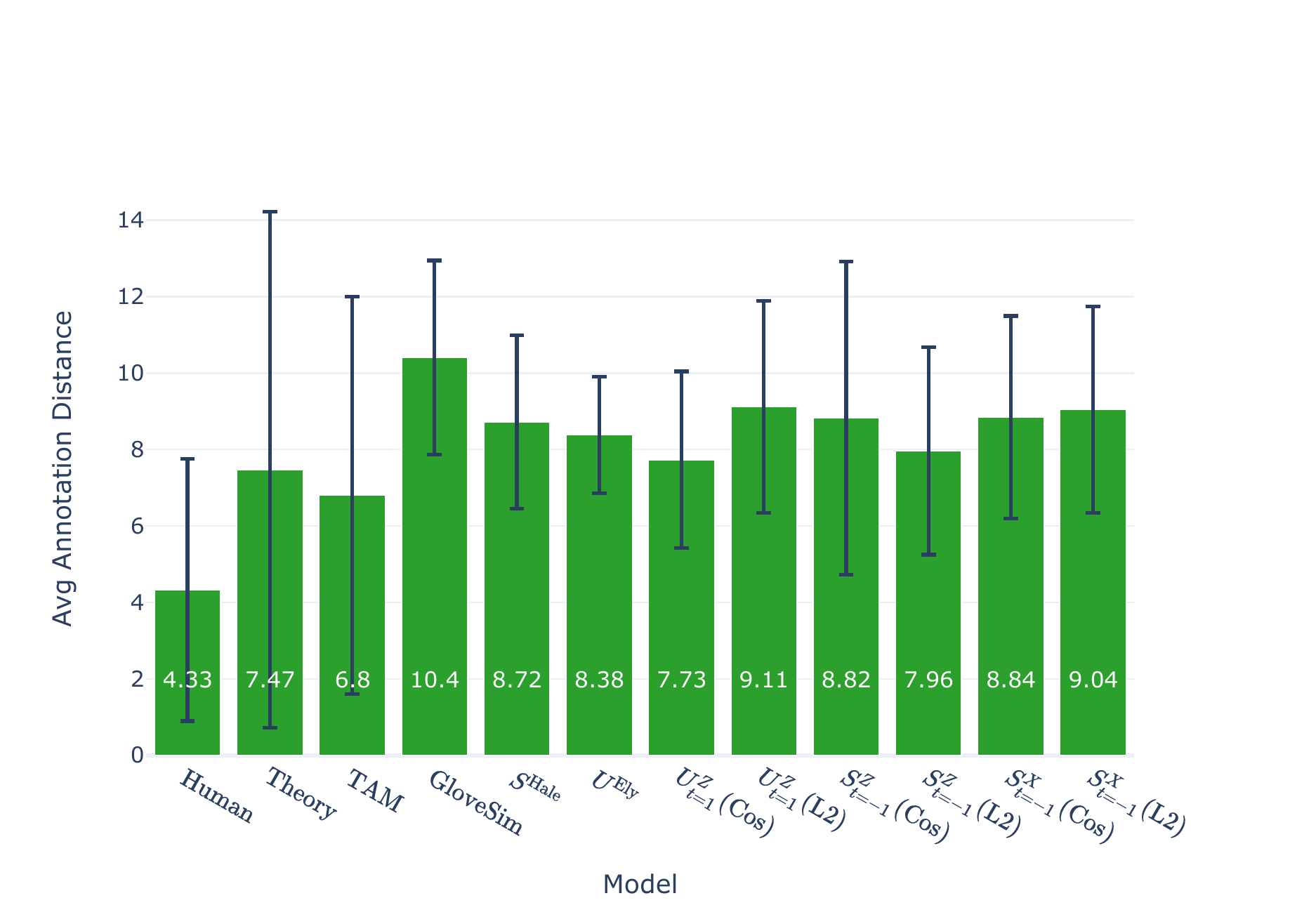}
\caption{The turning points result against the testset. The error bars are the standard deviation.}
 \label{fig:turningpointstest}
\end{figure}

Two results are presented in Table \ref{fig:turningpointstrain} for the training dataset, and Table \ref{fig:turningpointstest} for the test set. The tables show the main models from the TRIPOD papers, baselines such as \textit{GloveSIM} and the best results from the previous chapter, with results from the TD-VAE model. Once again, the main reported measure is the average normalised distance between each turning point and the highest predicted peaks, which means smaller distances are better. Also, as per the hierarchical rollout model, the task is tougher for the TD-VAE models as they are entirely unsupervised, whereas the TRIPOD models are supervised. Nevertheless, the models perform reasonably well. As was hinted at by observing the manual plots, the suspense metrics do perform better than the other models and are nearly identical to the best models for the hierarchical rollout. Once again, these models are worse than TAM, but this is expected as it is a supervised model. Crucially the suspense models are substantially better than the TRIPOD training baseline and far better than a random baseline. As per the hierarchical rollout model, surprise performs well and, though far better than random, is similar to the supervised TRIPOD baseline and better than baselines such as \textit{GPTSim}. So while overall, the TD-VAE model is not better on this task than the hierarchical rollout model, the task shows the model can perform well on the turning point task.

\section{Conclusion}

This chapter attempted to model suspense and surprise via variational models that can directly sample a latent space. To recap, the potential benefit is saving on the computational complexity of generating concrete continuations with a language model such as GPT-2. The results have shown the model for suspense had substantially worse performance than the hierarchical rollout model. Performance of surprise is similar to the hierarchical rollout model. Task performance is comparable and markedly better than random on the secondary movie turning points. The main question is why, as a suspense model, TD-VAE performs relatively poorly. The following chapter extends this one and conditions on TD-VAE latent vectors as a story generation system. Some of the problems encountered in text generation are related to problems with modelling suspense, and so will be discussed together in Chapter \ref{chap:tdvaegeneration}. As will possible technical changes to the model and future work to improve performance for modelling suspense, text generation, and other related tasks.

\chapter{Temporal VAE Generation}

\label{chap:tdvaegeneration}

\section{Introduction}

The previous chapter explored an alternative VAE-based method for modelling suspense. The rationale behind the approach is that latent vector spaces have been shown to represent rich semantic knowledge. VAE methods such as the TD-VAE model can project forward the semantic space. With the Ely model, the capability was used to model a distribution over the expected future state where the variance from the current state would model suspense and the difference between it and the actual state surprise. One of the justifications for the GPT-2 lower layer is that it would be simpler to generate text as GPT is an autoregressive model. While text generation is possible with bidirectional LM such as BERT as with \citet{wang-cho-2019-bert}, these methods are far more cumbersome and don't really improve generation performance. The intention behind adopting GPT-2 as the base was duel purpose: It would provide a base for the hierarchical suspense model. Second, the future projection forward of the latent state by TD-VAE would also function as a planning mechanism. The projected latent state can be both conditioned on directly for text generation or as part of a reranker for selecting generated sentences.

\section{Planning for Story Generation}

One of the problems with language models and story generation is maintaining the long term coherence of a plot \citep{fan-etal-2018-hierarchical}. The attention mechanism within LMs is powerful in attending to the relevant context. This enables LMs to generate text that fits stylistically, is grammatically correct and topically appropriate. However, they often generate coherent text but fail to progress the plot of the story. One answer is to increase the number of parameters in LMs (GPT-3 \citep{NEURIPS2020_1457c0d6} has  $175$ billion parameters); increase the amount of training time, data or diversity of training data; or increase the length of the context. While all these can improve text generation performance, there are diminishing returns for a massive increase in the models' size, memory requirements, and volume of data. The scale of the models is also too large for the ordinary researcher with more limited resources to put their fangs into. 

Chapter \ref{chap:backgroundml} reviewed the long history of both traditional planning systems and case-based reasoning systems in creating primarily story generation systems that can plan longer-term and generate coherent plots. Case-based reasoning is left for  \ref{chap:salience}, which adopts such methods in inferring salience, while this chapter focuses on planning methods.

\begin{figure}[htbp]
\centering
\includegraphics[width=0.80\textwidth]{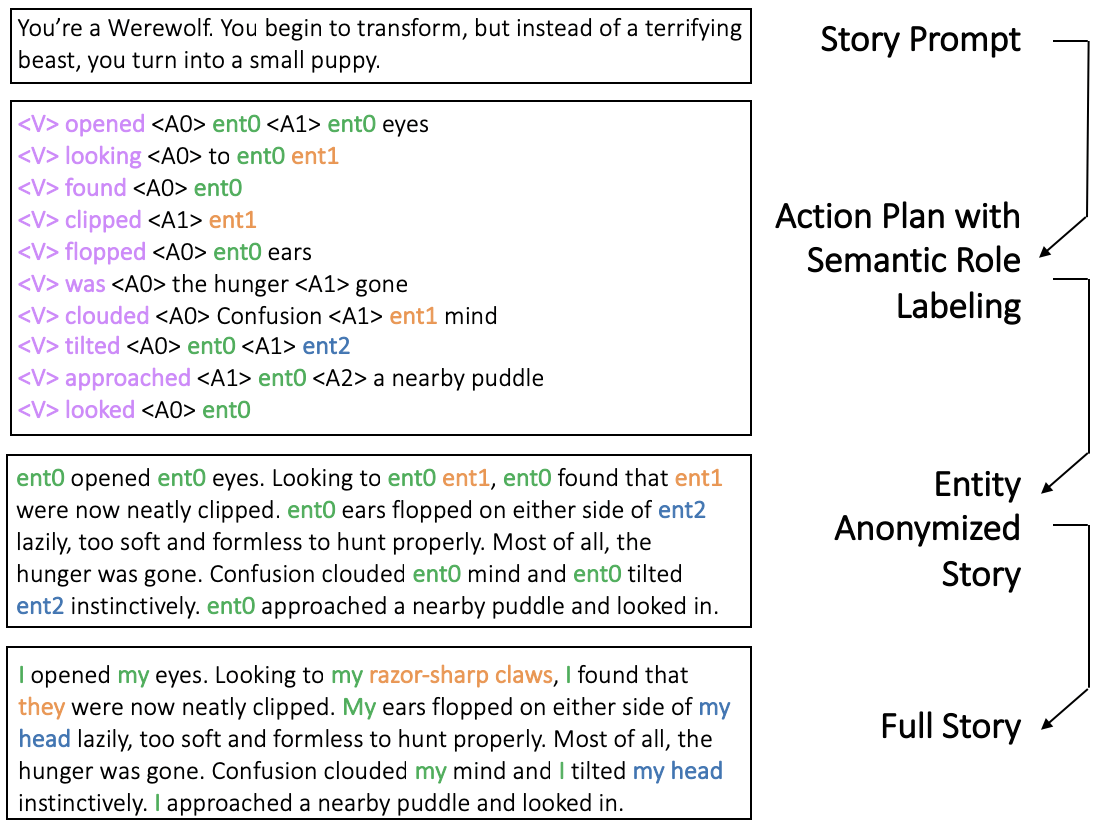}
\caption{A typical multistage pipeline approach reproduced from \citet{fan-etal-2019-strategies}.}
 \label{fig:srl_model}
\end{figure}

Planning methods for neural models have made a resurgence in text generation. The primary approach relevant to the one in the chapter is a multistage pipeline. The motivation is that planning allows for a pipeline with specialised components. The LM can be limited to realising a plan (plot) produced by other components. The text generation advantages of LMs is retained but under the control of the plan. Planning aims to push the model to generate more plotful text (in story generation) without requiring the considerable resource increases needed for the LM to learn the plot development implicitly. Typical of the approach are strategies adopted by \citet{fan-etal-2019-strategies} as illustrated in Figure \ref{fig:srl_model}. In this method, SRL (Semantic Role Labels) are extracted per sentence. Named entities and coreferences are de-lexicalised and replaced with codes, e.g. \textit{ent0} or \textit{ent1}. Then a model, then most typically a transformer, is trained to predict the simplified plot skeleton, and then a second model is trained to condition on and generate the surface level text from each of the skeleton sentences. In the case of Fan et al. there is an additional network to generate and substitute the coreferences generated in the surface text. The general principle behind the approach is to focus the transformer on a simplified event representation so the first network can specialise in generating plots without the distracting surface-level details a single model with a single end to end transformer would have. The method has been demonstrated to produce more interesting stories and cohesive plots. Numerous alternatives to SRL for story generation plans have been tried, including keyword extraction \citep{yao2019plan,goldfarb2019plan}, compression \citep{xu-etal-2018-skeleton}, event representation \citep{Martin2018EventRF,Ammanabrolu_Tien_Cheung_Luo_Ma_Martin_Riedl_2020}, and for non-neural models, graph-based generation \citep{10.5555/2891460.2891543}. An alternative to directly conditioning on the plan is cooperative discriminators \citep{holtzman-etal-2018-learning}, where a batch of generated sentences can be reranked with a separately trained discriminator so that the most coherent continuation can be selected. The approach adds a quality filter to the text generated via sampling from a transformer such as GPT.  \citet{goldfarb-tarrant-etal-2020-content} combine a multistage pipeline with a suite of individually trained re-rankers on top of a multistage SRL planning pipeline. Text is conditioned on a plan, and the re-rankers pick the best generated text.

The relevance of the planning approach is the novelty of the work is to apply a planning mechanism in the latent space with the TD-VAE for suspense from Chapter \ref{chap:tdvaesuspense}. Latent state-space models \citep{koller2009probabilistic} abstract away the relationship from the observed input and output into a probabilistic latent state. This is in principle similar to the story planning skeletons that use representations such as SRL. The limitation to using keywords or SRL is that the model must commit to a specific representation apriori. For example, keyword approaches have proved inferior to more structured SRL approaches because they provide better-structured information. However, as was reviewed in narrative theory connected with suspense, there are often more subtle cues or indicators for suspense. These softer clues are important in story comprehension. An intuition behind planning in the latent space is sentence vectors can represent many semantic and linguistic properties that aren't by a simplified keyword or SRL extraction. It is better to make use of the dense properties of the vector space as reviewed in Chapter \ref{chap:backgroundml} if similar performance can be achieved. The TD-VAE component should learn which are the most relevant to project forward in time, and the GPT model can learn to condition on these vectors to improve text generation. Or the vectors can be used to rerank generated text as per the cooperative discriminators method. The second issue with the SRL type multistage models is that they require significant pre-processing. Extracting named entities and coreference and running SRL is time-consuming and noisy even with the best models. While it is more feasible on short stories such as WritingPrompts it would be far more error-prone to try and apply the methods to a corpus consisting of long novels or screenplays that might have greater than twenty thousand works, each of which may be thousands of sentences. The same applies in inference, where unseen text must be passed through a more complicated preprocessing pipeline. While it is a secondary issue, it would be better if an end to end model could learn to infer the most relevant features implictly without a more complicated engineered pipeline.

\section{VAEs for Text Generation}

Conditioning on the latent space is not a unique idea, and there is a variety of papers that have applied abstract latent spaces: A pioneering approach in the area is \citet{bowman-etal-2016-generating}. The model employed has an encoding RNN model that then projects into a latent space with a VAE type projection. Then this space is directly input to an RNN decoder which can generate text via sampling or beam search methods. \citet{Shen_Su_Niu_Demberg_2018} adapts a similar architecture, with the main architecture difference being the latent space and the decoder, and the text generation are trained separately to resolve problems with the loss degenerating during training; Shen et al.'s work is focused on improving conversational dialogue. \citet{fang-etal-2019-implicit} is a similar model with a VAE layer between an LSTM encoder and decoder but adapted what they term an \textit{implicit VAE} to improve performance. \textsc{Optimus} \citep{li-etal-2020-optimus} is effectively a transformer update of these models in that it replaces LSTMs with BERT for the encoder and GPT-2 for the decoder. The principle is the same in that the BERT [CLS] token is projected into a latent space. The VAE is projected into the latent space of GPT-2; either this projection is added to GPT-2 own embeddings or projected as a history for GPT-2. More recently than the experimental work on TD-VAE. \citet{dai-etal-2021-apo} have adopted a similar approach but with non-linear Poincare embeddings rather than embeddings in the Euclidean space. In a separate task \citet{wang-etal-2019-topic} combines a neural topic model with a sequence model to be able to generate topic-related text where text can be generated from a mixture of topics. All the discussed papers demonstrate that VAE models can generate text that doesn't surpass the performance of non-generative models but can be competitive. The key to the methods, as is the rationale for TD-VAE model, is the latent space provides an intermediate representation between encoder and decoder that acts as a bottleneck and compresses the most salient information. For a non-conditioned generation, the models allow direct generation by sampling from the VAE without the encoder. 

The TD-VAE model specified in Chapter \ref{chap:tdvaesuspense} has a very different architecture to any of these papers. All the presented papers have the VAE as an intermediate layer between an encoder or decoder. The structure of the model is flat in that the LSTM encoder creates a latent state over the sequence of input word pieces, which is then compressed and conditioned on. In contrast, the TD-VAE model is hierarchical. The entire sentence states are fed in as individual sequence elements. Rather than the VAE component just compressing the last context provided, the whole TD-VAE layer projects forward the context in the latent space, which can then be conditioned on. The hierarchical structure should make the model a lot better for long term planning since the VAE model is operating in a higher sentence level of abstraction. The VAE component also has access to the whole sequence of sentences in the context. Parallel with the work on the TD-VAE generation model, Plan-CVAE \citep{wang-etal-2020-plan} was published, which also looks to apply VAEs to planning in text generation. However, the architecture is largely different. Plan-CVAE extracts per sentence keywords as per the planning approaches discussed earlier in the chapter. The keywords are then run through a GRU, and then a CVAE model generates a sentence from each keyword into a surface sentence. The primary difference is the planning mechanism is explicit and not implicit in the vector space. As such, as of when the experimental work was completed, the TD-VAE model represented a unique approach to text generation.

In the previous Chapter \ref{chap:tdvaesuspense} it was mentioned that the TD-VAE model was trained on a diverse set of short and long story corpora to try and improve the generalisation of the model. The same argument applies to story generation. The task evaluated in the chapter is to generate a coherent story continuation to an existing prompt which may be shorter or longer. Just as story comprehension requires anticipating what can and is likely to happen next, and is key to the model of inferring suspense. Generating a story continuation well requires being able to match what the reader thinks is plausible, as well as creating drama and interest. So a more diverse training set should also help improve generation tasks' performance as the model is more likely to have encountered similar examples. It should be noted that \textit{WritingPrompts} has some highly uncommon scenarios as prompts and stories, as does open domain storytelling as a whole. There can always be an unseen situation. The culture of all the datasets is also limited to western genres and works. It is likely to be lacking for typical Chinese stories such as the \textit{Dream of the Red Chamber} mentioned in the introduction. However, increasing the diversity of stories is likely to increase the models' exposure to different character stereotypes, interpersonal interactions, topics, themes and events that may improve generating plausible story continuations even though the story is unique.

\section{Method}

\subsection{Conditional Generation}
\label{sec:conditionalgeneration}

The TD-VAE model is applied in two ways: as a reranker and directly conditioning the latent vectors for text generation. The TD-VAE model, when used directly as a discriminator to rerank generated candidate sentences \citep{holtzman-etal-2018-learning} as part of a beam search of sampled continuations. To allow comparison with two-step planning methods, we also use PSA (Pseudo Self-Attention; \citealt{DBLP:journals/corr/abs-1908-06938}) to inject the expected latent state directly as history in the transformer model. This conditions directly on the latent vectors as a planning mechanism (\textit{TD-VAE Conditioning}). The reranker model can improve text generation and perform well on automated coherence evaluations. In the second approach, the conditioning model performs worse than other baselines because conditioning reduces the diversity of the text generated. The model is same as in Chapter \ref{chap:tdvaesuspense}. Please see Figure~\ref{fig:tdvaehierarchy} for an illustration of the architecture. As such, this section will focus on design decisions made to support text generation and the latent planning process.

In the previous chapter, the $\ell_{\text{cond}}$ was mentioned but not defined. The loss was conceived of as a means of conditionally generating text from the latent sentence vector. However, it was found to also improve performance slightly on suspense and on the turning points task and so was applied there as well. The loss is based on Pseudo Self-Attention (PSA; \citealt{DBLP:journals/corr/abs-1908-06938}) but also shares close similarities with  the latent conditioning from \textsc{Optimus} \citep{li-etal-2020-optimus}. The pseudo-self-attention is in eqn. \ref{eqn:psa}:

\begin{myequation}
 \text{PSA}(X, Y) = \\ \text{softmax}( ( Y W_q) \begin{bmatrix} X W_k'  \\ Y W_k \end{bmatrix}^\top) \begin{bmatrix} X W'_v \\  Y W_v\end{bmatrix}
\label{eqn:psa}
\end{myequation}

The intuition is simple in that the approach is to inject the latent vector state from the sentences into the vector space of the transformer before self-attention. In theory, it allows the transformer to condition on an arbitrary vector like any prepended input text. For the notation, $X$ represents the context sentence in the latent space $e_{t}$, $Y$ represents the embedding from the language model before attention is applied. $W_k$ and $W_v$ denote the normal key and value attention weights of the transformer's self-attention, GPT-2 for this model. $U_k$ and $V_v$ are linear layers to project $X$ into the GPT-2 attention mechanism space. The GPT-2 transformer has $16$ heads, and so the dimensions of the weights is scaled to project the input $X$ uniquely across each of the heads. However, following from Ziegler et al., each layer of GPT-2 shares the same projection.\footnote{Only projecting into the final layer would require more architectural changes since it would then mean the final layer is handling a sequence of a different length from the lower levels.} It is perhaps a weakness of the approach as each layer would be expected to function differently and would ideally have a separate projection. However, GPT-2 having $24$ layers means it massively increases the number of parameters beyond what is feasible. 

The pseudo attention mechanism is an intuitive way around being able to condition a transformer on a latent vector which is more complicated than an LSTM where the latent vector can be prepended and projected as just another element in the input sequence. Ziegler et al. improves on approaches that simply concatenate the latent represent $X$ with $Y$ and only finetune the output decoder, late fusion. It also improves on another method that adds a new \textit{context attention} mechanism on top of the transformer over the GPT-2 output and the latent vector $X$. The likely reason is that by injecting the state into the self-attention mechanism at an earlier stage, the more powerful self-attention mechanism of the transformer can better learn how to integrate the knowledge encoded in the latent vector.

In Ziegler et al's. original formulation, the idea was that the input vectors $X$ would be fixed. It would allow $X$ to represent any predetermined vectors. However, in the case of sentence representation $e_t$, which are the $X$ in the pseudo-self-attention equation, they are jointly trained with the rest of the model. $\ell_{\text{cond}}$ is simply the standard GPT-2 loss applied to a batch of sentences with the appropriate $e_t$ injected via PSA as $X$. The gradients are allowed to back-propagate through the sentence representations transformers. The intuition behind it is that it functions as a form of auto-encoder. The $e_t$ will be finetuned to improve the representation decrease the NLL loss of GPT-2. Suppose the model is able to generate more similar sentences to the real ones by conditioning on the latent vector. In that case, if TD-VAE is able to project forward a relevant future vector, it is a means of controlling future text generation and thus acts as a planning system. 

Given conditioning on a latent vector with a transformer is challenging and far from a solved problem, several other options were explored. One of the biggest problems with large LMs is that they are unfeasible for most research groups to finetune and, even when feasible, hugely costly. PPLM (Plug and Play Language Models, \citealt{DBLP:conf/iclr/DathathriMLHFMY20}) is a method that leaves the existing LM as is and instead relies on making posthoc adjustments to the gradients of the LM to \textit{steer} the output tokens generated by the LM. It decouples control of the text generation from the LM. A discriminator could be trained directly to infer whether the next sentence is coherent and follows from the previous sentence embeddings. The discriminator can directly guide the text generation without needing to fine-tune the weights of the underlying LM. The advantages is it would mean that an alternative sentence encoder could be used that is not able to generate text, which would simplify the model architecture. Larger LMs can be used and reasonably fine-tuned with the resources available. As the plug and play name suggests LMs could be swapped with alternative models. It also potentially reduce the parameter count required by the latent projection methods of PSA. Unfortunately, early experiments with this were not fruitful, but it is an area for future work.

Another approach from \citet{DBLP:journals/corr/RanzatoCAZ15} is reinforcement learning for text generation. In Ranzato et al., with the REINFORCE  \citep{10.1007/BF00992696}, non-differentiable metrics such as BLEU or ROUGE can be loss functions that add flexibility to the model training. The approach has been taken up and shows promising results for tasks like summarisation \citep{DBLP:conf/iclr/PaulusXS18}. The idea for controlling generation from latent vectors would be a sentence would be generated in full, and rather than NLL, the loss would be a distance metric from the expected upcoming sentence embedding. The loss should be more precise and relevant than NLL from individual tokens. The reinforcement learning approach is at least an order of magnitude slower and far more memory intensive, which made the approach unfeasible. It is another area for future work.

\subsection{Latent Planning Generation}

Text generation uses top-p sampling (or nucleus sampling) from \citet{Holtzman2020The} on the underlying GPT-2 model. Conventional language model sampling for text generation samples over the whole probability distribution of the output wordpiece tokens, which for GPT-2 is circa $50$k. The problem is that low probability tokens can cause text generation to degenerate quickly. Both top-p and top-k sampling improve text generation by filtering out the least probable tokens from the selectable output tokens. Top-k keeps only the most likely $k$ word pieces, and top-p keeps only word pieces above a sorted cumulative probability threshold. Keeping only the most probable tokens keeps the quality of the generation higher. Holtzman et al. found top-p sampling generates sentences with better coherence and diversity than top-k sampling \citep{fan-etal-2018-hierarchical,holtzman-etal-2018-learning,radford2019language} methods. For a given next token, the number of reasonable alternatives varies with the context. Therefore top-p is more naturally able to adapt to this and limit the sampling to only reasonable options. The default top-p threshold in our experiments is $0.925$. As it is a sampling method, the text generated will change for the same given context.

\begin{figure}[htbp]
  \centering
  \includegraphics[trim=0 0 180 0,clip,width=0.98\textwidth]{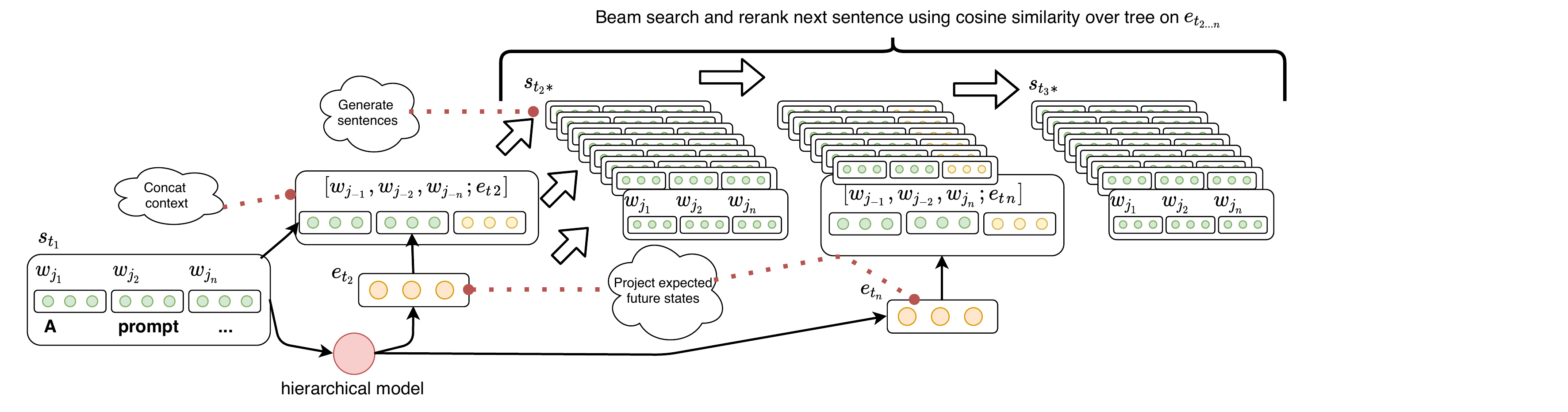}
  \caption{For conditional text generation, the expected $e_{t_2}$ vector is the TD-VAE inferred state for the following sentence. It is concatenated with the existing prompt, and sentences are sampled from the conditioned vector. This happens iteratively as part of a beam search (beam $10$) where the most likely sentences are kept based on how close they are to the expected $e_{t_2}$ vectors using cosine similarity. The final single story is the most like path through the beam search.}
  \label{fig:tdvae_writing}
\end{figure}

Most of the discrete plan-based methods generate text in two phases: The first generates a discrete plan for the stories, the second realises the generated text from the plan. Figure~\ref{fig:tdvae_writing} illustrates how this can be extended with TD-VAE and latent planning. Given a prompt, TD-VAE can project forward the most likely sentence vectors to $n$ steps in the future. These vectors can then be conditioned on using the projection layer described by splicing the projected vector into the hidden state history of the transformer. The advantage is that the latent state information is incorporated as a natural part of the GPT-2 history, and so architectural changes are not required for GPT. Most planning approaches generate a template for the whole story in advance, but this implementation is incremental; TD-VAE projects forward sentences over $n$ sentences incrementally, choosing the next sentence and then projecting forward again. 

In addition to conditioning on these latent vectors, the method employs beam search of $k$ (10 in evaluated stories) reranking approach that generates $n$ (100 per time step for the evaluated stories) candidates sentences and then reranks them by minimising the reconstruction error cosine similarity with the reconstruction from the latent state $z$. For text generation, both conditioning and reranking can be applied independently. However, conditioning worked better when combined with sampling multiple candidate sentences and reranking in practice. Only stories generated with either reranking or reranking with conditioning are judged in the human evaluation. The ranking only model is known as \textit{TD-VAE R}, and the combined is called \textit{TD-VAE C}. The following is more detailed step by step guide to the process:

\begin{enumerate}
  \item \textbf{Sentence Encoding}: The input tokens for each sentence in the context $w_{j_{1}}$, $w_{j_{2}}$, $w_{j_{3}}$ in sentence representations $e_{t_1}$, $e_{t_2}$, $e_{t_3}$ for each sentence in the context. One advantage of a hierarchical model and intermediate representations is the context can be $100$ sentences long which is far longer than a pure transformer with the same memory constraints can have. $100$ sentences is chosen purely as a technical constraint limited by GPU memory during training. The sentence context length was balanced against the number of model parameters such as size of the hidden dimensions and number of layers in trialling alternative model configurations.
\item \textbf{Context encoding:} All the sentence representations are parsed through the belief network ($b$) of the TD-VAE which aggregates the sentence states from $b_{t_{-3}}$, $b_{t_{-2}}$, $b_{t_{-1}}$ to the present. $b_{t}$ The $b_{t}$ represent all the knowledge about the story up until that point.
 \item \textbf{Projection:} $b_{t}$ is projected forward via latent state $z_{t_{1}}$, $z_{t_{2}}$, $z_{t_{3}}$ which is reconstructed to a future $t$, $e_{t_{1}}'$, $e_{t_{2}}'$, $e_{t_{3}}'$. The future states should represent the anticipated future sentence state. This is the planning for the story. In experiments the default is to project $5$ steps ahead.
\item \textbf{Concatenation:} The future $e_{t_{1}}$ are spliced into to word piece tokens from the context, up to $512$ tokens. When concatenated, it is projected with the pseudo-self-attention mechanism, and so it effectively becomes the history (or context) for GPT-2 generation.
\item \textbf{Generation:} $n$ sentences are generated from GPT-2 with the top-p method. $n$ is the number of generated sentences per step, which was $10$ on all the evaluated stories.
\item \textbf{Filtering:} The filtering of the beam search is done by comparing the cosine distance between  $e_{t_{1}}'$ against $e_{t_{1}}$ which is the generated sentence. Keeping only the top $n$ sequences at each step allows is comparable to the cooperative discriminators model where sentences that are more like the expected future state are retained, steering the model to what is expected by TD-VAE.

\end{enumerate}

The model has both a push and pull component where the latent vectors steer sentences generated by GPT-2. The pull is selecting those sampled sentences that more closely match the TD-VAE \textit{plan}. Iteratively applying the model from a prompt can create a story of arbitrary length. Selecting a single story is selecting the highest-ranking continuations from across the beam. Overall the model can function as a plan based system in the latent vector space and apply them with the transformer model. It thus represents a departure from previous work.

As per the scope discussion in the 

\section{Experiments}

\subsection{Automatic Evaluation}

Evaluating generated stories is inherently hard. A well-written story needs to be stylistically and grammatically correct. They also have to be able to advance a consistent plot while having elements of suspense and surprise that create dramatic tension. It is inefficient and expensive to rely solely on human feedback. Therefore it was felt necessary to conduct an automatic evaluation to verify the potential of the TD-VAE model before doing an entire evaluation exercise on generated stories. 

Often word overlap metrics such BLEU \citep{xu-etal-2018-skeleton} and ROUGE \citep{lin-2004-rouge}, which are common for evaluation in areas such as machine translation and summarization, have been applied to story evaluation. However, recent work by \citet{guan-huang-2020-union} has found that these and related measures correlate poorly with human judgement and not that much above human judgement. The same applies to more recent transformer metrics such as {BERT}Score \citep{DBLP:conf/iclr/ZhangKWWA20}. \citet{guan-etal-2021-openmeva} develop a framework for automatic meta-evaluation of metrics for story evaluation. The best performing measure is \textsc{UNION} \citep{guan-huang-2020-union}, but this only scores around $0.3$ in human judgement correlation and most other measures are far worse.\footnote{\textsc{UNION} would have been a relevant automatic evaluation metric, but the code was not available at the time this work was done.} It makes intuitive sense since most of the measures, especially BLEU, ROUGE or perplexity, measure how similar text is at a surface level. As taking generated stories from a common prompt beyond reusing characters of place names, there's no reason for thinking an interesting story would overlap much with a Gold standard. Well written stories are inherently likely to diverge from one another.

An alternative approach was adopted based on cloze and swapping. ROC cloze endings have been widely used \citep{schwartz-etal-2017-effect} for our evaluation. Rather than evaluate on artificial stories of only five sentences, evaluation is on the full stories of the \textit{WritingPrompts} dataset. For the swap,  the evaluation randomly swaps two sentences in each story. For mutation cloze, a randomly sampled sentence generated by GPT-2 at that point in the story, conditioned on the story up to that point. Another way of selecting cloze stories is to randomly select from the corpus. However, this is too easy a task for these more powerful neural models. Generating a sentence with the underlying GPT-2 model is far more challenging as the sentence will fit stylistically and is more of a test for semantic understanding and plot progression. There are two versions of the task: The easy version is to distinguish the original from the modified version of the story. For the harder version, the task is to identify which of the sentences have been modified (mutated or swapped) by determining if they are in the top $K$ least likely candidates across the whole story with varying $K$ values reported. The changed sentences in the hard task are identified by the least likely continuation just by the probability or vector distance depending on the model. The automatic evaluation is not an overall evaluation of a story but a more restricted evaluation of whether TD-VAE can judge coherence.

\begin{table}[]
\centering
\begin{tabular}{ccccccc}
\toprule
\textbf{Task} & \textbf{Model} &  \textbf{Easy} $\mathbf{\pm}$ \textbf{(CI)} & \textbf{K-1} $\mathbf{\pm}$ \textbf{(CI)} & \textbf{K-5} $\mathbf{\pm}$ \textbf{(CI)} & \textbf{K-10} $\mathbf{\pm}$ \textbf{(CI)} & \textbf{Avg} $\mathbf{\pm}$ \textbf{(CI)}  
 \\ \midrule
\textbf{Swap 2}  & Rand                              & .500 (.050)   & .037 (.020)        & .190 (.040)        & .377 (.049)  & .276 (.045)
\\
- & LM                           & .532 (.050)   & .045 (.022)        & .235 (.043)        & .479 (.050)   & .322 (.047)     
  \\

- & LSTM                               & .963 (.020)   & .053 (.023)         & .211 (.041)        & .419 (.049)   & .412 (.049)      
\\

- & Trans                               & .945 (.024)   & .044 (.022)        & .269 (.045)         & \textbf{.493 (.050)} & .438 (.050)   \\

- & TD-VAE                                & \textbf{.969 (.018)}   & \textbf{.054 (.024)}         & \textbf{.277 (.045)}         & .489 (.050) & \textbf{.447 (.050)}  \\
\midrule

\textbf{Mut 1} & Rand                              & .500 (.050)  & .018 (.014)         & .094 (.030)         & .188 (.039) & .200 (.040) \\

- & LM                               & .712 (.045)   & .030 (.018)         & .153 (.036)        & .293 (.056) & .297 (.046) \\      

- & LSTM                         & .613 (.049)
& .019 (.015)         & .149 (.036)         & .453 (.050)  & .309 (.046)
 \\

- & Trans                            & .951 (.023)   & .085 (.029)         & \textbf{.268 (.045)}         & .421 (.049)  & .431 (.050)     
 \\

- & TD-VAE    & \textbf{.974 (.017)}  &   \textbf{.068 (.026)}   & .257 (.044)          & \textbf{.471 (.050)}  & \textbf{.443 (.050)} \\

 \bottomrule
\end{tabular}
\caption{Results of the hard cloze and swap tasks for models trained on WritingPrompts. Confidence Interval at 0.05.}
\label{tab:cloze_res_hard_wp}
\end{table}

\begin{table}[]
\centering
\begin{tabular}{ccccccc}
\toprule
\textbf{Task} & \textbf{Model} &  \textbf{Easy} $\mathbf{\pm}$ \textbf{(CI)} & \textbf{K-1} $\mathbf{\pm}$ \textbf{(CI)} & \textbf{K-5} $\mathbf{\pm}$ \textbf{(CI)} & \textbf{K-10} $\mathbf{\pm}$ \textbf{(CI)} & \textbf{Avg} $\mathbf{\pm}$ \textbf{(CI)} 
 \\ \midrule
\textbf{Swap 2}  & Rand                        &   .500 (.050)   & .037 (.020)        & .190 (.040)        & .377 (.049)  & .276 (.045)
\\

- & LM                                 & .545 (.050)  & .045 (.022)         & .239 (.043)        & .473 (.050) & .326 (.047) \\
- & LSTM                                & .946 (.024)  & .035 (.019)        & .226 (.042)        & .441 (.050)  & .412 (.049)        \\

- & Trans                                  & .953 (.022)   & \textbf{.058 (.024)}         & .238 (.041)        & .427 (.049) & .419 (.049)  \\
- & TD-VAE                                 & \textbf{.979 (.000)}   & \textbf{.058 (.024)}         & \textbf{.279 (.045)}         & \textbf{.496 (.050)}         
&  \textbf{.453 (.050)} \\
\midrule

\textbf{Mut 1} & Rand                                 & .500 (.050)  & .018 (.014)         & .094 (.030)         & .188 (.039) & .200 (.040) \\

- & LM                                 & .640 (.048)   & .026 (.017)        & .121 (.033)        & .255 (.044)  & .261    (.044)     
 \\

- & LSTM                     & .932 (.026)   & .106 (.031)       & .187 (.039)        & .303 (.046)  & .382 (.049)
 \\ 

- & Trans                                 & \textbf{.966 (.019)}   & .020 (.015)        & .217 (.042)        & .379 (.049) & .396 (.049)    
 \\

- & TD-VAE                                 & .931 (.026)   & \textbf{.134 (.035)}         & \textbf{.356 (.048)}         & \textbf{.556 (.050)}  & \textbf{.494 (.050)}       \\

 \bottomrule
\end{tabular}
\caption{Results of the hard cloze and swap tasks for models trained on all datasets. Confidence Interval at 0.05}
\label{tab:cloze_res_hard_all}
\end{table}

The TD-VAE models evaluated are the same as the \textit{All} $5$ layer TD-VAE model from Chapter \ref{chap:tdvaesuspense}. The results are given in Table~\ref{tab:cloze_res_hard_wp} for the model trained only on the WritingPrompts dataset, and in Table~\ref{tab:cloze_res_hard_all} for the model trained on all datasets. Random results are obtained by randomly choosing the $K$ most likely sentences. The evaluation is on 400 random stories of between 25 and 75 sentences long from the WritingPrompts testset. The LM models results are calculated using the perplexity of a two-sentence sliding window. Each of the other models' probabilities are based on the softmax of a distance metric. Therefore, the mutated or swapped sentence should be the furthest away from expectations using cosine distance. For the hierarchical LSTM, the transformer discriminator is used to predict the correct answer as per conditional generation, and for the LM, lower perplexity is used. Though TD-VAE is capable of jumpy predictions and could try to predict multiple steps ahead, the evaluation for TD-VAE is only one step ahead to be consistent with the other models.

The easy story cloze task has strong results of greater than 0.9 for most LSTM, transformer and TD-VAE models in telling the original from the modified story, so this task is not that useful. The language model (LM) performs much worse in identifying the original story, so all the hierarchical models are improved. This is perhaps not surprising with cloze, as sampling from the same LM makes it unlikely this will perform well in detecting its own sampled sentences from the original. However, the comparison with the hierarchical models is tougher: For the hard task, K-1, K-5, and K-10, the TD-VAE model shows a clear improvement over both the transformer and LSTM models the swap and cloze tasks. That TD-VAE model is able to predict better the cloze sentence than either LSTM or transformer models with equivalent depth shows it performs strongly on coherence tasks. It doesn't necessarily mean generated stories with the model will be better but demonstrated that a human evaluation was worthwhile.

\subsection{Human Evaluation}

There were two human evaluations attempted for story generation. The first unsuccessful one was for the automated system to generate a relatively long story of $20$ sentences following a short prompt from the \textit{WritingPrompts} corpus. This is a fairly standard approach adopted by the original \textit{WritingPrompts} paper \citep{fan-etal-2018-hierarchical} and by more recent papers such as \citet{goldfarb-tarrant-etal-2020-content}. The second evaluation which is more successful is a long prompt followed by a short generated sequence. The first more unsuccessful evaluation is recapped first.

Common to both are the five models selected for evaluation:  \textit{Gold}, the actual human-written story from the dataset; \textit{LM}, the fine-tuned GPT-2 medium model: \textit{LSTM Reranking}, which generates multiple sentences using GPT and reranks and selects the most likely candidate using a beam search. LSTM Reranking is used instead of the transformer model because the automated benchmarks are similar, so evaluating both would be superfluous. \textit{TD-VAE Reranking} used beam search like the LSTM Reranking model but based on the TD-VAE. \textit{TD-VAE Conditioning} differs in that it directly conditions on the expected latent vectors. So the TD-VAE expectations change the text generated, unlike the Reranking model, which only filters out less suitable sentences generated by GPT-2. All the models are the \textit{all} datasets versions of the models as these had better performance overall on the automated evaluation task. Both human evaluation exercises are kept to only five models as evaluators will need to be able to read both a prompt, understand them and recall them when answering questions, so it is important to keep the numbers down. Together the models chosen provide a Gold standard, a strong LM baseline, and strong alternative hierarchical models to compliment the TD-VAE models. The advantage of evaluating all models on a common finetuned GPT-2 base is that it provides a more direct comparison of the influence of the upper hierarchical layers, which wouldn't be the case with SOTA models with largely different architectures.

The task for the evaluator is to read a prompt and five stories and then rate answers to various questions. One way of evaluating would be to score each story on an ordinal Likert rating scale-out of $5$ or $10$. Instead, the method for both studies is to force a ranking between $1$ and $5$ with one being the best story and five the worst. A forced ranking avoids situations where the evaluators might evaluate multiple stories similarly, and forces them to make a choice about which is better than a rating scale. \citet{kiritchenko-mohammad-2017-best} finds ranking (or best-worst scaling) to be more effective than absolute scales. For the exercise, $5$ MTurk annotators were asked to evaluate and similar quality control procedures were applied as in Chapter \ref{chap:suspenseannotation}. One-hundred stories in total were evaluated. There are also similar quality control questions. All stories had a randomised order within each task, so it is not possible to guess stories from a position. The stories themselves were generated to $25$ sentences. However, because different models had different average lengths per sentence, they were truncated to as near as possible $1500$ characters while keeping whole sentences. The truncation is so the evaluator judgement isn't affected by the length of the story.
Mechanical Turk workers were asked to rank the five options from best to worst using five criteria: \textit{Overall} (a subjective judgement on their overall impression of the story); \textit{coherence} (internal consistency of characters, places, actions and plot development); \textit{Fluency} (how well written the story is, including its use of expressive language); \textit{Relevance} (how close the story is to what is expressed in the prompt; automatic systems might generate a good story, but unrelated to what was specified in the prompt); \textit{Suspense} (how suspenseful was the story.)

No results are presented from this initial study for two reasons. The first is that the inter-annotator agreement is low, only $0.16$ with Krippendorf's $\alpha$. Stories are inherently subjective, and hence it should not be expected to be high, but it is far lower than expected and necessary to draw valid conclusions. Secondly, all the models are closely clustered on most of the questions between $2.8$ and $3.1$. There are few statistically significant results with a standard Tukey test at $0.05$. The Tukey test is favoured because, unlike a straight pairwise T-test, it controls for multiple comparisons. A natural answer could be that it is a failure of quality control, and some evaluators may be cheating. There may be some of this, but the quality control answers were generally reasonable. From inspection, it is also hard to judge which story is better. The working conclusion was that it was hard for evaluators to judge the stories because, with a short prompt followed by a long story, the story quickly diverges, making the judgment more subjective. The evaluators will also struggle to remember all the details when answering questions at the end, and so there may be a recency effect where, though the order is random, they recall more of the last story. The second evaluation attempted to fix these issues.

The problem with differentiating between model outputs is not unique to the discussed evaluation. In a recent paper, after both the evaluation exercises in this Chapter, \citet{clark-etal-2021-thats} reviewed the strength of Gold standards and the consistency of evaluation judgements. They find that with GPT-2 generated text, untrained MTurk evaluators can only correctly pick the human-generated content $57.9\%$ of the time. With a far large LM GPT-3 it reduces to $49.9\%$, random chance. Even adding well-written instructions, comparisons, or examples only made a small difference. The results are likely a combination of the much better generation of recent large transformer models, inherent subjectivity, and workers on crowdsourcing platforms tendency to complete tasks quickly without necessarily thoroughly understanding what is expected. Recent work by \citet{clark-smith-2021-choose} suggest interactive story generation tasks where users choose between alternative models that generate text dynamically is a better approach to evaluation, mainly because the evaluator is more engaged in the process if they choose how the story develops. Although this approach is not adopted, keeping the focus of the evaluator was part of the revised evaluation.

The approach taken in the improved evaluation is to have a long prompt of $20$ sentences in addition to the prompt for $50$ stories. Further stories were generated but didn't need to be evaluated because both agreement and the significance of results were much improved, so further collection would not have altered the results. The model generates continuations of $10$ sentences for each of these stories that are truncated with whole sentences to a length of $800$ characters to keep the lengths similar. The benefit of the long prompt is one of the most challenging aspects of story generation is maintaining the coherence of the plot. An extended context makes it more evident to the evaluator if the continuation isn't coherent, doesn't fit the story's style, or progresses the plot as expected.  Mechanical Turk evaluators received $1$ dollar per hit. To ensure quality we found it necessary to use the Master Worker option and filter for workers with $> 98$ approval and at least $2000$ successful hits.

\begin{figure}[htbp]
  \centering
  \includegraphics[width=1.0\textwidth]{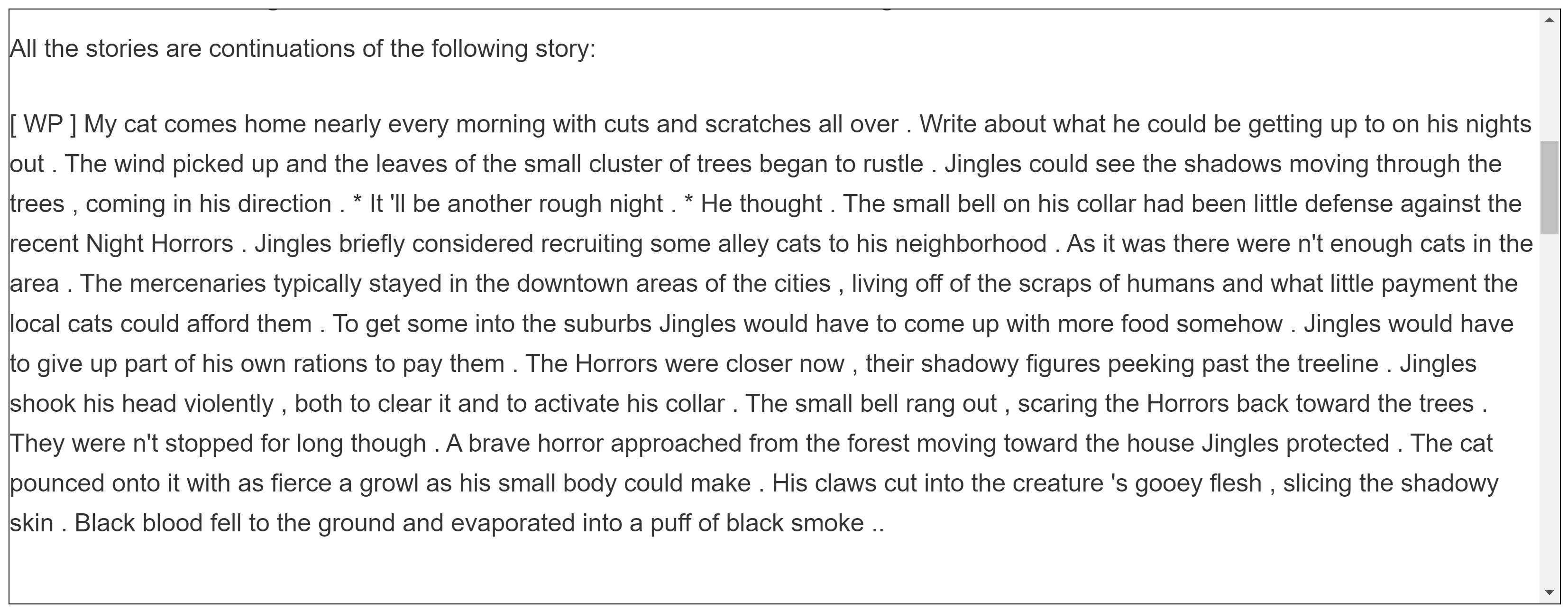}
  \caption{The story prompt show to AWS Turkers.}
  \label{fig:story_cont_story_prompt}
\end{figure}

\begin{figure}[htbp]
  \centering
  \includegraphics[width=1.0\textwidth]{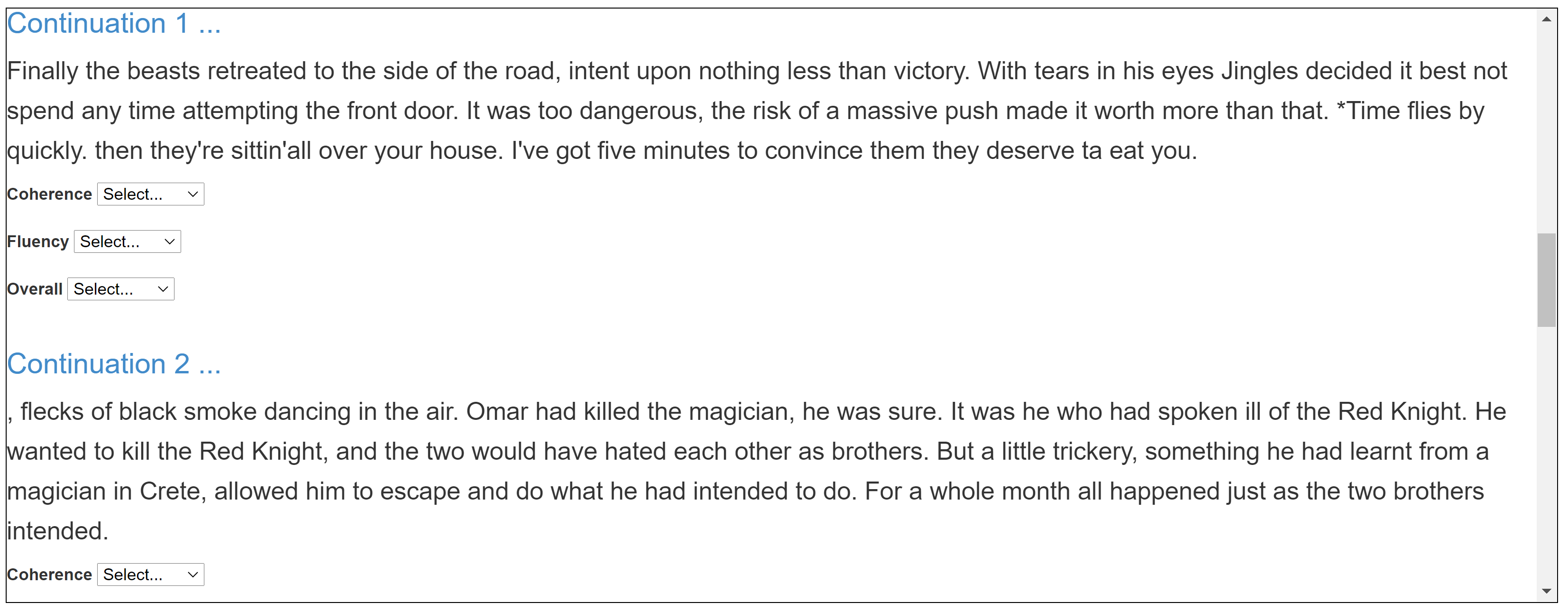}
  \caption{Showing the continuation and dropdown choices.}
  \label{fig:story_cont_select}
\end{figure}

The Mechanical Turk task is illustrated in three figures. Figure \ref{fig:story_cont_story_prompt} shows an example of a long prompt shown to the evaluators. Figure \ref{fig:story_cont_select} shows a continuation with a selection box per continuation. Validation is enforced to ensure the stories can only be ranked $1 - 5$. Figure \ref{fig:story_cont_submit} shows the freeform questions to verify quality and understand why evaluators are making particular rank orderings.

\begin{figure}[htbp]
  \centering
  \includegraphics[width=1.0\textwidth]{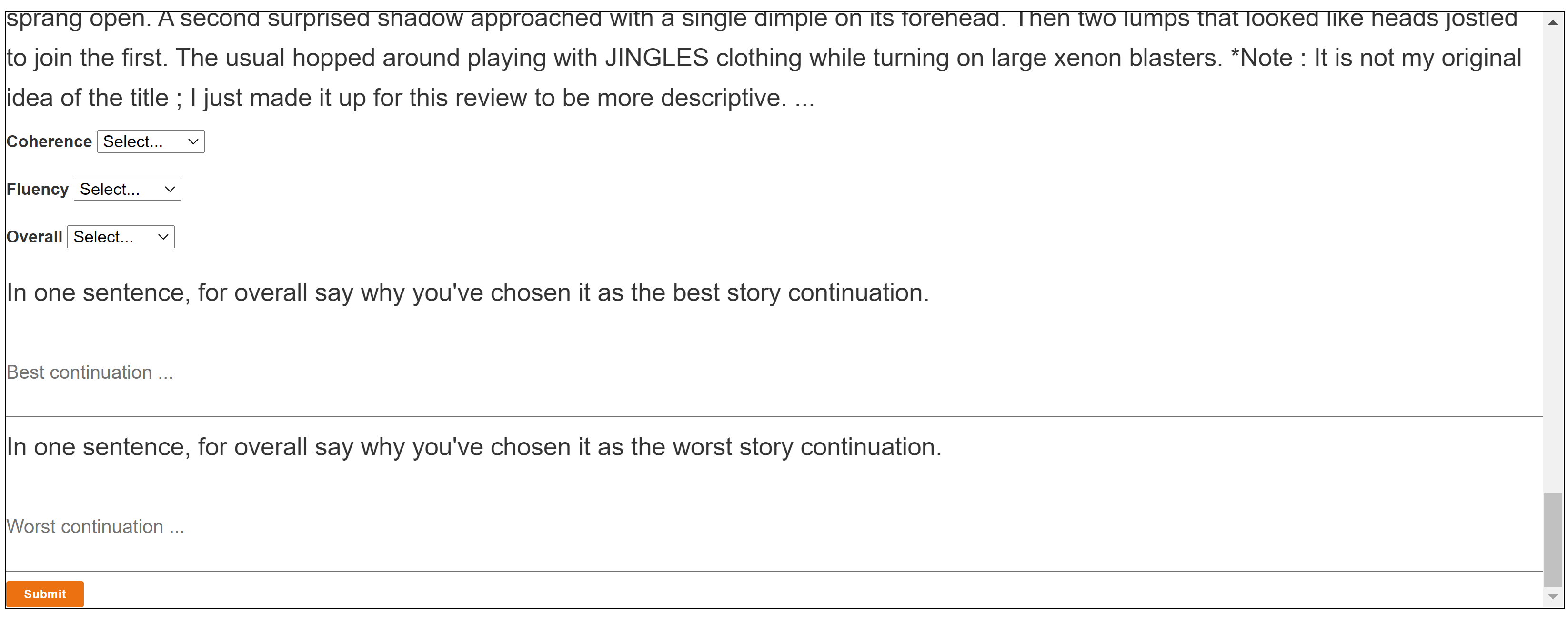}
  \caption{Checker validation question and the submit button.}
  \label{fig:story_cont_submit}
\end{figure}

The following are the three questions that were used for the continuation study. \textit{Suspense} and \textit{Relevance} are less appropriate to shorter continuations of longer stories since the context is more firmly established. So the questions that were asked in the study are the following, with each evaluator asked to rank the stories from best to worst, $1 - 5$:

\begin{itemize}
  \item \textbf{Coherence}: Narrative coherence is the degree to which the continuation makes sense given what has gone before. Coherent stories are internally consistent and involve a plot development and progression with characters, places and actions that are meaningful in the context of the prior story.
  \item \textbf{Fluency}: How well does the continuation follow on from the story, is it natural, does it follow in terms of style, grammar, use of literary devices, and expressive language.
  \item \textbf{Overall}: This is an overall judgement of how good and interesting you think the continuations are in terms of criteria such as character, plot and how well written they are.
\end{itemize}

Additionally, MTurk workers were asked to summarise the story and give reasons for their overall choice. This served to filter poor quality submissions and to give insight into the factors they considered important. Five annotators judged each story after screening out and re-collecting for low-quality annotations. Overall inter-annotator agreement for the task is $0.35$ using Krippendorff's alpha. The agreement is still relatively low but a marked improvement on the first study. It is also demonstrated that although there is a clearer ranking with the second study, there is still considerable disagreement about which systems are the best. It is perhaps because the difference between the models is comparatively small.

\textit{The following are two random examples of evaluated stories and the continuations generated by each model. These examples are presented in a fixed order, but in the real task the different model continuations were shuffled. All stories as per the task are generated with a single run and evaluated and example stories are not cherry picked.}

\paragraph*{Example One}

\paragraph*{Prompt:} \textit{[ WP ] A woman investigates her husband 's murder ... while receiving insufferable vague help from his ghost .} `` Would you like a blanket Mrs . Everson ? '' the police officer asked in a calm and collected voice . `` No , im fine thank you '' she responded , holding back her tears . The police officer nodded and made his way over to the group of officers surrounding the body on the floor . Mrs. Everson sat , quietly waiting for the next officer to ask her a question , `` leave '' a voice appeared in her ear as if emanating from the wall behind her . A shiver ran down her spine and she quickly turned to see the blank wall , as she looked back a tall man stood before her , `` Mrs. Eve Everson ? '' the man requested . `` Yes '' she responded looking up through watery eyes . The man ran through the standards , he introduced himself as the local DI , and began to present her with a series of questions . `` Did your husband have any enemies Mrs . Everson ? '' . `` No '' She responded quickly , `` yes '' came the voice from the wall , she quickly turned again to see nothing once again . `` Are you alright Mrs . Everson ? '' asked the DI `` ah , yes , sorry it was just ... nothing '' she responded nervously . `` Did you notice anything suspicious leading up to the incident ? '' The DI asked leaping right back into questioning mode , `` ah no , nothing I can thi.. ''.

\paragraph*{Gold:} she began before the voice cut her off `` Think Eve , THINK ! '' it screamed this time , and with the added volume the voice became recognizable . Eve leapt from the chair and fell back wards onto the ground , she scrambled away from the wall , staring at it intensely , screaming out to it `` DAVE ! DAVE ! '' . ...

\paragraph*{LM:} -'ve noticed at all. Must have been some kind of mistake on your part... or is this just another case of mistaken identity. '' Now that he was back in his own voice, the old wizard resumed his conversation : `` The wisest of us have fallen to thinking this over. Why have not the Wise Ones declared this thing a danger? One only knows that we are unfortunate, for it does not take much skill to hunt down the Bad One. ' ` We may at least care for our own lives, as the ogresses do,'thought they. So they agreed to go with the Bad One, and the two started on their way.

\paragraph*{LSTM:} - That's just normal stuff all the time these days. You don't talk too much after dark, what are you doing out here?! '' he said, heading to the door. Walking further into the old room the Man familiarly began to open her hair, dress and shoes, several tresses flying madly about her. He saw through the cloud, after all the time they had spent together the flame consumed him, but in all those years before, this woman still kept talking. She began to rub her hands together, feeling her hair dancing, her dresses floating, yet he remained so entranced. At last he ran them off as fast as thought, sure to leave no trail. ...

\paragraph*{TD-VAE R:} and rubbing his chin. The young woman began forcing past the police officers staring at her with wide eyed bewilderment and longing eyes, almost nauseous. She silently wandered over to where she went down the opposite corner next to the fence. This time however she stopped short of finding where Mrs. Everson had been hidden, very well probably upstairs though she might not be dead. She raised an eyebrow asking the man if the officer was lady like that, the young woman did not know, but he looked at her derisively for not actually knowing.

\paragraph*{TD-VAE C:} he replied angrily. Mrs. Everson could not hide her worry, trying to console herself that no one was paying any attention to her at all. ; Once upon a time, a magician walked down an avenue in our town. ; He asked a young woman what happened to her. ; At one point, he told the woman, you had knocked her head down.

\paragraph*{Example Two}

\paragraph*{Prompt:} \textit{[ WP ] My cat comes home nearly every morning with cuts and scratches all over . Write about what he could be getting up to on his nights out .} The wind picked up and the leaves of the small cluster of trees began to rustle . Jingles could see the shadows moving through the trees , coming in his direction . * It 'll be another rough night . * He thought . The small bell on his collar had been little defense against the recent Night Horrors . Jingles briefly considered recruiting some alley cats to his neighborhood . As it was there were n't enough cats in the area . The mercenaries typically stayed in the downtown areas of the cities , living off of the scraps of humans and what little payment the local cats could afford them . To get some into the suburbs Jingles would have to come up with more food somehow . Jingles would have to give up part of his own rations to pay them . The Horrors were closer now , their shadowy figures peeking past the treeline . Jingles shook his head violently , both to clear it and to activate his collar . The small bell rang out , scaring the Horrors back toward the trees . They were n't stopped for long though . A brave horror approached from the forest moving toward the house Jingles protected . The cat pounced onto it with as fierce a growl as his small body could make . His claws cut into the creature 's gooey flesh , slicing the shadowy skin . Black blood fell to the ground and evaporated into a puff of black smoke ..

\paragraph*{Gold:} More horrors came out , rushing for Jingles . Every young kitten knew about the Night Horrors , and as soon as they were old enough to fight they began training . Most kittens were lucky enough to have a sparing partner in their parent or their siblings . Jingles came from a large litter and he was no exception . He was ready for the horrors charging him .

\paragraph*{LM:} , flecks of black smoke dancing in the air. Omar had killed the magician, he was sure. It was he who had spoken ill of the Red Knight. He wanted to kill the Red Knight, and the two would have hated each other as brothers. But a little trickery, something he had learnt from a magician in Crete, allowed him to escape and do what he had intended to do. For a whole month all happened just as the two brothers intended.

\paragraph*{LSTM:} Jingle bells rang in his ears as he jumped away, brushing up against the blackened remains of his coat and fur Blurry lights blinded him once more and he cursed softly into his night ears, All because he feared The Grudge, the curse long gnawed on his bones They lured him back with stories about how *the streets were no place for bad men to roam wild and cruel man to terrorize a terrified girl. Little did he know they only made him afraid, and fed the paranoia with sad tales of candy shops filled with children and toddlers. He had grown so accustomed to life being the same as a petrified child, that eventually, the grudges abandoned him forever.

\paragraph*{TD-VAE R:} Disgusted with himself he began to throw open the door to the house. The household alarm woke him first when his private room window sprang open. A second surprised shadow approached with a single dimple on its forehead. Then two lumps that looked like heads jostled to join the first. The usual hopped around playing with JINGLES clothing while turning on large xenon blasters. *Note : It is not my original idea of the title ; I just made it up for this review to be more descriptive. ...

\paragraph*{TD-VAE C:} Finally the beasts retreated to the side of the road, intent upon nothing less than victory. With tears in his eyes Jingles decided it best not spend any time attempting the front door. It was too dangerous, the risk of a massive push made it worth more than that. *Time flies by quickly. then they're sittin'all over your house. I've got five minutes to convince them they deserve ta eat you.

\paragraph*{END EXAMPLES}

First, before analysing the models' results, it's worth delving into the examples themselves. Both examples are randomly selected. None of the automated stories are good continuations of the prompt. In \textit{Example One} about the \textit{Ghost} helping the murder investigation: In the \textit{LM} example, a \textit{Wizard} suddenly appears. Whereas the \textit{LSTM} generation talks about \textit{falling through a cloud} and the \textit{flame consuming him}. In \textit{Example Two} about \textit{Jingles} the cat, the \textit{LM} model starts talking about a \textit{Red Knight} and a \textit{Magician} that is utterly irrelevant to the plot, probably because of a reference to mercenaries and shadowy figures in the prompt. \textit{TD-VAE R} model mentions xenon blasters suggesting a sci-fi. There are clear discontinuities to the plot whether sudden new characters appear and events don't make sense given the prompt. On the other hand, some of it makes sense and flows. Stylistically the text fits.\footnote{Some of the examples contain meta commentary where the author is talking about the story. It is because \textit{WritingPrompts} is a creative writing forum and so some authors will include commentary as part of their story which the model is learning during training.} Some real characters are reoccurring characters, such as \textit{Jingles}, and \textit{Mrs Everson}, are mentioned correctly. 

The examples demonstrate the weaknesses of the underlying model in understanding real-world relations between the characters and their actions which was touched on the grounding problem and discussion of common sense events in Chapter \ref{chap:backgroundml}. The models can hallucinate characters and events that may be related topically (a wizard is connected as a ghost is to the supernatural) but are not causally related to what has already happened in the story. The problem is often masked by hand-picked examples in the literature. For example, in the GPT-2 paper \citep{radford2019language}, the \textit{Unicorn} story example attracted a much community interest. It was cherry-picked out of multiple samples and hides the underlying problem of the model. It was noted earlier that the initial strategy was to use only the prompts, which are in italic in the examples, followed by a long generated text. The problem with this is that the short prompts are fairly open to interpretation. It means the generated stories can be very different, making it hard for evaluators. It also means that the model can easily go off in its own direction without concern for consistency or coherence. After all, maybe the cat does live with a \textit{Magician} on \textit{Crete}, and it's a historical work, and there is a \textit{Red Knight} in the story. The \textit{ghost} investigation example could have fantasy dream sequences. With a short prompt, all of these can seem reasonable. It is a possible reason why the models in the short prompt evaluations were ranked nearly identically. The longer prompt is far more challenging as it's obvious to the reader if the continuation doesn't flow. Most of the papers that evaluate generation from \textit{WritingPrompts} corpus or other similar story generation tasks seem to use short prompts, which is a possible problem. It would suggest in further work that conditioning on far longer texts is a more demanding and preferable evaluation strategy.

On the problems themselves, the \textit{TD-VAE C} with the planning approach seems less susceptible to hallucinating or nonsensical continuations. However, as shall be seen with the results being more conservative or leads to less diverse and more boring stories. The planning approaches reviewed earlier in the chapter are also less susceptible; if the model is substituting \textit{ent6} with the real character at the end, then naturally, it won't invent implausible new ones. Also, although only used for analysis the generated examples from the knowledge and episodic memory models in Chapter \ref{chap:salience}, while imperfect, also appear better; the memory and external knowledge could ground the model better. Overall, the problems reflect the weaknesses of current LMs. The conclusion in Chapter \ref{chap:conclusion} proposes some further ideas around character and persona modelling and grounding with multimodality that, while discussed in the context of improving suspense and salience inference, should also improve story generation.

\begin{figure}[t]
  \centering
  \includegraphics[width=1.0\textwidth]{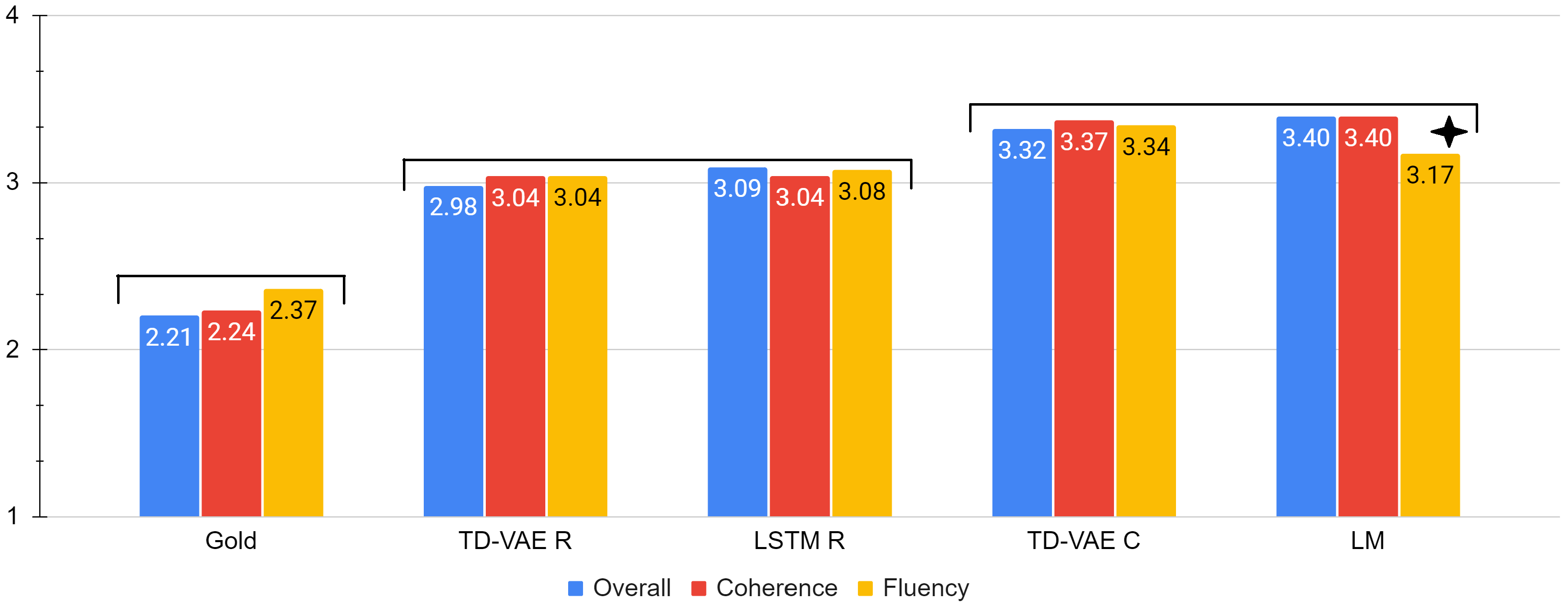}
  \caption{Continuation story generation results. Bar markers split into groups at $p < 0.05$ significance using a pairwise t-test. The exception is LM fluency marked with the star, which is statistically the same as LSTM R and TD-VAE R.}
  \label{fig:cont_bar}
\end{figure}

Figure~\ref{fig:cont_bar} shows how the models were ranked by annotators on a scale of 1 to~5, i.e., lower is better. There isn't much difference between the different questions (overall, coherence, and fluency), except that fluency is relatively better for the base GPT-2 language model than overall and coherence. The Gold standard stories are better than all model-generated continuations. The two reranking models LSTM and TD-VAE R, are significantly better than the base LM model, which means reranking improves the story over straight sampling. The TD-VAE C model that uses the planning mechanism doesn't improve the base LM, and fluency is worse.

\begin{table}[t]
\centering
\begin{tabular}{@{}lcccccc@{}}
\toprule
\textbf{Model} & \multicolumn{1}{l}{\textbf{Nouns}} & \multicolumn{1}{l}{\textbf{Verbs}} & \multicolumn{1}{l}{\textbf{Coreferences}} & \multicolumn{1}{l}{\textbf{Coreference Chains}} &
\multicolumn{1}{l}{\textbf{Meteor}} &
\multicolumn{1}{l}{\textbf{BLEU}}\\ \midrule
Gold           & \textbf{4.08}                      & \textbf{2.04}                      & \textbf{2.23}                      & 4.17       & - & -            \\
LM             & 2.48                               & 1.29                               & 2.06                               & 4.69          & .101                                & .628                                      \\
LSTM           & 2.59                               & 1.21                               & 2.18                               & \textbf{5.02}    & \textbf{.107}                                & .619                           \\
TD-VAE R       & 3.19                               & 1.41                               & 1.61                               & 4.00       & .090                                & \textbf{.705}                             \\
TD-VAE C       & 2.07                               & .885                               & 1.65                               & 3.99   & .083                                & .639                              \\ \bottomrule
\end{tabular}
\caption{Statistics with the average number of unique nouns and verbs (after stemming), coreferences per 100 tokens, and length of coreference mention chains per story. Meteor and BLEU are measured against the Gold stories. These stats are on a more extensive set of 400 stories and text of up to 2000 characters per generation.}
\label{tab:continuation_story_diversity}
\end{table}

The main question is why TD-VAE Conditioning can perform well on automated evaluation (cloze and swap) but not improve story generation compared to the LSTM-based reranking model. Inspection of the stories used for evaluation suggests that the conditioning model is more repetitive than other models and uses more repeated linguistic constructs in continuations. One approach from \citet{roemmele_evaluating_2017} is to run linguistic diagnostics to measure the qualities of the generated text. Analysis of the linguistic diversity of the generated continuations by counting unique nouns and verbs, coreference chains, and similarity measures with the Gold standard are on Table~\ref{tab:continuation_story_diversity}. The TD-VAE C model has far less diversity than any other models and differs markedly from the human-written stories. For both TD-VAE models, unique coreferences are lower than other models. In contrast, the coreference chain length is similar, implying that the TD-VAE models are more repetitive when generating coreferences and are less likely to introduce new entities, again reducing text diversity. On the other hand, TD-VAE R outperforms both the LM and LSTM model in terms of noun and verb diversity. 

For completeness, the Meteor and BLEU scores are also in Table~\ref{tab:continuation_story_diversity}. The TD-VAE models outperform the LM baseline and the LSTM-based hierarchical model in terms of BLEU, while the patterns is reversed for Meteor. As noted earlier, it's best to take measures with a pinch of salt since these automatic measures tend to correlate poorly with human judgement.

If a planning mechanism works, we would expect it to increase lexical diversity, as the planning would condition the LM to explore new areas and develop the plot. In contrast, it would appear the opposite is happening here. It has been noted by \citet{DBLP:conf/iclr/WelleckKRDCW20} amongst others that large language models can produce dull and repetitive text and often copy from the context. The failure of the conditioning model could be an analogue, where TD-VAE can train well by expecting the future text to be similar in topic and style to the text seen. This would lead conditioning in the latent vectors to produce more dull and repetitive text than the LM on its own, which appears to be the case for TD-VAE~C. There seems to be a clear split between TD-VAE used in a ranking model and when conditioned on. 

The strong automated performance and improvement on the base LM with human evaluation show that TD-VAE~R can determine coherent continuations and improve story generation. However, when conditioned on this, it produces text more similar to the existing text, not interesting and eventful plots. Part of this could come from the richness of the latent representations. In a keyword planning system, the plot skeleton conditioned on is quite loose, for example from \citet{wang-etal-2020-plan}: \textit{lake} $\rightarrow$ \textit{friends} $\rightarrow$ \textit{water} $\rightarrow$ \textit{swim} $\rightarrow$ \textit{shore}. It allows many possible realisations and is not that restrictive of the LM. In contrast, sentence embeddings represent many more linguistic and semantic properties, so learning to condition on them can cause generated text to be repetitive and limited.

If the model had been clearly better than the other baselines, then it would have been worthwhile comparing the model to more powerful planning systems such as \citet{goldfarb-tarrant-etal-2020-content}; or systems that try and combine commonsense reasoning in planning such as CAST \citep{DBLP:journals/corr/abs-2105-01311} or C2PO \citep{Ammanabrolu_Cheung_Broniec_Riedl_2021}; or approaches that combine neural methods with more traditional symbolic approaches such as \citet{martin2021thesis}. Given the mixed results, it was deemed not beneficial to do so. 

\section{Conclusion}

The current and previous chapters have tried to adapt a temporal VAE model operating in the latent space for comprehension tasks on inferring suspense and surprise and in story generation. The rationale for both models is the same: The Ely definition of suspense and surprise requires a model to predict a realistic distribution over future outcomes. The TD-VAE can do this in the latent space and is computationally much simpler than generating concrete alternatives for future continuations. Generating plausible continuation in a simplified latent space is what is required of a story planning system. This chapter adapted the model as a planning system via conditional generation and reranking.

In both comprehension and generation tasks, the models produced mixed results. For suspense and surprise modelling, both were better than simpler baselines. Surprise was comparable in performance to the hierarchical rollout model but was far worse with suspense. In the secondary turning points task, the model was similar in performance to hierarchical rollout. For the generation tasks, the model does well on technical measures such as cloze, and applying TD-VAE for reranking can help generation. However, conditioning on the latent vector degrades the performance of story generation. This chapter concludes by reviewing the reasons for the mixed performance and suggesting a direction for future work.

The final analysis in the story generation hinted at what the problem may be in that the diversity of the text generation is reduced when conditioning on the latent vectors. From reading the generated examples, it also seems that particularly the TD-VAE conditioning model seems to stick more to the existing situation. If the prompt ends with dialogue, then the dialogue carries on; if there's a description of the scene, the description carries on. While the content is generally coherent, there is less plot progression. Two possible related reasons: The first is that the latent vector represents many different linguistic and semantic properties which was the intended strength. The representations are less focused on the salient features required for plot progression. The second is that TD-VAE struggles to learn progressions or progressions for other latent vector features that do not advance the plot. Both reasons would account for the observed results. The model is weak on suspense which requires it to generate variations itself but strong on surprise and turning points which are the difference between expectations and what happens, and so sudden changes will be noticed. The model is also strong on the automated cloze and swap tasks which requires predicting the likelihood of the next sentence. In narratives, typically, the situation develops slowly and as expected, but it is more dramatic and sudden events that make the story interesting. A model can do well on cloze type automated tasks by learning normal progressions without the more dramatic shifts. Conditioning on the model would thus slow down the plot development. That also could be why reranking sentences generated by GPT-2 could help. The generated sentences are more diverse and go in a different direction, and then selecting the most coherent as a filter will improve the story versus just random sampling.

One option for solving the issues would be to move away from raw sentence inputs to more structured approaches. \citet{Martin2018EventRF} adopt an \textit{event2event} model where events extracted from sentences are the input to an event encoder that is similar in structure to the sentence encoder. The event can mean the main verb or include the subject or object. The event could also include more structured information such as the SRL planning systems or extracted coreferences. As per the original rationale for multi-step planning systems, the idea is that encoding only the most salient plot information will focus the model and help with higher-level plot structures. While it may improve the performance of the model, such approaches go against the original ethos of the approach, which is the model should implicitly learn the most relevant representations. 

An alternative future direction is to try and change the sentence encoding loss to prioritise the more salient plot information. One hybrid approach between extracting preprocessed events and a purely unsupervised latent vector is to incorporate more structure into the sentence embedding. In the current version, the loss inspired by Quick-Thoughts simply makes neighbouring sentence vectors more similar than other sentences in the batch. The loss can pick up on any cues that distinguish adjacent sentences from others, not necessarily those most pertinent to story plots. Some other losses tried to encode more commonsense information from NLI and ATOMIC, a commonsense reasoning dataset. A hybrid approach might be that the essential information in a sentence for a story plot is the main verb and the coreference. Predicting these directly via negative sampling or more direct in-filling could be alternative, or supplementary losses to make the sentence representations more plot-focused. For tracking entities and coreferences, there is plenty of prior work, including, for example, \citet{gupta-durrett-2019-effective} tried to explicitly incorporate learning these qualities into neural models.  Of course, there could be many such variants on the approach.  The main point would be the sentence is still represented in a single latent vector that would take raw text as input.  There would be no complicated preprocessing pipelines required for inference.  Instead, the latent vectors would be trained to pick up the most relevant details.

In Chapter \ref{chap:rolloutsuspense}, one suggestion for learning an $\alpha$ importance score weighting was to rely on an existing or new corpus containing aligned summaries. The approach is similar to that adopted for the Shmoop aligned summary evaluation of salience in Chapter \ref{chap:salience}. The intuition would be that if an event occurs in the summary or an extended synopsis, it is salient and hence more crucial to the plot. Instead of just learning an $\alpha$ value that would be applied posthoc, the alternative would be to build it into a model. For example, if a sentence closely matched a sentence from an aligned synopsis, possibly with a similar method to the following chapter, it can be flagged as salient. A sentence vector can then be trained to predict whether or not sentences are salient. The loss would be similar to the NLI loss in this chapter. This can obviously infer directly as a task which parts of stories are salient. With sentence embeddings, the loss would also train salient sentences to be more similar and non-salient sentences more dissimilar. Hence, the vector representations would be differentiated.

An alternative route may improve both sentence representations directly and the predictive power of the upper TD-VAE layer that is discrete representations. The current latent sentence vectors represent all the semantic and linguistic properties in a continuous space. One of the arguments favouring simplified planning systems, either keywords, events, or SRL, focuses on the central role of the sentence or more extended passage in the plot. The sentence vectors in the TD-VAE model represent a lot of information in a continuous compressed space. It is similar to image data, where complex information about different objects, colours, textures, etc. have usually been compressed via CNNs into simplified continuous representations of the feature space. Transformers were not really an option for image data since they needed discrete input tokens. Recently, however, there has been an enormous success of visual transformers on images \citep{DBLP:conf/iclr/DosovitskiyB0WZ21}. Not only have such models often outperformed CNN models but they have also been shown to improve the robustness \citep{DBLP:journals/corr/abs-2111-10493} on vision models. The method is conceptually pretty simple; a clustering method such as k-means can convert the continuous latent space into a discrete set of representations which can then be modelled as numeric code and looked up in a dictionary when needed. The idea behind improving the model by discretising sentence representations into $n$ codes is that it should focus the representations into functional roles. A simplified functional role should then be easier to infer into the future by the top level of the model and thus aid predictive power. For the same reason it could also aid the conditional generation since effectively it would then be learning to condition on a discrete number of control codes. Topic-VAE \citep{wang-etal-2019-topic} does this with a mixture of discrete topic codes.

Within the VAE space there has been a similar development with VQ-VAE \citep{NEURIPS2019_5f8e2fa1,NIPS2017_7a98af17}. VQ-VAEs have also found success in image domains. Conceptually the change between these are similar to the visual transformers in that the latent space uses clustering so that the latent space is represented by a discrete code. The intuition is that the representations are still in a continuous space. The clustering produces more stereotypical roles that are more cohesive and so simpler for a standard VAE to reconstruct the input as it relies on a more tightly defined latent space for a particular class of object. In the vision domain they have been found to be less sensitive to mode collapse and can better capture diversity. The second issue seems to be one of the main issues with the existing vanilla VAE architectural components. Replacing TD-VAEs standard VAEs blocks with VQ-VAE blocks would convert the temporal inference to effectively mapping between discrete codes or representations in the latent space. It is conceptually more similar to the simplified event systems as a planning system for modelling suspense or planning for story generation. Suppose the number of representations are limited by necessity. In that case, it will focus the representations of the primary role of the sentence while removing irrelevant details, which may improve the models performance in predicting plot development.

Other methods could be relevant for improving TD-VAE predictions. Earlier generative alternatives to VAEs such as GANs and invertible flows were reviewed, and both have potential. Invertible flows allow more flexibility in the probability distributions that can be modelled (i.e. non Gaussian), though usually at the cost of computational complexity. Likewise, adversarial methods could improve on VAEs reconstruction losses. The most recent and state of the art development for temporal prediction VAEs is Clockwork-VAE \citep{DBLP:journals/corr/abs-2102-09532}, which was proposed after the TD-VAE experimental work was completed. The Clockwork-VAE is inspired by earlier work on clockwork RNNs \citep{DBLP:conf/icml/KoutnikGGS14}. The concept behind clockwork networks is that each layer has a different cadence to improve performance on longer sequences. So a Clockwork-VAE that predicted future image frames in a video and with four layers might have cadences per layer going from the bottom of $\frac{1}{125}$,$\frac{1}{25}$, $\frac{1}{5}$,$\frac{1}{1}$. So the top layer sees every frame, and the bottom layer only every $125$ frames. The aim is by stacking layers of different cadences, the longer cadences will learn to predict more coarse-grained changes over long periods and upper layers fine-grained details. The result should lead to better overall predictions over long sequences. In principle, it could be applied to stories, but there is a problem. The flow in stories is not linear. There are asides, digressions, and long descriptions away from the main flow. This means that simply picking every $n$ sentence isn't likely to work since it will often pick out less relevant plots parts. Earlier, it was suggested that aligned summaries might improve sentence embeddings to take into account salience. Chapter \ref{chap:salience} uses such an approach for evaluation of salience. A similar approach could work with an equivalent of an aligned source approach with longer works of fiction. For example, suppose there were short summaries for whole novels or screenplays and then aligned summaries per act or chapter and full text. In that case, there could be a three-layer clockwork model where each layer only receives events sparsely at different levels of abstraction. There are publishers such as \href{https://www.shmoop.com/}{Shmoop}, \href{https://www.sparknotes.com/}{Sparknotes}, \href{https://www.cliffsnotes.com/}{Cliff Notes}, and fan maintained sources that would be suitable, but clearly, it requires more preparation than just a corpus of full-text stories.

Chapter \ref{chap:salience} extends an LM with an external knowledgebase and episodic memory. The model can generate text, although it is only evaluated for comprehension tasks. The knowledgebase and memory address limitations both with the TD-VAE model and the hierarchical rollout model from Chapter \ref{chap:rolloutsuspense}. Although both models have many parameters, they are forced to encode all their knowledge into their parameters; an external KB provides a more scalable and updatable alternative. Typically transformers have a context length of a few hundreds or in the low thousand wordpiece tokens that can be attended to in the context text. The advantage of a hierarchical model is that it extends the context as the input to the top level of the model is sentence embeddings. The context can be a hundred sentences or so. Nevertheless, the longest books are far longer than this, so an episodic memory provides a more efficient method of conditioning on the most salient information. Though not done in the thesis, both the knowledgebase and episodic memory are orthogonal to the models thus far and thus could be added to improve the existing models. 
\chapter{Knowledge and Salience}

\label{chap:salience}

\section{Introduction}

So far, the existing work has tried to model suspense and surprise with hierarchical supervised models, and the last chapter adopted one of the methods as a text generation system. There remain some limitations with the current work. The methods won't scale to books or whole screenplays, which are far longer than the \textit{WritingPrompts} stories. The average \textit{WritingPrompts} story has $53$ sentences and around $800$ words on average. In contrast, classic books such as \textit{Charlie and the Chocolate Factory} have around $30$k words, \textit{Great Expectations} around $183$K, and extremely long novels such as \textit{A Suitable Boy} can be over half a million words. Even with screenplays, \textit{Gone Girl} has $24$K words, and even for shorter classic plays such as the \textit{Importance of Being Ernest} has circa $14$K words. There are several reasons why the methods so far in the thesis are inadequate for works of this length. The first is just computationally. The hierarchical rollout is slow in generating alternative continuations, and generating a tree of concrete alternatives for thousands of sentences is infeasible. Even if the TD-VAE model outperformed hierarchical rollout; while it is much faster, sampling $100$ variational states per sentence is still a relatively slow process. A far bigger problem, though, is typically transformers are limited to a context of around $512$ or $1024$ word pieces which translates roughly to just under half the number of words because of bytepiece encoding. That means when inferring suspense on the typical book of over $100$K words, most of the book will not be in context and not conditioned on; the model will not be able to remember most of the story. Another more narratology-related issue is that short stories of $50$ sentences are brief and usually to the point, as are summaries of longer works. Longer form stories typically have more subplots or are filled with details that add colour to the story but may not be necessary for the plot. Any model that wants to infer over a longer work will need to track the most pertinent details from any point earlier in the work. To go back to the \textit{Great Expectations} example, \textit{Abel Magwitch} disappears for some $27$ chapters. Or in \textit{Gone Girl}, there are seemingly innocuous details the viewer is asked to recall later; these details are a central part of the mystery and tell the viewer something is amiss, and the situation is not as it seems. Even in a more straightforward plot like \textit{Rocky} the viewer needs to refer back much earlier in the story to relevant events, such as the first encounter with the champion \textit{Apollo Creed}, or the financial issues \textit{Rocky} has with his manager. This chapter aims to explore approaches that are scalable to much longer works while tackling these issues.

There is a second issue that the chapters on the TD-VAE model tried to tackle but not wholly successfully. The TD-VAE sentence vectors were trained on  ATOMIC, commonsense reasoning datasets, and NLI datasets. The idea behind both is that many real-world interactions are absent of narrative texts, but a system would need to know them for causal reasoning. One issue with the approach taken is that the knowledge must be directly encoded into the model's weights. Other commonsense knowledge could be encoded by adding more datasets, but it is not scalable. The size of the model must increase. Instead, this chapter explores alternative external knowledgebase systems that have had much recent success in question answering and factual domains.

The chapter puts these ideas together, combining an external knowledgebase that also acts as a memory mechanism. The knowledgebase allows the model to access external specialised narrative plot knowledge or general knowledge. The memory component allows the model to retrieve relevant plot elements from earlier in the story. As per the rest of this thesis, the model adopts a reader approach where the model reads sentences incrementally as a reader, or a viewer does, rather than analysing the story as a whole unit. The rationale is because lots of the expectations of storytelling is the gradual revealing of information and much of the drama is created as a result of this. Posthoc analysis where the model can look at the context in both directions is complimentary but quite different. Instead of surprise and suspense, the chapter tries to infer salience which is conceptually related but more general.

\citet{forster1985aspects} compared a story to a \textit{wriggling worm of time} that can be seen as a series of events arranged in order (see also \citealt{abbott2008cambridge}) --- dinner comes after breakfast,  night after day, nemesis follows hubris. Not all events are of equal importance, and some are far more \textit{salient} than others. For example, the beginning of Dickens' \textit{Great Expectations} ---
\textit{Keep still, you little devil, or I'll cut your throat!} --- is more \textit{salient} to the story than events such as \textit{my sister had a trenchant way of cutting our bread and butter for us}. \textit{Salient} events in storytelling are those that drive the plot forward or change the state in the story world, as opposed to descriptive details or non-consequential activities. As such, detecting \textit{salience} is an essential part of understanding narrative. Salient events are the core of plots and can aid storyline writing and story generation; they represent essential information and are relevant to question answering.

Chapter \ref{chap:introduction} discussed the scope of the stories to which the work in the thesis can be applied. All the chapters on suspense were limited by technical constraints with the models, which restricted them to shorter work. This chapter evaluates \textit{salience} on much longer works and so overcomes the constraint as the technical methods are computationally simpler. There is also a separate evaluation on much shorter Propp fairytales of a similar length to \textit{WritingPrompts} stories, so these methods also apply to shorter works. The evaluation later is also on classical works of fiction. So they are less likely to have the eccentricities and variability of the quality of \textit{WritingPrompts}'s stories. The other consideration is that short stories, by necessity, need to develop the plot quickly. In contrast, longer ones have sub-plots, more background, and scenes to fill in the colour and the detail. The models in the chapter are incrementally reading the story and making inferences for \textit{salience} over the windows of an LM that are limited to a couple of paragraphs of length. This limits the scope to be a more limited local scope within the context of a longer story. Events may be salient within a local scene but not important overall in the story. So the model is not inferring global salience across the whole story, though there is overlap. This idea is revisited later in the chapter when discussing the definition of salience, the technical introduction to the model, and the discussion of results.

Also, as per the introductory discussion on scope, what is salient heavily relies on the plot structure. The plot structure is particular to a genre, culture, and other social factors. So like the other work in the thesis, the predictive power of the models is determined by the training data. As per the TD-VAE model trained in Chapter \ref{chap:tdvaesuspense} the model is trained on a range of datasets to improve generalisability. The details of which are introduced later in the chapter. However, as the chapter relies on a retrieval component that accesses external knowledgebase, these contents are also relevant. The knowledgebases are the whole of Wikipedia and plots extracted from Wikipedia. Both sources contain a range of references to plots from worldwide literature and films, and Wikipedia includes a range of diverse information across cultures. Nevertheless, there is a slant towards the West and with plots of popular and classic works from Western literature. This isn't a problem for the evaluation as it is primarily on classic works of Western literature, it would impact transfer to other domains if applied. One advantage, though, of relying on an external knowledgebase is it may be possible to more easily adapt to new domains by changing the knowledgebase to be more relevant or finetuning the encoders that perform the retrieval without necessarily having to finetune the whole model on the new domain.

The model builds on the Barthes Cardinal Functions (BCF) definition of salience introduced in Chapter \ref{chap:backgroundtheory}. The computational work follows from \citet{otake-etal-2020-modeling}, who used Barthes Cardinal Functions (BCF) for unsupervised \textit{salience} detection. The chapter augments this approach with a knowledgebase (KB) and memory, and applies the method to book-length stories. In contrast, Otake et al. only evaluated the model on a small corpus of short fairytales. To recap,  Barthes Cardinal Functions \citep{Barthes1966AnIT} are hinge events that cannot be deleted without altering the story; they are the decision points between alternative consequential paths. Barthes and Duisit also define \textit{catalysers}, which are inconsequential events such as the bread and butter example, \textit{indices}, which are descriptive, referring to a character or situation, and \textit{informants}, which identify time and space. These latter types can be seen as \textit{satellites} around the \textit{nuclei}, or filling in gaps between cardinal functions. Hence to identify BCF is to identify the main skeleton of the plot. The model treats the BCF events as the salient events in a story. This scheme relates in narratology with \citet{chatman1980story}'s \textit{kernels} and \textit{satellites} model, as well as with discourse theory in RST \citep{Mann1988RhetoricalST}, which similarly has \textit{nuclei} and \textit{satellites}. The key to the Otake et al. method is that it can be implemented using any LM (Language Model) on any text and does not require a large corpus of annotated training data. In this sense, it follows on from all the earlier work in the thesis that reuses domain knowledge implicit in unsupervised neural network models with theory from narratology.

The BCF defines a cardinal function as one that can't be removed without impacting comprehensibility. They are the key events, not those that fill in the gaps between them. \textit{Salience} can be thought of in two ways. The first is looking retrospectively at the story as a whole. In this scenario, it is known what happens at the end. It is possible to tell which events were important retrospectively. These might be the events described in a plot summary or extended synopses. However, recall the theory of suspense discussed in Chapter \ref{chap:backgroundtheory} and of the reader as a problem solver. The writer tries to create a sense of suspense and mystery by playing with the reader's expectations. This generally occurs in stories but is more common in deliberately suspenseful narratives such as \textit{Gone Girl} and \textit{Great Expectations}. It means that there will be lots of moments of tension where something significant may happen but is not retrospectively. Or there can be the reverse, innocuous events which turn out to be important later. In \textit{Gone Girl}, there are small clues that \textit{Amy} may be missing and not murdered, that make not be picked up at the time, and are subsequently important. The model in this chapter is the same as the previous in adopting a reader model of incrementally reading the story. Therefore just like a human reader, it cannot know what will be important retrospectively. \textit{Salience} for the current occurring event is in the context of what the reader already knows and what happens immediately afterwards. False leads will be expected to be marked as \textit{salient}, and the small clues that turn out to be important, but don't seem significant at the time, will be expected to be inferred as not \textit{salient}. This is in contrast to the first view, where it is known as it can be inferred backwards from what happens at the end. Neither view is correct. The first view can be thought of as representing what is \textit{salient} when remembering a story. The second view of \textit{salience} is what is important while reading or viewing. The first view is more of what would be wanted for a summary written for a study guide, Wikipedia, or IMDB. The second is how the author or filmmaker would want the audience to, and do experience a story. There is like to an overlap in \textit{salient} events for both, but they are different concepts of \textit{salience}.

It should be noted that the Otake et al. implementation of BCF runs the method at each point in the story until the end, so this implementation fits more with the first view. However, this has a factorial complexity and is impractical for longer works beyond the short fairytales they evaluated. Therefore though the model is the same, the implementation means in practice, this chapter has a different view of \textit{salience} more in line with the second reader model. It is more local to the scene. The conclusion, Chapter \ref{chap:conclusion} presents some proposals for how the expectations can be extended further into the future with long book or movie script length works.

\begin{figure}[htbp]
\centering
\includegraphics[trim={1.0cm 4.0cm 0.5cm 0.2cm},clip,width=1.0\textwidth]{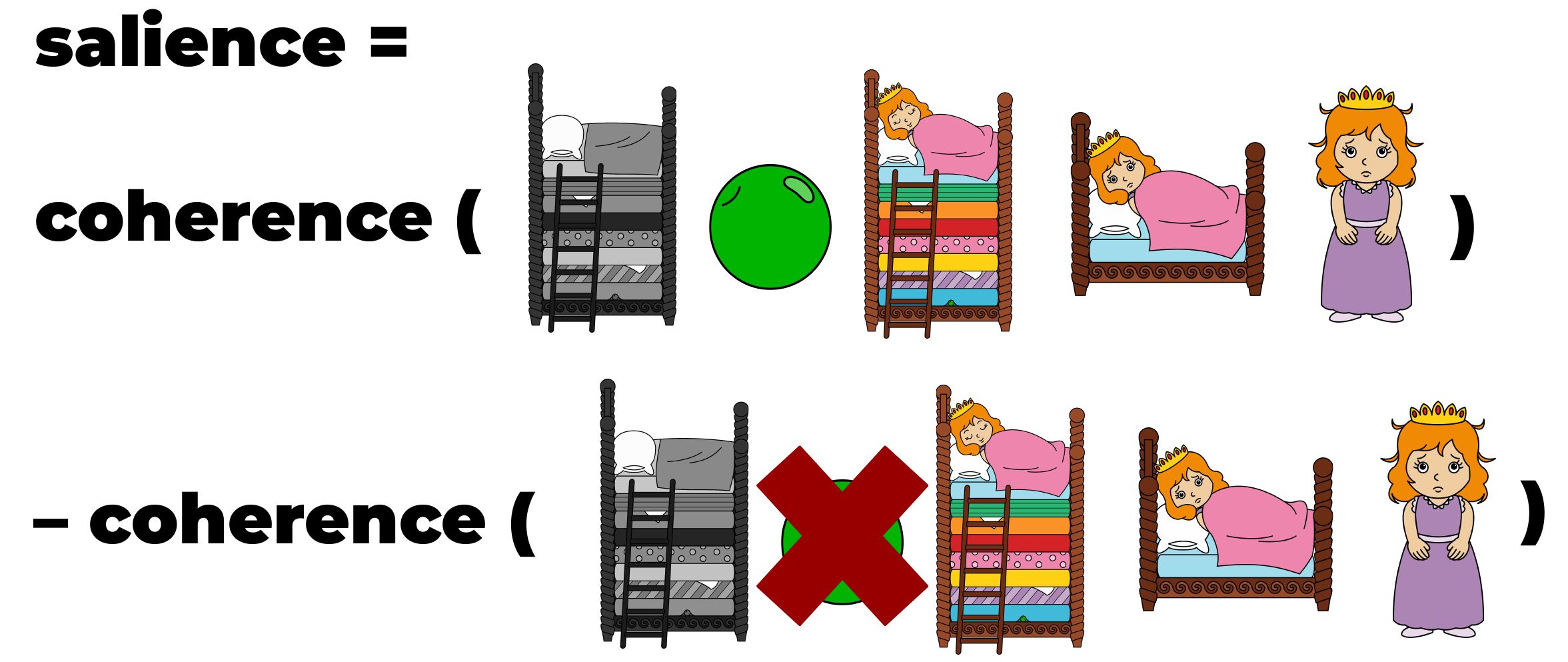}
\caption{Illustration of the BCF deletion method with an example from the fairytale \textit{The Princess and the Pea}. The greyscale represents the context or events that have been seen by the model, and the $x$ is the deleted event. The events are: There is a bed, a pea is placed on the bed, the princess tries to sleep, she is unable to sleep because of the discomfort caused by the pea, she is exhausted as she has been unable to sleep.}
 \label{fig:princess_and_the_pea}
\end{figure}

Conceptually the method is simple and illustrated in Figure \ref{fig:princess_and_the_pea}. Coherence in a story can be measured by any language model with average log-likelihood. \footnote{There are other vector-based approaches that are specified later.} Average log-likelihood will be lower probability when less expected events occur in the story. The BCF deletion method has a language model that conditions all of a story up until a given point. The past story is greyed out in the diagram. In the example, the current context is the pea. The method judges the salience by subtracting the coherence over the remainder of the story with or without the current event deleted. The idea is that deleting events will decrease coherence (or cause perplexity to go up) for ongoing events. The more significant an event is in a story, the more it will affect the log-likelihood of subsequent events, which will mean a higher salience score. In the \textit{Princess and the Pea} example, the pea event would be more salient since in the story causally the pea explains the Princess struggling to sleep later. As per Otake et al., it is assumed that each event is a sentence. Otake et al. experiment with methods other than event deletion, such as replacing specific with common verbs (\textit{do}, \textit{does}, or additionally with indefinite pronouns such as \textit{someone}. But as these methods performed slightly worse, only the deletion method is carried forward from their work. The exact formulation is later in Section \ref{sec:salience_methods}.

The method's appeal is that it perfectly aligns with Barthes' notion of a cardinal function. Any language model can also run it without any supervised \textit{salience} labels. Although salience is different from surprise or suspense, there are substantial overlaps from the earlier models in the thesis. For example, from Chapter \ref{chap:rolloutsuspense}, the Hale definition of surprisal is the negative likelihood of the following sentences. It measures surprise as how much a sentence changes the log-likelihood of a language model. The BCF deletion method is just this as a difference between the likelihood of when an event is present or not. With Ely surprise, there are also similarities as the model is the same as Hale, except it is a distance in the vector space and not a probability. There is a comparable difference metric based on vector states to parallel it. Ely suspense is more complicated since it is a probability distribution over future states, but, in principle, it has some overlap. A  BCF salient sentence is one in which deleting it causes a high level of future unexpectedness; an Ely suspenseful event is one where the expected variance of the future outcomes changes in relation to the current state. The biggest conceptual difference is that BCF measures the salience of the particular context event. In contrast, Ely suspense is the suspense of all the context, including the current state. So BCF salience would be expected to be spikier as it is the direct effect of the current event and doesn't slowly change with contextual changes. The advantage over Ely suspense is that running it only requires two passes with a language model and not generating a distribution over future states. It makes the method feasible for longer novels of greater than $100$k words. Finally, the model is also related to the turning points task. Turning points, by definition, are points in which the plot changes direction; if an event is deleted where the plot changes direction then it would have a much bigger effect on expected log-likelihood and hence be more salient.

This chapter also extends the BCF concept by exploring new measures of salience derived from structural manipulations. The model infers \textit{swap salience}, which is swapping rather than deleting an event within the BCF framework. \citet{Schmid+2008+17+34} discusses how an event can be salient if a reader expects it, but it is unexpected to the character in the story. The reader puts themselves into the character's shoes. \citet{zillmann1996psychology} emphasises how suspense is driven by anticipation and apprehension on behalf of characters the reader cares about. \citet{Bae2009SuspenseSO} propose to use this knowledge disparity between the reader and the character to create more suspenseful plots and hence more important events. \textit{Knowledge salience} is the difference between an expert-informed reader versus a naive one by taking the difference between the average log-likelihood of a base LM and an LM enriched with memory and a KB. Taking inspiration from the Ely surprise method, the model also infers  \textit{vector salience} which is the deletion method but relies on vector distance metrics rather than language model likelihood. The work also seeks to overcome the limitations of Otake et al.:

\begin{itemize}
  \item \textbf{Small and short dataset:} Their model is only evaluated on a short corpus of Russian fairytales. The aim is to scale the method to run on much longer book-length works. The evaluation is on aligned summaries to provide salience labels from the Shmoop corpus \citep{DBLP:journals/corr/abs-1912-13082}.
 \item \textbf{Limited Context:} Otake et al. model is based on GPT-2 which relies on the limited context of GPT-2 of $512$ word pieces. The aim is to extend the model with a memory mechanism.
\item \textbf{Knowledge}: In a similar way to the Ely suspense model, is limited by being able to generate plausible continuations. The limiting factor with BCF is how well the likelihood of the LM corresponds with the causal chain of events underlying it. While there are many ways to tackle this, the approach is an external case-based knowledgebase. 
\end{itemize}

While Otake et al. is also unsupervised the approach contrasts with related work on summarisation that relies on supervision using a proxy such as ROUGE score to align summaries with full text and train a model. For example in extractive summarisation \citep{Tsai2020ExtractiveSB}; abstractive summarisation \citep{tan-etal-2017-abstractive, pmlr-v119-zhang20ae}; combining both \citep{liu-lapata-2019-text}; salience for news \citep{liu-etal-2018-automatic-event,jindal-etal-2020-killed}; or using a probabilistic model for assigning labels for abstractive summarisation \citep{DBLP:journals/corr/abs-2004-03589}. In work more specific to the domain \citet{Papalampidi2019MoviePA} identifies supervised \textit{turning points}, and extends this to multimodal summarisation \citep{papalampidi-etal-2020-screenplay} with silver labels. As per the suspense work, the difficulty with a supervised approach is the cost of annotations, especially in a more subjective task, training annotators adequately and collecting enough annotations to have a high inter-annotator agreement. The evaluation is based on an aligned summary approach and is detailed in Section \ref{sec:salience_shmoop}. As with the suspense annotations, one advantage with an unsupervised model is the inferred salient events are completely independent of the labels. With supervised models, problems with the dataset construction can create shortcuts in training which can exaggerate test set performance.

\section{Extending an LM with Knowledge and Memory}

The limitation of context is addressed by incorporating an episodic knowledge retrieval mechanism (derived from RAG; \citealt{NEURIPS2020_6b493230}) and fusing this with a short-term memory mechanism that can extend the capabilities of a transformer LM. The intent is that the memory will learn to recall the most relevant parts of the story, act as an implicit index into these dimensions, and the KB will supplement this with typical plot knowledge. Characters, places, subplots ebb and flow in long stories, so the most relevant information may be hundreds of pages previous with mainly irrelevant information in-between, which suits indexed episodic memory rather than a transformer that must filter out the largely irrelevant details in-between.  The episodic memory can be thought of as acting as an index into crucial elements of the plot, which is essential for narrative comprehension. The intuition is similar to the five dimension indexing model \citep{1999-04086-005,doi:10.1207/s1532799xssr02032,Zwaan1995TheCO} where the reader needs to be able to index events and track \textit{time}, \textit{space}, \textit{causation}, \textit{motivation}, and \textit{protagonist} to comprehend a narrative. The event-indexing model has been supported by cognitive reading studies such as \citet{Rinck2003WhoWW}.

The main architectural innovation is to use an external knowledgebase, based on RAG (Retrieval Augmented Generation, \citealt{NEURIPS2020_6b493230}), and combine this seamlessly with a memory mechanism to improve the model's predictive performance. The structure of this model is to use a question and document encoder, both transformers, to learn and look up passages of text from a knowledgebase (based on DPR; \citealt{karpukhin-etal-2020-dense}) and then fuse this knowledge into a transformer encoder/decoder model such as BART \citep{lewis-etal-2020-bart} or T5 \citep{JMLR:v21:20-074}. Similar models, including REALM \citep{Guu2020RetrievalAL},
Hard EM \citep{min-etal-2019-discrete}, SpanSeqGen \citep{min-etal-2020-ambigqa}, and Fusion-in-Decoder \citep{izacard-grave-2021-leveraging}, have achieved state-of-the-art results in factual domains such as answering natural language questions, trivia or games such as Jeopardy. In these domains, the key insight is that offloading knowledge externally allows models to perform better than much larger transformers that need to encode all knowledge in their weights. All these models operate as a key value store with dense vectors to retrieve passages of text from a knowledgebase. The performance of the dense vector system surpasses that of earlier systems such as Wizard of Wikipedia \citep{DBLP:conf/iclr/DinanRSFAW19} and \citet{chen-etal-2017-reading} that have more traditional TF-IDF indices to a \textit{Wikipedia} dataset to improve the grounding of chat dialogue in facts, and question answering respectively. However, TF-IDF is fixed and adapting retrieval for the task is limited to retrieving more passages than needed and then applying an additional neural filter. The advantage the dense vector approaches have is that the question and answer encoders for the lookup can be trained to find the most relevant passages for the given task. 

The hypothesis, as supported by the reviewed literature in the Chapter \ref{chap:backgroundml}, is that knowledge is as crucial in comprehending and understanding the relationships between characters, and events in stories, as in factual domains. Therefore similar methods will improve the predictive power of the language model, which will enhance the performance of the BCF method. These methods that rely on retrieving raw text are also competitive with those that have tried to incorporate structured information such as GraphRetriever \citep{DBLP:journals/corr/abs-1911-03868} or PathRetriever \citep{Asai2020Learning}. The experiments are both with a \textit{Wikipedia} KB and \textit{Wikiplots}, a KB of story plot summaries. The motive for the latter is that these plot fragments or vignettes act as a planning system (or schema; \citealt{schank1977scripts}) guiding expectations. \citet{riedl2008story} used a similar concept in a rule-based system. \citet{sap-etal-2020-recollection} also use a bag-like episodic memory mechanism for inference in stories without the more sophisticated transformer encoders of the RAG model. 

After the experimental work in this chapter, a follow-up paper by \citet{DBLP:journals/corr/abs-2104-07567} on several RAG variants found that the KB was able to reduce the amount of hallucination in generating dialogue. The KB grounds the text generation in relevant facts retrieved from the KB. While the story domain is different intuitively, the same effect is desirable; inferring salience should be grounded either in plot knowledge from \textit{Wikiplots} or general knowledge from \textit{Wikipedia}, and also the memory of the previous character actions and plot developments. Another paper after this work that adopted a similar external KB approach is \textsc{RETRO} \citep{DBLP:journals/corr/abs-2112-04426}. The \textsc{RETRO} model is found to perform as well on a diverse set of tasks with $10$ times fewer parameters than the groups own comparable LM, Gopher \citep{DBLP:journals/corr/abs-2112-11446}. The model also performs similarly to two huge LMs, GPT-3 \citep{NEURIPS2020_1457c0d6} and Jurassic-1 \citep{J1WhitePaper}, both of which approach $200$ billion parameters with approximately $25$ fewer parameters and with substantially less training data, training time, and computational resources. Although \textsc{RETRO} is substantially larger, and post-dates the model in the chapter. It demonstrates the improvements across a range of tasks by augmenting an LM with an external knowledgebase. Another recent model, \textsc{KERN} \citep{DBLP:journals/corr/abs-2112-03254} that retrieves, from multiple external knowledgebases, achieves state of the art on commonsense reasoning Q\&A tasks.

The other main advantage of an external knowledgebase, as demonstrated in experiments later with alternative knowledgebases, is that they can be swapped and changed without retraining the underlying LMs. In the case of RAG, knowledgebase passages are retrieved via maximising the dot product of a vector encoded by the context and from the key of the stored passage, and so as long as the encoders remain the same, there is more flexibility for domain adaptation than a standard LM.

The problem with running transformers on longer texts is that the transformer attention mechanism has a complexity of $O(n^2 \cdot d)$ where $n$ is the length, and $d$ is the depth of the network. It is polynomial because each wordpiece attends to every other one. There has been a huge amount of recent work on faster and lighter transformers such as Longformer \citep{DBLP:journals/corr/abs-2004-05150}, see \citet{DBLP:journals/corr/abs-2009-06732} and \citet{DBLP:journals/corr/abs-2103-14636} for reviews of the many different variants. Through a variety of methods such as \textit{recurrence}, \textit{sparse attention}, \textit{Low-Rank factorization}, etc. they try to simplify the computational complexity so transformers can be run on sequences of $4$K, $8$K, and $16$K sequences. Nothing stops one of these models from being combined with RAG instead of BART or T5. However, just using a long-transformer on its own, even with $16$K wordpieces context, would fall far short of most of the works evaluated in the chapter. Also, as noted from discussing indexing and the background on narratology in any particular moment, most of the story is irrelevant. This leaves a lot of noise for the transformer to filter out; the episodic memory does it more naturally. It cannot also access external knowledge seamlessly with memory. Lastly, it is also far from cognitively plausible that readers are attending over every word \textit{Great Expectations} at each moment. Accessing a limited number of memories or knowledge fits far better with how the reader is likely to engage with the story.

\section{Methods}
\label{sec:salience_methods}

\subsection{Model}

The RAG model has been extended to incorporate a memory module, see Figure~\ref{fig:memory_rag}.\footnote{The code is provided via Github at \url{https://github.com/dwlmt/story-fragments/}}. Seen passages are added to the memory cache (configurable FIFO or LRU). The model retrieves $n$ passages, performs a lookup in both the KB and memory and then reranks them together using the dot product score between the question and document encoder vectors.  A significant benefit is that it naturally integrates both a short-term and long-term KB retrieval mechanism with a relatively simple design while allowing a powerful pre-trained LM (BART Large; \citealt{lewis-etal-2020-bart}) and retrieval systems (DPR; \citealt{karpukhin-etal-2020-dense}) from RAG to be retained. The architecture in more detail for both training and inference:

\begin{figure}[htbp]
\centering
\includegraphics[width=0.95\textwidth]{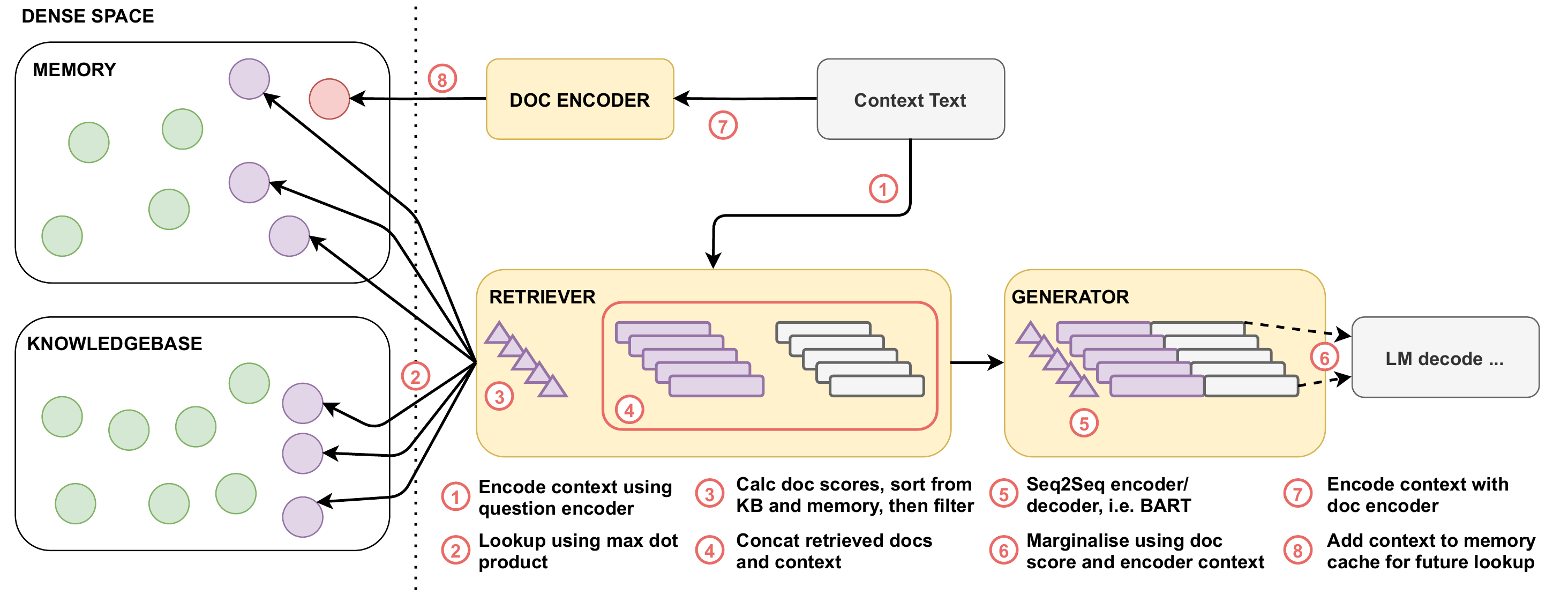}
\caption{Architecture of the memory RAG model: On the left-hand side are caches containing the permanent KB and transitory memory, which seen passages are added to. The Retriever encodes context text, looks up from both KB and memory, and concatenates the retrieved text to the context text. The generator, the BART encoder-decoder processes each passages concatenation, and marginalises over them to produce a distribution over output wordpieces.}
\label{fig:memory_rag}
\end{figure}

\begin{enumerate}
   \item \textbf{Encode Context:} The context is provided as a discrete number of sentences. The sentences can be pre-split for training datasets or dynamically for inference. The context sentences are encoded with the \textit{question encoder} from RAG. The encoder is a hierarchical bidirectional transformer that produces a vector of $768$ dimensions. The vector is the key for the lookup on both the knowledgebase and memory. 
   \item \textbf{Lookup:} The model allows either a KB and memory, either, or non (for ablation). Both memory and the KB rely on Faiss indices \citep{DBLP:journals/TBD/JohnsonDJ21}. Faiss is the library for the efficient search of dense vectors.\footnote{As per their own figures, the throughput and latency of Faiss can outperform traditional BM25 search, and the KB can scale to billions of vectors which makes it suitable for a large KB.} The KB is wrapped with the Datasets \citep{lhoest-etal-2021-datasets} library. The RAG KB and memory can be conceptually thought of as a dictionary where the query or question key and the stored passage answer keys are dense vectors. Search maximises the dot product between the query and the passage key. KB passages retrieved are blocks of sentences with associated metadata such as source or position in a story. The model retrieves a configurable number of $k$ from both memory and the KB separately if present, $k$ can be varied for each.
   \item \textbf{Calc doc scores and filter:} The extension for RAG to add the memory requires results for the knowledgebase and memory results be merged. It is simply ordering the dot product of the question and answer keys from the retrieved passages and taking the top $z$. So for example, if $10$ passages are retrieved from the KB and $10$ from memory, when $z = 10$ the passages are conditioned on are the highest scoring $10$ from both sources. The method allows the model to access the most relevant passages from memory or the KB, depending on the situation. This can vary as there might be a domain knowledge heavy context situation where more entries in the  \textit{Wikipedia} KB are relevant, or alternatively when events are referred to from earlier in the story, and so memory is more relevant. 
   \item \textbf{Concatenate:} The retrieved passages are each concatenated separately to the context text with a separator in between. The implementation reverses the order of the concatenation, so retrieved passages are after the context text. It makes truncation simpler, so only retrieved passages are truncated, as the combined context needs to fit within a length of $512$ wordpieces.
   \item \textbf{LM Encoder/Decoder:} The concatenated context and passages are run through the BART encoder/decoder. BART consists of a stacked bidirectional encoder, similar to BERT; an autoregressive transformer. Between the two are cross encoder so the decoder either when generating text or predicting tokens can attend to the whole context.
   \item \textbf{Marginalise:} Each passage retrieved are not weighted equally. The dot product from the question and answer key is normalised with softmax to produce a distribution over all the retrieved passages, and the probability weighting averages across all the outputs to produce a single output distribution across all the output wordpieces of the LM decoder. It means that high scoring passages are weighted more highly in the output. The process also allows the question and answer encoders to be finetuned via backpropagation. The marginalisation turns the retrieval into a form of hard attention mechanism.
   \item \textbf{Encode Context for Memory:} The whole point of a memory mechanism in long-running works is that memories can be retrieved later when relevant. Therefore, the model encodes all seen passages with the answer encoder (document encoder in RAGs usage).
   \item \textbf{Add Context to Memory:} The passage text and metadata such as position, title, source, etc., are saved to memory and are available when the next context lookup happens.  The default cache configuration is LRU (Least Recently Used), but the code also supports a FIFO (cache).  LRU is preferred because some passages are retrieved substantially more than others, and so if memory is limited, it is preferable they are in memory for longer.  There are two modes for the cache, training and inference:  In training, stories need to be interleaved, so there is only one passage for each story in the batch so that by the time the next one ($t_{1}$) is read, the previous context ($t_{0}$ is in memory. It means that multiple stories are being read at the same time.  One option would be to segregate the cache so that each story could only look up memories from the same story.  Instead, a single pool is used for all the training stories.  A single pool allows RAG to condition on a different story from the one being trained on.  The other stories act as negative examples since they provide distractors whereby the question encoders need to learn to distinguish between stories to recall useful memories.  From inspection, after a while of training, the models learn to mainly select memories from the correct stories since this is advantageous.  No comparison between the models was made on salience, but in perplexity on the validation set, the single pool performed similarly to the segregated model.  The training of the question encoder was longer, suggesting an ongoing training effect from the negative examples.  For inference, the memory cache is cleared between each work, and each inference is on a single story at a time.  It means only memories from the same work can be recalled, which is desirable in inference since distracting memories for other stories, while helpful for training the question encoder, would only degrade inference performance. 
\end{enumerate}

The notation follows from the RAG paper and the model derived from the RAG-Token model\footnote{There is an alternative RAG sequence model that doesn't marginalise over the all the passages but instead selects only the most relevant one. This may work for question answering where the answer may be in one place but for a story there are likely to be multiple pieces of knowledge from the KB and from memory that are relevant.}. Assuming $x$ is the original input or context text, and $y$ is the target label for upcoming text, and $z$ a passage of text from a retrieved document from the KB or memory, $t$ a time index from the place of the passage in the story, and $\theta$ the parameterisation of the model. The generation task, $p_{ \theta }(y_{t} \mid x,z,y_{1:t-1})$, is to predict $y_{t}$ by marginalising over the input, previously generated word pieces and the retrieved document, this is defined in eqn.~(\ref{eqn:rag_token}). Each next token is marginalised over all the retrieved $z$ passages. The respective probability varies for each $z$ at each step for each retrieved passage.
\begin{myequation}
\begin{split}
P(y \mid x)  &\approx   &\prod ^{T}_{t}  \sum_{z \in \text{Z}(p( \cdot \mid x))} p_{ \mu } (z \mid x)p_{ \theta }(y_{t} \mid x,z,y_{1:t-1})
\end{split}
\label{eqn:rag_token}
\end{myequation}

The top $z \in Z$, by default five, passages are retrieved by maximising the dot product, $p_{\mu}(z \mid x) \propto \exp(\mathbf{d}(z)^{T}\mathbf{q}(x))$, where $\mathbf{d}$ is the document encoder, and $\mathbf{q}$ the question encoder, both pre-trained from the DPR \textit{multiset} model, resulting in a bi-encoder setup. Only $\mathbf{q}$ is finetuned in training.  The text passages $z$, whether retrieved from the KB or memory, are then concatenated onto the original $x$ text and fed through the BART large encoder/decoder model. The memory mechanism for training is a single pool of up to $128$k vectors that operates as an LRU cache during training.

The principal training loss in eqn. ~(\ref{eqn:nll}) is simply the negative log likelihood over the batch as per the standard left-to-right decoder loss for BART. Normally BART also has a masked loss function on the encoder as well, but this isn't part of RAG since it would mask the text for retrieving from the KB.
Because the model marginalises the retrieved passages, back-propagation through this loss also updates the question encoder to retrieve more relevant passages. 

\begin{myequation}
 \mathcal{L}_{\text{nll}}(y) = \sum_{j} -\log  p(y_{j} \mid p_{j})
\label{eqn:nll}
\end{myequation}

In the default implementation, only the answer encoder is updated via back-propagation. It is necessary because, otherwise, the KB encoded before training begins would become stale. Other similar models such as REALM \citep{Guu2020RetrievalAL} iteratively update the KB during training, but this is not feasible with a large KB without vast GPU resources. Although with the memory extended RAG model, it is possible to train only with a memory and not a fixed external KB. As memory is a temporary cache in which entries are discarded, staleness is less of a problem. For the memory only models, both the question and answer encoders are finetuned. Being able to finetune the decoders gives the model an advantage over other models such as Fusion-in-Decoder \citep{izacard-grave-2021-leveraging} which integrate external passages in a different way. Fusion-in-decoder outperforms RAG on most factual domain tasks, but neither question nor answer encoders can be finetuned during a task.\footnote{Fusion-in-Decoder was released at the same time as the experimental salience work was being evaluated and so was not an option anyway.}

\subsection{Alternatives Losses}

As well as adapting RAG with memory, several other approaches tried to improve the performance of the default model. It is now standard to perform top-k or top-p sampling in text generation. The benefit of both methods is that by restricting the output to a limited number of wordpieces the probability mass is concentrated, which improves the quality of the generated text and reduces the chances of generation. The same problem applies to the standard softmax output for LMs where the probability mass is distributed over $50$K wordpieces in the case of BART or GPT. Likewise, the probability mass is distributed widely over all the tokens during training. It is because softmax doesn't allow zero probabilities. Various alternatives such as Sparsemax \citep{pmlr-v48-martins16} have been tried as a replacement which allow probabilities of zero. $\text{entmax}$ \citep{peters-etal-2019-sparse,correia-etal-2019-adaptively} is a generalisation of the model which allows the probability concentration to be changed with the $\alpha$ parameter. Softmax is when $\alpha = 1.0$ and Sparsemax is when $\alpha = 2.0$. The potential advantage of the method is that $\alpha$ should be trainable to identify the best probability concentration for the given task. $\text{entmax}$ is defined in eqn. \ref{eqn:define_entmax} and is based on the Tsallis family of entropies defined in \ref{eqn:tsallisdef}.

\begin{myequation}\label{eqn:define_entmax}
    \aentmax(\mathbf{z}) =
    \argmax_{\p \in \simplex^d} \mathbf{p}^\top\mathbf{z} + \HHt_{\alpha}(\mathbf{p}),
\end{myequation}

\begin{myequation}\label{eqn:tsallisdef}
     \HHt_{\alpha}(\mathbf{p}) =
\begin{cases}
\frac{1}{\alpha(\alpha-1)}\sum_j\!\left(p_j - p_j^\alpha\right)\!, &
\!\!\!\alpha \ne 1,\\
-\sum_j \pp_j \log \pp_j, &
\!\!\!\alpha = 1.
\end{cases}
\end{myequation}

The model with a revised loss should be a drop-in replacement for softmax with an appropriate changed loss function. The final softmax layer and loss were changed with the deep-spin library version.\footnote{\url{https://github.com/deep-spin/entmax}} The change was not successful, and the training proved unstable during change with different either fixed $\alpha$ or tunable values that allow sparsity. It is likely an artefact of combining the final output sparsity with the passage probabilities; a probability will be zero if any passages are zero of that particular wordpiece. 

The second extension is unlikelihood training. A problem with the TD-VAE generation model, when conditioned on and applied as a planning system, is that it produced less diverse stories. Generated stories tended to stick to the existing topic rather than advance the plot. It is not a unique problem, \citet{Holtzman2020The} has noted that transformer models can often produce dull and repetitive text. It's one of the problems planning systems are aiming to address. A recent approach is to reduce the repetitiveness in transformer generated text is unlikelihood training \citep{DBLP:conf/iclr/WelleckKRDCW20}. A log-likelihood loss increases the likelihood of an LM predicting the real token, and unlikelihood loss penalises the model if it generates sequences of tokens. The penalising loss is defined in eqn. \ref{eqn:unlikelihood_loss}. The set $C_t$ are sequences of tokens; the loss goes down if the LM doesn't generate each sequence with the set $C_t$. Welleck et al. apply the loss as shown in eqn. Eqn. \ref{eqn:unlikelihood_loss_add} where the are two sets of $C_t$, \textit{copy} and \textit{repeat}. The \textit{copy} set contains a set of all $4$ consecutive wordpiece tokens in the context. The idea is that the LM shouldn't generate sequences that have happened before except for shorter coreferences. \textit{Repeat} are all the $4$ token sets of wordpieces generated in the output of the LM, and so it should stop the LM from generating repeated sequences in the output.

\begin{myequation}
 \ell_{\text{un}} = \sum_{t=1}^{|y|}\sum_{y \in C_{t}}\beta(y_c) log (1 - p_{\theta}(y_{c} \mid \mathbf{x},y<{t} \And y \neq y_c))
\label{eqn:unlikelihood_loss}
\end{myequation}

\begin{myequation}
 \ell =  \ell_{\text{nll}} + \alpha( \ell_{\text{un}}(C^{\text{repeat}}) +  \ell_{\text{un}}(C^{\text{copy}}))
\label{eqn:unlikelihood_loss_add}
\end{myequation}

When the work in the chapter was completed, unlikelihood training had not been applied specifically to story tasks. The model has been applied successfully to dialogue tasks by  \citet{li-etal-2020-dont} and \citet{lagutin-etal-2021-implicit}. The motive is largely similar to narrative tasks; a dialogue where a model generates similar text to earlier in the dialogue or repeats passages from an external knowledgebase is boring. The characters and places may have been seen before within stories, but the story needs to develop. The intent is that unlikelihood training will encourage more plot development. The model was implemented as per Wellack et al.'s \textit{copy} and \textit{repeat} variants. One difference is, because RAG concatenates the KB and memory passages, all the $4$ n-gram token sequences are part of the excluded set of tokens. The model should draw on the passages but not repeat them. Improving plot progression and diversity should improve both salience  

Unlike $\alpha-\text{entmax}$, the models train well with unlikelihood training. However, the results are not reported in this chapter. The reason is that the results for salience are close to similar results for the non-likelihood models, within $1.0$ to $2.0$ performance points either way, with nearly being clearly better. Story generation is not evaluated in this chapter, but examples are given later for analysis. The generated examples show quite a big difference between style and progression between models with and without the likelihood of loss. The likely cause is that salience is more localised to the influence of a single sentence on a longer block of text and is less sensitive than a task that requires the generation of longer passages of text. The unlikelihood loss is revisited in the discussion and with future work.

\subsection{Knowledgebases}

Two knowledgebases are part of the evaluation:

\begin{itemize}
  \item \textbf{Wikipedia:} Is a pretrained KB from DPR (Dense Passage Retrieval; \citealt{karpukhin-etal-2020-dense}). The KB is $21$ million passages from the English language Wikipedia. The entries are created by splitting Wikipedia into contiguous $100$ word blocks. The DPR encoders are trained on supervised question and answer tasks to maximise the similarity of the correct question and answer pairs versus negative examples. As the KB is the whole of \textit{Wikipedia} it contains a huge amount of factual information about people, places and other specialised domains. The intent behind using the dataset is that this broad general knowledge will help a language model lookup material relevant to the context and improve the model's predictive performance. Examples in action from \textit{Great Expectations} are when \say{Pip turns out of the Temple onto Fleet Street} then entries from Wikipedia from Fleet Street or places on Fleet Street such as the Court of Chancellory or Pubs are looked up. When a character in the book is rowing, it brings up sources talking about rivers, boats, and marshlands.  
  \item \textbf{Wikiplots:} Is a new dataset encoded with the same encoders as the \textit{Wikipedia} dataset. The datasets are circa $135$K plot summaries extracted from books, movies, tv-series, plays, etc., from Wikipedia.\footnote{The dataset was recreated from code at \url{https://github.com/markriedl/WikiPlots}}. Each entry in the KB is $6$ sentence chunks from each plot. A sliding window means that each entry is overlapped by $3$ sentences which means the before and after contex for each part of the plot are available across multiple entries. \citet{riedl2008story} with a case-based story planning system, has a KB consisting of plot vignettes. The idea is the same in that genres such as \textit{action}, \textit{romances}, \textit{crime}, etc. often have similar plots. A KB that can look up similar plots can condition on typical events that occur in similar plots and it can guide both generation and with salience compprehending what is likely to happen. As summaries are by definition compressed, a single entry contains far more plot progression than the memory entries will have, and so acts more like a template for a longer section of a plot. 
\end{itemize}

It should be noted there are two different versions of RAG with differently trained encoders. All KBs and models are based on the pretrained multiset version of RAG, which is trained on multiple factual domain tasks. The choice is simply because the model trained on a wider range of tasks transfers better across domains.

\paragraph*{Wikipedia Entry Example (Fleet Street):} Fleet Street is a major street mostly in the City of London. It runs west to east from Temple Bar at the boundary with the City of Westminster to Ludgate Circus at the site of the London Wall and the River Fleet from which the street was named. The street has been an important through route since Roman times. During the Middle Ages, businesses were established and senior clergy lived there; several churches remain from this time including Temple Church and St Bride's. The street became known for printing and publishing at the start of the 16th century, and it ...

\paragraph*{Wikiplots Entry Example (Gone Girl Plot):} On their fifth wedding anniversary, writing teacher Nick Dunne returns home to find his wife Amy missing. Her disappearance receives press coverage, as Amy was the inspiration for her parents' popular Amazing Amy children's books. Detective Rhonda Boney finds poorly concealed evidence of a struggle in the house. Suspicion mounts around Nick, whose apathy is interpreted by the media as characteristic of a sociopath and even sows doubt in his twin sister Margot. In the past Amy revealed to Nick that Amazing Amy was a perfected version made up of the real Amy's failures. Their marriage disintegrated over time; both lost their jobs in the recession and moved from New York City to Nick's hometown of North Carthage, Missouri, to care for his dying mother.

\subsection{Datasets}

As the focus is on longer works, all the training datasets are longer-form narrative works datasets:

\begin{itemize}
  \item \textbf{BooksCorpus:} \citet{DBLP:conf/iccv/ZhuKZSUTF15} provides a large corpus of longer novel-length works and is used for training. However, \textit{BooksCorpus} consists of free books scraped from \href{https://www.smashwords.com/}{Smashwords}; these works are highly slanted towards particular genres such as romance and fantasy which are unlike the evaluated task, which is mainly classic works.
\item \textbf{Gutenberg:}  To supplement BooksCorpus an additional training dataset from Gutenberg using the \href{https://github.com/c-w/gutenberg}{c-w/gutenberg} library filtered to only English language fictional works. 
\item \textbf{Movie scripts:} Another important area of longer-form storytelling is movies or dramatic works.  So, to improve diversity, the Movie Scripts datasets \citep{ramakrishna-etal-2017-linguistic} is used.
\end{itemize}

Multi-dataset models performed better on the validation set in training than single corpus models, so only these are evaluated. The training set sizes are \textit{BooksCorpus} circa $18$k works, \textit{Gutenberg} $27$k, and \textit{Movie Scripts} $1.5$k. Sentences are split with Blingfire.

\subsection{Baselines}

The primary baselines for salience prediction come from \citet{otake-etal-2020-modeling}. \textit{Random} just randomly assigns a salience score to each sentence position. \textit{Ascending} assigns scores that increase per position. \textit{Descending}, the reverse, assigns decreasing scores per position. The intuition behind these benchmarks is that important information can be clustered at the beginning or end of a story or chapter.

Otake et al. use TF-IDF clustering approach as a baseline. Instead a BERT derived clustering summarisation approach is adopted from \citet{DBLP:journals/corr/abs-1906-04165}. The method uses k-means to cluster BERT sentence vectors according to the number of desired sentences and then selecting the sentences closest to the centroids. The rationale is that sentences should be clustered but topics or situation and so by choosing the centroids the model is picking the best representative of each cluster. It is a transformer neural network equivelant of the Otake et al. TF-IDF baseline. Since \textit{salience} scores are required, the method is adapted to output the cosine distance from the centroid as a salience score. $k$ is set so that there is one cluster for every $10$ sentences. One change from Miller is to use the \textit{stsb-roberta-large} sentence transformers model \citep{reimers-gurevych-2019-sentence}, which has sentence embeddings that perform much better on a range of semantic tasks than raw BERT. 

\subsection{Training}

Datasets are read in an interleaved or round-robin fashion so that only one $(x,y)$ pair from each story is in a batch. Batches are sliding windows of $12$ sentences for both $x$ and $y$ with a $k$ of five passages to retrieve. The combined context for the concatenated encoder text is truncated to $512$ word pieces, and the max length for the decoder is $128$. The model is trained with a batch size of $32$. RAG has a delimiter separating retrieved text when concatenating for BART. 

To allow the model to train on 12GB GPUs, we use the zero optimisation memory saving features of DeepSpeed \citep{10.1145/3394486.3406703}, which also necessitates using FP16, with gradient checkpointing for the model. Training uses the base version of the RAG multiset encoders and the original pre-trained BART Large. We finetune with Adam \citep{DBLP:journals/corr/KingmaB14} with a learning rate of $2^{-6}$.\footnote{The environment can be setup either via the \textit{requirements.txt} file with pip on the Anaconda \textit{environment.yaml} file, both in the Github repository.} 

The preprocessing for all datasets is the same:

\begin{enumerate}
  \item Sentence splitting using Blingfire.\footnote{Blingfire (\url{https://github.com/microsoft/BlingFire}) is preferred to Spacy for work in the chapter as it is much faster and so better for longer works. Also from inspection it seems to be better than Space 2.0 at splitting whole sentences.}
  \item Stories are randomly shuffled according to a fixed seed.
  \item There is a 80/10/10 training/validation/test split but this is only used for early stopping in training since evaluation is on separate datasets - ProppLearner and Shmoop. 
\end{enumerate}

\begin{itemize}
  \item \textbf{Training}
    \begin{itemize}
        \item \textbf{Config}: The config files are in the \textit{training\_config}. The reported models are the 12 sentence block variants without experiment entmax or unlikelihood training which isn't reported.
        \item \textbf{Models}: The model files will be made available via Github. BART Large has $400$M parameters, the question encoder has $110$M parameters and the doc encoder has $110$M parameters. In the baseline model all apart from the doc encoder is finetuned, and all are finetuned with the memory only models.
        \item \textbf{Policy}: All models were trained with batch size $20000$ instance per epoch in training and $2000$ for validation. The early stopping policy is to stop training after $3$ epochs without improvement.
        \item \textbf{Time}: For the baseline model, training took 9 days. Other models are comparable.
        \item \textbf{Epochs}: Baseline model training ran for 11 epochs, again other models are similar.
        \item \textbf{Validation Performance}: The best validation loss is $398$ (sum of the log likelihood) from $694$ on an untrained model.
    \end{itemize}
  \item \textbf{Inference}
     \begin{itemize}
     \item \textbf{Computation}: Inference computation depends on which salience measures are enabled. The main salience BCF method requires two passes through the text. Adding in knowledge or swap salience adds another pass for each. This is because the text must be passed through with an without each structural change. With all methods enabled for long works such as \textit{The Brothers Karamazov}, \textit{Emma}, \textit{Moby Dick}, or \textit{Great Expectations} all $>150K$ words inference time is typically $4-6$ hours running on a single GPU. This is pretty reasonable given the length of the works, and obviously much shorter novels and plays have proportionally shorter inference time. 
     \item \textbf{Memory}: The base configuration uses $28$GB of general purpose memory, this needs to be increased to $64$ is the full Wikipedia KB with $23$M passages is used.
     \end{itemize}
\end{itemize}

\subsection{Inference}

\textit{Salience} detection is based on the BCF method \citep{otake-etal-2020-modeling}. Let $S$ be the set of all sentences. The \textit{salience} is $\sigma$. For BCF this uses an event removal function $r$ and coherence evaluator $c$. $c$ is the difference in coherence between when the sentence $t$ is present and removed in eqn. (\ref{eqn:del}) for the following $n$ sentences. Note that $r$ can be used more broadly as a structural manipulation function. In this paper $r$ is also used for \textit{swap} function and a \textit{knowledge difference} function, these are described later.
\begin{myequation}
    \sigma(S_t,S_{\{1:n\}}) = c(S_{\{1:n\}}) - c(\tilde{S}_{\{1:n\}})
\label{eqn:del}
\end{myequation}
The coherence eqn. (\ref{eqn:coh}) and eqn. (\ref{eqn:coh2}) are the average log-likelihood of the word pieces following sentences up to the maximum word pieces of the label, normalised by the length, eqn. ~(\ref{eqn:z}).
\begin{myequation} 
\begin{split}
    &c(S_{\{1:n\}}) = Z  \log P^{}_{}(S_{\{t+1:n\}} \mid S_{\{1:t-1\}}, S_{t}) 
\end{split}
\label{eqn:coh}
\end{myequation}
\begin{myequation} 
\begin{split}
    &c(\tilde{S}_{\{1:n\}}) = Z \log P^{}_{}(S_{\{t+1:n\}} \mid S_{\{1:t-1\}}, r(S_t))
\end{split}
\label{eqn:coh2}
\end{myequation}
\begin{myequation} 
\begin{split}
    Z = \frac{1}{\lvert S_{\{t+1:n\}} \rvert} 
\end{split}
\label{eqn:z}
\end{myequation}
\textit{Salience} detection is based on the BCF method \citep{otake-etal-2020-modeling}. Let $S$ be the set of all sentences. The \textit{salience} is $\sigma$. For BCF this uses an event removal function $r$ and coherence evaluator $c$. $c$ is the difference in coherence between when the sentence $t$ is present and removed in (\ref{eqn:del}) for the following $n$ sentences. Note that $r$ can be used more broadly as a structural manipulation function. In this paper $r$ is also used for \textit{swap} function and a \textit{knowledge difference} function, these are described later.

One difference is Otake et al. run \textit{salience} from each deleted sentence to the end of the story, which is factorial complexity for the number of sentences. While it works for short stories, it is infeasible on novel-length works, so the re-implementation of \textit{salience} is more localised and run over the next $128$ LM wordpieces.

As well as BCF Salience, several other measures for \textit{salience} are explored. Additionally, there are experiments with \textit{knowledge salience}, which measures the difference between \textit{salience} with the RAG KB and Memory enabled versus with it disabled. \textit{Swap salience} follows the same structure as sentence deletion, but the $r$ function swaps the order of the sentences rather than deleting them and so tests order dependence as a form of salience. The sentiment is another relevant factor in whether something is salient; more emotional passages, either negative or positive, might be more salient. VADER sentiment \citep{Hutto_Gilbert_2014} is used as an adjustment factor for other salience measures $salience \cdot (1.0 + abs(\text{sentiment}))$ where sentiment is the absolute values of the sentiment in the range $0.0-1.0$. In addition, following from Chapter \ref{chap:rolloutsuspense}, measures are defined based on embeddings: $E$ is the average of the word piece vectors from the BART encoder after marginalisation. The first measure is the cosine distance from subsequent vectors, defined as \textit{Ely surprise} $\text{cos dist}(E_{t},E_{t-1})$. The second measure takes a vector distance rather than average log-likelihood in the sentence deletion BCF method to create a version based on an embedding, not LM decoding. The evaluated measures are:
\begin{itemize}
  \item \textbf{Clus-Sal}: The clustering baseline.
  \item \textbf{Like-Sal}: The main BCF measure described based on the negative log-likelihood of the BART decoder with the deletion method.
  \item \textbf{No-Know-Sal}: The same as but with both the memory and KB disabled as per Otake et al. It is an ablation measure since it tests BART on its own without any KB or memory augmentation.
  \item \textbf{Like-Imp-Sal}: Uses sentiment to adjust the salience from the \textit{Like-Sal} method. As per the suspense work, the hypothesis is that already salient sentences will be more salient if they have stronger absolute sentiment, as these are like to correspond with more dramatic parts of the story. 
  \item \textbf{Like-Clus-Sal}: Combining the \textit{Like-Sal} and {Clus-Sal} measures via weighted addition: $\text{Clus-Sal} + 2 \cdot \text{Like-Sal}$. The rationale for the measure is that Otake et al. found improvements by combing BCF with TF-IDF clustering method. This is the equivalent but with the neural clustering method.
  \item \textbf{Like-Clus-Imp-Sal}: The same as \textit{Like-Clus-Sal} except with addition of the \textit{Imp} adjustment.
  \item \textbf{Know-Sal}: The difference between average log-likelihood of the LM with the KB and memory on versus off, \textit{knowledge salience}. It is a simple attempt to model differences between a naive and informed reader. As reviewed in Chapter \ref{chap:backgroundtheory} some theories of suspense have posited that suspense can happen when the reader is aware of something the protagonist is not. The opening shower scene in \textit{Psycho} is the classic example. 
  \item \textbf{Swap-Sal}: Uses the same BCF approach but swaps rather than deletes a sentence to test structural ordering. It is a way of testing how order dependent events are within the text with the idea that salient events that are more central to the plot. 
  \item \textbf{Emb-Surp}: The \textit{Ely surprise} cosine distance measure. It is the same as Ely surprise in Chapter \ref{chap:rolloutsuspense} except that the LM has the assistance of memory and the KB. Also, there is no sentence embedding, and so the representation is the average of the encoder embeddings within BART marginalised over all the retrieved passages. As per the earlier work, Cosine distance is used. Other measures such as L1, L2, etc. were tried but performed worse and so are not reported.
  \item \textbf{Emb-Sal}: \textit{Salience} based on above embedding distance, not average log-likelihood. The same except as \textit{Like-Sal} it applies the deletion BCF method with the \textit{Emb-Surp} distances, and so it is a vector-based formulation of the BCF method.
\end{itemize}
The evaluation is run on the following model variants: With the \textit{Wikiplots} dataset, with the \textit{Wikipedia} dataset, and with just \textit{Memory} enabled and additional finetuning of the document encoder.

\section{Experiments}

\subsection{Model Perplexity Improvements}

The major innovation of the RAG derived model is incorporating the KB and memory mechanism into the LM, and therefore, it needs to be tested what impact it has as a general LM. Table~\ref{tab:perplexity} shows the baseline model median perplexity with combinations of KB and memory access turned off. When the perplexity is lower then the model is better at predicting what will happen next which should in theory make it better for inferring salience. The corpus for evaluation is the Shmoop corpus detailed in Section \ref{sec:salience_shmoop}.

\begin{table}[htbp]
\centering
\begin{tabular}{@{}lc@{}}
\toprule
\textbf{Model}     & \textbf{Perplexity $\downarrow$} \\ \midrule
LM+Mem+KB & 19.44         \\
LM+KB & 19.37         \\
LM+KB(Wikipedia) & 19.94 \\
LM+Mem & \textbf{15.95}         \\
LM & 66.00         \\
LM+Scram(Mem+KB) & 60.21 \\

\bottomrule
\end{tabular}
\caption{Median perplexity of the baseline model. Plus means that type of memory or KB is enabled. Scram means that random passages have been retrieved from the KB and memory. Wikiplots KB unless Wikipedia is specified. All over the first five stories in the dataset using $20$ retrieved passages.}
\label{tab:perplexity}
\end{table}

The best model is the baseline with only the memory, and the KB turned off. Both versions of the KB, on their own and memory combined, are slightly worse and around the same perplexity. The crucial difference is that \textit{LM}, the model with neither, is far worse, and scrambling, which retrieve random passages, is only slightly better. Scrambling shows that it is the ability to learn which are the relevant passages to retrieve that is important rather than just any additional context. Overall, these results validate that memory and KB are hugely improving the predictive power of the LM.

\subsection{ProppLearner}

Following on from the BCF paper \citep{otake-etal-2020-modeling}, there is an evaluation of the ProppLearner task derived from the Propp dataset \citep{Finlayson2017ProppLearnerDA}, a richly annotated corpus of 15 Russian fairytales translated into English. The Proppian functions \citep{Propp1958MorphologyOT} with which this corpus is annotated define stereotypically essential roles in the classic Russian fairytale. There are $31$ roles in total. They represent the key events of a story's plot. Examples of the functions include \textit{the villain receives information about his victim}, \textit{the villain causes harm or injury to a member of family}, \textit{the hero and the villain join in direct combat}, \textit{the villain is punished}. The roles are particular to the genre but were highly influential on later theories of narratology. As per Otake et al., all plot role sentences are treated as salient. The results are reported using MAP (Mean Average Precision; \citealt{Manning2008IntroductionTI}). Though the dataset is small, the evaluation is a useful regression test against Otake et al. 

\begin{table}[htbp]
\centering
\begin{tabular}{@{}lc@{}}
\toprule
\textbf{Model}     & \textbf{MAP $\uparrow$} \\ \midrule
Random             & .213         \\
Ascending          & .277         \\
Descending         & .185         \\
TF-IDF             & .279         \\ \midrule
Otake Sal       & .280         \\
Otake Comb Sal   & .301         \\ \midrule
Clus-Sal & .275         \\
Like-Sal & \textbf{.319}         \\
Like-Imp-Sal & .313         \\
Know-Diff-Sal & .309         \\
Swap-Sal & .236  \\
Emb-Surp & .247         \\
Emb-Sal & \textbf{.319}  \\\bottomrule
\end{tabular}
\caption{Compare Otake et al. baselines and models with RAG equivalents. }
\label{tab:propp_table}
\end{table}

 All the RAG models are the baseline used with different variants of the \textit{salience} measures. The best RAG models, see Table \ref{tab:propp_table}, measures \textit{Like-Sal} and \textit{Emb-Sal} score slightly better than Otake et al.'s model. However, this validation is limited as the Propp dataset is tiny, with only $15$ stories of less than $150$ sentences and limited annotations. It would be expected that the knowledgebase and memory, in particular, would have a much bigger impact on longer works of fiction, which are evaluated in the next section. It nevertheless is a replication that shows the method can work and surpass the baselines with an alternative LM and architecture. Other alternatives such as \textit{know-sal} and \textit{swap-sal} also perform better than the Otake et al. baselines but by a smaller margin. Also, as per the \textit{suspense} work, the \textit{Imp} factor doesn't seem to improve the model.
 
 \begin{figure}[htbp]
\centering
\includegraphics[trim={0.5cm 1.0cm 2.0cm 2.5cm},clip,width=1.0\textwidth]{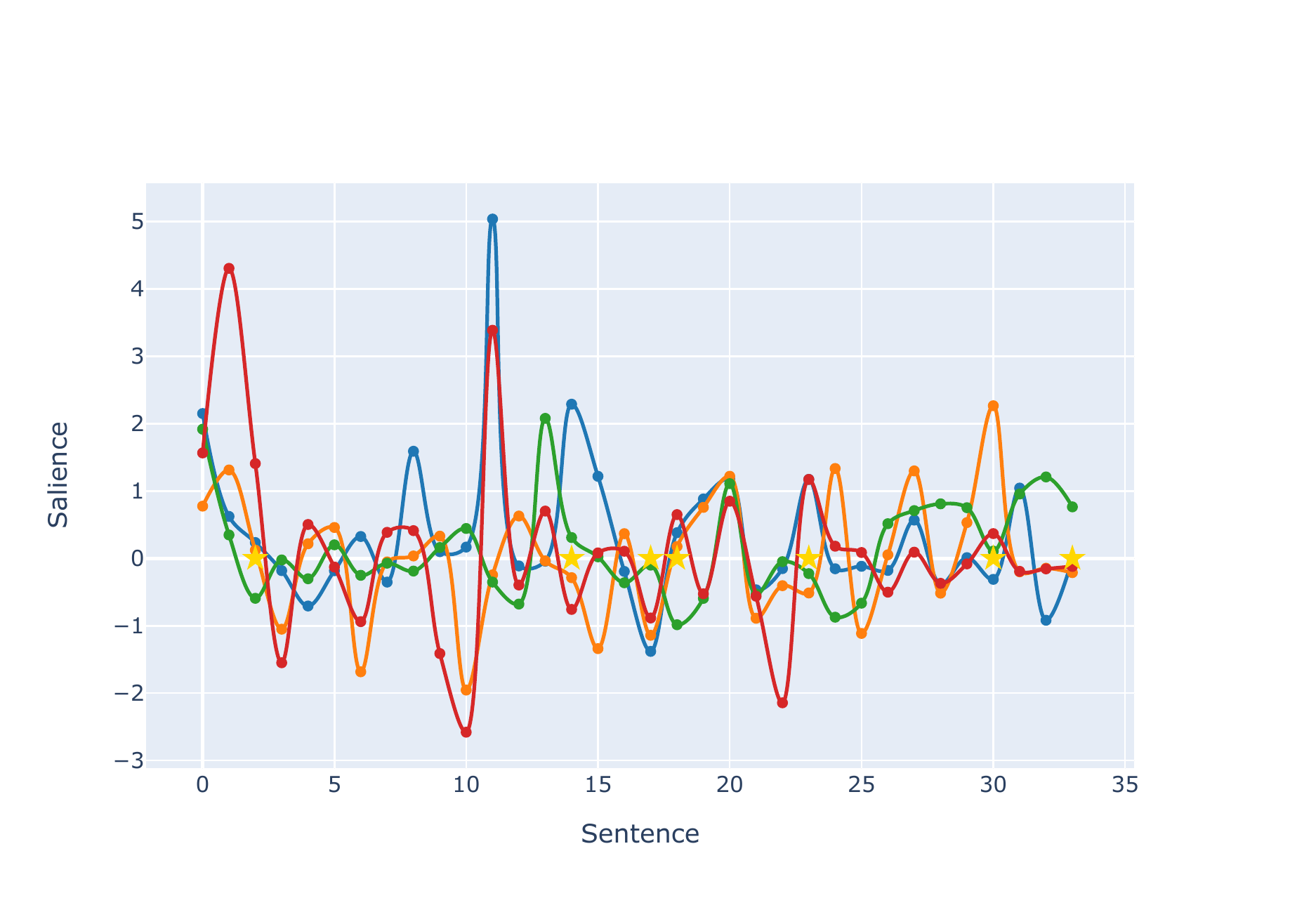}
\caption{A salience plot of the Propp story \textit{Nikita the Tanner}:
\textbf{\textcolor{plotacolor}{Like-Sal}}, \textbf{\textcolor{plotbcolor}{Clus-Sal}}, \textbf{\textcolor{plotccolor}{Otake-Sal}}, \textbf{\textcolor{plotdcolor}{Emb-Sal}}, {\color{yellow} $\medstar$} \textbf{Propp roles}. Plots are rescaled to unit variance and scaled.}
 \label{fig:propp_nikita}
\end{figure}

\begin{figure}[htbp]
\centering
\includegraphics[trim={0.3cm 1.0cm 2.0cm 2.5cm},clip,width=1.0\textwidth]{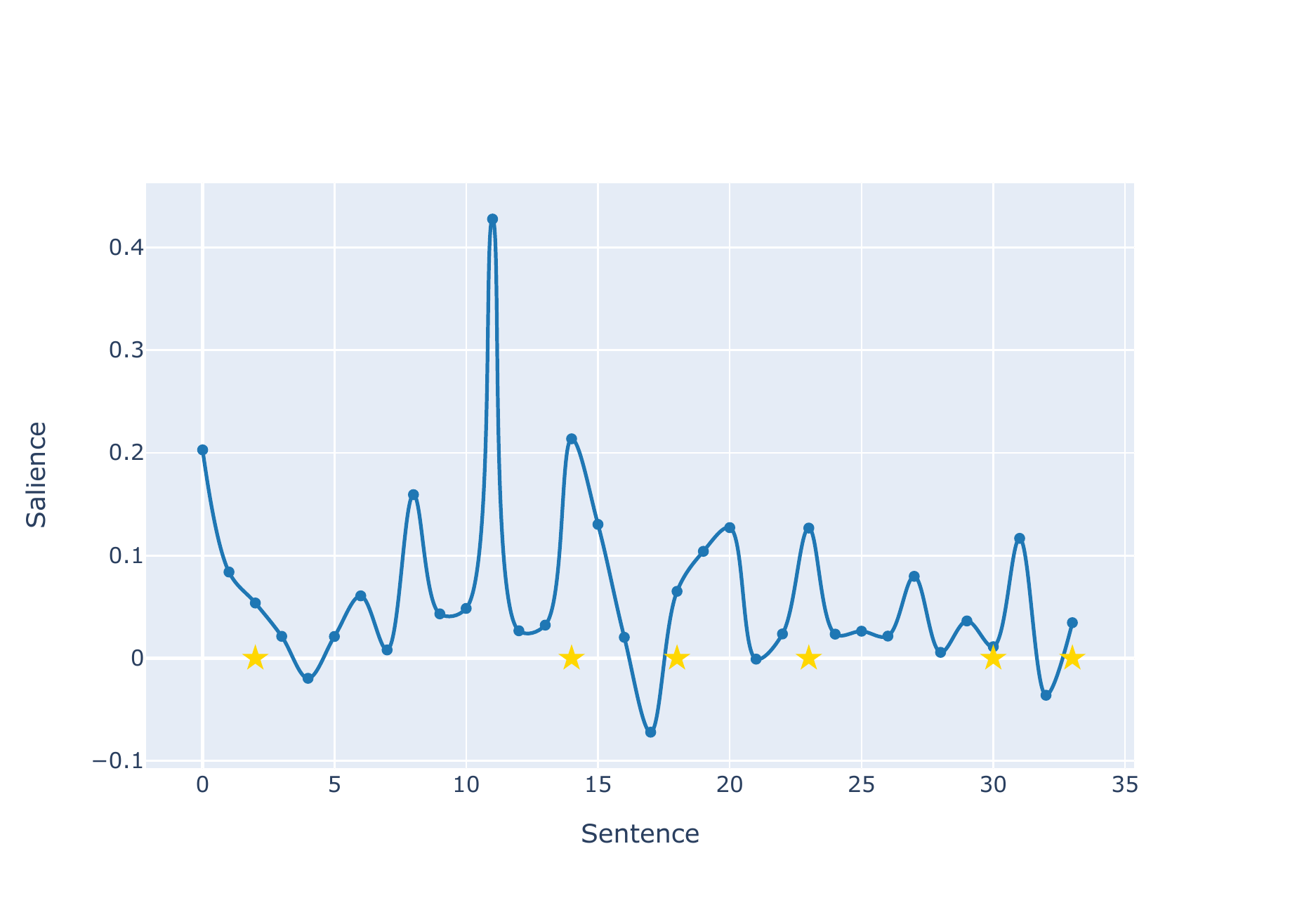}
\caption{Like-Sal plotted for \textit{Nikita the Tanner}. Unlike the multiplot version, the output is not scaled and so shows the real value of the average log likelihood difference.}
 \label{fig:propp_nikita_all}
\end{figure}

There is another benefit to the shorter form of the Propp fairytales over the Shmoop evaluation; the shortness of the stories makes it more straightforward to analyse what the salience method is doing than a long book of greater than $100$K words. To illustrate, there are two examples: Figure \ref{fig:propp_nikita} shows four different measures plotted against the story, \textit{Nikita the Tanner}. The plotted \textit{Otake-Sale} metric is the  \textit{No-know-sal} which is the closest equivalent with the new model. It is the same as Otake et al. but with BART large rather than GPT-2. Figure \ref{fig:propp_nikita_all} has the main \textit{Like-Sal} method on it's own. The second story \textit{The Magic Swan-Geese} is in Appendix \ref{appendix:salience}, Section \ref{sec:swan}. With both stories \textit{Like-Sal} and \textit{Emb-Sal} are closer at matching the annotations than the \textit{Clus-Sal} baseline or \textit{Otake-Sal}. There are a couple of cases where the annotations exactly match a peak, but it is more insightful to look at the missed examples. For example, in \textit{Nikita the Tanner} the highest peak in the whole story is sentence $11$, which is \textit{When the princess heard this , she wrote her father to find Nikita the Tanner in Kiev and to send him to deliver her from captivity}. There is no annotated function anywhere this peak. However, one functional role is \textit{the seeker agrees to or decides upon counteraction}. The seeker in this story is the Princess since she wants the help of the Hero. So arguably, the peak is really a Propp function, although it is not annotated as such. Coincidentally in the \textit{The Magic Swan-Geese} sentence $11$ is also the whole story peak for the story which is \textit{Stove , stove , tell me, whither have the geese flown ?} In the story, the annotated function is the sentence before, \textit{the girl guessed that they had carried off her brother, and rushed after them.} The predicted salient sentence is where the story situation and the girl asks the stove; the following sentences are the dialogue with the stove, and hence it makes sense it would be marked by the BCF method. The story is also a good indicator of what the BCF method is doing. In the same story, when she first asks the stove about the Apple tree, when she meets a Hedgehog, and when the Goose tries to attack her are all predicted as salient. The same happens in the Nikiti story, where there are salient peaks when he refuses to fight the Dragon and when he threatens to destroy the Dragon's lair. All the salient peaks are marked by changing topics, context, or direction in the immediate following few sentences.\footnote{One interesting aside, in both \ref{fig:propp_nikita_all} and \ref{fig:propp_swan_all} and later diagrams in the chapter there can be negative salience. Strictly speaking an informatic theoretic perspective new information would not reduce the predictive power and so negative salience would be impossible. However, in practice there are irrelevant asides it stories that do make predictions of future sentences worse, and so have a negative salience, as removing them helps comprehension.}

The full text for \textit{Nikita the Tanner}:

\begin{enumerate}
\setcounter{enumi}{-1}
\item A dragon appeared near Kiev ; he took heavy tribute from the people - a lovely maiden from every house , whom he then devoured .
\item Finally , it was the fate of the tsar 's daughter to go to the dragon .
\item He seized her and dragged her to his lair but did not devor her , because she was a beauty .
\item Instead , he took her to wife .
\item Whenever he went out , he boarded up his house to prevent the princess from escaping .
\item The princess had a little dog that had followed her to the dragon 's lair .
\item The princess often wrote to her father and mother .
\item She would attach her letter to the dog 's neck , and the dog would take it to them and even bring back the answer .
\item One day the tsar and tsarina wrote to their daughter , asking her to find out who in this world was stronger than the dragon .
\item The princess became kindlier toward the dragon and began to question him .
\item For a long time he did not answer , but one day he said inadvertently that a tanner in the city of Kiev was stronger than he .
\item When the princess heard this , she wrote her father to find Nikita the Tanner in Kiev and to send him to deliver her from captivity .
\item Upon receiving this letter , the tsar went in person to beg Nikita the Tanner to free his land from the wicked dragon and rescue the princess .
\item At that moment Nikita was currying hides and held twelve hides in his hands ; when he saw that the tsar in person had come to see him , he began to tremble with fear , his hands shook , and he tore the twelve hides .
\item But no matter how much the tsar and tsarina entreated him , he refused to go forth against the dragon .
\item So they gathered together five thousand little children and sent them to implore him , hoping that their tears would move him to pity .
\item The little children came to Nikita and begged him with tears to go fight the dragon .
\item Nikita himself began to shed tears when he saw theirs ."
\item He took twelve thousand pounds of hemp , tarred it with pitch , and wound it around himself so that the dragon could not devor him , then went forth to give him battle .
\item Nikita came to the dragon 's lair but the dragon locked himself in .
\item \say{ Better come out into the open field , } said Nikita , \say{ or I will destroy your lair together with you ! }
\item And he began to break down the door .
\item The dragon , seeing that he could not avoid trouble , went out to fight in the open field .
\item Nikita fought him for a long time or a short time ; in any event , he defeated him .
\item Then the dragon began to implore Nikita : \say{ Do not put me to death , Nikita the Tanner ; no one in the world is stronger than you and I .}
\item \say{Let us divide all the earth , all the world , into equal parts ; you shall live in one half , I in the other . }
\say{ Very well , } said Nikita , \say{ let us draw a boundary line .}
\item He made a plow that weighed twelve thousand pounds , harnessed the dragon to it , and the dragon began to plow a boundary from Kiev ; he plowed a furrow from Kiev to the Caspian Sea .
\item \say{ Now , } said the dragon , \say{ we have divided the whole earth . }
\item \say{ We have divided the earth , } said Nikita , \say{ now let us divide the sea ; else you will say that your water has been taken . }
\item The dragon crawled to the middle of the sea ; Nikita killed him and drowned him in the sea .
\item That furrow can be seen to this very day ; it is fourteen feet high .
\item Around it the fields are plowed , but the furrow is intact ; and those who do not know what it is , call it the rampart .
\item Nikita , having done his heroic deed , would not accept any reward , but returned to currying hides .
\end{enumerate}

\subsection{Shmoop Alignment Dataset}
\label{sec:salience_shmoop}

\begin{table}[htbp]
\centering
\begin{tabular}{@{}lll@{}}
\toprule
\textbf{Summary}                                                                                                                                                                                                                                             & \textbf{Full Text}                                                                                                                                                                                                                                                                                                                                                                                                                   & \textbf{Score} \\ \midrule
\begin{tabular}[c]{@{}l@{}}Then, two huge columns \\ of water shoot from it, \\ knocking the crew down.\end{tabular}                                                                                                                                         & \begin{tabular}[c]{@{}l@{}}The electric light suddenly went\\ out, and two enormous waterspouts\\ crashed onto the deck  of the frigate,\\ racing like a torrent from stem  to\\ stern, toppling crewmen, breaking\\ spare  masts and yardarms from\\ their lashings.\end{tabular}                                                                                                                                                       & 0.723          \\ \midrule
\begin{tabular}[c]{@{}l@{}}He grabs the extinguisher \\ cap thing and tries to smother\\ the kid/grandpa ghost with it.\end{tabular}                                                                                                                         & \begin{tabular}[c]{@{}l@{}}In the struggle, if that can be called\\ a struggle  in which the Ghost with\\ no visible resistance  on its own part\\ was undisturbed by any effort of its\\ adversary, Scrooge observed that its\\ light  was burning high and bright;\\ and dimly  connecting that with its\\ influence over him, he seized the\\ extinguisher-cap, and by a sudden\\ action pressed it down upon its head.\end{tabular} & 0.600          \\ \midrule
\begin{tabular}[c]{@{}l@{}}Sara wants to say that she\\ already knows French, but \\ she doesn't know how to say\\ so and ends up giving Miss \\ Minchin the impression that\\ she's being difficult and doesn't \\ want to learn the language.\end{tabular} & \begin{tabular}[c]{@{}l@{}}Miss Minchin was a very severe and\\ imposing person, and she seemed so\\ absolutely sure that Sara knew\\ nothing whatever of French that she\\ felt as if it would be almost rude to \\correct her.\end{tabular}                                                                                                                                                                                        & 0.722          \\ \midrule

\bottomrule
\end{tabular}
\caption{Example showing the alignment of summary with full-text sentences, the score is cosine similarity. The examples are all chosen because the previously used SRL event extraction fails with this approach and matches incorrect sentence and the examples show different strengthes of matches and types of sentence.}
\label{tab:saliency_alignment_short}
\end{table}

\begin{table}[htbp]
\centering
\begin{tabular}{@{}lll@{}}
\toprule
\textbf{Summary}                                                                                                                                                                                                                                             & \textbf{Full Text}                                                                                                                                                                                                                                                                                                                                                                                                                   & \textbf{Score} \\ \midrule
   
\begin{tabular}[c]{@{}l@{}}Emerson still feels rough\\ about ruining the lecturer’s talk \\ in the chapel.\end{tabular}                                                                                                                                      & \begin{tabular}[c]{@{}l@{}}But Mr. Emerson, contrite and unhappy, \\ hurried away to  apologize to the \\ Rev. Cuthbert Eager.\end{tabular}                                                                                                                                                                                                                                                                                          & 0.486          \\ \midrule
\begin{tabular}[c]{@{}l@{}}She thinks Dinah should find a \\ nice man and settle down.\end{tabular}                                                                                                                                      & \begin{tabular}[c]{@{}l@{}}And then you might get married to\\ some decent man, and there'd be plenty\\ ready to have you, if you'd only leave\\ off that preaching, as is ten times worse\\ than anything your Aunt Judith ever did.\end{tabular}                                                                                                                                                                                                                                                                                          & 0.334          \\ \midrule

\begin{tabular}[c]{@{}l@{}}According to him, the driftwood\\ is dry and ideal for starting a fire.\end{tabular}                                                                                                                                      & \begin{tabular}[c]{@{}l@{}}It is now dry and would\\ burn like tinder.\end{tabular}                                                                                                                                                                                                                                                                                          & 0.630          \\ \midrule

\begin{tabular}[c]{@{}l@{}}Detectives were sent to each port\\ in England to see if the money might \\be recovered.\end{tabular}                                                                                                                                      & \begin{tabular}[c]{@{}l@{}}As soon as the robbery was discovered,\\ picked detectives hastened off to \\Liverpool, Glasgow, Havre, Suez, Brindisi,\\ New York, and other ports, inspired by\\ the proffered reward of two thousand\\ pounds, and five per cent. on the sum\\ that might be recovered.\end{tabular}                                                                                                                                                                                                                                                                                          & 0.618          \\ 

\bottomrule
\end{tabular}
\caption{Continued example table showing the Shmoop full text alignment examples.}
\label{tab:saliency_alignment_short2}
\end{table}

Ideally, to evaluate this thesis on longer works, there would be a set of Gold standard annotations with the salient sentences. Typically even short novellas can be over $20$K words, and more normal novels longer than $50$K words. More sweeping works such as \textit{Anna Karenina}, \textit{Wuthering Heights}, \textit{The Fellowship of the Ring}, or \textit{David Copperfield} can be well over $100$K words. Per-sentence annotations for longer works such as novels and plays are prohibitively expensive. This is especially true when multiple annotators are required to ensure high inter-annotator agreement. It would also not be possible with insufficiently trained and lower cost crowdsourced workers. Reading a local passage would not be enough as it is only possible to judge \textit{salience} over the whole narrative, which can be tens of thousands of words. This requires strong comprehension and thus requires skilled annotators and is a daunting annotation task. Instead, this paper builds on a variant of an approach for event \textit{salience} in news articles \citep{liu-etal-2018-automatic-event,jindal-etal-2020-killed}. The method is to align expert-written summaries with the full text, tagging sentences that align with the summary as salient, thus turning the evaluation into a binary ranking problem. The intuition is that the summary will mention only salient events and themes.

The approach is to use the \href{https://github.com/achaudhury/shmoop-corpus}{Shmoop corpus} \citep{DBLP:journals/corr/abs-1912-13082}, which contains classic works of literature, such as \textit{Moby Dick}, but also plays such as \textit{A Midsummer Night's Dream}, and short stories including \textit{The Mask of the Red Death}. The Shmoop corpus has stories split into chapters with aligned summaries. These bullet point summaries, albeit colloquial in style, are professionally written as study guides for students. They are written with a deep understanding of the plots and the salient events in them, which can serve as a valid proxy for salience. Conceptually they are also similar to the \textit{ProppLearner} evaluation, although without specific Proppian roles, which are unused anyway for binary \textit{salience} classification. It also aligns with the BCF concept, as if events from the summary are removed, they would significantly alter the plot.\footnote{There are occasional exceptions, such as summary points that discuss themes of the overall work and not specific plot events, but these are rare.} \citet{jindal-etal-2020-killed} align summaries to text by using BERT \citep{devlin-etal-2019-bert} to match constituent parts of events extracted from semantic role labels (SRLs). However, in testing, this performed poorly. Unlike news, the story summaries are more loose descriptions of events, which the SRL method struggles with. The method seems to match common verbs such as \textit{give}, \textit{go}, \textit{walk}, etc. when a looser match across the whole sentence is better. Instead, the method employing an S-Bert transformer \citep{reimers-gurevych-2019-sentence} on the whole sentence worked much better in aligning summaries to the full text. The method is as follows:

\begin{enumerate}
  \item Split aligned chapters into sentences, $S_t$ for summaries and $F_t$ for the full text.
  \item Extract sentence embeddings using the \href{https://www.sbert.net/}{Sentence Transformers} model \textit{stsb-roberta-large} \citep{reimers-gurevych-2019-sentence} , $r(S_t)$ and $r(F_t)$ .
  \item Calculate cosine similarity for all pairwise $r(S_{t})$ and $r(F_{t})$ for $t \pm \rho$, where the range is $\rho = 10.0$\% and the valid range for $t$ is $x \in \mathcal{X}$ for $S_{t}$, and $y \in \mathcal{Y}$ for $F_{t}$.
  \item Mark up to $k$ as salient sentences for all sentence pairs in the alignment window $s(x, y) =cos\_sim(r(S_{t_x}),r(F_{t_y})$ where:
  \begin{itemize}
    \item $k = 3$
    \item $s(x, y) \ge \mu$, $\mu = 0.35$
    \item $s(x, y) \ge  \argmax_{x \in \mathcal{X}, y \in \mathcal{Y}} s(x, y) - \theta$, where $\theta = 0.05$
  \end{itemize} 
\end{enumerate}
The evaluated threshold of $\mu = 0.35$ is relatively low because the summary must match something in the text. From observation even if the similarity score is relatively low, the process still usually picks the correct sentence. Pairs of summary and full-text sentences are matched within a percentile range. The rationale is that matches are likely to occur in the full text in a roughly similar position to the summary. There are up to three targets per summary sentence, as the summary sentences often compress information with multiple clauses and sometimes there are near identically suitable matches. The $\theta$ of $0.05$ ensures that there are only multiple matches if the other matches are close to the best match.\footnote{An alternative to be the first match over a threshold. The problem with this is that whatever threshold is chosen it can be a worse match than the best one, and would mean in some circumstances the best match wouldn't be chosen.} It is to ensure that a summary sentence will only match the full text when they are close to being as good matches as each other, either because it's not clear which the best one is, or because the summary refers to multiple events. If there is one clear best match then only that is chosen. The advantage of this method is that it allows automated evaluation of \textit{salience}  to scale to longer works that test the memory and KB mechanism of the model without excessive annotation cost. The silver Shmoop annotations are on $226$ titles, spanning $6,939$ chapters with $214,617$ silver standard labels. Each chapter averages $148$ sentences with an average of $31$ labelled as salient using the criteria specified. 

Examples of the alignment analysis are in Table \ref{tab:saliency_alignment_short} and Table \ref{tab:saliency_alignment_short2}. The alignment example shows that across a diversity of different examples, despite the informality of Shmoop, there are largely correct alignments. However, there may be some noise and a limited number of errors in the labels because the training of the models and the inference methods are entirely unconnected with the labelling process. In contrast to some supervised methods, the model cannot take advantage of artefacts in the labels to exaggerate inference performance. While human evaluation is usually preferred, even with the resources available, annotating long novels and having a high-level inter-annotator agreement would be a daunting task. Expertly written summaries with a high-quality alignment provide the best option for evaluating a large corpus of texts. One note where there is a difference is that the salience model is a reader model. The process models the reader's expectations as they read the story. The salience model is also more localised in that it measures the influence of the sentence on the next half dozen sentences. On the other hand, the Shmoop corpus is a retrospective summary of the work. The difference is that details may be important locally but not included in the summary as they are not important within the whole story. Or vice versa, where seemingly innocuous details become important later. While the overall salience should be broadly similar, there is likely to be a tendency to underestimate performance.

\begin{figure}[htbp]
\centering
\includegraphics[trim={0.3cm 1.0cm 2.0cm 2.5cm},clip,width=1.0\textwidth]{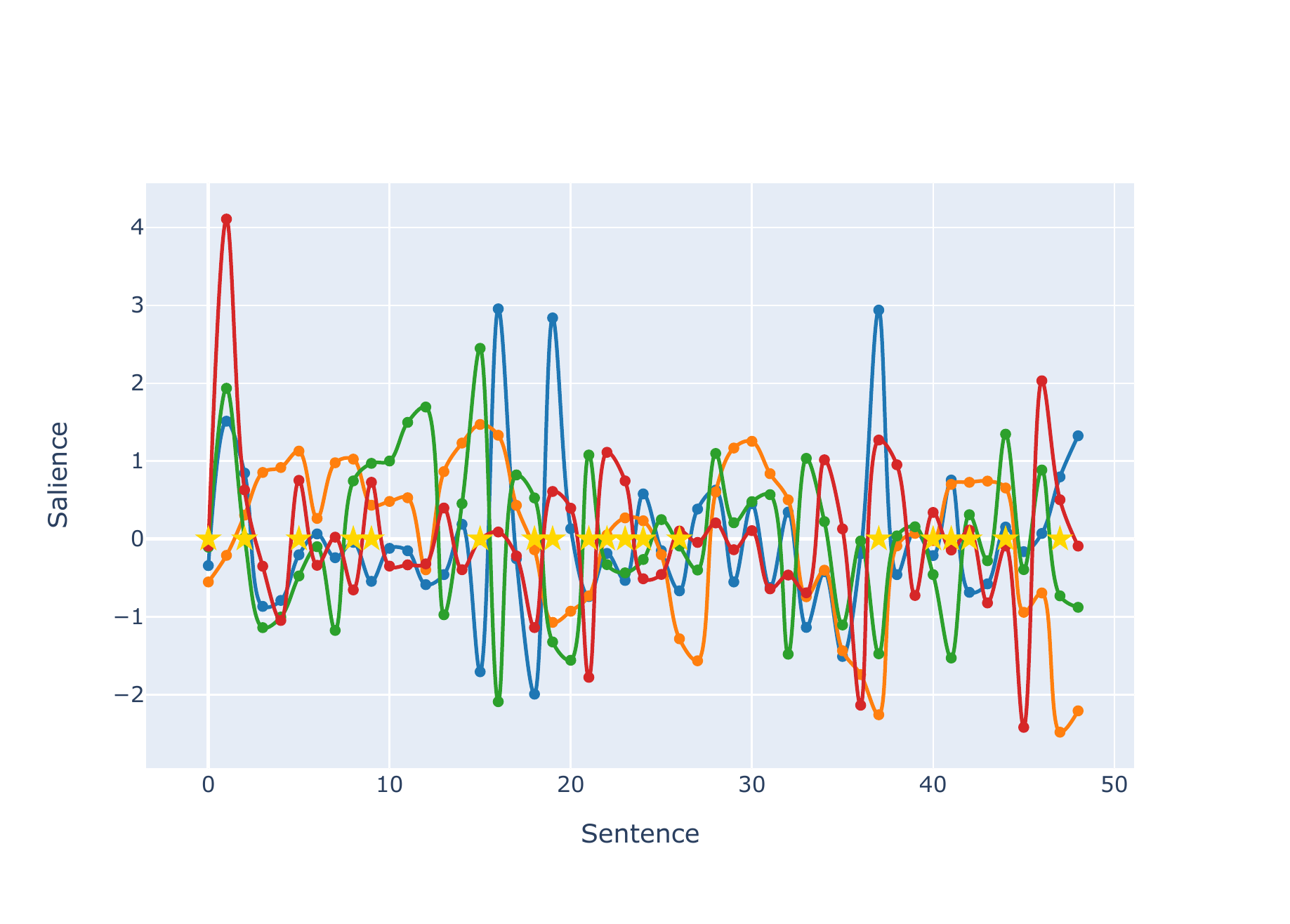}
\caption{The plot of \textit{Macbeth} Act 5, Scene 8 (the dagger soliloquy scene, \url{https://www.shmoop.com/study-guides/literature/macbeth/summary/act-5-scene-8}): \textbf{\textcolor{plotacolor}{Like-Sal}}, \textbf{\textcolor{plotbcolor}{Clus-Sal}}, \textbf{\textcolor{plotccolor}{Otake-Sal}}, \textbf{\textcolor{plotdcolor}{Emb-Sal}}, {\color{yellow} $\medstar$} \textbf{Shmoop aligned labels}. Plots are rescaled to unit variance and scaled.}
 \label{fig:macbeth_all}
\end{figure}

\begin{figure}[htbp]
\centering
\includegraphics[trim={0.2cm 1.0cm 2.0cm 2.5cm},clip,width=1.0\textwidth]{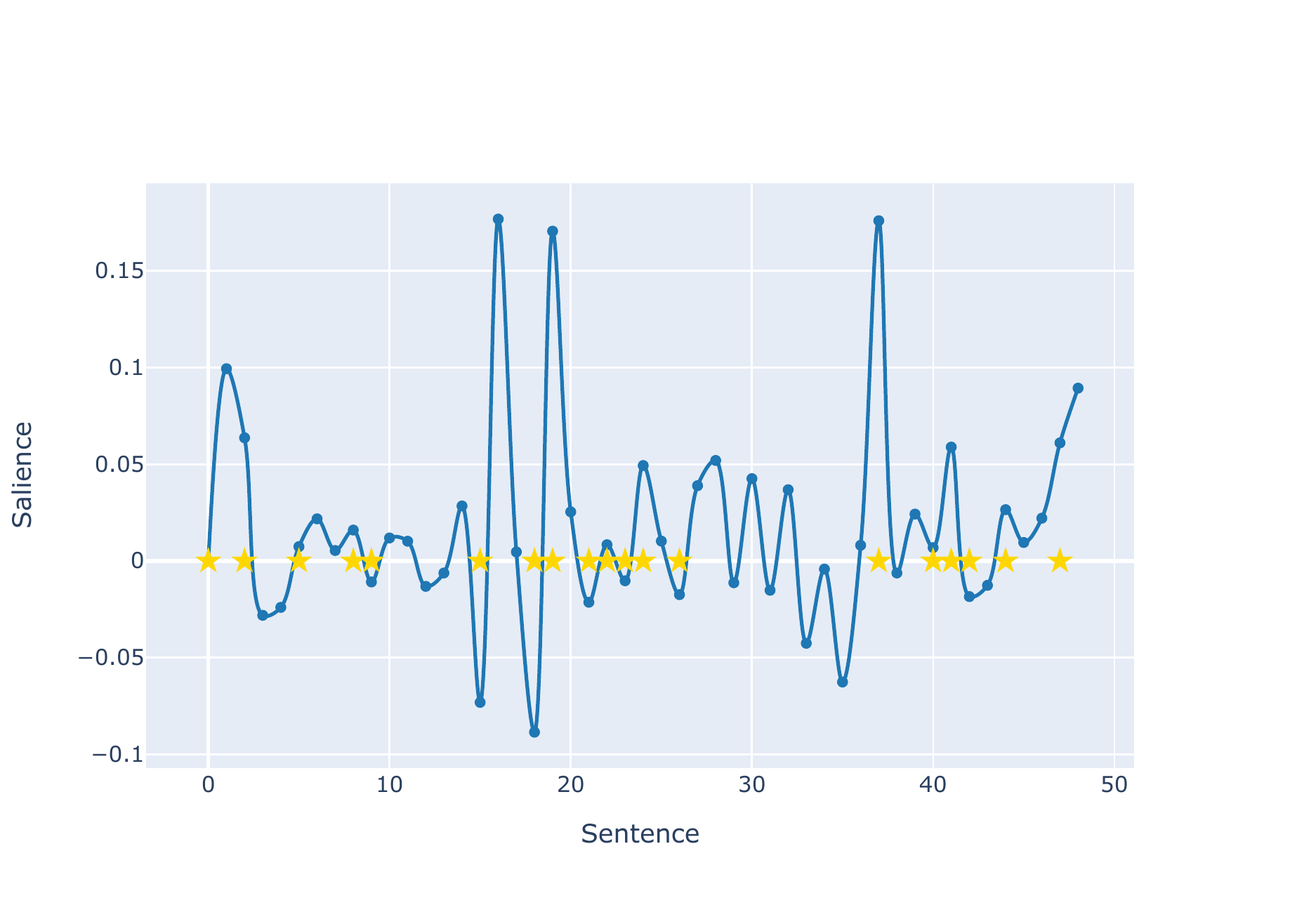}
\caption{The plot of \textit{Macbeth} Act 5, Scene 8 (the fight scene between Macbeth and Macduff, \url{https://www.shmoop.com/study-guides/literature/macbeth/summary/act-5-scene-8}): Only \textbf{Like-Sal} is plotted.}
 \label{fig:macbeth_like_sal}
\end{figure}

\begin{figure}[htbp]
\centering
\includegraphics[trim={0.3cm 1.0cm 2.0cm 2.5cm},clip,width=1.0\textwidth]{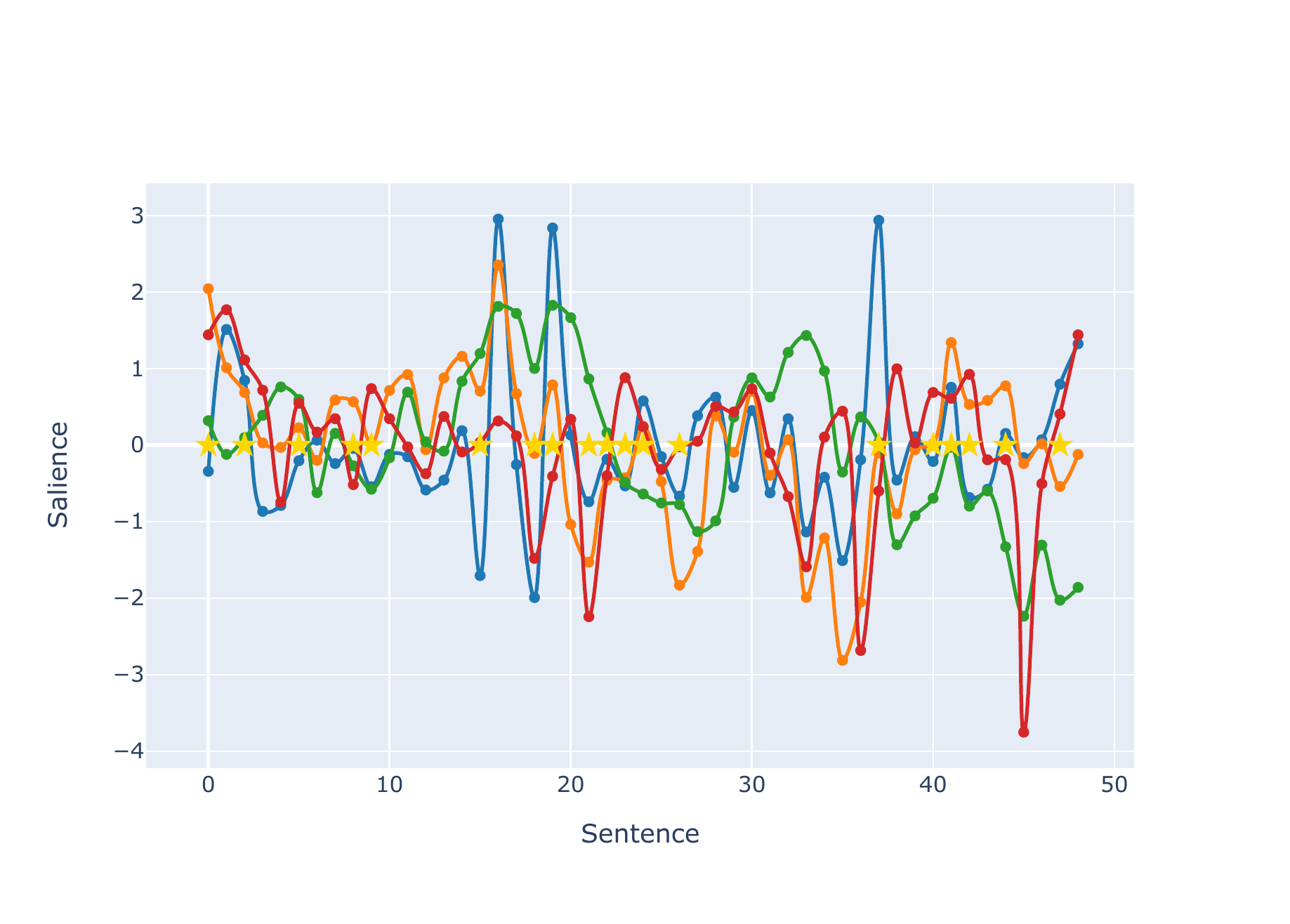}
\caption{The plot of \textit{Macbeth} Act 5, Scene 8 (the fight scene between Macbeth and Macduff, \url{https://www.shmoop.com/study-guides/literature/macbeth/summary/act-5-scene-8}): \textbf{\textcolor{plotacolor}{Like-Sal}}, \textbf{\textcolor{plotbcolor}{Like-Clus-Imp-Sal}}, \textbf{\textcolor{plotccolor}{Know-Diff-Sal}}, \textbf{\textcolor{plotdcolor}{Emb-Surp}}, {\color{yellow} $\medstar$} \textbf{Shmoop aligned labels}. The same plot as previously but with alternatives metrics plotted against \textit{Like-Sal}. Plots are rescaled to unit variance and scaled.}
 \label{fig:macbeth_alternative}
\end{figure}

\begin{figure}[htbp]
\centering
\includegraphics[trim={0.2cm 1.0cm 2.0cm 2.5cm},clip,width=1.0\textwidth]{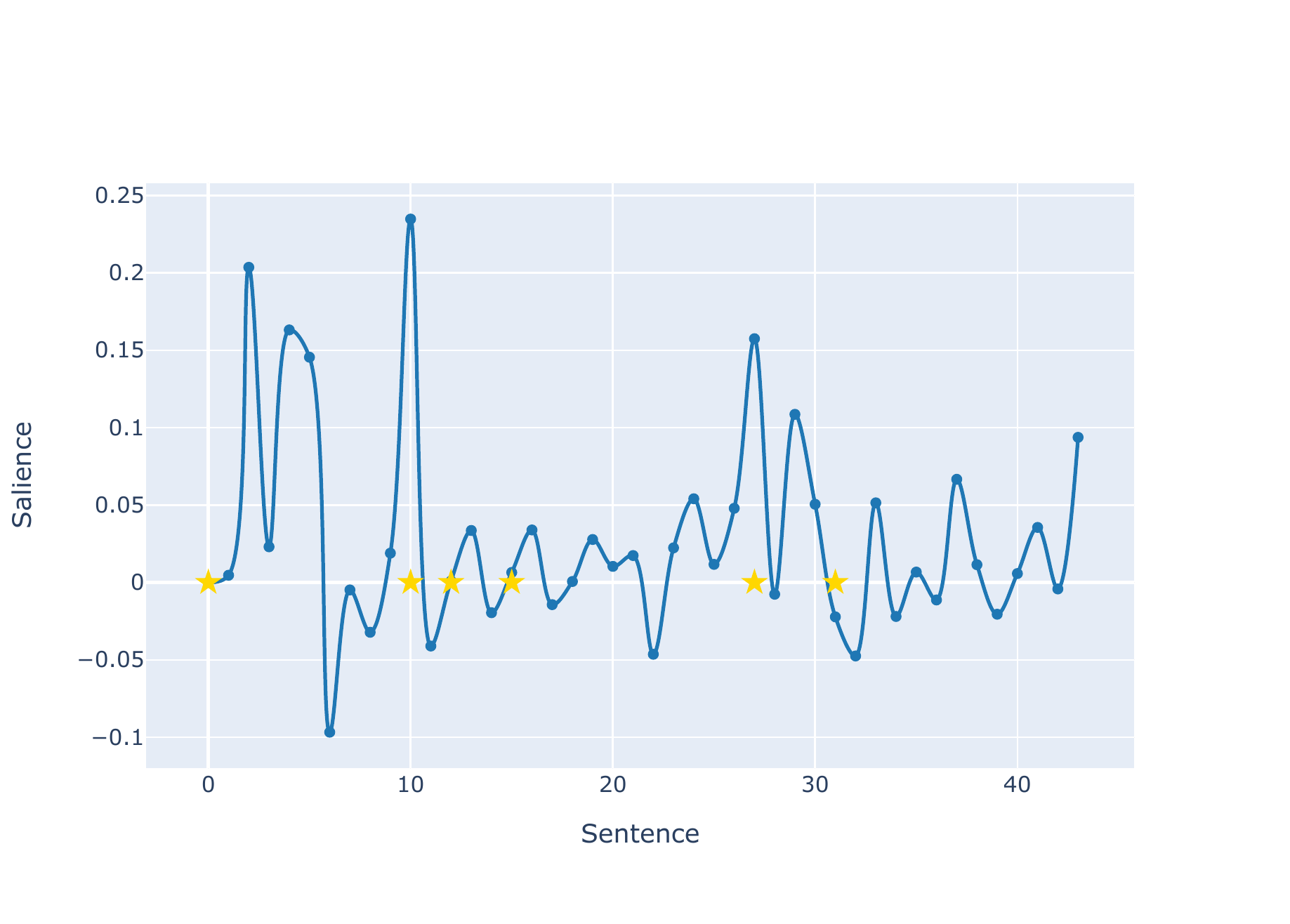}
\caption{The plot of \textit{Macbeth} Act 2, Scene 1 (the fight scene between Macbeth and Macduff, \url{https://www.shmoop.com/study-guides/literature/macbeth/summary/act-2-scene-1}): Only \textbf{Like-Sal} is plotted.}
 \label{fig:macbeth_8_like_sal}
\end{figure}

As per the Propp results, the salience can be plotted. For a single example of \textit{Macbeth}, Figure \ref{fig:macbeth_all} show the same metrics as the Propp results, Figure \ref{fig:macbeth_like_sal} shows \textit{Like-Sal} on it's own for clarity, and Figure \ref{fig:macbeth_alternative} shows the some alternative metrics with \textit{Like-Sal} for comparison. The Macbeth example is used for two reasons. First, the alignments are not given for space reasons, but the method works well even between a contemporary commentary and Shakespearean English, as is the RAG LM able to adapt. Second, the example of the final fight scene between Macbeth and Macduff is an example that illustrates how the method performs. In the scene, there is the entry of Macbeth and Macduff; following this the confrontation; then the fight; and finally, the announcement that Macbeth has been killed. Some of the salient events are missed in between, but the main peaks in the \textit{Like-Sal} are when the shifts happen between these main phases, and salience drops with continuing ongoing events in between. The three main peaks with the \textit{Like-Sal} are:

\begin{itemize}
	\item Though Birnam Wood be come to Dunsinane And thou opposed, being of no woman born, Yet I will try the last.
	\item And damned be him that first cries “Hold! Enough!”
<ACTION> They exit fighting. </ACTION> 
	\item Here comes newer comfort. Enter Macduff with Macbeth’s head. MACDUFF: Hail, King! for so thou art. 
\end{itemize}

To give a second example from \textit{Macbeth} in Figure \ref{fig:macbeth_8_like_sal}, though some peaks are missed, the two biggest peaks are when Macbeth is woken from his sleep and asks for his sword and the famous \textit{... Is there a dagger I see before me? ... }. Although Macbeth is the example used, this similarly seems to apply to other works; bigger shifts and dramatic effects seem more likely to be picked up.

\begin{table}[htbp]
\centering
\begin{tabular}{@{}lccccccccc@{}}
\toprule
 \multicolumn{1}{c}{\textbf{Measure}}                    & \multicolumn{3}{c}{\textbf{MAP $\uparrow$}}    & \multicolumn{3}{c}{\textbf{Rouge-L $\uparrow$}} & \multicolumn{3}{c}{\textbf{Recall K $\uparrow$}} \\ \midrule
                                     \multicolumn{1}{l}{}                  & \textbf{Plots} & \textbf{Mem} & \textbf{Pedia} & \textbf{Plots}  & \textbf{Mem} & \textbf{Pedia} & \textbf{Plots}  & \textbf{Mem}  & \textbf{Pedia} \\ \midrule
                Random                                & .178           & .178         & .178           & .250            & .250         & .250           & .132            & .132          & .132           \\
 Ascending                             & .152           & .152         & .152           & .243            & .243         & .243           & .163            & .163          & .163           \\
Descending                            & .207           & .207         & .207           & .180            & .180         & .180           & .109            & .109          & .109           \\
 Clus-Sal                              & .230           & .230         & .230           & .296            & .296         & .296           & .187            & .187          & .187           \\ \midrule

 No-Know-Sal                           & .246           & .246         & .246           & .336            & .336         & .336           & .205            & .205          & .205           \\ \midrule

 Like-Sal                              & .294           & .280         & .288           & .368            & .356         & .359           & .254            & .241          & .243           \\
                                   
                                Like-Imp-Sal                          & .291           & .276         & .287           & .367            & .352         & .369           & .253            & .238         & .251           \\
                                Like-Clus-Sal                         & .291           & .273         & .287           & .355            & .339         & .358           & .245            & .228          & .243           \\
                              Like-Clus-Imp-Sal & .289           & .276         & .285           & .351            & .336         & .355           & .240            & .225          & .251           \\
                                  Know-Diff-Sal                         & .246           & .242         & .243           & .301            & .300         & .306           & .199            & .194          & .200           \\
                                  Swap-Sal                              & .256           & .241         & .252           & .309            & .294         & .313           & .210            & .193          & .210 \\
 Emb-Surp                              & .249           & .196         & .243           & .311            & .315         & .315           & .201            & .245          & .200  \\
 
  Emb-Sal                              & \textbf{.312}           & \textbf{.311}         & \textbf{.309}           & \textbf{.413}            & \textbf{.371}         & \textbf{.419}           & \textbf{.271}            & \textbf{.271}          & .\textbf{272} \\

  \bottomrule
\end{tabular}
\caption{Shmoop results from the silver label evaluations.}
\label{tab:shmoop_results}
\end{table}

The results are presented in Table~\ref{tab:shmoop_results}: There are three metrics reported and chosen because they represent have different qualities: MAP (Mean Average Precision; \citealt{Manning2008IntroductionTI}) was reported in the Propp results and is fairly standard for ranking tasks. MAP is the used because it has become the standard in similar salience tasks on new corpora, see for example \citet{liu-etal-2018-automatic-event} and \citet{jindal-etal-2020-killed}. MAP averages precision over several different $K$ values and is just more robust than a single measure of precision. The gold standard labels are all selected as \textit{salient} by the \textit{Shmoop}. The predictions to evaluate are the most salient $K$ where $K$ is the number of Gold labelled salient sentences.

 Recall $k$ is a compliment to precision orientated MAP. The Recall $K$ record is the number of aligned \textit{Shmoop} salient sentences. Other $K$'s of $1$, $3$, $5$, and $10$ were evaluated but largely following the same pattern of results across models and so are not reported. At per the chapter level, the results at lower $K$s are also noisier. 

The disadvantage of MAP is that it relies on exact matches. ROUGE-L \citep{lin-2004-rouge} is a word overlap metric that is widely employed in summarization evaluation, for example, by \citet{pmlr-v119-zhang20ae}. Two nearby sentences can describe similar events but may not match with MAP or Recall as the model, and \textit{Shmoop} alignment picks different sentences. ROUGE-L will give partial credit for these close matches. The method for applying ROUGE is to take the top $K$ most salient sentences where $K$ is the number \textit{Shmoop} salient sentences. ROUGE-L is then applied to the concatenated text of both as if they were both summaries of the chapter. ROUGE and similar methods (BLEU, METEOR) have long been critiqued for automatic evaluation of summaries \citep{graham-2015-evaluating}. The problem with stories is they rely on matches of token n-grams. N-grams can be different but convey meaning and thus fail to match. Or the opposite where they match, but the sentences have very different meanings from each other.

Because of the problems with ROUGE and other similar measures methods such as BERTScore \citep{DBLP:conf/iclr/ZhangKWWA20} have become popular recently. The concept is similar to ROUGE, but rather compare token n-grams, the model is comparing semantic vector space embeddings. This should help with the n-gram matching problem and better encapsulate the meaning of the sentences and not just word overlap. BERTScore was trialled. It is not reported because the \textit{Shmoop} alignment method for identifying salient sentences uses a similar vector matching. A second level of looser vector matches between the \textit{Shmoop} and the predicted \textit{salient} sentences may artificially inflate performance. The reported evaluation metrics and those trialled and not reported are all different but have a similar pattern of results across models and prediction metrics, which adds overall confidence to the results.

The \textit{Clus-Sal} baseline measure improves on all the other baselines but only by a comparatively small margin with the best of each, by $0.03$ compared with the best MAP baseline, $0.04$ with Rouge-L, and $0.02$ with recall. The baseline is a centroid based extractive summarisation model that uses a powerful transformer; the relatively small performance improvement increase shows that the task is challenging.

The main \textit{Like-Sal} measure shows an improvement of around $0.05$ over \textit{Clus-Sal}, and  $0.10$--$0.15$ over the baseline. This is a reasonable improvement given the model is unsupervised. The \textit{No-Know-Sal} (without memory and KB) is about $0.03$--$0.04$ worse on MAP and recall, which indicates that the RAG enhancements are helping improve \textit{salience} detection. The theoretical reason would be that BCF detects shifts in state and the informed model with the KB and memory is more likely to predict more obvious events. So salient events are more likely to be significant plot shifts. The biggest finding is that \textit{salience} based on the embedding, \textit{Emb-Sal}, is the strongest measure. This shows the merit of using the BART model more flexibly as a general-purpose sentence encoding model. The \textit{Emb-Surp} measure is a slight improvement on the baselines, indicating that it is mainly the BCF method that causes an improvement in \textit{salience} detection, rather than a simple measure of how much the story changes from sentence to sentence.

One difference from the Otake et al. finding is that combining the \textit{Clus} measures makes little difference. Neither do the \textit{Imp} measures that use absolute sentiment score. While worth exploring further, this is consistent with \citet{wilmot-keller-2020-modelling} findings when adjusting sentiment with inferring \textit{surprise} and \textit{suspense}.

Of the more esoteric measures, both \textit{Swap-Sal} and \textit{Know-Sal} improve on the baseline, although not by much. The more interesting is \textit{Know-Diff-Sal}, which performs similarly to the \textit{Clus-Sal} baseline. The measure as a proxy to exploit the difference between reader and character is quite crude. There may be a more sophisticated way to develop this idea by modelling character knowledge explicitly.

Largely speaking, there does not seem to be much of a difference between the different memory and KB configurations. With the best measure \textit{Emb-Sal}, the results are nearly identical. With the original BCF measure \textit{Like-Sal} and its variants, both the \textit{Wikiplots} dataset (plot summaries from Wikipedia) and the full \textit{Wikipedia} dataset only result in a tiny improvement. It might be expected that a KB would improve performance for salience prediction, but recall that in the perplexity evaluation, memory-only performed better. The present results also suggest that the memory mechanism is the main reason for the improvement over \textit{No-Know-Sal}.

\subsection{Further Analysis}

The memory and KB  access pattern of the model is highly non-linear and references the earlier mentions of the same characters, places, or moods. There are two examples of memory lookup to follow, one from \textit{Great Expectations}, the other from \textit{Kim}. In each example, the context text is given with the three top-ranking memory lookups with their position and probabilities given.\footnote{10 memories are accessed in total, but only the top 3 are shown for brevity.} The first example of this is from \textit{Great Expectations}, final chapter, where \textit{Pip} and \textit{Estella} have their last meeting. The meeting is uneasy, and the memory recalls other periods in the novel where there are tension or quarrels between the pair. The memory focuses on the characters and their relationship rather than many irrelevant details and subplots occurring in between. It illustrates how the episodic memory is acting as an index to similar situations far in the work. The second example is similar in that it describes Kim's service for his long-term master, the Lama, a religious figure who talks about their ongoing quest in the mountains. The recalled memories are earlier of the same quest at an earlier stage, the Lama sending Kim away to School (a sign of their local commitment to each other), and an earlier description of their trips over the mountains together. So once again, the memory is looking up contexts from similar contexts and when interactions between the two characters mentioned. Episodic memory is an aid for understanding longer-form narrative texts. 

\paragraph*{Memory Lookup - Great Expectations (774)}: "You have always held your place in my heart", I answered. And we were silent again until she spoke. "I little thought", said Estella, "that I should take leave of you in taking leave of this spot. I am very glad 
to do so." "Glad to part again, Estella? To me, parting is a painful thing. To me, the remembrance of our last parting has been ever mournful and painful." "But you said to me" returned Estella, very earnestly, "God bless you, God forgive you!" And if you could say that to me then, you will not hesitate to say that to me now,--now, when suffering has been stronger than all other teaching, and has taught me to understand what your heart used to be. I have been bent and broken, but--I hope--into a better shape. Be as considerate and good to me as you were, and tell me we are friends." "We are friends" said I, rising and bending over her, as she rose from the bench.

\label{mem_great}

\paragraph*{Memory One - 497 (0.146):} "Oh! don't be so proud, Estella, and so inflexible." "Calls me proud and inflexible in this breath!" said Estella, opening her hands. "And in his last breath reproached me for stooping to a boor!" "There is no doubt you do," said I, something hurriedly, \"for I have seen you give him looks and smiles this very night, such as you never give to--me." "Do you want me then," said Estella, turning suddenly with a fixed and serious, if not angry, look, "to deceive and entrap you?" "Do you deceive and entrap him, Estella?" "Yes, and many others,--all of them but you. Here is Mrs. Brandley. I'll say no more." And now that I have given the one chapter to the theme that so filled my heart, and so often made it ache and ache again, I pass on unhindered, to the event that had impended over me longer yet; the event that had begun to be prepared for, before I knew that the world held Estella, and in the days when her baby intelligence was receiving its first distortions from Miss Havisham's wasting hands. In the Eastern story, the heavy slab that was to fall on the bed of state in the flush of conquest was slowly wrought out of the quarry, the tunnel for the rope to hold it in its place was slowly carried through the leagues of rock, the slab was slowly raised and fitted in the roof, the rope was rove to it and slowly taken through the miles of hollow to the great iron ring.

\paragraph*{Memory Two - 582 (0.116):} "Why must it be done without his knowledge?" she asked, settling her hands upon her stick, that she might regard me the more attentively. "Because," said I, "I began the service myself, more than two years ago, without his knowledge, and I don't want to be betrayed. Why I fail in my ability to finish it, I cannot explain. It is a part of the secret which is another person's and not mine." She gradually withdrew her eyes from me, and turned them on the fire. After watching it for what appeared in the silence and by the light of the slowly wasting candles to be a long time, she was roused by the collapse of some of the red coals, and looked towards me again--at first, vacantly--then, with a gradually concentrating attention. All this time Estella knitted on. When Miss Havisham had fixed her attention on me, she said, speaking as if there had been no lapse in our dialogue,--  "What else?" "Estella," said I, turning to her now, and trying to command my trembling voice, "you know I love you. You know that I have loved you long and dearly." She raised her eyes to my face, on being thus addressed, and her fingers plied their work, and she looked at me with an unmoved countenance.

\paragraph*{Memory Three - 389 (0.105):} Let us make one more round of the garden, and then go in. Come! You shall not shed tears for my cruelty to-day; you shall be my Page, and give me your shoulder." Her handsome dress had trailed upon the ground. She held it in one hand now, and with the other lightly touched my shoulder as we walked. We walked round the ruined garden twice or thrice more, and it was all in bloom for me. If the green and yellow growth of weed in the chinks of the old wall had been the most precious flowers that ever blew, it could not have been more cherished in my remembrance. There was no discrepancy of years between us to remove her far from me; we were of nearly the same age, though of course the age told for more in her case than in mine; but the air of inaccessibility which her beauty and her manner gave her, tormented me in the midst of my delight, and at the height of the assurance I felt that our patroness had chosen us for one another. Wretched boy! At last we went back into the house, and there I heard, with surprise, that my guardian had come down to see Miss Havisham on business, and would come back to dinner. The old wintry branches of chandeliers in the room where the mouldering table was spread had been lighted while we were out, and Miss Havisham was in her chair and waiting for me. It was like pushing the chair itself back into the past, when we began the old slow circuit round about the ashes of the bridal feast. But, in the funereal room, with that figure of the grave fallen back in the chair fixing its eyes upon her, Estella looked more bright and beautiful than before, and I was under stronger enchantment.

\paragraph*{END EXAMPLE}

\label{mem_kim}
\paragraph*{Memory Lookup - Kim (524):} The lama raises a hand toward the rampart of the Himalayas. 'Not with you, O blessed among all hills, fell the Arrow of Our Lord! And never shall I breathe your airs again!'   'But thou art ten times the stronger man in this good air,' says Kim, for to his wearied soul appeal the well-cropped, kindly Plains. 'Here, or hereabouts, fell the Arrow, yes. We will go very softly, perhaps, a koss a day, for the Search is sure. But the bag weighs heavy.'   'Ay, our Search is sure. I have come out of great temptation.' It was never more than a couple of miles a day now, and Kim's shoulders bore all the weight of it--the burden of an old man, the burden of the heavy food-bag with the locked books, the load of the writings on his heart, and the details of the daily routine. He begged in the dawn, set blankets for the lama's meditation, held the weary head on his lap through the noonday heats, fanning away the flies till his wrists ached, begged again in the evenings, and rubbed the lama's feet, who rewarded him with promise of Freedom--today, tomorrow, or, at furthest, the next day. 'Never was such a chela. I doubt at times whether Ananda more faithfully nursed Our Lord. And thou art a Sahib?

\paragraph*{Memory One - 438 (0.117):} And Kim, as interested in the life of this world as she soon to leave it, squatted with his feet under the hem of his robe, drinking all in, while the lama demolished one after another every theory of body-curing put forward by Hurree Babu. At noon the Babu strapped up his brass-bound drug-box, took his patent-leather shoes of ceremony in one hand, a gay blue-and-white umbrella in the other, and set off northwards to the Doon, where, he said, he was in demand among the lesser kings of those parts. 'We will go in the cool of the evening, chela,' said the lama. 'That doctor, learned in physic and courtesy, affirms that the people among these lower hills are devout, generous, and much in need of a teacher. In a very short time--so says the hakim--we come to cool air and the smell of pines.'   'Ye go to the Hills? And by Kulu road? Oh, thrice happy!'  shrilled the old lady. 'But that I am a little pressed with the care of the homestead I would take palanquin ...  but that would be shameless, and my reputation would be cracked. Ho! Ho! I know the road--every march of the road I know. Ye will find charity throughout--it is not denied to the well-looking.

\paragraph*{Memory Two - 217 (0.105):} So the lama also loved the Friend of all the World?'   'Ay; and he did not tell lies, or return me to captivity.'   'Small wonder the Padre does not know how to unravel the thread. How fast he talks to the Colonel Sahib!' Mahbub Ali chuckled. 'By Allah!' the keen eyes swept the veranda for an Instant--'thy lama has sent what to me looks like a note of hand. I have had some few dealings in hoondis. The Colonel Sahib is looking at it.'   'What good is all this to me?'  said Kim wearily. 'Thou wilt go away, and they will return me to those empty rooms where there is no good place to sleep and where the boys beat me.'   'I do not think that. Have patience, child. All Pathans are not faithless--except in horseflesh.' Five--ten--fifteen minutes passed, Father Victor talking energetically or asking questions which the Colonel answered. 'Now I've told you everything that I know about the boy from beginnin to end; and it's a blessed relief to me. Did ye ever hear the like?'   'At any rate, the old man has sent the money.

\paragraph*{Memory Three - 217 (0.104):} Villages have almost come to blows over the honour of bearing it, for not only has the lama given them blessings, but his disciple good money--full one-third Sahibs' prices. Twelve miles a day has the dooli travelled, as the greasy, rubbed pole-ends show, and by roads that few Sahibs use. Over the Nilang Pass in storm when the driven snow-dust filled every fold of the impassive lama's drapery; between the black horns of Raieng where they heard the whistle of the wild goats through the clouds; pitching and strained on the shale below; hard-held between shoulder and clenched jaw when they rounded the hideous curves of the Cut Road under Bhagirati; swinging and creaking to the steady jog-trot of the descent into the Valley of the Waters; pressed along the steamy levels of that locked valley; up, up and out again, to meet the roaring gusts off Kedarnath; set down of mid-days in the dun gloom of kindly oak-forests; passed from village to village in dawn-chill, when even devotees may be forgiven for swearing at impatient holy men; or by torchlight, when the least fearful think of ghosts--the dooli has reached her last stage.The little hill-folk sweat in the modified heat of the lower Siwaliks, and gather round the priests for their blessing and their wage. 'Ye have acquired merit,' says the lama. 'Merit greater than your knowing. And ye will return to the Hills,' he sighs. 'Surely. The high Hills as soon as may be.' The bearer rubs his shoulder, drinks water, spits it out again, and readjusts his grass sandal. Kim--his face is drawn and tired--pays very small silver from his belt, heaves out the food-bag, crams an oilskin packet--they are holy writings--into his bosom, and helps the lama to his feet. The peace has come again into the old man's eyes, and he does not look for the hills to fall down and crush him as he did that terrible night when they were delayed by the flooded river. The men pick up the dooli and swing out of sight between the scrub clumps

\textbf{END EXAMPLE}

The second part of the analysis is on text generation. The extended RAG model is like any language model and can also generate text. An initial intention was to generate stories and perform a secondary evaluation as a story generation system. The idea would be that the external KB, Wikiplots, would represent fragments and plots and act as a case-based planning system. The work was not completed because of time constraints and the evaluation difficulty. However, code for story generation has been created for the RAG model, both for writing a long story from a prompt and for writing alternative continuations. The idea behind alternative continuations is that generating a passage to fit into a particular place in the story is a test of comprehension of previous events and actions. It is extremely difficult to evaluate for a long work as the evaluator needs to know and comprehend the story to tell if the generated text is plausible. However, examples of generated text are still useful for diagnosing the strengths and weaknesses of the model.

The examples are as follows. The first is the direct continuation from the \textit{Great Expectations} end given in the memory example. For each example the context is given, with the real continuation and a generated continuation. Highlighted in bold are mistakes.

\paragraph*{Great Expectations Example:}

\paragraph*{Context:} ... \say{We are friends,} said I, rising and bending over her, as she rose from the bench. \say{And will continue friends apart,} said Estella.

\paragraph*{Real Continuation:} I took her hand in mine, and we went out of the ruined place; and, as the morning mists had risen long ago when I first left the forge, so the evening mists were rising now, and in all the broad expanse of tranquil light they showed to me, I saw no shadow of another parting from her.

\paragraph*{Generated:} “I have had so much suffering myself, that I feel as if my heart has not had what it deserves, and my heart has been broken into pieces, so that it has not recovered, but has been twisted, broken, and torn apart.” “And you, dear Estella, have had a good deal more suffering than I have,” I said, as I stood over her. “It has taught me, dear Estelle, to be very good to others.” She seemed to remember that she had said that to me once, and gave an answer which I had not heard ...

\paragraph*{Rocky Example:}

\paragraph*{Context:} Mickey nods to Benny. He secretly places a small ring knife between his index and middle finger. In one smooth movement he drags the razor over the blood welts and quickly covers the draining wounds with gauze... It is done so quickly no one is ever aware of the operation. ANOTHER ANGLE At the rear of the arena Adrian looks transfixed at the ring. She is caught up in the heat of the battle. ANOTHER ANGLE The BELL RINGS for round fourteen. Rocky bores in close, but Apollo still has spring in his legs. He seems determined to end it this round... Apollo catches Rocky flush on the jaw. Rocky is staggered. Like a wolf, Apollo cuts loose with pure savagery.

\paragraph*{Real Continuation:} Rocky is driven against the ropes and receives a devastating beating from the Champion... Rocky is dropped. Rocky sits stunned in the middle of the ring... Everything is a distortion. He looks for familiar faces as though to ask for help... Mickey and Mike scream frantically for him to stay down. REFEREE Six -- Seven -- Eight -- Rocky gets to his feet and tenses with renewed energy. He is like a wounded wild animal. The tide suddenly turns. Rocky drops low and catches Apollo with a pair of terrific body punches that seem to drive Apollo's diaphragm up to his throat... A CRACK is HEARD. A glaze of pain covers Apollo's eyes.

\paragraph*{Generated:} Rocky is thrown backwards into the ropes. Rocky is out of the fight. AN INTERCUT (TO ADRIAN)... She is kneeling at the ringside. ANOTHER INTERCUT Rocky is trying to crawl out from under the ropes but his back is to the ropes. Apollo leaps out to catch Rocky with his right hand and slams him against the ropes. ROCKY (bleeding profusely) is gone. \textbf{ANOTHER INTERCEDIT (TO ADRIAN)...She is lying in the ring...} Rocky's body is limp. His head is on the ropes. ANOTHER ANGLE The crowd is transfixed ...

Please see further referred to generation examples for \textit{Pretty Woman} and \textit{Gone Girl} in the appendix, Section \ref{sec:genexamples}.

Overall the examples show the model can stylistically generate excellent text. The \textit{Gatsby} and \textit{Great Expectations} examples particularly fit well. The events and characters generated are grounded in the story. Circumstances not mentioned in the context are referred to from being looked up in memory. There are weaknesses: For example, the \textit{Pretty Woman} example says they are waiting for police when they aren't. Probably because of another mention of police earlier in the film. The \textit{Gone Girl} example says they have a baby when they don't, although possibly having a baby is mentioned earlier in the film. In \textit{Rocky}, \textit{Adrian} is described as lying in the ring when she is clearly watching. In the \textit{Gatsby} example, while the style fits perfectly, it appears as if he wakes up twice and is a confusing read.
The errors that are relatively common are the common sense reasoning errors highlighted in the ML background common to language models or confusion as to how events relate causally. It seems the model can look up topics that have happened but not necessarily relations between them. In another example from testing generation code with \textit{Gatsby}, the generated text says he died in a car accident when he was injured and killed someone else. The generation examples with the earlier results are relevant for future work to improve the model discussed in the conclusion.

\section{Conclusion}

The main overall finding is that the BCF method can infer \textit{salience} over and above baselines with an improvement on much longer works. Augmenting an LM with memory and an external KB can improve the detection of salience and increase the predictive power of the LM on narrative text. The vector-based version of the concept can perform slightly better than using the log-likelihood from an LM. Therefore, this paper demonstrates that it is feasible to run an unsupervised method on novels from Dickens or plays by Shakespeare and correlate with an automated silver label benchmark. Nevertheless, the MAP results are around $0.3$, and ROUGE-L is $0.4$, which leaves room for improvement. There are three areas for further work that build on the work of the chapter: Further applications for the existing model; improvements to the model;  and narratology related developments.

On further applications, as has been noted, code was already developed for text generation but not evaluated. A text generation model with episodic memory has a different use case from most story generation work. Most story generation work focuses on continuing from a prompt on conditioning on latent topics such as Horror or Crime. The examples show that a model that can store a much longer work in memory can also generate alternatives for any point in a much longer story. As discussed in relation to the TD-VAE generation work, human evaluation is challenging, see \citet{karpinska-etal-2021-perils} and \citet{clark-etal-2021-thats}. The best approach would be an interactive one like in \citet{clark-smith-2021-choose}, but with the additional requirement that generated text would start after a long prompt of multiple chapters, and evaluators would need to be familiar with the book. The unlikelihood and entmax methods that made little difference to salience results from observing generated outputs seem to make a more considerable difference in generation and could also be evaluated in further work.

A second application that was experimented with but not evaluated was extractive summarization. The BCF salient method, as shown by the plots, can rank all sentences for salience. An experimental method was implemented in the code for abridging text\footnote{The code is defined in \url{https://github.com/dwlmt/story-fragments/} in the \textbf{ABRIDGE} env variable parameters within the BCF predictor.}. The method runs the BCF method over individual sentences (or longer sliding windows). The least salient $x$ percentage is removed, and the process is run iteratively to shorten the text. Iterative running seems to improve the performance over just filtering on the most salient $x$ as there are changing dependencies in the remaining text as non-salient sentences are removed. Some examples of 10\% summaries of long works are pretty readable in abridged form so that further evaluation would be worthwhile. Some examples are given in the appendix in Section \ref{sec:abridgeexamples}. As per the model's rationale, it also has the benefit of being a completely unsupervised approach.

For improving the model, there are several directions for future work. There are simple improvements, such as the inference only looks $128$ word pieces ahead; the length was chosen because it is the BART large model label size. Simply extending the inference length via iteration may improve the BCF performance. However, the central gap identified by the generation example is on long-term causality, and so-called commonsense, such as that people do not wake up twice from the same sleep. One approach discussed earlier with story planning approaches such as extracting simplified events, and coreferences may also help with BCF. The simplified event structure could aid in disentangling causal relations in the model and improve inference. As the rationale is the same as earlier, it won't be discussed further. There are other approaches to improving model performance.

One problem with the extended RAG model is that no temporal or structural relations exist between any memories or KB entries retrieved; they are weighted for relevance only. It is a problem, though, when thinking about causality. The story has no idea of which order earlier events happened in or, with external KB elements, how they relate to the text. It can lead to the model confusing the causal chain of events. For example,  \textit{Gatsby} is in a car accident and then later is murdered by someone else; just retrieving the relevant passages in isolation can lead the model to assume \textit{Gatsby} was killed in the accident. In \textit{Great Expectations}, \textit{Pip} for most of the story thinks his mysterious benefactor is \textit{Miss Haversham} but later finds out it isn't. Both memories are retrieved later, leading to ambiguity even after it is known. The same applies in \textit{Pretty Woman} when the police are expected to be involved because of much earlier erroneous memories, or in \textit{Gone Girl} where the model can still generate text saying \textit{Amy} is missing after she has returned.

Transformers inherently don't have positional information; they are positionally invariant. It is only with the positional embeddings that transformers can learn the relative or absolute position of the tokens they attend to. \citet{wang-chen-2020-position} find that sinusoidal embeddings can perfectly encode absolute position and relative position with a relatively small error. Sinusoidal embeddings are based on the varying periodicity of the \textit{sin} and \textit{cos} functions. One approach to introducing temporality is to record positionality embeddings alongside the passage's text and key embedding. The positional embedding can then be fused into the passage embedding when conditioned on. The periodicity can be adjusted to sentences or extended paragraphs so the method would scale to longer works. It should allow the model to learn to condition the order and the relative recency of passages retrieved from memory without explicitly building a causal chain of events. Another comparatively simple approach with time-aware language models \citep{DBLP:journals/corr/abs-2106-15110} is to prepend a valid date range into the KB passage proves surprisingly effective. The method is aimed at factual question answering such as \textit{Who did Cristiano Ronaldo play for in 2011?} but in principle, it could apply to an episodic memory as well as a KB.

Another approach to improving causality is multi-hop reasoning. Loosely defined, multi-hop reasoning involves tasks where the answer must be from multiple places in a text or sources, and some level of reasoning is required across both. For example, from HotpotQA \citep{yang-etal-2018-hotpotqa}, \textit{Are Chumbawamba and Spin Doctors from the same country?} Two separate Wikipedia passages are required to know the first is a British band and the second is from the US. The task has often been used for question answering and reasoning over explicit graph structures from facts extracted from a source such as Wikipedia, see for example \citet{lin-etal-2018-multi}. There are other applications beyond direct question answering, for example, \citet{jhamtani-clark-2020-learning} develop a model for explanations that attempts to answer questions such as \textit{what causes forest fires?}, and the answer is static electricity causes sparks, and sparks cause forest fires. Within the realm of question answering \citet{saxena-etal-2020-improving} use graph embeddings over a structured graph knowledgebase and \citet{saxena-etal-2021-question} extend this to temporal embeddings for reasoning over questions that require temporal comprehension. \textsc{CLUE} \citep{arabshahi-etal-2021-conversational} propose a multi-hop prover based on commonsense reasoning datasets to support task-orientated conversational agents. Explicit commonsense reasoning structured have been proposed in story planning, including \textsc{C2PO} \citep{Ammanabrolu_Cheung_Broniec_Riedl_2021}, CAST \citep{DBLP:journals/corr/abs-2105-01311}, and by \citet{martin2021thesis}. These methods, roughly speaking, build structured graphs of the causes and effects of commonsense actions with some reasoning such as pre and post conditions to keep consistent state and plausibility.

Clearly, multi-hop reasoning is fundamental to story comprehension. It is up to BART to piece together causality from the most relevant passages in the current architecture. There is also the weakness in that the memory and KB will only retrieve the most relevant passages to the current context. It will not look up chains of events that might be relevant. For example, when \textit{Pip} finds out \textit{Magwitch} is the benefactor, it may look up the latest communication with his lawyer but not link the sequence of events where he was transported to Australia. He was the escaped convict at the beginning. He is grateful because \textit{Pip} helped him get food and whittles. This requires retrieving a sequence of memories from each context, not a one-off lookup. The same could be said for knowledge from external sources where maybe say Britain's transportation to Australia policy is relevant. One approach is a more structured graph structure with explicit reasoning as per many of the mentioned multi-hop approaches; the architectures are however highly complex pipelines and usually require extensive pre-processing. An alternative more in the spirit of the current architecture is by \citet{yuan-etal-2019-multi}; the dialogue models recursively retrieves previous chat utterances and then fuses together the memories over several hops. MDR \citep{DBLP:conf/iclr/XiongLIDLWMY0KO21} does multi-hop QA via iterative dense vector retrieval similar to RAG and ranking of the strength of retrievals. Though the domain is different, as per RAG a similar method is adaptable to storytelling. It would allow multi-hops of passages from both the memory to be retrieved, and separately attended to. A loss can be engineered so that predictions in the context can train the multi-hop retrieval, reranking and filtering components as part of improving the predictive power of the model.

Lastly, this chapter includes looking at some possible narratology related future work. The \textit{know-sal} is a crude attempt to model some of the ideas expressed in Chapter \ref{chap:backgroundtheory} that when reading a story, the readers try to put themselves in the position of the characters and imagine how they think about things. The idea behind knowledge salience is that if the average log-likelihood difference was large between the naive and informed reader, it would indicate that the event was predictable for the story as a whole but not locally and so relied on previous plot events, and hence more likely to be salient. A better model would be one that explicitly tracks the key characters and their state. The model required would require that the reader have their own model of the story and have a similar model for each of the primary protagonists. It would then be able to infer via different events that are expected by one character and not the other, or expected by the reader but not the character. It is a concept central to some models of suspense. A relevant recent work is \citet{sims-bamman-2020-measuring} that explicitly tracks whole information propagation through a narrative; unlike the models in the thesis though it relies on the model of the entire story being present and not incremental reading. There are many alternative ways per character tracking could be modelled. Most relevant to the work in the chapter, though, would be to actively track coreferences as part of reading and indexing when specific characters or events happened. As part of the multihop reasoning process, per character models could then be restricted to retrieve only memories of when a character is present, or relevant KB entries to their situation. 

\chapter{Conclusion}

\label{chap:conclusion}

\section{Introduction}

This thesis started by introducing the importance of storytelling within our cultures. Of central interest is that stories are not just sequences of everyday events  — someone making a cup of tea or inflating a bicycle tyre is a sequence of events but doesn't have a plot that makes them compelling stories. Though understanding sequences of events is a prerequisite for comprehension of stories. As \citet{forster1985aspects} said of a story: \textit{It can only have one merit: that of making the audience want to know what happens next. And conversely, it can only have one fault: that of making the audience not want to know what happens next.} As was reviewed in Chapter \ref{chap:backgroundtheory}, concepts such as suspense and surprise are essential aspects of an exciting plot, and there are various theories of both. As per the theory implemented in this thesis from \citet{ely2015suspense}, the concepts also have more widespread applications in games, sports, financial markets, auctions, political elections, etc. Within the structural analysis of plots, there are theories such as \citet{Barthes1966AnIT} or \citet{chatman1980story} that decompose events into their structural roles. Top-down theories of plot development that started with work by \citet{freytag1894freytag} and was touched on with the turning points evaluation. Later, more cognitive narratology theories that are more about human limitations and processes in watching or reading that were reviewed in Chapter \ref{chap:backgroundtheory}. While there has been much interest particularly in the structural theories with symbolic AI systems as reviewed in Chapter \ref{chap:backgroundml}, there has been negligible work with more recent neural network architectures and models. Particularly with more powerful neural language models. Much of the work on the state of the art ML systems has been focused on narrow improvement on specific task performance and leaderboards. The goal of the thesis was to take advances in unsupervised LM methods, combine them with narrative theory, and test if they could meaningfully infer higher-level concepts within stories.

\section{Contributions}

The thesis has made several contributions to pursue these goals. The first contribution is annotating a small corpus of short stories to evaluate suspense on a per sentence basis. Evaluated on the corpus is a hierarchical rollout model, an unsupervised model built on a GPT base with a sentence encoder and an RNN story context encoder. The model is trained unsupervised on only story data. Several theories for inference of surprise and suspense from the literature were adapted from \citet{hale2006uncertainty} and \citet{ely2015suspense}. The theories rely on two qualities of the model: The first is the ability to infer a probability distribution over possible continuations. The second is to have the latent sentence representation state represent the outcome or the situation in the story. This is a novelty from prior work as previous studies were in simple game domains where the state is defined only by the final output of the games over all possible states. The approximation of using the latent semantic state to represent alternative outcomes and a GPT language model to generate plausible continuations make implementing the Ely theory feasible, whereas otherwise, it would be intractable. A combination of both Hale and Ely versions of surprise and suspense were evaluated, with the $\alpha$ extension to weight importance. 

The results supported the underlying hypothesis that the combination of the theory and the unsupervised model does correlate with human judgements, whereas the baselines were barely better than random. Furthermore, the Ely models for both surprise and suspense that incorporate both latent state and probabilities perform better than the Hale probability only models, demonstrating that the latent state can be a useful proxy for story state. In a secondary analysis on movie turning points, the model though inferior to a supervised method, was better than baselines. The overall contribution of the hierarchical rollout model is to demonstrate that recent advances in unsupervised methods possess the semantic qualities to be able to infer high-level concepts in stories, subject to limitations reviewed in the next section. The contribution is that rules-based and procedural methods can be synthesized via approximations with these models successfully. The hybrid can be a competitive alternative to relying on supervised labels for domain-specific tasks. 

The second approach to modelling surprise and suspense is with the TD-VAE model. The method is to sample directly over future continuations in the latent space. It converts the Ely process to one that can operate via sampling entirely in the latent space rather than generating concrete sentences. It is also in Bayesian terms a better approximation of the Ely method as the TD-VAE is a variational Bayesian method. The model was also novel at the time of the completion of the work within the domain. Certainly, within NLP, VAE methods had been used for text generation and topic modelling, for example. Outside of the domain, in reinforcement learning and video prediction, similar state-space models have been applied to long-range generative prediction. However, the combination is novel. The results are more mixed. Surprise performs well, but Ely suspense measures are far worse than the hierarchical rollout, although they do still more weakly correlate with human judgements. The performance will be revisited in limitations. However, the model can still be seen as a limited success, especially with surprisal measures. However, a contribution of the thesis is that the latent space approach does seem viable to tasks such as surprise and suspense inference (and other higher order tasks) via future work and improvements to the latent representations and temporal projection mechanisms.  

The second contribution with the TD-VAE model was to apply it as a planning system in text generation. The rationale is to replicate planning systems but in the latent space and not with simplified plots such as keywords or SRL. The TD-VAE model for story generation was evaluated on technical cloze measures. The TD-VAE model performs strongly on the technical cloze and swap tests which test coherence. However, in human evaluation, the reranking version performs similarly to the LSTM reranking model. Conditioning directly on the latent vectors performs worse. The reason for the worsening performance seems to be because conditioning is restricting the story to be less diverse and interesting rather than pushing the plot in different directions. The contribution from this work is that it is an alternative model for a planning system for story generation. The results are not wholly successful. However, it seems as if a latent space representation can be more tailored than in the current architecture for more significant plot events. If engineered correctly, then it should have advantages in implicitly learning the plot progression rather than relying on a fixed apriori representation.

The work on salience made a number of contributions. It was demonstrated that the BCF methods could scale and achieve promising results on longer-form stories with salience. Extensions of the method such as the \textit{Emb-Sal} performed better and demonstrate the method can work in the latent vector space. On the technical side, the episodic memory mechanism and knowledgebase was crucial for the model working on the longer-form stories. The significance is that while RAG-like dense retrieval models have recently become commonplace in factual question answering, they were not widely applied for more general comprehension tasks. As was argued in the rationale for the model within cognitive narratology, indexing of events, characters, motivations, consequences, as well as mood and less concrete cues, has been seen as essential to human story comprehension. It is evident from the results that the same models finetuned do work as expected. Being able to externalise representations allows better performance to be achieved with a smaller model. Relying on external memory and knowledge also makes models more adaptable to new domains and tasks, which is revisited in future work.

The thesis's core is to integrate unsupervised models with narrative theory. The contribution as a whole is that there are positive correlations across a range of theories and models supporting the original thesis. The work shows how narrative theories can be applied with softer latent semantic representations. Unsupervised methods are continuing to improve and develop. The second related idea introduced in Chapter \ref{chap:backgroundml} is that of multitask training where a model can be trained on many tasks at once, such as NER, SRL, coreferences, entailment, semantic similarity, etc. It is more secondary in the thesis, but there are examples of it in the training of the TD-VAE model sentence encoder on entailment and the ATOMIC commonsense reasoning. Both unsupervised methods and multitask supervised methods can have similar overall effects in that they can train rich latent representations that improve performance across tasks and can transfer across domains. The theory applied in the thesis is only a tiny fraction of what is in narratology. Lots of domains have similar insights. It may be more promising in many applications to add heuristics or rule-based reasoning through neuro-symbolic methods than blindly create a set of labels for a new task.

\section{Limitations}

The first limitation is the moderate performance of the results and limitations in the evaluation set. For the first set of suspense results, the hierarchical model performed well, while the TD-VAE model was more mixed. One limitation of the work is that this thesis has focused on unsupervised methods; as such, there was no evaluation versus a supervised model. It was the intent of the approach, but nevertheless, it is still likely, as is seen by the turning points tasks, that a supervised method would outperform if given enough data. With the suspense annotations themselves, the dataset is relatively small, with $100$ annotations for each of the validation and test set stories. A larger evaluation set would be better. In Chapter \ref{chap:suspenseannotation} one decision taken to try and improve the inter-annotator agreement was to annotate from the central protagonists perspective. It was the only metric recorded for reasons of time and resources. Annotation and comparison from the reader's perspective are more typical in suspense theory and so is lacking from the thesis. It would make a natural extension for further work. The surprise measures are also modelled directly against suspense. The reasonable results of this cross-comparison is expected as there are some overlaps between the concepts. However, separate annotations for reader surprise would be an improvement on the current evaluation. 

In Chapter \ref{chap:tdvaegeneration} the main limitation is the failure of the latent conditioning model to improve over baselines in human evaluation. As noted in the chapter, the main issue is that conditioning restricts the story from developing the plot. In the chapter, several areas were suggested for future work to improve the model that will not be repeated here. The primary finding of interest from the evaluation is that the conditioning can make stories worse while it performs well on technical benchmarks such as the cloze and swapping tasks. It demonstrates the dangers of technical measures as a proxy for story quality. As \citet{guan-etal-2021-openmeva} and \citet{guan-huang-2020-union} find that for story generation, technical metrics have a low correlation with human judgement. Technical measures such as BLEU can score highly if there is a high word overlap, but a high word overlap means the same topics, characters and places are mentioned but does not mean it's a good story. There is a similar problem with cloze, where a continuation can be highly coherent but dull. \citet{Schmid+2017+229+246} argues that stories are \textit{eventful} where there are transgressions of what is normally expected in the narrative world. The conception is what the characters expect in their world and not what the reader expects, having read many similar books. There is also \textit{repetitiveness} where stories can often follow predefined scripts, and hence the reader knows what to expect. Schmid argues they both occur in various combinations. It is a higher-level concept of a story than most automated systems operate at. Nevertheless, the relevance is that technical systems can fall into the trap of learning what's likely to come next and generating dull and repetitive text. Or alternatively, they could be trained to make bigger dramatic leaps which would likely be disconcerting to a reader and miss the filling in details they would expect to see between more dramatic events. The TD-VAE model is too much of the first. The same issues are also relevant to why suspense modelling is worse with the same model.

Chapter \ref{chap:salience} on \textit{salience} had promising results in that the BCF deletion method outperformed baselines on longer length novels. The method was only a moderate improvement with the best results in the $0.3$ to $0.4$ range for MAP, leaving much room for improvement. Part of the performance relates to evaluation construction. The Shmoop alignment was designed to provide a scalable way to evaluate salience on a larger corpus of longer works but has weaknesses: Inevitably though the alignment method is generally good, there is noise and errors. There are also cases where arguably, there are multiple good matches for the summary. As noted in the chapter, the Shmoop annotations are retrospective, but the reader model is incremental. It creates a conflict between events that may seem important at the time and retrospectively. All of these factors will underestimate performance. The second limitation is the model itself, which as per the future work section of the chapter, is limited by the model's ability to infer events. The implementation of BCF is also only to use the output context of BART, which is $128$ wordpieces. It is a limitation of the work because it makes the salience detection method far more local as the model judges in the context of the next handful of sentences how influential the sentence is. It means it is more of a salience measure within the scene than of the plot as a whole. 

The major limitation of all the separate studies in the thesis is what they are modelling. The theories behind Hale, Ely or BCF require that a model comprehends a typical common sense sequence of developments and those more specific to narrative plots. But for example, the GPT model that generates the concrete continuations is nowhere near producing human quality stories. The whole rationale for planning systems or the more complicated commonsense reasoning or reranking architectures is to improve the performance of base language models. The same can be said of the TD-VAE, where all the generated story models (including the baselines) were judged worse than the Gold labelled human stories. The same with the generation examples illustrated from the salience work where the model could generate text that confused the situation. If the models are worse at predicting the sequence of events in stories, how do they perform well on task evaluation?

As was covered in the background in Chapter \ref{chap:backgroundml}, transformer models can have a tendency to learn surface-level features \citep{ribeiro-etal-2020-beyond} and take statistical shortcuts \citep{geirhos2020shortcut}. The models in the thesis are mainly trained to autoregressively predict the next work token or correct continuation sentence. The models will likely learn from surface-level linguistics cues more than the underlying plot structure. The domain is highly conducive to this working in the comprehension evaluations for suspense and salience. The Ely model of suspense depends on how much the generated or with TD-VAE sampled alternative differs from the existing context.\footnote{The context is not just the current event since the LSTM in the hierarchical rollout or the TD-VAE represents all the state up until that point in the story and expectations over the future state.} Suspenseful events are when there is a higher variance in the expected outcome. Surface-level linguistic cues will likely be highly correlated with more major plot shifts. A lexical shift in topic, mood, the mention of new named entities will cause the model to generate more diverse continuations and increase suspense. This similarly applies to the surprise definition, where surprise will be higher when the surface elements differ from the immediate expectations. To give some examples from the Gold standard training examples from Chapter \ref{chap:suspenseannotation}: In the story of the young girl at the beach, given in Table \ref{tab:train_annotations}, the most suspenseful sentences are when she is suddenly swept off her feet and screams and swallows water. In the \textit{Clancy} example, Table \ref{tab:guide_annotations}, the biggest increase of suspense is when a man jumps into the trench, and they start fighting. In the \textit{Pretty Woman} example, plotted in Figure \ref{fig:tp_train_36} and Figure \ref{fig:turningpoint_pretty_woman}, the climactic turning point is when \textit{Edward} changes his mind about going to the airport and instead visits \textit{Vivian's} apartment. In all these, both dramatic events and the lexicon change together. A model able to infer either or both will thus generate more suspenseful continuations or find the shift more surprising.

The salience work on longer-form texts is similar but on a longer scale. In the longer work, the RAG extended model will retrieve relevant similar plots from \textit{Wikiplots} or general knowledge from \textit{Wikipedia} depending on the configuration. The memory typically references earlier occurrences of the same characters, places, action, or atmosphere. The BCF deletion method picks up where a context sentence is salient because if it is missing, the subsequent text is less predictable. In practice, this means that if a new character, place, situation or action is mentioned, it is nearly always likely to be salient. Similarly, if one of these reoccurs, then the context of previous occurrences will be recalled and anything that shifts from this will be salient. In the same way,  anything that differed from what's typical in the situation retrieved from the KB will be salient. Examples from the salience examples in Chapter \ref{chap:salience}, from the Propp examples the most salient sentences are in the \textit{Nikita} story, plotted in Figure \ref{fig:propp_nikita}, when the \textit{Princess} first writes to \textit{Nikita} asking for his help on the quest. In the \textit{Swan Geese} story, plotted in Figure \ref{appendix:salience}, the most salient sentence is when the talking stove that is introduced. In both cases, the following context sentences are all related to the introduced elements. With \textit{Macbeth}, plotted in Figure \ref{fig:macbeth_all} and Figure \ref{fig:macbeth_alternative}, salience picks up \textit{Is this a dagger which I see before me} where the following speech in the soliloquy is about the knife. In the final battle, there are salient peaks when the quarrel turns into a fight and when \textit{Macbeth's} head is presented as a trophy. The difference in the longer form is that when characters are introduced, they are salient. When the characters reoccur, they are not, and salience depends on the changes of what has previously happened. The same happens in the memory lookup examples for \textit{Great Expectations} (Page \pageref{mem_great}) and \textit{Kim} (Page \pageref{mem_kim}), the salience can either be based on learning events or surface-level cues. The surface-level cues serving as a proxy for a more complex comprehension may be one reason the methods can perform reasonably well despite the known shortcomings of the model.

\begin{figure}[htbp]
\centering
\includegraphics[trim={2cm 1.0cm 1.5cm 1.0cm},clip,width=1.0\textwidth]{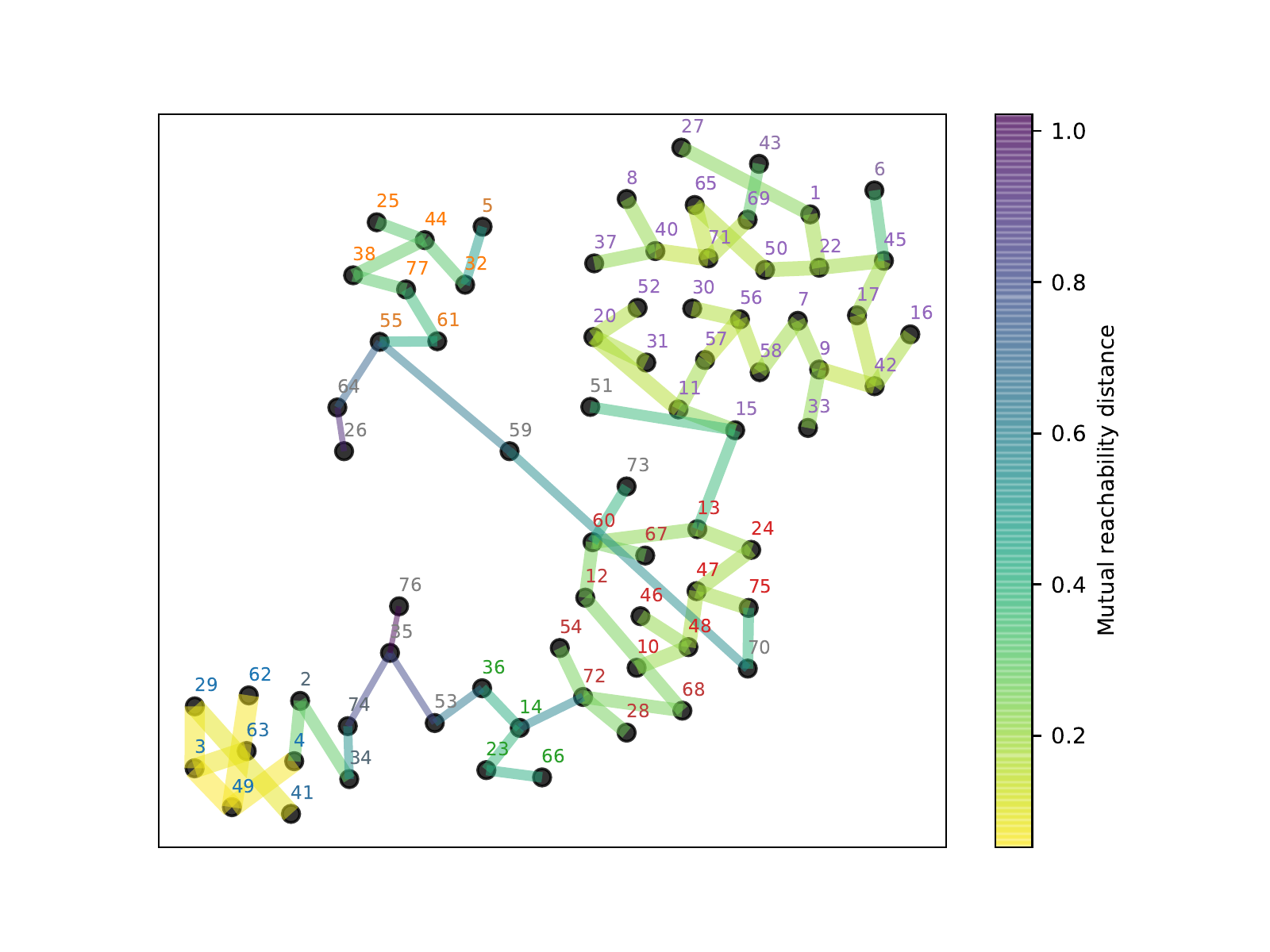}
\caption{Minimum spanning tree example from the Hierarchical Rollout model on \textit{WritingPrompts}. The colours represent the distances between embeddings. The numbers are the sentences and the number colours are HDBSCAN clusters. This is story $48$ from the test set with a dimensionality reduction in UMAP to $48$ and with cosine similarity from the MST and UMAP.}
 \label{fig:mst_example}
\end{figure}

\begin{figure}[htbp]
\centering
\includegraphics[trim={4.0cm 3.5cm 4.0cm 4.0cm},clip,width=1.0\textwidth]{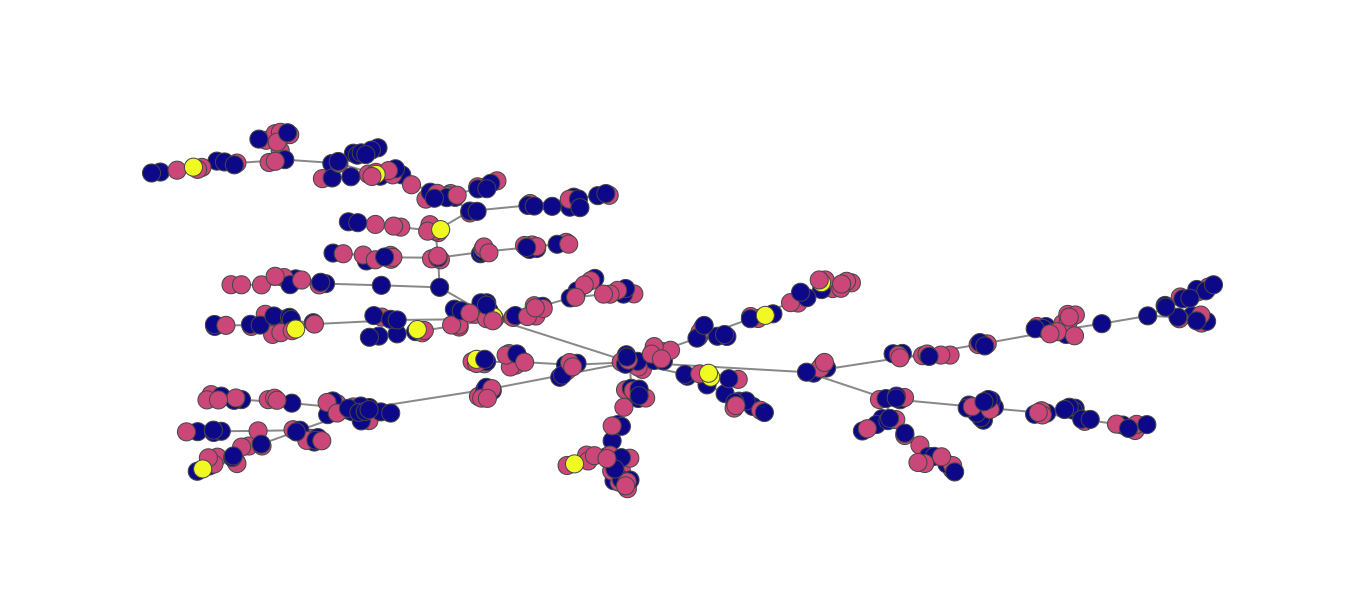}
\caption{Cosine similarity clustering of context passages from the \textit{Great Gatsby}. The colours are HDBSCAN clusters. The code produces interactive examples where the text is shown on mouseover hovers.}
 \label{fig:gatsby_network_plot}
\end{figure}

One of the other central limitations of the study relying on sentence embeddings and predictions in the vector space is that it is hard to interpret what the model is doing. It could be learning surface-level cues, more complicated events, or named entity coreferences. In the TD-VAE chapters, some work was done on the linguistic and semantic properties of the sentence representations. Other analysis techniques have been developed to analyse the qualities of embeddings and internal neural networks. One alternative is clustering algorithms on embeddings. For all the work in the thesis, clustering methods were developed for the sentence representation or, for RAG, the averages of encoder or decoder workpiece embeddings and the question encoder embeddings.\footnote{The code is still available in the respective Github repositories.} The embeddings were clustered directly with both K-means \citep{DBLP:journals/tit/Lloyd82} and HDBSCAN \citep{McInnes2017}, a hierarchical clustering method. Because of the high dimensionality of the embeddings, UMAP \citep{McInnes2018} or PCA \citep{doi:10.1080/14786440109462720} is used to reduce the dimensionality of representations based on L2 and Cosine distances. Unfortunately, although all the clustering produced output, it didn't help with analysis of either with per story clustering or the whole dataset. Figure \ref{fig:mst_example} is an example of a minimum spanning tree trying to show the clusters of sentence embeddings for a single story, and Figure \ref{fig:gatsby_network_plot} is for all the context passages in great expectations. For all of the work, the clustering doesn't seem to produce clean topical groupings for the cluster, possibly because the space is contiguous. With the spanning trees, while similar topics, events and characters occur close to each other as would be expected, it's hard to draw analytical insights beyond that. Similarly, NetworkX \citep{paper:hagberg:2008} was used to try and graph the memory lookup for the RAG models. Various network centrality methods \citep{DBLP:journals/corr/abs-2011-07190} were applied to try and determine the most significant passages. Once again, beyond understanding that the model does seem to index the relevant named entities and situations. A follow up where external models such as SRL, coreference, and NER can be cross-referenced against the lookup or clustering could be beneficial. Another analysis area is methods for introspecting deep learning models. There are gradient methods such as Saliency Maps \citep{DBLP:journals/corr/SimonyanVZ13} or SmoothGrad \citep{DBLP:journals/corr/SmilkovTKVW17}. Various adversarial methods \citep{DBLP:journals/corr/abs-1902-07285} can also be applied to probe models by identifying the types of predictions that have a larger impact on output predictions. AllenNLP Interpret \citep{wallace-etal-2019-allennlp} was experimented with to try and apply these methods, but they didn't produce interpretable results. With any follow-up work, better analysis techniques will be needed to understand the models. Another option could be further structural manipulation as per the TD-VAE cloze and swap tasks or further analysis with multitask benchmarks beyond SentEval.

\section{Future Work}

The theme of the thesis has been to try and model suspense, surprise, and salience with deep learning models as expectations over future outcomes with structural manipulations and rules to infer these measures. Along these lines, there are a number of directions for future research: Improving the core capability of the model, extending the applications, multimodality, developing closer connections with cognitive science, and analytical insights from the models. 

To consider improvements it is worth revisiting the examples given in the introductions and how these relate to the theories discussed and the model's limitations. This well-known quote from \textit{Great Expectations} sums up much of the theory:

\begin{displayquote}
That was a memorable day to me, for it made great changes in me. But it is the same with any life. Imagine one selected day struck out of it, and think how different its course would have been. Pause you who read this, and think for a moment of the long chain of iron or gold, of thorns or flowers, that would never have bound you, but for the formation of the first link on one memorable day. \textit{— Chapter 9, Great Expectations, Charles Dickens} 
\end{displayquote}

The quote sums more poetically the concepts of both Ely suspense and BCF salience in the thesis. There are lots of events that occur in a story. The Ely suspense method relies on identifying the crucial points at which alternative consequential paths can occur. Ely surprise is where the reality of the projected paths differ from what is expected. BCF recognises where these points of divergence are but without directly inferring the outcome paths. The other introductory examples are relevant as well: \textit{Rocky} and \textit{Pretty Woman} were selected because they are stereotypical of the so-called \textit{Hero's Journey} where the overall plot is formulaic and predictable, but the path along it is where the interest lies. The comedy examples from \textit{Only Fools and Horses} and  \textit{Mr Bean} emphasise how dramatic elements can be local to a scene and do not have to develop the plot as a whole.

Each of the specific proposals for improving the three main models in the thesis is discussed in the relevant chapters. This section will focus on commonalities. The examples all require a model to better comprehend commonsense events and particularly plot knowledge. One simple extension could be to combine the models of the thesis so there would be an improved TD-VAE latent state type model with access to episodic memory, and external knowledgebases would perform on longer-works. In the salience work, the \textit{Emb-Sal} method performed best, and the BCF work would be possible to produce via sampling a generative TD-VAE like the model and relying on vector distances and as per Ely surprise. It would also be possible to sample suspense, although it would depend on improving the performance of suspense through architectural improvements first.

The losses employed throughout this thesis are auto-regressive in that the model predicts a future sentence or token or in the form where the correct option is maximised vis-a-vis negative examples. \citet{goldfarb-tarrant-etal-2020-content}, in story generation and \citet{guan-huang-2020-union}, in evaluation, have demonstrated that many different structural manipulations (e.g. reordering, repetition, relative ordering, verb and coreference subsitutions, negation) can improve story generation, which could be incorporated, and more can be developed.

The TD-VAE work tried to integrate common sense reasoning via directly training on the entailment datasets and the ATOMIC text. Learning of plots was implicit or via a case-based \textit{Wikiplots} KB for the RAG work. Obviously, this can be extended simply by adding more datasets as a new training data such as \textsc{GLUCOSE} \citep{mostafazadeh-etal-2020-glucose}, or the extended \textsc{ATOMIC 2020} \citep{Hwang2021COMETATOMIC2O} amongst many other recent ones. Or with the RAG plot approach, adding more relevant text such as character profiles or story analysis via Shmoop type sources. An alternative would be to abandon raw text and adopt a more structured approach to prediction. Each of \citet{DBLP:journals/corr/abs-2105-01311} or C2PO \citep{Ammanabrolu_Cheung_Broniec_Riedl_2021} and \citet{martin2021thesis} are story generation systems that use reasoning over graphs of simplified event structures on common sense or plot scripts. StoRM \citep{DBLP:journals/corr/abs-2112-08596} dynamically builds a reader model consisting of a graph structure of events and linked entities from the story that are joined via inference to common sense graph knolwedgebases such as  \textsc{ATOMIC 2020} and ConceptNet \citep{conceptnet2}.  The change to the existing models would be to split stories and common sense datasets into graph event chains. Inferring in the story is then linked to predictions or inferring the events within the structure. Likewise, common sense knowledge would be learnt over branched chains of events from common sense datasets. Methods like Ely or BCF could be adapted to measure divergence in graph structures or the impact of node deletion. A more structured graph approach could be applied to the external KB, perhaps inspired by the causal chains of \citet{jhamtani-clark-2020-learning} in learning explanations. Or for plot knowledge, a graph structure such as \citet{sims-bamman-2020-measuring} built around chains of coreferences and events could be suitable. The approach is well worth pursuing for future work. However, with common sense reasoning (though not specifically in the story domains), neural network models only models can be competitive to hybrid models with reasoning \citep{DBLP:conf/cikm/BrancoBS021}. Language models with adaptions can learn common sense in generalisable ways \citep{wang-etal-2021-language-models}, albeit with some limitations. As of writing, dense vector models such as \textsc{RETRO} \citep{DBLP:journals/corr/abs-2112-04426} that lookup KB entries via dense vectors and integrate via attention mechanisms are state of the art on many factual domain tasks. So it is not a given that a structured graphical reasoning model would outperform an implicit text input model.

In the future work sections for both the TD-VAE models in Chapter \ref{chap:tdvaegeneration} and the RAG Chapter \ref{chap:salience} the idea of cadence was mentioned. The idea is that some events in a story are just not important, others may be important locally but not within the plot as a whole, while others are crucial for the plot as a whole. It would be infeasible to try and annotate a long novel or screenplay with a hierarchy of importance for every event. It would still be better to have a model that could learn to distinguish between the level of significance or a hierarchy where different levels of the models could focus on separate levels of abstraction. In Chapter \ref{chap:tdvaegeneration} it was noted that a simple clockwork model where each layer viewed a different $n$ of sentences wouldn't work as stories are not contiguous in the way a video or game playing environment might be. The Shmoop method was based on chapter aligned summaries and the full text. The full resource of a site such as Shmoop also contains high-level synopses of the whole work, including the full text. It also contains key quotes that may or may not be part of the summary, character profiles, themes and analysis, quizzes, etc. There are also other commercial competitor study guides and various Wiki or community-maintained resources with many of the same resources. It is fairly standard in an abstractive summarisation system such as \textsc{PEGASUS} \citep{pmlr-v119-zhang20ae} to auto-align summary text with the main text to train the abstractive text generation process. The intent is essentially to train the attention in transformers or where to look in creating the summary. A simple proposal would be to use an alignment process to tag and train a classifier for the different levels in the summarisation hierarchy and possibly orthogonal aspects such as character personas or themes. The tags could then be fed in as features and training targets for events as, for example, a model such as Semantically-Aware BERT \citep{DBLP:conf/aaai/0001WZLZZZ20} does with SRL labels, or GPT2E \citep{stylianou-vlahavas-2020-e} does with named entities and coreferences. A more ambitious alternative is to use the classifier to filter the inputs to different levels of abstraction and then feed only each level of abstraction into a single layer of the architecture as per a Clockwork-VAE \citep{DBLP:journals/corr/abs-2102-09532} type model. The motive is that each layer would specialise in a level of abstraction, and so the most recent important high level, and intermediate plot points would remain in context with lower scene level details. It may also be possible to integrate this with a KB as per the multi-hop suggestions in Chapter \ref{chap:salience} where multi-levels of lookups could be retrieved, combined with filtering and reranking. The obvious downside is that the architecture would be limited to training on data where aligned summaries and other analyses are available or feasible to collect.

All of the models read the text incrementally and try to infer what will happen next. It is a model introduced as the reader as a problem solver from \citet{Gerrig1994ReadersAP}. Any model of the narrative world is implicit to the model. From \citet{Lotman1977TheSO} and \citet{Schmid+2008+17+34} the idea of the character perspective is crucial since what makes narrative events is transgressing expectations the characters have within their world. For example, in \textit{Rocky}, people in the story think of him as a washed-up journeyman who doesn't have much of a chance. In \textit{Pretty Woman}, \textit{Vivian} is not expecting a long-term relationship with \textit{Edward}. In \textit{Great Expectations}, \textit{Pip}, a poor orphaned boy, is not expecting to be in line for a great fortune. In \textit{Gone Girl}, the town doesn't think \textit{Amy}, a respectable middle-class professional, will fake her own death. In suspense theory, \citet{smuts2008desire} credits the difference between reader's expectations and the character's own perceptions as mainly responsible for suspense. The future work section of Chapter \ref{chap:salience} touched on the idea of an explicit character perspective. The idea would be to have more limited alternative views that only represent the experiences of the main protagonists or are limited to only the knowledge in that world (no external KB). The inference process would be to maintain a per major protagonist state, a world state without external knowledgebase, and a richer reader model that integrates external knowledge. Theoretically, the model would be measure differences in expectations between these. Following the RAG work could be perplexity or operate in the vector space. Following in the spirit of this thesis, a simple implementation is to tag and track coreferences and optionally events and propagation graphs as per \citet{sims-bamman-2020-measuring}. A memory mechanism like RAG can then be altered to match dense vectors as now, but only from memories or external KB related to events, the character is aware of. Interactive role-playing story work trained on the LIGHT \citep{urbanek-etal-2019-learning}  corpus by \citet{ammanabrolu-etal-2021-motivate} adopts a more similar implicit approach where character personas, goals, room objects, etc. are retrieved when relevant and implicitly learnt by the model via concatenation or cross attention. In personalised chat, a model by \citet{zhong-etal-2020-towards} integrates personas into chat via co-attention to integrate the response, context and personality of the speaker into generating a response. The model would be similar, except the character models would be personalised from their experiences and via context-relevant retrieval and not fixed. A more graph-based alternative could be inspired by story generation systems such as CAST \citep{DBLP:journals/corr/abs-2105-01311},  StoRM \citep{DBLP:journals/corr/abs-2112-08596} or 
\citet{DBLP:journals/corr/abs-2112-08593} where separate graph structures could be maintained for each character and world model, and the inference would then need to be adapted to measure differences in graph structures. An extension of either work is that \citet{gorinski-lapata-2015-movie} summarises movies by evaluating scenes as being important when major characters co-occur. \citet{Papalampidi_Keller_Lapata_2021} extends the approach to multimodal movie summarisation. Differences between character models and reader or world models don't need to be modelled in a binary way. They could be modelled n-way or differently depending on how many significant characters co-occur in a scene.

A completely separate line of theoretical work could be inspired by \citet{Toubiae2011695118}. The model is to model successive sentences or passages in the vector space. Measurements of the velocity of changes in the vectors space, the total volume of a story in the vectors space, and circuits in the space infer the \textit{shape of the story} which is correlated with human judgement. The work only uses a relatively old model of averaged Word2Vec vectors. The nice thing about the geometrically inspired approach is it can be run on any vector-based representation model and is also unsupervised. Any models from the thesis or proposed extensions could be adapted to the method.

The whole focus of the stories has been on text-only even when the form itself, such as plays or screenplays, are multimodal in the original form. There are also comics, tv series, etc., that are also obviously multimodal stories. The decision was taken to stick to only text as a modality as there are many different challenges with preprocessing, aligning, integrating multiple modalities. A form such as movies can have images from the video stream, the audio track, subtitles, audio descriptions, sometimes alignments to a written screenplay, and links to metadata from sites such as IMDB or analysis, comparable to Shmoop. Similar methods to those applied in the thesis can be applied to multimodality. Multimodality has long been a central concern of machine learning research, see \citet{DBLP:journals/corr/abs-2105-11087} for a recent review and \citet{10.1109/TPAMI.2018.2798607} for an earlier review. Multimodality is a tougher challenge in its own right because of the complexities of handling multimodalities, but also a way of addressing the grounding issues with language discussed in Chapter \ref{chap:backgroundtheory}. The goal of this section is not a systematic review of the literature but to pick out recent trends that support the approach of the thesis for further multimodal work. 

One trend that is not multimodal but for solely vision is a move towards vision transformers \citep{DBLP:journals/corr/abs-2006-03677}, see  \citet{Khan2022TransformersIV} for a review of many recent variants. These were mentioned in the context of future work around discrete VAE models for improvements to the TD-VAE models. Vision transformers apply clustering to produce discrete codes that allow the transformer to work with what would otherwise be continuous representations. Transformers have achieved state of the art results in many computer vision tasks. One intuition behind their success is similar to the motive for improving the TD-VAE model; a discrete representation simplifies the latent space to stereotypical roles, reduces noise and allows the model to learn improved transformations in the simplified latent space.

One of the benefits of having vision transformers adopting a similar architecture to recent NLP models is integrating the two approaches for multimodal models. Recent multimodal models that have attracted much attention include Dall-E \citep{pmlr-v139-ramesh21a} and
CLIP \citep{pmlr-v139-radford21a}. CLIP is a contrastively trained model with a vision and language encode that maximises the similarity of co-occurring text and images vis a vis negatively sampled examples. It jointly trains a semantic space where images and their description occur at similar locations and thus allows a two-way search trained in an unsupervised way. Dall-E is an encoder/decoder model where a vision transformer acts as the encoder and GPT-3 as the decoder and allows descriptions to be generated from images. As per a VAE, the models have a shared intermediate latent space. There is much recent work in this area: UNiT \citep{DBLP:journals/corr/abs-2102-10772} is a multimodal and multitask that jointly trains transformer models via concatenation of image features. UNITER \citep{DBLP:conf/eccv/ChenLYK0G0020} jointly trains a joint embedding space across multimodal datasets with masked image and text training.\footnote{UNITER users a masked CNN for the image encoder and not a vision transformer.} UNIMO \citep{li-etal-2021-unimo} relies on contrastive learning but separately trains on unimodal text and image-only datasets as well as multimodal datasets; it means the model can be trained on a much wider range of corpora, improving overall performance. SLIP \citep{DBLP:journals/corr/abs-2112-12750} combines CLIP with multitask self-supervised training. \citet{pmlr-v139-cho21a} integrate vision and language encoders to a single model, which composed into an encoder/decoder architecture can generate either or both modalities. SimVLM \citep{DBLP:journals/corr/abs-2108-10904} adopts a weak supervision approach to allow training on progressively more noisy examples. There is more structured work such as \citet{zhong2021SGGfromNLS} who train on graph-based scene descriptions. Most of the cited papers with self-supervision are trained with contrastive (or negative sampling) losses as the sentences encoded in this thesis have been. In vision, \textsc{SEER} \citep{DBLP:journals/corr/abs-2103-01988} performs self-supervision via training the model to cluster different views of the same images in the same cluster.\footnote{\textsc{SEER} is a convolutional neural network model, and not a transformer.} \citet{DBLP:journals/corr/abs-2202-08360} finds that such methods, that parallel the self-supervised LMs such as GPT or BART, can be as accurate as supervised models, but can also be more robust than supervised models for shifts in domain and to unseen examples. 

The self-supervision paradigm has been extended further to other kinds of data: Work by \citet{DBLP:conf/nips/GrillSATRBDPGAP20} and data2vec \citep{DBLP:journals/corr/abs-2202-03555} create latent representations via filling-in masks to new types of data. The innovation is to have a student model that recreates a masked input from a teacher model that is a moving average of the student. The process effectively makes a masked BERT missing token loss applicable to any type of data and an alternative to contrastive losses. 

The area of multimodal models is advancing rapidly with new models combining ideas from the previous generation of models while improving the architecture. For example, Flamingo \citep{DBLP:journals/corr/abs-2204-14198} is a multimodal transformer that uses an existing frozen large LM and frozen image encoders. Rather than finetune the whole model, small samplers for the vision model are combined with an interleaving adapter to perform multimodal tasks such as question answering or captioning by retraining only a tiny number of weights compared to the source pre-trained models. \footnote{Later on in future work, using adapters rather than finetuning for new domains is discussed.} OFA \citep{DBLP:journals/corr/abs-2202-03052} is a multimodal sequence-to-sequence transformer that combines multiple multimodal sequence-to-sequence tasks (e.g. captioning, question answering, image generation) with self-supervision masking of image and text tokens. OFA achieves several state-of-the-art benchmarks which much fewer data than previous models. COCA \citep{DBLP:journals/corr/abs-2205-01917} combines a hybrid image to text sequence to sequence transformer architecture with contrastive learning as per CLIP as achieves better results on both kinds of tasks. mPlug \citep{DBLP:journals/corr/abs-2205-12005} has a more efficient skipping attention for for combining modalities. Cogview \citep{DBLP:journals/corr/abs-2204-14217,DBLP:conf/nips/DingYHZZYLZSYT21} adopt a hierarchical multimodal transformer architecture to hugely improve image generation from captions. All these models support the approach advocated in Chapter \ref{chap:backgroundml} that multitask learning, and self-supervised learning can improve performance across tasks. The latest generation of models requires far less data to achieve better performance than even recent models such as Dall-E and CLIP. It makes the possibility of applying methods like those in this thesis to multimodal stories much more feasible soon.

On the application side, apart from more standard tasks, there have been a variety of interesting applications: Understanding representations differ across languages and culture \citep{liu-etal-2021-visually}; recognising rare and unseen object classes in BALLAD \citep{DBLP:journals/corr/abs-2111-14745}; multimodal video representations in MDMMT \citep{Dzabraev2021MDMMTMM}; multimodal question answering \citep{Yuan2021AdversarialMN}; multimodal open domain chat and dialogue \citep{shuster-etal-2021-multi}; or multimodal neural script learning and temporal reasoning with MERLOT \citep{DBLP:journals/corr/abs-2201-02639}.

Although there are differences in the architecture and domains for the cited models, they all follow the same trends as the justification for the thesis approach in Chapter \ref{chap:backgroundml}. Whether trained through separate unimodal or multimodal datasets, contrastive or masked losses, combining multimodal features through early or later fusion; the trend is towards self-supervised (or unsupervised) methods to learn powerful latent state representations. There is also a trend towards multimodal multitask learning where the same model trained on lots of tasks outperforms task-specific models and generalises better to new tasks and domains. The two primary models of the thesis are firstly to project forward possible alternatives in a lot either through concrete continuations or directly in the latent space. Or with the RAG methods, episodic memory and knowledgebase retrieving the most relevant details with dense vector representations. Either one is suitable to be adapted to multimodal models like those discussed that learn a single joint representation in the vector space, or maintains them separately and integrates via \textit{late fusion}. With movies, there would be expected to be similar surface-level cues in the non-dialogue features, such as the music changing when there is a shift in the mood or more rapid transitions in the shots. The surface cues would, of course, be picked up alongside a deeper understanding of objects and relations. Recall, from a multimodal episodic memory or KB of other cultural references or works is likely to be just as beneficial as in the unimodal domain. Pretrained multimodal models can be finetuned to reuse the representations captured from training on a diverse set of tasks and datasets. The finetuning would need to be engineered to induce particular plot elements, entities, visual or audio cues that are most relevant for the downstream. Likewise, there is a similar opportunity to apply theories and rules via structural manipulations and heuristics with multimodality rather than labelling a task-specific dataset. There are comparable structural theories for visual narratives \citep{Bateman2014AMD,wildfeuer2014film,Cohn2015NarrativeCJ,10.3389/fpsyg.2014.00680,cogs.12016} that could suit a similar method. It could be a more productive approach than trying to annotate the various modalities of a comic or movie with a complex theoretical scheme. It is similar to how a full annotation of Barthes' theory would require each event on the story to be labelled as a cardinal function, catalysers, indices, or informers with structure as to how they related. 

In the area of text to image generation as well as the aforementioned Cogview models, Dall-E 2 \citep{DBLP:journals/corr/abs-2204-06125} and Imagen \citep{DBLP:journals/corr/abs-2205-11487} achieve similar outstanding result relying on diffusion models rather than a transformer for best results.\footnote{Imagen, like Flamingo, also relies on a frozen LM and only trains the multimodal image generation diffusion part of the model.} Diffusion models, developed by \citet{pmlr-v37-sohl-dickstein15}, can be thought of as layered models where noise is progressively added to the input at each layer. Each layer is trained to try and reverse the noise added by the previous layer. \citet{NEURIPS2020_4c5bcfec} simplified the learning objective to allow faster and improved training. They can be thought of as similar to denoising autoencoders but without the bottleneck. Sampling from the space when also conditioning on a caption vector will thus generate an original image as the model converts the noise layer by layer into an image. Like VAEs the latent space can be a Gaussian distribution \citep{NEURIPS2021_b578f2a5}. The output of Imagen and Dall-E 2 is a low-resolution image that can progressively be upscaled into a high-resolution image via methods such as SR3 \citep{DBLP:journals/jmlr/HoSCFNS22} and
cascaded diffusion \citep{DBLP:journals/corr/abs-2106-15282}. Diffusion networks can be applied to continuous vector fields or discrete or structured representations as per \citet{NEURIPS2021_958c5305}. In this recent image generation work, they have proved themselves equal to transformers and far superior to previous models relying on VAEs. I Chapter \ref{chap:tdvaesuspense} and  \ref{chap:tdvaegeneration} the model fell down on the TD-VAE model lack the capability to learn plot progressions. There are temporal autoregressive diffusion models \citep{hoogeboom2022autoregressive} that can learn to infer future states, although not currently with jumpy predictions as with TD-VAE. As diffusion models have shown their recent superiority in image generation tasks over similar VAE models, they may likewise show big improvements in modelling story state as an improvement over TD-VAE. This could either be by applying them directly in the continuous vector space as per the existing model or on discrete representations as is recommended in the relevant chapters.

The approach that has been taken throughout the thesis with training models is to finetune existing models where possible with domain-specific story data. The sizing of models has been selected to make it feasible given the available resources. However, many of the best-performing language models and newer multimodal models mentioned earlier in the chapter are far bigger than the models in this thesis. They can have tens of billions of parameters. Finetuning these models is not feasible for most labs, given the resources required. Lots of recent attention has therefore been on adapting existing pre-trained models to new tasks and domains by finetuning only a smaller fraction of the original model's weights in adapter layers on new data. \citet{DBLP:journals/corr/abs-2204-10019} evaluate three such approaches for adapting large LMs: A \textit{prompt tuning} method is a small LM encoder that injects the input text in the latent space of the larger model. A retrieval model similar to RAG that finetunes answer encoders and applies cross attention across the retrieved passages again into the latent space of an LM; an existing LM can be infused with external knowledge without finetuning the main encoder/decoder LM. A model that can connect the same LMs can be run recursively so that each recursion provides a better prompt for the next. All of the methods with parameter sizes of approximately $1-2$\% of the original models can achieve similar performance to the whole model of $> 10$ billion parameters being finetuned on the new domain and task. The explanation is that large models, when tuned on a new domain or task can overfit the new task, forget training from other tasks, and lose their generality and out of domain robustness. Smaller adapter components retain the overall power of the model, making the best use of it for the new task or domain while keeping retraining to a minimum. The same problem occurs in machine translation where there can be language with large volumes of data and parallel translation, say English and French, and large numbers with hardly any data. Smaller adapter modules that interleave existing layers in the model \citep{bapna-firat-2019-simple} have proved effective in adapting existing models to new language with small amounts of training data, \citet{cooper-stickland-etal-2021-recipes} provides some \textit{recipes} for doing this. The thesis has discussed how culturally specific story forms can be. Adapting to local cultures of stories is not so different in principle from adapting to new languages. There are additional approaches that adapt existing LMs to be able to adapt multimodal input without finetuning, such as MAGMA \citep{DBLP:journals/corr/abs-2112-05253} (Flamingo has a similar architecture), or can adapt large multimodal models such as CLIP \citet{DBLP:conf/cvpr/ZhangG0021}. The basic concept where the adapters are applied upstream of the frozen model, interleaves it, or downstream is to train a small number of weights to adapt the much large model for the new setting. This is orthogonal to all the other ideas in future work. These frozen model adaption methods could be more a productive alternative for reusing and integrating larger and better performing models for story-specific applications, with a much lower training cost.

Some of the further application areas have been touched on already. For example, in Chapter \ref{chap:salience} it was mentioned the method of identifying salience can be iteratively applied to abridged text. Both the suspense and the salience have only been tested on fictional narrative form. So clearly, a further evaluation would be to try the method compared to summary systems. In principle, the same could apply to identifying the most suspenseful events as well, but unlike summarisation, there wouldn't be suitable existing extractive summarisation datasets available. Barthes' scheme includes other non \textit{cardinal functional} elements such as \textit{catlysers}, \textit{indices}, and \textit{informants}; there are other theoretical models. Further work could try and fill out more complex structures with reasoning applied over latent representations, as per the BCF deletion method. In the same chapter, it was also mentioned that the advantages of an episodic memory model are that it should generate alternative text at any point in the story. The focus of the model should be on generation in context rather than the short prompts typical of most generation tasks. An interactive evaluation similar to \citet{clark-smith-2021-choose} with an extended context well-known to the reader is a favourable approach.

 An interesting comparison would be to apply the same approach on News corpora. News is typically written in a pyramidic style, with the most salient information presented near the beginning. This should be reflected in the salience method. With surprise and suspense, it could be hypothesised that it would considerably change the models' behaviour. If the main details are known upfront and conditioned on, there should be less surprise and suspense later on. Other domains such as comedy also rely on expectation, suspense and surprise, and similar methods could be applied. The \textit{Mr Bean} and \textit{Only Fools and Horses} examples rely entirely on breaking audience expectation; stand-up material and classic jokes are different but also rely a comedic version of suspense. In the analysis for both the Ely model work and BCF, it was discussed that the models might well be picking up boundaries between scenes, events like the entrance or exit of characters, or a shift in topic or mood. The cues could also correlate at a lower level with discourse models such as RST \citep{Mann1988RhetoricalST} and SDRT \citep{Asher2005LogicsOC}, and there is some overlap with the structural narrative theories and these models. Therefore it could be another avenue for future work.
 
 In the background, Chapter \ref{chap:backgroundtheory}, the \textit{paradox of suspense} was reviewed and how there are contradictory alternative views of suspense. The general results from suspense and surprise work support the idea that both uncertainty over state and outcome matter, and the outcome seems a more important contributor. However, the studies only test if a computational model correlates with annotations. A whole other line of work would be more explicitly cognitive in annotating a range of factors that influence suspense and then trying to replicate those factors in the model. For example, if recall is an important factor in surprise, through manipulating the context length, episodic memory can be truncated, masked, etc., to test correspondence with the cognitive phenomenon. There is recent work on reading strategies of fiction for longer books \citep{Farsi2020ReadingSA}; a hypothesis or the study is that reading is not linear, but readers typically reread the section or refer back to earlier parts to recall events. There are high tracking studies for reading difficulties \citep{GranEkstrand2021ScreeningFR} and Children's reading \citep{AlKhalefah2021ReadingPO}. A novel use for an extended RAG type model would be to correlate memory with patterns of reading in eye-tracking studies of fiction or non-fiction. Are the passages that human readers refer back to the same as those most looked up by the models? Are sections that need to be reread or are read more slowly correlated with high perplexity or bigger distance shifts in embeddings suggesting they are similarly less comprehensible to the model?
 
 The last decade has seen a considerable increase in computational methods for literary analysis and humanities research. \citet{underwood2019distant} has advocated it as a complement to traditional methods. Automated tagging of suspense (assuming it could be scaled to works of interest) or salience would clearly interest digital humanities scholars. Secondary analysis such as the memory graphs from the RAG model or which KB knowledge is most relevant could be suitable for mapping structure in novels or as inputs to further graph-based methods. Clustering analysis of latent vector could annotate sections according to identified types.
 
 The final future work section assumes a composite hybrid model of all of those in the thesis with the suggested future enhancements. The model would be a hierarchical one that as per the Clockwork-VAE suggestion could capture different levels of planning at each level; forecasting of the future state would be within the latent state possible with the discrete VQ-VAE or transformer like vocabulary code representations; projection could be with a temporal diffusion model; conditional generation is improved via better adapters; there is an external KB and episodic memory; the memory has a better model of temporal ordering and causality; the memory can conditionally track characters.\footnote{There are alternative architecture that would have extracted event structure or graph-based event chains that would also work with many of the suggestions.} The conceit is that it allows future work that isn't feasible with the individual components or existing limitations. It would be feasible to run the Ely surprise or suspense methods of longer works of the kind found in Shmoop with the memory aiding with the much longer plots. Models such as \textit{Dramatis} \citep{DBLP:conf/aaai/ONeillR14} build a system to write more suspenseful stories. The integrated system would make it feasible to try and do something similar in an open domain. The model would sample over possible alternative states and measure the levels of surprise and suspense and conditionally generate for the desired paths. The conditional generation would only be run on the final chosen path, saving the computational complexity of generating for lots of branches and allowing planning to look further ahead. The generation system could also exploit the difference between reader expectations and that of major characters to create more suspenseful stories. For example, choosing a path the reader expects but the characters don't leading to suspense. Or another possibility is to generate a character specific history or back story, or generate character events during the story that do not appear in the generated text but not in the reader script. It will create more of a mystery, the reader as a problem solver paradigm, where the readers is trying to filling in the gaps in their understanding. Alternatively, selecting paths that lead to surprising events for one character but not others. It is a minor theme in the thesis because obviously events not in the text cannot be trained on by a model, but would be crucial in generating interesting stories of more than a few paragraphs. After all the plot of \textit{Gone Girl} and \textit{Great Expectations} depends on events that are unknown to both the reader and other keys characters for most of the story, and it is the inconsistencies from being unaware of those events that creates most of the drama.
 
The thesis started with the equation of \citet{gottschall2012storytelling}, that $\text{Story} = \text{Character} + \text{Predicament} + \text{Attempted Extrication}$, and the quote that storytelling can be thought of as a cognitive simulator. The challenge is that modelling stories computationally requires not just comprehension of everyday and normally expected sequences of events but also unexpected events, unusual circumstances and characters' motivations. There are limitations to the results in the thesis and the models in it to comprehend all these elements. However, the novel approach in this thesis of operationalising narrative theory, whether with Hale, Ely, or Barthes' models, with recent advances in LMs and unsupervised learning, shows much promise. As innovation continues with ML models and improvements such as the characters' perspectives is likely to lead to improved modelling for better comprehension and story generation. Likewise, exploring the capabilities implicit in models with narrative theory will lead to further insights. In the \textit{Hitchhiker's Guide to the Galaxy} \citep{adams1980hitchhiker} the \textit{philsophers} threaten to strike saying \textit{we demand rigidly defined areas of doubt and uncertainty!} They fear that the \textit{stupendous super computer which was so amazingly intelligent} will put them out of work. With suspense and the dramatic tension in storytelling, the challenge is the opposite in modelling the jeopardy and doubt inherent to the form, even if it's someone faking their own death, a troubled relationship with a childhood companion, or drying their lettuce in a sock.


\appendix

\chapter{Sentence-By-Sentence Annotation Instructions} 

\label{appendex:annotationinstructions}

\textit{The following are the written instructions giving to the annotators. The training examples are given in Chapter \ref{chap:suspenseannotation}}

For the first HIT there will be an additional training step to pass. This will take about 5 minutes. After this you will receive a code which you can enter in the code box to bypass the training for subsequent HITS. Other stories are in separate HITS, please search for "Story dramatic tension, reading sentence by sentence" to find them. The training completion code will work for all related HITS.

You will read a short story and for each sentence be asked to assess how the dramatic tension increases, decreases or stays the same. Each story will take an estimated 8-10 minutes. Judge each sentence on how the dramatic tension has changed as felt by the main characters in the story, not what you as a reader feel. Dramatic tension is the excitement or anxiousness over what will happen to the characters next, it is anticipation.

Increasing levels of each of the following increase the level of dramatic tension:
\begin{itemize}
\item \textbf{Uncertainty:} How uncertain are the characters involved about what will happen next? Put yourself in the characters shoes; judge the change in the tension based on how the characters perceive the situation.
\item \textbf{Significance:} How significant are the consequences of what will happen to the central characters of the story?
\end{itemize}
An Example: Take a dramatic moment in a story such as a character that needs to walk along a dangerous cliff path. When the character first realises they will encounter danger the tension will rise, then tension will increase further. Other details such as falling rocks or slips will increase the tension further to a peak. When the cliff edge has been navigated safely the tension will drop. The pattern will be the same with a dramatic event such as a fight, argument, accident, romantic moment, where the tension will rise to a peak and then fall away as the tension is resolved.

You will be presented with one sentence at a time. Once you have read the sentence, you will press one of five keys to judge the increase or decrease in dramatic tension that this sentence caused. You will use five levels (with keyboard shortcuts in brackets):

\begin{itemize}

\item \textbf{Big Decrease (A):} A sudden decrease in dramatic tension of the situation. In the cliff example the person reaching the other side safely.
\item \textbf{Decrease (S):} A slow decrease in the level of tension, a more gradual drop. For example the cliff walker sees an easier route out.
\item \textbf{Same (Space):} Stays at a similar level. In the cliff example an ongoing description of the event.
\item \textbf{Increase (K):} A gradual increase in the tension. Loose rocks fall nearby the cliff walker.
\item \textbf{Big Increase (L):} A more sudden dramatic increase such as an argument. The cliff walker suddenly slips and falls.
\end{itemize}
\chapter{Further Salience Examples}

\label{appendix:salience}

\textit{Contains further examples from Chapter} 

\section{Propp Magic Swan Example}
\label{sec:swan}

\begin{figure}[htbp]
\centering
\includegraphics[trim={0.5cm 1.0cm 2.0cm 2.5cm},clip,width=1.0\textwidth]{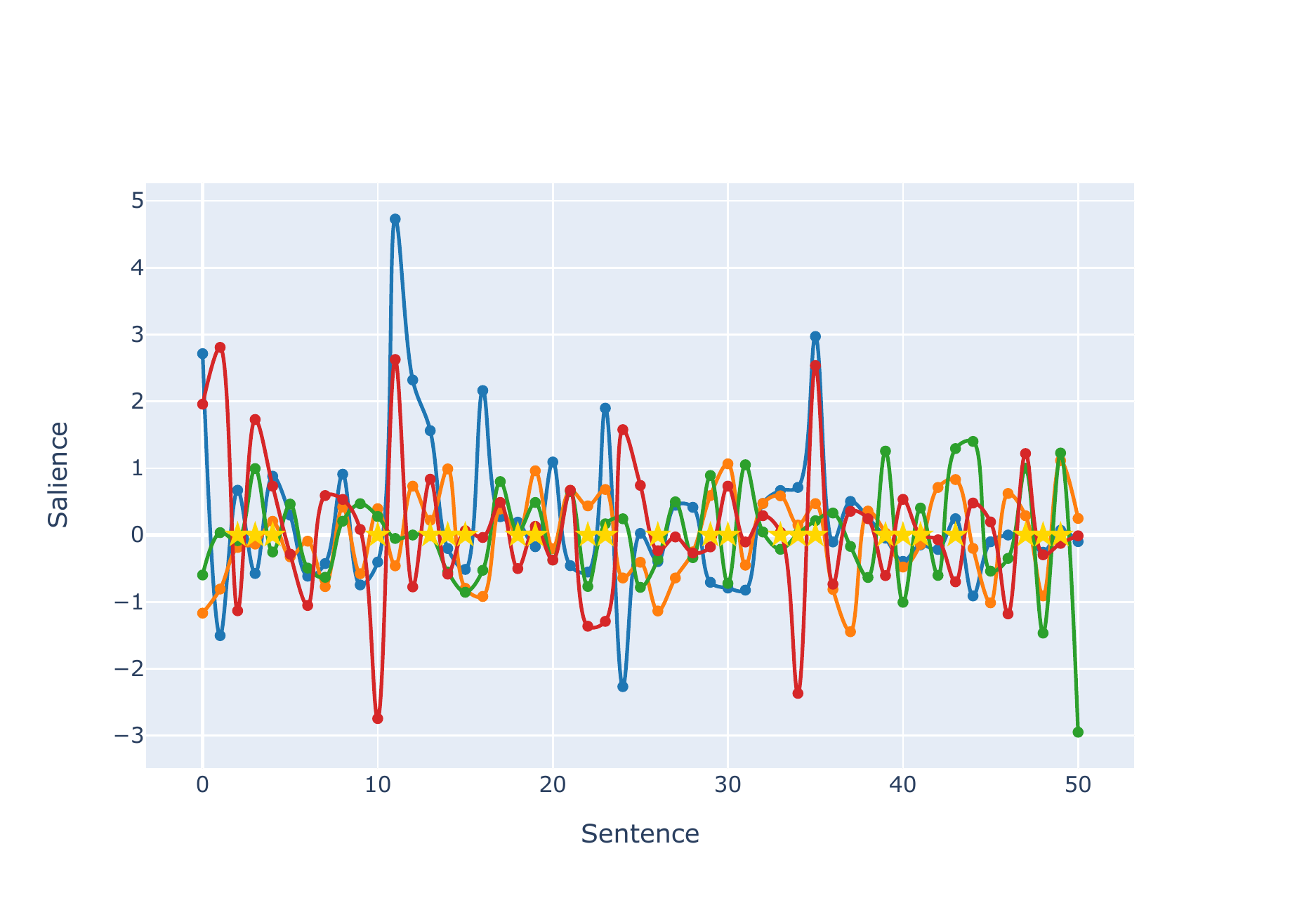}
\caption{A salience plot of the Propp story \textit{The Magic Swan-Geese}:
\textbf{\textcolor{plotacolor}{Like-Sal}}, \textbf{\textcolor{plotbcolor}{Clus-Sal}}, \textbf{\textcolor{plotccolor}{Otake-Sal}}, \textbf{\textcolor{plotdcolor}{Emb-Sal}}, {\color{yellow} $\medstar$} \textbf{Propp roles}. Plots are rescaled to unit variance and scaled.}
 \label{fig:propp_swan}
\end{figure}

\begin{figure}[htbp]
\centering
\includegraphics[trim={0.3cm 1.0cm 2.0cm 2.5cm},clip,width=1.0\textwidth]{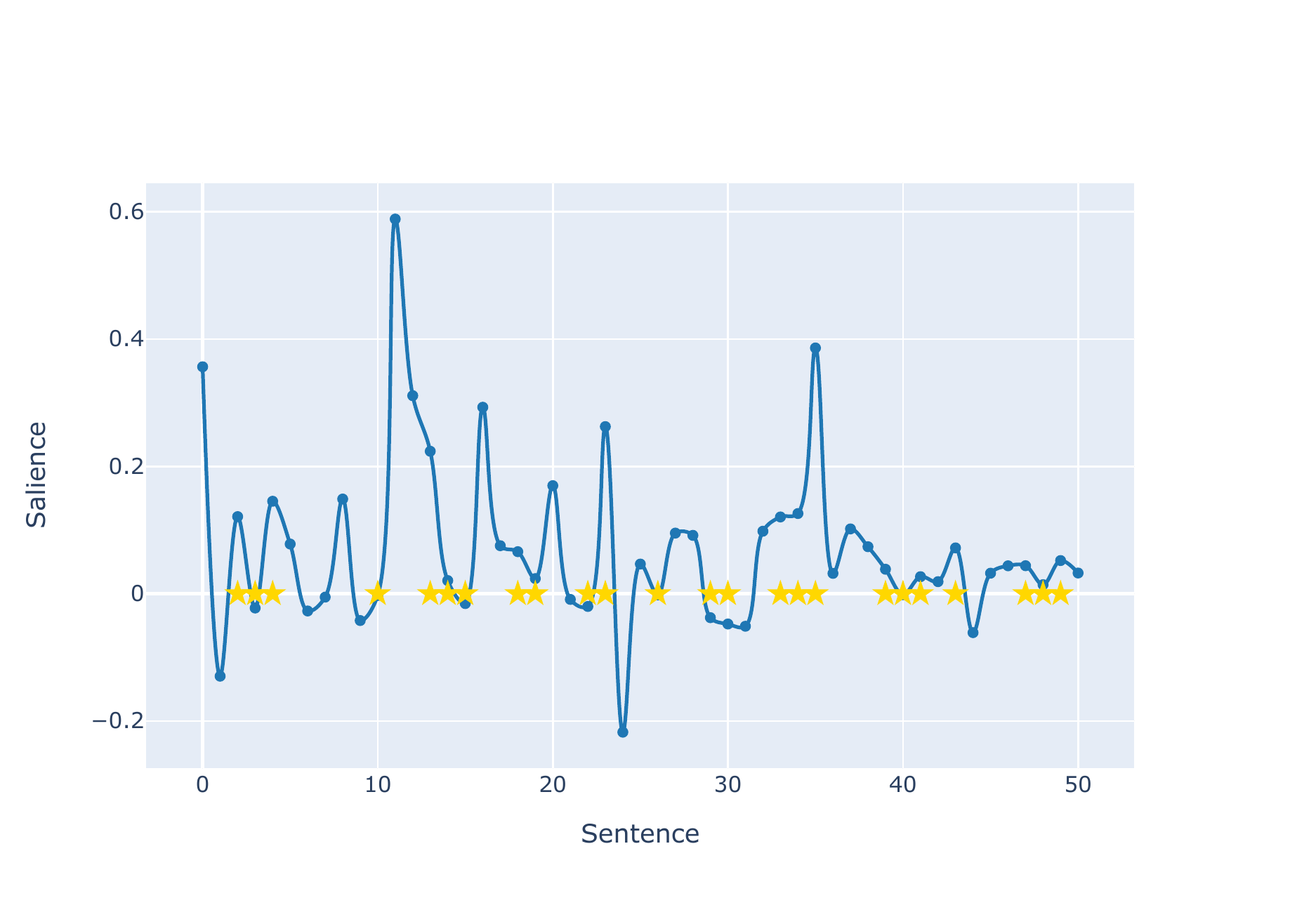}
\caption{Like-Sal plotted for \textit{The Magic Swan-Geese}. Unlike the multiplot version, the output is not scaled and so shows the real value of the average log likelihood difference.}
 \label{fig:propp_swan_all}
\end{figure}

The \textit{Magic Swan-Geese} plots and text for the salience work in Chapter \ref{chap:salience}.  Figure \ref{fig:propp_swan} is the multiple plot and Figure \ref{fig:propp_swan_all} is the single plot. 

The full text for  \textit{The Magic Swan-Geese}:

\begin{enumerate}
\setcounter{enumi}{-1}
\item An old man lived with his old wife ; they had a daughter and a little son .
\item \say{ Daughter , daughter , } said the mother , \say{ we are going to work ; we shall bring you back a bun , sew you a dress , and buy you a kerchief .}
\item Be careful , watch over your little brother , do not leave the house . ''
\item The parents went away and the daughter forgot what they had told her ; she put her brother on the grass beneath the window , ran out into the street , and became absorbed in games .
\item Some magic swan geese came , seized the little boy , and carried him off on their wings .
\item The girl came back and found her brother gone .
\item She gasped , and rushed to look in every corner , but could not find him .
\item She called him , wept , and lamented that her father and mother would scold her severely ; still her little brother did not answer .
\item She ran into the open field ; the swan geese flashed in the distance and vanished behind a dark forest .
\item The swan geese had long had a bad reputation ; they had done a great deal of damage and stolen many little children .
\item The girl guessed that they had carried off her brother , and rushed after them .
\item \say{Stove , stove , tell me , whither have the geese flown ?}
\item  \say{If you eat my cake of rye I will tell you .}
\item  \say{Oh , in my father 's house we do n't even eat cakes of wheat !}
\item The stove did not tell her .
\item She ran farther and saw an apple tree .
\item  \say{Apple tree , apple tree , tell me , whither have the geese flown ?}
\item \say{If you eat one of my wild apples , I will tell you .} 
\item  \say{Oh , in my father 's house we do n't even eat sweet apples .}
\item She ran farther and saw a river of milk with shores of pudding .
\item \say{River of milk , shores of pudding , whither have the geese flown ?}
\item \say{If you eat of my simple pudding with milk , I will tell you .}
\item \say{Oh , in my father 's house we do n't even eat cream .}
\item She would have run in the fields and wandered in the woods for a long time , if she had not luckily met a hedgehog .
\item She wanted to nudge him , but was afraid that he would prick her , and she asked : \say{ Hedgehog , hedgehog , have you not seen whither the geese have flown ?}
\item \say{Thither , } he said , and showed her .
\item She ran and saw a little hut that stood on chicken legs and turned round and round .
\item In the little hut lay Baba Yaga with veined snout and clay legs , and the little brother was sitting on a bench , playing with golden apples .
\item His sister saw him , crept near him , seized him , and carried him away .
\item But the geese flew after her : if the robbers overtook her , where would she hide ?
\item There flowed the river of milk with shores of pudding .
\item \say{Little mother river , hide me ! } she begged .
\item \say{If you eat my pudding .}
\item There was nothing to be done ; she ate it , and the river hid her beneath the shore , and the geese flew by .
\item She went out , said \say{ Thank you , } and ran on , carrying her brother ; and the geese turned back and flew toward her .
\item What could she do in this trouble ?
\item There was the apple tree .
\item \say{Apple tree , apple tree , little mother , hide me ! '' she begged .}
\item \say{If you eat my wild apple .}
\item She ate it quickly .
\item The apple tree covered her with branches and leaves ; the geese flew by .
\item She went out again and ran on with her brother .
\item The geese saw her and flew after her .
\item They came quite close , they began to strike her with their wings ; at any moment they would tear her brother from her hands .
\item Luckily there was the stove on her path .
\item \say{Madam Stove , hide me ! } she begged .
\item \say{If you eat my cake of rye .} 
\item The girl quickly stuck the cake in her mouth , went into the stove , and sat there .
\item The geese whirred and whirred , quacked and quacked , and finally flew away without recovering their prey .
\item And the girl ran home , and it was a good thing that she came when she did , for soon afterward her mother and father arrived .
\end{enumerate}

\section{Text Generation Examples}

\label{sec:genexamples}

\textit{Further text generation example for Chapter \ref{chap:salience}.}

\paragraph*{Gone Girl Example:}

\paragraph*{Context:}  NICK I did not kill my wife, Sharon. I am not a murderer. SHARON But you were unfaithful. NICK I was. I am not proud of my actions. SHARON You allowed Amy’s parents, Amy’s friends, the people of your hometown, to believe that you were a loving husband who was desperate to find his missing wife. GG Yellow Revisions 9/27/13 133A. NICK I am desperate to find my missing wife. SHARON I just wonder how you can ask us to believe you, now that we know.. .you’re a liar? NICK I did not come forward about my affair because I knew it would make me look very, very bad. I don’t care anymore about how I look. That’s done.

\paragraph*{Real Continuation:} I care about finding my wife. SHARON I’m just trying to get clear- NICK Let me say this: I am not a killer. But I’m far from being a good guy. I was a bad husband to a wonderful wife. I broke the vow I made to her. SHARON What does that mean to you, Nick? NICK It means I was basically a con artist. (MORE) GG Yellow Revisions 9/27/13 NICK (CONT’D) I met Amy Elliott seven years ago and I was transfixed. Amy does that.

\paragraph*{Generated:}  We had a horrible marriage. And, when we thought she was gone, I knew that it was all over for me. And I knew that the police would be looking for me and I knew that they would not find Amy. And it would be my fault, if they found Amy. I knew that. I had to do something, if I was going to find her. \textbf{If I was going find her, it would be because I had a new baby}. I could not have left Amy for a man like Nick Dunne. I was only going to have my child because I wanted to be with her.

\paragraph*{Great Gatsby Example:}

\paragraph*{Context:} ... One night I did hear a material car there, and saw its lights stop at his front steps. But I didn’t investigate. Probably it was some final guest who had been away at the ends of the earth and didn’t know that the party was over. On the last night, with my trunk packed and my car sold to the grocer, I went over and looked at that huge incoherent failure of a house once more. On the white steps an obscene word, scrawled by some boy with a piece of brick, stood out clearly in the moonlight, and I erased it, drawing my shoe raspingly along the stone. Then I wandered down to the beach and sprawled out on the sand.  

\paragraph*{Real Continuation:} Most of the big shore places were closed now and there were hardly any lights except the shadowy, moving glow of a ferryboat across the Sound. And as the moon rose higher the inessential houses began to melt away until gradually I became aware of the old island here that flowered once for Dutch sailors’ eyes—a fresh, green breast of the new world. Its vanished trees, the trees that had made way for Gatsby’s house, had once pandered in whispers to the last and greatest of all human dreams ...

\paragraph*{Generated:} I laid down beside a big rock and watched the night sky. I slept till nearly dawn, when a soft splash woke me. I looked up, and there was the ghost of a bird lying on the sand near the shore. It was a white bird that fluttered lazily across the light breeze, then sank into the sand and lay motionless. I slept for hours and hours until I dreamed of Daisy, but \textbf{when I awoke in the morning} I could hear nothing but the gentle murmur of the waves and the soft sighs of the gulls. I rose with a start, and looked out at the dark horizon.

\paragraph*{Pretty Woman Example:}

\paragraph*{Context:} EDWARD Well, I'm only halfway out. VIVIAN Didn't say much in the car on the way home. You thinking about dinner ? I was a maniac. PENTHOUSE - LIVING ROOM - INTERIOR NIGHT Medium shot of the terrace seen from the open French window. VIVIAN I mean, the business was good, I think. You know ? Vivian sits down on the balustrade of the terrace. VIVIAN He's in trouble. You want his company. He doesn't want to let it go. EDWARD Thanks for the recap.

\paragraph*{Real Continuation:} VIVIAN The problem is, I think you liked Mr. Morse. PENTHOUSE - TERRACE - EXTERIOR NIGHT Edward points his finger toward Vivian. EDWARD What I would like is for you to get down from there. You're making me very nervous. Please come down. VIVIAN It's making you nervous ? What if I just leaned back a little bit like this ? She starts leaning back a little. VIVIAN Would you... Would you rescue me if I fell ? EDWARD Vivian, I'm serious. Come...

\paragraph*{Generated:} EDWARD Sure, let's talk about that. Vivian leaves the kitchen. Edward looks over the balcony at her, and then looks at his watch. \textbf{EDWARD We still have some time before the police show up.} Vivian turns to face him. EDWARD Yeah, I've got to get a lot of work done tonight. Vivian shakes her head. EDWARD I know, I know. Vivian looks at him a bit longer. EDWARD Can you go get some clothes? Vivian stares at him a moment. EDWARD No, I'll come back to get them when I'm done with work tonight.

\section{Abridgement Examples}

\label{sec:abridgeexamples}

The following are some unevaluated abridgement examples. Each one has twenty sentences and is not the complete work. The method is to iterate with the BCF method from Chapter \ref{chap:salience} where there are $4$ iterations and at each iteration retains the most salient $75\%$ of sentences which give approximately a $10\%$ abridgement.

\paragraph*{Great Expectations}

\begin{enumerate}
\item My father’s family name being Pirrip, and my Christian name Philip, my infant tongue could make of both names nothing longer or more explicit than Pip.
\item As I never saw my father or my mother, and never saw any likeness of either of them (for their days were long before the days of photographs), my first fancies regarding what they were like were unreasonably derived from their tombstones.
\item “Hold your noise!” cried a terrible voice, as a man started up from among the graves at the side of the church porch.
\item “Quick!” “Pip, sir.” “Once more,” said the man, staring at me.
\item “Where’s your mother?” “There, sir!” said I.
\item You know what a file is?” “Yes, sir.” “And you know what wittles is?” “Yes, sir.” “You get me a file.”
\item He tilted me again.
\item You fail, or you go from my words in any partickler, no matter how small it is, and your heart and your liver shall be tore out, roasted, and ate.
\item There’s a young man hid with me, in comparison with which young man I am a Angel.
\item That young man has a secret way pecooliar to himself, of getting at a boy, and at his heart, and at his liver.
\item At the same time, he hugged his shuddering body in both his arms,—clasping himself, as if to hold himself together,—and limped towards the low church wall.
\item My sister, Mrs. Joe Gargery, was more than twenty years older than I, and had established a great reputation with herself and the neighbours because she had brought me up “by hand.”
\item When I ran home from the churchyard, the forge was shut up, and Joe was sitting alone in the kitchen.
\item “What’s the matter now?” said she, smartly, as she put down her cup.
\item Joe put his mouth into the forms of returning such a highly elaborate answer, that I could make out nothing of it but the single word “Pip.” “There was a conwict off last night,” said Joe, aloud, “after sunset-gun.
\item And they fired warning of him.
\item Hulks are prison-ships, right ’cross th’ meshes.”
\item Now, you get along to bed!”
\item Then, as I looked up at it, while it dripped, it seemed to my oppressed conscience like a phantom devoting me to the Hulks.
\item The gates and dikes and banks came bursting at me through the mist, as if they cried as plainly as could be, “A boy with somebody else’s pork pie!
\end{enumerate}

\paragraph*{Rocky}

\begin{enumerate}
\item BLUE DOOR FIGHT CLUB - NIGHT SUPERIMPOSE OVER ACTION...
\item In Rocky's corner he is being assisted by a shriveled, balding CORNERMAN, who is an employee of the club...
\item CORNERMAN (lackluster) ...Ya waltzin' -- Give the suckers some action.
\item The BELL RINGS...
\item The fighter on the stretcher passes behind him.
\item FIGHTER 1 ...Tomorrow me an' my woman are gonna tip on down to Atlantic City, man.
\item TROLLEY - NIGHT Rocky is on the trolley heading to South Philly...
\item Rocky shrugs and moves away...
\item He nears a heavy man working the crane.
\item FATS (terror-filled) Don't hit the face!
\item ROCKY Mr. Gazzo wants the two hundred now!
\item STREET - DAY Later that morning Rocky passes \\Animal Town Pet Shop\ in South Philly...
\item He stares at a litter of Lhasa Apsa puppies.
\item ROCKY ...How's the turtle food this week?
\item ROCKY The last food I got here had more moths than flies -- An' the moths get caught in my turtle's throat -- That makes them cough -- The OWNER, a squat woman of forty, steps out of the back and waves at Rocky.
\item ROCKY (continuing) Yo, Gloria -- I was talkin' about the turtle food -- Like I was sayin', the moths get caught in the turtle's throat an' makes 'em cough... (coughs) A little cough an' I gotta smack 'em on the shell -- An' whatta think they get?
\item (Adrian nods) Rocky waves goodbye to Adrian and exits the shop. EXT. LEHIGH ST.
\item TRAIN TRESTLE - DAY Gazzo picks up Rocky.
\item Gazzo's Bodyguard looks at Rocky's bruised face in the mirror and smiles.
\item They pull away. EXT. GYM - DAY Gazzo drives off and Rocky strolls across the street to Goldmill's Gym.
\end{enumerate}


\bibliography{thesis}


\end{document}